\documentclass[english, PhD, oneside]{sapthesis}
\usepackage[utf8]{inputenx}
\usepackage{indentfirst}
\usepackage{microtype}
\usepackage{multirow}
\usepackage{tabularx}
\usepackage{xspace} 
\usepackage{tabu}
\usepackage{enumitem}
\usepackage{float}
\usepackage{url}

\usepackage{lettrine}
\usepackage{epsfig}
\usepackage{pifont}
\usepackage{dsfont}
\usepackage{placeins}
\usepackage[nottoc, notlof, notlot]{tocbibind}
\usepackage{amsmath} 
\usepackage{amssymb}  
\usepackage{algorithm,algcompatible}
\usepackage{bm}
\usepackage{mathtools}

\usepackage{graphicx}
\usepackage{subfig}
\usepackage{subfloat}
\usepackage{adjustbox}
\usepackage{titlesec}
\usepackage{xspace}
\usepackage{bbold}
\usepackage{xcolor}

\usepackage[hyperfootnotes=true,			
			bookmarks=true]{hyperref}
\newcommand{\myTitle}{Towards Recognizing New Semantic Concepts in New Visual Domains}

\newcommand{\vct}[1]{\ensuremath{\boldsymbol{#1}}}
\newcommand{\mat}[1]{\mathtt{#1}}
\newcommand{\set}[1]{\ensuremath{\mathcal{#1}}}
\newcommand{\con}[1]{\ensuremath{\mathsf{#1}}}

\newcommand{\ind}[1]{\ensuremath{\mathbb 1_{#1}}}
\newcommand{\argmax}{\operatornamewithlimits{\arg\,\max}}
\newcommand{\argmin}{\operatornamewithlimits{\arg\,\min}}

\def\etal{\textit{et al.}~}

\DeclareMathOperator{\Var}{Var}
\DeclareMathOperator{\E}{E}
\DeclareMathOperator{\DA}{DA}
\DeclareMathOperator{\mDA}{mDA}

\algnewcommand\INPUT{\item[\textbf{Input:}]}%
\algnewcommand\OUTPUT{\item[\textbf{Output:}]}%

\newcommand*{\eg}{e.g.\@\xspace}
\newcommand*{\ie}{i.e.\@\xspace}

\newcommand{\cmark}{\ding{51}}%

\newcommand{\myparagraph}[1]{\vspace{5pt}\noindent\textbf{#1}}

\newcommand{\methodName}{{CuMix} }
\newcommand{\methodNameFull}{\textbf{Cu}rriculum \textbf{Mix}up for recognizing unseen categores in unseen domains}
\newcommand{\sota}{state of the art }
\newcommand{\icl}{ICL}
\newcommand{\ours}{MiB }
\newcommand{\expandednick}{\textbf{M}odel\textbf{i}ng the \textbf{B}ackground for incremental learning in semantic segmentation}

\newcommand{\hypers}{hyper-parameters }
\newcommand{\hyper}{hyper-parameter }
\newcommand{\owr}{B-DOC }
\newcommand{\owrFull}{B-DOC }
\newcommand{\real}{{\rm I\!R}}

\DeclareMathOperator{\GBN}{GBN}

\DeclareMathOperator\sign{sgn}
\makeatletter
\newlength\eqcol@newlen
\newlength\eqcol@oldlen
\let\eqcol@bc\hfil
\let\eqcol@ec\hfil
\let\eqcol@br\hfil
\let\eqcol@el\hfil
\newcolumntype{e}[1]{%
  >{\setbox0\hbox\bgroup}#1%
  <{\egroup
    \ifdim\wd0<\eqcol@newlen\else\global\eqcol@newlen\wd0\fi
    \ifdim\wd0<\eqcol@oldlen\else\global\eqcol@oldlen\wd0\fi
    \hbox to \eqcol@oldlen{%
      \csname eqcol@b#1\endcsname
      \box0 %
      \csname eqcol@e#1\endcsname
    }%
  }%
}
\newcount\eqcol@count
\def\eqcolRead{%
  \global\advance\eqcol@count1 %
  \eqcol@oldlen5em\relax
  \csname eqcol@def@\romannumeral\eqcol@count\endcsname
}
\def\eqcolWrite{%
  \immediate\write\@auxout{%
  \gdef\expandafter\noexpand\csname eqcol@def@\romannumeral\eqcol@count\endcsname
    {\global\eqcol@oldlen\the\eqcol@newlen\relax}%
  }%
  \global\eqcol@newlen0pt\relax
}
\let\eqcol@old@tabular\tabular
\def\tabular{\eqcolRead\eqcol@old@tabular}
\let\eqcol@old@endtabular\endtabular
\def\endtabular{\eqcol@old@endtabular\eqcolWrite}
\makeatother

\hypersetup{
			colorlinks=true,
			linkcolor=red,
                        linktoc=page,
			anchorcolor=black,
			citecolor=red,
			urlcolor=blue,
			pdftitle={\myTitle},
			pdfauthor={Massimiliano Mancini},
			pdfkeywords={thesis, sapienza, roma, university, domain adaptation, computer vision, 
            			recognition, object classification, deep learning, transfer learning, depth, rgb}
 }

\title{\myTitle}
\author{Massimiliano Mancini}
\IDnumber{1646014}
\course[]{Ingegneria Informatica}
\courseorganizer{Department of Computer, Control and Management Engineering ”Antonio
Ruberti“, Sapienza - University of Rome}
\submitdate{2019/2020}
\copyyear{2020}
\advisor{Prof. Barbara Caputo}
\coadvisor{Prof. Elisa Ricci}
\authoremail{massimiliano.mancini8@gmail.com}
\cycle{XXXII}

\begin{document}

\frontmatter
\maketitle

\dedication{To Monte Santa Maria Tiberina, my home.}
\newpage\null\thispagestyle{empty}\newpage

\begin{abstract}
Deep learning is the leading paradigm in computer vision. However, deep models heavily rely on large scale annotated datasets for training. Unfortunately, labeling data is a costly and time-consuming process and datasets cannot capture the infinite variability of the real world. Therefore, deep neural networks are inherently limited by the restricted visual and semantic information contained in their training set. In this thesis, we argue that it is crucial to design deep neural architectures that can operate in previously unseen visual domains and recognize novel semantic concepts. In the first part of the thesis, we describe different solutions to enable deep models to generalize to new visual domains, by transferring knowledge from a labeled source domain(s) to a domain (target) where no labeled data are available. We first address the problem of unsupervised domain adaptation assuming that both source and target datasets are available but as mixtures of multiple latent domains. In this scenario, we propose to discover the multiple domains by introducing in the deep architecture a domain prediction branch and to perform adaptation by considering a weighted version of batch-normalization (BN). We also show how variants of this approach can be effectively applied to other scenarios such as domain generalization and continuous domain adaptation, where we have no access to target data but we can exploit either multiple sources or a stream of target images at test time. Finally, we demonstrate that deep models equipped with graph-based BN layers are effective in predictive domain adaptation, where information about the target domain is available only in the form of metadata. In the second part of the thesis, we show how to extend the knowledge of a pre-trained deep model incorporating new semantic concepts, without having access to the original training set. We first consider the problem of adding new tasks to a given network and we show that using simple task-specific binary masks to modify the pre-trained filters suffices to achieve performance comparable to those of task-specific models. We then focus on the open-world recognition scenario, where we are interested not only in learning new concepts but also in detecting unseen ones, and we demonstrate that end-to-end training and clustering are fundamental components to address this task. Finally, we study the problem of incremental class learning in semantic segmentation and we discover that the performances of standard approaches are hampered by the fact that the semantic of the background changes across different learning steps. We then show that a simple modification of standard entropy-based losses can largely mitigate this problem. In the final part of the thesis, we tackle a more challenging problem: given images of multiple domains and semantic categories (with their attributes), how to build a model that recognizes images of unseen concepts in unseen domains? We also propose an approach based on domain and semantic mixing of inputs and features, which is a first, promising step towards solving this problem.

\end{abstract}

\newpage\null\thispagestyle{empty}\newpage

\section*{Acknowledgements}
I would like to thank all who contributed to achieving this amazing goal.
Heartfelt thanks to my advisors, Prof. Barbara Caputo, Prof. Elisa Ricci, and Samuel Rota Bul\`o. At the beginning of the Ph.D. I was a tenacious but pretty messy and badly organized student. Day by day, with huge patience, countless suggestions, and precious advice, they turned that student into a researcher. I am deeply thankful to Barbara, for introducing me to research, for transferring me her passion and dedication, and for teaching me that a clear plan is better than a bunch of ideas. A huge thanks to Elisa for inspiring me with her behavior, making me understand the importance of stubbornness, and how to face deadlines and pressure while always keeping the same positive attitude. A special thanks to Samuel: discussing problems and ideas with you, trying to follow your thoughts has been an amazing, advanced school for growing my scientific perspectives. All of you taught me how to identify interesting research questions, and how to think for answering them. You showed me how to make the best out of all experiences, how to celebrate successes and how to embrace and react to failures. I enjoyed every moment of this Ph.D. and I will always be grateful to you for shaping me as the researcher I am today.

 I would like to express my gratitude to Prof. Bernt Schiele and Prof. Timothy Hospedales for having taken the time to accurately read this thesis. It was a great honor for me to receive their positive and valuable feedback.

I am grateful to Stefano Messelodi, for welcoming me to Fondazione Bruno Kessler, allowing me to work in such an engaging environment. Appreciation is also due to Hakan Karaoguz and Prof. Patric Jensfelt for hosting me in the RPL lab in Stockholm, introducing me to the challenges of robot vision. Additionally, I wish to thank Prof. Zeynep Akata and all members of the EML lab in T\"ubingen for showing me different perspectives and, recently, welcoming me for a new exciting experience.

This journey wouldn't have been the same without some good fellows sharing the way. I thank all members of the VANDAL lab in Rome and Turin, with a big thanks to Fabio and Dario for bearing me in my attempt to become a better supervisor. Thanks also to Fabio (the first), Paolo, Valentina, Antonio, Silvia, and Mirco for sharing with me lab life and conference adventures. 
 I am grateful to all members of TeV and MHUG labs in Trento, with a special mention to Pilz, Simo, Swathi, Levi, Enrico, Aliaks, and Sub: thanks for sharing with me lab life, stressful and joyful times, conferences, and beers. Heartfelt thanks to all my co-authors, Lorenzo in particular for his fundamental support, smart insights, and nice moments together. 

Research is a part but not all of my life. I wish to thank all my long-time friends in Monte, with extra gratitude to Robi, Alex and Diego. Whenever I return to my hometown you always make me feel as if I have never been away. I love that feeling.

I would like to thank my family, from my cousins to my grandparents, for never making me feel alone. To my parents, Rinaldo and Anna: thank you for always supporting me and for the values you taught me. I do not think I can express in words how much I owe you.
Thanks to my sister, Serena, for understanding me and making me always remember what really matters. I am proud of you.

Finally, I want to thank Elisa, my girlfriend. These years were not easy for us: a long distance in between, occasional stress, pressures. You have always been patient, helping me, pushing me, and believing in me far more than what I do. I love you.

\newpage\null\thispagestyle{empty}\newpage

\tableofcontents
\listoffigures
\listoftables

\mainmatter
\chapter{Introduction}
\label{chap:intro}

\section{Overview}
\label{sec:intro}
A long-standing goal of artificial intelligence and robotics
is the implementation of agents that are able to interact in the real world.
In order to achieve this goal, a crucial step lays in making the agents understand the current state of the surrounding
environment, by providing them with both powerful sensors and the ability to process the information the sensors give them. To this extent, visual cameras are one
of the most powerful and information-rich sensors. Indeed, applications requiring visual abilities are countless: from self-driving cars to detecting and handling objects for service
robots in homes, from kitting in industrial workshops, to
robots filling shelves and shopping baskets in supermarkets,
etc., they all imply interacting with a wide variety of objects, which requires a deep understanding of how these objects
look like, their visual properties and associated functionalities. 

Due to the central role that vision has in the path towards developing agents with intelligent, autonomous behaviors, a lot of research efforts have been spent on improving computer and robot
vision systems. Within this context, in recent years these fields have seen unprecedented advancements thanks to deep learning architectures \cite{Goodfellow-et-al-2016}. Deep models are very effective in learning discriminative representations from input data, and their applications touch on many different fields, such as natural language processing \cite{mikolov2013distributed,collobert2008unified,deng2018nlp,young2018nlp}, speech recognition \cite{hinton2012deep,deng2013speech,deng2013speechmicr} and reinforcement learning \cite{lillicrap2015rl,mnih2015rl,gu2017rl}. In the context of computer vision, Convolutional Neural Networks (CNNs) \cite{lecun1998gradient} are the leading paradigm. These networks are particularly effective in processing grid-like input data \cite{Goodfellow-et-al-2016} a category to which images belong. The successes of CNNs in computer vision are countless: they have achieved outstanding results in many visual tasks, ranging from object classification \cite{krizhevsky2012imagenet,he2016deep} and detection \cite{girshick2014rich,ren2017faster}, to more complex ones such as image captioning \cite{karpathy2015deep,you2016image}, visual question answering \cite{antol2015vqa,xu2016ask} and motion transfer \cite{siarohin2019animating,chan2019everybody}.

Despite their effectiveness, CNNs have some drawbacks. First, they are data-hungry, i.e. very
large labeled datasets are usually required for training them \cite{russakovsky2015imagenet}. This
is a major issue since it is hard to obtain a large amount of labeled data for any possible application scenario. For instance, this often happens in robotics, where data acquisition and annotation are especially time-consuming and often infeasible.

\begin{figure}[t]
    \centering
    \includegraphics[width=\textwidth]{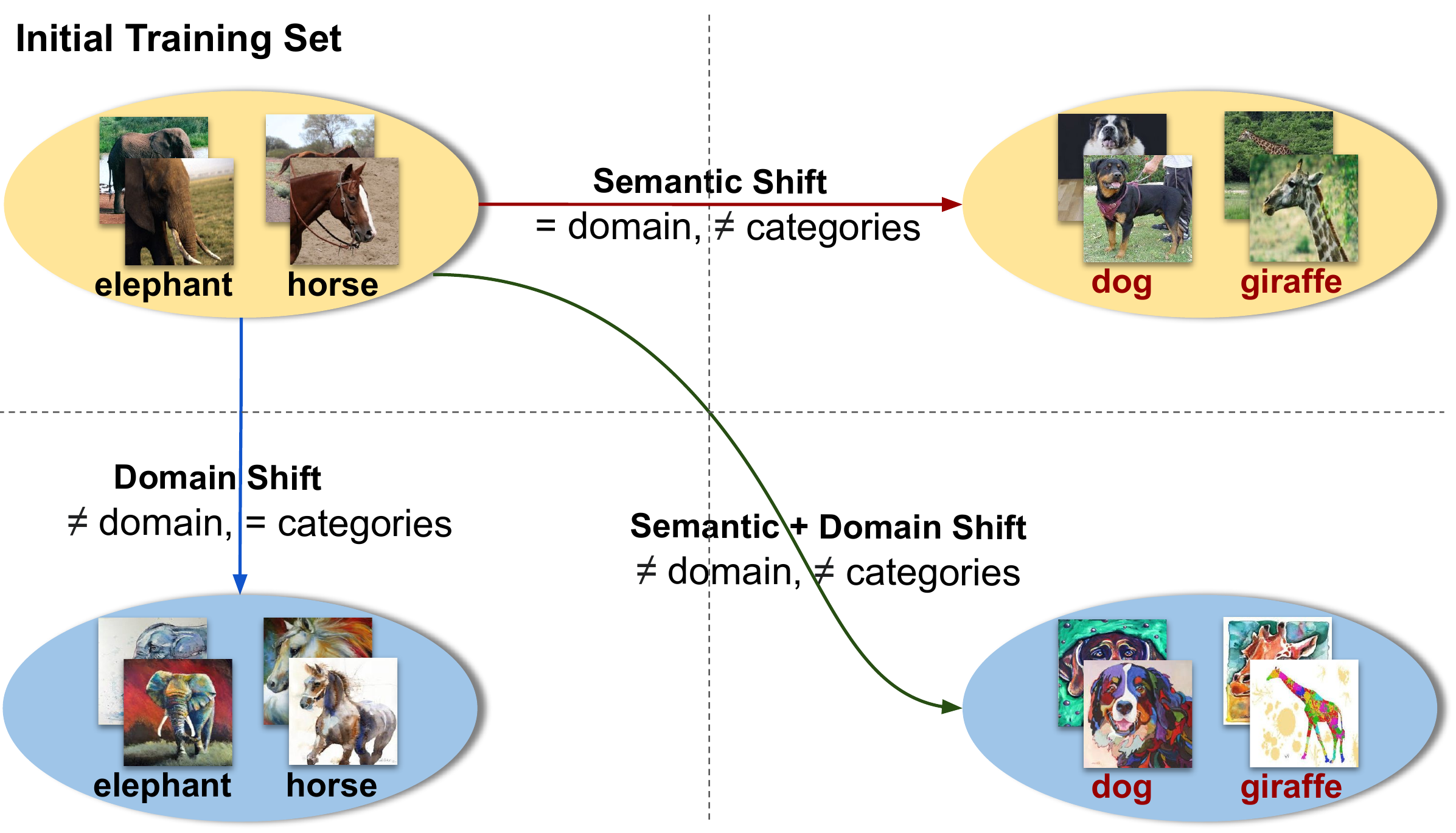} 
    \caption{
    Overview of our research problem. Suppose we are given an initial training set composed of images of a set of classes (e.g. \textit{elephant}, \textit{horse}) acquired in a given domain (e.g. \textit{real photos}). Two main discrepancies can occur at test time: either images contain the same semantics but in different domains (e.g. \textit{paintings}, bottom-left) or they contain images of the same domain but depicting different semantic concepts (e.g. \textit{dog} and \textit{giraffe}, top-right). In the first case we talk about \textit{domain shift} problem, while the second considers the \textit{semantic shift} problem. The goal of this thesis is to address the two problems together (bottom-right), i.e. recognizing new semantic concepts (e.g. \textit{dog}, \textit{giraffe}) in new visual domains (\textit{paintings}).} 
    \label{fig:thesis-teaser}
\end{figure}

Another major limitation of deep architectures is that their effectiveness is limited to the particular set of knowledge present in their training set, relying on the closed world assumption (CWA) \cite{sunderhauf2018limits}. This assumption rarely holds in practice and, due to the large variability of the real world, training and test images may differ significantly in terms of visual appearance, or may even contain different semantic categories. As a simple example, let us consider the scenario represented in Figure \ref{fig:thesis-teaser}. If we train a system to recognize animals (e.g. \textit{elephant} and \textit{horses}) in a given visual domain (e.g. \textit{real photos}) it will inherently assume that (i) those animals are the only animals we want to recognize and (ii) that they will always appear under the distribution of real images. What will eventually come as no surprise is that the model will struggle in distinguishing the same animals in a different visual domain (e.g. \textit{paintings}) and it will never be able to recognize animals (e.g. \textit{dog} and \textit{giraffe}) not present in its initial training set. This was a toy example but, in reality, applications where we would like to adapt a model to new input distributions and/or semantics, are countless. For example, given a robot manipulation task we cannot forecast a priori all the possible conditions (e.g. environments, lighting) it will be employed in. Moreover, we might have data only for a subset of objects we would like to recognize, at least initially. Similar reasoning applies to autonomous driving, where it is nearly impossible to collect data for every possible driving condition (e.g. weather, road), and the semantic categories we want to recognize might change with the location (e.g. region-specific animals) or purpose of the vehicle (e.g. garbage collector). 

The goal of this thesis is to address these two problems together. In particular, we want to extend the effectiveness of deep architectures to visual domains and semantic concepts \textit{not included} in the initial training set, with the long-term goal of building visual recognition systems capable of recognizing new semantic concepts in new visual domains.


\subsection{Domain shift: generalizing to new visual domains}
\label{sec:intro-da}
To recognize new semantic concepts in new visual domains, the first problem we must face is generalizing to new visual domains, by overcoming the \textit{domain shift} problem. To this extent, Domain Adaptation (DA) methods \cite{csurka2017domain,wang2018deep} are specifically designed to transfer knowledge from a \textit{source}
domain, where a large amount of labeled data are available, to a domain of interest, i.e. the \textit{target} domain where few or no labeled data are available. While standard approaches usually focus on a single-source and single-target scenario \cite{ganin2014unsupervised,long2017deep}, a large variety of settings exist depending on the information we have about our source and target domains. For instance, we might have multiple sources and/or multiple target domains, as in multi-source DA \cite{xu2018deep,MDAN_ICLRW18}, and multi-target DA \cite{chen2019blending,gholami2020unsupervised}. In these cases, a naive application of single-
source/target domain adaptation algorithms would not suffice, consequently leading to poor results. Moreover, the domains might be either explicitly divided or unified in a mixed dataset. Thus, we must \textit{discover} the various domains required for effectively addressing the domain shift problem \cite{gong2013reshaping,xiong2014latent,hoffman2012discovering}.  
While standard DA assumes that data of the target domain are available during the initial training phase, a more realistic scenario is that, initially, we do not have any image of the target domain at all. This problem arises in practice every time our systems are employed in unseen environments such as novel viewpoints, illumination, or weather conditions. There are three possible ways to tackle this problem, depending on the information we have on our target domain.

In case we have no information about our target but we have multiple source domains, we can address this problem by disentangling domain-specific and domain-agnostic components, thereby building a model robust to \textit{any} possible target domain shift. This is the goal of domain generalization (DG) that has recently raised a lot of interest in the community \cite{li2017deeper,li2019episodic,carlucci2019domain}. Differently, if we have no information about our target and a single source domain, we cannot disentangle domain and semantic specific components. In this scenario, the only feasible strategy is to dynamically adapt our model as we receive target domain data at test time, in a continuous fashion. This setting is called Continuous DA and multiple works tried to address it before the deep learning breakthrough by e.g. manifold-based techniques \cite{hoffman2014continuous} and low-rank exemplar SVM \cite{li2018domain}. 

Eventually, we could have information about the target domain shift in the form of metadata describing the visual inputs we should expect. This scenario is called Predictive DA (PDA) and assumes the presence of a single source domain and multiple auxiliary ones and that each domain has its own respective metadata \cite{yang2016multivariate}. Understanding how a metadata links to the domain-specific parameters, allows us to infer a model for any target domain given its respective description. 

The first part of this thesis describes how we provided solutions for the domain-shift problem, regardless of the information we have about our source/target domain. We started from the latent domain discovery problem, where we assume to have data of both source and target domains but with the two being mixtures of multiple hidden domains. In this particular scenario, we show how a weighted version of batch-normalization (BN) \cite{ioffe2015batch}, coupled with a domain discovery branch can equip a deep architecture with the ability to discover latent domains for DA \cite{mancini2018boosting,mancini2019inferring}. We will show how, the same domain classifier can be applied to the more complex DG task, where no data is available about our target domain. In particular, the similarity among the domains can be used either within the network (i.e. through BN layers \cite{mancini2018robust}) or at classification level \cite{mancini2018best} to effectively tackle DG. Finally, we will extend BN-based DA algorithms to the PDA scenario by relating domains and their specific parameters through a graph, where each node is a domain (with attached parameters) and the weight of each edge depends on the similarity among the domains, as given by the available metadata \cite{mancini2019adagraph}. Moreover, we provide a simple extension of BN to tackle the Continuous DA problem, showing the effectiveness of this algorithm both on challenging robotics scenarios \cite{mancini2018kitting} and as a tool to refine the target model predicted by our PDA algorithm \cite{mancini2019adagraph}.


\subsection{Semantic shift: breaking model's semantic limits}
\label{sec:intro-icl}
The second major problem we must tackle, if we want to recognize new semantic concepts in unseen domains, is to understand how to integrate novel knowledge within our deep architecture, thereby overcoming the semantic shift problem. To this extent, multiple works have tried to extend the knowledge base of a pre-trained deep model, and, depending on the information we have regarding the new concepts, we can split them into three main categories. 

In the case where we have data available for our new concepts, we are in the incremental learning scenario \cite{rebuffi2017icarl,kirkpatrick2017overcoming,li2017learning}. In incremental learning (IL), we have a pre-trained model and we receive data of the new classes/tasks in successive learning stages \textit{without} having access to the original training set. The goal is to sequentially learn new classes/tasks as new data are available while not forgetting previous knowledge, thereby addressing the catastrophic forgetting problem. 

A special case is when we want our model to not only acquire new knowledge but also to \textit{detect} unseen concepts. This is the goal of open-world recognition (OWR), where the task is to classify images if they belong to the categories of the training set, to spot samples corresponding
to unknown classes, and based on such unknown
class detections update the model to progressively
include the novel categories \cite{bendale2015towards}.

A second scenario assumes that just one or few samples are available for the novel semantic concepts. This is the case of one and few-shot learning \cite{fei2006one,vinyals2016onematching,snell2017prototypical,sung2018fewlearning}, where we make use of the available training data to build a model capable of inferring the classifier for the novel classes, given a little amount of data. Solutions to this problem usually rely on classifier regression \cite{kozerawski2018clear}, weight imprinting \cite{qi2018wi1,snell2017prototypical} and meta-learning techniques \cite{finn2017model,sun2019fewmeta}.

Finally, we might face the extreme case where no training data is available for the new categories we want to recognize. This research thread is Zero-Shot Learning \cite{lampert2013awa,akata2013label,xian2018zeroshotgood} where the goal
is to recognize semantic concepts that were not seen during training, given external information about
the novel classes. This information is available either in the form of manually annotated attributes, visual descriptions, or word embeddings \cite{akata2015evaluation,xian2018zeroshotgood}.

In the second part of the thesis, we explore ways to include novel semantic concepts within a pre-trained architecture. In particular, we start by considering multi-task/domain learning, where the goal is to sequentially learn multiple classifiers for different domains/tasks from a single pre-trained model. To this extent, we propose an algorithm based on task-specific binary masks applied on top of the parameters of the pre-trained model. We show how while requiring very few additional parameters, our algorithm achieves performance comparable to task-specific fine-tuned models. 

Furthermore, we move towards the incremental class learning scenario, considering OWR. For this, we develop the first end-to-end trainable architecture for OWR \cite{mancini2019knowledge}, based on a deep extension of non-parametric classifiers, i.e. NCM and NNO \cite{mensink2012metric,guerriero2018deep,bendale2015towards}. We also show how we can improve the performances of this algorithm by considering clustering strategies that can push samples closer to their class-specific centroid while distancing them from the ones of other classes \cite{fontanel2020boosting}.  

Finally, we explore the application of incremental class learning (ICL) techniques in the task of semantic segmentation \cite{cermelli2020modeling}. Here we discover that the performance of standard approaches is hampered by the semantic content of the background class, which changes among different incremental steps. We call this problem \textit{background semantic shift} and we provide the first solution to it through a simple yet effective modification of the logits used within standard distillation and entropy-based losses. 



\subsection{Recognizing unseen categories in unseen domains}
\label{sec:intro-dgzsl}
An open research question is whether we can address the domain and semantic shift problems together, producing a deep model able to recognize new semantic concepts in possibly unseen domains. In the third part of this thesis, we will start analyzing how we can merge these two worlds, providing a first attempt in this direction in an offline but quite extreme setting. In particular, we consider a scenario where, during training, we are given a set of images of multiple domains
and semantic categories and our goal is to build a model that can to recognize images of unseen concepts, as in ZSL, in unseen domains, as in DG. This new problem, which we called ZSL+DG, poses novel research questions which go beyond the ones of DG and ZSL problems, if taken in isolation. For instance, we can rely on the fact that multiple source domains permit to disentangle semantic and domain-specific information, as in DG. Despite this, we have no guarantee that the disentanglement will hold for the unseen semantic categories at test time. Additionally, while in ZSL it is reasonable to assume that the learned mapping between images and semantic attributes will generalize also to images of unseen concepts, in ZSL+DG we have no guarantee that this will happen for images of unseen domains. 

To tackle this problem, we propose a solution based on a variant of the well-known \textit{mixup} regularization strategy \cite{zhang2017mixup}. In particular, we show how we can use \textit{mixup} to simulate features of novel domains and semantic concepts during training, achieving state-of-the-art performances in both DG, ZSL, and in the novel ZSL+DG scenario \cite{mancini2020dgzsl}. Up to our knowledge, this is the first algorithm able to work in both worlds, recognizing unseen semantic concepts in unseen domains. 

\section{Contributions}
\label{sec:intro-contributions}
Focusing on visual recognition, this thesis contributes towards developing deep learning architectures able to cope with test images containing both different visual domains (\ie domain shift) as well as new semantic concepts (\ie semantic shift) unseen during the initial training phase. To this extent, we can divide the main contributions into three parts. The first contains techniques able to attack the well-known domain shift problem of classical DA by considering non-canonical scenarios where the amount of information regarding either the source(s) or the target(s) domains varies. The second part contains algorithms that are able to extend pre-trained architectures with new semantic concepts (i.e. tasks or classes) using external datasets not available during the initial training phase. The goal of these algorithms is to produce models capable of recognizing previously unseen concepts without hampering the performances on old ones. In the third part, we start exploring the recognition of unseen semantic concepts in unseen visual domains, presenting one of the first works merging these two worlds. In the following, we will describe the specific contributions presented in each part.

\myparagraph{Modeling the Domain Shift}
In the context of attacking the domain shift problem, we will present:
\begin{itemize}
    \item The first deep learning model capable of discovering latent domains in unsupervised domain adaptation, when the source domain is composed of a mixture of multiple visual domains~\cite{mancini2018boosting,mancini2019inferring,mancini2019discovering}. Specifically, the architecture is based on two main components, i.e. a side branch that automatically computes
the assignment of each sample to its latent domain and novel layers that exploit domain membership information to appropriately align the distribution of the CNN internal feature representations to a reference distribution.
    \item Two domain similarity-based frameworks for Domain Generalization \cite{mancini2018robust,mancini2018best}. The frameworks rely on the idea that, given a set of
different classification models associated with known domains
(e.g. corresponding to multiple environments, robots), the best
model for a new sample in the novel domain can be computed
directly at test time by optimally combining the known models. While in \cite{mancini2018robust} the combination is held out through the statistics of batch-normalization layers \cite{ioffe2015batch}, in \cite{mancini2018best} a similar principle is applied at classification level.
    \item A simple yet effective algorithm for Continuous DA in Robotics \cite{mancini2018kitting}. The algorithm is based on an online update of standard batch-normalization layers. We show the effectiveness of our algorithm on a newly collected dataset with challenging robotic scenarios, containing various illumination conditions, backgrounds, and viewpoints.
    \item The first deep learning model that can tackle Predictive DA \cite{mancini2019adagraph}. In this scenario no target data are
available and the system has to learn to generalize from
annotated source images plus unlabeled samples with associated metadata from auxiliary domains. We
inject metadata information within a deep architecture by encoding the relation between different domains
through a graph. Given the target domain metadata, our approach produces the target model by a weighted combination of the domain-specific parameters associated to the graph nodes. We also propose to refine the predicted target model through the incoming
stream of target data directly at test time, extending \cite{mancini2018kitting}. 
\end{itemize}

\myparagraph{Modeling the Semantic Shift}.
In the context of including new semantic concepts to a pre-trained architecture, we will present:
\begin{itemize}
    \item An effective algorithm performing multi-domain learning \cite{mancini2018adding,mancini2020boostingmva}. The algorithm builds on previous works by masking the weights of a pre-trained architecture through task/domain-specific binary filters \cite{mallya2018piggyback}. However, we take into account more
elaborated affine transformations of the binary masks, showing that our generalization achieves significantly higher levels of adaptation to new tasks, with performances comparable to fine-tuning strategies while requiring slightly more than 1 bit per network
parameter per additional task. With this strategy, we achieve results close to the state of the art in the Visual Domain Decathlon challenge \cite{rebuffi2017learning}.
    \item An incremental class learning algorithm for semantic segmentation which explicitly models the background semantic shift problem \cite{cermelli2020modeling}. In particular, we identify and analyze the problem of semantic shift of the background class in incremental learning for semantic segmentation. This problem arises since the background class might contain both old as well as still unseen classes. This exacerbates the catastrophic forgetting problem and hampers the ability to learn novel concepts. To tackle this issue, we propose a new distillation-based algorithm with an objective function and a classifier initialization strategy that explicitly model the semantic shift of the background class. The proposed algorithm largely outperforms standard incremental learning methods in different benchmarks.
    \item The first deep architecture that can to perform open-world recognition (OWR) \cite{mancini2019knowledge}. The proposed deep network is based on a deep extension
of a non-parametric model, \cite{bendale2015towards} and it can detect whether a perceived object
belongs to the set of categories known by the system and learns
without the need to retrain the whole system from scratch.
In a first study \cite{mancini2019knowledge}, we considered both the cases where annotated images about the new category can be provided by
an ’oracle’ (i.e. human supervision), or by autonomous mining
of the Web. In a second instance \cite{fontanel2020boosting}, we show how clustering-based techniques can boost the performances of this OWR framework.  
\end{itemize}

\myparagraph{Modeling the Semantic and Domain Shift together}.
In the context of merging the two worlds, we will describe the new ZSL+DG problem \cite{mancini2020dgzsl} where, at test time, images of unseen domains as well as unseen classes must be correctly classified. Additionally, we will present the first holistic method capable of addressing ZSL and DG individually and both combined together (ZSL+DG).
Our method is based on simulating new domains and categories during training by mixing the available training domains and classes both at the image and feature levels. The mixing strategy becomes increasingly more challenging during training, in a curriculum fashion. The extensive experimental analysis show the effectiveness of our approach in all settings: ZSL, DG, and ZSL+DG.

\section{Outline}
\label{sec:intro-outline}
\textbf{Chapter \ref{chap:da}} will discuss the domain shift problem. It will first give an overview of the problems (Section \ref{sec:da-problem}) and the related works (Section \ref{sec:da-related}), delving into the details of the Domain Alignment Layers of \cite{carlucci2017just,carlucci2017autodial}, which serve as a starting point for our works. In Section \ref{sec:da-latent}, we will describe our \textbf{multi-domain Alignment Layers} which allows us to model multiple but mixed source domains through weighted normalization and a domain classifier for unsupervised domain adaptation. In Section \ref{sec:da-dg},\ref{sec:da-continuous} and \ref{sec:da-predictive} we will consider the case where no target data are available. In particular, in Section \ref{sec:da-dg} we will extend the multi-domain Alignment Layers to the domain generalization scenario and we show how the domain classifier can be used as a proxy to merge activations from layers beyond normalization ones for effective DG. In Section \ref{sec:da-continuous}, we present \textbf{ONDA}, a continuous DA approach which makes use of continuous update of normalization statistics as target data arrive. Finally, in Section \ref{sec:da-predictive}, we present \textbf{AdaGraph}, a first deep learning-based approach for predictive domain adaptation which merges normalization statistics of different layers based on the given vectorized description of the target domain. 

\vspace{8pt}\textbf{Chapter \ref{chap:icl}} will lead us to the semantic shift problem. It will start by presenting a general problem definition (Section \ref{sec:IL-ps}) with an overview of the related works (Section \ref{sec:IL-relateds}). It will then describe \textbf{BAT} (Section \ref{sec:IL-multitask}), an approach for multi-domain learning where task-specific binary masks are affinely transformed to obtain a good trade-off among performances and parameters. In Section \ref{sec:IL-semantic-seg}, we identify the \textbf{background-shift} problem on incremental class learning for semantic segmentation and we describe \textbf{MiB}, the first method addressing it, by changing how background probabilities are treated in standard entropy losses. Finally, in Section \ref{sec:IL-owr}, we will describe the \textbf{DeepNNO}, a first deep approach for Open World Recognition, and how we can improve this model with clustering and learned rejection thresholds.

\vspace{8pt}
\textbf{Chapter \ref{chap:both}} will discuss the importance of tackling both domain and semantic shift together (Section \ref{both:problem}) and the works that pushed towards this direction (Section \ref{both:related}). We will then present a new task, \textbf{zero-shot learning under domain generalization} and a first holistic method, \textbf{CuMix}, addressing domain and semantic shift together, using increasingly more complex mixing of samples and features.

\vspace{8pt}The thesis concludes by summarizing the findings, open problems, and possible future direction of research in \textbf{Chapter \ref{chap:conclusions}}.

\section{Publications}
\label{sec:intro-publications}
In the following, the author's publications are listed in chronological order. Note that some articles (marked with *) have not been included in the thesis. 
\begin{itemize}
\item * M. Mancini, S. Rota Bul\`o, E. Ricci, B. Caputo\\ {\sl Learning Deep NBNN Representations for Robust Place Categorization}\\
       IEEE Robotics and Automation Letters, May 2017, vol. 3, n. 2., pp. 1794-1801. Presented at IEEE/RSJ International Conference on Intelligent Robots and Systems (IROS) 2017.
       
\item M. Mancini, L. Porzi, S. Rota Bul\`o, B. Caputo, E. Ricci\\ {\sl Boosting Domain Adaptation by Discovering Latent Domains}\\
       IEEE International Conference on Computer Vision and Pattern Recognition (CVPR) 2018. \textbf{(spotlight)}
       
\item M. Mancini, S. Rota Bul\`o, B. Caputo, E. Ricci\\ {\sl Robust Place Categorization with Deep Domain Generalization}\\
       IEEE Robotics and Automation Letters, July 2018, vol. 3, n. 3., pp. 2093-2100.
       
\item M. Mancini, E.Ricci, B. Caputo, S. Rota Bul\`o\\ {\sl Adding New Tasks to a Single Network with Weight Transformations using Binary Masks}\\European Computer Vision Conference Workshop on Transferring and Adapting Source Knowledge in Computer Vision 2018. \textbf{(best paper award honorable mention)}

\item M. Mancini, S. Rota Bul\`o, B. Caputo, E. Ricci\\ {\sl Best sources forward: domain generalization through source-specific nets}\\
IEEE International Conference on Image Processing (ICIP) 2018.

\item M. Mancini, H. Karaoguz, E. Ricci, P. Jensfelt, B. Caputo\\ {\sl Kitting in the Wild through Online Domain Adaptation}\\
       IEEE/RSJ International Conference on Intelligent Robots and Systems (IROS) 2018.

\item M. Mancini, H. Karaoguz, E. Ricci, P. Jensfelt, B. Caputo\\ {\sl Knowledge is Never Enough: Towards Web Aided Deep Open World Recognition}\\IEEE International Conference on Robotics and Automation (ICRA) 2019.

\item M. Mancini, S. Rota Bul\`o, B. Caputo, E. Ricci\\ {\sl AdaGraph: Unifying Predictive and Continuous Domain Adaptation through Graphs}\\IEEE/CVF International Conference on Computer Vision and Pattern Recognition (CVPR) 2019. (\textbf{oral})

\item * M. Mancini, L. Porzi, F. Cermelli, B. Caputo \\ {\sl Discovering Latent Domains for Unsupervised Domain Adaptation through Consistency}\\International Conference on Image Analysis and Processing (ICIAP) 2019.

\item * F. Cermelli, M. Mancini, E. Ricci, B. Caputo\\ {\sl The RGB-D Triathlon: Towards Agile Visual Toolboxes for Robots}\\ IEEE/RSJ International Conference on Intelligent Robots and Systems (IROS) 2019.

\item M. Mancini, L. Porzi, S. Rota Bul\`o, B. Caputo, E. Ricci\\ {\sl Inferring Latent Domains for Unsupervised Deep Domain Adaptation}\\IEEE Transactions on Pattern Analysis \& Machine Intelligence 2019.

\item * L. O. Vasconcelos, M. Mancini, D. Boscaini, B. Caputo, E. Ricci\\ {\sl Structured Domain Adaptation for 3D Keypoint Estimation}
\\International Conference on 3D Vision (3DV) 2019. (\textbf{(oral})

\item F. Cermelli, M. Mancini, E. Ricci, B. Caputo\\ {\sl Modeling the Background for Incremental Learning in Semantic Segmentation}\\ IEEE/CVF International Conference on Computer Vision and Pattern Recognition (CVPR) 2020.

\item M. Mancini, E.Ricci, B. Caputo, S. Rota Bul\`o\\ {\sl Boosting Binary Masks for Multi-Domain Learning through Affine Transformations}.\\ Machine Vision and Applications, June 2020, vol. 31, n. 6, pp. 1-14.

\item D. Fontanel, F. Cermelli, M. Mancini, S. Rota Bul\'o, E. Ricci, B. Caputo\\ {\sl Boosting Deep Open World Recognition by Clustering}\\ IEEE Robotics and Automation Letters, October 2020, vol. 5, no. 4, pp. 5985-5992. Presented at IEEE/RSJ International Conference on Intelligent Robots and Systems (IROS) 2020.

\item M. Mancini, Z. Akata, E. Ricci, B. Caputo\\ {\sl Towards Recognizing Unseen Categories in Unseen Domains}. \\European Computer Vision Conference (ECCV) 2020.

\item * L. O. Vasconcelos, M. Mancini, D. Boscaini, S. Rota Bul\'o, B. Caputo, E. Ricci\\ {\sl Shape Consistent 2D Keypoint estimation under Unsupervised Domain Adaptation}. \\International Conference on Pattern Recognition (ICPR) 2020.

\end{itemize}

\chapter{Recognition across New Visual Domains}
\label{chap:da}
\textit{This chapter presents various strategies to tackle the domain shift problem in the presence of different information regarding the source and target domains. We start by providing a general formulation of the problem (Sec.~\ref{sec:da-problem}). We then review related literature (Sec.~\ref{sec:da-related}), analyzing the Domain Alignment layers for DA (Sec.~\ref{sec:da-preliminaries}), introduced in previous works \cite{carlucci2017just,carlucci2017autodial,li2016revisiting}. In the remaining sections, we describe how we extended the Domain Alignment layers to address non-canonical DA settings. We start with the latent-domain discovery problem (Sec.~\ref{sec:da-latent}), where we have multiple source/target domains but mixed, \ie we do not know to which domain each sample belongs to. We describe the first deep learning solution to this problem  \cite{mancini2018boosting,mancini2019inferring} based on a weighted computation of the batch-normalization statistics \cite{ioffe2015batch} both at training (in case of mixed source domains) and at inference time (in case of mixed targets). In Sec.~\ref{sec:da-dg}, we show how a similar approach can be applied to tackle the domain generalization problem \cite{mancini2018robust}, removing the assumption of having target data at training time. Additionally, we show how to extend the same idea beyond batch-normalization layers, mixing activations of domain-specific classification modules \cite{mancini2018best}. In Sec.~\ref{sec:da-continuous}, we take a step further, removing the assumption of having multiple source domains during training, developing a model able to adapt to arbitrary target domains at inference time, dynamically updating its internal knowledge, in a continuous fashion \cite{mancini2018kitting}. Finally, in Sec.~\ref{sec:da-predictive}, we provide a solution to the Predictive DA scenario, where we must use multiple auxiliary domains with associated metadata during training to learn the relationship among metadata and domains. We then exploit this knowledge to generate a model for the target domain given just its description in terms of metadata. Our solution, called AdaGraph \cite{mancini2019adagraph}, is based on multiple domain-specific batch-normalization layers connected through a graph that we use at inference time to produce a model for the target domain. AdaGraph is the first deep learning-based approach to tackle the Predictive DA problem. In \cite{mancini2019adagraph}, we also extend the continuous DA approach in \cite{mancini2018kitting} to dynamically refine the predicted models at test time.}

\section{Problem statement}
\label{sec:da-problem}
As described in Section \ref{sec:intro-da}, the goal of DA algorithms is to transfer knowledge from a large labeled dataset, \ie the \textit{source} domain, to a small and/or unlabeled one, \ie the \textit{target}. In particular, throughout this work, we will focus on the case where the target domain is either fully unsupervised or not present at all during training. 

The first case is the Unsupervised Domain Adaptation problem (UDA). Formally, we can define the UDA problem as follows. Let us denote with $\mathcal{X}$ our input space (e.g. the image space), with $\mathcal{Y}$ our output space (e.g. the set of possible semantic classes) and with $\mathcal{D}$ the set of possible visual domains (e.g. environments, illumination conditions). Denoting with $\mathcal{D}^s\subset\mathcal{D}$ the set of our source domain(s), we can define our supervised training set as $\mathcal{S}=\{(x^s_i,y^s_i,s_i)\}_{i=1}^n$ where $x^s_i\in\mathcal{X}$, $y^s_i \in \mathcal{Y}$ and $s_i \in \mathcal{D}^s$. Moreover, let us define our unsupervised target dataset as $\mathcal{T}=\{(x^t_j,t_j)\}_{j=1}^m$, with $x^t_j\in\mathcal{X}$, and $t_j\in\mathcal{D}^t\subset\mathcal{D}$. Note that we assume source and target domains to differ, \ie $\mathcal{D}^t\cap\mathcal{D}^s\equiv\emptyset$. Moreover, due to the domain shift, each domain has different joint distribution defined over $\mathcal{X}\times\mathcal{Y}$: we have $p(\mathtt{x},\mathtt{y}|d_i)\neq p(\mathtt{x},\mathtt{y}|d_j)$ with $d_i\in \mathcal{D}^s\cup\mathcal{D}^t$,$d_j\in \mathcal{D}^s\cup\mathcal{D}^t$ and $d_i\neq d_j$. Our goal is to learn a mapping $f:\mathcal{X}\rightarrow \mathcal{Y}$ which is effective for each of our target domain(s) $\mathcal{D}^t$.

From our formulation, we have the standard single-source/target scenario when $|\mathcal{D}^s|=|\mathcal{D}^t|=1$, while the multi-source scenario when $|\mathcal{D}^s|>1$. In both cases, $\mathcal{T}$ is assumed available \textit{during} training. In case both $\mathcal{S}$ and $\mathcal{T}$ are available but at least one of them is composed of an unknown mixture of domains (i.e. $|\mathcal{D}^s|=\con{k_s}\geq1,|\mathcal{D}^t|=\con{k_t}\geq1$ with unknown $\con{k_s}$ and/or $\con{k_t}$), we are in the \textbf{latent domain discovery} scenario and we have no domain identifier $d$ in the triplets of $\mathcal{S}$ and $\mathcal{T}$.


In case $\mathcal{T}$ is not available during training but $|\mathcal{D}^s|>1$, we are in the \textbf{Domain Generalization (DG)} scenario. In this setting, we can exploit the presence of multiple source domains, even latent, to disentangle domain and semantic specific components from our inputs, producing a model robust to \textit{any} possible target domain.

In $\mathcal{T}$ is not available during training and $|\mathcal{D}^s|=1$, we cannot disentangle domain-specific and semantic-specific information. However, we can still cope with the domain shift problem in different ways, depending on the information we have about our target. If no information is available, we can only adapt our model at test time, while classifying samples of the target domain. This is known as the \textbf{Continuous/Online DA} scenario. 

Lastly, another scenario is \textbf{Predictive DA (PDA)}. In this case, we have a set of auxiliary domains $\mathcal{D}^a$ forming an additional training dataset $\mathcal{A}=\{(x^a_i,d^a_i)\}_{i=1}^r$. Moreover, the domain identifiers $d\in\mathcal{D}^s\cup\mathcal{D}^a$ are expressed as \textit{metadata}. Using the auxiliary set $\mathcal{A}$ and the domain metadata, we can learn a mapping among metadata and domain-specific parameters. Then, given target metadata $d^t$, we can infer its domain-specific parameters, reducing the domain shift problem.

In the following section, we will review the relevant literature for DA and each of the previously mentioned problem. As a final remark, it is worth highlighting that, in this chapter, we assume source and target domains {sharing} the \textit{same} output space $\mathcal{Y}$. In Chapter \ref{chap:icl} we will consider the case where the visual domains are shared among train and test data (i.e. $\mathcal{D}^s=\mathcal{D}^t$) but the semantic classes differ and/or varies over time. Finally, in Chapter \ref{chap:both} we will consider the scenario where both the output and the domain space differ among train and test.

\section{Related Works}
\label{sec:da-related}
In this section we will review previous works on DA. We start by reviewing DA methods, based on both hand-crafted and deep features, in standard scenarios where target domain data are available. We then review previous works tackling the domain shift problem without target domain data, starting from DG techniques and covering less explored directions, such as Continuous and Predictive DA.

\myparagraph{DA methods with hand-crafted features.} Earlier DA approaches operate on hand-crafted features and attempt to reduce
the discrepancy between the source and the target domains by adopting different strategies. For instance, instance-based methods \cite{huang2007correcting,yamada2012no,gong2013connecting} develop from the idea of learning classification/regression models by re-weighting source samples according to their similarity with the target data. A different strategy is exploited by feature-based methods, coping with domain shift by learning a common subspace for source and target data such as to obtain domain-invariant representations \cite{gong2012geodesic,long2013transfer,fernando2013unsupervised}. Parameter-based methods \cite{yang2007adapting} address the domain shift problem by discovering a set of shared weights between the source and the target models. However, they usually require labeled target data which is not always available.

While most earlier DA approaches focus on a single-source and single-target setting, some works have considered the related problem of learning classification models when the training data spans multiple domains \cite{mansour2009domain,duan2009domain,sun2011two}. The common idea behind these methods is that when source data arises from multiple distributions, adopting a single source classifier is suboptimal and improved performance can be obtained by leveraging information about multiple domains. However, these methods assume that the domain labels for all source samples are known in advance. In practice, in many applications the information about domains is hidden and latent domains must be discovered into the large training set.
Few works have considered this problem in the literature. Hoffman \etal \cite{hoffman2012discovering} address this task by modeling domains as  Gaussian distributions in the feature space and by estimating the membership of each training sample to a source domain using an iterative approach. 
Gong \etal \cite{gong2013reshaping} discover latent domains by devising a nonparametric approach which aims at
simultaneously achieving maximum distinctiveness among domains 
and ensuring that strong discriminative models are learned for each latent domain. In \cite{xiong2014latent} 
domains are modeled as manifolds and source images representations are learned decoupling information about semantic category and domain. By exploiting these representations the domain assignment labels are inferred using a mutual information based clustering method. 


\myparagraph{Deep Domain Adaptation.} Most recent works on DA consider deep architectures and robust domain-invariant features are learned using either supervised neural networks \cite{long2015learning,tzeng2015simultaneous,ganin2014unsupervised,ghifary2016deep,bousmalis2016domain,carlucci2017autodial}, deep autoencoders \cite{zeng2014deep} or generative adversarial networks \cite{bousmalis2016unsupervised,shrivastava2017learning}. Research efforts can be grouped in terms of the number of source domains available at training time.

In the single-source DA setting, we can identify two main 
 strategies. The first deals with \emph{features} and aims at learning deep domain invariant
representations. 
The idea is to introduce in the 
learning 
architecture different measures of domain
distribution shift at a single or multiple levels \cite{LongZ0J17,sun2016return,carlucci2017autodial,carlucci2017just}
and then train the network to minimize these measures while also reducing a task-specific loss, for instance for classification or detection. 
In this way the network produces 
features invariant to the domain shift, but still discriminative
for the task at hand. 
Besides distribution evaluations, other domain shift measures used similarly 
are the error in the target sample reconstruction \cite{ghifary2016deep}, or 
various 
coherence metrics 
on the pseudo-labels assigned by the source models to the target data \cite{sener2016learning,haeusser2017associative,saito2017asymmetric}. Finally, a different
group of feature-based methods rely on adversarial loss functions \cite{tzeng2015simultaneous,ganin2016domain}. The method proposed in \cite{sankaranarayanan2018generate},
that push the network to be
unable to discriminate whether a sample coming from the source or 
from the target, 
is an interesting variant of \cite{ganin2016domain}, where the domain difference is still measured at the feature 
level but passing through an image reconstruction step.  
Besides integrating the domain discrimination objective
into end-to-end classification networks, 
it has 
also 
been shown that 
two-step networks may 
have practical advantages \cite{tzeng2017adversarial,angeletti2018adaptive}.

The second popular 
deep adaptive 
strategy focuses on \emph{images}.
The described adversarial logic that demonstrated its effectiveness for feature-based methods, has also
been extended to the goal of reducing the visual domain gap. Powerful GAN \cite{goodfellow2014generative} 
methods have been exploited to generate new images or perturb existing ones to resemble the visual style of a 
certain domain, thus reducing the discrepancy at pixel level \cite{bousmalis2017unsupervised,shrivastava2017learning}.
Most of the works based on image adaptation aim at generating either target-like source images
or source-like target images, but it has been recently shown that integrating both the
transformation directions is highly beneficial \cite{russo2018sbadagan}.

In practical applications one may be offered more than one source domain. This has triggered the study of multi-sources DA algorithms.
The multi-source setting 
was initially studied from a theoretical point of view, focusing on theorems indicating how to 
optimally sub-select the data to be used in learning the source models \cite{crammer2008learning}, or 
proposing principled rules for combining the source-specific 
classifiers and obtain the ideal target class prediction \cite{mansour2009domain}. 
Several other works followed this direction in the shallow learning framework. 
When dealing with shallow-methods the na\"{\i}ve model learned by collecting all
the source data in single domain without any adaptation was usually showing low performance on the target.
It has been noticed that this behavior changes when moving to deep learning, where the larger number of 
samples as well as their variability supports generalization and usually provides good results on the target.
Only very recently two methods presented multi-source deep learning approaches that improve over this 
reference. The approach proposed in \cite{xu2018deep} builds over \cite{ganin2016domain}
by replicating the adversarial domain discriminator branch for each available source. Moreover
these discriminators are also used to get a perplexity score that indicates how the multiple
sources should be combined at test time, according to the rule in \cite{mansour2009domain}. 
A similar multi-way adversarial strategy is used in \cite{MDAN_ICLRW18}, but this work comes 
with a theoretical support that frees it from the need of respecting a specific optimal source 
combination and thus from the need of learning the source weights.

While recent deep DA methods significantly outperform approaches based on hand-crafted features, the
vast majority of them only consider single-source, single-target settings.
Moreover, almost all work presented in the literature so far assume to have direct access to multiple source domains, where in many practical applications such knowledge might not be directly available, or costly to obtain in terms of time and human annotators. 
To our knowledge, our works \cite{mancini2018boosting,mancini2019inferring} are the first works proposing a deep architecture for discovering latent source domains and exploiting them for improving classification performance on target data.

\myparagraph{Domain Generalization.} Opposite to domain adaptation \cite{csurka2017domain}, where it is assumed that target data are available in the training phase, the key idea behind DG is to learn a domain agnostic model to be applied to any unseen target domain. Although less researched than domain adaptation, the need for DG algorithms has been recognized for quite some time in the literature \cite{muandet2013domain}.

Previous DG methods can be broadly grouped into four main categories.
The first category comprises methods which attempt to learn
domain-invariant feature representations by considering specific alignment losses, such as maximum mean
discrepancy (MMD), adversarial loss or self-supervised losses. Notable approaches in this category are \cite{muandet2013domain,li2018domainadv,carlucci2019domain}. 
The second category of methods \cite{li2017deeper,khosla2012undoing} 
develop from the idea of creating deep architectures where both domain-agnostic and domain-specific parameters are learned on source domains. After training, only the domain-agnostic part is retained and used for processing target data. 
The third category devise specific optimization strategies or training procedures in order to enhance the generalization ability of the source model to unseen target data. For instance, in \cite{li2018learning} a meta-learning approach is proposed for DG. Differently, in \cite{li2019episodic} an episodic training procedure is presented to learn models robust to the domain shift. 
The latter category comprises methods which introduce 
data and feature augmentation strategies to synthesise novel samples and improve the generalization capability of the learned model \cite{shankar2018generalizing,volpi2018generalizing,volpi2019addressing}. These strategies are mostly based either on adversarial training \cite{shankar2018generalizing,volpi2018generalizing} or data augmentation \cite{volpi2019addressing}.


\myparagraph{Beyond DG: Domain Adaptation without Target Data.}  DG assumes that multiple source domains are available, in some applications this assumption might not hold. This calls for DA methods able to cope with the domain shift when i) only one source domain is available and ii) no target data are available in the training phase. Depending on their available information, these methods can work by exploiting \eg the stream of incoming target samples, or side information describing possible future target domains. Note that, differently from DG, these methods produce models which are not robust to any possible target domain, but must be re-adapted if the target domain changes

The first scenario is typically referred as {continuous} \cite{hoffman2014continuous} or {online} DA \cite{mancini2018kitting}. To address this problem, in \cite{hoffman2014continuous} a manifold-based DA technique is employed, such as to model an evolving target data distribution. In \cite{li2018domain} Li \etal propose to sequentially update a low-rank exemplar SVM classifier as data of the target domain become available. In \cite{lampert2015predicting}, the authors propose to extrapolate the target data dynamics within a reproducing kernel Hilbert space. 

The second scenario corresponds to the problem of Predictive DA (PDA). PDA is introduced in \cite{yang2016multivariate}, where a multivariate regression approach is described for learning a mapping between domain metadata and points in a Grassmanian manifold. Given this mapping and the metadata for the target domain, two different strategies are proposed to infer the target classifier. In Section \ref{sec:da-predictive}, we show how it is possible to address this task with deep architectures, using batch-normalization layers \cite{ioffe2015batch}. 

Other closely related tasks are the problems of zero shot domain adaptation and domain generalization. 
In zero-shot domain adaptation \cite{peng2018zero} the task is to learn a prediction model in the target domain 
under the assumption that task-relevant source-domain data and task-irrelevant
dual-domain paired data are available.
Domain generalization methods \cite{muandet2013domain,li2017deeper,dinnocente2018domain,motiian2017unified} attempt to learn domain-agnostic classification models by exploiting labeled source samples from multiple domains but without having access to target data. Similarly to Predictive DA, in domain generalization multiple datasets are available during training. However, in PDA data from auxiliary source domains are not labeled. 


\section{Preliminaries: Domain Alignment Layers}
\label{sec:da-preliminaries}
Batch-normalization \cite{ioffe2015batch} (BN) is a common strategy used in deep architectures for stabilizing the optimization problem, making the gradients more well-behaved, and enabling a faster and more effective training \cite{santurkar2018helpbn1,bjorck2018helpbbn2}. 
BN works by normalizing the input features to a fixed, target distribution, \ie a standard Gaussian. 
Recent works \cite{li2016revisiting,carlucci2017just,carlucci2017autodial} have shown how we can use BN layers to perform domain adaptation in a traditional batch setting. In the following, we will denote BN layers with domain-specific statistics as Domain Alignment layers (DA-layers).

DA-layers~\cite{li2016revisiting,carlucci2017just,carlucci2017autodial} are motivated by the observation that, in general, activations within a neural network follow domain-dependent distributions.
As a way to reduce domain shift, the activations are thus normalized in a domain-specific way, shifting them according to a parameterized transformation in order to match their first and second-order moments to those of a reference distribution, which is generally chosen to be normal with zero mean and unit standard deviation. While most previous works only considered settings with two domains, \ie source and target, the basic idea can be applied to any number of domains, as long as the domain membership of each sample point is known.
Specifically, denoting as $q^d_\mathtt{x}$ the distribution of activations for a given feature channel and domain $d$, an input $x^d\sim q^d_\mathtt{x}$ to the DA-layer can be normalized according to
\begin{equation}
\label{eq:da-domainalignment}
    \DA(x^d; \mu_d, \sigma_d) = \frac{x^d - \mu_d}{\sqrt{\sigma_d^2 + \epsilon}}
\end{equation}
where $\mu_d = \E_{x\sim q^d_\mathtt{x}}[x]$, $\sigma^2_d = \Var_{x\sim q^d_\mathtt{x}}[x]$ are mean and variance of the input distribution, respectively, and $\epsilon>0$ is a small constant to avoid numerical issues.
In practice, when the statistics $\mu_d$ and $\sigma^2_d$ are computed over the current mini-batch, we obtain the application of standard batch normalization separately to the sample points of each domain.

The main idea behind these works is to create a deep architecture with one parallel branch per domain, where all branches share the same parameters but embed different, domain-specific, BN layers (i.e. different statistics within DA-layers). The domain-specific BN layers align the distributions of features of different domains to the same reference distribution, achieving the desired domain adaptation effect. In the following sections, we will show how variants of DA-layers can be successfully applied in multiple distinct DA scenarios, even without the presence of target domain data during the initial training phase.

\section[Latent Domain Discovery]{Latent Domain Discovery \footnotemark\footnotetext{M. Mancini, L. Porzi, S. Rota Bul\`o, B. Caputo, E. Ricci. {\sl Boosting Domain Adaptation by Discovering Latent Domains}. IEEE International Conference on Computer Vision and Pattern Recognition (CVPR) 2018.}\footnotemark\footnotetext{M. Mancini, L. Porzi, S. Rota Bul\`o, B. Caputo, E. Ricci. {\sl Inferring Latent Domains for Unsupervised Deep Domain Adaptation}. IEEE Transactions on Pattern Analysis \& Machine Intelligence 2019.
}} 
\label{sec:da-latent}
As stated in Section \ref{sec:da-related}, the problem of Unsupervised DA has been widely studied and both theoretical results \cite{ben2010theory,mansour2009domain} and several algorithms have been developed, both considering  shallow models \cite{huang2007correcting,gong2013connecting,gong2012geodesic,long2013transfer,fernando2013unsupervised} and deep architectures \cite{long2015learning,tzeng2015simultaneous,ganin2014unsupervised,long2016unsupervised,ghifary2016deep,carlucci2017autodial,bousmalis2016domain}. While deep neural networks tend to produce more transferable and domain-invariant features with respect to shallow models, previous works have shown that the domain shift is only alleviated but not entirely removed \cite{donahue2014decaf}.

\begin{figure}[t]
  \centering
  \includegraphics[width=\columnwidth, height=9cm,trim={0cm 0cm 0cm 0cm},clip]{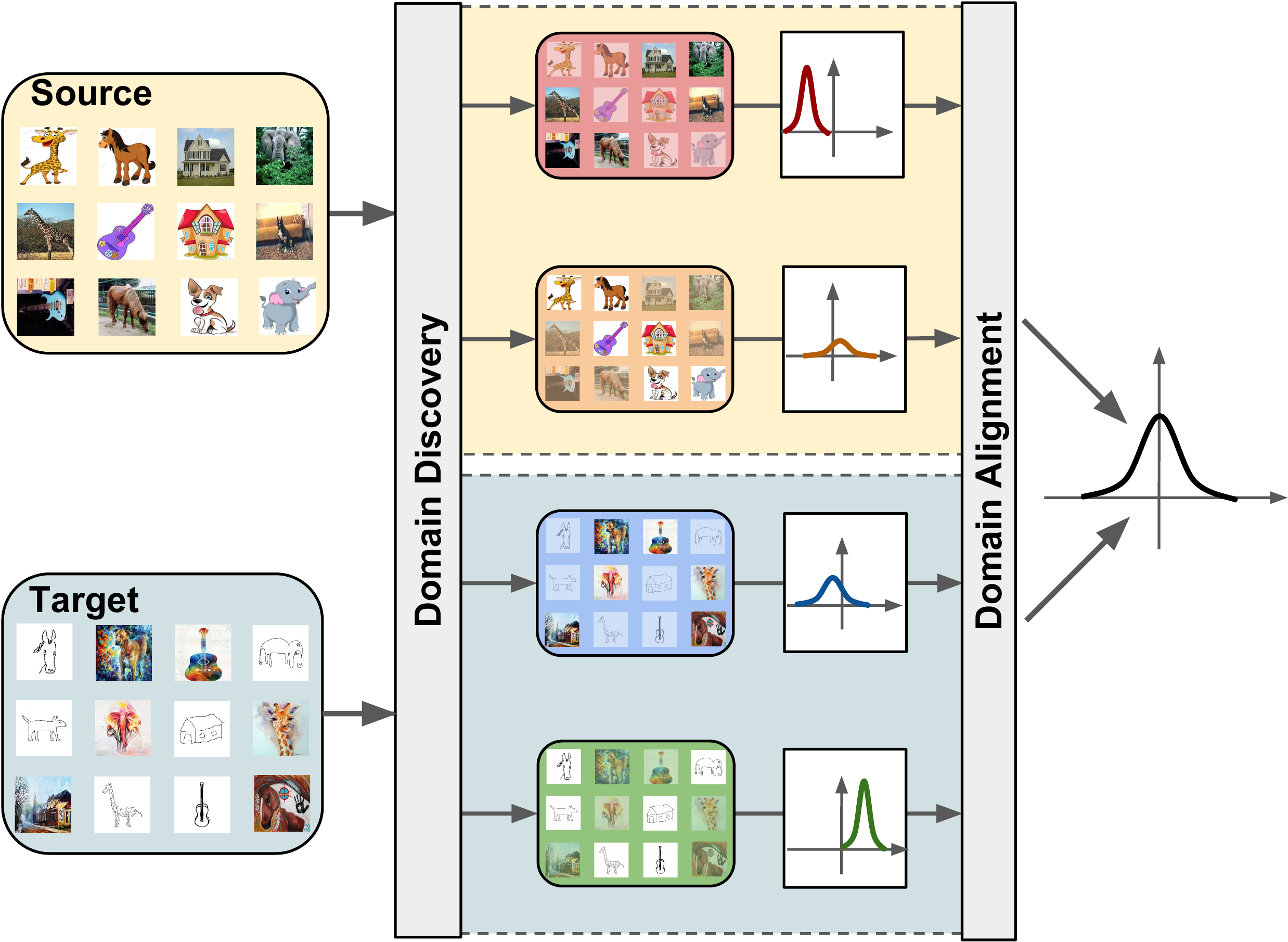}
  \caption{The idea behind the proposed framework for latent domain discovery. 
In this section, we introduce a novel deep architecture which, given a set of images, automatically discovers multiple latent 
domains
and use this information to align the distributions of the internal CNN feature representations of sources and target domains
for the purpose of domain adaptation. In this way, more accurate target classifiers can be learned. 
  }
  \label{fig:latentDA-teaser}
\end{figure}

Most previous works on UDA focus on a single-source and single-target scenario. However, in many computer vision applications 
labeled training data are often generated from multiple distributions, \ie there are multiple source domains. Examples of multi-source DA problems arise when the source set corresponds to images taken with different cameras, collected from the web or associated to multiple points of views. In these cases, a naive application of single-source domain adaptation algorithms would not suffice, leading to poor results. Analogously, target samples may arise from more than a single distribution and learning multiple target-specific models may improve significantly the performance. Therefore, in the past several research efforts have been devoted to develop domain adaptation methods considering multiple source and target domains \cite{mansour2009domain,duan2009domain,sun2011two,xu2018deep}. However, these approaches assume that the multiple domains are known. A more challenging problem arises when training data correspond to latent domains, \ie we can make a reasonable estimate on the number of source and target domains available, but we have no information, or only partial, about domain labels.
To address this problem, known in the literature as \emph{latent domain discovery}, previous works have proposed methods which simultaneously discover hidden source domains and use them to learn the target classification models~\cite{hoffman2012discovering,gong2013reshaping,xiong2014latent}.

{This section introduces the first approaches \cite{mancini2018boosting,mancini2019inferring} based on deep neural networks able to automatically discover latent domains in multi-source, multi-target UDA setting.}
Our method is inspired from the Domain Alignment Layers described in Section \ref{sec:da-preliminaries}, introduced by ~\cite{carlucci2017autodial,carlucci2017just}. 
Our approach develops from the same intuition of Domain Alignment Layers, i.e. aligning representations of source and target distributions to a reference Gaussian. However, to address the additional challenges of discovering and handling multiple latent domains, we propose a novel architecture which is able to (i) learn a set of assignment variables which associate source and target samples to a latent domain and (ii) exploit this information for aligning the distributions of the internal CNN feature representations and learn robust target classifiers (Fig.\ref{fig:latentDA-teaser}).
Our experimental evaluation shows that the proposed approach alleviates the domain discrepancy and outperforms previous UDA techniques on popular benchmarks, such as Office-31~\cite{saenko2010adapting}, PACS \cite{li2017domain} and Office-Caltech~\cite{gong2012geodesic}.

To summarize, the contributions presented in this section are threefold. Firstly, we introduce a novel deep learning approach for unsupervised domain adaptation which operates in a multi-source, multi-target setting. Secondly, we describe a novel architecture which is not only able to handle multiple domains, but also permits to automatically discover them by grouping source and target samples. Thirdly, our  experiments demonstrate that this framework is superior to 
many state-of-the-art single- and multi-source/target UDA methods.


\subsection{Problem Formulation}
We assume to have data belonging to one of several domains. Specifically, as in Section \ref{sec:da-problem}, we consider  $\con{k_s}$ \emph{source} domains, characterized by unknown probability distributions $p_{\mathtt{xy}}^{s_1},\dots,p_{\mathtt{xy}}^{s_\con{k_s}}$ defined over $\set{X}\times\set{Y}$, where $\set X$ is the input space (\eg images) and $\set Y$ the output space (\eg object or scene categories) and, similarly, we assume $\con{k_t}$ \emph{target} domains characterized by $p_{\mathtt{xy}}^{t_1},\ldots,p_{\mathtt{xy}}^{t_\con{k_t}}$. Note that, for simplicity, we wrote $=p(x,y|d)$ as $p_{\mathtt{xy}}^{d}$.
The numbers of source and target domains are not necessarily known a-priori, and are left as hyperparameters of our method.

During training we are given a set of labeled sample points from the source domains, and a set of unlabeled sample points from the target domains, while we can have partial or no information about the domain of the source sample points.
We model the source data as a set $\set{S}=\{(x_1^s,y_1^s),\dots,(x_\con{n}^s,y_\con{n}^s)\}$ of i.i.d. observations from a mixture distribution $p_{\mathtt{xy}}^s=\sum_{i=1}^\con{k_s} \pi_{s_i} p_{\mathtt{xy}}^{s_i}$, where $\pi_{s_i}$ is the unknown probability of sampling from a source domain $s_i$.
Similarly, the target sample $\set{T}=\{x_1^t,\dots,x_\con{m}^t\}$ consists of i.i.d. observations from the marginal $p_\mathtt{x}^t$ of the mixture distribution over target domains.
Furthermore, we denote by $x_\set{S}=\{x_1^s,\dots,x_\con{n}^s\}$ and $y_\set{S}=\{y_1^s,\dots,y_\con{n}^s\}$, the source data and label sets, respectively.  
We assume to know the domain label for a (possibly empty) sub-sample $\hat {\set S}\subset \set S$ from the source domains and we denote by $d_{\hat {\set S}}$ the domain labels in $\set D^s=\{s_1.\ldots,s_{\con k_s}\}$ of the sample points in $x_{\hat S}$. 
Note that, differently from the general formulation in Section \ref{sec:da-problem}, here neither $\mathcal{S}$ and $\mathcal{T}$ might have domain labels available.

Our goal is to learn a predictor that is able to classify data from the target domains. The major difficulties that this problem poses, and that we have to deal with, are: (i) the distributions of source and target domains can be drastically different, making it hard to apply a classifier learned on one domain to the others, (ii) we lack direct observation of target labels, and (iii) the assignment of each source and target sample point to its domain is unknown, or known for a very limited number of source sample points.

Several previous works~\cite{long2015learning,tzeng2015simultaneous,ganin2014unsupervised,ghifary2016deep,bousmalis2016domain,carlucci2017autodial} have tackled the related problem of domain adaptation in the context of deep neural networks, dealing with (i) and (ii) in the single domain case for both source and target data (\ie $\con {k_s}=1$ and $\con {k_t}=1$).
In particular, some recent works have demonstrated a simple yet effective approach based on the replacement of standard BN layers with specific \textit{Domain Alignment layers} ~\cite{carlucci2017just,carlucci2017autodial}.
These layers reduce internal domain shift at different levels within the network by normalizing features in a domain-dependent way, matching their distributions to a pre-determined one.
We revisit this idea in the context of multiple, unknown source and target domains and introduce a novel Multi-domain DA layer (mDA-layer) in Section~\ref{sec:latentDA-dalayers}, which is able to normalize the multi-modal feature distributions encountered in our setting.
To do this, our mDA-layers exploit a side-output branch attached to the main network (see Section~\ref{sec:latentDA-domain-prediction}), which predicts domain assignment probabilities for each input sample.
Finally, in Section~\ref{sec:latentDA-loss} we show how the predicted domain probabilities can be exploited, together with the unlabeled target samples, to construct a prior distribution over the network's parameters which is then used to define the training objective for our network.

\subsection{Multi-domain DA-layers}
\label{sec:latentDA-dalayers}

\begin{figure*}[t]
  \centering
  \includegraphics[width=1.0\textwidth,trim={0 5.8cm 0 0},clip]{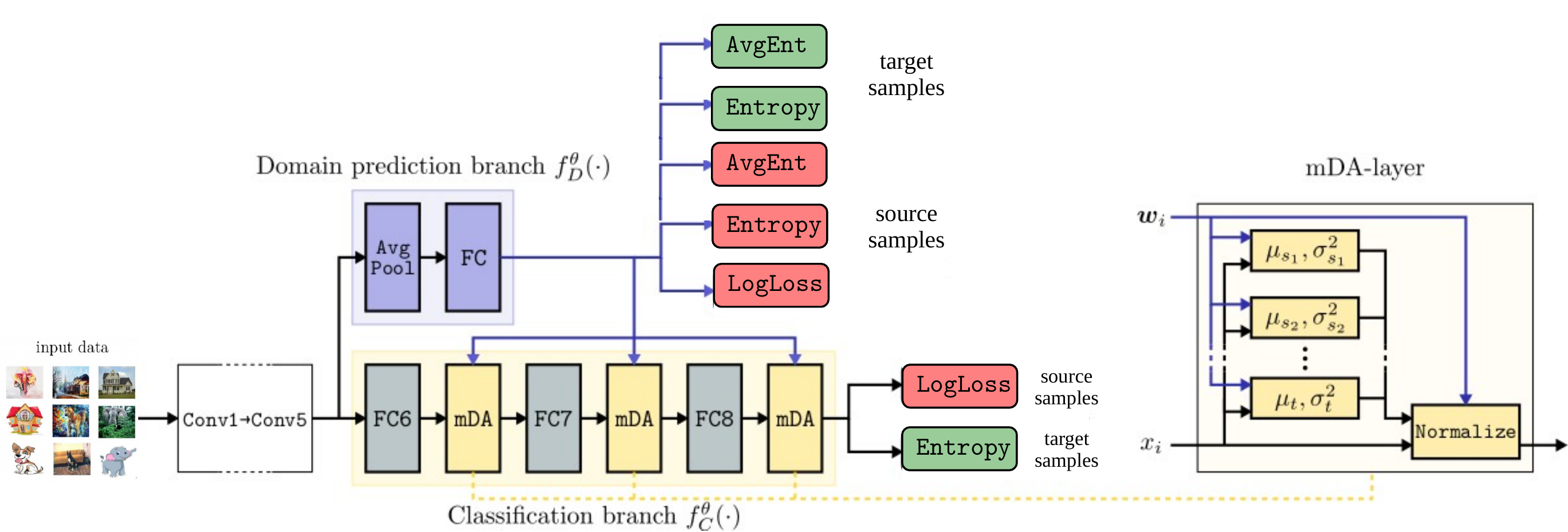}
  \caption{Schematic representation of our method applied to the AlexNet architecture (left) and of an mDA-layer (right).}
  \label{fig:latentDA-method}
\end{figure*}

In Section \ref{sec:da-preliminaries}, we described Domain Alignment Layers and how they are a simple yet effective solution for doman adaptation. However, applying them as described in Eq.~\eqref{eq:da-domainalignment} 
 requires full domain knowledge, because for each domain $d$, $\mu_d$ and $\sigma^2_d$ need to be calculated on a data sample belonging to the specific domain $d$.
In our case, however, we do not know the domain of the source/target sample points, or we have only partial knowledge about that.
To tackle this issue, we propose to model the layer's input distribution as a mixture of Gaussians, with one component per domain\footnote{Interestingly, \cite{deecke2018mode} showed how a similar strategy can be effective even within a single domain.}.
Specifically, we define a global input distribution $q_\mathtt{x} = \sum_d \pi_d q^d_\mathtt{x}$, where $\pi_d$ is the probability of sampling from domain $d$, and $q^d_\mathtt{x} = \mathcal{N}(\mu_d, \sigma^2_d)$ is the domain-specific distribution for $d$, namely a normal distribution with mean $\mu_d$ and variance $\sigma_d^2$.
Given a mini-batch $\set{B}=\{x_i\}_{i=1}^\con{b}$, a maximum likelihood estimate of the parameters $\mu_d$ and $\sigma_d^2$ is given by
\begin{equation}
\label{eqn:mixture-params}
\begin{aligned}
  \mu_d &= \sum_{i=1}^\con{b} \alpha_{i,d} x_i, &
  \sigma_d^2 &= \sum_{i=1}^\con{b} \alpha_{i,d} (x_i - \mu_d)^2,
\end{aligned}
\end{equation}
where
\begin{equation}
\label{eqn:weights}
  \alpha_{i,d} = \frac{q_\mathtt{d|x}(d\mid x_i)}{\sum_{j=1}^\con{b} q_\mathtt{d|x}(d\mid x_j)},
\end{equation}
and $q_\mathtt{d|x}(d\mid x_i)$ is the conditional probability of $x_i$ belonging to domain $d$, given $x_i$.
Clearly, the value of $q_\mathtt{d|x}$ is known for all sample points for which we have domain information.
In all other cases, the missing domain assignment probabilities are inferred from data, using the \emph{domain prediction} network branch which will be detailed in Section~\ref{sec:latentDA-domain-prediction}.
Thus, from the perspective of the alignment layer, these probabilities become an additional input, which we denote as $w_{i,d}$ for the predicted probability of $x_i$ belonging to $d$.

By substituting $w_{i,d}$ for $q_\mathtt{d|x}(d\mid x_i)$ in 
\eqref{eqn:weights}, we obtain a new set of empirical estimates for the mixture parameters, which we denote as $\hat{\mu}_d$ and $\hat{\sigma}^2_d$.
These parameters are used to normalize the layer's inputs according to
\begin{equation}
\label{eqn:normalization}
  \mDA(x_i, \vct{w}_i; \vct{\hat{\mu}}, \vct{\hat{\sigma}}) = \sum_{d\in \set D} w_{i,d} \frac{x_i - \hat{\mu}_d}{\sqrt{\hat{\sigma}_d^2 + \epsilon}},
\end{equation}
where $\vct{w}_i=\{w_{i,d}\}_{d\in\set D}$, $\vct{\hat{\mu}}=\{\hat{\mu}_d\}_{d\in\set D}$, $\vct{\hat{\sigma}}=\{\hat{\sigma}^2_d\}_{d\in\set D}$ and $\set D$ is the set of source/target latent domains.
As in previous works \cite{carlucci2017autodial,carlucci2017just,ioffe2015batch}, during back-propagation we calculate the derivatives through the statistics and weights, propagating the gradients to both the main input and the domain assignment probabilities.

\subsection{Domain prediction}
\label{sec:latentDA-domain-prediction}
Our mDA-layers receive a set of domain assignment probabilities for each input sample point, which needs to be predicted, and different mDA-layers in the network, despite having different input distributions, share consistently the same domain assignment for the sample points. 
As a practical example, in the typical case in which mDA-layers are used in a CNN to normalize convolutional activations, the network would predict a single set of domain assignment probabilities for each input image, which would then be fed to all mDA-layers and broadcasted across all spatial locations and feature channels corresponding to that image.
We compute domain assignment probabilities using a distinct section of the network, which we call the \emph{domain prediction} branch, while we refer to the main section of the network as the \emph{classification} branch.
The two branches share the bottom-most layers and parameters as depicted in Figure~\ref{fig:latentDA-method}.

The domain prediction branch is implemented as a minimal set of layers followed by two softmax operations with $\con{k_s}$ and $\con{k_t}$ outputs for the source and target latent domains, respectively (more details follow in Section~\ref{sec:latentDA-experiments}). The rationale of keeping the domain prediction separated between source and target derives from the knowledge that we have about the source/target membership of a sample point that we receive in input, while it remains unknown the specific source or target domain it belongs to.
Furthermore, for each sample point $x_i$ with known domain membership $\hat d$, we fix in each mDA-layer $w_{i,d}=1$ if $d=\hat d$, otherwise $w_{i,d}=0$.

We split the network into a domain prediction branch and classification branch at some low level layer.
This choice is motivated by the observation~\cite{aljundi2016lightweight} that features tend to become increasingly more domain invariant going deeper into the network, meaning that it becomes increasingly harder to compute a domain membership as a function of deeper features.
In fact, as pointed out in~\cite{carlucci2017autodial}, this phenomenon is even more evident in networks that include DA-layers.

\subsection{Training the network}
\label{sec:latentDA-loss}
{In order to exploit unlabeled data within our discriminative setting, we follow the approach sketched in~\cite{carlucci2017autodial}, where unlabeled data is used to define a regularizer over the network's parameters.
By doing so, we obtain a loss for $\theta$ that takes the following form:
\begin{equation}
\label{eq:latent-loss_general}
\begin{aligned}
  L(\theta) = L_\text{cls}(\theta)+L_\text{dom}(\theta)\,,
\end{aligned}
\end{equation}
where $L_\text{cls}$ is a loss term that penalizes based on the final classification task, while $L_\text{dom}$ accounts for the domain classification task.}

{\myparagraph{Classification loss $L_\text{cls}$.}
The classification loss consists of two components, accounting for the the supervised sample from the source domain $\set S$ and the unlabeled target sample $\set T$, respectively:
\begin{equation}
\label{eqn:loss_cls}
\begin{aligned}
  L_\text{cls}(\theta)=&- \frac{1}{\con{n}} \sum_{i=1}^\con{n} \log f_C^\theta(y_i^s; x_i^s)+\frac{\lambda_C}{\con{m}} \sum_{i=1}^\con{m} H(f_C^\theta(\cdot;x_i^t)).
\end{aligned}
\end{equation}
The first term on the right-hand-side is the average log-loss related to the supervised examples in $\set S$, where $f_C^\theta(y_i^s; x_i^s)$ denotes the output of the \emph{classification branch} of the network for a source sample, \ie the predicted probability of $x_i^s$ having class $y_i^s$. The second term on the right-hand-side of \eqref{eqn:loss_cls} is the entropy $H$ of the classification distribution $f_C^\theta(\cdot; x_i^t)$, averaged over all unlabeled target examples $x_i^t$ in $\set T$, scaled by a positive hyperparameter $\lambda_C$.}

{\myparagraph{Domain loss $L_\text{dom}$.}
Akin to the classification loss, the domain loss presents a component exploiting the supervision deriving from the known domain labels in $\hat{\set S}$ and a component exploiting the domain classification distribution on all sample points lacking supervision. However, the domain loss has in addition a term that tries to balance the distribution of sample points across domains, in order to avoid predictions to collapse into trivial solutions such as constant assignments to a single domain. 
Accordingly, the loss takes the following form:
\begin{multline}
\label{eqn:loss_dom}
  L_\text{dom}(\theta)=-\frac{\lambda_D}{|\hat{\set{S}}|} \sum_{x_i \in x_{\hat {\set{S}}}} \log f_{D_s}^\theta(d_i; x_i)\\
  -\lambda_B H(\bar f_{D_s}^\theta(\cdot)) +\frac{\lambda_E}{|\set S\setminus\hat{\set S}|} \sum_{x\in x_{\set{S}\setminus\hat{\set S}}} H(f_{D_s}^\theta(\cdot; x))\\
  -\lambda_B H(\bar f_{D_t}^\theta(\cdot)) +\frac{\lambda_E}{\con m} \sum_{i=1}^\con m H(f_{D_t}^\theta(\cdot; x_i^t)).
\end{multline}
Here, $f_{D_s}^\theta$ and $f_{D_t}^\theta$ denote the outputs of the \emph{domain prediction branch} for data points from the source and target domains, respectively, while
$\bar f_{D_s}^\theta$  and $\bar f_{D_t}^\theta$ denote the distributions of predicted domain classes across $\set S$ and $\set T$, respectively, \ie
\[
\bar f_{D_s}^\theta(y) = \frac{1}{\con n}\sum_{i=1}^\con n f_{D_s}^\theta(y; x^s_i),\quad
\bar f_{D_t}^\theta(y) = \frac{1}{\con m}\sum_{i=1}^\con m f_{D_t}^\theta(y; x^t_i)\,.
\]
The first term in \eqref{eqn:loss_dom} enforces the correct domain prediction on the sample points with known domain and it is scaled by a positive hyperparameter $\lambda_D$.
The terms scaled by the positive hyperparameter $\lambda_E$ enforce domain predictions with low uncertainty for the data points with unknown domain labels, by minimizing the entropy of the output distribution. 
Finally, the terms scaled by the positive hyperparameter $\lambda_B$ enforce balanced distributions of predicted domain classes across the source and target sample, by maximizing the entropy of the averaged distribution of domain predictions. 
Interestingly, since the classification branch has a dependence on the domain prediction branch via the mDA-layers, by optimizing the proposed loss, the network learns to predict domain assignment probabilities that result in a low classification loss.
In other words, the network is free to predict domain memberships that do not necessarily reflect the real ones, as long as this helps improving its classification performance.}

{We optimize the loss in \eqref{eq:latent-loss_general} with stochastic gradient descent. Hence, the samples $\set S$, $\set T$, $\hat{\set S}$ that are considered in the computation of the gradients are restricted to a random subsets contained in the mini-batch. In Section~\ref{sec:latentDA-experiments} we provide more details on how each mini-batch is sampled. We call our model \textbf{m}ulti-\textbf{D}omain \textbf{A}lignment layers for latent domain discovery (mDA).}

\subsection{Experimental results}
\label{sec:latentDA-experiments}

\subsubsection{Datasets}
\label{sec:latentDA-experiments-datasets}
In our evaluation we consider several common DA benchmarks: the combination of USPS~\cite{friedman2001elements}, MNIST~\cite{lecun1998gradient} and MNIST-m~\cite{ganin2014unsupervised}; the Digits-five benchmark in~\cite{xu2018deep}; Office-31~\cite{saenko2010adapting}; Office-Caltech~\cite{gong2012geodesic} and PACS~\cite{li2017deeper}.

\myparagraph{MNIST, MNIST-m and USPS}
are three standard datasets for digits recognition.
USPS \cite{friedman2001elements} is a dataset of digits scanned from U.S. envelopes, MNIST~\cite{lecun1998gradient} is a popular benchmark for digits recognition and MNIST-m~\cite{ganin2014unsupervised} its counterpart obtained by blending the original images with colored patches extracted from BSD500 photos~\cite{arbelaez2011contour}. 
Due to their different representations (\eg colored vs gray-scale), these datasets have been adopted as a DA benchmark by many previous works \cite{ganin2014unsupervised,bousmalis2016domain,bousmalis2016unsupervised}.
Here, we consider a multi source DA setting, using MNIST and MNIST-m as sources and USPS as target, training on the union of the training sets and testing on the test set of USPS.

\myparagraph{Digits-five} is an experimental setting proposed in \cite{xu2018deep} which considers 5 datasets of digits recognition.
In addition to MNIST, MNST-m and USPS, it includes SVHN~\cite{netzer2011reading} and Synthetic numbers datasets~\cite{ganin2016domain}.
SVHN~\cite{netzer2011reading} contains pictures of real-world house numbers, collected from Google Street View.
Synthetic numbers~\cite{ganin2016domain} is built from computer generated digits, including multiple sources of variations (\ie position, orientation, background, color and amount of blur), for a total of 500 thousands images.
We follow the experimental setting described in~\cite{xu2018deep}: the train\slash{}test split comprises a subset of 25000 images for training and 9000 for testing for each of the domains, except for USPS for which the entire dataset is used.
As in~\cite{xu2018deep}, we report the results when either SVHN or MNIST-m are used as targets and all the other domains are taken as sources.

\myparagraph{Office-31} is a standard DA benchmark which contains images of 31 object categories collected from 3 different sources: Webcam (W), DSLR camera (D) and the Amazon website (A).
Following~\cite{xiong2014latent}, we perform our tests in the multi-source setting, where each domain is in turn considered as target, while the others are used as source.

\myparagraph{Office-Caltech}~\cite{gong2012geodesic} is obtained by selecting the subset of $10$ common categories in the Office31 and the Caltech256~\cite{griffin2007caltech} datasets.
It contains $2533$ images, about half of which belong to Caltech256.
The different domains are Amazon (A), DSLR (D), Webcam (W) and Caltech256 (C).
In our experiments we consider the set of source\slash{}target combinations used in~\cite{gong2013reshaping}.

\myparagraph{PACS}~\cite{li2017deeper} is a recently proposed DA benchmark which is especially interesting due to the significant domain shift between its domains.
It contains images of 7 categories (\textit{dog, elephant, giraffe, guitar, horse}) and 4 different visual styles: \ie Photo (P), Art paintings (A), Cartoon (C) and Sketch (S).
We employ the dataset in two different settings.
First, following the experimental protocol in~\cite{li2017deeper}, we train our model considering 3 domains as sources and the remaining as target, using all the images of each domain.
Differently from \cite{li2017deeper} we consider a DA setting (\ie target data is available at training time) and we do not address the problem of domain generalization.
Second, we use 2 domains as sources and the remaining 2 as targets, in a multi-source multi-target scenario.
In this setting the results are reported as average accuracy between the 2 target domains.

{In all experiments and settings, we assume to have no domain labels (\ie $\hat{\set{S}}=\emptyset$), unless otherwise stated.}

\subsubsection{Networks and training protocols}
\label{sec:latentDA-experiments-networks}
We apply our approach to four different CNN architectures: the MNIST and SVHN networks described in~\cite{ganin2014unsupervised,ganin2016domain}, AlexNet~\cite{krizhevsky2012imagenet} and ResNet~\cite{he2016deep}.
We choose AlexNet due to its widespread use in many relevant DA works~\cite{ganin2014unsupervised,carlucci2017autodial,long2015learning,long2016unsupervised}, while ResNet is taken as an exemplar for modern state-of-the-art architectures employing batch-normalization layers.
Both AlexNet and ResNet are first pre-trained on ImageNet and then fine-tuned on the datasets of interest.
The MNIST and SVHN architectures are chosen for fair comparison with previous works considering digits datasets~\cite{ganin2016domain,xu2018deep}. 
Unless otherwise noted, we optimize our networks using Stochastic Gradient Descent with momentum $0.9$ and weight decay $5\times10^{-4}$.

For the evaluation on MNIST, MNIST-m and USPS datasets, we employ the MNIST network described in~\cite{ganin2014unsupervised}, adding an mDA-layer after each convolutional and fully-connected layer.
The domain prediction branch is attached to the output of \texttt{conv1}, and is composed of a convolution with the same meta-parameters as \texttt{conv2}, a global average pooling, a fully-connected layer with 100 output channels and finally a fully-connected classifier.
Following the protocol described in~\cite{carlucci2017autodial,ganin2014unsupervised}, we set the initial learning rate $l_0$ to 0.01 and we anneal it through a schedule $l_p$ defined by $l_p= \frac{l_0}{(1+\gamma p)^\beta}$ where $\beta=0.75$, $\gamma=10$ and $p$ is the training progress increasing linearly from 0 to 1.
We rescale the input images to $32\times 32$ pixels, subtract the per-pixel image mean of the dataset and feed the networks with random crops of size $28\times 28$.
A batch size of 128 images per domain is used.

For the Digits-five experiments we employ the SVHN architecture of~\cite{ganin2016domain}, which is the same architecture adopted by~\cite{xu2018deep}, augmented with mDA-layers and a domain prediction branch in the same way as the MNIST network described in the previous paragraph.
We train the architecture for 44000 iterations, with a batch size of 32 images per domain, an initial learning rate of $10^{-4}$ which is decayed by a factor of 10 after 80\% of the training process.
We use Adam as optimizer with a weight decay $5\times10^{-5}$, and pre-process the input images like in the MNIST, MNIST-m, USPS experiments.

For the experiments on Office-31 and Office-Caltech we employ the AlexNet architecture.
We follow a setup similar to the one proposed in~\cite{carlucci2017autodial,carlucci2017just}, fixing the parameters of all convolutional layers and inserting mDA-layers after each fully-connected layer and before their corresponding activation functions.
The domain prediction branch is attached to the last pooling layer \texttt{pool5}, and is composed of a global average pooling, followed by a fully connected classifier to produce the final domain probabilities.
The training schedule and hyperparameters are set following~\cite{carlucci2017autodial}. 

For the experiments on the PACS dataset we consider the ResNet architecture in the 18-layers setup described in~\cite{he2016deep}, denoted as ResNet18.
This architecture comprises an initial $7\times 7$ convolution, denoted as \texttt{conv1}, followed by 4 main modules, denoted as \texttt{conv2} -- \texttt{conv5}, each containing two residual blocks.
To apply our approach, we replace each Batch Normalization layer in the residual blocks of the network with an mDA-layer.
The domain prediction branch is attached to \texttt{conv1}, after the pooling operation.
The branch is composed of a residual block with the same structure as \texttt{conv2}, followed by global average pooling and a fully connected classifier.
In the multi-target experiments we add a second, identical domain prediction branch to discriminate between target domains.
We also add a standard BN layer after the final domain classifiers, which we found leads to a more stable training process in the multi-target case.
In both cases, we adopt the same training meta-parameters as for AlexNet, with the exception of weight-decay which is set to $10^{-6}$ and learning rate which is set to $5\cdot10^{-4}$.
The network is trained for 600 iterations with a batch size of 48, equally divided between the domains, and the learning rate is scaled by a factor 0.1 after 75\% of the iterations.

Regarding the hyperparameters of our method,  we set the number of source domains $\con{k}$ equal to $Q-1$, where $Q$ is the number of different datasets used in each single experiment.
In the multi-source multi-target scenarios, since we always have the domains equally split between source and target, we consider $\con{k}$ equal $Q/2$ for both source and target.
Following \cite{carlucci2017autodial}, in the experiments with AlexNet we fix $\lambda_C=\lambda_E=0.2$ with $\lambda_B=0.1$.
Similarly, for the experiments on digits classification, we set $\lambda_C=\lambda_E=0.1$ and $\lambda_B=0.05$ for MNIST, MNIST-m and USPS, and $\lambda_C=0.01$ and $\lambda_E=\lambda_B=0.05$ for Digits-five, with $\lambda_E=0.01$ if $\lambda_B=0$, which we found leading to a more stable minimization of the loss of the domain branch.
In the experiments involving ResNet18 we select the values $\lambda_C=0.1$ and $\lambda_E=\lambda_B=0.0001$ through cross-validation, following the procedure adopted in \cite{long2013transfer,carlucci2017autodial}.
Similarly, in the multi-target ResNet18 experiments we select $\lambda_C=\lambda_E=\lambda_B=0.1$. 
When domain labels are available for a subset of source samples, we fix $\lambda_D=0.5$. 

We implement\footnote{Code available at: \url{https://github.com/mancinimassimiliano/latent_domains_DA.git}} all the models with the Caffe~\cite{jia2014caffe} framework and our evaluation is performed using an NVIDIA GeForce 1070 GTX GPU. 
We initialize both AlexNet and ResNet18 from models pre-trained on ImageNet, taking AlexNet from the Caffe model zoo, and converting ResNet18 from the original Torch model\footnote{\scriptsize\url{ https://github.com/HolmesShuan/ResNet-18-Caffemodel-on-ImageNet}}. 
{For all the networks and experiments, we add mDA layers and their variants in place of standard BN layers}.

\subsubsection{Results}
In this section, we first analyze the proposed approach, demonstrating the advantages of considering multiple sources\slash{}targets and discovering latent domains.
We then compare the proposed method with state-of-the-art approaches.
For all the experiments we report the results in terms of accuracy, repeating the experiments at least 5 times and averaging the results.
In the multi-target experiments, the reported accuracy is the average of the accuracies over the target domains. {As for standard deviations, since we do not tune the hyperparameters of our model and baselines by employing the accuracy on the target domain, their values can be high in some settings. For this reason, in order to provide a more appropriate analysis of the significance of our results, we propose to adopt the following approach. In particular, let us model the accuracy of an algorithm as a random variable $X_a$ with unknown distribution. The accuracy of a single run of the algorithm is an observation from this distribution. Therefore, in order to compare two algorithms we consider the two sets of associated observations $A=\{a_1,\dots,a_n\}$ and $B=\{b_1,\dots,b_m\}$ and estimate the probability that one algorithm is better than the other as: 
\[
p(X_a>X_b)=\frac{\sum_{a \in A}\sum_{b \in B} \delta(a>b)}{|A|\cdot|B|}
\] 
where $\delta$ is the Dirac function. 
In the following we use this metric to compare our approach with respect to a baseline where no latent domain discovery process is implemented (specifically, the method DIAL~\cite{carlucci2017just}, see below) considering five runs for each experiment. For sake of clarity, we denote this probability estimate as $p^*$. 
}

{In the following we first analyze the performances of the proposed approach with $\lambda_B=0$ (denoted as mDA $\lambda_B=0$), \ie the algorithm we presented in \cite{mancini2018boosting}, and then we describe the impact of the loss term we introduce in this section setting $\lambda_B>0$ (denoted simply as mDA).}

\myparagraph{Experiments on the Digits datasets}

In a first series of experiments, reported in Table~\ref{tab:latentDA-digits}, we test the performance of our approach on the MNIST, MNIST-m to USPS benchmark (M-Mm to U).
The comparison includes: (i) the baseline network trained on the union of all source domains (\textit{Unified sources}); (ii) training separate networks for each source, and selecting the one the performs the best on the target (\textit{Best single source}); (iii) DIAL~\cite{carlucci2017just}, trained on the union of the sources (\textit{DIAL \cite{carlucci2017just} - Unified sources}); (iv) DIAL, trained separately on each source and selecting the best performing model on the target (\textit{DIAL \cite{carlucci2017just} - Best single source}).
We also report the results of our approach in the ideal case where the multiple source domains are known and we do not need to discover them (\textit{Multi-source DA}).
For our approach with $\lambda_B=0$, we consider several different values of $\con{k}$, \ie the number of discovered source domains.

By looking at the table several observations can be made.
First, there is a large performance gap between models trained only on source data and DA methods, confirming that deep architectures by themselves are not enough to solve the domain shift problem~\cite{donahue2014decaf}.
Second, in analogy with previous works on DA~\cite{mansour2009domain,duan2009domain,sun2011two}, we found that considering multiple sources is beneficial for reducing the domain shift with respect to learning a model on the unified source set.
Finally, and more importantly, when the domain labels are not available, our approach is successful in discovering latent domains and in exploiting this information for improving accuracy on target data, partially filling the performance gap between the single source models and \textit{Multi-source DA}. 
Interestingly, the performance of our algorithm changes only slightly for different values of $\con{k}$, motivating our choice to always fix $\con{k}$ to the known number of domains in the next experiments. {Importantly, comparing our approach with DIAL we achieve higher accuracy in most of the runs, \ie  $p^*=0.65$}. {In this experiment, the introduction of the loss term forcing a uniform assignment among clusters (denoted as mDA) leads to comparable performances to our method with $\lambda_B=0$. This behaviour can be ascribed to the fact that the separation among different domains is quite clear in this case and adding constraints to the domain discovery process is not required. In the following, we show that the proposed loss is beneficial in more challenging datasets.} 



\begin{table}[t]
			\caption{Digits datasets: comparison of different models in the multi-source scenario. MNIST (M) and MNIST-m (Mm) are taken as source domains, USPS (U) as target.} 
		\centering
		\scalebox{1.}{
		\begin{tabular}{ l | c} 
			\hline
			Method & M-Mm to U \\
            	\hline
                Unified sources &  57.1 \\
                Best single source & 59.8  \\
			DIAL \cite{carlucci2017just} - Unified sources &81.7 
            \\
          {DIAL \cite{carlucci2017just} - Best single source} & 81.9 \\
mDA $\lambda_B=0$ $\con{k}=2$& 	82.5 \\
mDA $\lambda_B=0$ $\con{k}=3$& 	82.2 \\
mDA $\lambda_B=0$ $\con{k}=4$& 	82.7 \\
mDA $\lambda_B=0$ $\con{k}=5$& 82.4 \\
mDA ($\con{k}=2$)&82.4  \\\hline\hline
{Multi-source DA}&84.2
\\ \hline 
		\end{tabular}
        }
		\label{tab:latentDA-digits}
\end{table}

In a second set of experiments (Table~\ref{tab:latentDA-digits-5}), we compare our approach with previous and recently proposed single and multi-source unsupervised DA approaches.
Following~\cite{xu2018deep}, we perform experiments on the Digits-five dataset, considering two settings with SVHN and MNIST-m as targets.
As in the previous case, we evaluate the performance of the baseline network (with and without BN layers) and of DIAL when trained on the union of the sources, and, as an upper bound, our Multi-source DA with perfect domain knowledge.
Moreover, we consider the Deep Cocktail Network (DCTN)~\cite{xu2018deep} multi-source DA model, as well as the ``source only'' baseline and the single source DA models reported in~\cite{xu2018deep}: Reverse gradient (RevGrad) \cite{ganin2014unsupervised} and Domain Adaptation Networks (DAN) \cite{long2015learning}.
For all single source DA models we consider two settings: ``Unified Sources'', where all source domains are merged, and ``Multi-Source'', where a separate model is trained for each source domain, and the final prediction is computed as an ensemble.
As we can see, the Unified Sources DIAL already achieves remarkable results in this setting, outperforming DCTN, and Multi-source DA only provides a modest performance increase.
As expected, the performance of our approach lies between these two ($p^*$ equal to 0.56 and 0.64 for SVHN and MNIST-m respectively, with $\lambda_B=0$).


\begin{table}[t]
			\caption{Digits-five~\cite{xu2018deep} setting, comparison of different single source and multi-source DA models. The first row indicates the target domain with the others used as sources.} 
		\centering
		\scalebox{1.}{
		\begin{tabular}{l | l | c  c | c } 
			\hline
			&Method  & SVHN & MNIST-m & Mean\\
            	\hline
               \multirow{6}{*}{Unified sources} & Source only)&74.1&64.4&69.3\\ 
                &Source only+BN &77.7&59.4&68.6\\
                &Source only from~\cite{xu2018deep}&72.2&64.1&68.2\\
            	&RevGrad \cite{ganin2014unsupervised}&68.9&71.6&70.3\\
                &DAN \cite{long2015learning}&71.0&66.6&68.8\\
                &DIAL \cite{carlucci2017just}&82.2&68.8&75.5\\
                \hline
              &mDA $\lambda_B=0$ &{82.4}&{69.1}&{75.8}\\
               &mDA &\textbf{82.6}&\textbf{70.1}&\textbf{76.4}\\
                \hline\hline
                 \multirow{6}{*}{Multi-source} &Only Source \cite{xu2018deep}&64.6&60.7&62.7\\
               & RevGRAD \cite{ganin2014unsupervised}&61.4&71.1&66.3\\
                &DAN \cite{long2015learning}&62.9&62.6&62.8\\
                &DCTN \cite{xu2018deep}&77.5&70.9&74.2\\
                &Multi-source DA&84.1&69.4&76.8\\
                \hline
		\end{tabular}
        }
		\label{tab:latentDA-digits-5}
\end{table}
\subsubsection{Experiments on PACS}
         \vspace{-10pt}
\myparagraph{Comparison with state of the art.}
In our main PACS experiments we compare the proposed approach with the baseline ResNet18 network, and with ResNet18 + DIAL~\cite{carlucci2017just}, both trained on the union of source sets.
As in the digits experiments, we also report the performance of our method when perfect domain knowledge is available (Multi-source DA).
Table \ref{tab:latentDA-pacs} shows our results.
In general, DA models are especially beneficial when considering the PACS dataset, and multi-source DA networks significantly outperform the single source one.
Remarkably, our model is able to infer domain information automatically without supervision.
In fact, its accuracy is either comparable with Multi-source DA (Photo, Art and Cartoon) or in between DIAL and Multi-source DA (Sketch). {The average $p^*$ is $0.67$.
Looking at the partial results, it is interesting to note that the improvements of our approach and Multi-source DA w.r.t. DIAL are more significant when either the Sketch or the Cartoon domains are employed as target set (average $p^*=0.81$).
Since these domains are less represented in the ImageNet database, we believe that the corresponding features derived from the pre-trained model are less discriminative, and DA methods based on multiple sources become more effective. Setting $\lambda_B>0$, allows to obtain a further boost of performances in the Sketch scenario, where the source domains are closer in appearances. In the other settings, the domain shift is mostly among the Sketch domain and all the others and it can be easily captured by our original formulation in \cite{mancini2018boosting}.}

\begin{table}[t]
			\caption{PACS dataset: comparison of different methods using the ResNet architecture. The first row indicates the target domain, while all the others are considered as sources. } 
		\centering
		\scalebox{1.}{
		\begin{tabular}{ l | c  c  c  c | c  } 
			\hline
			Method & Sketch & Photo & Art & Cartoon & Mean \\
            	\hline
                ResNet \cite{he2016deep} &60.1&92.9&74.7&72.4&75.0\\
DIAL \cite{carlucci2017just} &66.8&\textbf{97.0} &87.3 &85.5&84.2 \\
mDA $\lambda_B=0$ &{69.6}&\textbf{97.0}&\textbf{87.7}&\textbf{86.9}&{85.3}\\
mDA   &\textbf{70.7} &\textbf{97.0} &87.4 &86.3 &\textbf{85.4} \\\hline\hline
Multi-source DA & 71.6 & 96.6  & 87.5 & 87.0 & 85.7 \\ \hline 
\hline
		\end{tabular}
        }
		\label{tab:latentDA-pacs}
\end{table}

To analyze the performances of our approach in a multi-source multi-target scenario, we perform a second set of experiments on the PACS dataset considering 2 domains as sources and the other 2 as targets.
The results, shown in Table~\ref{tab:latentDA-pacs-multitarget}, comprise the same baselines as in Table~\ref{tab:latentDA-pacs}.
Note that, apart from the difficulty of providing useful domain assignments both in the source and target sets during training, the domain prediction step is required even at test time, thus having a larger impact on the final performances of the model.
The performance gap between DIAL and our approach increases in this setting compared to Table~\ref{tab:latentDA-pacs}.
Our hypothesis is that not accounting for multiple domains has a larger impact on the \emph{unlabeled} target than on the \emph{labeled} source.
Looking at the partial results, when Photo is considered as one of the target domains there are no particular differences in the final performances of the various DA models: this may be caused by the bias of the pre-trained network towards this domain. However, when the other domains are considered as targets, the gain in performances produced by our model are remarkable. When Sketch is one of the target domains, our model 
completely fills the gap between the unified source\slash{}target DA method and the multi-source multi-target upper bound with a gain of more then 7\% when Art and Cartoon considered as other target. {Setting $\lambda_B > 0$ in this setting allows to obtain a further boost of performances. This is evident in the scenario where Photo and Art are both the source or target domains, with Cartoon-Sketch correspond to the other pair. In this scenario the source/target pairs are quite close and enforcing a uniform assignment among the latent domains provides a better estimate of each of them.}

\begin{table*}[t]
			\caption{PACS dataset: comparison of different methods using the ResNet architecture on the multi-source multi-target setting. The first row indicates the two target domains. }
		\centering
		\scalebox{.88}{
		\begin{tabular}{ l | c c c c c c | c } 
			\hline
			 \multirow{2}{*}{Method/Targets} & Photo& Photo& Photo& Art& Art& Cartoon& \multirow{2}{*}{Mean}\\
			  & Art & Cartoon & Sketch&Cartoon&Sketch&Sketch &\\
            	\hline
                ResNet \cite{he2016deep} &71.4&84.2&81.4&62.2&70.3&54.2&70.6\\
DIAL \cite{carlucci2017just} &{86.7} &86.5 &86.8 &77.1 &72.1 &67.7 &79.5 \\
Random assignment &86.6&86.7&85.9&76.2&69.1&69.4&79.1\\
mDA $\lambda_E=\lambda_B=0$ &86.8&86.5&86.7&78.6&73.8&68.7&80.2\\
mDA $\lambda_B=\lambda_C=0$& 82.4&85.0&83.7&71.7&74.0&68.8&76.4\\
mDA $\lambda_B=0$ &86.1&{87.9}&{87.9}&\textbf{79.3}&{79.9}&{74.9}&{82.6}\\


mDA &\textbf{87.2}&\textbf{88.1} &\textbf{88.7} &77.7&\textbf{81.3}&\textbf{77.0}&\textbf{83.3} \\\hline\hline
Multi-source/target DA & 87.7&88.9&86.8&79.0&79.8 &75.6&83.0\\ \hline 
		\end{tabular}
        }
		\label{tab:latentDA-pacs-multitarget}
\end{table*}


\myparagraph{Ablation study.}
We exploit the challenging multisource-multitarget scenario of Table \ref{tab:latentDA-pacs-multitarget} in order to assess the impact of the various components of our algorithm. In particular we show how the performance are affected if (i) a random domain is assigned to each sample; (ii) no loss is applied to the domain prediction branch; (iii) no entropy loss is applied to the classification of unlabeled target samples. From Table \ref{tab:latentDA-pacs-multitarget}  we can easily notice that if we drop either the domain prediction branch (random assignment) or the losses on top of it ($\lambda_E=\lambda_B=0$), the performances of the model become comparable to the ones obtain by the DIAL baseline. This shows not only the importance of discovering latent domains, but also that both the domain branch and our losses allow to extract meaningful subsets from the data. Moreover, this demonstrates the fact that our improvements are not only due to the introduction of multiple normalization layers, but also to the latent domain discovering procedure.
{For what concerns the classification branch, without the entropy component on unlabelled target samples ($\lambda_C=0$), the performance of the model significantly decreases (\ie from 82.6 to 76.4 in average). This confirms the findings of previous works \cite{carlucci2017autodial,carlucci2017just} about the impact that this loss for normalization based DA approaches. In particular, assuming that source and target samples of different domains are independently normalized, the entropy loss generates a gradient flow through unlabeled samples based in the direction of its most confident prediction. This is particularly important to learn useful features even for the target domain/s, for which no supervision is available.}

\begin{figure*}[t]
 \centering
  \subfloat[Photo as target]{\includegraphics[width=0.47\textwidth,height=0.2\textheight]{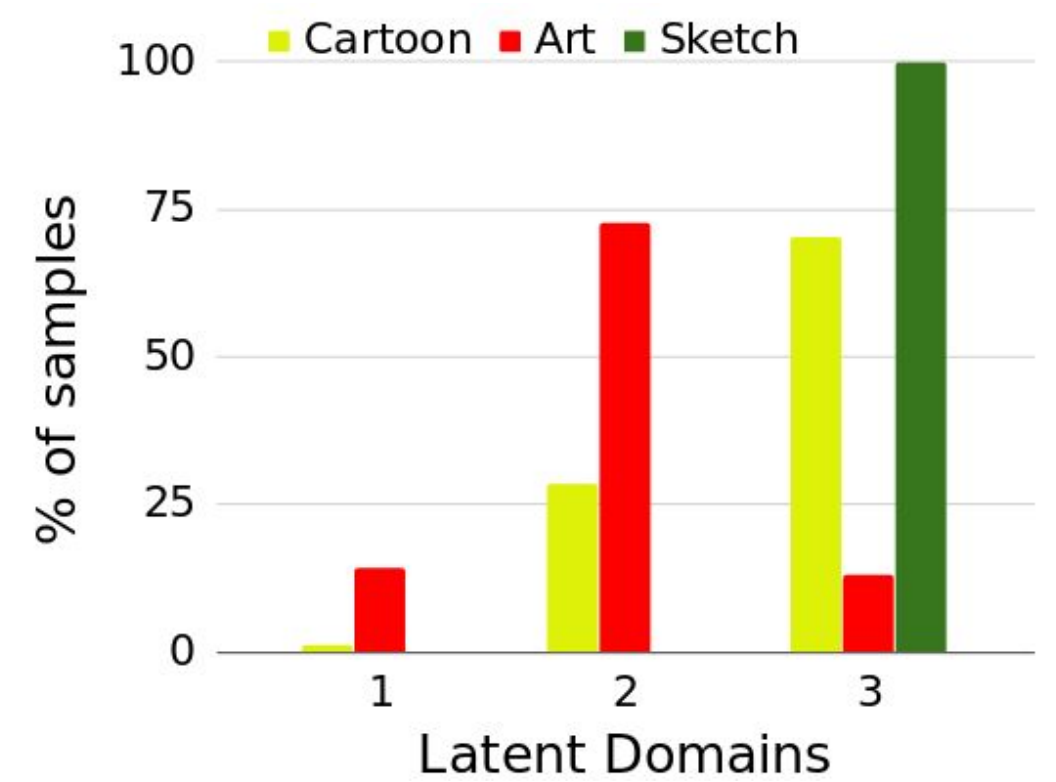}\label{fig:latentDA-assignment-photo}}\hfill
  \subfloat[Art as target]
  {\includegraphics[width=0.47\textwidth,height=0.2\textheight]{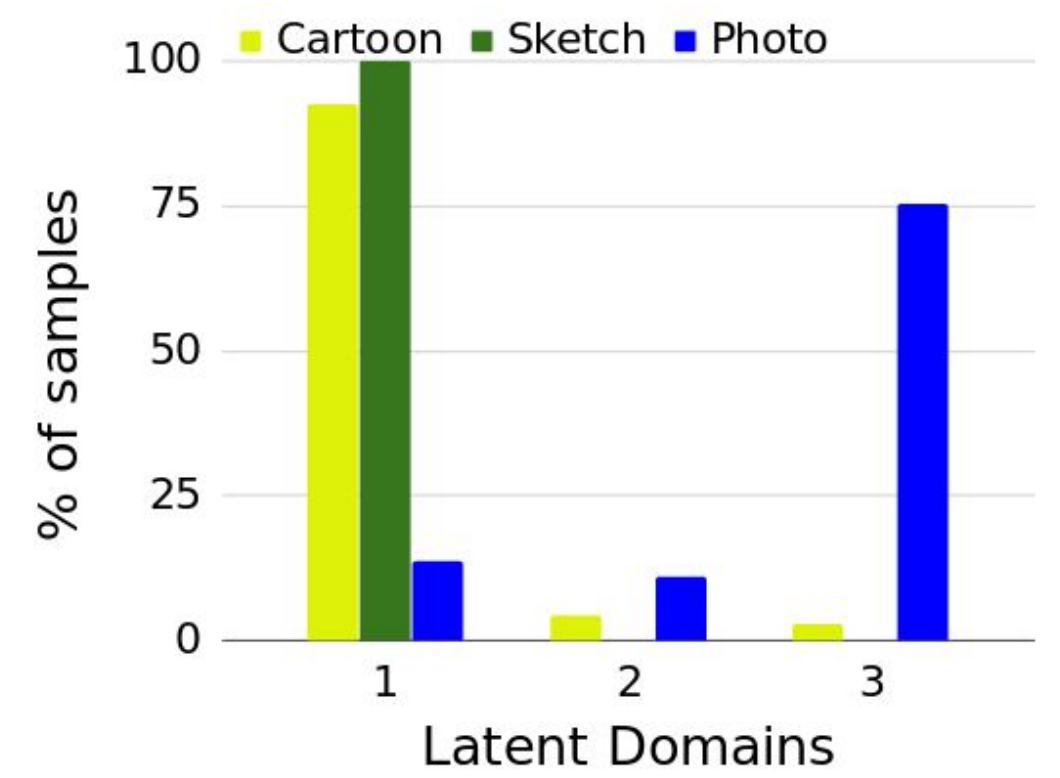}\label{fig:latentDA-assignment-art}}\\
  \subfloat[Cartoon as target]{\includegraphics[width=0.47\textwidth,height=0.2\textheight]{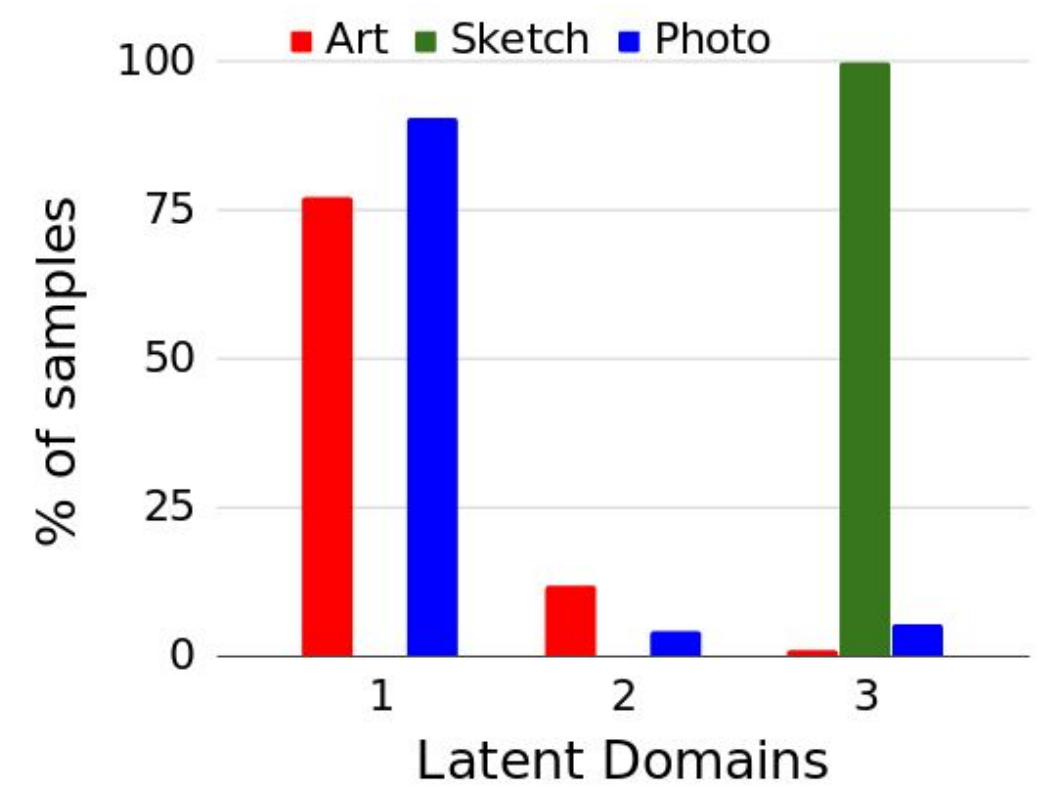}
  \label{fig:latentDA-assignment-cartoon}} \hfill
  \subfloat[Sketch as target]{\includegraphics[width=0.47\textwidth,height=0.2\textheight]{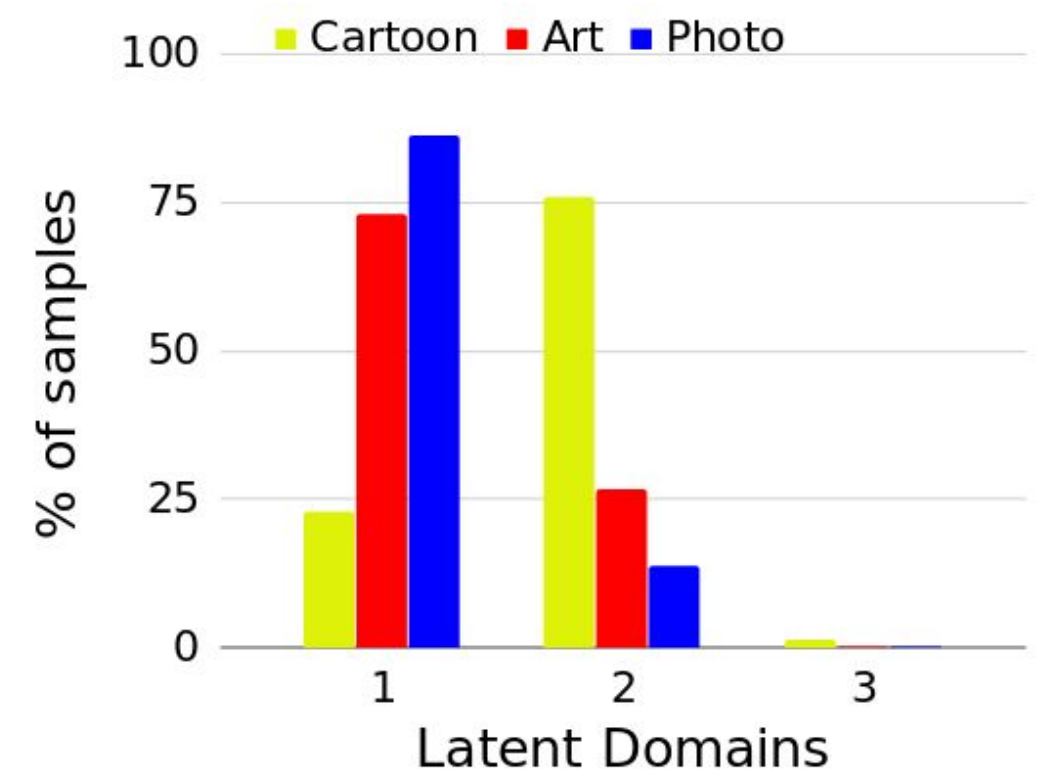}\label{fig:latentDA-assignment-sketch}}
  \caption{Distribution of the assignments produced by the domain prediction branch for each latent domain in all possible settings of the PACS dataset. Different colors denote different source domains.
}\vspace{-10pt}
  \label{fig:latentDA-max-assignment}
\end{figure*}

\begin{figure*}[t!]
  \centering
  \subfloat[Photo as target]{\includegraphics[width=0.48\textwidth,trim={0cm 0cm 36.1cm 0cm},clip]{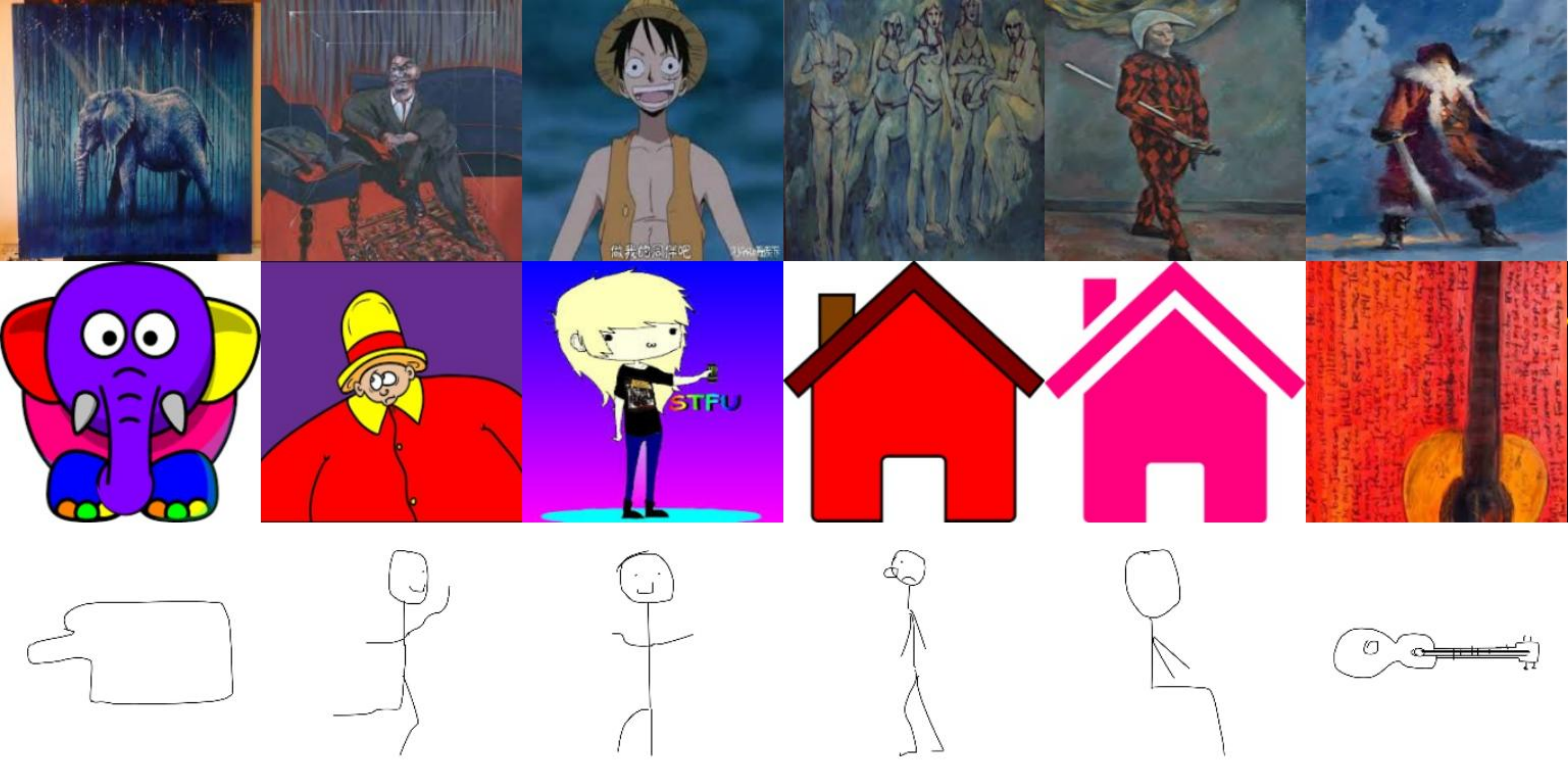}\label{fig:latentDA-topk-photo}}\hfill
  \subfloat[Art as target]
  {\includegraphics[width=0.48\textwidth,trim={0cm 0cm 36.1cm 0cm},clip]{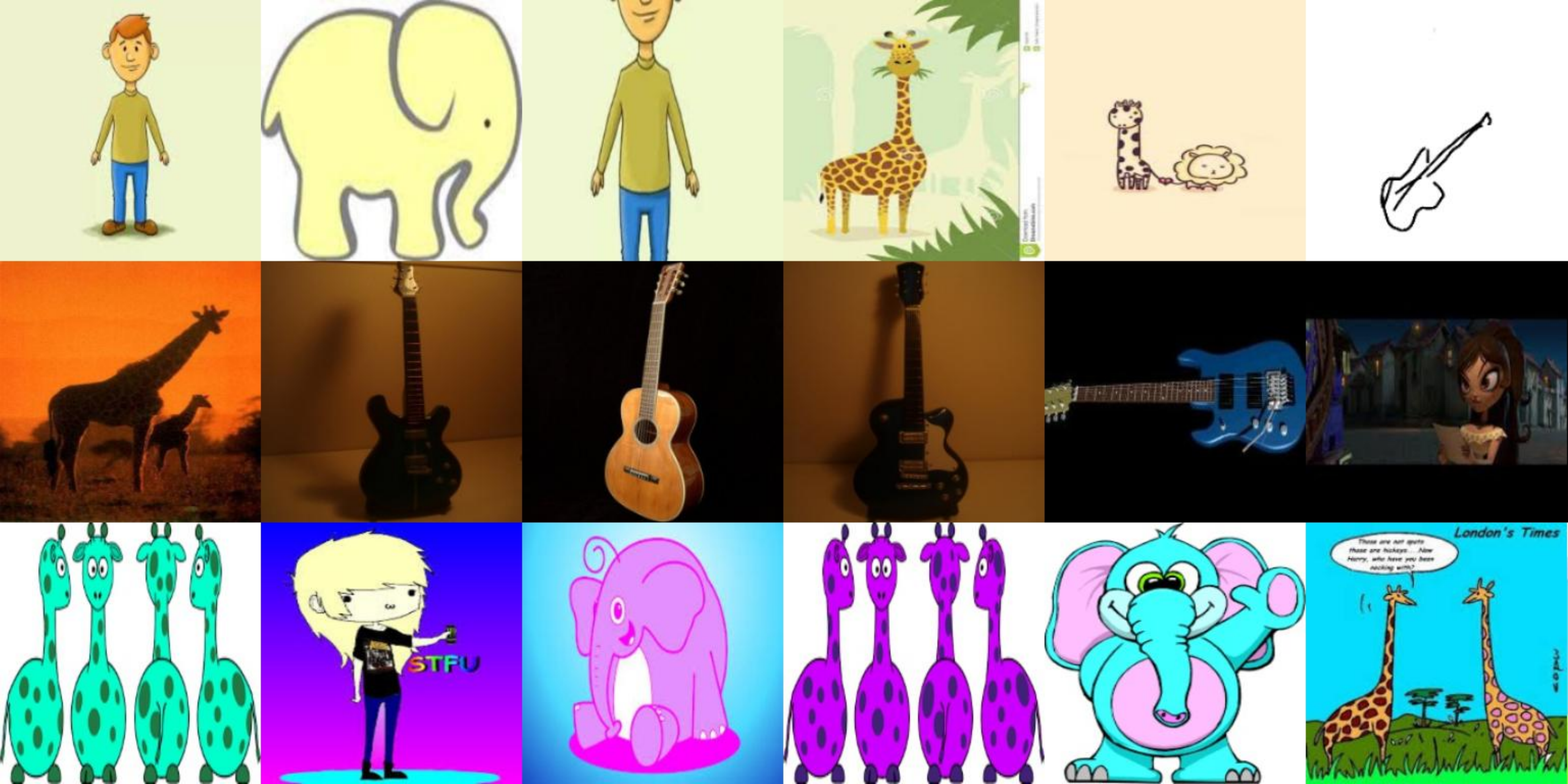}\label{fig:latentDA-topk-art}}\\  
  \subfloat[Cartoon as target]{\includegraphics[width=0.48\textwidth,trim={0cm 0cm 36.1cm 0cm},clip]{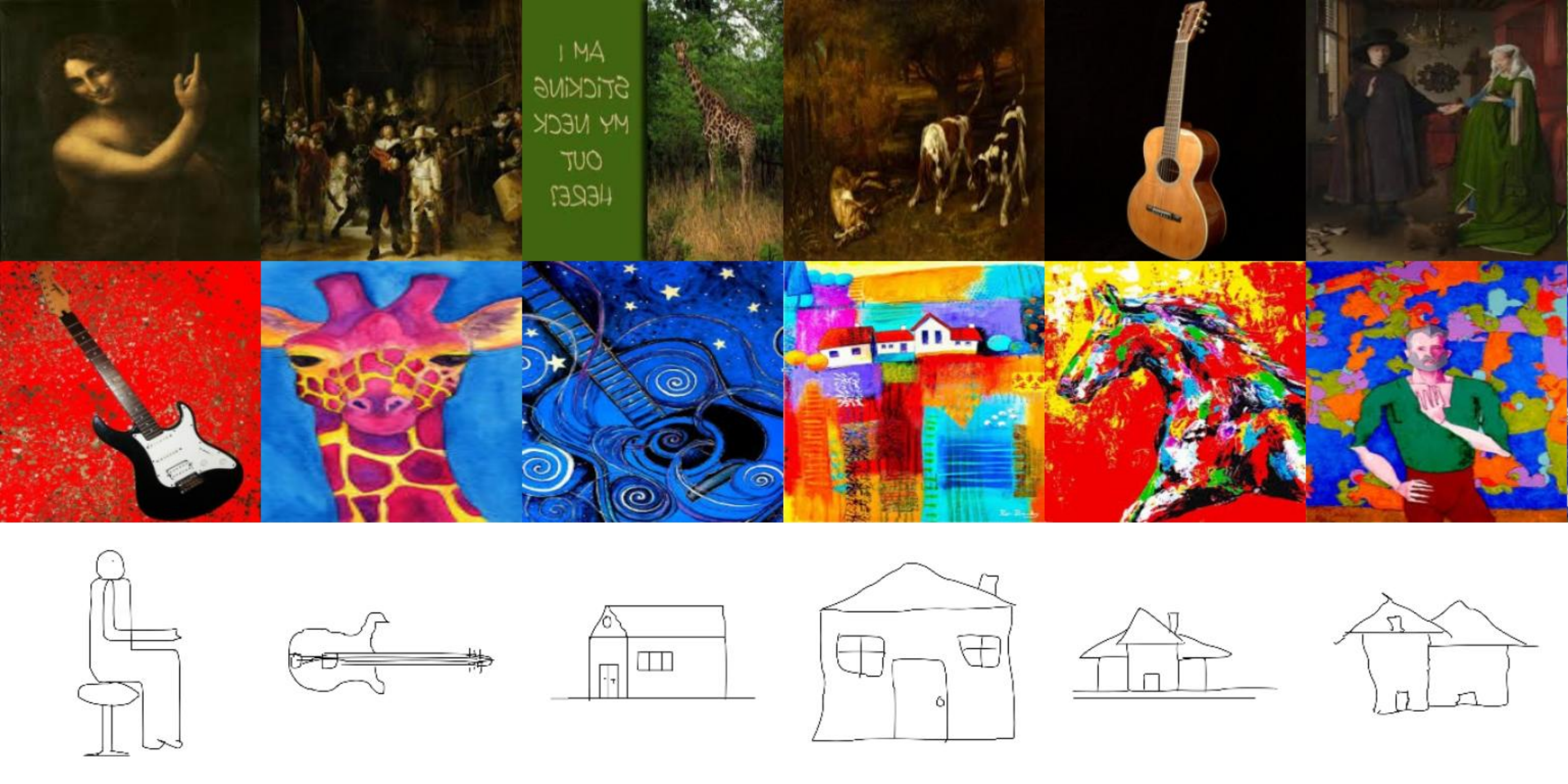}\label{fig:latentDA-top-k-cartoon}}   \hfill
  \subfloat[Sketch as target]{\includegraphics[width=0.48\textwidth,trim={0cm 0cm 36.1cm 0cm},clip]{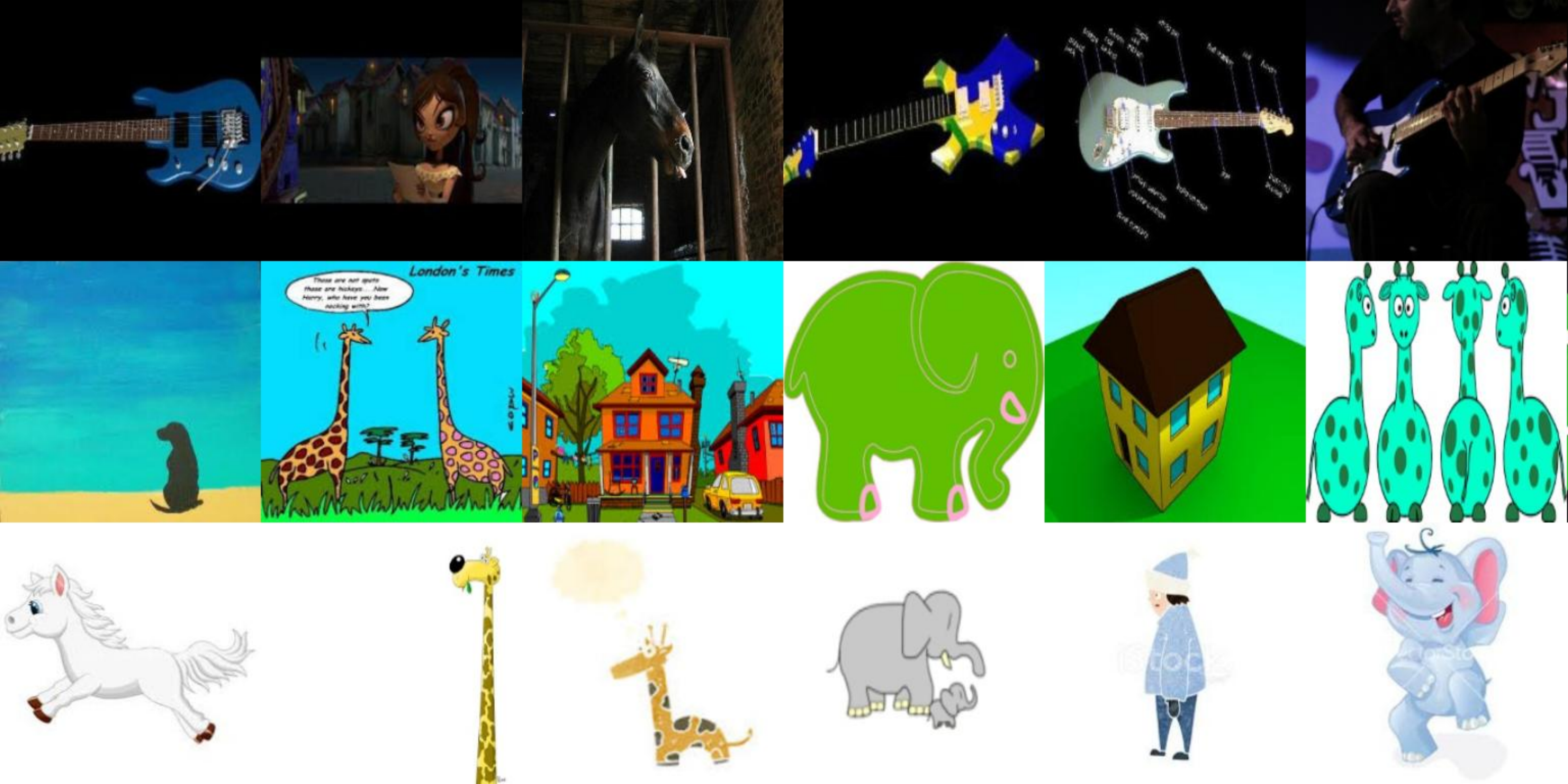}\label{fig:latentDA-topk-sketch}}
  \caption{Top-6 images associated to each latent domain for the different sources\slash{}target combinations. Each row corresponds to a different latent domain.}
  \label{fig:latentDA-top-k}
  \vspace{-15pt}
\end{figure*}

\myparagraph{In-depth analysis.}
The ability of our approach to discover latent domains is further investigated on PACS.
First, in Figure~\ref{fig:latentDA-max-assignment}, we show how our approach assigns source samples to different latent domains in the single target setting.
The four plots correspond to a single run of the experiments of Table \ref{tab:latentDA-pacs}.
Interestingly, when either Cartoon (Figure~\ref{fig:latentDA-assignment-cartoon}) or Sketch (Figure~\ref{fig:latentDA-assignment-sketch}) is the target, samples from Photo and Art tend to be associated to the same latent domain and, similarly, when either Photo (Figure~\ref{fig:latentDA-assignment-photo}) or Art (Figure~\ref{fig:latentDA-assignment-art}) is the target, samples from Cartoon and Sketch are mostly grouped together.
These results confirms the ability of our approach to automatically assign images of similar visual appearance to the same latent distribution.
In Figure~\ref{fig:latentDA-top-k}, we show the top-6 images associated to each latent domain for each sources\slash{}target setting.
In most cases, images associated to the same latent domain have similar appearance, while there is high dissimilarity between images associated to different latent domains.
Moreover, images assigned to the same latent domain tend to be associated with one of the original domains.
For instance, the first row of Figure~\ref{fig:latentDA-topk-photo} contains only images from Art, while the third contains only images from Sketch.
Note that no explicit domain supervision is ever given to our method in this setting.

\begin{figure*}[ht!]
 \centering
 \subfloat[Cartoon and Sketch as sources]
  {\includegraphics[width=0.48\textwidth]{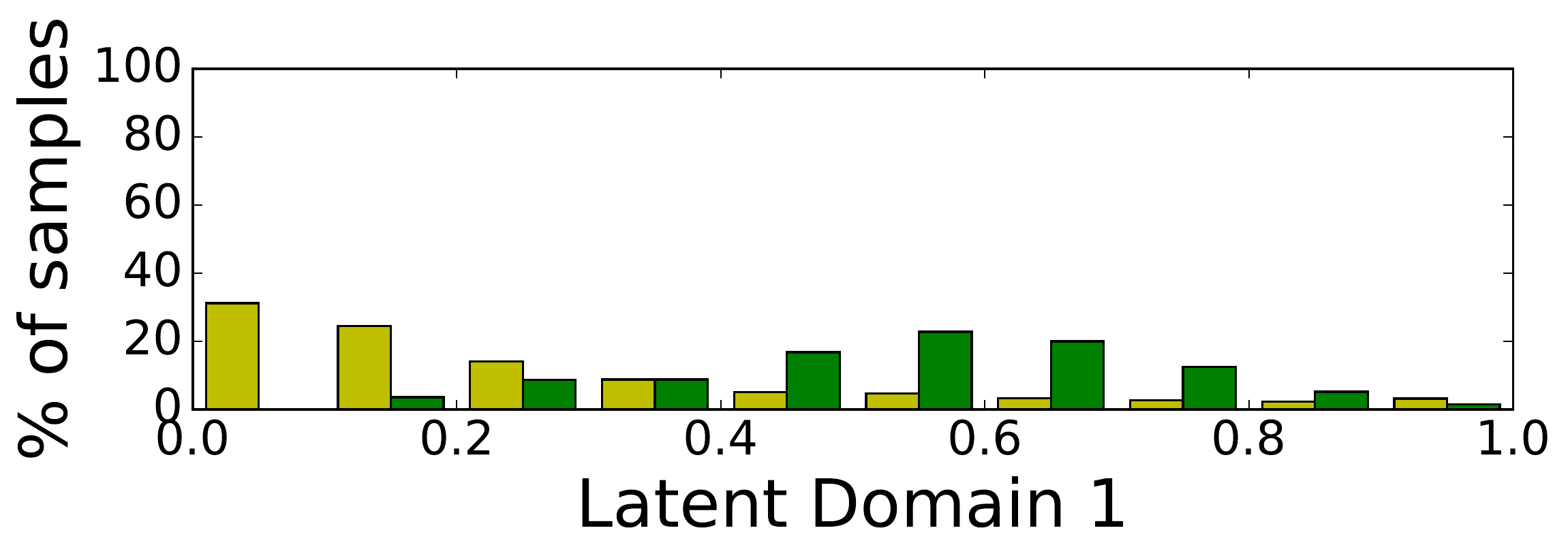}\label{fig:latentDA-assignment-pa-s}} \hfill
  \subfloat[Art and Photo as targets]{\includegraphics[width=0.48\textwidth]{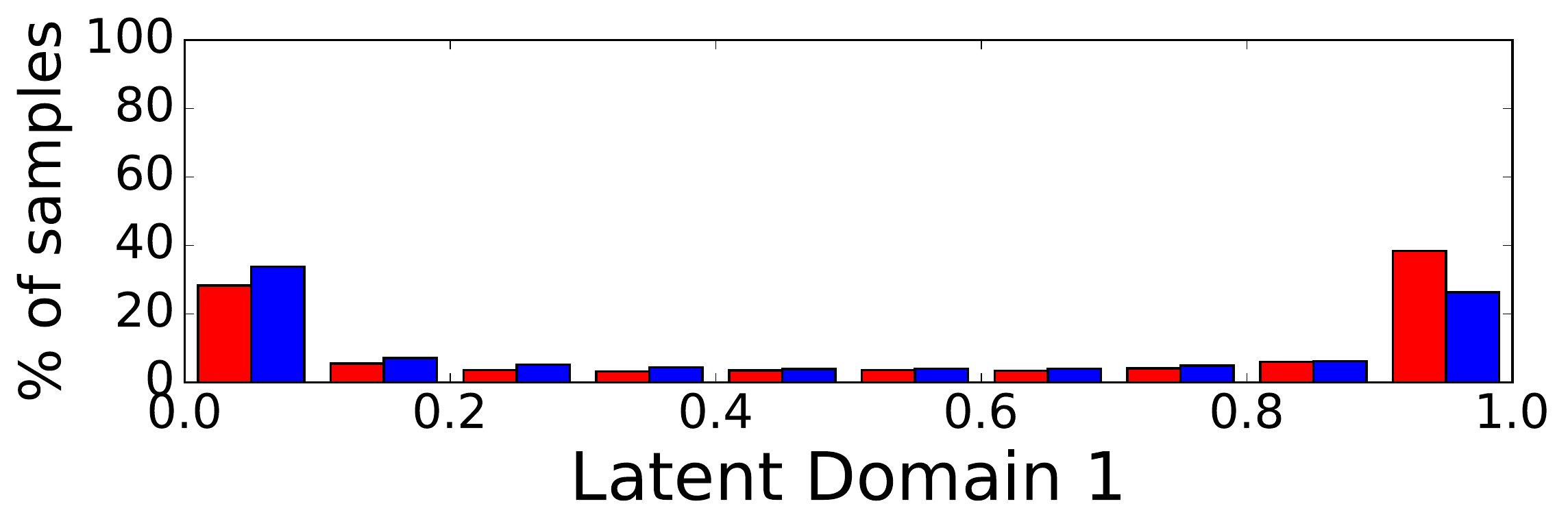}\label{fig:latentDA-assignment-pa-t}}\\
  \subfloat[Art and Sketch as sources]
  {\includegraphics[width=0.48\textwidth]{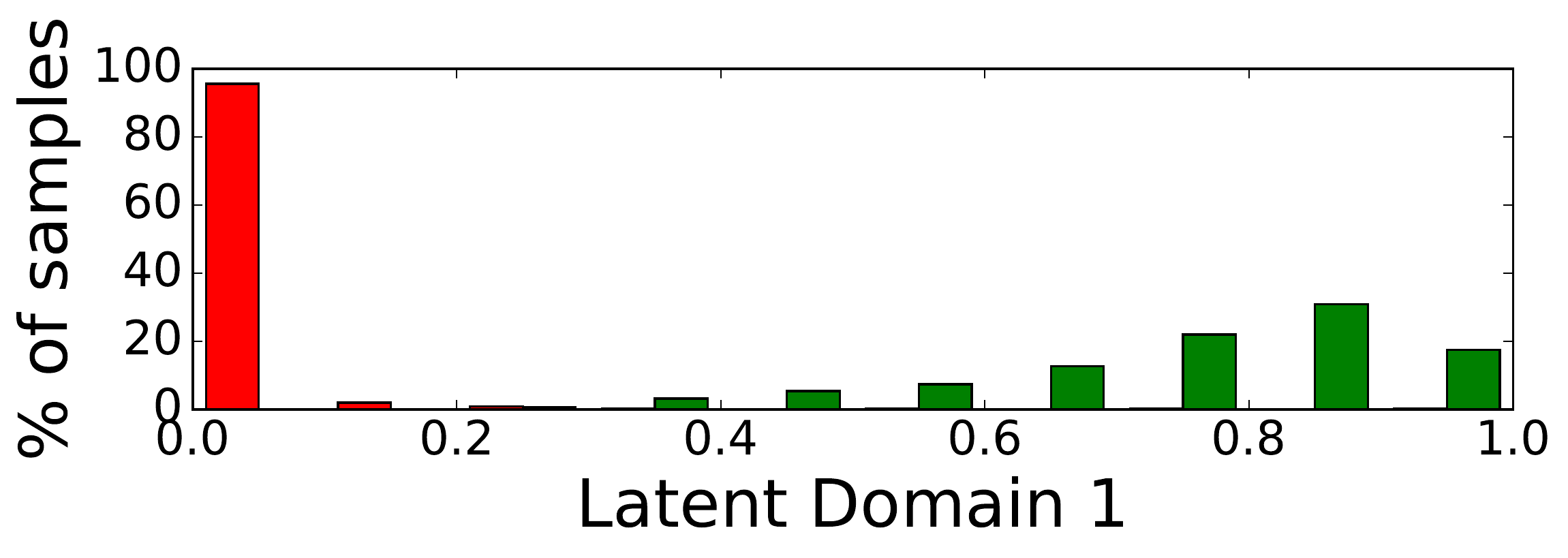}\label{fig:latentDA-assignment-pc-s}}\hfill
    \subfloat[Cartoon and Photo as targets]{\includegraphics[width=0.48\textwidth]{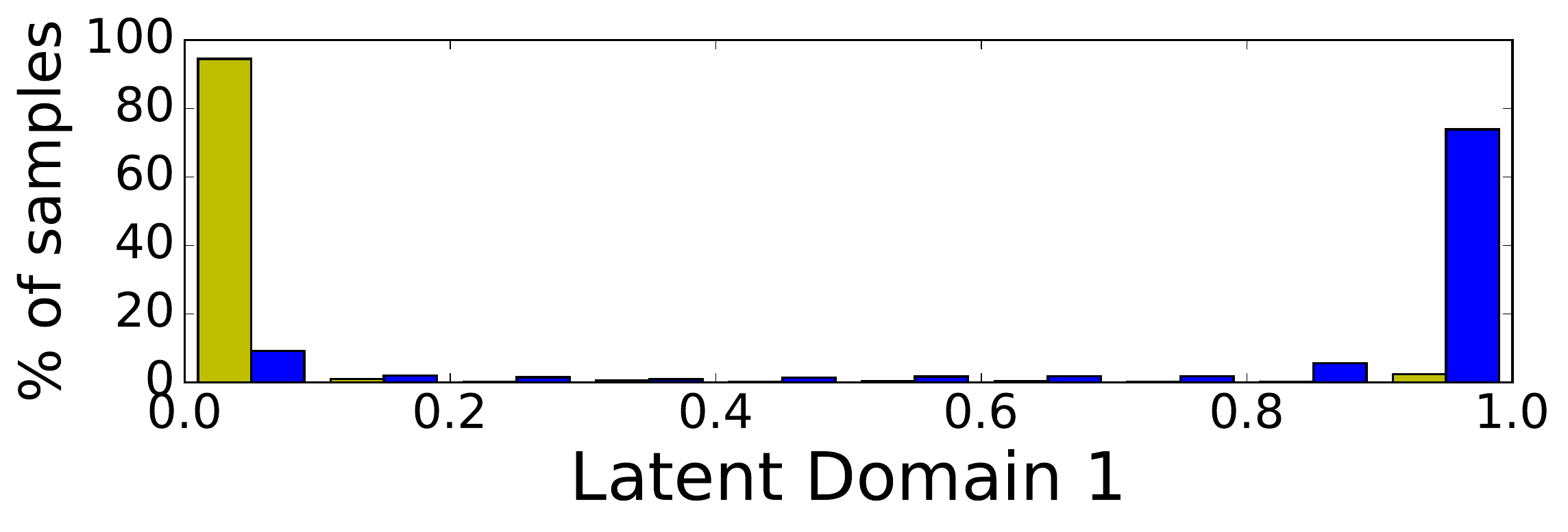}\label{fig:latentDA-assignment-pc-t}}\\
    \subfloat[Art and Cartoon as sources]{\includegraphics[width=0.48\textwidth,]{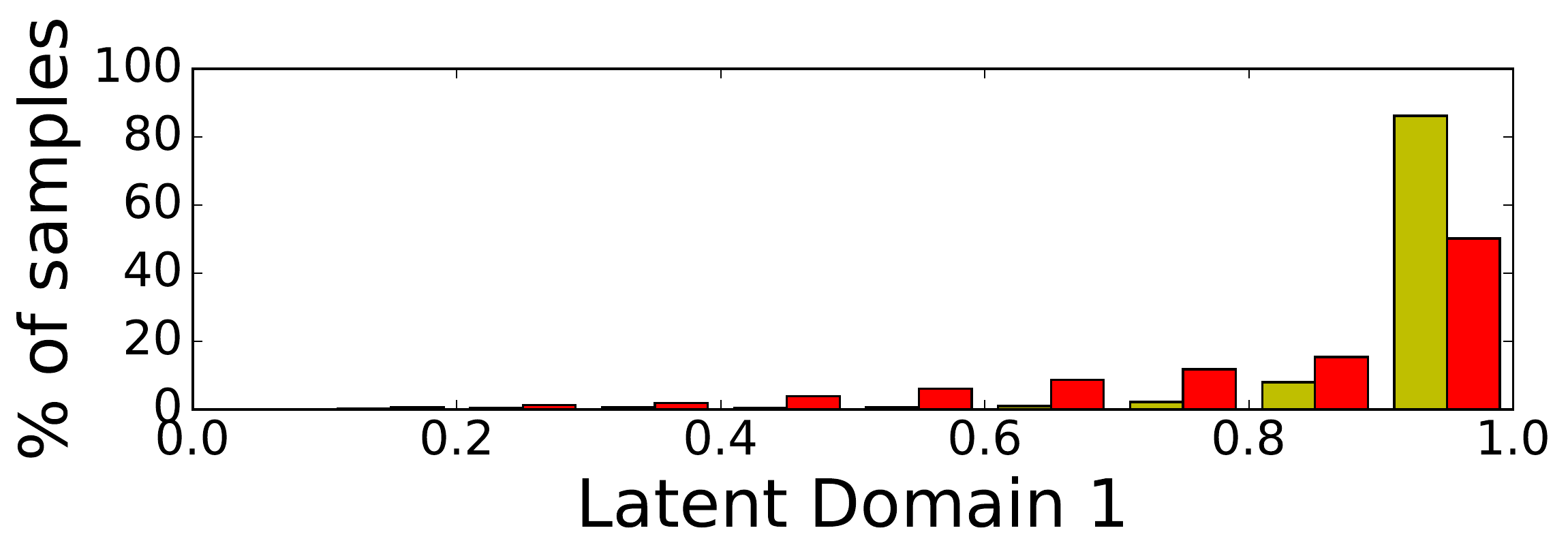}\label{fig:latentDA-assignment-ps-s}}\hfill
  \subfloat[Photo and Sketch as targets]
  {\includegraphics[width=0.48\textwidth,]{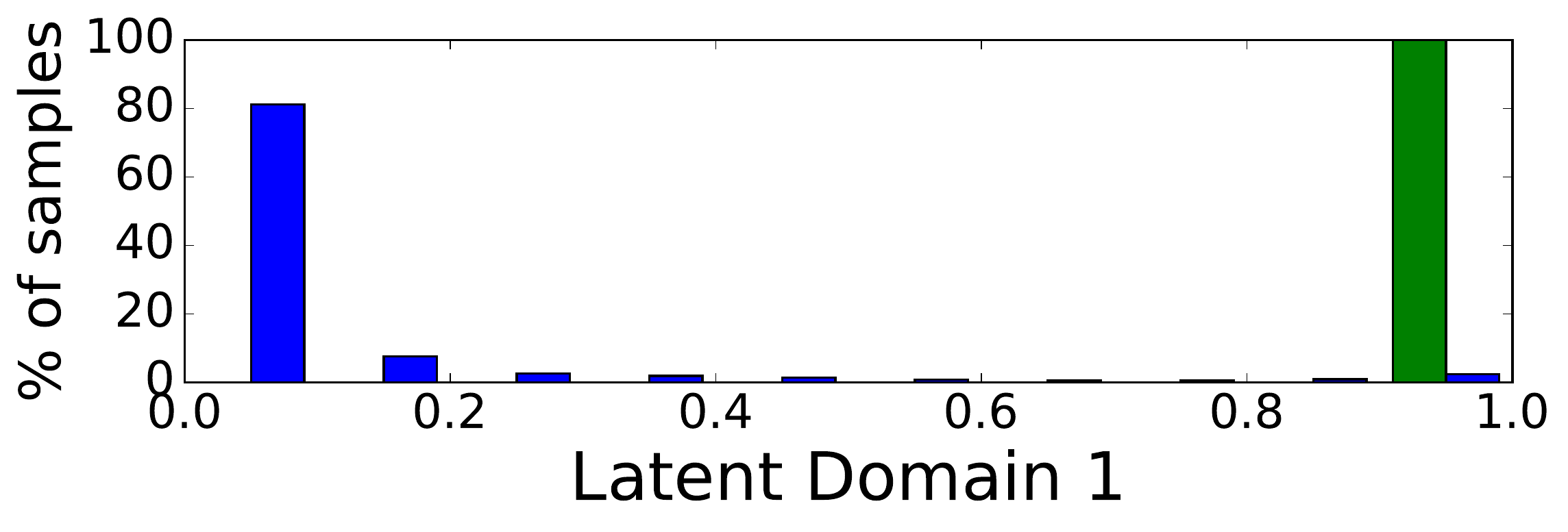}\label{fig:latentDA-assignment-ps-t}}\\
   \subfloat[Photo and Sketch as sources]
  {\includegraphics[width=0.48\textwidth,]{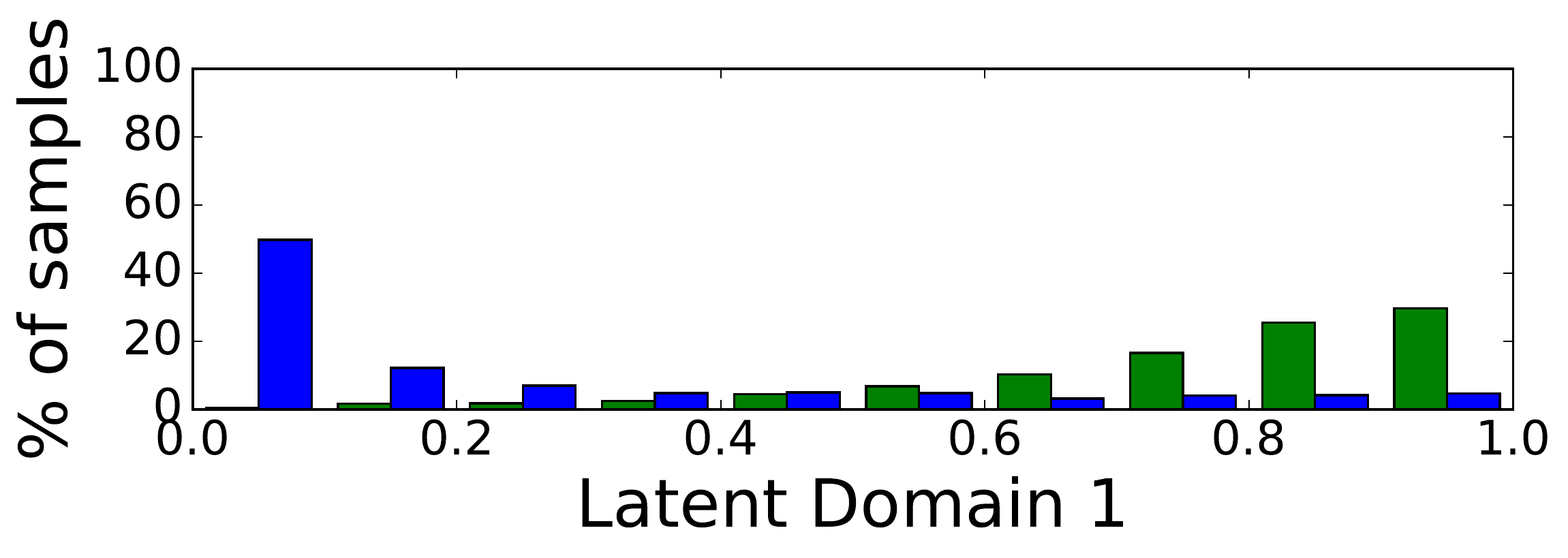}\label{fig:latentDA-assignment-ac-s}}\hfill
    \subfloat[Art and Cartoon as targets]{\includegraphics[width=0.48\textwidth,]{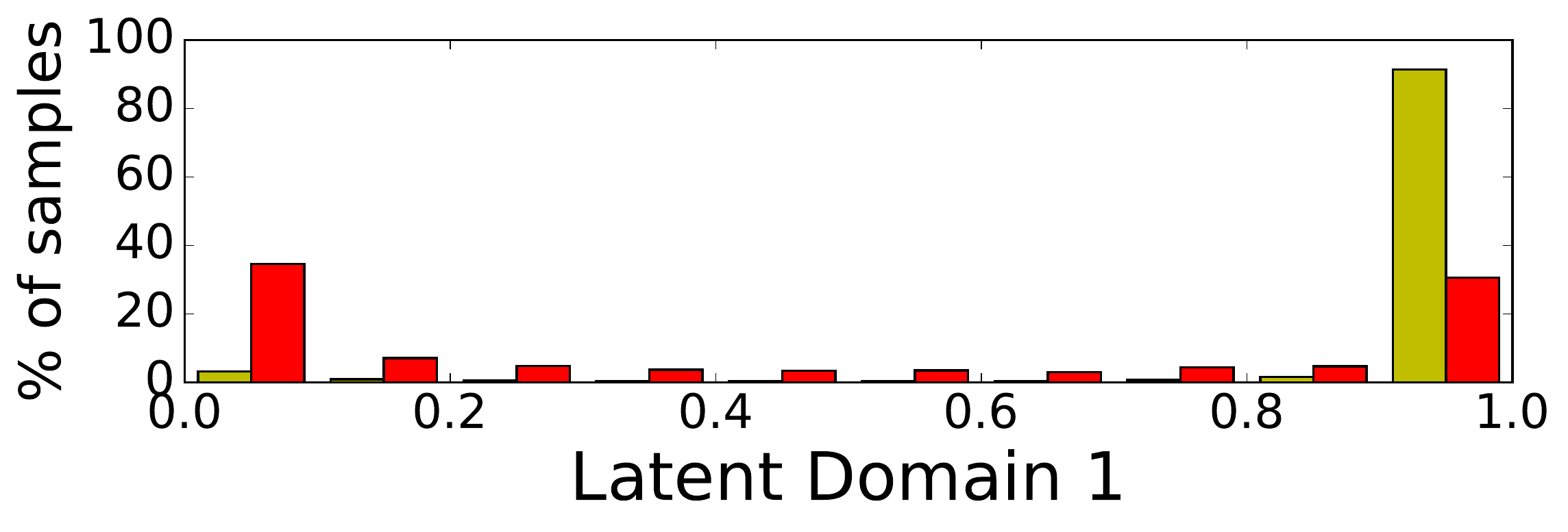}\label{fig:latentDA-assignment-ac-t}}\\
 \subfloat[Cartoon and Photo as sources]
  {\includegraphics[width=0.48\textwidth,]{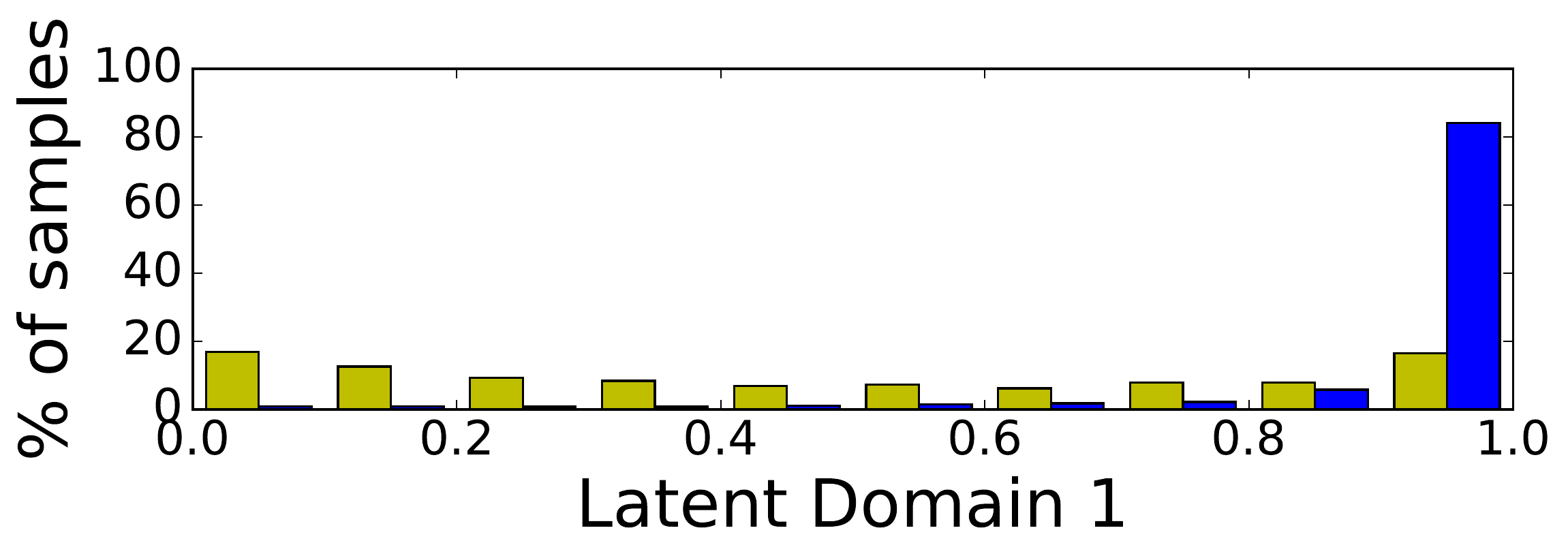}\label{fig:latentDA-assignment-as-s}}\hfill
    \subfloat[Art and Sketch as targets]{\includegraphics[width=0.48\textwidth,]{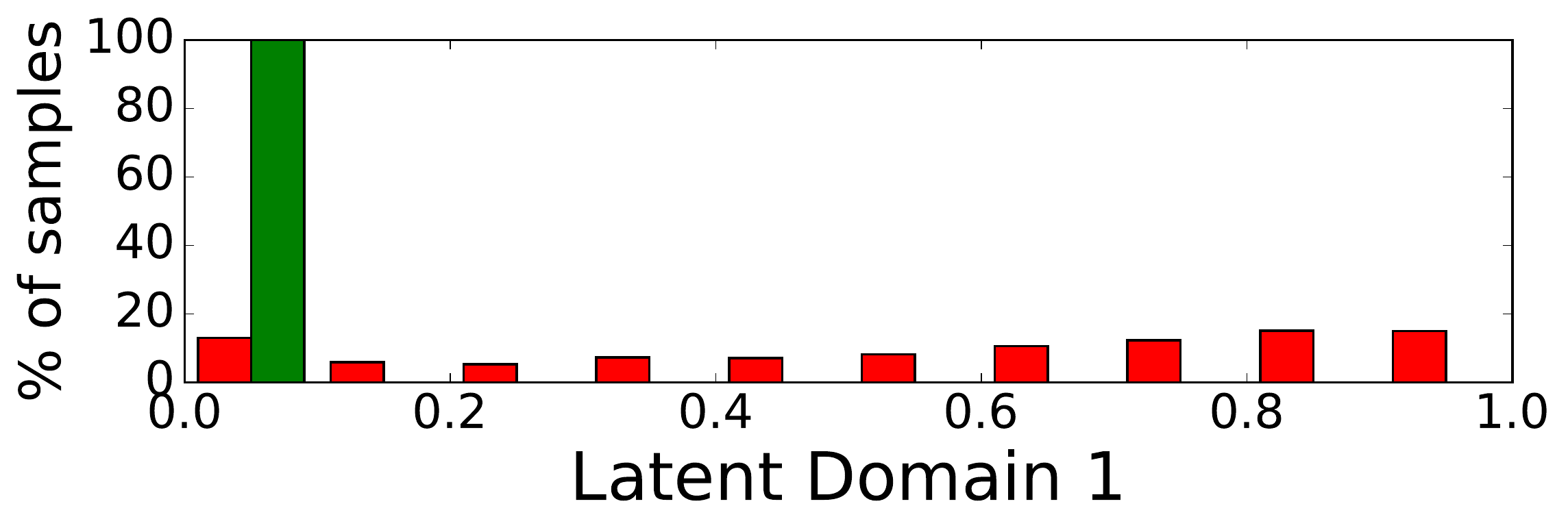}\label{fig:latentDA-assignment-as-t}}\\
  \subfloat[Art and Photo as sources]
  {\includegraphics[width=0.48\textwidth,]{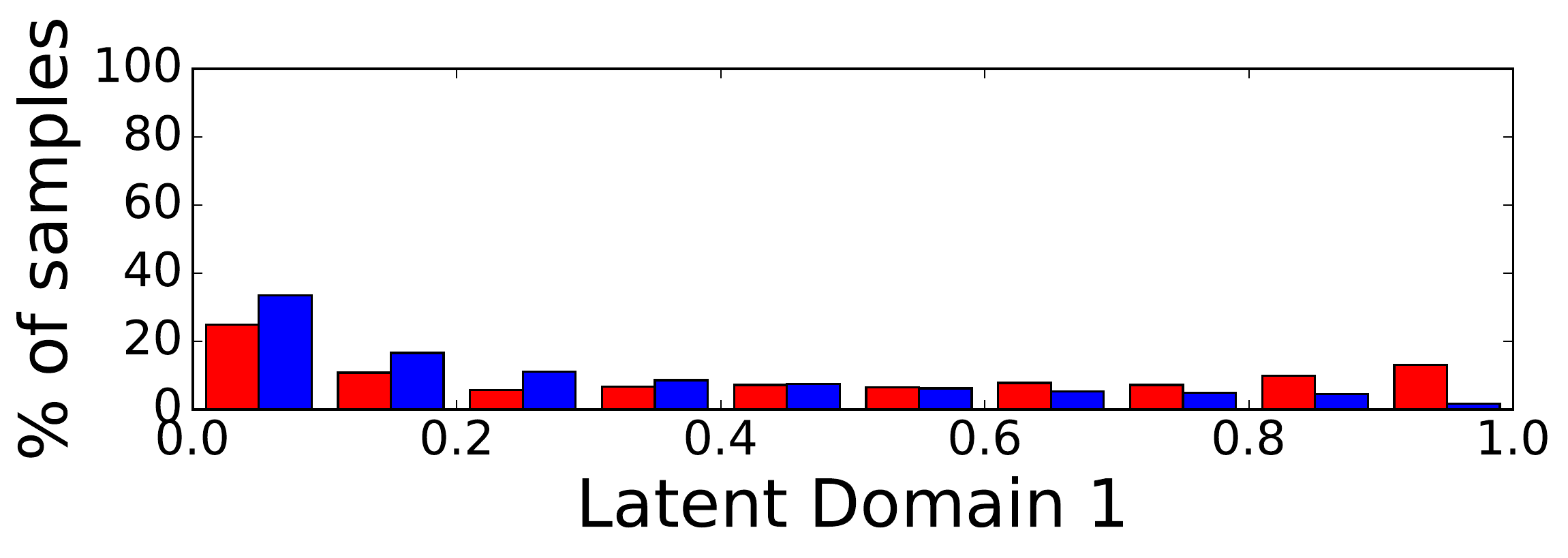}\label{fig:latentDA-assignment-cs-s}}\hfill
    \subfloat[Cartoon and Sketch as targets]{\includegraphics[width=0.48\textwidth,]{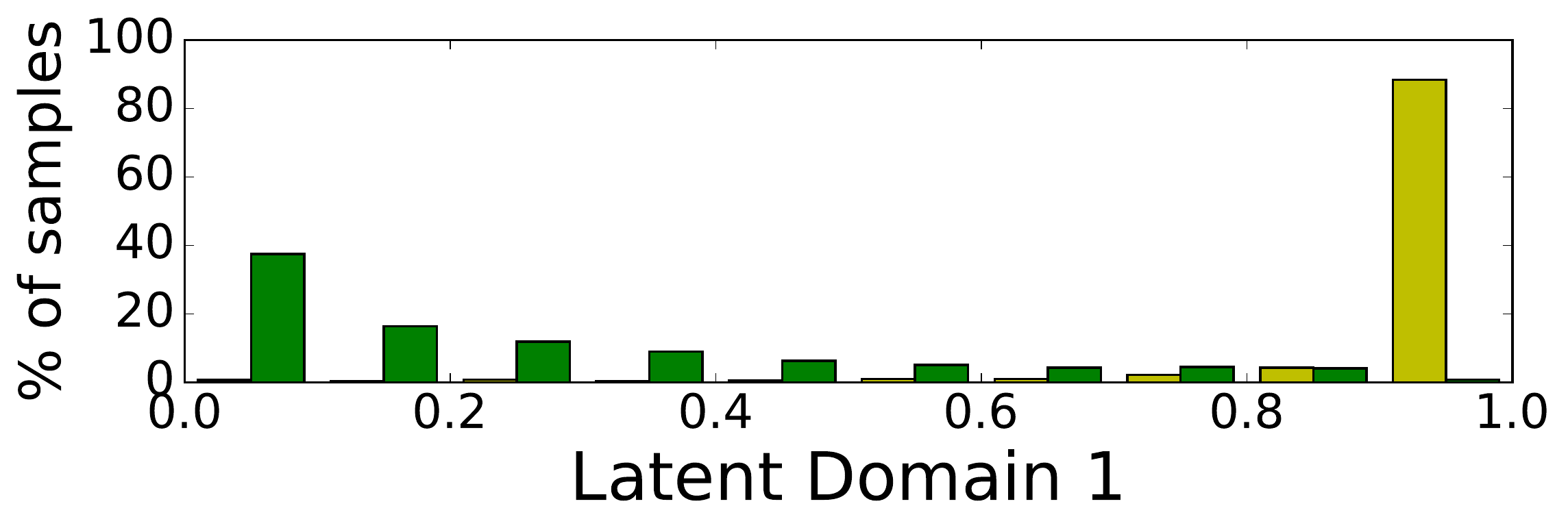}\label{fig:latentDA-assignment-cs-t}}
  \\
  
  \caption{Distribution of the assignments produced by the domain prediction branch in all possible multi-target settings of the PACS dataset. Different colors denote different source domains (red: Art, yellow: Cartoon, blue: Photo, green: Sketch).
  }
  \label{fig:latentDA-soft-assignment}
\end{figure*}

In Figure~\ref{fig:latentDA-soft-assignment}, we show the histograms of the domain assignment probabilities predicted by our model with $\lambda_B=0$ in the various multi-source, multi-target settings of Table~\ref{tab:latentDA-pacs}.
As the figures shows, in most cases the various pairs of target domains tend to be very well separated: this justifies the large gain of performances produced by our model in this scenario.
The only cases where the separation is less marked is when Art and Photo, which have very similar visual appearance, are considered as targets.
On the other hand, source domains are not always as clearly separated as the targets.
In particular the pairs Photo-Cartoon, Art-Photo and Art-Cartoon, tend to receive similar assignments when they are considered as source.
A possible explanation is that the supervised source loss could have a stronger influence on the domain assignment than the unsupervised target one.
In any case, note that these results do not detract from the validity of our approach.
In fact, our main objective is to obtain a good classification model for the target set, independently from the actual domain assignments we learn.

\begin{figure*}[t!]
 \centering
 \subfloat[Cartoon and Sketch as sources]
  {\includegraphics[width=0.48\textwidth]{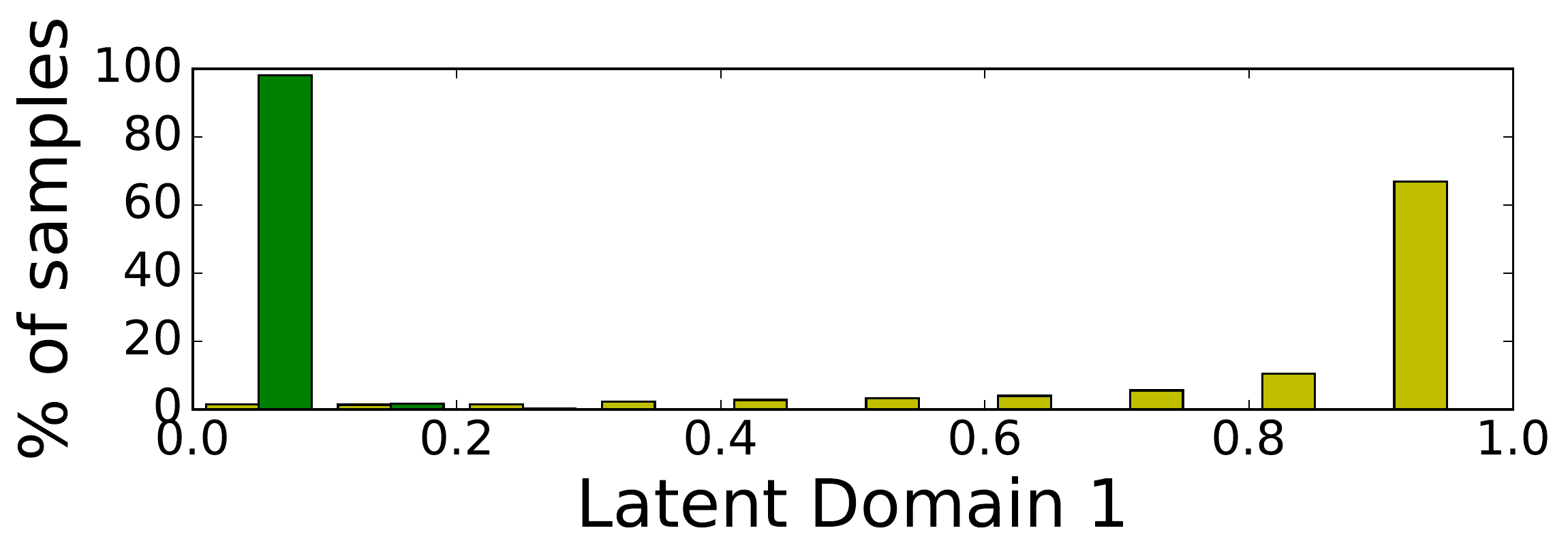}\label{fig:latentDA-assignment-cluster-pa-s}} \hfill
  \subfloat[Art and Photo as targets]{\includegraphics[width=0.48\textwidth,]{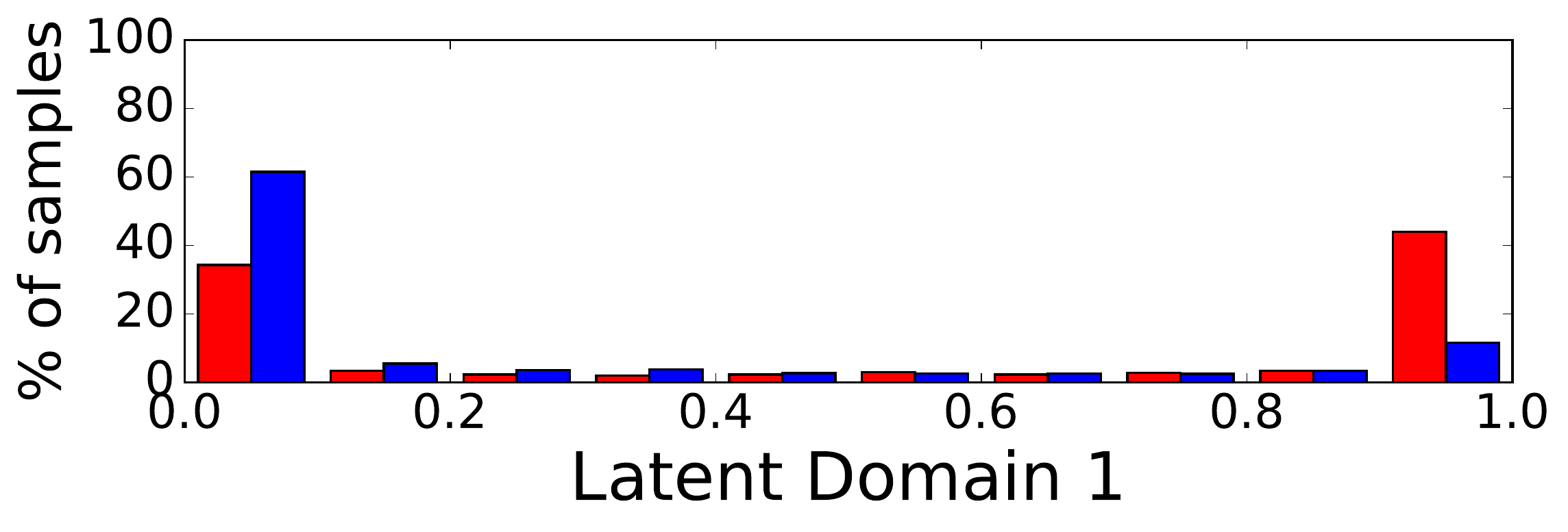}\label{fig:latentDA-assignment-cluster-pa-t}}\\
  \subfloat[Art and Sketch as sources]
  {\includegraphics[width=0.48\textwidth]{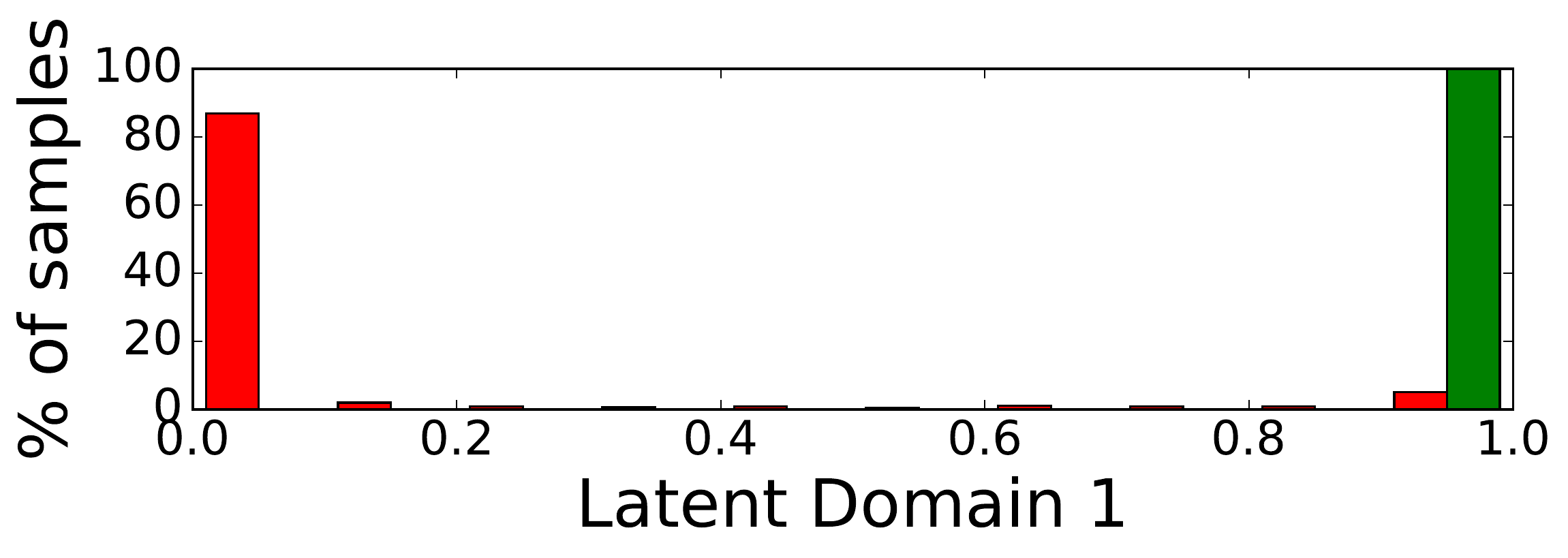}\label{fig:latentDA-assignment-cluster-pc-s}}\hfill
    \subfloat[Cartoon and Photo as targets]{\includegraphics[width=0.48\textwidth,]{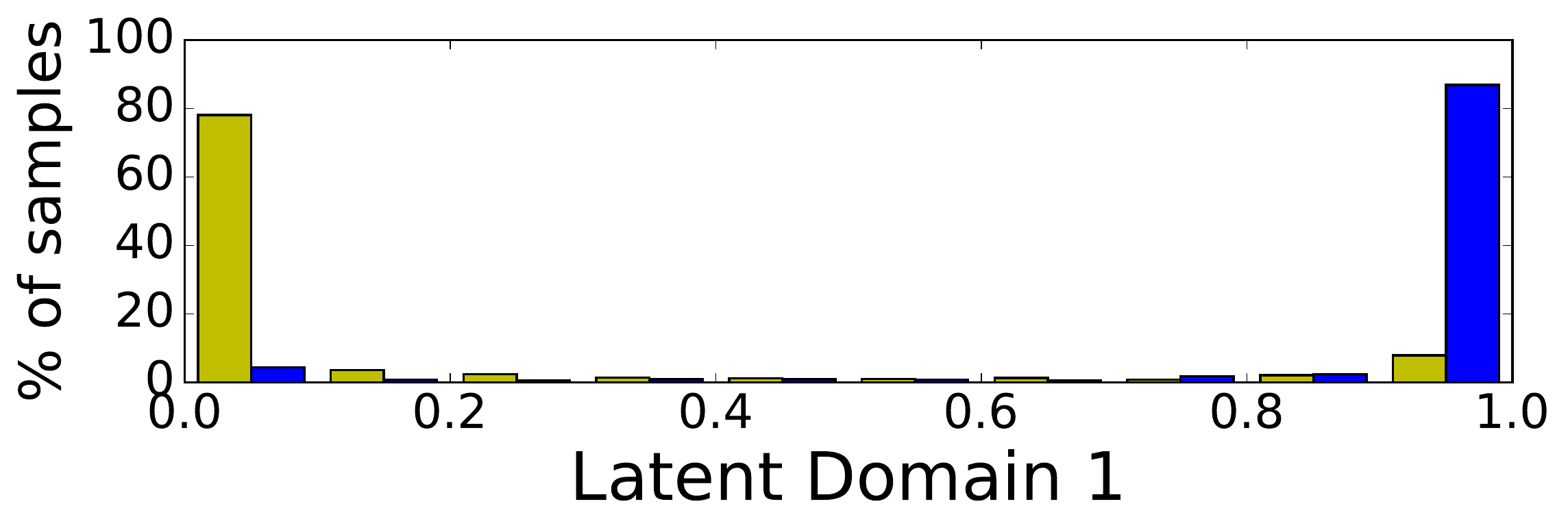}\label{fig:latentDA-assignment-cluster-pc-t}}
  \\
    \subfloat[Art and Cartoon as sources]{\includegraphics[width=0.48\textwidth,]{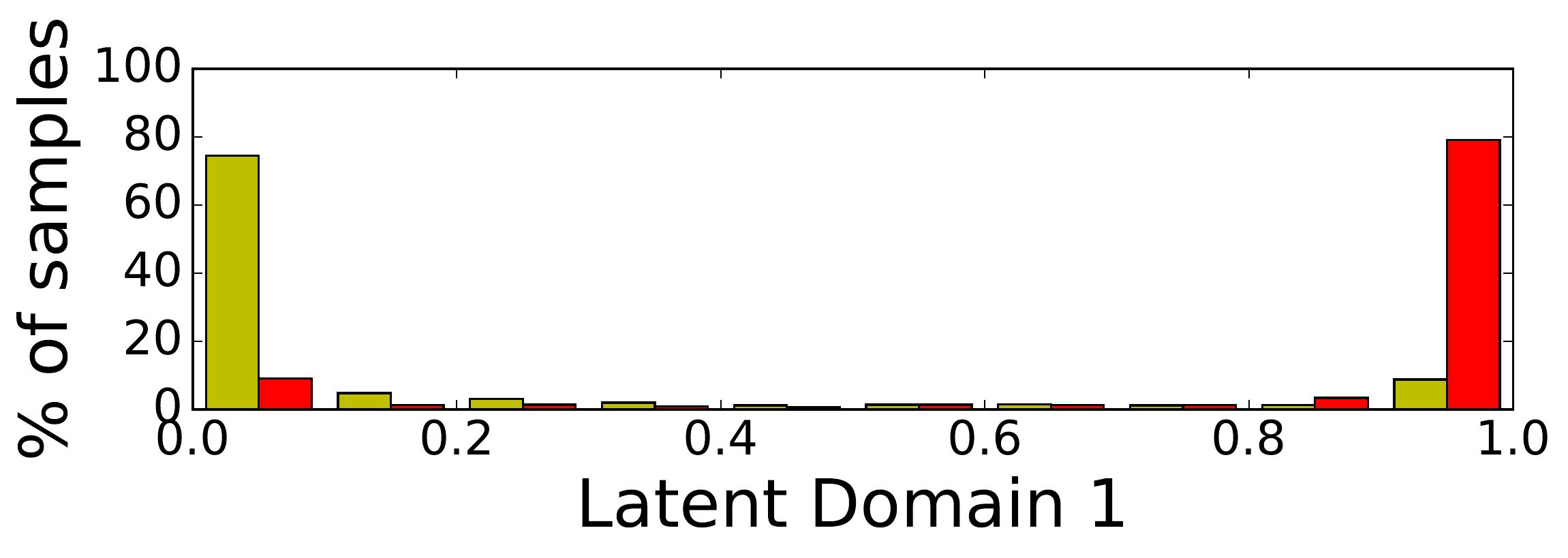}\label{fig:latentDA-assignment-cluster-ps-s}}\hfill
  \subfloat[Photo and Sketch as targets]
  {\includegraphics[width=0.48\textwidth,]{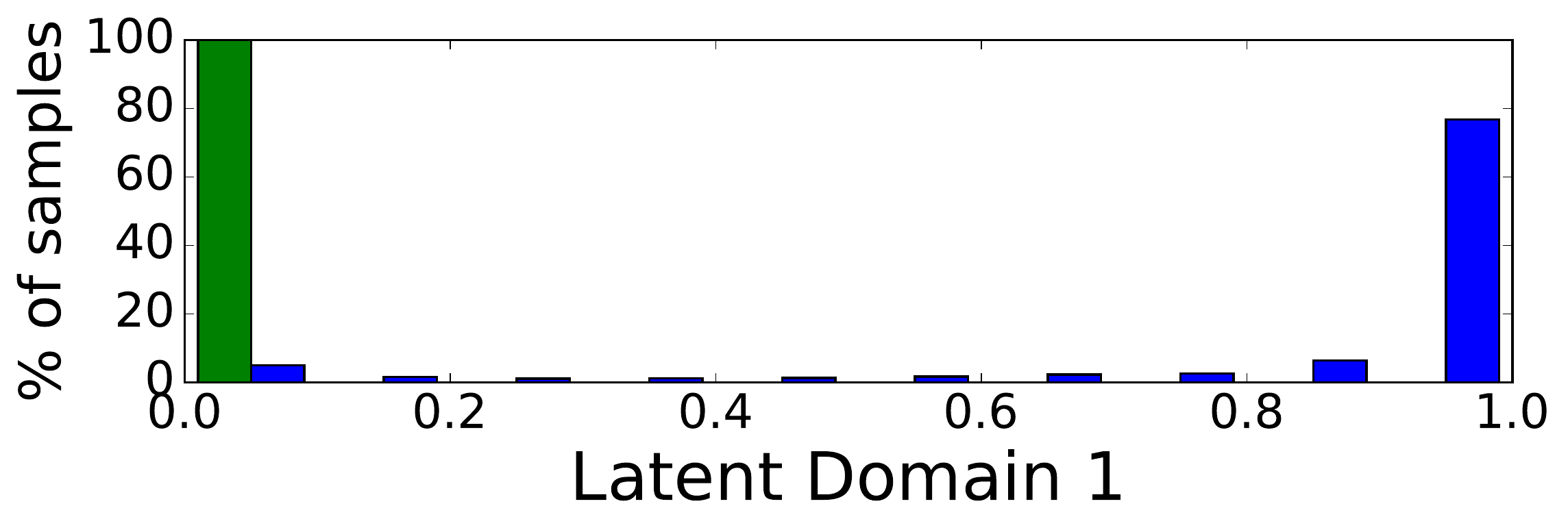}\label{fig:latentDA-assignment-cluster-ps-t}}\\
   \subfloat[Photo and Sketch as sources]
  {\includegraphics[width=0.48\textwidth,]{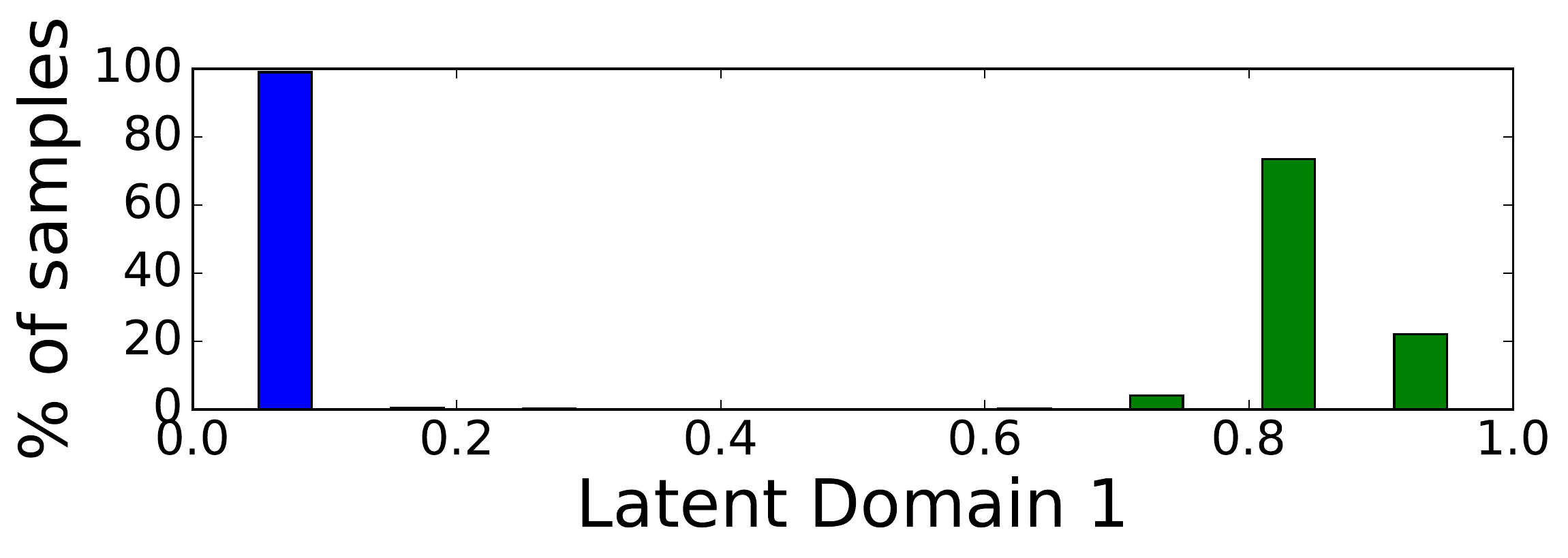}\label{fig:latentDA-assignment-cluster-ac-s}}\hfill
    \subfloat[Art and Cartoon as targets]{\includegraphics[width=0.48\textwidth,]{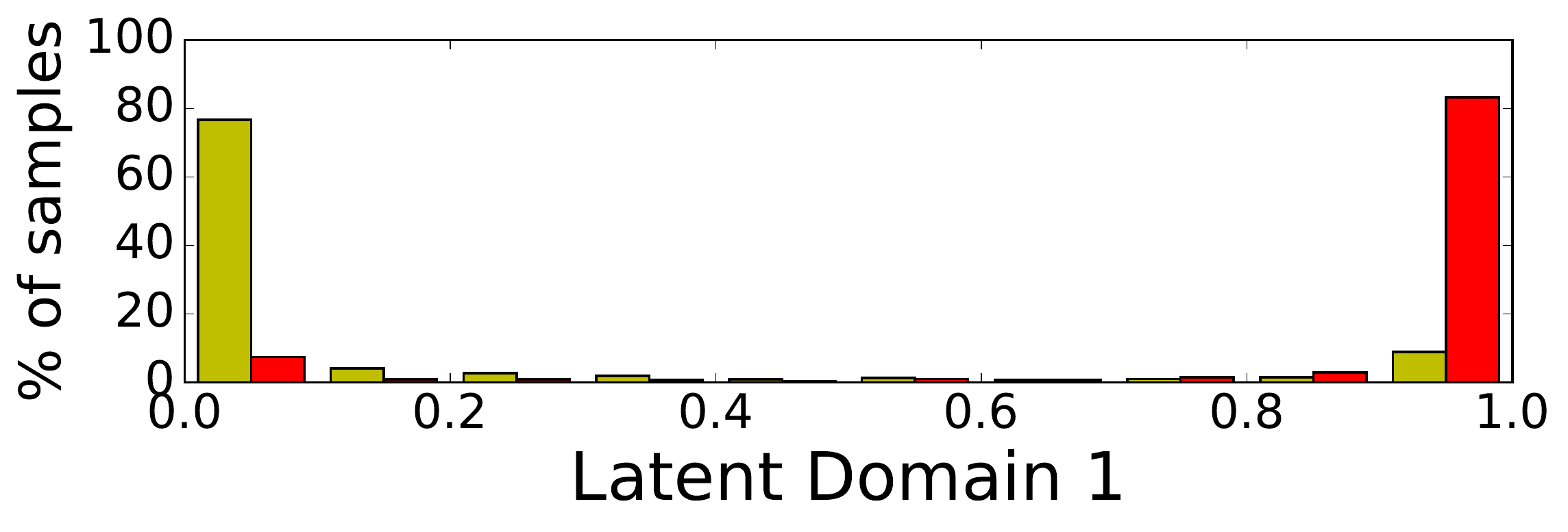}\label{fig:latentDA-assignment-cluster-ac-t}}
 \\
 \subfloat[Cartoon and Photo as sources]
  {\includegraphics[width=0.48\textwidth,]{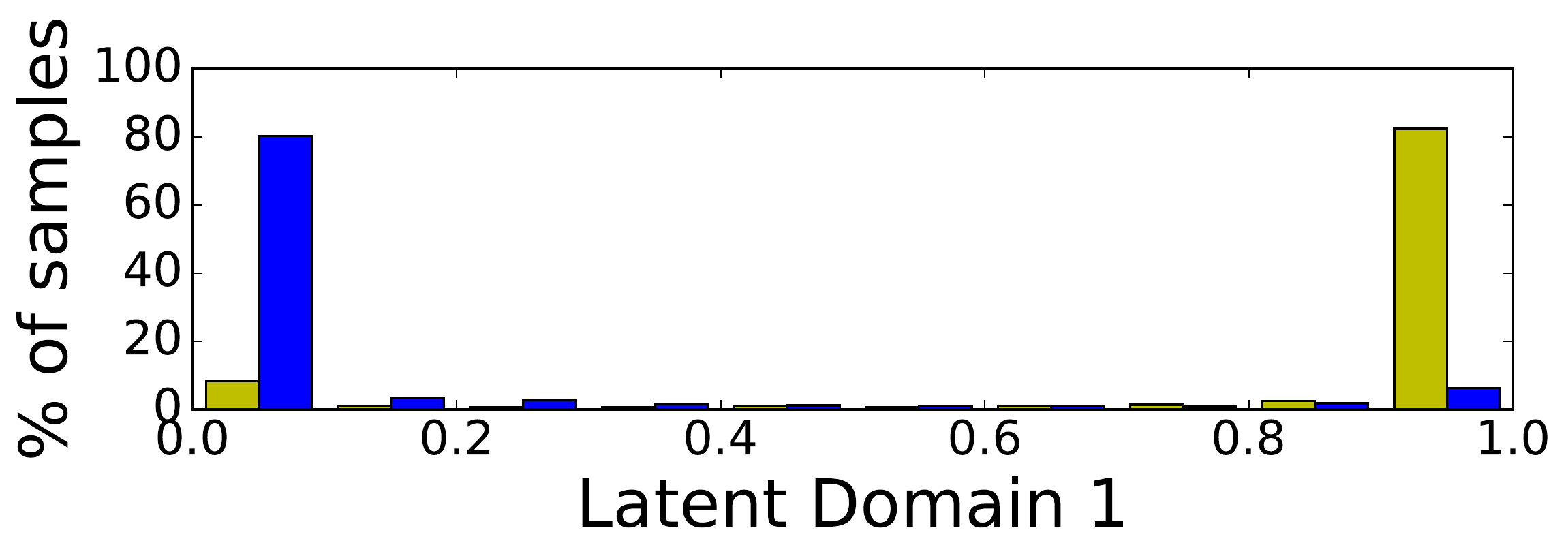}\label{fig:latentDA-assignment-cluster-as-s}}\hfill
    \subfloat[Art and Sketch as targets]{\includegraphics[width=0.48\textwidth,]{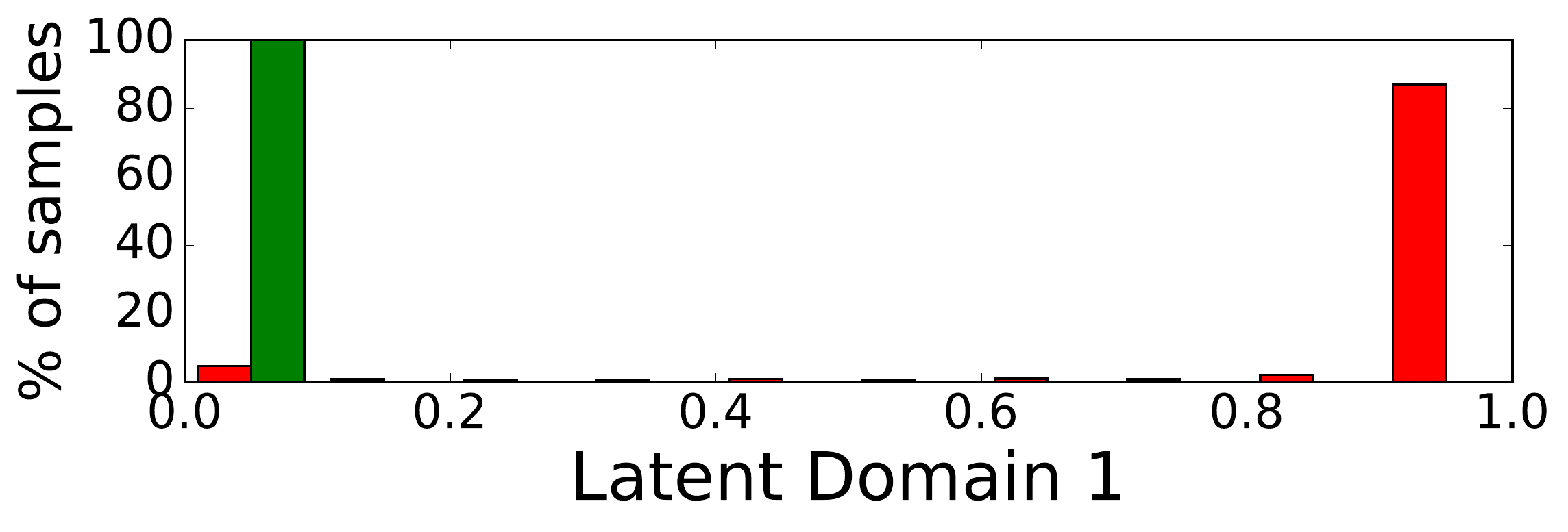}\label{fig:latentDA-assignment-cluster-as-t}}\\
  \subfloat[Art and Photo as sources]
  {\includegraphics[width=0.48\textwidth,]{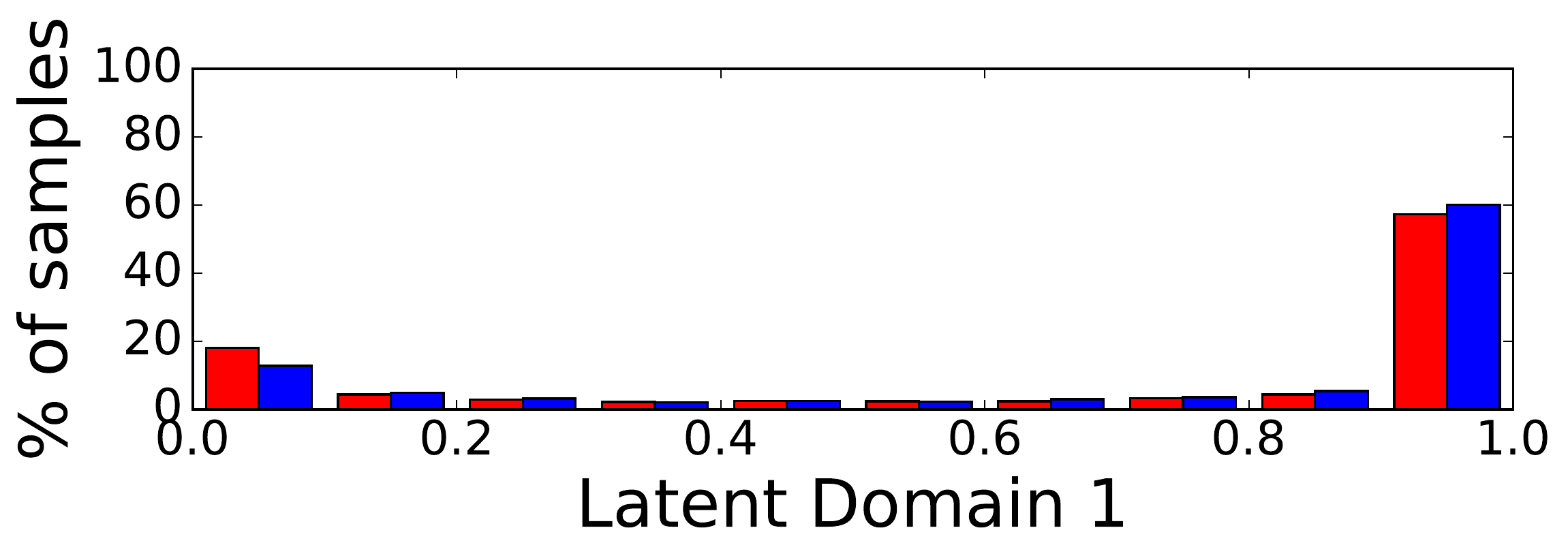}\label{fig:latentDA-assignment-cluster-cs-s}}\hfill
    \subfloat[Cartoon and Sketch as targets]{\includegraphics[width=0.48\textwidth,]{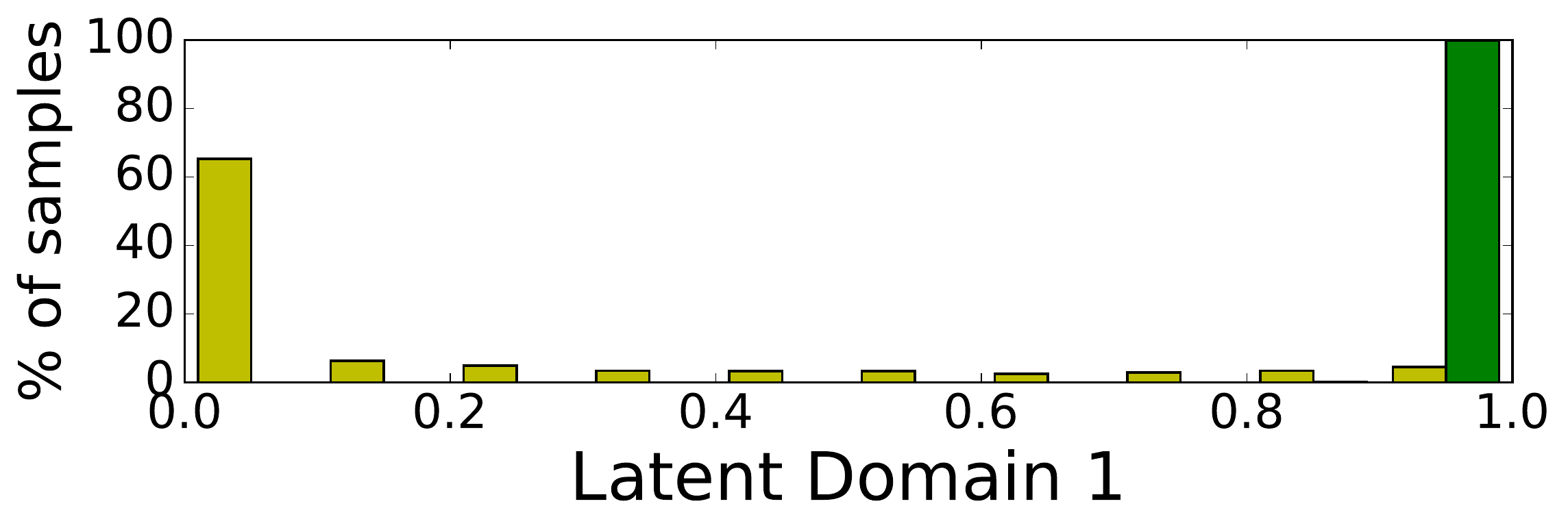}\label{fig:latentDA-assignment-cluster-cs-t}}
  \\
  \caption{{Distribution of the assignments produced by the domain prediction branch trained with the additional constraint on the entropy loss in all possible multi-target settings of the PACS dataset. Different colors denote different source domains (red: Art, yellow: Cartoon, blue: Photo, green: Sketch).}
  }
  \label{fig:latentDA-soft-assignment-cluster}
\end{figure*}

{In Figure~\ref{fig:latentDA-soft-assignment-cluster}, the same analysis is performed on our method with the additional constraint of having a uniform assignment distribution among domains. As the figure shows, this constraint allows to obtain a clearer domain separation in most of the cases, overcoming the difficulties that the domain prediction branch experienced in separating domain pairs such as Photo-Cartoon and Photo-Art.}

\begin{figure*}[t!]
 \centering {\includegraphics[width=0.48\textwidth]{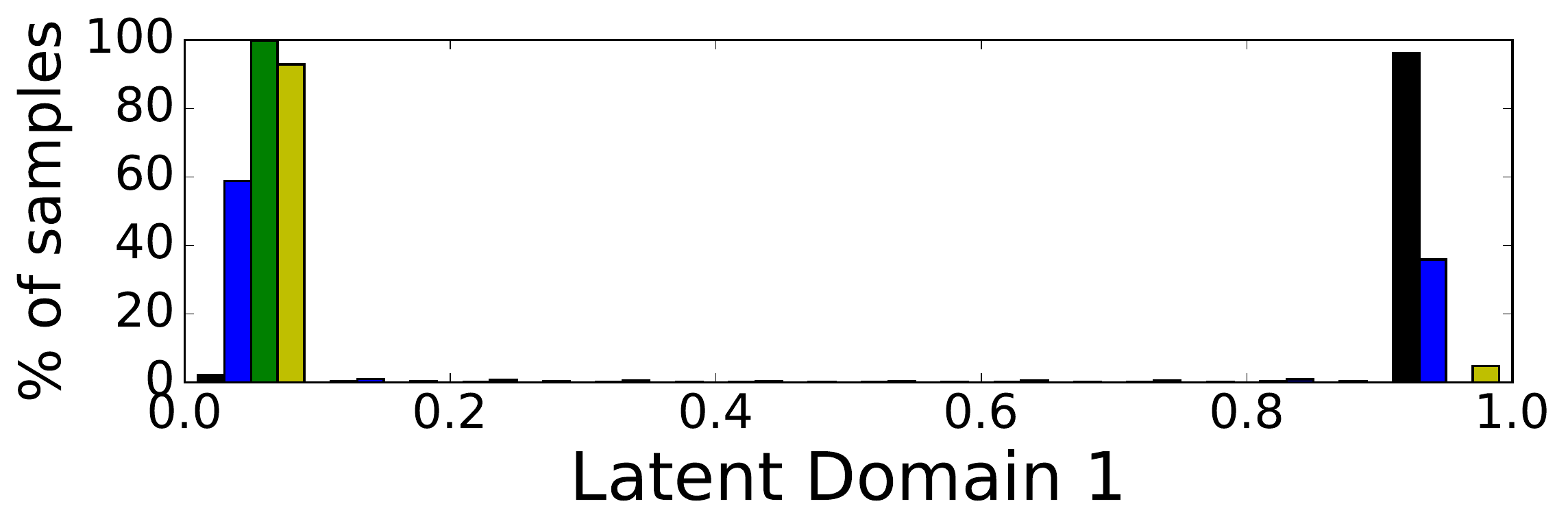}\hfill
  \includegraphics[width=0.48\textwidth,]{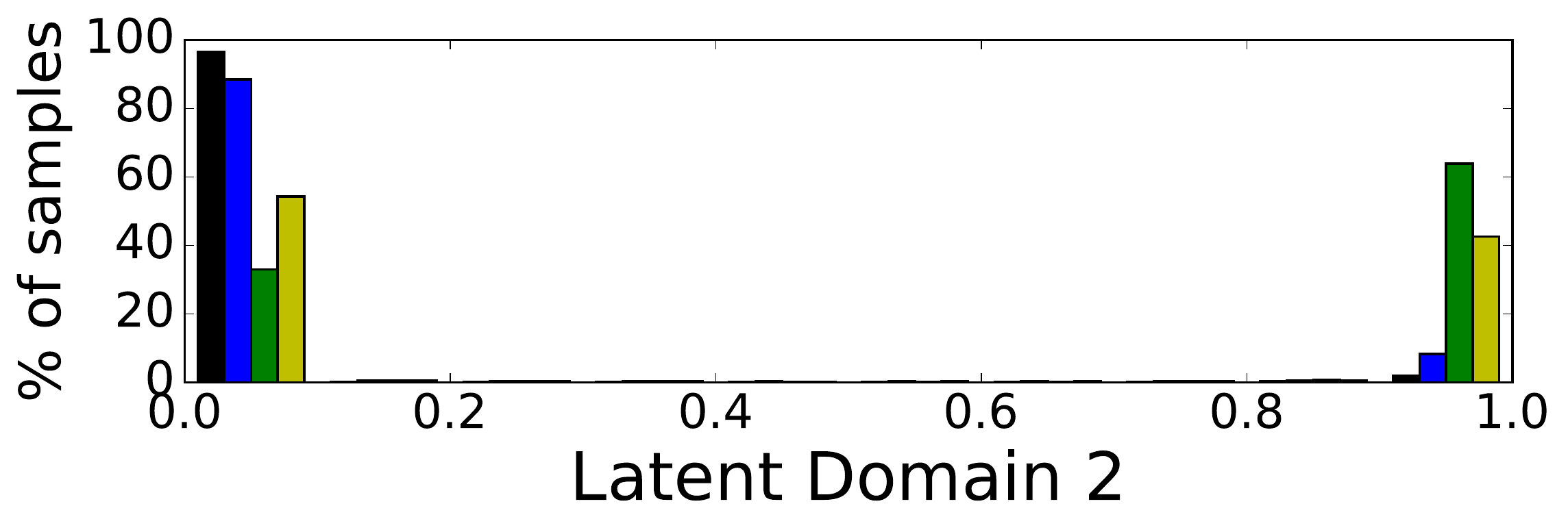}}
  \subfloat[SVHN as target]{\includegraphics[width=0.48\textwidth]{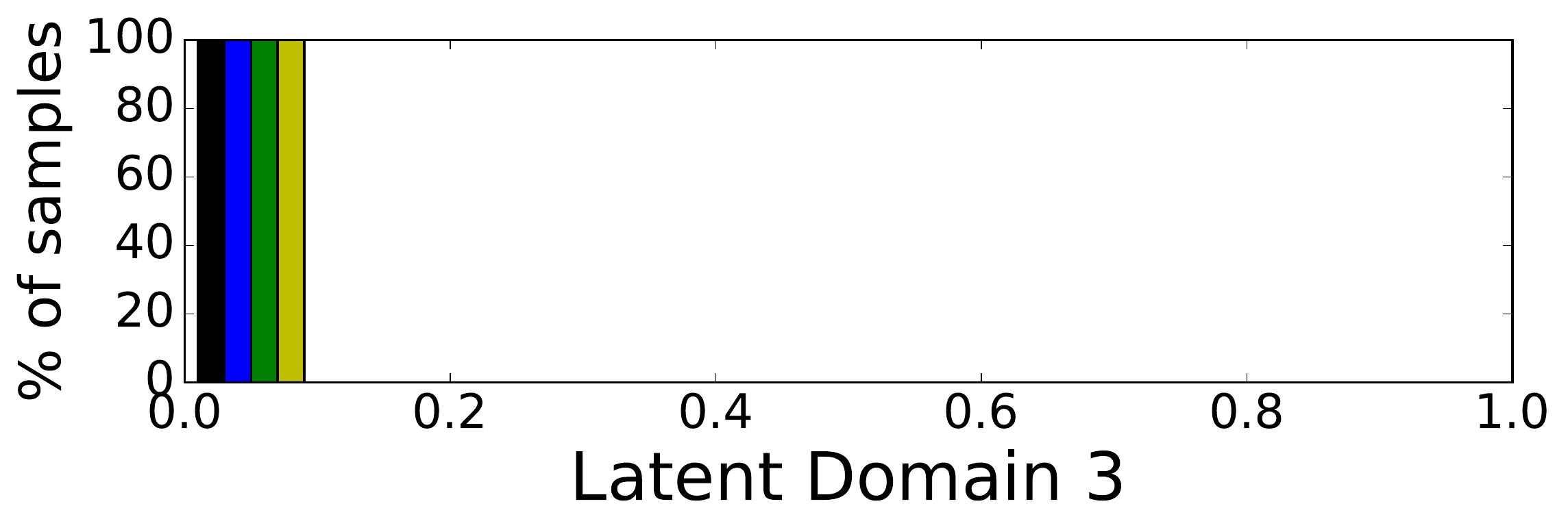}\hspace{12pt}
   \includegraphics[width=0.48\textwidth,]{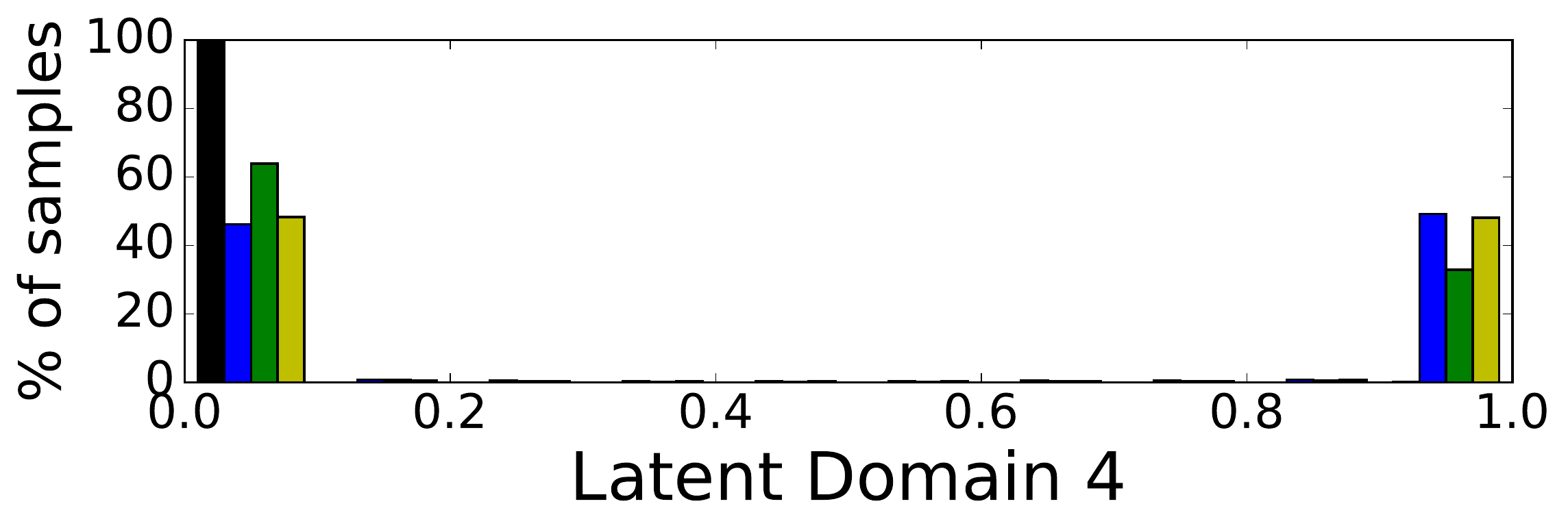}}

   \includegraphics[width=0.48\textwidth]{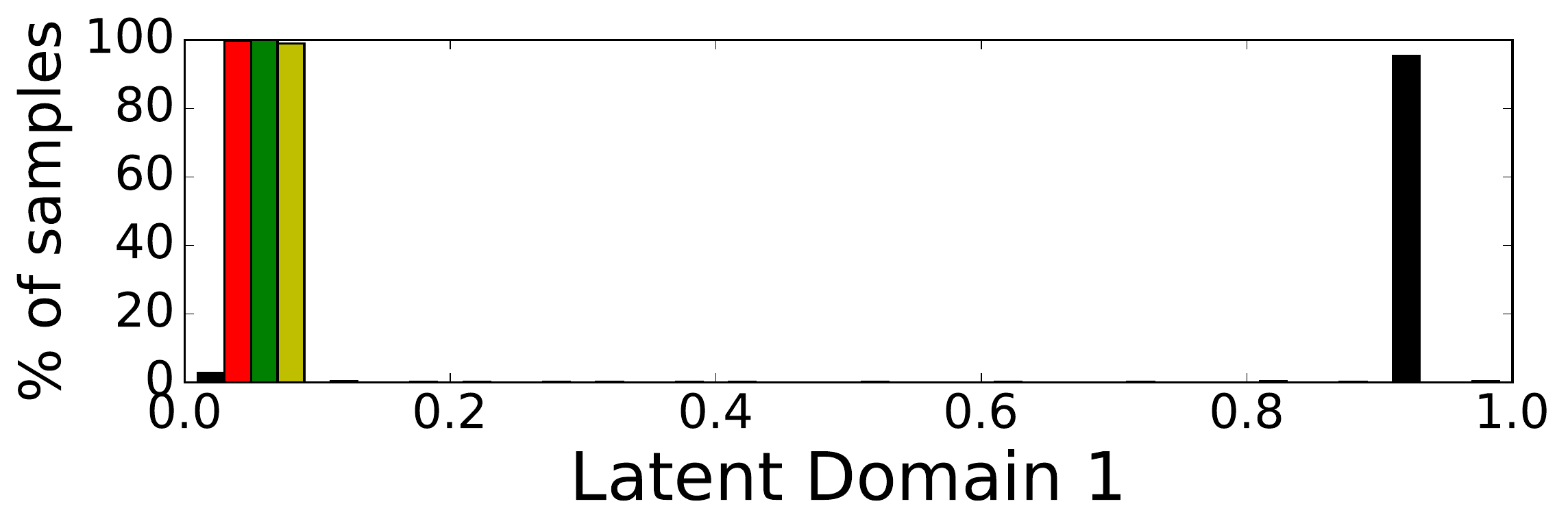}\hfill
  \includegraphics[width=0.48\textwidth,]{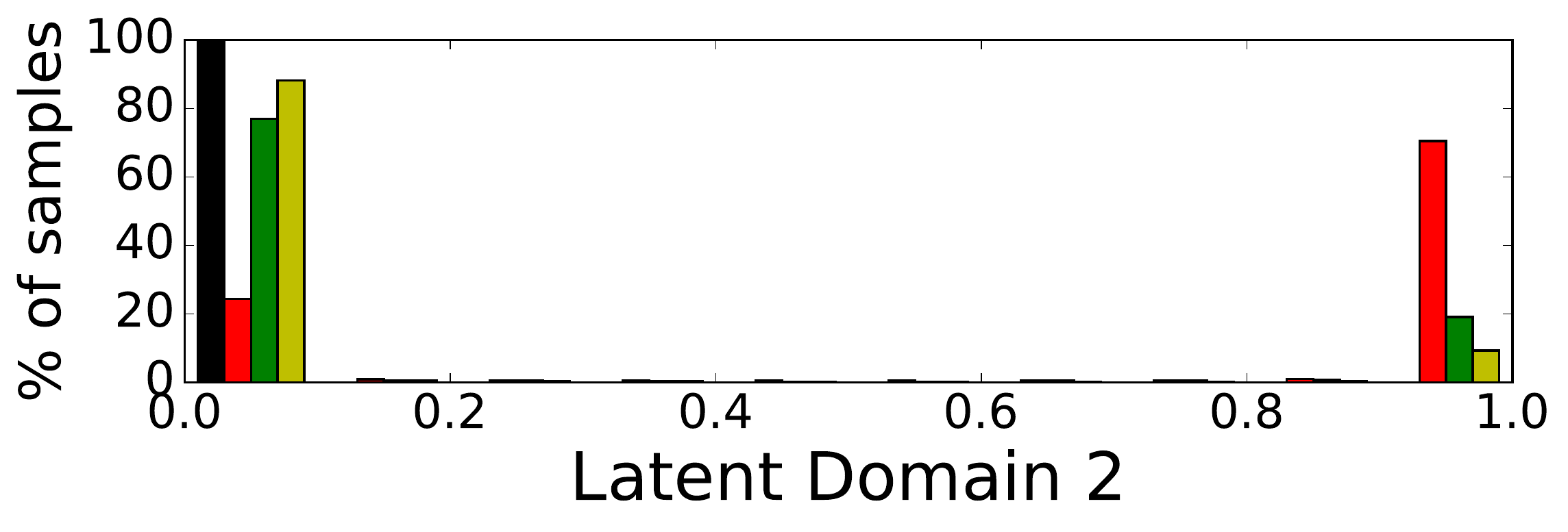}
  \subfloat[MNIST-m as target]{\includegraphics[width=0.48\textwidth]{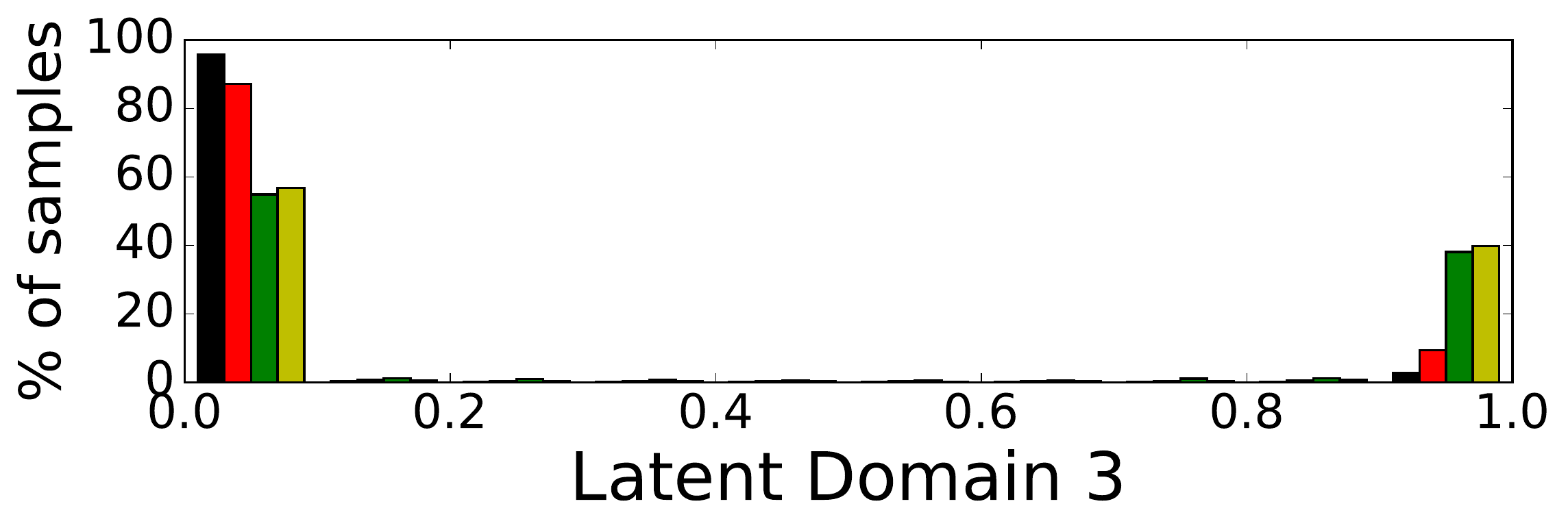}\hspace{12pt}
   \includegraphics[width=0.48\textwidth,]{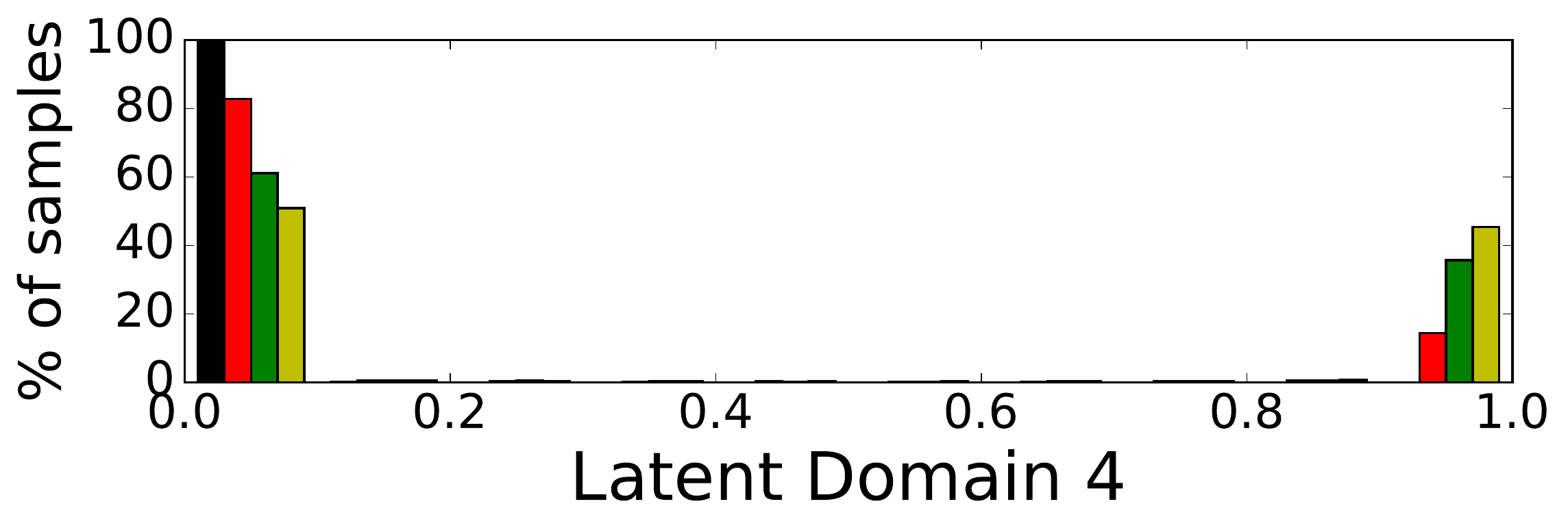}}
  
  \caption{{Distribution of the assignments produced by the domain prediction branch for each latent domain in all target settings of the Digits-five dataset. Different colors denote different source domains (black: MNIST, blue: MNIST-m, green: USPS, red: SVHN,  yellow: Synthetic numbers).}
  }
  \label{fig:latentDA-soft-assignment-digits}
\end{figure*}

{We perform a similar analysis in another dataset, Digits-five. The results are reported in Figure~\ref{fig:latentDA-soft-assignment-digits}. As the figure shows, when SVHN is the target domain, one of the latent domains (latent domain 1) receives very confident assignments for the samples of the MNIST dataset. The samples of the other source datasets receive assignment spread through all the latent domains, with the exceptions of USPS which receives the most confident predictions for the second latent domain and MNIST-m, which partially influences the first latent domain, the one with confidence assignments to MNIST. One latent domain does not receive assignments form any of the sources (latent domain three): this might happen if the entropy term overcomes the uniform assignment constraints in the early stages of training. Similarly, when MNIST-m is the target domain, the first two latent domains receive confident assignments for samples belonging to MNIST and SVHN datasets respectively, while the third and the fourth receive higher assignments for samples of the remaining source domains.}

\FloatBarrier


\FloatBarrier

\subsubsection{Experiments on Office-31}

In our Office-31 experiments we consider the following baselines, trained on the union of the source sets: (i) a plain AlexNet network; (ii) AlexNet with BN inserted after each fully-connected layer; and (iii) AlexNet + DIAL~\cite{carlucci2017just}.
Additionally, we consider single source domain adaptation approaches, using the results reported in~\cite{xu2018deep}.
The methods are Transfer Component Analysis (TCA) \cite{pan2010domain},  Geodesic Flow Kernel (GFK) \cite{gong2012geodesic}, Deep Domain Confusion (DDC) \cite{tzeng2015simultaneous}, Deep Reconstruction Classification  Networks (DRCN) \cite{ghifary2016deep} and  Residual  Transfer  Network  (RTN) \cite{long2016unsupervised}, as well as the  Reversed  Gradient (RevGrad) \cite{ganin2014unsupervised} and Domain Adaptation Network (DAN) \cite{long2015learning} algorithms considered in the digits experiments.
For these algorithms we report the performances obtained in the ``Best single source'' and ``Unified sources`` settings, as available from~\cite{xu2018deep}.
As in the previous experiments, Multi-source DA with perfect domain knowledge can be regarded as a performance upper bound for our method.
Finally, we include results reported in~\cite{xu2018deep} for different multi-source DA models: Deep Cocktail Network (DCTN) \cite{xu2018deep}, the two shallow methods in~\cite{xie2015learning} (sFRAME) and~\cite{gopalan2011domain} (SGF), and an ensemble of baseline networks trained on each source domain separately (Source only).
These results are summarized in Table~\ref{tab:latentDA-ablation-office}.

We note that, in this dataset, the improvements obtained by adopting a multi-source model instead of a single-source one are small.
This is in accordance with findings in~\cite{li2017deeper}, where it is shown that the domain shift in Office-31, when considering deep features, is indeed quite limited if compared to PACS, and it is mostly linked to changes in the background (Webcam-Amazon, DSLR-Amazon) or acquisition camera (DSLR-Webcam).
This is further supported by the smaller gap between DIAL and our method in this case compared to the previous experiments  {(average $p^*$ of 0.54)}. {In this setting, introducing our uniform loss term does not provides boost in performances. We ascribe this behaviour to the fact that in this scenario, each batch is built with a non-uniform number of samples per domain (following \cite{carlucci2017autodial}) while our current objective assumes a balanced sampling among domains. }

\begin{table}[t]
			\caption{Office-31 dataset: comparison of different methods using AlexNet. In the first row we indicate the source (top) and the target domains (bottom).} 
		\centering
		\scalebox{1.}{
		\begin{tabular}{ l | l r | c | c | c | c  } 
			\hline
			& \multirow{2}{*}{Method}& Source& A-W & A-D & W-D  & \multirow{2}{*}{{Mean}}\\
           & & Target & D & W & A  & \\\hline
           \multirow{7}{*}{\parbox{1.5cm}{Best single source \cite{xu2018deep}}} 
           &TCA\cite{pan2010domain}&&95.2&93.2&51.6&68.8\\
           &GFK\cite{gong2012geodesic}&&95.0&95.6&52.4&68.7\\
           &DDC\cite{tzeng2015simultaneous}&&98.5&95.0&52.2&70.7\\
           &DRCN\cite{ghifary2016deep}&&99.0&96.4&56.0&73.6\\
           &RevGrad\cite{ganin2014unsupervised}&&99.2&96.4&53.4&74.3\\
           &DAN\cite{long2015learning}&&99.0&96.0&54.0&72.9\\
           &RTN\cite{long2016unsupervised}&&99.6&96.8&51.0&73.7\\\hline\hline
           \multirow{6}{*}{\parbox{1.5cm}{Unified sources}} &
           Source only from \cite{xu2018deep}&&98.1 &93.2 &50.2&80.5\\
           &Source only  &		&94.6 &89.1	&49.1	&77.6\\
&{Source only+BN}  &	&91.9&92.7	&46.5	&77.0\\
&RevGrad\cite{xu2018deep}&&\textbf{98.8}&\textbf{96.2}&54.6&83.2\\
&DAN\cite{xu2018deep}&&\textbf{98.8}&95.2&53.4&82.5\\
&Single BN &	&92.9&95.2	&60.1	&82.7\\
&DIAL \cite{carlucci2017just} &
&93.8 &94.3	&62.5	&83.5\\\hline
&mDA $\lambda_B=0$ &
	&93.7	&94.6 &\textbf{62.6}	&83.6\\
	&mDA &&93.6&93.6&62.4&83.2
	\\\hline\hline






            \hline
            \multirow{5}{*}{\parbox{1.5cm}{Multi-source}}&Source only \cite{xu2018deep}&&98.2&92.7&51.6&80.8\\
            &sFRAME\cite{xie2015learning}&&54.5&52.2&32.1&46.3\\
            &SGF\cite{gopalan2011domain}&&39.0&52.0&28.0&39.7\\
            &DCTN \cite{xu2018deep}&&99.6&96.9&54.9&83.8\\            
           &Multi-source DA	&	&94.8&95.8	&62.9	&84.5\\ \hline
		\end{tabular}
        }
        \label{tab:latentDA-ablation-office}
\end{table}

\begin{figure}[ht!]
  \centering
  \includegraphics[width=0.75\textwidth,height=0.22\textheight]{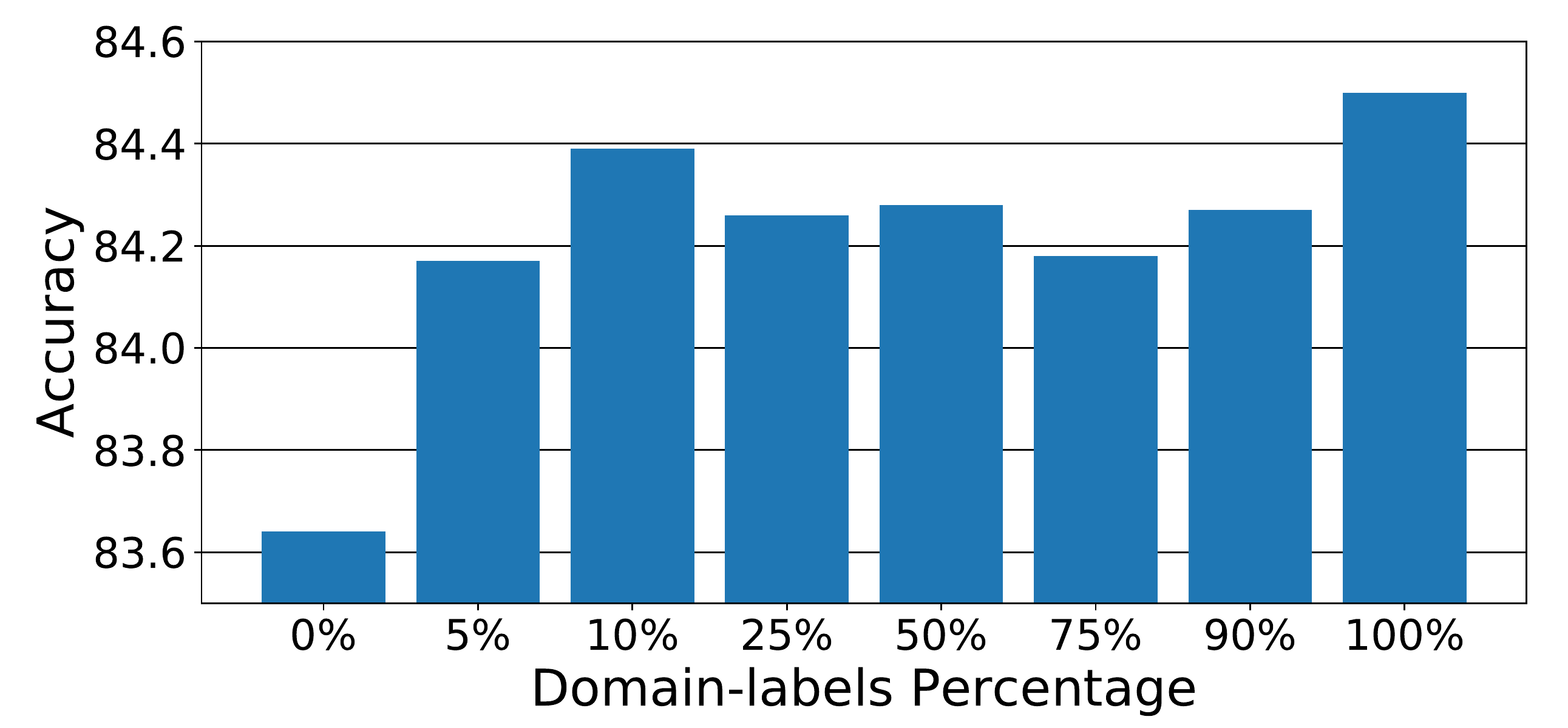}
  \caption{Office31 dataset. Performance at varying number of domain labels ($\%$) for source samples. 
  }
  \label{fig:latentDA-office-bars}
\end{figure}

In a final Office-31 experiment, we consider a setting where the true domain of a subset of the source samples is known at training time. Figure~\ref{fig:latentDA-office-bars} shows the average accuracy obtained when a different amount of domain labels are available.
Interestingly, by increasing the level of domain supervision the accuracy quickly saturates towards the value of Multi-source DA, completely filling the gap with as few as $5\%$ of the source samples.

\subsubsection{Comparison with S.o.t.A. on inferring latent domains}
In this section we compare the performance of our approach with previous works on DA which also consider the problem of inferring latent domains~\cite{hoffman2012discovering,xiong2014latent,gong2013reshaping}.
Since there are no previous works adopting deep learning models (i) in a multi-source setting and (ii) discovering hidden domains.
Therefore, the methods we compare to all employ handcrafted features.
For these approaches we report results taken from the original papers.
Furthermore, we evaluate the method of Gong~\etal~\cite{gong2013reshaping} using features from the last layer of the AlexNet architecture.
For a fair comparison, when applying our method we freeze AlexNet up to \texttt{fc7}, and apply mDA layers only after \texttt{fc7} and the classifier.

We first consider the Office-31 dataset, as this benchmark has been used in~\cite{hoffman2012discovering,xiong2014latent}, showing the results in Table~\ref{tab:latentDA-sota-office}.
Our model outperforms all the baselines, with a clear margin in terms of accuracy.
Importantly, even when the method in~\cite{gong2013reshaping} is applied to features derived from AlexNet, still our approach leads to higher accuracy.  
For the sake of completeness, in the same table we also report results from previous multi-source DA methods~\cite{gopalan2013unsupervised,nguyen2015dash,lin2017cross} based on shallow models.
While these approaches significantly outperform \cite{hoffman2012discovering} and \cite{xiong2014latent}, still their accuracy is much lower than ours. 
{Moreover, introducing our novel loss term provides higher performances with respect to the our approach with $\lambda_B=0$. }

\begin{table}[t]
			\caption{Office-31: comparison with state-of-the-art algorithms. In the first row we indicate the source (top) and the target domains (bottom).} 
		\centering
		\scalebox{.99}{
		\begin{tabular}{ l@{\hspace{-6ex}} r | c | c | c | c } 
			\hline
			 \multirow{2}{*}{Method}& Sources &A-D & A-W & W-D  & \multirow{2}{*}{Mean}\\          
             &Target & W & D & A  & \\  \hline
            Hoffman \etal \cite{hoffman2012discovering}&&24.8	&42.7	&12.8	&26.8\\
            Xiong \etal \cite{xiong2014latent}&&29.3	&43.6	&13.3	&28.7\\
            \hline
            Gong \etal (AlexNet) \cite{gong2013reshaping}	&&91.8&{94.6}	&48.9	&78.4\\
         mDA $\lambda_B=0$	&&{93.1}&
         94.3&{64.2}& {83.9}\\
                  mDA	&&\textbf{94.5}&\textbf{94.9}&\textbf{64.9}&\textbf{84.8}\\
            \hline\hline
                             Gopalan \etal  \cite{gopalan2013unsupervised} &&51.3	&36.1	&35.8	&41.1\\
          Nguyen \etal  \cite{nguyen2015dash} &&64.5	&68.6	&41.8	&58.3\\
Lin \etal \cite{lin2017cross} &&73.2	&81.3	&41.1	& 65.2\\ \hline
		\end{tabular}
        }
		\label{tab:latentDA-sota-office}
          
\end{table}

To provide a comparison in a multi-target scenario, we also consider the Office-Caltech dataset, comparing our model with \cite{hoffman2012discovering,gong2013reshaping}.
Following \cite{gong2013reshaping}, we test both single target (Amazon) and multi-target (Amazon-Caltech and Webcam-DSLR) scenarios.
As for the PACS multi-source\slash{}multi-target case, the assignment of each sample to the source or target set is assumed to be known, while the assignment to the specific domain is unknown.
We again want to remark that, since we do not assume to know the target domain to which a sample belongs, the task is even harder since we require a domain prediction step also at test time.
As in the Office-31 experiments, our approach outperforms all baselines, including the method in~\cite{gong2013reshaping} applied to AlexNet features.
{In this scenario, introducing our uniform loss provides a boost in performances in the multi-target setting, where the two source/target pairs have similar appearance. This is inline to what reported for the multitarget experiments on PACS (Table \ref{tab:latentDA-pacs-multitarget}). }

\begin{table}[t]
 			\caption{Office-Caltech dataset: comparison with state-of-the-art algorithms. In the first row we indicate the source (top) and the target domains (bottom).
            } 
 		\centering
 		\scalebox{.99}{
 		\begin{tabular}{ l@{\hspace{-6ex}} r | c | c | c | c } 
        \hline
        		\multirow{2}{*}{Method}	 & Source& A-C & W-D & C-W-D  & \multirow{2}{*}{Mean}\\          
            & Target& W-D & A-C & A  & \\
 			\hline
                Gong \etal \cite{gong2013reshaping} - original& &41.7	&35.8	&41.0	&39.5\\
                Hoffman \etal \cite{hoffman2012discovering} - ensemble	&&31.7	&34.4	&38.9	& 35.0\\
               Hoffman \etal \cite{hoffman2012discovering} - matching &&39.6	&34.0	&34.6	&36.1\\
               Gong \etal \cite{gong2013reshaping} - ensemble &&38.7	&35.8	&42.8	&39.1\\
                Gong \etal \cite{gong2013reshaping}	- matching &&42.6	&35.5	&44.6	&40.9\\
             \hline
           Gong \etal (AlexNet) \cite{gong2013reshaping} - ensemble&&{87.8}	&87.9	&93.6	&89.8\\
{mDA $\lambda_B=0$}&&{93.5}&	{88.2}&{93.7}&{91.8}\\
{mDA}&&\textbf{95.0}&	\textbf{88.7}&\textbf{93.9}&\textbf{92.5}\\
             \hline
 		\end{tabular}
         }
 		\label{tab:latentDA-sota-office-caltech}
 \end{table}

%
%

\subsection{Conclusions}
\label{sec:latentDA-conclusions}
In this section, we presented a novel deep DA model for automatically discovering latent domains within visual datasets. 
The proposed deep architecture is based on a side-branch that computes the assignment of source and target samples to their associated latent domain. These assignments are then used within the main network by novel domain alignment layers which reduce the domain shift by aligning the feature distributions of the discovered sources and the target domains.
Our experimental results demonstrate the ability of our model to efficiently exploit the discovered latent domains for addressing challenging domain adaptation tasks. 
Future works could investigate other architectural design choices for the domain prediction branch, as well as the possibility to integrate it into other CNN models for unsupervised domain adaptation \cite{ganin2014unsupervised}. In the next section, we will remove the assumption of having target data during training, focusing on the domain generalization scenario. We will show how mDA layers can be extended to effectively address the domain generalization problem.

\section{Domain Generalization}
\label{sec:da-dg}
In the previous section, we showed how it is possible to overcome the domain shift problem effectively even when our source/target domain is a mixture of multiple ones. However, it relies on a fundamental assumption: the presence of target data during training. Unfortunately, this assumption is not always satisfied in practice. 

Let us consider the problem of semantic place categorization from visual data \cite{wu2009visual}. This task is important in robotics, since correctly
identifying the semantic category of a place allows the robot to improve its localization, mapping and exploration \cite{stachniss2006speeding,kostavelis2015semantic} capabilities. We have three strategies to address this problem. The first is using labeled datasets of training images \cite{wu2011centrist,fazl2012histogram,urvsivc2016part,mancini2017learning}. 
While the resulting models are very accurate when test samples are similar to training data, 
their performance significantly degrade when the robot collects images with very different visual appearance \cite{pronobis2010realistic}. 

A second strategy could be exploiting \emph{domain adaptation (DA)} techniques \cite{prasath2012transfer,kira2014transfer,costante2013transfer}. These methods develop models which are meant to be effective in the scenario where the robot will operate, \ie the \emph{target} domain.
While domain adaptation algorithms provide effective solutions, they require some prior knowledge of the target domain  at training time, \eg to have access to target data. 
Unfortunately, this information may not always be available. Consider for instance an household robot: since the number of possible customers is huge, it is inconceivable to collect data for each possible house and application scenario. 

In this context, a more relevant problem to address is \emph{domain generalization (DG)}. As described in previous sections, opposite to DA, where target data are exploited to produce a classifier accurate under specific working conditions, the idea behind DG is to learn a domain agnostic model applicable to any unseen target domain. In other words, the goal of DG is building a model which is 
as general as possible, e.g. employable by different robots and in various environmental conditions. 
 

\begin{figure}[t]
\centering
\includegraphics[width=0.9\columnwidth,trim={0 0.75cm 3.5cm 0},clip]{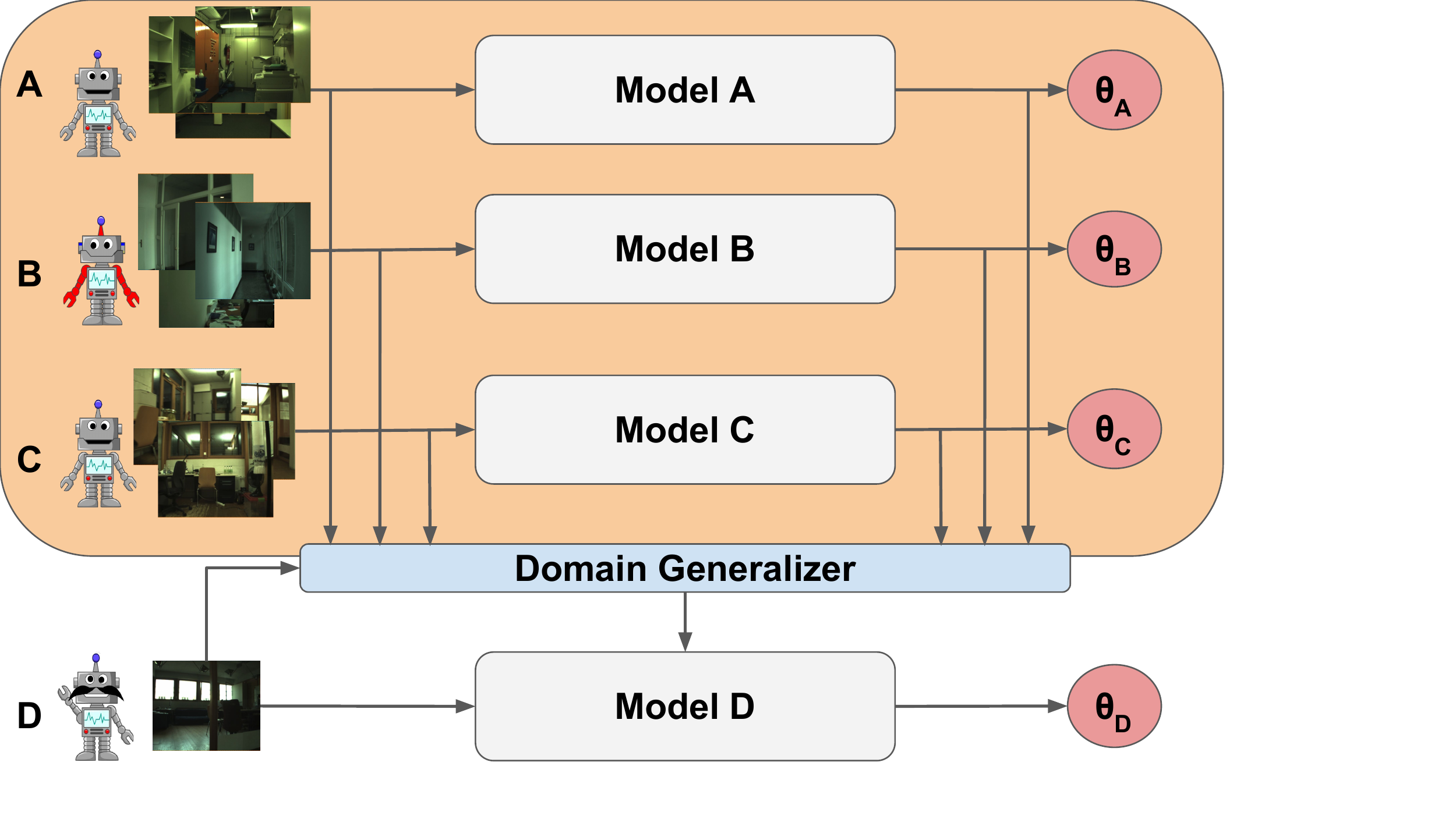}
\caption{The domain generalization problem. At training time (orange block) images of multiple source domains (\eg 
A,B,C) are available. These images are used to train different models with parameters $\theta_i$. 
Our approach automatically computes a model D which accurately classifies images of a novel domain (not available during training)
by combining the models of the known domains.}
   \label{fig:wbn-teaser}
\end{figure}

In this section, we build on the mDA layers presented in Section \ref{sec:da-latent} and we first propose a novel deep learning framework for DG, namely We call this approach WBN (\textbf{W}eighted \textbf{B}atch \textbf{N}ormalization for Domain Generalization) \cite{mancini2018robust}. The approach develops from the idea that, given data from multiple source domains and the associated models, 
the best model for the target domain can be generated on-the-fly when a novel sample arrives by optimally combining the precomputed models from source domains (see Fig.\ref{fig:wbn-teaser}).  To implement this idea we design a novel CNN architecture which relies on two main components. First, inspired by recent works on domain adaptation \cite{carlucci2017just,li2016revisiting}, we construct multiple source models by embedding into a common CNN few domain-specific Batch Normalization layers. In this way, different classifiers can be built keeping the number of parameters limited. Second, we design a lateral network branch which computes the likelihood that a certain instance belongs to a given domain. When applied to a novel target sample, this branch calculates 
its 
probabilities to be part of the different source domains. These values are used to construct the target classifier performing a combination of known source models. This is similar to the idea of mDA layers, with the difference that (i) no target data are available during training and (ii) the domain assignment branch is used to compute the similarity of target samples with source domains. 

In the second part of this section, we extend this approach by considering domain-specific classifiers, and classifying each incoming target image by optimally fusing the prediction scores of the source-specific classifiers. 
As in WBN, this is achieved through an end-to-end trainable deep architecture with two main components. The first implements the source-specific classifiers, while the second module is a network branch which computes the similarities of an input sample to all source domains, such as to assign weights to the source classifiers and properly merge their predictions. The second module is also designed in order to easily permit, if needed, the integration of a domain agnostic classifier which, acting in synergy with the domain-specific models, can further improve generalization. We call this approach \textbf{B}est \textbf{S}ources \textbf{F}orward for Domain Generalization (BSF) \cite{mancini2018best}.

 To this aim, the novel Weighted Batch Normalization (WBN) layers are introduced. We demonstrate the effectiveness of the proposed DG approach with extensive experiments on three 
datasets, namely the COsy Localization Database (COLD) \cite{pronobis2009ijrr}, the Visual Place Categorization (VPC) dataset \cite{wu2009visual} {and the Specific PlacEs Dataset (SPED) \cite{chen2017deep}}. Moreover, we show how the proposed framework can be employed where no prior information about source domains is available at training time: given a training set, our model can be used to automatically cluster training data and learning multiple models, discovering latent domains and associated classifiers.

To summarize, the main contributions of this section are:
(i) an extension of the mDA framework, WBN, which exploits the similarity of target samples with the given source domains to address the DG problem; (ii) we introduce the problem of domain generalization for semantic place recognition, showing how WBN is effective in addressing it, even without exact domain knowledge; (iii) we extend of WBN, by considers source-specific classifiers in place of domain-specific alignment layers, showing its effectiveness in standard DG benchmarks in computer vision. 

\subsection{Problem Formulation}
The goal of DG is to extend the knowledge acquired from a set of source domains to any unknown target domain. In this context, the source sets correspond, \eg, to data acquired by multiple robots in different environments while the unknown target to any unseen environment. Formally, following the notation in Section \ref{sec:da-problem}, we have our training set defined as $\mathcal{S}=\{(x^s_i,y^s_i,s_i)\}_{i=1}^n$ where $x^s_i\in\mathcal{X}$, $y^s_i \in \mathcal{Y}$ and $s_i \in \mathcal{D}^s$, with $\mathcal{D}^s \subset \mathcal{D}$. Note that \textit{no target domain data} $\mathcal{T}$ is available during training. Moreover, we assume $|\mathcal{D}^s|=\con k_s>1$, analyzing in Sections \ref{sec:da-continuous} and \ref{sec:da-predictive} the case where $|\mathcal{D}^s|=1$ but other information is available. Our goal is to learn a predictor $f:\mathcal{X}\rightarrow \mathcal{Y}$ able to work in any possible target domain $\mathcal{D}^t$ \textit{unseen} during training, i.e. $\mathcal{D}^s\neq \mathcal{D}^t$. It is worth highlighting that, differently from the Latent Domain Discovery problem presented in Section \ref{sec:da-latent}, here we might have full knowledge about the domain labels. 

\begin{figure}[t]
  \centering
  \subfloat[AlexNet+BN]{\includegraphics[width=0.8\columnwidth,trim={0 5cm 0 4cm},clip]{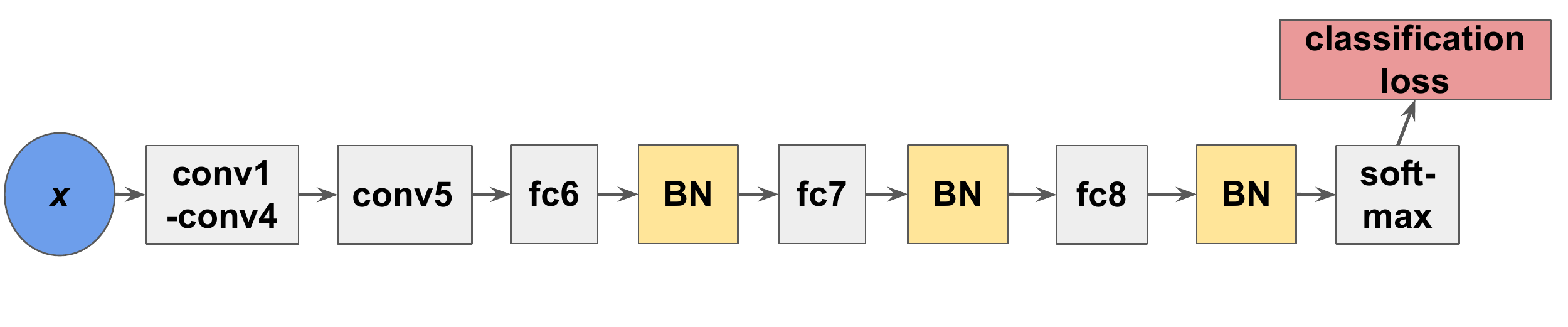}\label{fig:bn}}
  \hfill
  \subfloat[AlexNet+DA layers]{\includegraphics[width=0.8\columnwidth,trim={0 4cm 0 3.2cm},clip]{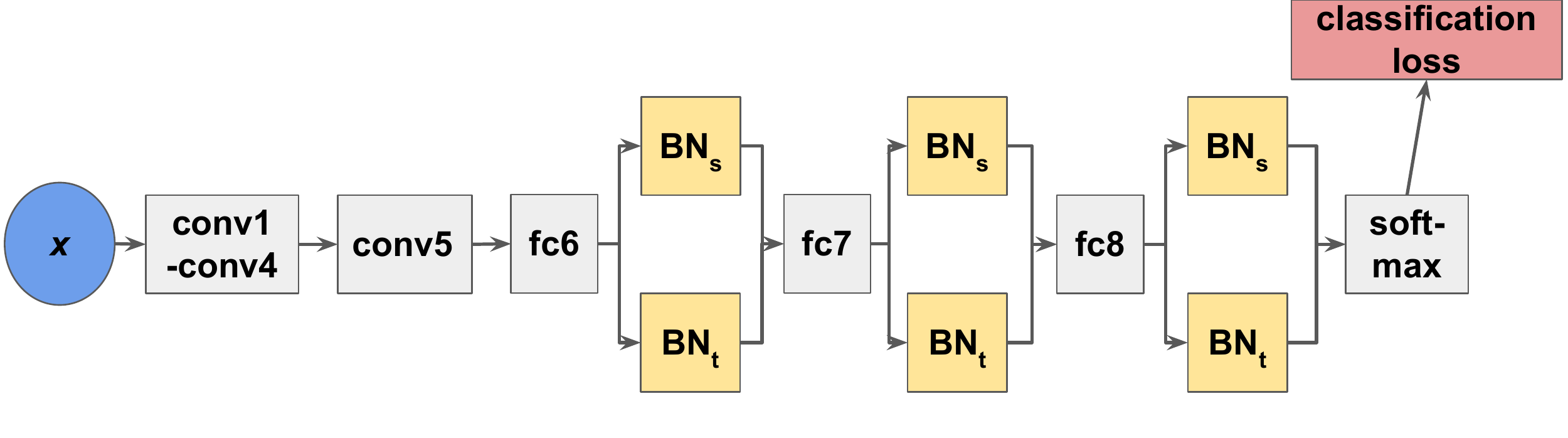}\label{fig:da-bn}}
  \hfill
  \subfloat[AlexNet+WBN]{\includegraphics[width=0.8\columnwidth,trim={0 1.2cm 0 0},clip]{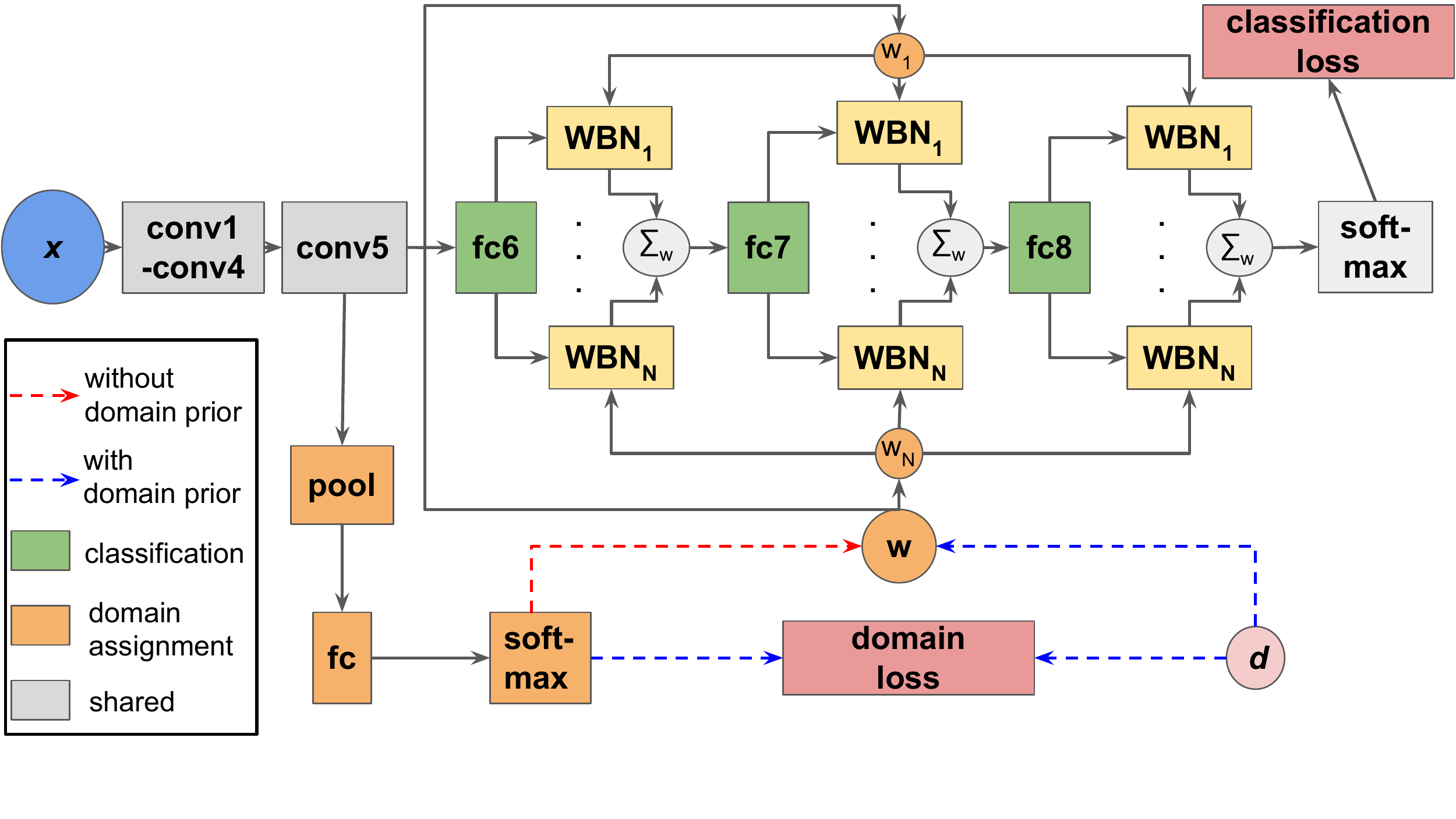}\label{fig:wbn}}
  \caption{Example of the proposed WBN framework. (a) AlexNet with BN layers after each fully connected. (b) The same network employing Domain Alignment layers for domain adaptation, where different BN are used for source and target domains. (c) Our approach for DG with WBN layers. 
  }
  \label{fig:wbn-method}
\end{figure}

\subsection[Starting point: Domain Generalization with Weighted BN]{Starting point: Domain Generalization with Weighted BN \footnotemark\footnotetext{M. Mancini, S. Rota Bul\`o, B. Caputo, E. Ricci. {\sl Robust Place Categorization with Deep Domain Generalization}.
       IEEE Robotics and Automation Letters, July 2018, vol. 3, n. 3., pp. 2093-2100.}}
\label{sec:wbn}
A clear issue with DA methods, including DA-layers and mDA, is that they require the presence of a target set $\mathcal{X}_t$ in the training phase. This implies that data collected in the scenario of interest should be available for learning the classification model. However, a more realistic situation, especially in robotics, is when we employ our system in completely unseen environments/domains. As an example, consider a service robot: it is unfeasible to collect data for all possible working environments. Therefore, it is important to drop the assumption of having target data beforehand while designing deep models addressing the domain shift problem. In this subsection, we start by removing target data from DA-layers and mDA layers.

From the formulation of DA-layers defined in Eq.~\eqref{eq:da-domainalignment}, we can obtain multiple, domain-specific models by considering separate BN statistics for each of the source domains during training. In particular, given the features of a sample $x_i$ at a given layer and spatial location (omitted for simplicity) as well as its domain label $s_i$, we can apply the domain-specific BN as follows: 
\begin{equation}
\label{eq:dg-hard1}
\hat{x}_i =\gamma \frac{x_i-\mu_{s_i}}{\sqrt{\sigma_{s_i}^2 + \epsilon}} + \beta\,.
\end{equation}

The problem with this formulation is that, at test time, no statistics from the unseen target domains are available. To solve this problem, we restore to a soft-version of Eq.~\eqref{eq:dg-hard1}. Let us write Eq.~\eqref{eq:dg-hard1} as:  
\begin{equation}
\label{eq:dg-hard2}
\hat{x}_i =\gamma \sum_{j=1}^{\con k_s}\mathds{1}_{s_i=s_j}\frac{x_i-\mu_{s_j}}{\sqrt{\sigma_{s_j}^2 + \epsilon}} + \beta\,.
\end{equation}
The hard-assignment in Eq.~\eqref{eq:dg-hard2}, used at training time, can be replaced with a weighted version at test time, modeling the uncertainty we have about our target domain. In particular, we can write:
\begin{equation}
\label{eq:dg-soft}
\hat{x}^t_i =\gamma\sum_{j=1}^{\con k_s} w_{i,j}\frac{x_i-\mu_{s_j}}{\sqrt{\sigma_{s_j}^2 + \epsilon}} + \beta\,,
\end{equation}
where $w_{i,j}$ is the probability of sample $i$ to belong to domain $j$, with $\sum_{j=1}^{N}w_{i,j} = 1$ and $\forall j\; w_{i,j}\geq 0$. {The intuition behind this choice is deriving a classification model for the target domain as a combination of models from the source domains, with the weights derived from the similarity of the target domain data to the source domains.}

In order to compute the weights $w_{i,j}$, we restore to the same domain classification module described in Section \ref{sec:latentDA-domain-prediction}, employing a separate network branch which originates from the first few convolutional layers of the main network (see Fig. \ref{fig:wbn}). This choice is motivated by the fact that end-to-end training is allowed and the number of parameters is kept limited. The specific architecture of the branch may be variable, with the only restriction that its final output must be a probability vector of dimension $\con k_s$, corresponding to the number of known domains.

Denoting the classification branch as $f_C^\theta$ and $f_D^\theta$, during training we minimize a simplified version of Eq.~\eqref{eq:latent-loss_general}, namely:
\begin{equation}
\label{eq:dg-wbn-stdloss}
L(\theta)=-\frac{1}{\con n}\sum_{i=1}^{\con n} \log f_C^\theta(y_i^s; x_i^s) + \lambda \log f_{D}^\theta(d_i; x_i)
\end{equation}
the loss is the sum of two terms, one considering place label information for accurate recognition, the other enforcing the lateral branch to successfully compute the correct domain, with $\lambda$ balances the contribution of the semantic classification and the domain prediction terms. At test time, the domain assignment produced by $f_D^\theta$ for target samples will be used to obtain the domain similarity $w_{.,j}$ of Eq.~\eqref{eq:dg-soft}.

Finally, we would like to highlight that this framework can be easily extended to perform DG in the lack of domain labels, following what described in Section \ref{sec:da-latent}. In particular, we can rely on the soft-assignment strategy to compute the latent domain statistics, as in Eq.~\eqref{eqn:mixture-params}. As in the previous section, the intuition is that, since similar input images will tend to produce similar outputs in the lateral network branch, implicitly visual data will be automatically clustered, enabling a latent domain discovery process. In this scenario, we let the domain assignment network be guided by the semantic loss while computing the statistics using Eq.~\eqref{eqn:mixture-params}.  In Fig.\ref{fig:wbn-method} we show the difference between this model and standard DA-layers.

\subsection{WBN Experiments: Domain Generalization in Semantic Place Categorization}
\myparagraph{Datasets.} In our experiments we use three robot vision datasets, namely the widely adopted COLD \cite{pronobis2009ijrr} and VPC  \cite{wu2009visual} datasets, {and the recent SPED dataset \cite{chen2017deep}}. 

The \textbf{COLD} Database contains three datasets of indoor scenes acquired in different laboratories and from different robots. The COLD-Freiburg (Fr) has 26 image sequences collected in the Autonomous Intelligent Systems Laboratory at the University of Freiburg, with a camera mounted on an ActivMedia Pioneer-3 robot. COLD-Ljubljana (Lj) contains 18 sequences acquired from 
an iRobot ATRV-Mini platform at the Visual Cognitive Systems Laboratory of University of Ljubljana. In the COLD-Saarbr\"ucken (Sa) an ActivMedia PeopleBot has been employed to gather 29 sequences inside the Language Technology Laboratory at the German Research Center for Artificial Intelligence in Saarbr\"ucken. 

The \textbf{VPC} dataset contains images acquired from several rooms of 6 different houses with multiple floors. The images are acquired by means of a camcorder placed on a rolling tripod, simulating a mobile robotic platform. The dataset contains 11 semantic categories, but only 5 are common to all houses: bedroom, bathroom, kitchen, living room and dining-room. Following previous works \cite{wu2009visual,fazl2012histogram,yang2012object}, we use the common categories in our experiments. 

\textbf{SPED} is a large scale dataset introduced in the context of place recognition. It contains images of 2543 outdoor cameras collected from the Archive of Many Outdoor Scenes (AMOS) \cite{jacobs2007consistent} during February and August 2014\footnote{The full dataset was not available at the time we proposed WBN, but the authors provided us a subset with about 500 images per camera corresponding to 900 categories.}

\myparagraph{Networks and training protocols.} For COLD and VPC we perform experiments with two common architectures: AlexNet \cite{krizhevsky2012imagenet} and ResNet \cite{he2016deep}. For AlexNet we use the standard architecture pre-trained on Imagenet \cite{deng2009imagenet}. In all the experiments, we fine-tune the last two fully-connected layers, rescaling the input images to 227 $\times$ 227 pixels. For ResNet we consider the 10 layers version of the architecture, again pre-trained on ImageNet. In all the experiments, we rescale the input images to 224x224 pixels, fine-tuning the network starting from the last residual block.  Both the networks are trained with a weight decay of 0.0005 and an initial learning rate of 0.001, while the initial learning-rate of the final classifier is set to 0.01. The learning rate is dropped of a 0.1 factor after $90\%$ of the iterations. For the experiments on COLD, we use a batch size of 256 for AlexNet and 64 for ResNet, 
training the networks for 1000 iterations. For VPC, we set the batch size to 128 and 64 for AlexNet and ResNet respectively, training the networks for 2000 iterations. The training parameters are the same for our method and the baselines and fine-tuning is performed for all the models.

WBN can be applied to common CNNs by simply replacing standard BN layers with our WBN layers. While for ResNet BN layers are already employed, this is not true for AlexNet. For these experiments we employ a variant of AlexNet where BN layers are inserted after each fully-connected layer.

{For SPED we 
use AlexNet and the AMOSNet architecture, following \cite{chen2017deep}. AMOSNet is very similar to {AlexNet}, with the first fully-connected layer replaced by a convolutional layer and a pooling operation. We follow the same protocol of \cite{chen2017deep}, using the same hyperparameters for training. 
We train both networks from scratch, applying BN or WBN layers after each layer with parameters, except the classifier. } The implementation details of the domain assignment branch follows the one of Section \ref{sec:latentDA-experiments} and we set $\lambda=1$ for all the experiments.  


The evaluation was performed using a NVIDIA GeForce 1070 GTX GPU, implementing all the models with the popular Caffe \cite{jia2014caffe} framework. For the baseline AlexNet architecture we take the pre-trained model available in Caffe, while for ResNet we consider the model from \cite{simon2016cnnmodels}. The code implementing the WBN layers is publicly available\footnote{https://github.com/mancinimassimiliano/caffe}

\myparagraph{Results on COLD.} We first perform experiments on the COLD database, where the goal is to demonstrate the effectiveness of WBN in learning effective classification models in case of varying environmental conditions (\eg illuminations, laboratories).. For each laboratory and illumination condition we consider the standard sequences 1 of part A, except for Saarbr\"ucken Cloudy, for which we take sequence 2 due to known acquisition issues\footnote{http://www.cas.kth.se/COLD/bugs.php} and Saarbr\"ucken Sunny, for which we take part B since sunny sequences for part A are not available. We consider the 4 classes shared between the sequences: printer area, corridor, bathroom and office (obtained by merging 1-person and 2-persons office). We report the results as the average accuracy per class. In these experiments we consider both AlexNet and ResNet comparing WBN with baseline models obtained adding traditional BN layers to the same architectures.

We test two different variants of the proposed approach. In the first case (WBN$ ^*$) we consider the presence of domain priors at training time, as in Section \ref{sec:wbn}. In the second variant, WBN, we do not assume to have knowledge about domains at training time, thus our model just relies on the soft-assignment. We highlight that WBN with soft-assignment is similar to the mDA layers of Section \ref{sec:da-latent} except that (i) no loss is applied on the domain prediction branch and (ii) no target data are available during training, thus no statistics are available for them and we must rely on the domain prediction branch also at test time.

Firstly, we consider different lighting conditions, \ie we assume that the domain shift is due to changes of illuminations. To this extent we train the network on sequences of the same laboratory, training on two lighting conditions (\eg \textit{sunny} and \textit{cloudy}) and testing on the third (\eg \textit{night}). The results are reported in Table \ref{tab:cold-dg-lab}. 

{As expected, when knowledge about domains is available (WBN$ ^*$), improved classification accuracy can be obtained, in general, with respect to a domain agnostic classifier. Interestingly, for both networks the result of WBN without domain priors is either comparable or surpasses the baseline in almost all settings. This suggests that the network is able to latently discover clusters of samples and effectively using this information for learning robust classification models. 
}

\begin{table}[t]
			\caption{DG accuracy on COLD over different lighting conditions. First row indicates the target sequence, with the first letters denoting the laboratory and the last the illumination condition (C=Cloudy, S=Sunny, N=Night). Vertical lines separate domains of the same laboratory. * indicates the algorithm uses domain knowledge.} 
		\centering
		\scalebox{.98}{
		\begin{tabular}{| c | l@{\hskip3pt} | c@{\hskip5pt}  c@{\hskip5pt}  c@{\hskip5pt} | c@{\hskip5pt} c@{\hskip5pt} c@{\hskip5pt} | c@{\hskip5pt} c@{\hskip5pt} c@{\hskip5pt} | c | } 
			\hline
			Net& Norm. & Fr.C & Fr.N & Fr.S & Lj.C & Lj.N & Lj.S &Sa.C &Sa.N & Sa.S& avg.\\
			\hline
					
 \multirow{3}{*}{\rotatebox[origin=c]{90}{AlexNet}}&BN&97.3	&89.1	&97.4	&92.9	&64.4	&94.2	&75.6	&69.7	&44.0	&80.5\\
 &WBN&\textbf{98.1}	&91.3	&97.1	&93.1	&65.1	&94.1	&\textbf{77.7}	&68.8	&\textbf{50.2}	&81.7\\
 &WBN$ ^*$&97.1	&\textbf{91.9}	&\textbf{98.0}	&\textbf{93.9}	&\textbf{65.6}	&\textbf{95.0}	&77.2	&\textbf{69.9}	&49.9	&\textbf{82.1}\\
\hline
\multirow{3}{*}{\rotatebox[origin=c]{90}{ResNet}}&BN&97.7	&\textbf{82.2}	&90.7	&89.5	&61.2	&90.3	&70.7	&73.0	&\textbf{38.7}	&77.1\\
&WBN&\textbf{98.1}	&81.8	&\textbf{94.1}	&94.5	&61.7	&93.7	&75.8	&76.9	&37.8	&79.4\\
&WBN$ ^*$&97.9	&81.3	&93.4	&\textbf{94.7}	&\textbf{65.1}	&\textbf{94.6}	&\textbf{78.1}	&\textbf{76.5}	&38.5	&\textbf{80.0}\\
            \hline
		\end{tabular}
        }
		\label{tab:cold-dg-lab}
\end{table}

Secondly, we perform a similar analysis to Table \ref{tab:cold-dg-lab} but considering changes of robotic platform\slash{}environment. We keep constant the lighting condition, training on two laboratories and testing on the third. Table \ref{tab:cold-dg-lig} shows the obtained results. Again, in most cases exploiting domain priors brings benefits in term of performances, for both networks. The results of Tables \ref{tab:cold-dg-lab} and \ref{tab:cold-dg-lig} show that the benefits of our WBN layer, with and without domain loss, are not limited to a particular type of domain shift (\ie changes in robots, environment or illumination condition), demonstrating that our approach provides a general and effective strategy to address domain variations. 
{In both experiments, there are few cases in which standard BN achieves comparable or slightly superior results w.r.t. WBN. A possible reason is that in some situations the ability of our model to generalize to novel settings may be hindered by the small number or by the specific characteristics of the available source domains.

{In order to verify the ability of WBN to discover latent domains, Fig. \ref{fig:vis-cold} shows the distribution of the values $\hat{w}_{i,j}$ computed for the images of the original source domains associated to one of the experiments in Table \ref{tab:cold-dg-lig}. The plots associated to other experiments are similar and we do not report them due to lack of space. Since we consider two latent domains in these experiments and $\hat{w}_{i,1}+\hat{w}_{i,2}=1$, we report only the values computed for $\hat{w}_{i,1}$
. Different colors represent the original source domains. As the figure shows, the lateral branch computes different assignments for the samples of the different original source domains. As a result, the latent source domains extracted by WBN tend to correspond to the original ones used by WBN$^*$.} 

\begin{figure}[t!]
   \begin{minipage}{0.48\textwidth}
     \centering
     \includegraphics[width=1.0\columnwidth,trim={0cm 0.6cm 0cm 9.5cm},clip]{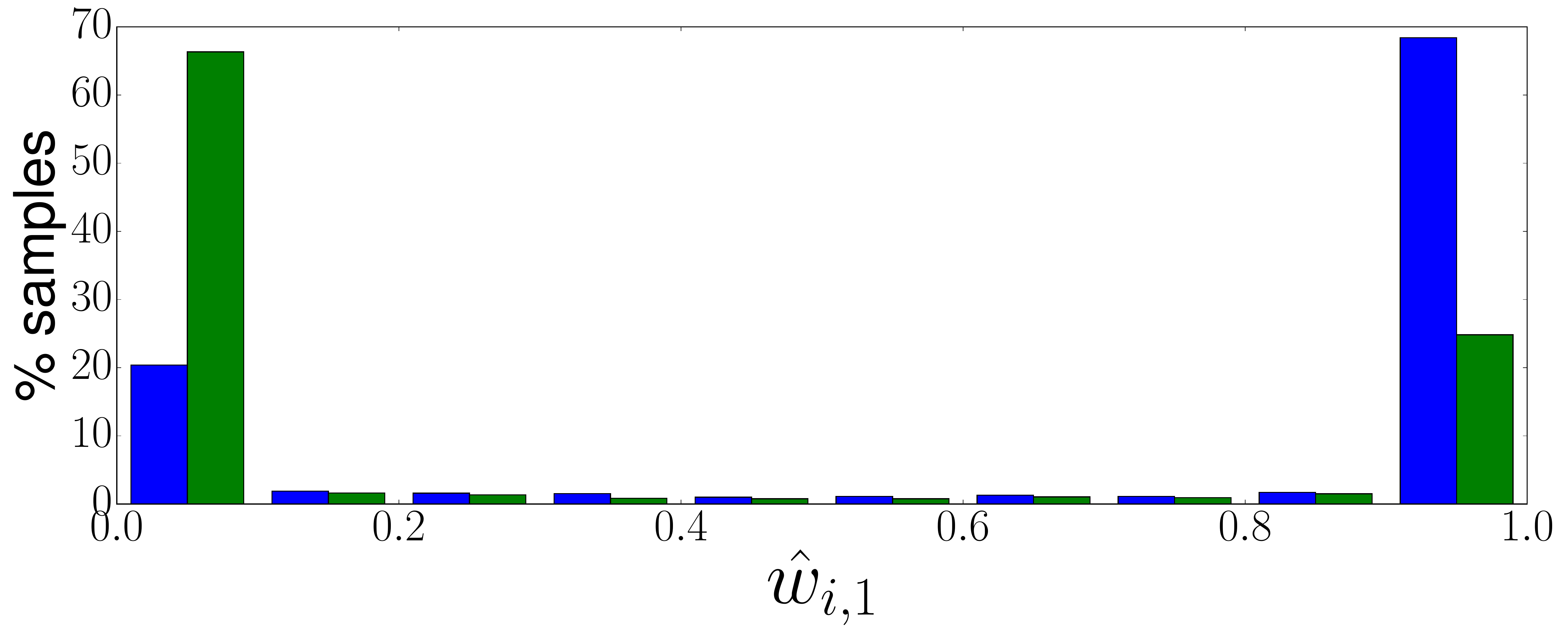}
  \caption{Distribution of the values of the weights computed with AlexNet+WBN for the scenario Lj.N as target in Table \ref{tab:cold-dg-lig}. Different colors represent different original source domains.}
   \label{fig:vis-cold}
   \end{minipage}\hfill
   \begin{minipage}{0.48\textwidth}
     \centering
     \includegraphics[width=1.0\columnwidth,trim={0cm 0.48cm 0cm 8.8cm},clip]{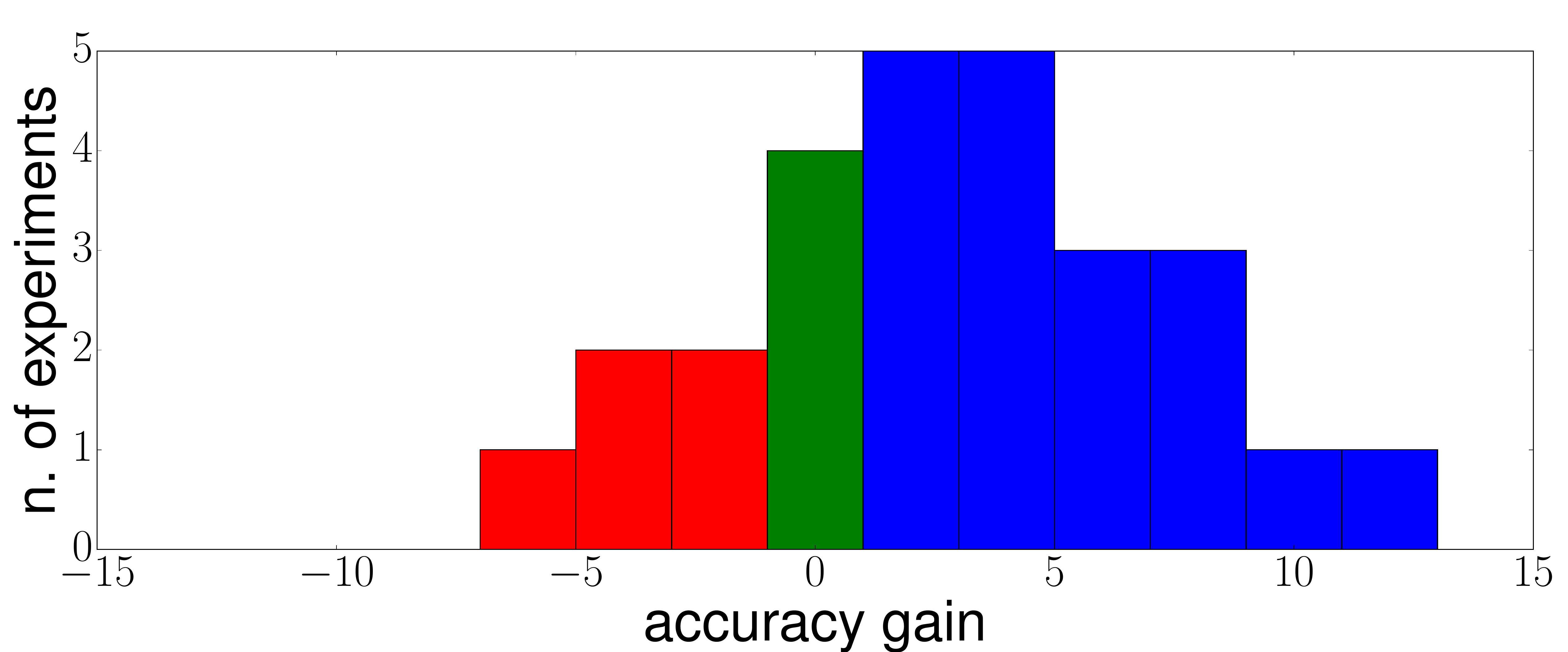}
\caption{Distribution of accuracy gains of AlexNet+WBN* w.r.t. AlexNet+BN considering Saarbr\"ucken as target, varying both laboratory and illumination. Colors indicate larger (blue), lower (red) and comparable (green) performances.}
   \label{fig:hist-81-cold}
   \end{minipage}
\end{figure}


\begin{table}[t]
			\caption{DG accuracy on COLD over different environments\slash{}sensors. First row indicates the target sequence, with the first letters denoting the laboratory and the last the illumination condition (C=Cloudy, S=Sunny, N=Night). Vertical lines separate domains with same illumination condition. * indicates the algorithm uses domain knowledge.} 
		\centering
		\scalebox{.98}{
		\begin{tabular}{| c | l@{\hskip3pt} | c@{\hskip5pt}  c@{\hskip5pt}  c@{\hskip5pt} | c@{\hskip5pt} c@{\hskip5pt} c@{\hskip5pt} | c@{\hskip5pt} c@{\hskip5pt} c@{\hskip5pt} | c | } 
			\hline
			Net& Norm.& Fr.C & Sa.C & Lj.C & Fr.N & Sa.N & Lj.N &Fr.S &Lj.S & Sa.S& avg.\\
			\hline
					
	\multirow{3}{*}{\rotatebox[origin=c]{90}{AlexNet}}&BN&\textbf{26.0}	&38.4	&\textbf{34.4}	&27.9	&26.6	&33.1	&28.8	&34.2	&25.1	&30.5\\
&WBN&25.8	&38.2	&33.0	&\textbf{29.4}	&26.6	&34.8	&30.3	&36.9	&25.1	&31.1\\
&WBN$ ^*$&25.9	&\textbf{40.3}	&33.4	&28.0	&\textbf{27.6}	&\textbf{34.9}	&\textbf{31.5}	&\textbf{44.3}	&\textbf{28.6}	&\textbf{32.7}\\
\hline
\multirow{3}{*}{\rotatebox[origin=c]{90}{ResNet}}&BN&\textbf{37.9}	&\textbf{40.9}	&39.3	&30.8	&48.3	&41.2	&30.6	&\textbf{40.6}	&27.6	&37.5\\
&WBN&37.3	&39.5	&\textbf{42.6}	&40.4	&51.8	&41.0	&33.8	&39.6	&\textbf{30.8}	&39.6\\
&WBN$ ^*$&36.6	&40.3	&40.0	&\textbf{41.2}	&\textbf{56.2}	&\textbf{45.2}	&\textbf{35.4}	&39.4	&25.6	&\textbf{40.0}\\
            \hline
		\end{tabular}
        }
		\label{tab:cold-dg-lig}
\end{table}

{In another series of experiments we consider the scenario where both illumination and laboratory change. We performed 27 different experiments, corresponding to the case where Saarbr\"ucken is considered as target domain. Figure \ref{fig:hist-81-cold} report the histogram of the gains in accuracy of our approach AlexNet+WBN* w.r.t. AlexNet+BN. As shown in Fig. \ref{fig:hist-81-cold}, in most of the cases our model leads to an increase in accuracy between 1-5\%. In only 5 out of 27 experiments, our model does not produce benefits.} 


\myparagraph{Comparison with SOTA on VPC.} In order to compare our model with the state-of-the art approaches in robotics, we  consider the VPC dataset. VPC has been used in previous works to test the DG abilities of different methods. Following the standard experimental protocol of \cite{wu2009visual}, we evaluate our model using 5 houses for training and 1 for test, averaging the results between the 6 configurations. For each house we report the average accuracy per class. 
Table \ref{tab:vpc-comparison} compares the result of our models with baseline deep architectures, with and without traditional BN layers. We consider both the case where domain information is available (WBN$ ^*$) and where it is not (WBN). Analogously to what observed in the experiments on COLD dataset, the accuracy increases when WBN is adopted, both in case of AlexNet and ResNet architectures. 
Interestingly, having domain priors during training produce a boost of performances for ResNet, while for AlexNet this is not the case. This suggests that different features have a different impact on our model. Features of the very last layers, as in AlexNet, may not be enough domain discriminative, especially in case of limited shift within the source domains. In those cases, a soft-assignment can provide a more effective strategy for clustering samples}.

\begin{table}[t]
			\caption{VPC dataset: average accuracy per class. } 
		\centering
		\scalebox{.98}{
		\begin{tabular}{| l | c c c c c c | c |} 
			\hline
			Net & H1 & H2& H3&H4 &H5 &H6& avg.\\
            \hline
            
			AlexNet & 49.8 & 53.4&	49.2&	64.4 &41.0	&43.4	& 50.2\\
AlexNet + BN& 54.5 &  \textbf{54.6} &  55.6 &  69.7   &  41.8 &  45.9 & 53.7\\
AlexNet + WBN &\textbf{54.7}& 51.9&	\textbf{61.8}&	\textbf{70.6}&	43.9& 46.5 &\textbf{54.9}\\
AlexNet + WBN$ ^*$ &53.5& \textbf{54.6} & 55.7 &	68.1&	\textbf{44.3}& \textbf{49.9} & 54.3 \\
\hline
ResNet & 55.8&  47.4 &  64.0 &  69.9 &  42.8 & 50.4 & 55.0\\
ResNet + WBN &55.7 &49.5 &\textbf{64.7 }&\textbf{70.2} &42.1 &\textbf{52.0} &55.7\\
ResNet + WBN$ ^*$& \textbf{56.8} & \textbf{50.9} & 64.1 &  69.3  &   \textbf{45.1} & 51.6&\textbf{56.5}\\
            \hline
            
		\end{tabular}
        }
		\label{tab:vpc-comparison}
\end{table}

Finally, Table \ref{tab:vpc-sota} compares the results obtained with WBN with those of state-of-the-art methods. Specifically we consider the method in \cite{wu2009visual}, where SIFT \cite{lowe2004distinctive} 
and CENTRIST (CE) features \cite{wu2011centrist} are provided as input to a nearest neighbor classifier, and the approach in \cite{fazl2012histogram}, where the same classifier is employed but using Histogram of Oriented Uniform Patterns (HOUP) as input. For sake of completeness, we also report the results obtained by exploiting also the temporal information between images. For this setting, we report the performances of the CENTRIST-based approach of \cite{wu2011centrist} coupled with Bayesian Filtering (BF) and the results of \cite{yang2012object} which used again a Bayesian Filter together with object templates. 
As shown in the Table, applying deep-learning techniques already guarantees an increase in performances of about $4\%$ with respect to the state of the art. Introducing WBN inside the network, allows a further accuracy gain. 

\begin{table}[t]
			\caption{VPC dataset: comparison with state of the art.} 
		\centering
		\scalebox{.95}{
		\begin{tabular}{| l | @{\hskip2pt}c@{\hskip3pt}  @{\hskip2pt}c@{\hskip3pt}  @{\hskip2pt}c@{\hskip3pt} | @{\hskip3pt}c@{\hskip3pt} | @{\hskip3pt}c@{\hskip3pt} | @{\hskip3pt}c@{\hskip2pt} @{\hskip2pt}c@{\hskip2pt} @{\hskip2pt}c@{\hskip3pt} | @{\hskip3pt}c@{\hskip2pt}  @{\hskip2pt}c@{\hskip3pt}|} 
			\hline
			Method & &\cite{wu2009visual}& & \cite{fazl2012histogram} & \cite{yang2012object} & \multicolumn{3}{@{\hskip3pt}c|@{\hskip3pt}}{AlexNet}&\multicolumn{2}{@{\hskip3pt}c@{\hskip3pt}|}{ResNet}\\ \hline
            Config. & SIFT & CE & BF & - & - & Base & BN &WBN$^*$& BN &WBN$^*$\\
            \hline
            Acc.& 35.0&41.9&45.6&45.9&50.0&50.2&53.7&54.3&55.0&\textbf{56.5}\\
            \hline
		\end{tabular}
        }
		\label{tab:vpc-sota}
\end{table}

\myparagraph{Experiments on a large-scale scenario: SPED.} {In this section we show the results obtained when WBN is applied to a large scale dataset of outdoor scenes, \ie the SPED dataset. In order to utilize SPED as a DG benchmark, we split the dataset in two sets, February and August, considering the months of data acquisition. Since no other automatic training data splits are possible using timestamps, in these experiments we do not use domain supervision and only consider WBN with two latent domains. The choice of having two domains is motivated by the fact that the dataset contains images collected at different times of the day and thus we assume that the latent domains automatically discovered by our method correspond to "night" and "day".} 

{Results are shown in Table \ref{tab:sped}. WBN provides a clear gain in all considered settings and for all considered architectures. The improvement of 4\% obtained in the case "August-to-February" for both networks is remarkable given the very large number of classes and the lack of domain supervision.}

\begin{table}[t!]
			\caption{SPED dataset: comparison of different models.} 
		\centering
		\scalebox{1.0}{
		\begin{tabular}{| l | c  c  c | c  c  c |} 
			\hline
			Net & \multicolumn{3}{c|}{AMOSNet}  & \multicolumn{3}{c|}{AlexNet}\\
            \hline
            Config. & Base & BN & WBN & Base & BN & WBN\\
            \hline
            February-to-August & 83.7 & 88.8 & \textbf{90.3} &83.6&88.9 & \textbf{90.5 }\\
            August-to-February & 71.2&82.7&\textbf{86.1}&73.9&83.1&\textbf{87.0}\\
            \hline
		\end{tabular}
        }
		\label{tab:sped}
\end{table}

\subsection[From BN to Classifiers: Best Sources Forward]{From BN to Classifiers: Best Sources Forward \footnotemark\footnotetext{M. Mancini, S. Rota Bul\`o, B. Caputo, E. Ricci. {\sl Best sources forward: domain generalization through source-specific nets}. IEEE International Conference on Image Processing (ICIP) 2018.}}
\label{sec:bsf}

\begin{figure}[t]
\centering
\includegraphics[width=0.8\columnwidth]{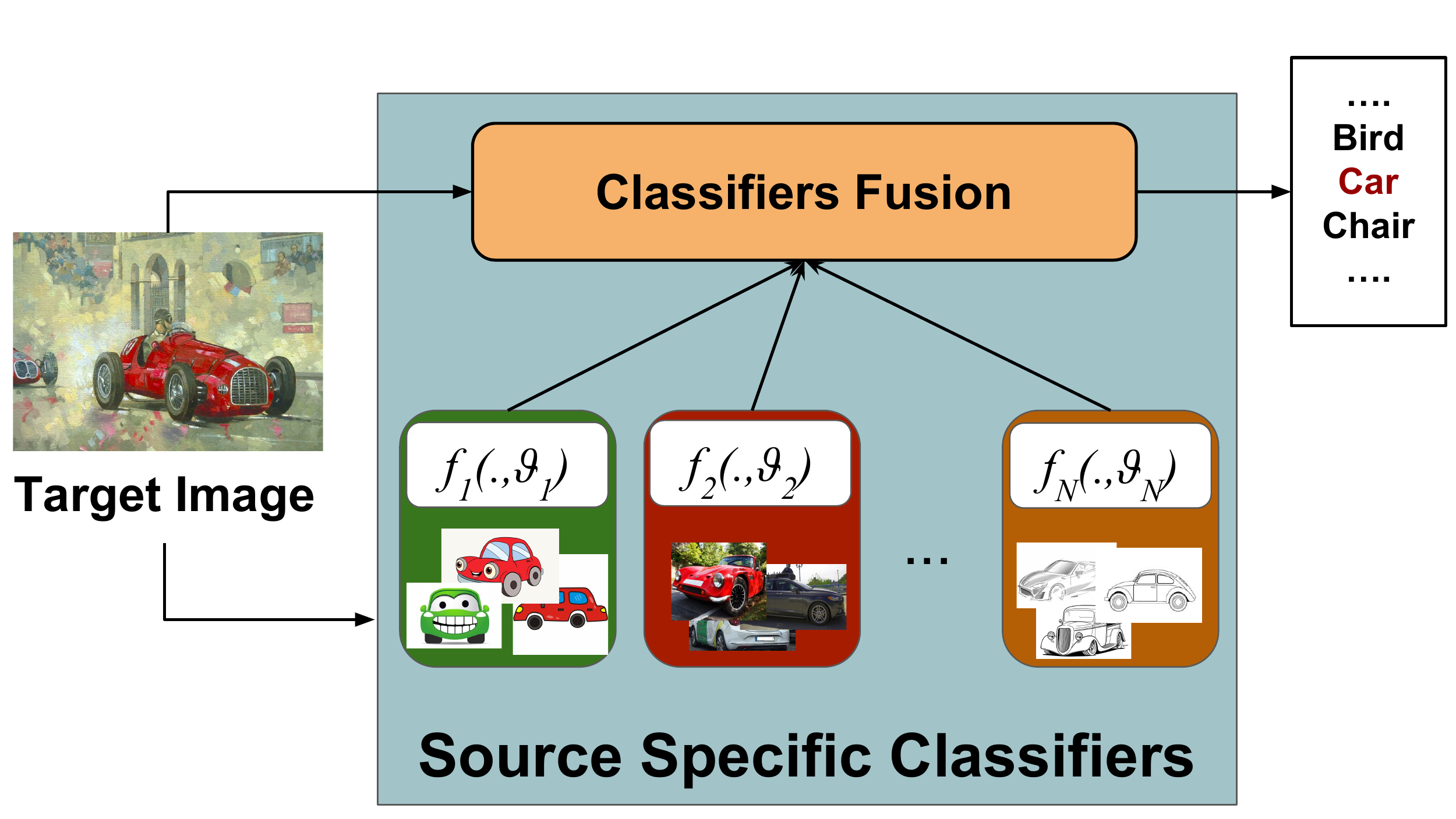}
    \caption{Intuition behind the proposed BSF framework. Different domain-specific classifiers and the classifiers fusion are learned at training time on source domains, \textit{in a single end-to-end trainable architecture}. When a target image is processed, our deep model optimally combines the source models in order to compute the final prediction.
} 
   \label{fig:bsf-idea}
\end{figure}

In Section \ref{sec:wbn}, we discussed how to address DG given a domain classification branch and domain-specific (either latent or explicit) normalization layers. However, the same approach can be applied, in principle, to other parts of the network. In this subsection, we describe how the same methodology can be applied to domain-specific classification layers (Fig. \ref{fig:bsf-idea}).

The approach devised in the previous section requires three components: (i) a way to estimate domain membership of a sample both at training and at test time, (ii) a distinction between domain-specific and domain-agnostic network elements,  and (iii) a strategy to merge domain-specific activations within the network. The first point can be easily addressed through a domain classifier, as described in Sections \ref{sec:wbn} and \ref{sec:latentDA-domain-prediction}. 

For what concerns the second point, we can write our classification model as $f_C^\Theta$, where $\Theta=\{\theta_j\}_{j=1}^{\con k_s}$ denotes the set of parameters to learn and each $\theta_j$ are the parameters corresponding to a specific domain $j$. Moreover, let us consider $\theta_j=\{\hat{\theta}_s,\hat{\theta}_j\}$, where $\hat{\theta}_s$ indicates the parameters shared by all domain-specific models and $\hat{\theta}_j$ the domain-specific ones. Under this formulation, in Section \ref{sec:wbn}, $\hat{\theta}_s$ were all the parameters of the network while $\hat{\theta}_j$ the domain-specific BN statistics. In this section, we change perspective and we assume $\hat{\theta}_s$ to be a feature extractor and $\hat{\theta}_j$ to be domain-specific semantic classification heads. Note that the formulation is general and can be applied to multiple/different levels of the network.

Now that we have defined the domain-specific component we must define how to merge activations of the domain-specific layers. During training, the most simple strategy would be to rely on the domain-label of the sample, namely:
\begin{equation}
\label{eq:bsf-classifier}
f_C^\Theta(x_i)= \sum_{j=1}^{\con k_s} \mathds{1}_{s_i=s_j} f_j(x_i)
 \end{equation}
 where we wrote $f_C^{\theta_j}$ as $f_j$ for simplicity. Similarly to Eq.~\eqref{eq:dg-hard2}, also this equation cannot be applied at test time, when the domain membership of a sample is unknown and falls out the space of the available source domains $\mathcal{D}^s$. Similarly to what we did in Section \ref{sec:wbn}, we can use a soft version of Eq.~\eqref{eq:bsf-classifier} at test time: 
 \begin{equation}
\label{eq:bsf-classifier-soft}
f_C^\Theta(x_i)= \sum_{j=1}^{\con k_s} w_{i,j} f_j(x_i,\theta_j)
 \end{equation}
 where $w_{i,j}=f_D^\Theta$ is the probability of sample $i$ to belong to domain $j$, as computed by our domain classifier. The model is trained with the same semantic and domain classification loss defined Eq.~\eqref{eq:dg-wbn-stdloss}.

\begin{figure}[t]
\centering
\includegraphics[width=0.8\columnwidth]{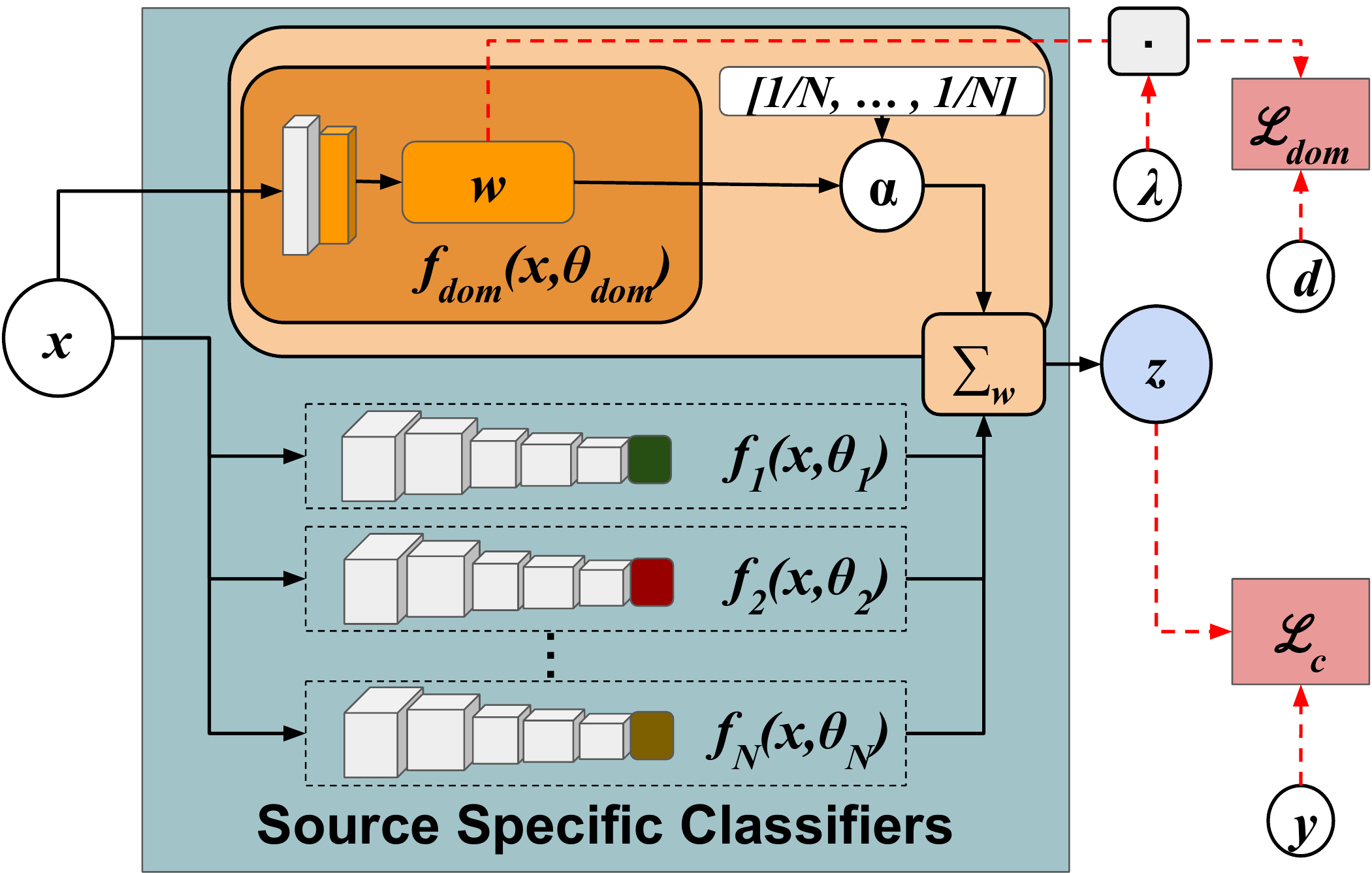}
    \caption{Simplified architecture of the proposed BSF framework. The input image is fed to a series of domain-specific classifiers and to the domain prediction branch. The latter produces the assignment $w$ which is fed to the domain prediction loss. The same $w$ is modulated by $\alpha$ before being used to combine the output of each classifier. The final output of the architecture $z$, is fed to the classification loss. 
} 
   \label{fig:bsf-method}
\end{figure}

 We highlight that, differently from Sections \ref{sec:da-latent} and \ref{sec:wbn}, here the merging of the domain-specific activations/components is held-out \textit{after} and not \textit{within} the feature extraction process. Moreover, we found a simple modification being beneficial in this scenario.
  In particular, we introduce an hyperparameter $0<\alpha<1$ and we re-write the classification function as follows: 
   \begin{equation}
 \label{eq:bsffinal-classifier}
f_C^\Theta(x_i)= (1-\alpha) \sum_{j=1}^{\con k_s} w_{i,j} f_{j}(x_i) + \frac{\alpha}{\con k_s}\sum_{j=1}^{\con k_s} f_{j}(x_i)\,.
 \end{equation}
 In practice, $\alpha$ allows to merge domain-specific component both exploiting the similarity among domains with its first term (as in Eq.~\eqref{eq:bsf-classifier}), while considering domain agreements on the predictions, weighting them equally with the second term. This allows the model to be robust to inaccurate domain assignment at test time while increasing the feedback to domain-specific models for source sets with few samples. In practice, during training we randomly switch 
with probability $\alpha$ between using the given domain label as $w_{ij}$ or assigning to all domain-classifier a uniform weight $1/\con k_s$. At test time, we use Eq.~\eqref{eq:bsffinal-classifier} with $w_{i,j}$ obtained from the domain prediction branch. As the experiments show, this choice allows us to obtain a more robust final classification model. Figure \ref{fig:bsf-method} provides an overview of our model.

\subsection{Experiments: Domain Generalization in Computer Vision}

\myparagraph{Datasets.}
We test the performance of BSF on two publicly available benchmarks. The first is \textbf{rotated-MNIST} \cite{ghifary2015domain}, a dataset composed by different domains originated applying different degrees of rotations to images of the original MNIST digits dataset \cite{lecun1998gradient}. We follow the experimental protocol of \cite{motiian2017unified}, randomly extracting 1000 images per class from the dataset and rotating them respectively of 0, 15, 30, 45, 60 and 75 degrees counterclockwise. As previous works, we consider one domain as target and the rest as sources. 

The second is \textbf{PACS} \cite{li2017deeper} the same database we used on the latent domain discovery section. Differently from Section \ref{sec:latentDA-experiments-datasets}, we consider a domain generalization setting (i.e. no target data available during training). Following the experimental protocol of \cite{li2017deeper}, we train our model considering three domains as source datasets and the remaining one as target. 

\myparagraph{Networks and training protocols.}
In our evaluation we set the parameters $\alpha=0.25$ and $\lambda=0.5$.
For the experiments on the rotated-MNIST dataset, we employ the LeNet architecture \cite{lecun1998gradient} following \cite{motiian2017unified}. The network is trained from scratch, using a batch size of 250 with an equal number of samples for each source domain.  We train the network for 10000 iterations, using Stochastic Gradient Descent (SGD) with an initial learning rate of 0.01, momentum 0.9 and weight decay 0.0005. The learning rate is decayed through an inverse schedule, following previous works \cite{ganin2014unsupervised}. For the domain prediction branch, we take as input the image and perform two convolutions, with the same parameters of the first two convolutional layers of the main network. Each convolution is followed by a ReLU non linearity and a pooling operation. 
The domain prediction branch follows the implementations of the previous sections. It terminates with a global average pooling followed by a fully connected layer which outputs the final weights. To ensure that $\sum_{j=1}^N w_{i,j}=1$, we apply the softmax operator after the fully connected layer. 
 
For PACS, we trained the standard AlexNet architecture, starting from the ImageNet pre-trained model. 
We use a batch size of 192, with 64 samples for each source domain. The initial learning rate is set to $5\cdot10^{-4}$ with a weight decay of $10^{-6}$ and a momentum of 0.9. We train the network for 3000 iterations, decaying the initial learning rate by a factor of 10 after 2500 iterations, using SGD. For the domain prediction branch, we use the features of \texttt{pool5} as input, performing a global average pooling followed by a fully-connected layer and a softmax operator which outputs the domain weights.

Our evaluation is performed using a NVIDIA GeForce 1070 GTX GPU, implementing all the models with the popular Caffe \cite{jia2014caffe} framework. For the baseline AlexNet architecture we take the pre-trained model available in Caffe.

\begin{figure}[t]
\centering
\includegraphics[width=0.8\columnwidth,trim=0.0cm 2cm 0cm 0cm]{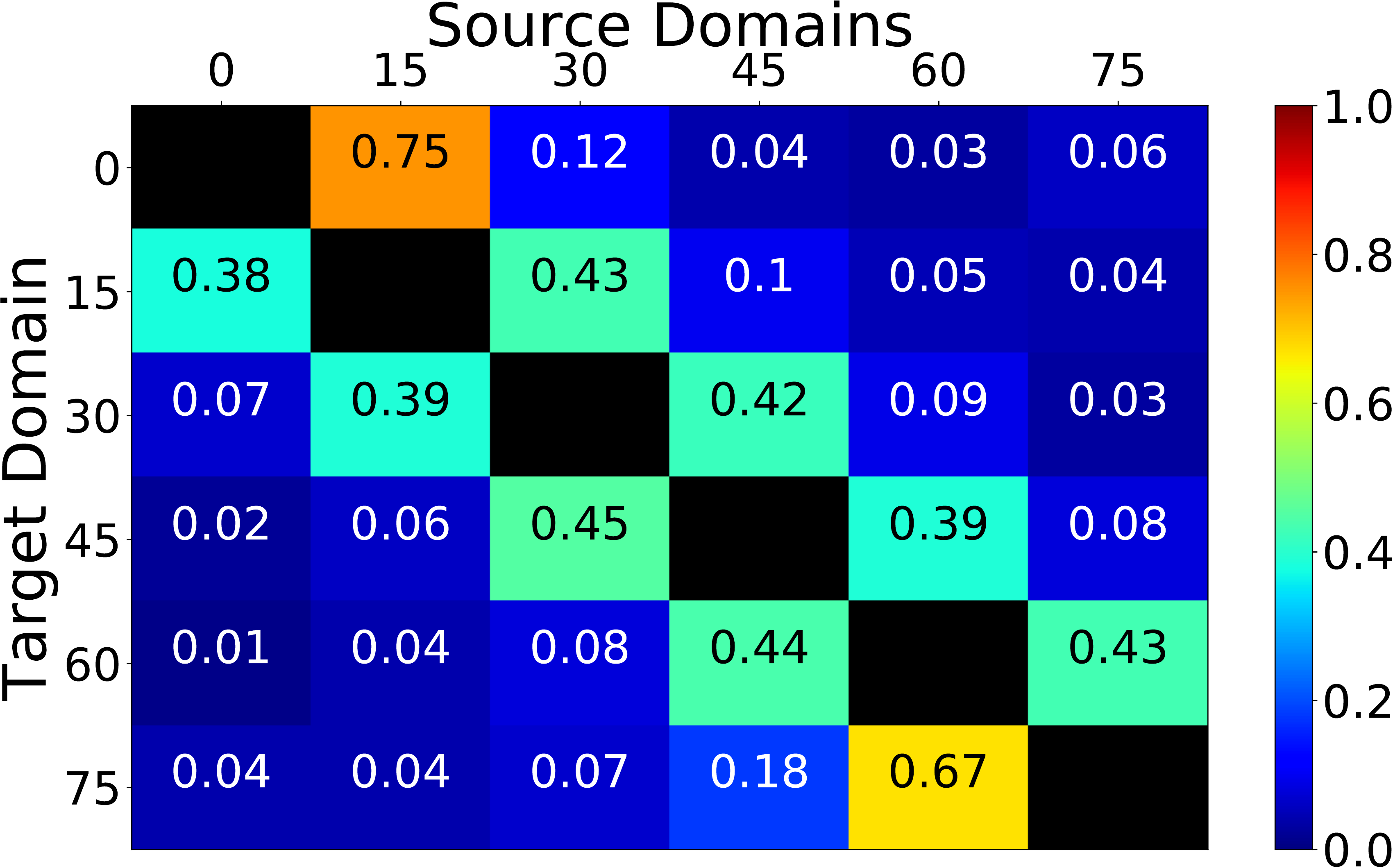}\vspace{20pt}
    \caption{Rotated-MNIST dataset: analysis of the assignments computed by the domain prediction branch.} 
   \label{fig:cm-mnist}
\end{figure}

\myparagraph{Results on Rotated-MNIST.} We first test the effectiveness of our model on the rotated-MNIST benchmark. 
We compare BSF with the CCSA method in \cite{motiian2017unified} and the multi-task autoencoders in \cite{ghifary2015domain} (MTAE) and \cite{rifaiexplicit} (CAE). The results from baseline methods are taken directly from \cite{motiian2017unified}.
\begin{table}[t]
			\caption{Rotated-MNIST dataset: comparison with previous methods. } 
		\centering
		\scalebox{.98}{
		\begin{tabular}{ l | c  c  c  c  c  c | c } 
			\hline
			Method & 0 & 15 & 30 & 45 & 60 & 75& Mean\\
            \hline
             CAE\cite{rifaiexplicit} & 72.1 & 95.3 & 92.6 & 81.5& 92.7&79.3&85.5 \\ 
             MTAE \cite{ghifary2015domain} & 82.5 & \textbf{96.3} & 93.4 & 78.6& 94.2&80.5&87.5 \\ 
             CCSA \cite{motiian2017unified} & 84.6 & 95.6 & 94.6 & 82.9& 94.8&82.1&89.1 \\ 
             BSF & \textbf{85.6}& 95.0 & \textbf{95.6}& \textbf{95.5}& \textbf{95.9}&\textbf{84.3}&\textbf{92.0} \\ \hline
		\end{tabular}
        }
		\label{tab:mnist}
\end{table}
As shown in Table \ref{tab:mnist}, BSF outperforms all the baselines. A remarkable gain in accuracy is achieved in the $45^o$ case. 
We ascribe this gain to the capability of our deep network to assign, for each target image, more importance to the source domains corresponding to the closest orientations, increasing the weights of the associated classifiers. Indeed, since $45^o$ is in the middle of the range between all possible orientations, it is likely that a stronger classifier can be constructed since we can exploit all the source models appropriately re-weighted. To further verify the effectiveness of our framework and its ability to properly combine source-specific models, we also compute for target samples with different orientations the number of assignments to each source domain. In this experiment one target sample $x_i$ is assigned to a source domain by computing the $\arg \max_j w_{i,j}$. The results are shown in Fig. \ref{fig:cm-mnist} (the number of assignments are normalized for each row). The figure clearly shows that the proposed domain prediction branch tends to associate a target sample to the source domains corresponding to the closest orientations. Consequently, our deep network classifies target samples constructing a model from the most related source classifiers. This results into more accurate predictions than previous domain-agnostic models due to the specialization of source classifiers on specific orientations. 

\myparagraph{Results on PACS.} We also perform experiments on the PACS dataset. We compare BSF with both previous approaches using precomputed features (in this case DECAF-6 features \cite{donahue2014decaf}) as input and end-to-end trainable deep models. For the baselines with pre-computed features, we report the results of MTAE \cite{ghifary2015domain}, low-rank exemplar SVMs (LRE-SVM) \cite{xu2014exploiting}, and uDICA \cite{muandet2013domain}), while for the end-to-end trainable deep models, we report the results of the domain agnostic model coupled with tensor factorization of \cite{li2017deeper} (TF-CNN) and the meta-learning approach MLDG \cite{li2017learning}. For a fair comparison the deep models \cite{li2017deeper,li2017learning} and our network are all based on the same architecture, \ie AlexNet.  Table \ref{tab:pacs} shows the results of our comparison. The performance of previous methods are taken directly from previous papers \cite{li2017deeper,li2017learning}. For our approach and \cite{li2017deeper} we also report results obtained without fine-tuning.
Our model outperforms all previous methods. 
These results are remarkable because, differently from the rotated-MNIST dataset, in PACS the domain shift is significant and it is not originated by simple image perturbations. Therefore, the association between a target sample and the given source domains is more subtle to capture. For sake of completeness we also report the performances obtained with the standard AlexNet network. These results shows that state of the art deep models have excellent generalization abilities, typically outperforming shallow models. However, designing deep networks specifically addressing the DG problem as we do leads to higher accuracy.

\begin{table}[t]
			\caption{PACS dataset: comparison with previous methods.} 
		\centering
		\scalebox{.98}{
		\begin{tabular}{ l | c c c c | c } 
			\hline
			Model & Art & Cartoon & Photo & Sketch & Mean\\\hline
             MTAE \cite{ghifary2015domain}& 60.3 & 58.7 & 91.1 & 47.9 & 64.5  \\ 
             LRE-SVM \cite{xu2014exploiting}& 59.7 & 52.8 & 85.5 & 37.9 & 59.0  \\  
               uDICA \cite{muandet2013domain}& 64.6 & 64.5 & 91.8 & 51.1 & 68.0  \\ 
               \hline
            TF-CNN\cite{li2017deeper} (no ft) & 62.7 & 52.7 & 88.8 & 52.2& 64.1 \\ 
       TF-CNN\cite{li2017deeper}& 62.9&\textbf{67.0}&89.5&57.5&69.2\\
            MLDG \cite{li2017learning}&\textbf{66.2}&66.9&88.0&59.0&70.0\\   
        BSF (no ft) & 64.1 & 60.6 & \textbf{90.4} & 49.4 & {66.1} \\
          BSF &64.1& 66.8 & 90.2&\textbf{60.1} & \textbf{70.3}  \\ \hline
         AlexNet  \cite{li2017deeper}   &63.3&63.1&87.7&54.1&67.1\\ \hline

		\end{tabular}
        }
		\label{tab:pacs}
\end{table}

We also perform a sensitivity analysis to study the impact of the parameter $\alpha$ on the performance and demonstrate the benefit of adding a domain-agnostic classifier. 
We consider the proposed approach without fine-tuning.
\begin{table}[t]
			\caption{PACS dataset: sensitivity analysis. 
            } 
		\centering
		\scalebox{.98}{
		\begin{tabular}{ l | c  c  c  c  } 
			\hline
			$\alpha$ & Art & Cartoon & Photo & Sketch \\\hline
             0 & \textbf{65.2} & 54.5 & \textbf{90.7} & \textbf{52.4} \\ 
 0.25 & 64.1 & 60.6 & 90.4 & 49.4 \\ 
 0.5 & 63.8 & \textbf{61.0} & 90.4 & 49.1 \\ 
             0.75 & 64.0 & 60.9 & 90.5 & 47.8\\
             1 & 63.0 & 60.1 & 90.5 & 47.5 \\ \hline
		\end{tabular}
        }
		\label{tab:ablation-alpha}
\end{table}
As shown in Table \ref{tab:ablation-alpha}, considering only the source-specific classifiers ($\alpha=0$) leads, on average, to the best performances, surpassing in the majority of the cases a domain agnostic classifier obtained by setting $\alpha=1$. This confirms our original intuition that addressing DG by fusing multiple source models is an effective strategy. However, there are few situations where using only source models can lead to a decrease in accuracy (\textit{e.g.} in the setting Cartoon) and incorporating a domain-agnostic component, even with reduced weight as $\alpha=0.25$, improves generalization accuracy.

\subsection{Conclusions}

In this section, we presented two deep learning models for addressing DG. The first, WBN, exploits a weighted formulation of BN to learn robust classifiers that can be applied to previously unseen target domains. We showed how this approach is effective in the context of semantic place categorization in robotics, achieving state-of-the-art performance on the VPC benchmark. The effectiveness of WBN is also confirmed by experiments on a large scale dataset of outdoor scenes.

The second, BSF, addresses the problem of DG by exploiting multiple domain-specific classifiers. In particular, it extends the principles of WBN, with a domain prediction branch choosing the optimal combination of source classifiers to use at test time, based on the similarity between the input image and the samples from the source domains. Differently from WBN, it goes beyond domain-specific BN layers but explores domain-specific classification modules. Moreover, a domain agnostic component is also introduced in our framework further improving the performance of the method. Experiments demonstrate the effectiveness of BSF which outperformed the state-of-the-art models on two benchmarks (at the time of submission).

With WBN and BSF, we have merged domain-specific models either at the BNs-level or at the classifiers one, due to the ease of linearly combining their parameters/statistic (WBN) and predictions (BSF). In future works, it would be interesting to blend domain-specific models at different levels of the networks, as explored in other works in contexts such as multi-task learning \cite{misra2016cross}, life-long learning \cite{aljundi2017expert} and motion control \cite{zhao2017blending1,zhang2018blending2}.

A drawback of both WBN and BSF is the assumption that multiple and diverse source domains are available at training time. This may be not always possible due to costly or even unfeasible data collection processes. Other recent approaches overcome this issue by considering external sources of knowledge such as automatically-generated training data \cite{mccormac2017scenenet} and online annotators \cite{song2015robot}. Generating synthetic data for the target task could be a huge advantage for training deep models, but requires the knowledge of the target task beforehand, something that is not assumed by our model. A possible solution to this issue consists of endowing the robot with the ability of access, on-demand, additional information about the target data.  Indeed, the generality of our framework allows the integration of external sources of knowledge (\eg generating multiple domains through web queries or synthetic data). Finally, a major drawback of DG models is the need for multiple labeled source domains during training. In the next sections, we will show how we can drop the assumption of having multiple source domains by extending the DA-layers models to the Continuous DA and Predictive DA scenarios.

\section[Continuous Domain Adaptation]{Continuous Domain Adaptation\footnotemark\footnotetext{M. Mancini, H. Karaoguz, E. Ricci, P. Jensfelt, B. Caputo. {\sl Kitting in the Wild through Online Domain Adaptation}.
       IEEE/RSJ International Conference on Intelligent Robots and Systems (IROS) 2018.}}
\label{sec:da-continuous}
Despite the remarkable performances achieved by DA algorithms in computer vision \cite{li2016revisiting,carlucci2017autodial}, and their growing popularity in robot vision \cite{angeletti2018adaptive} they require the presence of images from the target domain in advance during training. This is a huge limitation, especially in robotics, due to the likely unpredictable conditions of the environment in which a robot will be employed. In the previous section, we have seen how we can sidestep the need for target data during training in case we are given a set of multiple labeled source domains, addressing the DG scenario. However, this setting has also a limitation: the need of collecting (and labeling) data of multiple source domains. In this section, we want to overcome this issue, performing adaption given just a single source domain during training, without any target domain data. This setting, Continuous DA \cite{hoffman2014continuous}, requires to cope with the domain shift directly at test time, as the model processes data of the target domain.

Here, we consider a realistic application scenario for Continuous DA algorithms: the task of kitting. This task is the process of grouping related parts such as gathering components of a personal computer (PC) into one bin for assembly~\cite{banerjee2015ontology}. 
The kitting task requires the recognition of the parts in the environment, the ability to pick objects from the bins, and placing them at the correct location \cite{holz2015skill}. All of these subtasks are very challenging on their own but the recognition of the parts is crucial for the robot to sequentially perform the other subtasks. Already in today's factory settings, object recognition tasks possess challenges such as environmental effects (illumination, viewpoint, etc), varying object material properties, and cluttered scenes \cite{liu2012fast}. In order to simplify the recognition task, some approaches use machine vision in rather isolated settings for decreasing the environmental variability \cite{schyja2012modular}. Liu et. al \cite{liu2012fast} proposed a specially designed camera system and estimation based on 3D CAD models to robustly detect and verify the type and the pose of the object. Kaiba et. al. \cite{kaipa2015resolving} proposed an interactive method where a remote human operator resolves ambiguities in the perception system. Unfortunately, none of the above methods are generic enough to be applied in a truly unconstrained setting. In this section, we are primarily concerned with solving the object recognition problem for kitting using vision in the wild, i.e. in non-isolated settings exhibiting large variations. Right now, most of the robots in the manufacturing industry are operating in isolation, primarily because of safety concerns. However, many future scenarios have robots and humans working closer together, moving robots into new areas of applications, beyond mass production and preprogrammed behavior. For this to happen, not only safety but perception will be a major challenge. 

In this section, we describe two main contributions. The first is a kitting dataset that contains images of objects taken under varying illumination, viewpoint, and background conditions from a robotic platform. This dataset provides the community with a novel tool for studying the robustness of robot vision algorithms to drastic changes in the appearance of the input images and assess progress in the field. We are not aware of existing, publicly available kitting databases covering this range of visual variability. 

Second, we describe an approach for achieving \textit{online} adaptation of a deep model \cite{mancini2018kitting}. Differently from classical DA approaches, this algorithm can adapt a deep model to any target domain on the fly, without requiring any target domain data before-hand. We benchmark the performance of our algorithm on the proposed dataset, showing how this model is able to produce large improvements on the target domain performances compared to the base architecture trained on the source domain, and matching what would be achieved by having all data from the target available beforehand.

\begin{figure}[t]
\centering
\includegraphics[width=1\columnwidth]{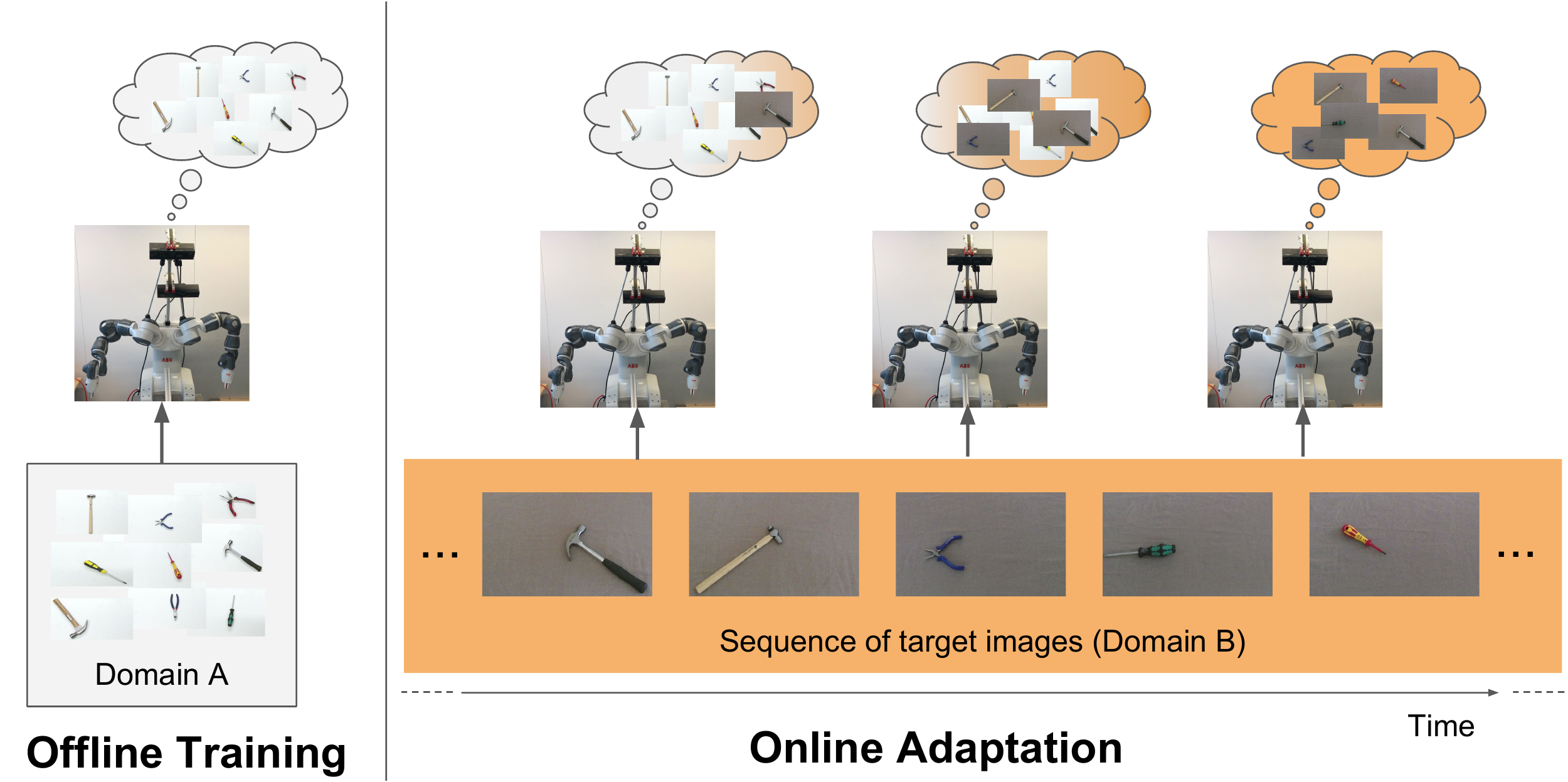}
    \caption{Our ONDA approach for performing kitting in arbitrary conditions. Given a training set, we can train a robot vision model offline. As the robot performs the task, we gradually adapt the visual model to the current working conditions, in an online fashion and without requiring target data during the offline training phase. } 
    \label{fig:cDA-teaser}
\end{figure}

\subsection{The KTH Handtool Dataset}

\begin{figure}[htb!]
      \centering
      \includegraphics[width=0.9\columnwidth]{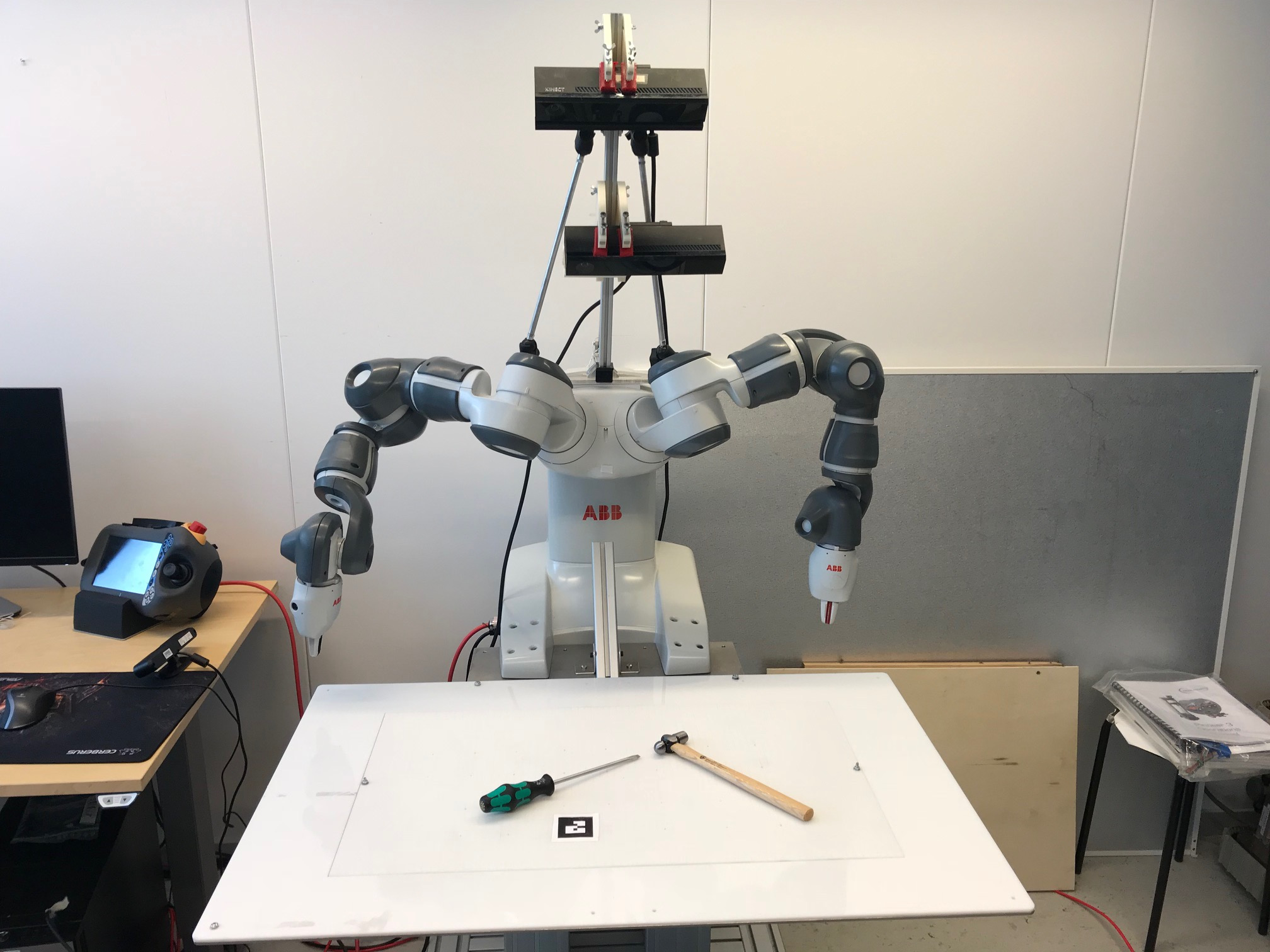}
      \caption{The 2-arm stationary robot platform. }
      \label{fig:cDA-robot}
 \end{figure}

The KTH Handtool Dataset\footnote{https://www.nada.kth.se/cas/data/handtool/} is collected for evaluating the object recognition/detection performance of robot vision methods in varying viewpoint, illumination and background settings, all crucial abilities for robot kitting in unconstrained, real-world settings. Instead of having general household objects, the dataset consists of hand tools in order to represent a workshop setting in a factory. It consists of 9 different hand tools for 3 different categories; hammer, plier and screwdriver. The images are collected with a 2-arm stationary robot platform shown in Fig. \ref{fig:cDA-robot}. Dataset consists of 3 different illuminations, 2 different cameras (One Kinect camera and one webcam) with different viewpoints and 2 different background settings that correspond to 12 (3x3x2) domains in total. For each hand tool, approximately 40 images with different poses are collected for each camera and domain setting. Table~\ref{tab:cDA-dataset} shows example images from different domains. In total, approximately 4500 RGB images are available in the dataset.

\begin{table*}[ht]
\begin{center}
 \caption{Example Images from KTH Handtool Dataset} \label{tab:cDA-dataset}
  \begin{tabularx}{1.0\textwidth}{XXXX} 
  \hline  \multirow{2}{*}{Camera Type} & \multicolumn{3}{c}{Illumination} \\ \cline{2-4}
        &  \hspace{6mm} Artificial & \hspace{6mm} Cloudy & \hspace{6mm} Directed  \\ 
      \hline   \vspace{5mm} Kinect & \vspace{0.5mm} \parbox[c]{1em}{\includegraphics[width=1.2in]{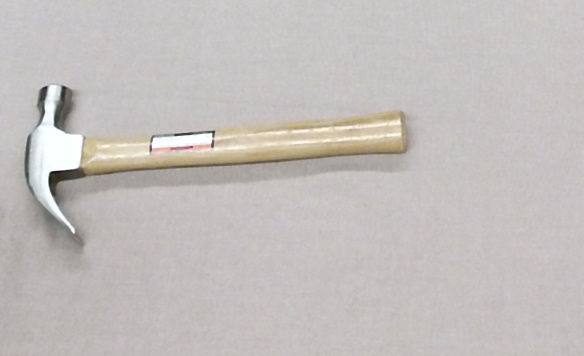}} & 
      \vspace{0.5mm} \parbox[c]{1em}{\includegraphics[width=1.2in]{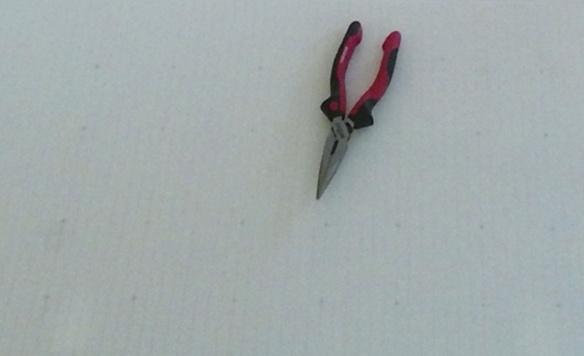}} & 
      \vspace{0.5mm} \parbox[c]{1em}{\includegraphics[width=1.2in]{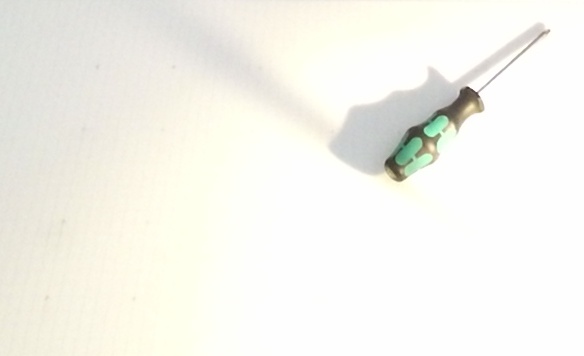}} \\
       \vspace{5mm} Webcam & \vspace{0.5mm} \parbox[c]{1em}{\includegraphics[width=1.2in]{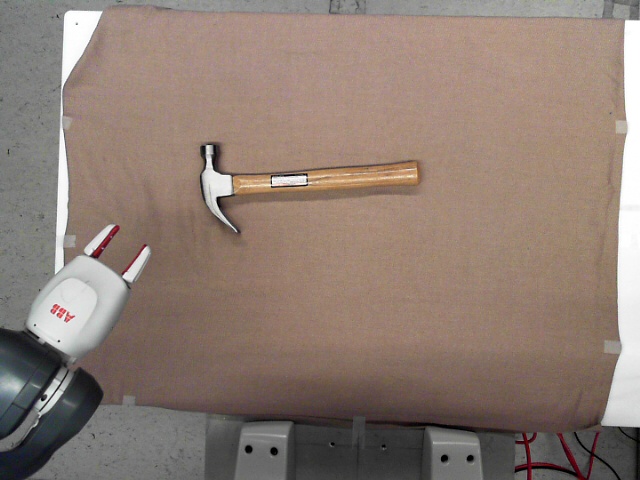}}  & \vspace{0.5mm} \parbox[c]{1em}{\includegraphics[width=1.2in]{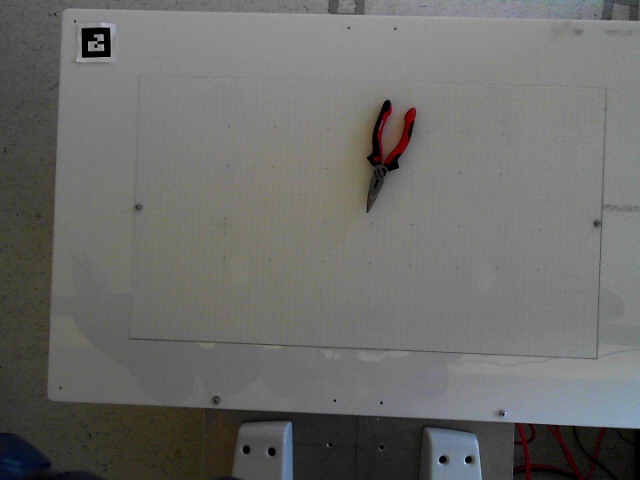}} & \vspace{0.5mm} \parbox[c]{1em}{\includegraphics[width=1.2in]{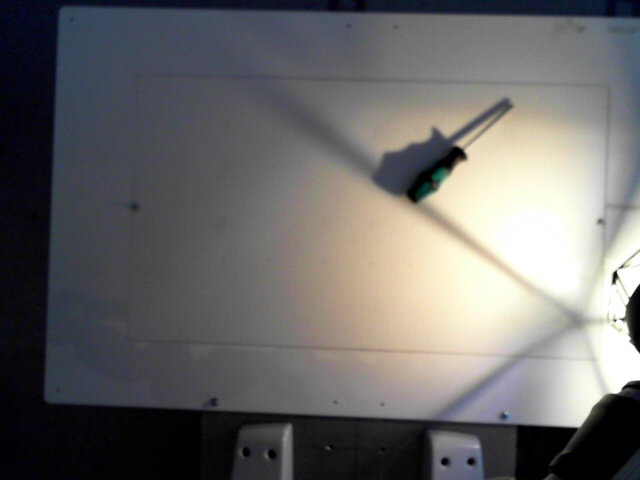}} \vspace{2mm}\\  
       \hline
  \end{tabularx}
\end{center}
\end{table*}

\subsection{Problem Formulation}
\label{sec:cDA-problem}
Suppose we collected a set of images using a robotic platform with the aim of training a robot vision model with it. Since the image collection has been acquired in the real world, the resulting model will be biased towards the particular conditions (\eg illumination, environmental) under which the images have been acquired. Because of this, if we employ such a system and the current working conditions are different from those of the training set, the performances of the model will degrade due to the presence of a substantial \textit{shift} between the training and test data. In this situation, to increase the generalization capabilities of the robot we can remove the acquisition bias either by collecting more training data in a large variety of conditions, which is extremely expensive, or by developing algorithms  able to bridge the gap between the training and test data, aligning the original model to the novel scenario. The latter is the goal of domain adaptation. Formally, we assume to have a source domain $\mathcal{S}=\{x^s_i,y^s_i\}_{i=1}^{\con n}$, where $x^s_i$ is an image and $y^s_i \in \{1,\cdots,C\}$ the associated semantic label. Opposite to traditional domain adaptation in a batch setting, during training we have only access to the source domain $\mathcal{S}$ and \textit{we do not have any data or prior information about the target domain $\mathcal{T}$}, apart from the set of semantic labels which is assumed to be shared. When the robot is active, the current working conditions will compose the target domain and we will have access to the automatically acquired sequence of images  $\mathcal{T}=\{x^t_1,\cdots,x^t_{T}\}$. In this scenario, in order to adapt the network parameters $\theta$ to this novel domain, we must exploit the incoming test images collected by the robot on the fly.

\subsection{ONDA: ONline Domain Adaptation with Batch-Normalization}
\begin{figure}[t]
\centering
\includegraphics[width=1\columnwidth]{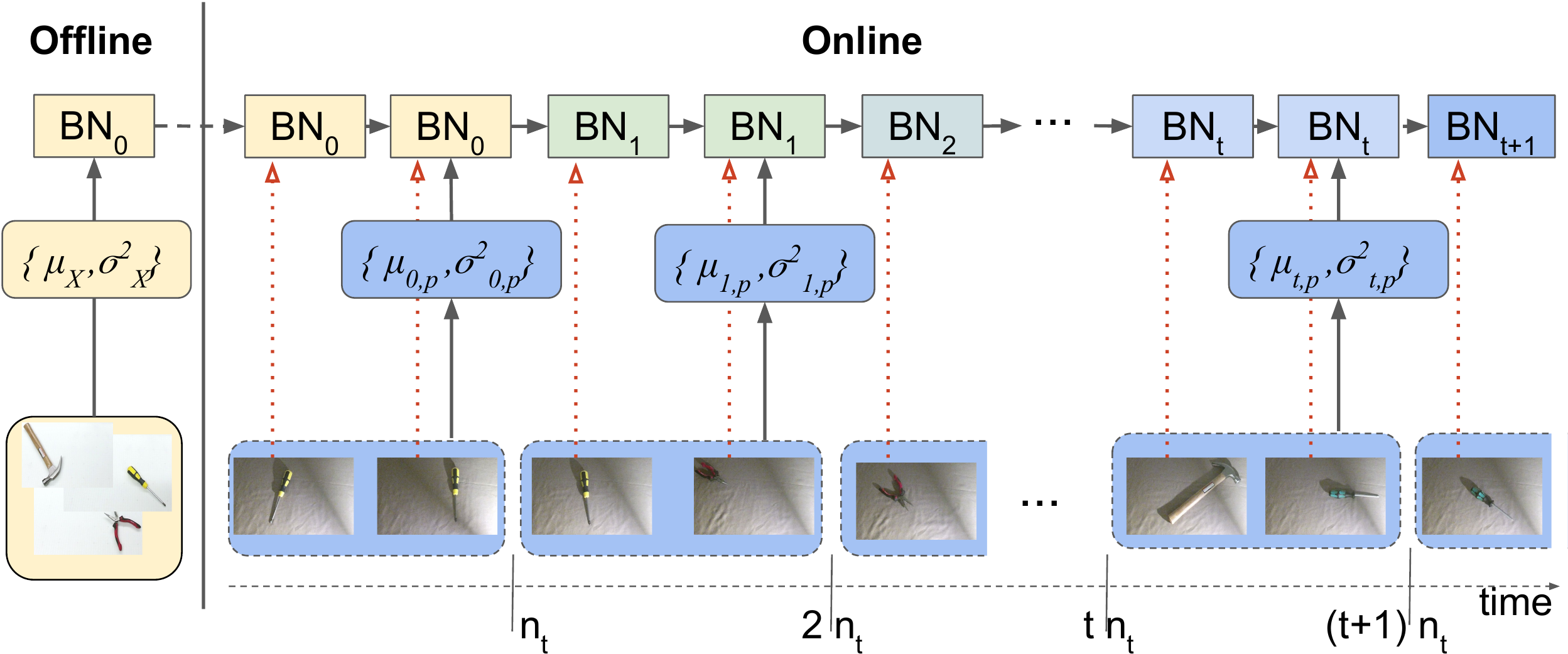}
    \caption{The statistics of the BN layers are initialized offline, by training the network on the images of the source domain. At deployment time, the input frames are processed using the global estimate of the statistics (red lines). However the robot collects each $n_t$ input frames to compute partial BN statistics, using these estimated values to gradually update the BN statistics for the current scenario. } 
   \label{fig:cDA-method}
\end{figure}
\label{sec:cDA-ONDA}

Starting from the idea of Domain Alignment Layers (Section \ref{sec:da-preliminaries}), we can follow the same principle of obtaining a target-specific model but considering an online setting. In particular, instead of having a fixed target set available during training, we propose to exploit the stream of data acquired while the robot is acting in the environment and continuously update the BN statistics. In this way, we can gradually adapt the deep network to a novel scenario. 

 Specifically, we start by training the network on the source domain $\mathcal{S}$, initializing the BN statistics at time $t=0$ as $\{\mu_0,\sigma^2_0\}=\{\mu_\mathcal{S},\sigma^2_\mathcal{S}\}$. 
Assuming that the set of network parameters $\theta$ are shared between the source and target domain except for the BN statistics, we can adapt the network classifier $f_\theta$ by updating the BN statistics with the estimates computed from the sequence $\mathcal{T}$. Defining as $n_t$ the number of target images to use for updating online the BN statistics, we can compute a partial estimate $\{\hat{\mu}_t,\hat{\sigma}^2_t\}$ of the BN statistics as: 
\begin{align*}
\hat{\mu_{t}} = \frac{1}{n_t}\sum_{i=1}^{n_t} x^t_i \ \ \ \ \hat{\sigma_{t}}^2 =\frac{1}{n_t}\sum_{i=1}^{n_t} 
(x^t_i-\hat{\mu}_{t})^2 
\end{align*} 
where $x^t_i$ is the distribution of activations for a given feature channel and domain $t$, following the notation in Section \ref{sec:da-preliminaries}. 
The global statistics at time $t$ can be updated as follows:
\begin{align*}
\sigma_{t}^2&=(1-\alpha)\sigma_{t-1}^2+\alpha \frac{n_t}{n_t-1}\hat{\sigma}_{t}^2\\
\mu_{t} &= (1-\alpha) \mu_{t-1} +\alpha\hat{\mu}_{t}
\end{align*}  
where $\alpha$ is the hyperparameter regulating the decay of the moving average.

The above formulation achieves a similar adaptation effect of DAL layers \cite{li2016revisiting,carlucci2017just,carlucci2017autodial} but with three main advantages. First, no samples of the target domain, neither labeled nor unlabeled, are used during training. Thus, no further data acquisition and annotation efforts are required. Second, since we do not exploit target data for training, contrary to standard DA algorithms, we have no bias towards a particular target domain. 
Third, since the adaptation process is online, the model can adapt itself to multiple sequential changes of the working conditions, being able to tackle unexpected environmental variations (\eg sudden illumination changes).

The reader might wonder if other possible choices may be considered for initializing $\{\mu_0,\sigma^2_0\}$, such as exploiting a first calibration phase where the robot collects images of the target domain in order to produce a first estimate of the BN statistics. Here we choose to use the statistics estimated on the source domain because 1) we want a model ready to be employed, without requiring any additional preparation at test time; 2) the robot may occur in multiple domains during employment and if a shift occurs (\eg illumination condition changes) our method will automatically adapt the visual model to the novel domain starting from the current estimated statistics: initializing $\{\mu_0,\sigma^2_0\}=\{\mu_\mathcal{S},\sigma^2_\mathcal{S}\}$ allows to check the performance of the algorithm even for multiple sequential shifts and long-term applications. Obviously our method can benefit from a calibration phase of the statistics closer to the target working conditions: we plan to analyze these aspects in the future.

\subsection{Experimental results}
\myparagraph{Networks and training protocols.} We perform our experiments with the AlexNet \cite{krizhevsky2012imagenet} architecture pre-trained on ImageNet \cite{deng2009imagenet}. We train 3 additional models: a variant of AlexNet with BN, the DA architecture DIAL from \cite{carlucci2017just} and our ONline DA model (ONDA). Following \cite{carlucci2017just}, we add BN layers or its variants after each fully-connected layer. Both the standard AlexNet, AlexNet with BN and DIAL are trained with a batch size of 128. We implemented \cite{carlucci2017just} by splitting the batch size between images of source and target domain proportionally to the number of images for each set, as in \cite{carlucci2017just}, without employing the entropy-loss for target images \cite{carlucci2017just,carlucci2017autodial}. We highlight that DIAL is our upper-bound in this case, since it shares the same philosophy of ONDA but with the assumption that images of the target domain are available at training time.

As preprocessing, we rescale all the images in order to ensure a shortest side of 256 pixels, preserving the aspect ratio and subtracting the mean value per channel computed over the ImageNet database. As input to the network we use a random crop of 227$\times$227 at training time, employing a central crop with the same dimensions during test. No additional data augmentation is performed. For all the variants of the architecture, we fine-tune the last layers for 30 epochs with an initial learning rate of 0.001 for \texttt{fc6}, \texttt{fc7} and of 0.01 for the classifier, with a weight decay of 0.0005 and momentum 0.9. We scale the initial learning rates by a factor of 0.1 after 25 epochs. 

In order to apply ONDA, we start from the weights of AlexNet with BN, training on the given source domain. Then, we perform one iteration over the target domain, without updating any parameter other than the BN statistics. As a trade-off between stability of the statistics and fastness of adaptation we set $n_t=10$ and $\alpha=0.1$. We will detail the impact of these choices in the following subsections.

In all the experiments, we consider the task of object recognition in the \textit{fine-grained} setting, with all the 9 classes considered as classification objective. We report the average accuracy between 5 runs, shuffling the order of the input images in each run of our model.

\subsubsection{Domain Adaptation results}
\begin{figure*}[t]
  \centering
  \subfloat[Source Domain: Artificial light, Kinect camera and White background]{\includegraphics[width=1\textwidth]{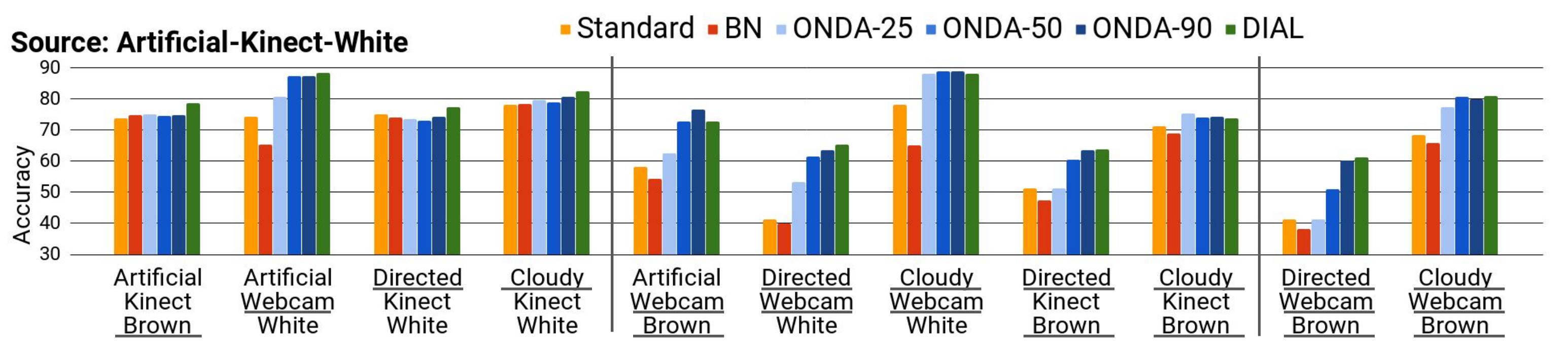}\label{fig:awk-hist}}
    \hfill
  \subfloat[Source Domain: Cloudy light, Webcam camera and Brown background]{\includegraphics[width=1\textwidth]{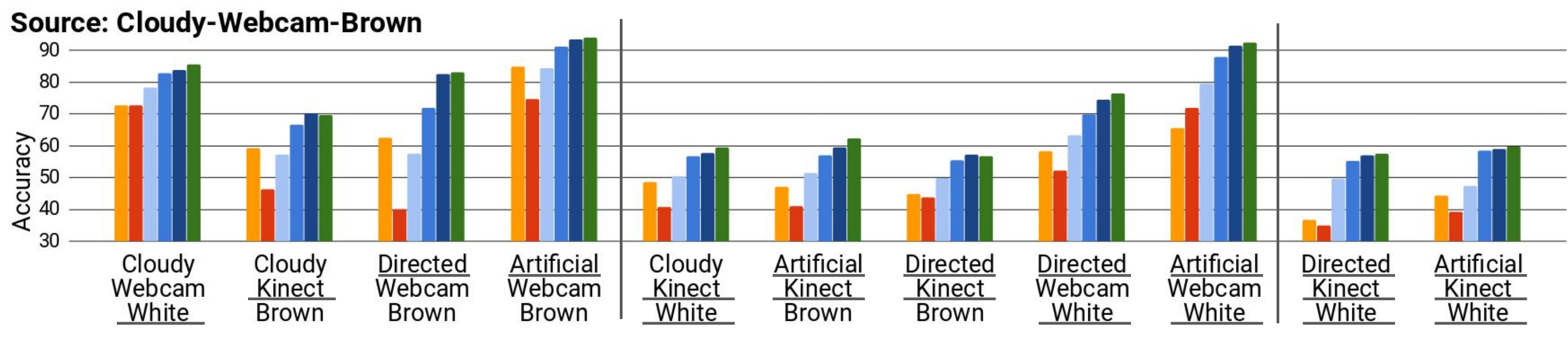}\label{fig:cwb-hist}}
  \caption{Experiments on isolated shifts. The labels of the x-axes denote the conditions of target domain, with the first line indicating the light condition, the second the camera and the third the background. We underlined the changes between the source and target domains. 
  }
  \label{fig:isolated-shifts}
\end{figure*}

In this subsection, we will present the results of our algorithm. In order to analyze the particular effect that each possible change may have to the adaptation capabilities of our model, we isolate the sources of shift. To this extent, we consider two sample starting source domains: in the first case (Figure \ref{fig:awk-hist}), the acquisition conditions are artificial light, Kinect camera and white background; in the second case we consider cloudy illumination, webcam and brown background (Figure \ref{fig:cwb-hist}). From these source domains we start by changing only one of the acquisition conditions (left part of the figures) and gradually increasing the number of changes to 2 and 3 conditions (middle and right parts respectively). We report the results for our model after 25\%, 50\% and 90\% of the target data processed. 

As the figures show, 
our model is able to fill the gap between the BN baseline (red bars) and the DA upper bound DIAL (green bars) in almost all settings. 
Only in few cases, where the shift between the performances of BN and DIAL is lower, this does not happen (\ie Figure \ref{fig:awk-hist}, target artificial-Kinect-brown and directed-Kinect-White). In all the other settings the gains are remarkable: considering both figures, the average difference between the performance of BN and ONDA-90 are of 
15\%, 18\% and 20\% for the single, double and triple shift cases respectively. We stress that the gain increases with the amount of shift between the source and target domains, underlying the importance of applying DA adaptation methods in changing environments. 
As expected, the statistics computed in the first stages (\ie ONDA-25) are not always sufficiently representative of the true estimate since they may be still biased by the statistics computed over the source domain. However the estimate becomes more precise as more images of the target domain are processed (\ie ONDA-50 and ONDA-90), gradually covering the gap with the estimate computed by DIAL. The fastness of adaptation and the quality of the estimates depend on the two hyperparameters $\alpha$ and $n_t$. In the next subsection we will analyze their impact to the final performances of the algorithm.

\myparagraph{Ablation study.} In this subsection we analyze the impact of the two hyperparameters, the update frequency $n_t$ and the decay $\alpha$, on the number of images needed by ONDA to estimate the statistics for the target domain. We use a sample scenario of Figure \ref{fig:cwb-hist}, where cloudy illumination, webcam camera and brown background are the source domain conditions and artificial light, Kinect camera and white background are the target domain ones. In the first experiment, we fix $n_t$ to 10, varying the value of $\alpha$. We start by a single pre-trained model of AlexNet with BN repeating the experiments for 5 runs, shuffling the order of the input data, and reporting the average accuracy for each update step.

\begin{figure*}[t]
  \centering
  \subfloat[Ablation on $\alpha$ with $n_t=10$]{\includegraphics[width=0.49\textwidth]{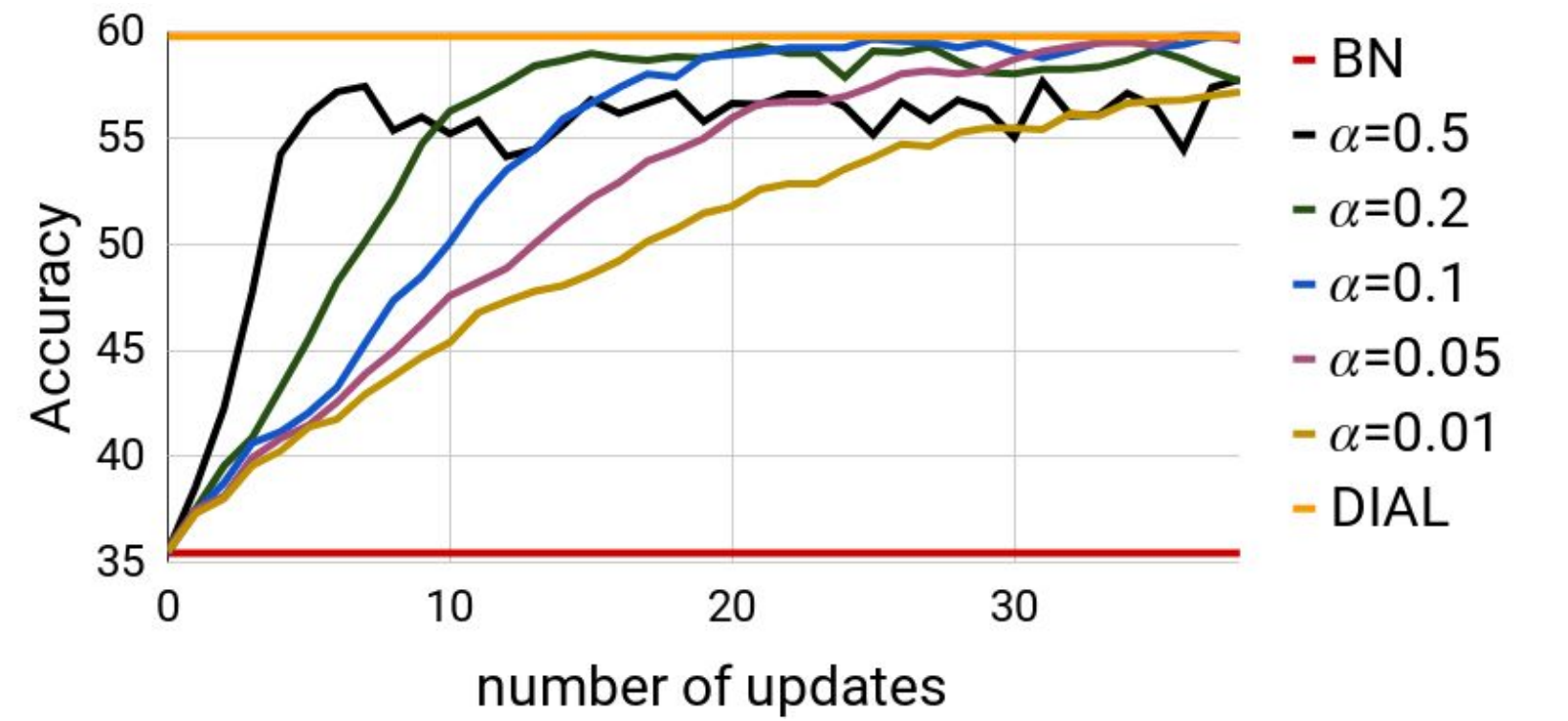}\label{fig:cDA-graph-alpha}}
    \hfill
  \subfloat[Ablation on $n_t$ with $\alpha=0.1$]{\includegraphics[width=0.49\textwidth]{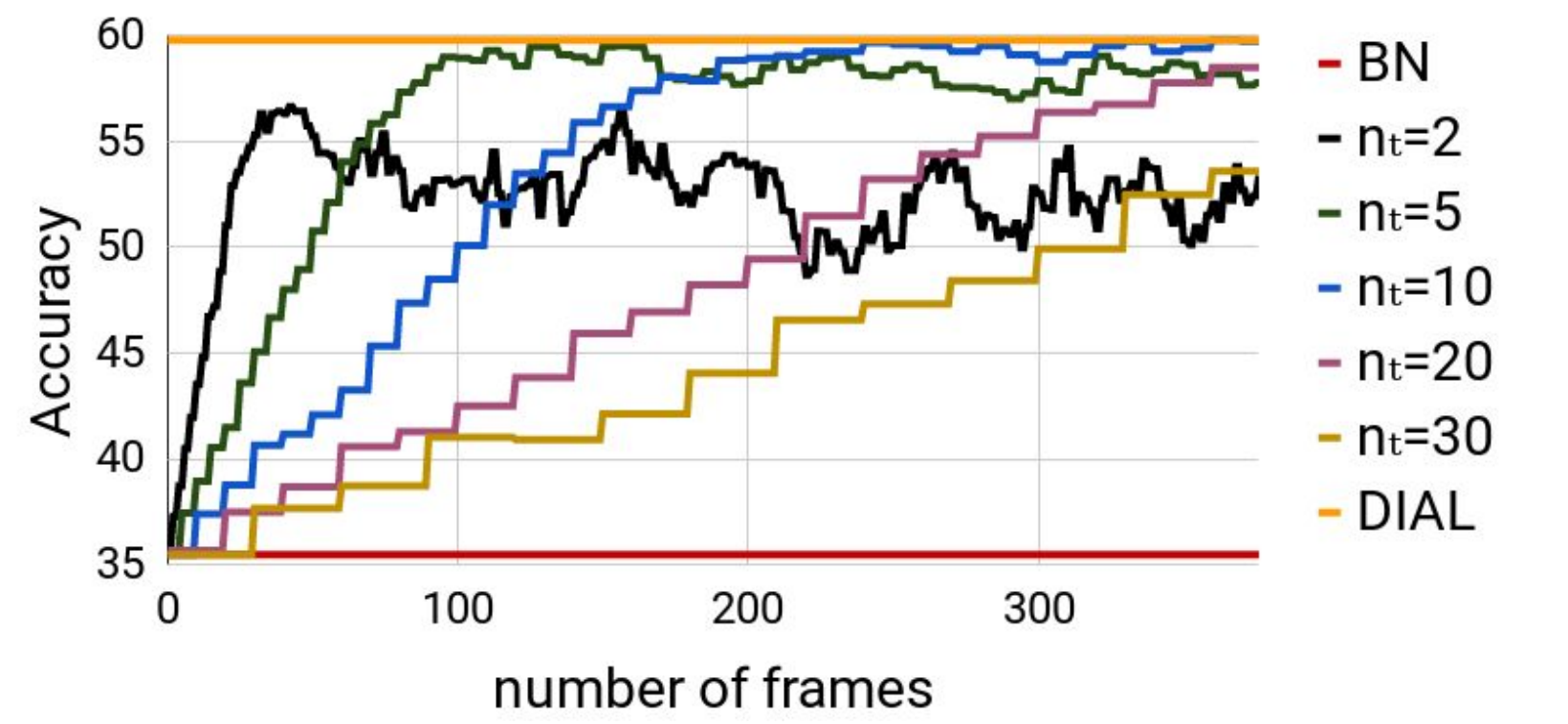}\label{fig:cDA-graph-nt}}
  \caption{Accuracy vs number of updates of ONDA for different values of (a) $\alpha$ and (b) $n_t$ in a sample scenario. The red line denotes the BN lower bound of the starting model, while the yellow line the DIAL upper bound.}
  \label{fig:cDA-ablation-study}
\end{figure*}

Results are shown in Figure \ref{fig:cDA-graph-alpha}: increasing the value of $\alpha$ to 0.2 (green line) or 0.5 (black line) allows the model to achieve a faster adaptation to the target conditions, with the drawback of a noisier estimation of the statistics. Thus, increasing $\alpha$ leads to an unstable convergence of the performance. On the other hand, choosing too low values of $\alpha$ (\eg 0.05 or 0.01, purple and gold lines respectively) allows a more stable convergence of the model, but with the drawback of slower adaptation to the novel conditions. 

Regarding the hyperparameter $n_t$, we follow the same protocol of the first experiment, fixing $\alpha$ to 0.1 and varying the number of images collected before updating the statistics, $n_t$, reporting how the accuracy changes with respect to the number of frames processed. As Figure \ref{fig:cDA-graph-nt} shows, low values of $n_t$ (\eg $n_t=2$) allows a faster adaptation, due to the higher update frequency, but at the price of a noisier estimation of the statistics, which is harmful to the final accuracy achieved by the model. At the same time, high values of $n_t$ (\eg 20, 30) allow for a more precise estimate of the statistics, highlighted by the smoothness of the respective lines in the graph, with the drawback of a lower speed of adaptation to the novel domain, caused by the lower update frequency.

The speed of adaptation and the final quality of the BN statistics is obviously a consequence of the values chosen for both hyperparameters. Obviously $\alpha$ and $n_t$ are not independent from each other: for a lower $n_t$ a lower $\alpha$ should be selected in order to preserve the final performance of the algorithm and conversely for a higher $n_t$, a higher $\alpha$ will allow a faster adaptation of the model. As a trade-off between fast adaptation and good results, we found experimentally that choosing $n_t=\{5,10,20\}$ and $\alpha=\{0.05,0.1\}$ worked well for both short and long term experiments. 

\subsection{Conclusions}
In this section, we presented a novel dataset for addressing the kitting task in robotics. The dataset takes into account multiple variations of acquisition conditions such as camera, illumination, and background changes which may occur during the robot employment. This dataset is intended for testing the robustness of robot vision algorithms to changing environments, providing a novel benchmark for assessing the robustness of robot vision systems.

Additionally, we described ONDA, an algorithm capable of performing online adaptation of deep models to any unseen visual domain. The algorithm, based on the update of the statistics of batch-normalization layers, can continuously adapt the model to the current environmental conditions of the robot, providing more robustness to unexpected working conditions. Experiments on the newly proposed dataset, confirm the ability of ONDA to fill the gap between a standard architecture, trained only on source data, and its domain adapted counterpart, without requiring any additional target data during training.

It is worth highlighting how, despite its effectiveness and the fact of requiring a single source domain (differently from the DG approaches in Section \ref{sec:da-dg}), the method has two main drawbacks. Since it adapts to the stream of the target samples, its adaptation is gradual and it cannot work under abrupt changes of the input distribution. As a consequence, it can only address one target domain shift at the time, contrary to DG approaches, which build a single model for multiple target domains. In the next section, we will show how we can merge the benefits of DG and Continuous DA, proposing the first deep model for the task of Predictive DA. 

Finally, as future works, we plan to enlarge the dataset, including more sources of variations and more objects. We further plan to provide a deeper analysis of our algorithm with more architectures, as well as exploring possible extensions that could exploit knowledge coming from previously met scenarios.

\section[Predictive Domain Adaptation]{Predictive Domain Adaptation \footnotemark\footnotetext{M. Mancini, S. Rota Bul\`o, B. Caputo, E. Ricci. {\sl AdaGraph: Unifying Predictive and Continuous Domain Adaptation through Graphs}. IEEE/CVF International Conference on Computer Vision and Pattern Recognition (CVPR) 2019.}}
\label{sec:da-predictive}
An underline common thread linking the sections of this chapter is the importance of being able to overcome the domain shift problem even under incomplete (Section \ref{sec:da-latent}) or absent (Sections \ref{sec:da-dg} and \ref{sec:da-continuous}) information about our target domain during training.  In particular, although it might be reasonable for some applications to have target samples available during training, in most cases data collection and labeling might be too costly (e.g. robotics) or even unfeasible (e.g. hazardous environments). Therefore, we argued that it is important to build models able to perform domain adaptation even without target data at training time. 

For this reason,  in Sections \ref{sec:da-dg} and \ref{sec:da-continuous} we focused on scenarios that do not assume the presence of target data during training, namely DG and Continuous DA.  In both scenarios, different information is used to overcome the domain shift. In the first, DG, the presence of multiple labeled source domains allows us to build models disentangling domain-specific and semantic-specific information, possibly generalizing to unseen input distributions. In the second, Continuous DA, target data received at test time are used to update the model gradually. Both scenarios have some inherent drawbacks. In DG, we require the presence of multiple labeled source domain, something that might be hard to obtain. In Continuous DA, instead, the model gradually adapts to the target distribution and, consequently, (i) it cannot work under abrupt changes of domains and (ii) it can address only one target domain shift at the time.

In this section, we want to take a step forward by (i) dropping the assumption of having multiple labeled source domains (opposite to DG) and (ii) adding the possibility to rapidly adapt the model to multiple target domains (opposite to Continuous DA).  
Following this idea, previous studies proposed the Predictive Domain Adaptation (PDA) scenario \cite{yang2016multivariate}, where neither the data nor the labels from the target are available during training. Only annotated source samples are available, together with additional information from a set of auxiliary domains, in form of unlabeled samples and associated metadata (\textit{e.g.} corresponding to the image timestamp or camera pose, etc).

\begin{figure}[t]
\begin{center}
   \includegraphics[width=1.0\columnwidth]{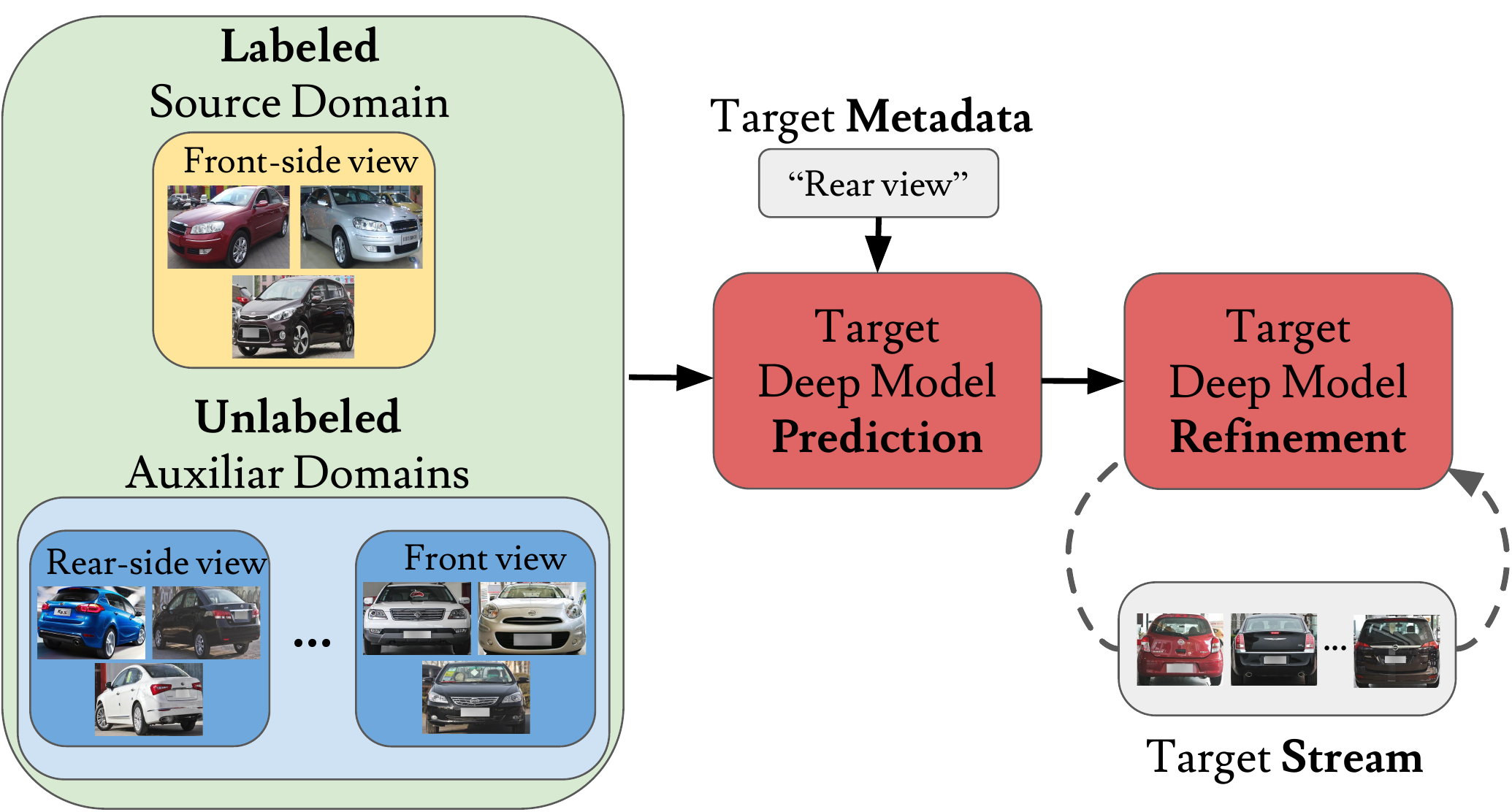}
\end{center}
   \caption{Predictive Domain Adaptation. During training we have access to a labeled source domain (yellow block) and a set of unlabeled auxiliary domains (blue blocks), all with associated metadata. At test time, given the metadata corresponding to the unknown target domain, we predict the parameters associated to the target model. 
   This predicted model is further refined during test, while continuously receiving data of the target domain.}
\label{fig:PDA-teaser}
\end{figure}

In this section we describe AdaGraph \cite{mancini2019adagraph}, a deep architecture for PDA.
As for the works presented in previous sections,  we learn a set of domain-specific models by considering a common backbone network with domain-specific alignment layers embedded into it \cite{carlucci2017autodial,carlucci2017just,li2016revisiting}. However, differently from the previous works, we propose to exploit metadata and auxiliary samples by
 building a graph which explicitly describes the dependencies among domains. 
Within the graph, nodes represent domains, while edges encode relations between domains, imposed by their metadata. 
Thanks to this construction,  when metadata for the target domain are available at test time, the domain-specific model can be recovered. We further exploit target data directly at test time by devising an approach for continuously updating the deep network parameters once target samples are made available {(Figure \ref{fig:PDA-teaser})}.  We demonstrate the effectiveness of our method with experiments on three datasets: the \textit{Comprehensive Cars} (CompCars) \cite{yang2015large}, the \textit{Century of Portraits} \cite{ginosar2015century}  and the \textit{CarEvolution} datasets \cite{RematasICCVWS13}, showing that our method outperforms state-of-the-art PDA approaches. Finally, we show that the proposed approach for continuous updating of the network parameters can be used for continuous domain adaptation, producing more accurate predictions than previous methods \cite{hoffman2014continuous,li2018domain}.

To summarize, the contributions presented in this section are: (i) the first deep architecture for addressing the problem of PDA; (ii) a strategy for injecting metadata information within a deep network architecture by encoding the relation between different domains through a graph; (iii) a simple strategy for refining the predicted target model which exploits the incoming stream of target data directly at test time.


\subsection{Problem Formulation} 
\label{sec:PDA-problem-formulation}
Our goal is to produce a model that is able to accomplish a task in a \textit{target} domain $\mathcal{T}$ for which \textit{no data} is available during training, neither labeled nor unlabeled. The only information we can exploit is a characterization of the content of the target domain in the form of \textit{metadata} $m_t$ plus a set of known domains $\mathcal{K}$, each of them having associated metadata. 
All the domains in $\mathcal{K}$ carry information about the task we want to accomplish in the target domain. In particular, since in this work we focus on classification tasks, we assume that images from the domains in $\mathcal{K} $ and $\mathcal{T}$ can be classified with semantic labels from a same set $\mathcal{Y}$. 
As opposed to standard DA scenarios, the target domain $\mathcal T$ does not necessarily belong to the set of known domains $\mathcal{K}$. Also, we assume that $\mathcal{K}$ can be partitioned into a \textit{labeled} \textit{source} domain $\mathcal{S}$ and $N$ \textit{unlabeled} \textit{auxiliary} domains $\mathcal{A}=\{A_1,\cdots,A_N\}$. 

In the specific, in this section we focus on the predictive DA (PDA) problem, aimed at regressing the target model parameters using data from the domains in $\mathcal K$. We achieve this objective by (i) interconnecting each domain in $\mathcal{K}$ using the given domain metadata; (ii) building domain-specific models from the data available in each domain in $\mathcal{K}$;  (iii) exploiting the connection between the target domain and the domains in $\mathcal{K}$, inferred from the respective metadata, to regress the model for $\mathcal{T}$.

A schematic representation of the method is shown in Figure 2. We propose to use a graph because of its seamless ability to encode relationships within a set of elements (domains in our case). Moreover, it  can be easily manipulated to include novel elements (such as the target domain $\mathcal{T}$).

\subsection{AdaGraph: Graph-based Predictive DA}
\label{sec:PDA-graph-da}

We model the dependencies between the various domains by instantiating a graph composed of nodes and edges. Each node represents a different domain and each edge measures the relatedness of two domains.  Each edge of the graph is weighted, and the strength of the connection is computed as a function of the domain-specific metadata. At the same time, in order to extract one model for each available domain, we employ recent advances in domain adaptation involving the use of domain-specific batch-normalization layers \cite{li2018adaptive,carlucci2017just}.  With the domain-specific models and the graph we are able to predict the parameters for a novel domain that lacks data by simply (i) instantiating a new node in the graph and (ii) propagating the parameters from nearby nodes, exploiting the graph connections.

\myparagraph{Connecting domains through a graph.}
Let us denote the space of domains as $\mathcal{D}$ and the space of metadata as $\mathcal{M}$. As stated in Section \ref{sec:PDA-problem-formulation}, in the PDA scenario, we have a set of known domains $\mathcal{K}=\{k_1,\cdots,k_n\}\subset\mathcal{D}$ and a bijective mapping $\phi:\mathcal{D}\mapsto\mathcal{M}$ relating domains and metadata. 
For simplicity, we regard as \textit{unknown} a metadata $m$ that is not associated to domains in $\set K$, \ie such that $\phi^{-1}(m)\notin \set K$.

Here we structure the domains as a \textit{graph} $\mathcal{G}=(\mathcal{V},\mathcal{E})$, where $\set V\subset\set D$ represents the set of vertices corresponding to domains and $\set E\subseteq\set V\times\set V$ the set of edges, \ie relations between domains. Initially the graph contains only the known domains so $\set V=\set K$. In addition, we define an edge weight $\omega:\set E\to\mathbb R$ that measures the relation strength between two domains $(v_1,v_2)\in \set E$ by computing a distance between the respective metadata, \ie
\begin{equation}
    \label{eq:PDA-edges}
    \omega(v_1,v_2)=e^{-d(\phi(v_1),\phi(v_2))}\,,
\end{equation}
where $d:\set M^2\to\mathbb R$ is a distance function on $\set M$.


Let $\Theta$ be the space of possible model parameters and assume we have properly exploited the domain data from each domain in $k\in\mathcal{K}$ to learn a set of domain-specific models (we will detail this procedure in the next subsection). We can then define a mapping $\psi: \mathcal{K}\mapsto\Theta$, relating each domain to its set of domain-specific parameters. 
Given some metadata $m\in\set M$ we can recover an associated set of parameters via the mapping $\psi\circ\phi^{-1}(m)$ provided that $\phi^{-1}(m)\in\set K$.
In order to deal with metadata that is unknown, we introduce the concept of \textit{virtual} node. Basically, a virtual node $\tilde{v} \in \mathcal{V}$ is a domain for which no data is available but we have metadata $\tilde m$ associated to it, namely $\tilde{m}=\phi(\tilde v)$. For simplicity, let us directly consider the target domain $\mathcal{T}$. We have $\mathcal{T}\in \mathcal{D}$ and we know $\phi(\mathcal{T})=m_t$. Since no data of $\mathcal{T}$ is available, we have no parameters that can be directly assigned to the domain. However, we can estimate parameters for $\set T$ by using the domain graph $\mathcal{G}$. Indeed, we can relate $\set T$ to other domains $v\in\set V$ using $\omega(\mathcal{T},v)$ defined in \eqref{eq:PDA-edges} by opportunely extending $\set E$ with new edges $(\set T,v)$ for all or some $v\in\set V$ (\eg we could connect all $v$ that satisfy $\omega(\mathcal{T},v)>\tau$ for some $\tau$). The extended graph $\set G'=(\set V\cup\{\set T\},\set E')$ with the additional node $\set T$ and the new edge set $\set E'$ can then be exploited to estimate parameters for $\set T$ 
by \textit{propagating} the model parameters from nearby domains. Formally we regress the parameters $\hat{\theta}_{\mathcal{T}}$ through the formula
\begin{equation}
    \label{eq:PDA-regression-metadata}
    \hat{\theta}_{\mathcal{T}}=\psi(\mathcal{T})=\frac{\sum_{(\set T,v)\in \set E'}\omega(\mathcal{T},v)\psi(v)}
    {\sum_{(\set T,v)\in \set E'}\omega(\mathcal{T},v)}\,,
\end{equation}
where we normalize the contribution of each edge by the sum of the weights of the edges connecting node $\mathcal{T}$. 
With this formula we are able to provide model parameters for the target domain $\mathcal{T}$ and, in general, for any unknown domain by just exploiting the corresponding metadata.

We want to highlight that this strategy only depends extending the graph with a virtual node $\tilde v$ and computing the relative edges. While the relations of $\tilde{v}$ with other domains can be inferred from given metadata, as in \eqref{eq:PDA-edges}, there could be cases in which no metadata is available for the target domain. 
In such situations, we can still exploit the incoming target image $x$ to build a probability distribution over nodes in $\mathcal{V}$, in order to assign the new data point to a mixture of known domains. To this end, let use define $p(v|x)$ the conditional probability of an image $x\in\mathcal{X}$, where $\mathcal{X}$ is the image space, to be associated with a domain $v\in\mathcal{V}$. From this probability distribution, we can infer the parameters of a classification model for $x$ through:
\begin{equation}
    \label{eq:PDA-regression-image}
    \hat{\theta}_x=\sum_{v\in \mathcal{V}}p(v|x)\cdot\psi(v)
\end{equation}
where $\psi(v)$ is well defined for each node linked to a known domain, while it must be estimated with \eqref{eq:PDA-regression-metadata} for each virtual domain $\tilde{v}\in\mathcal{V}$ for which $p(\tilde{v}|x)>0$.

In practice, the probability $p(v|x)$ is constructed from a metadata classifier $\mu$, trained on the available data, that provides a probability distribution over $\set M$ given $x$, which can be turned into a probability over $\set D$ through the inverse mapping $\phi^{-1}$.

 \begin{figure*}[t]
\begin{center}
   \includegraphics[width=1.0\linewidth]{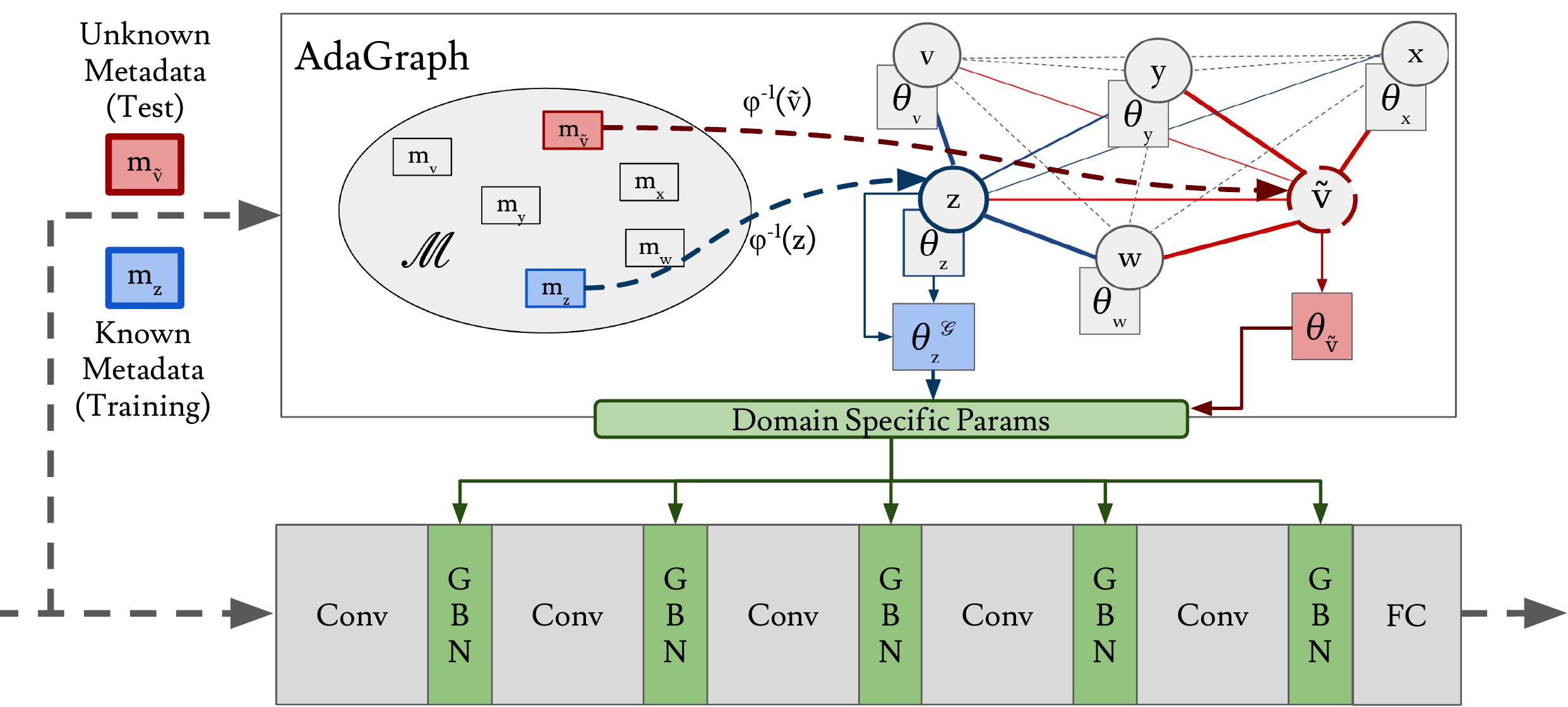}
\end{center}
   \caption{AdaGraph framework (Best viewed in color). Each BN layer is replaced by its GBN counterpart. The parameters used in a GBN layer are computed from a given metadata and the graph. 
   Each domain in the graph (circles) contains its specific parameters (rectangular blocks). During the training phase (blue part), a metadata (\ie $m_z$, blue block) is mapped to its domain (z). While the statistics of GBN are determined only by the one of $z$ ($\theta_z$), scale and bias are computed considering also the graph edges. During test, we receive the metadata for the target domain ($m_{\tilde{v}}$, red block) to which no node is linked. Thus we initialize $\tilde{v}$ and we compute its parameters and statistics exploiting the connection with the other nodes in the graph ($\theta_{\tilde{v}}$).}
\label{fig:PDA-method}
\end{figure*}

\myparagraph{Extracting node specific models.} 
We have described how to regress model parameters for an unknown domain by exploiting the domain graph.
Now, we focus on the actual problem of training domain-specific models using data available from the known domains $\set K$. 
Since $\mathcal{K}$ entails a labeled source domain $\mathcal{S}$ and a set of auxiliary domains $\mathcal{A}$,  we cannot simply train
independent models with data from each available domain due to the lack of supervision on domains in $\mathcal{A}$ for the target classification task. 
For this reason, we need to estimate the model parameters for the unlabeled domains $\mathcal{A}$ by exploiting DA techniques.

To achieve this, we start from the domain alignment layers presented in \cite{li2018adaptive,carlucci2017autodial,carlucci2017just} and described in Section \ref{sec:da-preliminaries}. 
In this scenario, the set of parameters for a domain $k$, $\psi(k)=\theta_k$ is composed of different parts. Formally for each domain we have $\psi(k)=\{\theta^{a},\theta^{s}_k\}$, where $\theta^a$ holds the domain-agnostic components and $\theta^s_k$ the domain-specific ones. In our case $\theta^a$ comprises parameters from standard layers (\ie the convolutional and fully connected layers of the architecture), while $\theta^s_k$ comprises parameters and statistics of the domain-specific BN layers. 
 
We start by using the labeled source domain $\mathcal{S}$ to estimate $\theta^a$ and initialize $\theta^s_\mathcal{S}$. In particular, we obtain $\theta_{\mathcal{S}}$ 
by minimizing the standard cross-entropy loss:
\begin{equation}
    \label{eq:PDA-loss-source}
    L(\theta_\mathcal{S}) = -\frac{1}{|\mathcal{S}|} \sum_{(x,y)\in\mathcal{S}} \log (f_{\theta_\mathcal{S}}(y;x)) \,,
\end{equation}
where $f_{\theta_{\mathcal{S}}}$ is the classification model for the source domain, with parameters $\theta_\mathcal{S}$. 

To extract the domain-specific parameters $\theta^s_k$ for each $k \in \mathcal{K}$, we employ 2 steps: the first is a selective forward pass for estimating the domain-specific statistics while the second is the application of a loss to further refine the scale and bias parameters. Formally, we replace each BN layer in the network with a \textit{GraphBN} counterpart (GBN), where the forward pass is defined as follows:
\begin{equation}
  \label{eq:PDA-graph-forward-std}
  \GBN(x,v)=\gamma_v \cdot\frac{x-\mu_v}{\sqrt{\sigma_v^2+\epsilon}}\; +\; \beta_v\,.
\end{equation}
where $\gamma_v$ and $\beta_v$ are the node specific scale and bias parameters of the BN layers. Basically in a GBN layer, the set of BN parameters and statistics to apply is conditioned on the node/domain to which $x$ belongs.  While this equation is similar to Eq.~\eqref{eq:da-domainalignment}, we highlight that, differently from it and \cite{carlucci2017just,carlucci2017autodial}, here we use domain-specific scale and bias parameters, not only statistics. During training, as for standard BN, we update the statistics by leveraging their estimate obtained from the current batch $\mathcal{B}$:
\begin{equation}
  \label{eq:PDA-stats-forward-std}
  \hat{\mu_v}=\frac{1}{|\mathcal{B}_v|} \sum_{x\in \mathcal{B}_v} x\;\;\;\text{and}\;\;\;
  \hat{\sigma^2_v}=\frac{1}{|\mathcal{B}_v|} \sum_{x\in \mathcal{B}_v} (x-\mu_v)^2\,,
\end{equation}
where $\mathcal{B}_v$ is the set of elements in the batch belonging to domain $v$. As for the scale and bias parameters, we optimize them by means of a loss on the model output. For the auxiliary domains, since the data are unlabeled, we use an entropy loss, while a cross-entropy loss is used for the source domain: 
\begin{align}
    \label{eq:PDA-loss-combined}
    L(\Theta^s) =-\frac{1}{|\mathcal{S}|} & \sum_{(x,y)\in\mathcal{S}} \log (f_{\theta_{\mathcal{S}}}(y;x))\\
   -\lambda\cdot\sum_{A_i\in\mathcal{A}}\frac{1}{|A_i|}  &\sum_{x\in A_i} \sum_{y\in \mathcal{Y}} f_{\theta_{A_i}}(y;x)\log \big(f_{\theta_{A_i}}(y;x)\big) 
\end{align}
where $\Theta^s=\{\theta^s_k \,|\, k \in \mathcal{K}\}$ represents the whole set of domain-specific parameters and $\lambda$ the trade off between the cross-entropy and the entropy loss.

While \eqref{eq:PDA-loss-combined} allows to optimize the domain-specific scale and bias parameters, it does not take into account the presence of the relationship between the domains, as imposed by the graph. A way to include the graph within the optimization procedure is to modify \eqref{eq:PDA-graph-forward-std} as follows:
\begin{equation}
  \label{eq:PDA-graph-forward-collaborative}
  \GBN(x,v,\mathcal{G})={\gamma}_v^\mathcal{G} \cdot\frac{x-\mu_v}{\sqrt{\sigma_v^2+\epsilon}} + {\beta}_v^\mathcal{G}
\end{equation}
with:
\begin{equation}
    \label{eq:PDA-training-combination-sb}
    \nu_v^\mathcal{G}=\frac{\sum_{k\in \mathcal{K}}\omega(v,k)\cdot \nu_{k}}{\sum_{k\in \mathcal{K}}\omega(v,k)}\,,
\end{equation}
 for $\nu\in\{\beta,\gamma\}$. Basically we use a scale and bias parameters during the forward pass which are influenced by the graph edges, as described in \eqref{eq:PDA-training-combination-sb}.
 
 Taking into account the presence of $\mathcal{G}$ during the forward pass is beneficial for mainly two reasons. First, it allows to keep a consistency between how those parameters are computed at test time and how they are used at training time. Second, it allows to regularize the optimization of $\gamma_v$ and $\beta_v$, which may be beneficial in cases where a domain contains few data. While the same procedure may be applied also for ${\mu_v,\sigma_v}$, in our current design we avoid mixing them during training. This choice is linked to the fact that each image belongs to a single domain and keeping the statistics separate allows us to estimate them more precisely.
 
 At test time, once we have initialized the domain-specific parameters of $\mathcal{T}$ using either \eqref{eq:PDA-regression-metadata} or \eqref{eq:PDA-regression-image}, the forward pass of each GBN layer is computed through \eqref{eq:PDA-graph-forward-collaborative}. In Figure \ref{fig:PDA-method}, we sketch the behaviour of AdaGraph both at training and test time.

\subsection{Model Refinement through Joint Prediction and Adaptation}
\label{sec:PDA-continuous-da}
While the approach described in the previous section allows to perform a blind adaptation of a model to a target domain given metadata, it is not completely true that we have no information about the images of the target domain. In fact, while at training time we have no access to target data, at test time target samples are gradually made available. 
While we could passively classify the target data stream, this would not be an effective choice, since the information coming directly from target images is valuable and can be leveraged to refine our model. This will be extremely important, \eg in the case of an inaccurate estimate of the target model parameters or in the presence of noisy metadata. In those cases, exploiting the stream of incoming images would compensate for the initial error.

To this end, we equip our model with a simple yet effective strategy to perform continuous domain adaptation. Following recent works \cite{li2018adaptive} and our ONDA framework, we start with the observation that a simple way to continuously adapt a model to the incoming stream of target data, is just by updating the BN statistics. Formally, let us suppose our target domain is composed by a set of $T$ observations $\mathcal{T}=\{x_1,\cdots,x_T\}$. Since we will receive one data sample at time, we provide our model with a memory. This memory has a fixed size $M$ (\eg $M=16$ in all our experiments) and allows to store a sequence of $M$ target samples. Once these samples have been collected, we use them to compute a local estimate of the GBN statistics for the target domain. This estimate will be added to the global estimation of the statistics used by the GBN layers of our model, in the same way BN statistics are updated during training. After the update, we free the memory and restart collecting samples of the target domain. Obviously the presence of a memory can be used not only to estimate the statistics for updating the GBN layers, but also as a starting point for more complex optimization strategies. In this work, we exploit the memory to further refine the regressed scale and bias parameters. In particular, we follow recent BN-based DA algorithms \cite{carlucci2017autodial,carlucci2017just} and employ an entropy loss on the target domain data collected in the memory. This loss is applied to the "output" normalized through the statistics computed using the samples within the memory, in order to ensure a consistency with the training phase for the update of the statistics and the parameters of each GBN layer. 

\subsection{Experimental results}
\label{sec:pda-epxs}

 \subsubsection{Experimental setting}
\myparagraph{Datasets.}
We analyze the performance of AdaGraph on three datasets: the \textit{Comprehensive Cars} (CompCars) \cite{yang2015large}, the \textit{Century of Portraits}  \cite{ginosar2015century} and the \textit{CarEvolution} \cite{RematasICCVWS13}.

The \textit{Comprehensive Cars} (CompCars) \cite{yang2015large} dataset is a large-scale database composed of 136,726 images spanning a time range between 2004 and 2015. As in \cite{yang2016multivariate}, we use a subset of 24,151 images with 4 types of cars (\texttt{MPV}, \texttt{SUV}, \texttt{sedan} and \texttt{hatchback}) produced between 2009 and 2014 and taken under 5 different view points (front, front-side, side, rear, rear-side). Considering each view point and each manufacturing year as a separate domain we have a total of 30 domains. As in \cite{yang2016multivariate} we use a PDA setting where 1 domain is considered as source, 1 as target and the remaining 28 as auxiliary sets, for a total of 870 experiments. In this scenario, the metadata are represented as vectors of two elements, one corresponding to the year and the other to the view point, encoding the latter as in \cite{yang2016multivariate}.

\textit{Century of Portraits} (Portraits) \cite{ginosar2015century} is a large scale collection of images taken from American high school yearbooks. The portraits are taken over 108 years (1905-2013) across 26 states. We employ this dataset in a gender classification task, in two different settings. In the first setting we test our PDA model in a leave-one-out scenario, with a similar protocol to the tests on the \textit{CompCars} dataset. In particular, to define domains we consider spatio-temporal information and we cluster images according to decades and to spatial regions (we use 6 USA regions, as defined in \cite{ginosar2015century}). Filtering out the sets where there are less than 150 images, we obtain 40 domains, corresponding to 8 decades (from 1934 on) and 5 regions (\textit{New England}, \textit{Mid Atlantic}, \textit{Mid West}, \textit{Pacific}, \textit{Southern}). We follow the same experimental protocol of the \textit{CompCars} experiments, \ie we use one domain as source, one as target and the remaining 38 as auxiliaries. We encode the domain metadata as a vector of 3 elements, denoting the decade, the latitude (0 or 1, indicating north/south) and the east-west location (from 0 to 3), respectively. 
Additional details can be found in the appendix. 
In a second scenario, we use this dataset for assessing the performance of our continuous refinement strategy. In this case we employ all the portraits before 1950 as source samples and those after 1950 as target data.  

\textit{CarEvolution} \cite{yang2015large} is composed of car images collected between 1972 and 2013. It contains 1008 images of cars produced by three different manufacturers with two car models each, following the evolution of the production of those models during the years. We choose this dataset in order to assess the effectiveness of our continuous domain adaptation strategy. A similar evaluation has been employed in recent works considering online DA \cite{li2018domain}. As in \cite{li2018domain}, we consider the task of manufacturer prediction where there are three categories: \texttt{Mercedes}, \texttt{BMW} and \texttt{Volkswagen}. Images of cars before 1980 are considered as the source set and the remaining are used as target samples.

\myparagraph{Networks and Training Protocols.}
To analyze the impact on performance of our main contributions we consider the ResNet-18 architecture \cite{he2016deep} and perform experiments on the Portraits dataset. In particular, we apply our model by replacing each BN layer with its AdaGraph counterpart. We start with the network pre-trained on ImageNet, training it for 1 epoch on the source dataset, employing Adam as optimizer with a weight decay of $10^{-6}$ and a batch size of 16. We choose a learning rate of $10^{-3}$ for the classifier and $10^{-4}$ for the rest of the architecture. We train the network for 1 epoch on the union of source and auxiliary domains to extract domain-specific parameters. We keep the same optimizer and hyperparameters except for the learning rates, decayed by a factor of 10. The batch size is kept to 16, but each batch is composed by elements of a single pair year-region belonging to one of the available domains (either auxiliary or source). The order of the pairs is randomly sampled within the set of allowed ones.

In order to fairly compare with previous methods we also consider Decaf features \cite{donahue2014decaf}.  In particular, in the experiments on the CompCars dataset, we use Decaf features extracted at the \texttt{fc7} layer. Note that these features are comparable to the ones used in \cite{yang2016multivariate} (\ie penultimate layer of the VGG-F model in \cite{chatfield2014return}).  Similarly, for the experiments on CarEvolution, we follow \cite{li2018domain} and use Decaf features extracted at the \texttt{fc6} layer. In both cases, we apply our model by adding either a BN layer or our AdaGraph approach directly to the features, followed by a ReLU activation and a linear classifier. For these experiments 
we train the model on the source domain for 10 epochs using Adam as optimizer with a learning rate of $10^{-3}$, a batch size of 16 and a weight decay of $10^{-6}$. The learning rate is decayed by a factor of 10 after 7 epochs. For CompCars, when training with the auxiliary set, we use the same optimizer, batch size and weight decay, with a learning rate $10^{-4}$ for 1 epoch. Domain-specific batches are randomly sampled, as for the experiments on Portraits.

For all the experiments we use as distance measure 
{$d(x,y)=\frac{1}{2\cdot\sigma}\cdot||x-y||_2^2$  with $\sigma=0.1$ and set 
$\lambda$ equal to $1.0$, both in the training and in the refinement stage.}
At test time, we classify each input image as it arrives, performing the refinement step after the classification. {The buffer size in the refinement phase is equal to 16 and we set $\alpha=0.1$, the same used for updating the GBN components while training with the auxiliar domains.

We implemented\footnote{The code is available at \url{https://github.com/mancinimassimiliano/adagraph}} our method with the PyTorch~\cite{paszke2019pytorch} framework  and  our  evaluation  is  performed  using  a  NVIDIA GeForce 1080 Ti GTX GPU.} 

\subsubsection{Results}
In this section we report the results of our evaluation, showing both an empirical analysis of the proposed contributions and a comparison with state-of-the-art-approaches. 

\myparagraph{Analysis of AdaGraph.}
We first analyze the performance of our approach by employing the Portraits dataset. In particular, we evaluate the impact of (i) introducing a graph to predict the target domain BN statistics (\textit{\textit{AdaGraph BN}}), (ii) adding scale and bias parameters trained in isolation (\textit{AdaGraph SB}) or jointly (\textit{AdaGraph Full}) and (iii) adopting the proposed refinement strategy (\textit{AdaGraph + Refinement}). As baseline\footnote{We do not report the results of previous approaches \cite{yang2016multivariate} since the code is not publicly available.} we consider the model trained only on the source domain and, as an upper bound, a corresponding DA method which is allowed to use target data during training. In our case, the upper bound corresponds to a model similar to the method proposed in \cite{carlucci2017autodial}. 

The results of our ablation are reported in Table \ref{tab:PDA-portraits-PDA}, where we report the average classification accuracy corresponding to two scenarios: \textit{across decades} (considering the same region for source and target domains) and \textit{across regions} (considering the same decade for source and target dataset). The first scenario corresponds to 280 experiments, while the second to 160 tests.
As shown in the table, by simply replacing the statistics of BN layers of the source model with those predicted through our graph a large boost in accuracy is achieved ($+4\%$ in the \textit{across decades} scenario and $+2.4\%$ in the \textit{across regions} one). At the same time, estimating the scale and bias parameters without considering the graph is suboptimal. In fact there is a misalignment between the forward pass of the training phase (\ie considering only domain-specific parameters) and how these parameters will be combined at test time (\ie considering also the connection with the other nodes of the graph). 
Interestingly, in the \textit{across regions} setting, our full model slightly drops in performance with respect to predicting only the BN statistics. 
This is probably due to how regions 
are encoded in the metadata (\ie considering geographical location), making it difficult to capture factors (\eg cultural, historical) which can be more discriminative to characterize the population of a region or a state. However, as stated in Section \ref{sec:PDA-continuous-da}, employing a continuous refinement strategy allows the method to compensate for prediction errors. As shown in Table \ref{tab:PDA-portraits-PDA}, with a refinement step (\textit{AdaGraph + Refinement}) the accuracy constantly increases, 
filling the gap between the performance of the initial model and our DA upper bound.

It is worth noting that applying the refinement procedure to the source model (\textit{Baseline + Refinement}) leads to better performance (about $+4\%$ in the \textit{across decades} scenario and $+2.1\%$ for \textit{across regions} one).
More importantly, the performance of the \textit{Baseline + Refinement} method is always worse than what obtained by \textit{AdaGraph + Refinement}, 
because our model provides, on average, a better starting point for the refinement procedure.

\begin{figure*}[t]
   \includegraphics[width=1.0\textwidth]{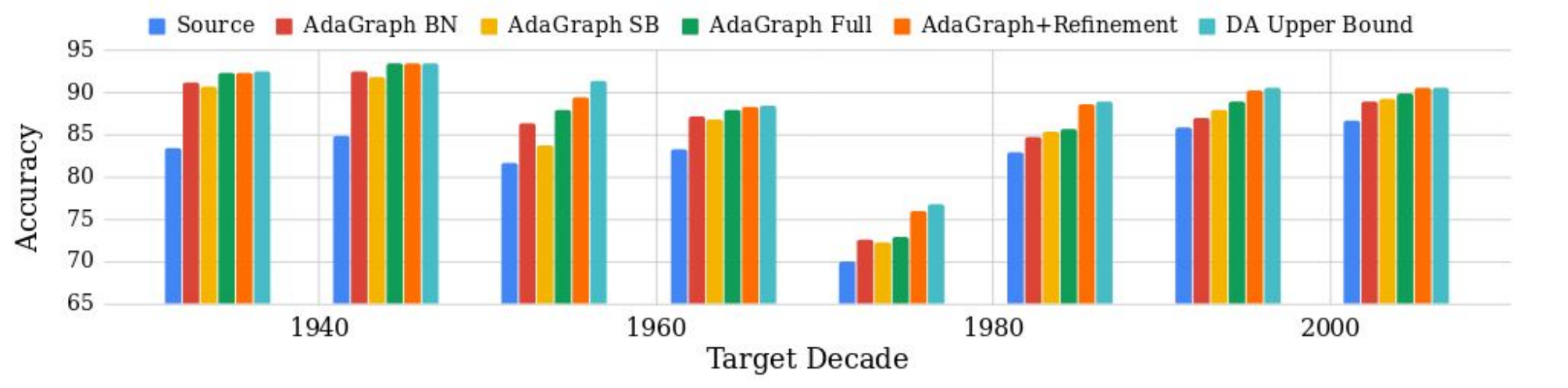}
   \caption{Portraits dataset: comparison of different models in the PDA scenario with respect to the average accuracy on a target decade, fixed the same region of source and target domains. The models are based on ResNet-18.}
\label{fig:PDA-ablation}
\end{figure*}

Figure \ref{fig:PDA-ablation} shows the results associated to the \textit{across decades} scenario. Each bar plot corresponds to experiments where the target domain is associated to a specific year. 
As shown in the figure, on average, our full model outperforms both \textit{AdaGraph BN} and \textit{AdaGraph SB}, showing the benefit of the proposed graph strategy. 
The results in the figure clearly also show that our refinement strategy always leads to a boost in performance. 

\begin{table}[t]
			\caption{Portraits dataset. Ablation study.} 
		\centering
		\scalebox{.99}{
		\begin{tabular}{ l  | c  c} 
			\hline
			Method & Across Decades & Across Regions \\\hline
           Baseline & 82.3& 89.2\\
           AdaGraph BN & 86.3&\textbf{91.6}\\
           AdaGraph SB & 86.0&90.5\\
           AdaGraph Full & \textbf{87.0}&91.0\\\hline
           Baseline + Refinement & 86.2&91.3\\
           AdaGraph + Refinement & \textbf{88.6} &\textbf{91.9}\\
           \hline\hline
           DA upper bound &89.1&92.1 \\ \hline
        \end{tabular}
        }
        \label{tab:PDA-portraits-PDA}
\end{table}

\myparagraph{Comparison with the state of the art.}
Here we compare the performances of our model with state-of-the-art PDA approaches. We use the CompCars dataset and we benchmark against the Multivariate Regression (MRG) methods proposed in \cite{yang2016multivariate}. 
We apply our model in the same setting as \cite{yang2016multivariate} and 
 perform 870 different experiments, computing the average accuracy (Table \ref{tab:PDA-compcars-sota}). Our model outperforms the two methods proposed in \cite{yang2016multivariate} by improving the performances of the Baseline network by $4\%$. {AdaGraph} alone outperforms the Baseline model when it is updated with our refinement strategy and target data (\textit{Baseline + Refinement}). 
When coupled with a refinement strategy, our graph-based model further improves the performances, {filling the gap between AdaGraph and our DA upper bound}. It is interesting to note that our model is also effective when there are no metadata available in the target domain. In the table, \textit{AdaGraph (images)} corresponds to our approach when, instead of initializing the BN layer for the target exploiting metadata, we employ the current input image and a domain classifier to obtain a probability distribution over the graph nodes, as described in Section \ref{sec:PDA-graph-da}.
The results in the Table show that \textit{AdaGraph (images)} is more accurate than \textit{AdaGraph (metadata)}. 

\begin{table}[t]
			\caption{CompCars dataset \cite{yang2015large}. Comparison with state of the art.  $\dagger$ denotes Decaf features as input, $\ddagger$ denotes VGG-Full.} 
		\centering
		\scalebox{.99}{
		\begin{tabular}{ l  | c } 
			\hline
			Method & Avg. Accuracy\\\hline
           Baseline$^\ddagger$ \cite{yang2016multivariate}& 54.0\\
           Baseline + BN$^\dagger$& 56.1\\
           MRG-Direct$^\ddagger$ \cite{yang2016multivariate}& 58.1\\
           MRG-Indirect$^\ddagger$ \cite{yang2016multivariate}& 58.2\\
           AdaGraph (metadata)$^\dagger$ & 60.1 \\
           AdaGraph (images)$^\dagger$ & \textbf{60.8} \\
           \hline
           Baseline + Refinement$^\dagger$ & 59.5\\
           AdaGraph + Refinement$^\dagger$ & \textbf{60.9}\\
        \hline\hline
        DA upper bound$^\dagger$ & {60.9}\\	\hline
        \end{tabular}
        }
        \label{tab:PDA-compcars-sota}
\end{table}

\myparagraph{Exploiting AdaGraph Refinement for Continous Domain Adaptation.} In Section \ref{sec:PDA-continuous-da}, we have shown a way to boost the performances of our model by leveraging the stream of incoming target data and refine the estimates of the target BN statistics and parameters. Throughout the experimental section, we have also demonstrated how this strategy improves 
the target classification model, with performances close to DA methods which exploit target data during training.

In this section we show how this approach can be employed as a competitive method in the case of continuous domain adaptation \cite{hoffman2014continuous}. We consider the CarEvolution dataset and compare the performances of our proposed strategy with two state of the art algorithms: the manifold-based adaptation method in \cite{hoffman2014continuous} and the low-rank SVM strategy presented in \cite{li2018domain}. 
As in \cite{li2018domain} and \cite{hoffman2014continuous}, we apply our adaptation strategy after classifying each novel image and compute the overall accuracy. {The images of the target domain are presented to the network in a chronological order \ie from 1980 to 2013}. 
The results are shown in Table \ref{tab:PDA-ced-continuous}. While the integration of a BN layer alone leads to better performances over the baseline, our refinement strategy produces an additional boost of about 3\%. If scale and bias parameters are refined considering the entropy loss,
accuracy further increases. 

We also test the proposed model on a similar task considering the Portraits dataset. 
The results of our experiments are shown in Table \ref{tab:PDA-faces-continuous}. Similarly to what observed on the previous experiments, continuously adapting our deep model as target data become available leads to better performance with respect to the baseline. The refinement of scale and bias parameters contributes to a further boost in accuracy.

\begin{table}[t]
			\caption{CarEvolution \cite{RematasICCVWS13}: comparison with state of the art.} 
		\centering
		\scalebox{.99}{
		\begin{tabular}{ l  | c } 
			\hline
			Method & Accuracy\\\hline
           Baseline SVM \cite{li2018domain}& 39.7\\
           Baseline + BN & 43.7\\
           CMA+GFK \cite{hoffman2014continuous}& 43.0\\
           CMA+SA \cite{hoffman2014continuous}& 42.7\\
           LLRESVM \cite{li2018domain}& 43.6\\
           LLRESVM+EDA\cite{li2018domain}& 44.3\\
           ONDA (Baseline+Refinement Stats) \cite{mancini2018kitting} & 46.5 \\
           Baseline + Refinement Full& \textbf{47.3} \\ \hline
        \end{tabular}
        }
        \label{tab:PDA-ced-continuous}
\end{table}

\begin{table}[t]
			\caption{Portraits dataset \cite{yang2015large}: performances of the refinement strategy on the continuous adaptation scenario} 
		\centering
		\scalebox{.99}{
		\begin{tabular}{ l  | c c c } 
			\hline
			Method & Baseline & Refinement Stats \cite{mancini2018kitting} & Refinement Full\\\hline
           Accuracy& 81.9& 87.3 & \textbf{88.1}\\ \hline
        \end{tabular}
        }
        \label{tab:PDA-faces-continuous}
\end{table}

\subsection{Conclusions}
We present the first deep architecture for Predictive Domain Adaptation, AdaGraph. We leverage metadata information to build a graph where each node represents a domain, while the strength of an edge models the similarity among two domains according to their metadata. We then propose to exploit the graph for the purpose of DA and we design novel domain-alignment layers.
This framework yields the new state of the art on standard PDA benchmarks.
We further present an approach to exploit the stream of incoming target data such as to refine the target model. 
We show that this strategy itself is also an effective method for continuous DA, outperforming state-of-the-art approaches, and our previous ONDA model.
 In future works, it would be interesting to explore methodologies to incrementally update the graph and to automatically infer relations among domains, even in the absence of metadata. Moreover, the connections among the nodes can be used in few-shot scenarios, using the relations among domains to provide additional feedback to nodes of domains with few samples.
 
 This section concludes our works which considered the domain shift problem in isolation both in the presence and in absence of target data and under different settings. In the next chapters, we will describe our works that tackled the semantic shift problem in isolation first (Chapter \ref{chap:icl}) and coupled with the domain shift one lately (Chapter \ref{chap:both}).

\chapter{Recognizing New Semantic Concepts}
\label{chap:icl}
\textit{This chapter analyzes different problems concerning the extension of a pre-trained architecture to new visual concepts in an incremental fashion, varying the knowledge we want to add and what we want to recognize. As in the previous chapter, we start by providing a general formulation of the problem (Sec. \ref{sec:IL-ps}) and reviewing previous works on incremental learning of classes/tasks and in an open world (Sec. \ref{sec:IL-relateds}). In Sec. \ref{sec:IL-multitask} we show how we can extend a model perform the same task (i.e. classification) across multiple visual domains with different output spaces (e.g. digits recognition, street signal classification) through affinely transformed binary mask \cite{mancini2020boostingmva}. This approach extends previous works on multi-domain learning \cite{mallya2018piggyback}, achieving the highest (at the time of acceptance) trade-off among learning new tasks effectively, and using a low number of parameters. In Sec. \ref{sec:IL-semantic-seg} we focus on the incremental class learning problem when new classes are added to the same classification head of old ones but in the context of semantic segmentation \cite{cermelli2020modeling}. Here we show how there is an inherent problem in this setting caused by the semantic shift of the background class across different incremental steps. We show how this problem can be addressed by a simple modification of the cross-entropy and distillation losses employed in previous approaches \cite{li2017learning}. Finally, in Sec. \ref{sec:IL-owr}, we analyze the problem of open-world recognition, where the goal is to not only include new classes incrementally but also to detect if an image belongs to an unknown category. We analyze the problem in robotics scenarios, starting by implementing the first deep approach for this problem \cite{mancini2019knowledge}. The approach extends standard non-parametric methods \cite{bendale2015towards} and is further improved in a subsequent work by clustering-based losses and class-specific rejection options \cite{fontanel2020boosting}. In \cite{mancini2019knowledge} we also discuss how the approach could be employed in a realistic scenario by obtaining datasets with new knowledge directly from the web, a first step towards having agents able to automatically expand their visual recognition capabilities by reasoning on what they see in the real world.}

\section{Problem statement}
\label{sec:IL-ps}
\myparagraph{Overview.} In Chapter \ref{chap:da} we have analyzed multiple algorithms being able to overcome the domain shift problem in various scenarios. However, while the domain shift is a crucial issue for the applicability of visual systems in real scenarios, it deals with one side of the problem: changes in the input distribution without changes in the semantic space. In this chapter, we are interested in tackling the opposite problem. Given a model trained to recognize a set of classes in a given domain, we want to extend its output space, equipping it with the ability to recognize semantic concepts not included in the initial training set.  

The methodologies used to add new knowledge to a pre-trained model can be roughly divided into three main categories, depending on the information we have about the training classes. In the first category, we receive data for the novel concepts we want our model to recognize. This scenario is usually called \textit{continual/incremental learning} \cite{de2019continual} and requires to add new knowledge to the model without access to the initial training set and, more importantly, without forgetting previous knowledge \cite{mccloskey1989catastrophic,kemker2018measuring,goodfellow2013empirical}. In the second category, we have models using few sample images of the classes of interests at test time, using the initial training set to learn how to compare this support set composed of few images with a query image. These models fall in the \textit{few-shot learning} paradigm and require to receive, at test time, sample images of the classes we want to recognize \cite{snell2017prototypical,finn2017model}. The last category of methods learn to recognize concepts beyond the initial training set without any image available but using \textit{class descriptions} (e.g. binary attributes \cite{lampert2013awa}, word embeddings \cite{mikolov2013distributed}). In this scenario, called \textit{zero-shot learning} \cite{xian2018zeroshotgood}, a model has to map images in a given semantic embedding space where all classes (seen and unseen) are projected. In this way, it is possible to compare images with unseen and/or seen concepts to perform the final classification.

In this thesis, we will consider both incremental and zero-shot learning models. In particular, in this chapter, we will consider scenarios where the domain shift is not present, i.e. training and test domains are the same, but the semantic space of the model is incrementally extended over time, as for the first category. In Chapter \ref{chap:both}, we will show how we can obtain a model attacking both domain- and semantic shift, recognizing unseen categories (as in zero-shot learning) in unseen domains (as in domain generalization).

\myparagraph{Incremental Learning.} Let us formalize the incremental learning problem. Assuming we have a model pre-trained on a set $\mathcal{T}_0=\{(x^0_i,y^0_i)\}_{i=1}^{n_0}$ with $x_i^0 \in \mathcal{X}$ and $y_i^0 \in \mathcal{C}_0$. Note that $\mathcal{X}$ is the input space, as in Section \ref{sec:da-problem}, while $\mathcal{C}_0$ is the output space of the initial training set (i.e. the set of classes in $\mathcal{T}_0$). Using this set we can obtain a function $f_0^\theta:\mathcal{X} \rightarrow \mathcal{C}_0$ parametrized by $\theta$ and mapping images into the initial output space. To include new concepts in $f_0^\theta$, we receive a new dataset containing the new concepts of interests. Since we might perform multiple training steps, let us denote with $\mathcal{T}_t=\{(x^t_i,y^t_i)\}_{i=1}^{n_t}$ the dataset we receive at time $t$. Note that, while the input space does not change (i.e. $x_i^t \in \mathcal{X}$) the output space does and we have $y_i^t \in \mathcal{C}_t$ with $\mathcal{C}_i\bigcap \mathcal{C}_j = \emptyset$ if $i\neq j$. After $T$ training learning steps, our goal is to obtain a model  $f_T^\theta:\mathcal{X} \rightarrow \mathcal{Y}_T$ where the output space $\mathcal{Y}_T$ comprises all the concepts seen until the training step T, i.e. $\mathcal{Y}_T=\bigcup {C_t}_{t=0}^{T}$. 

Under this definition, we have different problems, depending on how the output space is built \cite{chaudhry2018riemannian}. The first distinction is on the number of classification heads. We have \textit{single-head} models, where there is a single classification head for all the concepts in $\mathcal{Y}^T$, and \textit{multi-head}, in case we have one head per set of classes $\mathcal{C}_t$. In this latter scenario, despite some exceptions \cite{aljundi2019task,rao2019continual}, it is common to give as input to the prediction function the information about the output space of interests, i.e. $f_T^\theta:\mathcal{X}\times \mathcal{Z} \rightarrow \mathcal{Y}_T$ with $\mathcal{Z}=\{0,\cdots,T\}$. We will analyze this scenario in the context of \textbf{Multi-Domain Learning} \cite{rebuffi2017learning,bilen2017universal}, in Section \ref{sec:IL-multitask}. 

Considering the single-head scenario, the second distinction relates to the limits of $\mathcal{Y}^T$. In case $\mathcal{Y}^T$ is \textit{closed-ended}, we have the standard \textit{incremental class learning} scenario and we ask our model to recognize which to which class in $\mathcal{Y}^T$ our image belongs. We will analyze this setting in Section \ref{sec:IL-semantic-seg}, in the task of semantic segmentation. In case $\mathcal{Y}^T$ is \textbf{open-ended}, i.e. the model includes a rejection option for known classes, we are in the \textbf{open world recognition} one, and our model is asked to recognize the class of an image and, eventually, detecting if it belongs to an unknown concept. This scenario will be the focus of Section \ref{sec:IL-owr}.

In the following section, we will report the relevant literature for incremental learning, multi-domain learning and open world recognition.

\section{Related Works}
\label{sec:IL-relateds}

\myparagraph{Incremental Learning.}
The problem of catastrophic forgetting \cite{mccloskey1989catastrophic} has been extensively studied for image classification tasks \cite{de2019continual}. 
{Previous works can be grouped in three categories \cite{de2019continual}: replay-based \cite{rebuffi2017icarl, castro2018end, shin2017continual, hou2019learning, wu2018memory, ostapenko2019learning}, regularization-based \cite{kirkpatrick2017overcoming,chaudhry2018riemannian,zenke2017continual,li2017learning, dhar2019learning}, and parameter isolation-based \cite{mallya2018packnet, mallya2018piggyback, rusu2016progressive}.
In replay-based methods, examples of previous tasks are either stored \cite{rebuffi2017icarl, castro2018end, hou2019learning, wu2019large} or generated \cite{shin2017continual, wu2018memory, ostapenko2019learning} and then replayed while learning the new task. 
Parameter isolation-based methods \cite{mallya2018packnet, mallya2018piggyback, rusu2016progressive} assign a subset of the parameters to each task to prevent forgetting.} 
Regularization-based methods can be divided in prior-focused and data-focused. 
The former \cite{zenke2017continual, chaudhry2018riemannian, kirkpatrick2017overcoming, aljundi2018memory} define knowledge as the parameters value, constraining the learning of new tasks by penalizing changes of important parameters for old ones. 
The latter \cite{li2017learning, dhar2019learning} exploit distillation \cite{hinton2015distilling} and use the distance between the activations produced by the old network and the new one as a regularization term to prevent catastrophic
forgetting. 

Despite these progresses, very few works have gone beyond image-level classification. A first work in this direction is \cite{shmelkov2017incremental} which considers \icl\ in object detection 
proposing a distillation-based method adapted from \cite{li2017learning} for tackling novel class recognition and bounding box proposals generation. 
In this work we also take a similar approach to \cite{shmelkov2017incremental} and we resort on distillation. However, here we propose to address the problem of modeling the background shift which is peculiar of the semantic segmentation setting. 

To our knowledge, the problem of \icl\ in semantic segmentation has been addressed only in
\cite{ozdemir2018learn,ozdemir2019extending,tasar2019incremental,michieli2019incremental}. Ozdemir \etal \cite{ozdemir2018learn,ozdemir2019extending} describe an \icl\ approach for medical imaging, extending a standard image-level classification method \cite{li2017learning} to segmentation and devising a strategy to select relevant samples of old datasets for rehearsal. Taras 
\etal proposed a similar approach for segmenting remote sensing data. 
Differently, Michieli \etal
\cite{michieli2019incremental} consider \icl\ for semantic segmentation 
in a particular setting where labels are provided for old classes while learning new ones. Moreover, they assume the novel classes to be never present as background in pixels of previous learning steps. These assumptions strongly limit the applicability of their method. 

Here we propose a more principled formulation of the \icl\ problem in semantic segmentation. 
In contrast with previous works, we do not limit our analysis to medical \cite{ozdemir2018learn} or remote sensing data \cite{tasar2019incremental} and we do not impose any restrictions on how the label space should change across different learning steps \cite{michieli2019incremental}.  Moreover, 
we are the first to provide a comprehensive  experimental evaluation of \sota \icl\ methods on commonly used semantic segmentation benchmarks 
and to explicitly introduce and tackle the semantic shift of the background class, 
a problem recognized but largely overseen by previous works \cite{michieli2019incremental}.

\myparagraph{Multi-domain Learning.}
{Another challenge in incremental learning is extending a pre-trained model to address new tasks, each with  different output space. Indeed, the need for visual models capable of addressing multiple domains received a lot of attention in recent years for what concerns both multi-task learning \cite{zamir2018taskonomy,liu2019end,cermelli2019rgb} and multi-domain learning \cite{rebuffi2017learning,rosenfeld2017incremental}. Multi-task learning focuses on learning multiple visual tasks (\eg semantic segmentation, depth estimation \cite{liu2019end}) with a single architecture. On the other hand, the goal of multi-domain learning is building a model able to address a task (e.g. classification) in multiple visual domains (e.g. real photos, digits) \textit{without forgetting} previous domains and by using \textit{fewer parameters} possible. An important work in this context is \cite{bilen2017universal}, where the authors showed how multi-domain learning can be addressed by using a network sharing all parameters except for batch-normalization (BN) layers \cite{ioffe2015batch}. In \cite{rebuffi2017learning}, the authors introduced the Visual Domain Decathlon Challenge, a first multi-domain learning benchmark. The first attempts in addressing this challenge involved domain-specific residual components added in standard residual blocks, either in series \cite{rebuffi2017learning} or in parallel \cite{rebuffi2018efficient}, 
In \cite{rosenfeld2017incremental} the authors propose to use controller modules where the parameters of the base architecture are recombined channel-wise, while in \cite{liu2019end} exploits domain-specific attention modules.  
Other effective approaches include devising instance-specific fine-tuning strategies \cite{guo2019spottune}, target-specific architectures \cite{morgado2019nettailor} and learning covariance normalization layers \cite{li2019efficient}.

In \cite{mallya2018packnet} only a reserved subset of network parameters is considered for each domain.} The intersection of the parameters used by different domains is empty, thus the network can be trained end-to-end for each domain. Obviously, as the number of domain increases, fewer parameters are available for each domain, with a consequent limitation on the performances of the network. {To overcome this issue, in \cite{mallya2018piggyback} the authors proposed a more compact and effective solution based on directly learning domain-specific binary masks. 
The binary masks determine which of the network parameters are useful for the new domain and which are not, changing the actual composition of the features extracted by the network. This approach inspired subsequent works, improving both either the power of the binary masks \cite{mancini2018adding} or the number of bits required, masking directly an entire channel \cite{berriel2019budget}. 

In our work \cite{mancini2020boostingmva}, we take inspiration from these last research trends. In particular, we generalize the design of the binary masks employed in \cite{mallya2018piggyback} and \cite{mancini2018adding} considering neither simple multiplicative binary masks nor simple affine transformations of the original weights \cite{mancini2018adding} but a general and flexible formulation capturing both cases. 
Experiments show how our approach in \cite{mancini2020boostingmva} leads to a boost in the performances while using a comparable number of parameters per domain. Moreover, our approach achieves performances comparable to more complex models \cite{rebuffi2018efficient,morgado2019nettailor,li2019efficient,li2019efficient} in the challenging Visual Domain Decathlon challenge, largely reducing the gap of binary-mask based methods with the current state of the art. Note that while learning to address the same task (i.e. classification) in multiple visual domains, these trend of works addresses catastrophic forgetting by adding domain-specific parameters, extending the semantic extent of a pre-trained model by exploiting isolated set of parameters. In fact, if the initial network parameters remain untouched, the catastrophic forgetting problem is avoided but at the cost of the additional parameters required. The extreme case is the work of \cite{rusu2016progressive} in the context of reinforcement learning, where a parallel network is added each time a new domain is presented with side domain connections, exploited to improve the performances on novel domains. Differently to \cite{rusu2016progressive}, the mask-based approaches \cite{mallya2018packnet,mallya2018piggyback, mancini2018adding,mancini2020boostingmva} require a much lower overhead in terms of total parameters, showing comparable or even superior results to task-specific fine-tuned models \cite{mallya2018packnet,mallya2018piggyback,mancini2020boostingmva}.

}

\myparagraph{Open World Recognition.}
The necessity of breaking the closed-world assumption (CWA) for robot vision systems \cite{sunderhauf2018limits} has lead various research efforts on understanding how to extend pre-trained models with new semantic concepts while retain previous knowledge and detecting possibly unknown ones. There are two components towards this goal: the first is incrementally adding new categories to the pre-trained model, while the second is maintaining a right estimation of the uncertainty on the predictions allowing to reject inputs of unseen classes. Due to the central role this task has in real-world applications, recent years have seen a growing interests among robotic vision researches on topics such as continual \cite{lesort2020continual} and incremental learning \cite{valipour2017incremental,camoriano2017incremental,cermelli2019rgb}. In \cite{pasquale2015teaching}, the authors study how to update the visual recognition system of a humanoid robot on multiple training sessions. 
In \cite{camoriano2017incremental}, a variant of the Regularized Least Squares algorithm is introduced to add new classes to a pre-trained model. 
In \cite{parisi2018lifelong}, a growing dual-memory is proposed to dynamically learn novel object instances and categories. 
In \cite{lagunes2019learning} the authors proposed to learn an embedding in order to perform fast incremental learning of new objects. Another solution to this problem can exploit the help of a human-robot interaction, as in \cite{valipour2017incremental} where a robot incrementally learns to detect new objects as they are manually pointed by a human. 

While these approaches focus on incremental and continual learning, acting in the open world requires both detecting unknown concepts automatically and adding them in subsequent learning stages. 
Towards this objective, in \cite{bendale2015towards} the authors introduced the OWR setting, as a more general and realistic scenario for agents acting in the real world. 
In \cite{bendale2015towards}, the authors extend the Nearest Class Mean (NCM) classifier \cite{mensink2012metric,guerriero2018deep}
 to act in the open set scenario, proposing the Nearest Non-Outlier algorithm (NNO). In order to estimate whereas a test sample belongs to the known or unknown set of categories, this method introduces a {rejection} threshold that, after the first initialization phase, is kept fixed for subsequent learning episodes. In \cite{de2016online}, the authors proposed to tackle OWR with the Nearest Ball Classifier, with a rejection threshold based on the confidence of the predictions. In \cite{mancini2019knowledge}, we extended the NNO algorithm of \cite{bendale2015towards} by employing an end-to-end trainable deep architecture as feature extractor, with a dynamic update strategy for the rejection threshold. Moreover, our work was the first to consider the collection of datasets containing new knowledge using web resources, towards agent able to automatically include new knowledge with limited to none human supervision.
In the subsequent work \cite{fontanel2020boosting}, we showed how we can improve the performances of NCM based classifier for OWR through a global to local clustering loss. Moreover, differently for previous works, we designed class-specific rejection threshold rather are explicitly learned rather than fixed based on heuristic strategies. 

\section[Sequential and Memory Efficient Learning of New Datasets]{Sequential and Memory Efficient Learning of \\New Datasets \footnotemark\footnotetext{M. Mancini, E.Ricci, B. Caputo, S. Rota Bul\`o. {\sl Adding New Tasks to a Single Network with Weight Transformations using Binary Masks}. European Computer Vision Conference Workshop on Transferring and Adapting Source Knowledge in Computer Vision 2018.}\footnotemark\footnotetext{M. Mancini, E.Ricci, B. Caputo, S. Rota Bul\`o. {\sl Boosting Binary Masks for Multi-Domain Learning through Affine Transformations}. Machine Vision and Applications 2020.}}
\label{sec:IL-multitask}
In this section, we focus on the problem of multi-domain learning \cite{rebuffi2017learning,bilen2017universal}. Following the problem statement of \cite{rebuffi2017learning}, the goal of multi-domain learning is to train a model to address multiple classification tasks using as few parameters per each of them. In the following, we focus on the case, considered also in \cite{rebuffi2017learning} where we adapt an initial pre-trained model to address novel tasks sequentially. 
This capability is crucial for increasing the knowledge of an intelligent system and developing effective incremental \cite{ring1997child,kuzborskij2013n}, life-long \cite{thrun1995lifelong,thrun2012learning,silver2013lifelong} learning algorithms. While fascinating, achieving this goal requires facing multiple challenges. First, 
learning a new task should not negatively affect the performance on old tasks. Second, it should be avoided adding multiple parameters to the model for each new task learned, as it would lead to poor scalability of the framework. 
In this context, while deep learning algorithms have achieved impressive results on many computer vision benchmarks \cite{krizhevsky2012imagenet,he2016deep,girshick2014rich,long2015fully}, mainstream approaches for adapting deep models to novel tasks tend to suffer from the  problems  mentioned above. In fact, fine-tuning a given architecture to new data does produce a powerful model on the novel task, at the expense of a degraded performance on the old ones, resulting in the well-known phenomenon  of  catastrophic forgetting \cite{french1999catastrophic,goodfellow2013empirical}. At the same time, replicating the network parameters and training a separate network for each task is a powerful approach that preserves performances on old tasks, but at the cost of an explosion of the network parameters \cite{rebuffi2017learning}.

{Different works addressed these problems by either considering losses encouraging the preservation of the current weights \cite{li2017learning,kirkpatrick2017overcoming} 
or by designing task-specific network parameters \cite{rusu2016progressive,rebuffi2017learning,rosenfeld2017incremental,mallya2018packnet,mallya2018piggyback}. Interestingly, in \cite{mallya2018piggyback} the authors showed that an effective strategy for achieving good sequential multi-task learning performances with a minimal increase in term of network size
is to create a binary mask for each task. In particular, this mask is then multiplied by the main network weights, determining which of them are useful for addressing the new task. } 

\begin{figure*}[t]
\centering
 \includegraphics[width=1.
 \textwidth]{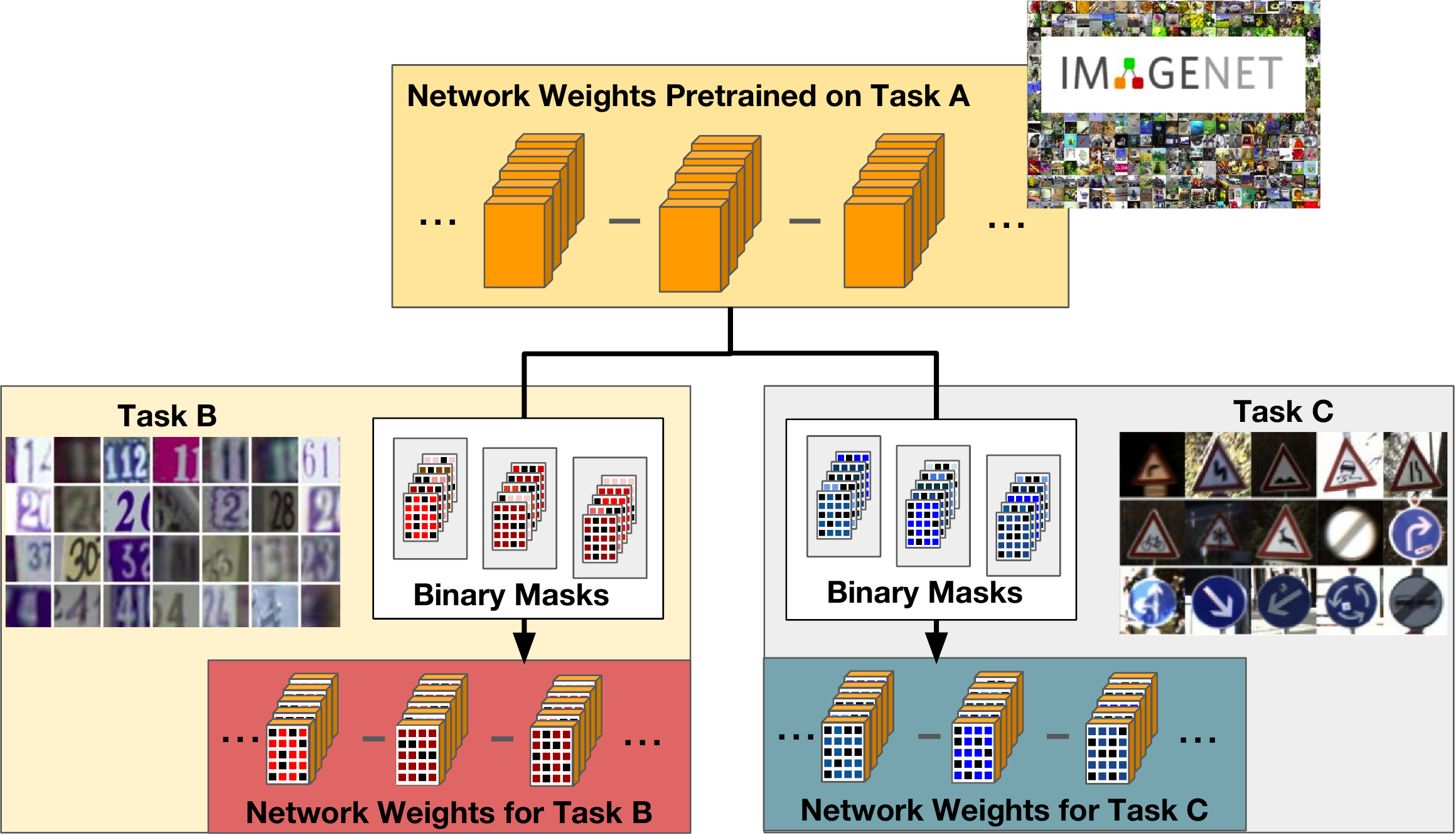} 
  \caption{Idea behind our BAT approach. A network pre-trained on a given recognition task A (\ie ImageNet) can be extended to tackle other recognition tasks B (\eg digits) and C (\eg traffic sign)  by simply transforming the network weights (orange cubes) through task-specific binary masks (colored grids).}
  \label{MT-fig:teaser-masks}
  \end{figure*}

{
In this section, we take inspiration from these last work. 
and we formulate the sequential multi-task learning as the problem of learning a perturbation of a \emph{baseline}, pre-trained network, in a way to maximize the performance on a new task. Importantly, the perturbation should be compact in the sense of limiting the number of additional parameters required with respect to the baseline network.
To this extent, we apply an affine transformation to each convolutional weight of the baseline network, which involves both a learned binary mask and few additional parameters. The binary mask 
is used as a scaled and shifted additive component and as a multiplicative filter to the original weights.  {Figure \ref{MT-fig:teaser-masks} shows an example application of our proposed algorithm. Given a network pre-trained on a particular task (\ie ImageNet \cite{russakovsky2015imagenet}, orange blocks) we can transform its original weights through binary masks (colored grids) and obtain a network which effectively addresses a novel tasks (\eg digit \cite{netzer2011reading} or traffic sign \cite{stallkamp2012man} recognition)}  
We name our solution \textbf{BAT} (Binary-mask Affinely Transformed for multi-domain learning).  This solution allows us to achieve two main goals: 1) boosting the performance of each task-specific network that we train, by leveraging the higher degree of freedom in perturbing the baseline network, while 2) keeping a low per-task overhead in terms of additional parameters (slightly more than 1 bit per parameter per task).}

We assess the validity of BAT, and some variants thereof, on standard benchmarks including the Visual Decathlon Challenge~\cite{rebuffi2017learning}. The experimental results show that our model achieves performances comparable with fine-tuning separate networks for each recognition task on all benchmarks, while retaining a very small overhead in terms of additional parameters per task. {Notably, we achieve results comparable to state-of-the-art models on the Visual Decathlon Challenge \cite{rebuffi2017learning} but without requiring multiple training stages \cite{li2019efficient} or a large number of task-specific parameters \cite{guo2019spottune,rebuffi2018efficient}.}

\subsection{Problem Formulation}
We address the problem of sequential 
learning of new tasks, \ie we modify a \emph{baseline} network such as, \eg ResNet-50 pre-trained on the ImageNet classification task, so to maximize its performance on a new task, while limiting the amount of additional parameters needed. The solution we propose exploits the key idea from Piggyback~\cite{mallya2018piggyback} of learning task-specific masks, but instead of pursuing the simple multiplicative transformation of the parameters of the baseline network, we define a parametrized, affine transformation mixing a binary mask and real parameters
that significantly increases the expressiveness of the approach, leading to a rich and nuanced ability to adapt the old parameters to the needs of the new tasks. This in turn brings considerable improvements on all the conducted experiments, as we will show in the experimental section, while retaining a reduced, per-task overhead.

{Let us assume to be given a pre-trained, \emph{baseline} network $f_0(\cdot; \Theta, \Omega_0):\set X\to\set Y_0$ assigning a class label in $\set Y_0$ to elements of an input space $\set X$ (\eg images)}.\footnote{We focus on classification tasks, but the proposed method applies also to other tasks.}
{The parameters of the baseline network are partitioned into two sets: $\Theta$ comprises parameters that will be shared for other domains, whereas $\Omega_0$ entails the rest of the parameters (\eg the classifier).
Our goal is to learn for each domain $i\in\{1,\ldots,\con m\}$, with a possibly different output space $\set Y_i$,
a classifier $f_i(\cdot;\Theta,\Omega_i):\set X\to\set Y_i$. Here, $\Omega_i$ entails the parameters specific for the $i$th domain, while $\Theta$ holds the shareable parameters of the baseline network mentioned above.}
\begin{figure*}[t]
 \centering
 \includegraphics[width=0.98\textwidth]{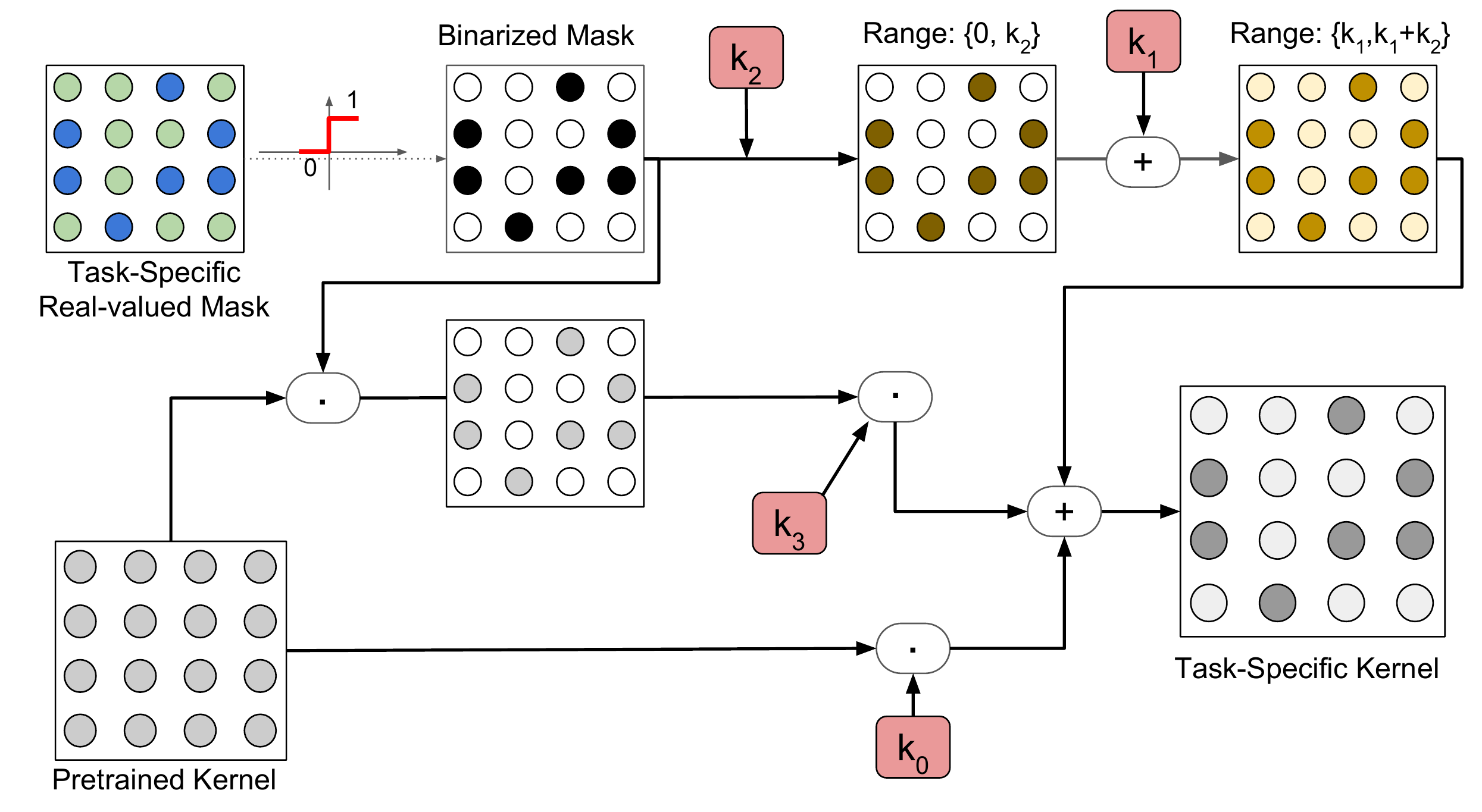}
     \caption{Overview of the proposed BAT model (best viewed in color). Given a convolutional kernel, we exploit a real-valued mask to generate a domain-specific binary mask. 
An affine transformation directly applied to the binary masks, which changes their range (through a scale parameter $k_2$) and their minimum value (through $k_1$). A multiplicative mask applied to the original kernels and the pre-trained kernel themselves are scaled by the factors $k_3$ and $k_0$ respectively. All the different masks are summed to produce the final domain-specific kernel. }
    \label{fig:method-full}
 \end{figure*}

 Each domain-specific network $f_i$ shares the same structure of the baseline network $f_0$, except for having a possibly differently sized classification layer.
{ For each convolutional layer\footnote{Fully-connected layers are a special case.} of $f_0$ with parameters $\mat W$, the domain-specific network $f_i$ holds a binary mask $\mat M$, with the same shape of $\mat W$, that is used to mask original filters. The way the mask is exploited to specialize the network filters produces different variants of our model, which we describe in the following.} 

\subsection{Affine Weight Transformation through Binary Masks}
\label{MT-sec:method}
{Following previous works \cite{mallya2018piggyback}, we consider domain-specific networks $f_i$ that are shaped as the baseline network $f_0$ and we store in $\Omega_i$ a binary mask $\mat M$ for each convolutional kernel $\mat W$ in the shared set $\Theta$. 
However, differently from \cite{mallya2018piggyback}, we consider a more general affine transformation of the base convolutional kernel $\mat W$ that depends on a binary mask $\mat M$ as well as additional parameters. Specifically, we transform $\mat W$ into}
\begin{equation}\label{eq:ours}
\tilde {\mat W}=k_0\mat W+k_1 \mathtt 1+k_2\mat M +k_3\mat W\circ\mat M\,,
\end{equation}
where $k_j\in\mathbb R$ are additional domain-specific parameters in $\Omega_i$ that we learn along with the binary mask $\mat M$, $\mat 1$ is an opportunely sized tensor of $1$s, and $\circ$ is the Hadamard (or element-wise) product.  The transformed parameters $\hat{\mat W}$ are then used in the convolutional layer of $f_i$. 
{We highlight that the domain-specific parameters that are stored in $\Omega_i$ amount to just a single bit per parameter in each convolutional layer plus a few scalars per layer, yielding a low overhead per additional domain while retaining a sufficient degree of freedom to build new convolutional weights. Figure~\ref{fig:method-full} provides an overview of the transformation in \eqref{eq:ours}.
 
Our model, can be regarded as a parametrized generalization of~\cite{mallya2018piggyback}, since we can recover the formulation of~\cite{mallya2018piggyback} by setting $k_{0,1,2}=0$ and $k_3=1$. }
Similarly, if we get rid of the multiplicative component, \ie we set $k_3=0$, we obtained the following transformation
\begin{equation}\label{eq:simple}
\check {\mat W}=k_0\mat W+k_1 \mathtt 1+k_2\mat M\,,
\end{equation}
which corresponds to a simpler but still effective version of our method (presented in \cite{mancini2018adding}) and will be taken into account in our analysis.

{We want to highlight that each model (i.e. \cite{mallya2018piggyback}, BAT, and its simplified version) has different representation capabilities. In fact, in \cite{mallya2018piggyback}, the domain-specific parameters can take only two possible values: either $0$ (i.e. if $m=0$) or the original pre-trained weights (i.e. if $m=1$). On the other hand, the scalar components of our simple model \cite{mancini2018adding} allow both scaling (i.e. with $k_0$) and shifting (i.e. with $k_1$) the original network weights, with the additive binary mask adding a bias term (i.e. $k_2$) selectively to a group of parameters (\ie the one with $m=1$). BAT generalizes \cite{mallya2018piggyback} and \cite{mancini2018adding} by considering the multiplicative binary-mask term $\mat W \circ \mat M$ as an additional bias component scaled by the scalar $k_3$. In this way, our model has the possibility to obtain \textit{parameter-specific} bias components, something that was not possible neither in \cite{mallya2018piggyback} nor in \cite{mancini2018adding}. The additional degrees of freedom makes the search space of our method larger with respect to \cite{mallya2018piggyback,mancini2018adding}, with the possibility to express more complex (and tailored) domain-specific transformations. Thus, as we show in the experimental section, the additional parameters that we introduce with our method bring a negligible per-domain overhead compared to \cite{mallya2018piggyback} and \cite{mancini2018adding}, which is nevertheless generously balanced out by a significant boost of the performance of the domain-specific classifiers.}

{Finally, following \cite{bilen2017universal}, we opt also for domain-specific batch-normalization parameters (\ie mean, variance, scale and bias), unless otherwise stated. 
Those parameters will not be fixed (i.e. they do not belong to $\Theta$) but are part of $\Omega_i$, and thus optimized for each domain. 
In the cases where we have a convolutional layer followed by batch normalization, we keep the corresponding parameter $k_0$ fixed to $1$, because the output of batch normalization is invariant to the scale of the convolutional weights.
}

\subsection{Learning Binary Masks}
{Given the training set of the $i_{th}$ domain, we learn the domain-specific parameters $\Omega_i$ 
by 
minimizing a standard supervised loss, \ie the classification log-loss. However, while the domain-specific batch-normalization parameters can be learned by employing standard stochastic optimization methods, the same is not feasible for the binary masks. Indeed, optimizing the binary masks directly would turn the learning into a combinatorial problem. {To address} this issue, we follow the solution adopted in~\cite{mallya2018piggyback}, \ie we
replace each binary mask $\mat M$ with a thresholded real matrix $\mat R$. By doing so, we shift from optimizing discrete variables in $\mat M$ to continuous ones in $\mat R$. However, the gradient of the hard threshold function $h(r)=1_{r\geq 0}$ is zero almost everywhere, making this solution apparently incompatible with gradient-based optimization approaches. To address this issue we consider a strictly increasing, surrogate function $\tilde h$ that will be used in place of $h$ \emph{only} for the gradient computation, \ie \[
h'(r)\approx \tilde h'(r)\,,
\]
where $h'$ denotes the derivative of $h$ with respect to its argument.
The gradient that we obtain via the surrogate function has the property that it always points in the right down hill direction in the error surface.} {Let $r$ be a single entry of $\mat R$, with $m=h(r)$} and let $E(m)$ be the error function. Then 
\[
\sign((E\circ h)'(r))=\sign(E'(m)h'(r))=\sign\left(E'(m)\tilde h'(r)\right)
\]
and, since $\tilde h'(r)>0$ by construction of $\tilde h$, we obtain the sign agreement
\[
\sign\left((E\circ h)'(r)\right)=\sign\left(E'(m)\right)\,.
\]
Accordingly, when the gradient of $E(h(r))$ with respect to $r$ is positive (negative), this induces a decrease (increase) of $r$. By the monotonicity of $h$ this eventually induces a decrease (increase) of $m$, which is compatible with the direction pointed by the gradient of $E$ with respect to $m$.

{In the experiments, we set $\tilde h(x)=x$, \ie the identity function, recovering the workaround suggested in~\cite{hin12} and employed also in~\cite{mallya2018piggyback}. 
However, other choices are possible. For instance, by taking $\tilde h(x)=(1+e^{-x})^{-1}$, \ie the sigmoid function, we obtain a better approximation that has been suggested in~\cite{goodman1994learning,bengio2013estimating}}. We test different choices for $\tilde h(x)$ in the experimental section.

\subsection{Experimental results}
\myparagraph{Datasets.}
{In the following we test our method on two different multi-task benchmarks, where the multiple tasks regard different classification objectives and/or domains}. For the first benchmark we follow \cite{mallya2018piggyback}, and we use 6 datasets: ImageNet \cite{russakovsky2015imagenet}, VGG-Flowers \cite{nilsback2008automated}, Stanford Cars \cite{krause20133d}, Caltech-UCSD Birds (CUBS) \cite{wah2011caltech}, Sketches \cite{eitz2012humans} and WikiArt \cite{saleh2015large}. VGG-Flowers \cite{nilsback2008automated} is a dataset of fine-grained recognition containing images of 102  categories, corresponding to different kind of flowers. There are 2'040 images for training and 6'149 for testing. Stanford Cars \cite{krause20133d} contains images of 196 different types of cars with approximately 8 thousand images for training and 8 thousands for testing. Caltech-UCSD Birds \cite{wah2011caltech} is another dataset of fine-grained recognition containing images of 200 different species of birds, with approximately 6 thousands images for training and 6 thousands for testing. Sketches \cite{eitz2012humans} is a dataset composed of 20 thousands sketch drawings, 16 thousands for training and 4 thousands for testing. It contains images of 250 different objects in their sketched representations. WikiArt \cite{saleh2015large} contains painting from 195 different artists. The dataset has 42'129 images for training and 10628 images for testing. These datasets contain a lot of variations both from the category addressed (\ie cars \cite{krause20133d} vs birds \cite{wah2011caltech}) and the appearance of their instances (from natural images \cite{russakovsky2015imagenet} to paintings \cite{saleh2015large} and sketches \cite{eitz2012humans}), thus representing a challenging benchmark for sequential multi-task learning techniques. 

The second benchmark 
is the Visual Decathlon Challenge \cite{rebuffi2017learning}. This challenge has been introduced in order to check the capability of a single algorithm to tackle 10 different classification tasks. The tasks are taken from the following datasets: ImageNet \cite{russakovsky2015imagenet}, CIFAR-100 \cite{krizhevsky2009learning}, Aircraft \cite{maji2013fine}, Daimler pedestrian classification (DP) \cite{munder2006experimental}, Describable textures (DTD) \cite{cimpoi2014describing}, German traffic signs (GTS) \cite{stallkamp2012man} , Omniglot \cite{lake2015human}, SVHN \cite{netzer2011reading}, UCF101 Dynamic Images \cite{bilen2016dynamic,soomro2012ucf101} and VGG-Flowers \cite{nilsback2008automated}. A more detailed description of the challenge and the datasets can be found in \cite{rebuffi2017learning}. For this challenge, an independent scoring function is defined \cite{rebuffi2017learning}. This function $S$ is expressed as:
\begin{equation}\label{MT-eq:S}
S=\sum^{10}_{d=1}\alpha_d\text{max}\{0,E_d^{\text{max}}-E_d\}^{2}
\end{equation}
where $E_d^{\text{max}}$ is the test error of the baseline in the domain $d$, $E_d$ is the test error of the submitted model and $\alpha$ is a scaling parameter ensuring that the perfect score for each task is 1000, thus with a maximum score of 10000 for the whole challenge. The baseline error is computed doubling the error of 10 independent models fine-tuned on the single tasks.
This score function takes into account the performances of a model on all 10 classes, preferring models with good performances on all of them compared to models outperforming by a large margin the baseline in just few of them. {Following \cite{berriel2019budget}, we use this metric also for the first benchmark, keeping the same upper-bound of 1000 points for each task. Moreover, as in \cite{berriel2019budget}, we report the ratio among the score obtained and the parameters used, denoting it as ${S}_p$. This metric allows to capture the trade-off among the performances and model size.}

\myparagraph{Networks and training protocols.} For the first benchmark, we use 3 networks: ResNet-50 \cite{he2016deep},  DenseNet-121 \cite{huang2017densely} and VGG-16 \cite{simonyan2014very}, reporting the results of Piggyback \cite{mallya2018piggyback}, PackNet \cite{mallya2018packnet} and both the simple \cite{mancini2018adding} and full version of our model (BAT). 

Following the protocol of \cite{mallya2018piggyback}, for all the models we start from the networks pre-trained on ImageNet and train the task-specific networks using Adam \cite{kingma2014adam} as optimizer except for the classifiers where SGD \cite{bottou2010large} with momentum is used. The networks are trained with a batch size of 32 and an initial learning rate of 0.0001 for Adam and 0.001 for SGD with momentum 0.9. Both the learning rates are decayed by a factor of 10 after 15 epochs. In this scenario we use input images of size $224\times224$ pixels, with the same data augmentation (\ie mirroring and random rescaling) of \cite{mallya2018packnet,mallya2018piggyback}. The real-valued masks are initialized with random values drawn from a uniform distribution with values between $0.0001$ and $0.0002$. Since our model is independent on the order of the tasks, we do not take into account different possible orders, reporting the results as accuracy averaged across multiple runs. For simplicity, in the following we will denote this scenario as \textit{ImageNet-to-Sketch}.

For the Visual Decathlon we employ the Wide ResNet-28 \cite{zagoruyko2016wide} adopted by previous methods \cite{rebuffi2017learning,rosenfeld2017incremental,mallya2018piggyback}, with a widening factor of 4 (\ie 64, 128 and 256 channels in each residual block). Following \cite{rebuffi2017learning} we rescale the input images to $72\times72$ pixels giving as input to the network images cropped to $64\times64$. We follow the protocol in \cite{mallya2018piggyback}, by training the simple and full versions of our model for 60 epochs for each task, with a batch size of 32, and using again Adam for the entire architecture but the classifier, where SGD with momentum is used. The same learning rates of the first benchmark are adopted and are decayed by a factor of 10 after 45 epochs. The same initialization scheme is used for the real-valued masks. \textit{No hyperparameter tuning} has been performed as we used a single training schedule for all the 10 tasks, except for the ImageNet pre-trained model, which was trained following the schedule of \cite{rebuffi2017learning}. As for data augmentation, mirroring has been performed, except for the datasets with digits (\ie SVHN), signs (Omniglot, GTS) and textures (\ie DTD) as it may be rather harmful (as in the first 2 cases) or unnecessary.

{In both benchmarks, we train our network on one task at the time, sequentially for all tasks. For each task we introduce the task specific binary masks and additional scalar parameters, as described in section \ref{MT-sec:method}. Moreover, following previous approaches \cite{rebuffi2017learning,rebuffi2018efficient,mallya2018piggyback,rosenfeld2017incremental}, we consider a separate classification layers for each task. This is reflected also in the computation of the parameters overhead required by our model, we do not consider the separate classification layers, following comparison systems \cite{rebuffi2017learning,rebuffi2018efficient,mallya2018piggyback,rosenfeld2017incremental}.}

\subsubsection{Results}
\myparagraph{ImageNet-to-Sketch.} In the following we discuss the results obtained by our model on the ImageNet-to-Sketch scenario. We compare our method with Piggyback \cite{mallya2018piggyback}, PackNet \cite{mallya2018packnet} and two baselines considering the network only as feature extractor, training only the task-specific classifier, and individual networks separately fine-tuned on each task. PackNet \cite{mallya2018packnet} adds a new task to an architecture by identifying important weights for the task, optimizing the architecture through alternated pruning and re-training steps. Since this algorithm is dependent on the order of the task, we report the performances for two different orderings \cite{mallya2018piggyback}: starting from the model pre-trained on ImageNet, in the first setting ($\downarrow$) the order is CUBS-Cars-Flowers-WikiArt-Sketch while for the second ($\uparrow$) the order is reversed. For our model, we evaluate both the full and the simple version, including task-specific batch-normalization layers. Since including batch-normalizatin layers affects the performances, for the sake of presenting a fair comparison, we report also the results of Piggyback \cite{mallya2018piggyback} obtained as a special case of our model with separate BN parameters per task for ResNet-50 and DenseNet-121. {Moreover, we report the results of the Budget-Aware adapters ($\text{BA}^2$) method in \cite{berriel2019budget}. This method relies on binary masks applied not per-parameter but per-channel, with a budget constraint allowing to further squeeze the network complexity. As in our method, also in \cite{berriel2019budget} task-specific BN layers are used.}

Results are reported in Tables \ref{MT-tab:resnet-ImageNet}, \ref{MT-tab:densenet-ImageNet} and \ref{MT-tab:vgg-ImageNet}. We see that both versions of our model are able to fill the gap between the classifier only baseline and the individual fine-tuned architectures, almost entirely in all settings. For larger and more diverse datasets such as Sketch and WikiArt, the gap is not completely covered, but the distance between our models and the individual architectures is always less than 1\%. These results are remarkable given the simplicity of our method, not involving any assumption of the optimal weights per task \cite{mallya2018packnet,li2017learning}, and the small overhead in terms of parameters that we report in the row "\#~Params" (\ie $1.17$ for ResNet-50, $1.21$ for DenseNet-121 and $1.16$ for VGG-16), which represents the total number of parameters (counting all tasks and excluding the classifiers) relative to the ones in the baseline network\footnote{If the base architecture contains $N_p$ parameters and the additional bits introduced per task are $A_p$ then ${\text{\#~Params}=1+\frac{A_p\cdot (T-1)}{32\cdot N_p}}$,
where $T$ denotes the number of tasks (included the one used for pre-training the network) and the 32 factor comes from the bits required for each real number. The classifiers are not included in the computation.}.

For what concerns the comparison with the other algorithms, our model consistently outperforms both the basic version of Piggyback and PackNet in all the settings and architectures, with the exception of Sketch for the DenseNet and VGG-16 architectures and CUBS for VGG-16, in which the performances are comparable with those of Piggyback. When task-specific BN parameters are introduced also for Piggyback (Tables \ref{MT-tab:resnet-ImageNet} and \ref{MT-tab:densenet-ImageNet}), the gap in performances is reduced, with comparable performances in some settings (\ie CUBS) but with still large gaps in others (\ie Flowers, Stanford Cars and WikiArt). These results show that the advantages of our model are not only due to the additional BN parameters, but also to the more flexible and powerful affine transformation introduced. 

This statement is further confirmed with the VGG-16 experiments of Table \ref{MT-tab:vgg-ImageNet}.  For this network, when the standard Piggyback model is already able to fill the gap between the feature extractor baseline and the individual architectures, our model achieves either comparable or slightly superior performances (\ie CUBS, WikiArt and Sketch). However in the scenarios where Piggyback does not reach the performances of the independently fine-tuned models (\ie Stanford Cars and Flowers), our model consistently outperform the baseline, either halving (Flowers) or removing (Stanford Cars) the remained gap. Since this network does not contain batch-normalization layers, it confirms the generality of our model, showing the advantages of both our simple and full versions, even without task-specific BN layers. 

{For what concerns the comparison with $\text{BA}^2$, the performances of our model are either comparable or superior in most of the settings. Remarkable are the gaps in the WikiArt dataset, with our full model surpassing $\text{BA}^2$ by 3\% with ResNet-50 and 4\% for DenseNet-121. Despite both Piggyback and $\text{BA}^2$ use fewer parameters than our approach, our full model outperforms both of them in terms of the final score (Score row) and the ratio among the score and the parameters used (Score/Params row). This shows that our model is the most powerful in making use of the binary masks, achieving not only higher performances but also a more favorable trade-off with the model size.} 

Finally, both Piggyback, $\text{BA}^2$ and our model outperform PackNet and, as opposed to the latter method, do not suffer from the heavily dependence on the ordering of the tasks. This advantage stems from having a sequential multi-task learning strategy that is task independent, with the base network not affected by the new tasks that are learned.

\begin{table*}[t]
			\caption{Accuracy of ResNet-50 architectures in the ImageNet-to-Sketch scenario.} 
                   
		\centering
        \scalebox{.75}{
		\begin{tabular}{ l | c || ec | ec | ec | ec | ec | ec | ec || c  } 
			\multirow{2}{*}{Dataset} & Classifier & \multicolumn{2}{ c | }{PackNet\cite{mallya2018piggyback}} &\multicolumn{2}{c |}{Piggyback}& $\text{BA}^2$ & \multicolumn{2}{ c ||}{BAT} & Individual\\
          & Only \cite{mallya2018piggyback} &$\downarrow$&$\uparrow$ & \cite{mallya2018piggyback}&BN& \cite{berriel2019budget}&Simple& Full  & \cite{mallya2018piggyback}\\ \hline
             \# Params&1&\multicolumn{2}{c |}{1.10}&1.16&1.17& 1.03&1.17&1.17&6\\  \hline
       ImageNet &76.2&75.7&75.7&\textbf{76.2} &\textbf{76.2}&\textbf{76.2}&\textbf{76.2}&\textbf{76.2} &76.2\\
       CUBS&70.7 &80.4&71.4&80.4 &82.1&81.2&\textbf{82.6} &82.4&82.8 \\
       Stanford Cars&52.8&86.1&80.0 &88.1 &90.6&\textbf{92.1}&{91.5}&91.4&91.8 \\
       Flowers&86.0&93.0&90.6&93.5&95.2&95.7&96.5&\textbf{96.7} &96.6\\
       WikiArt&55.6&69.4&70.3&73.4 &74.1&72.3&74.8&\textbf{75.3}&75.6\\
       Sketch&50.9&76.2&78.7&79.4 &79.4&79.3&\textbf{80.2}&\textbf{80.2} &80.8 \\
            \hline
        Score&533&732&620&934&1184&1265&1430&\textbf{1458}&1500\\
        Score/Params&533&665&534&805&1012&1228&1222&\textbf{1246}&250\\
		\end{tabular}
        }
		\label{MT-tab:resnet-ImageNet}
\end{table*}
\begin{table*}[t]
			\caption{Accuracy of DenseNet-121 architectures in the ImageNet-to-Sketch scenario.} 
\centering
        \scalebox{.75}{
		\begin{tabular}{ l | c || ec | ec| ec | ec | ec | ec | ec || c  } 
			\multirow{2}{*}{Dataset} & Classifier &  \multicolumn{2}{c |}{PackNet\cite{mallya2018piggyback}}&\multicolumn{2}{c |}{Piggyback} & $\text{BA}^2$ &\multicolumn{2}{c ||}{BAT} & Individual\\
          & Only \cite{mallya2018piggyback}&$\downarrow$ & $\uparrow$&\cite{mallya2018piggyback} &BN&\cite{berriel2019budget}&Simple&Full & \cite{mallya2018piggyback}\\ 
            \hline
           \# Params&1&\multicolumn{2}{c |}{1.11}&1.15& 1.21 &1.17&1.21&1.21&6 \\  \hline
       ImageNet &74.4 &\textbf{74.4}&\textbf{74.4}&\textbf{74.4} &\textbf{74.4}&\textbf{74.4} & \textbf{74.4}& \textbf{74.4}&74.4\\
       CUBS &73.5 &80.7&69.6&79.7 &81.4&\textbf{82.4}&81.5&{81.7}&81.9\\
       Stanford Cars&56.8 &84.7&77.9&87.2 &90.1&\textbf{92.9}&{91.7} &91.6&91.4\\
       Flowers&83.4 &91.1&91.5&94.3 &95.5&96.0&96.7&\textbf{96.9} &96.5\\
       WikiArt&54.9 &66.3&69.2 &72.0&73.9&71.5&75.5&\textbf{75.7} &76.4\\
       Sketch&53.1 &74.7&78.9&\textbf{80.0} &79.1&79.9&79.9&79.8 &80.5 \\
       \hline
        Score&324&685&607&946&1209&1434&1506&\textbf{1534}&1500\\
        Score/Params&324&617&547&822&999&1226&1245&\textbf{1268}&250\\
		\end{tabular}}
		\label{MT-tab:densenet-ImageNet}
\end{table*}

\begin{table*}[t]
			\caption{Accuracy of VGG-16 architectures in the ImageNet-to-Sketch scenario.}       
		\centering
        \scalebox{.9}{
		\begin{tabular}{ l | c || ec | ec | c | ec | ec || c  } 
			\multirow{2}{*}{Dataset} & Classifier & \multicolumn{2}{ c | }{PackNet\cite{mallya2018piggyback}} &Piggyback & \multicolumn{2}{ c ||}{BAT} & Individual\\
          & Only \cite{mallya2018piggyback} &$\downarrow$&$\uparrow$ & \cite{mallya2018piggyback}& Simple & Full  & \cite{mallya2018piggyback}\\ \hline
             \# Params&1&\multicolumn{2}{c |}{1.09}&1.16&1.16&1.16&6\\\hline
       ImageNet &71.6&70.7&70.7&\textbf{71.6} &\textbf{71.6}&\textbf{71.6} &71.6\\
       CUBS&63.5 &77.7&70.3&\textbf{77.8} &77.4 &77.4&77.4 \\
       Stanford Cars&45.3&84.2&78.3 &86.1&87.2&\textbf{87.3}&87.0 \\
       Flowers&80.6&89.7&89.8&90.7&\textbf{91.6}&91.5&92.3\\
       WikiArt&50.5&67.2&68.5&71.2 &71.6&\textbf{71.9}&67.7\\
       Sketch&41.5&71.4&75.1&76.5&76.5&\textbf{76.7 }&76.4 \\
            \hline
        Score&342&1152&979&1441&1530&\textbf{1538}&1500\\
        Score/Params&342&1057&898&1243&1319&\textbf{1326}&250\\
		\end{tabular}
        }
		\label{MT-tab:vgg-ImageNet}
\end{table*}

\myparagraph{Visual Decathlon Challenge.}
In this section we report the results obtained on the Visual Decathlon Challenge. {We compare our model with the baseline method Piggyback \cite{mallya2018piggyback} (PB), the budget-aware adapters of \cite{berriel2019budget} ($\text{BA}^2$), the improved version of the winner entry of the 2017 edition of the challenge \cite{rosenfeld2017incremental} (DAN), the network with task-specific parallel adapters \cite{rebuffi2018efficient} (PA), the task-specific attention modules of \cite{liu2019end} (MTAN), the covariance normalization approach \cite{li2019efficient} (CovNorm) and SpotTune \cite{guo2019spottune}}. We additionally report the baselines proposed by the authors of the challenge \cite{rebuffi2017learning}. For the latter, we report the results of 5 models: the network used as feature extractor (Feature), 10 different models fine-tuned on each single task (Fine-tune), the network with task-specific residual adapter modules \cite{rebuffi2017learning} (RA), the same model with increased weight decay (RA-decay) and the same architecture jointly trained on all 10 tasks, in a round-robin fashion (RA-joint). The first two models are considered as references. For the parallel adapters approach \cite{rebuffi2018efficient} we report also the version with a post training low-rank decomposition of the adapters (PA-SVD). This approach extracts a task specific and a task agnostic component from the learned adapters with the task specific components which are further fine-tuned on each task. Additionally we report the novel results of the residual adapters \cite{rebuffi2017learning} as reported in \cite{rebuffi2018efficient} (RA-N).

Similarly to \cite{rosenfeld2017incremental} we tune the training schedule, jointly for the 10 tasks, using the validation set, and evaluate the results obtained on the test set (via the challenge evaluation server) by a model trained on the union of the training and validation sets, using the validated schedule. As opposed to methods like~\cite{rebuffi2017learning} we use the same schedule for the 9 tasks (except for the baseline pre-trained on ImageNet), without adopting task-specific strategies for setting the hyperparameters. Moreover, we do not employ our algorithm while pre-training the ImageNet architecture as in~\cite{rebuffi2017learning}. For fairness, we additionally report the results obtained by our implementation of \cite{mallya2018piggyback} using the same pre-trained model, training schedule and data augmentation adopted for our algorithm (PB ours).

The results are reported in Table 
\ref{MT-tab:vdc-accuracy} 
{in terms of the $S$-score (see, Eq. \eqref{MT-eq:S}) and $S_p$}. In the first part of the table are shown the baselines (\ie fine-tuned architectures and using the network as feature extractor) while in the middle the sequential learning models against which we compare. In the last part of the table we report, for fairness, the methods that do not consider a sequential learning setting since they either train on all the datasets jointly (RA-joint) or have a multi-process step considering the all tasks (PA-SVD).

From the table we can see that the full form of our model (F) {achieves very high results, being the third best performing method in terms of $S$-score, behind only CovNorm  and SpotTune and being comparable to PA. However, SpotTune uses a large amount of parameters (11x) and PA doubles the parameters of the original model. CovNorm uses a very low number of parameters, but requires a two-stage pipeline. On the other hand, our model does not require neither a large number of parameters (such as SpotTune and PA) nor a two-stage pipeline (as CovNorm) while achieving results close to the state of the art (215 points below CovNorm in terms of $S$-score). Compared to binary mask based approaches, our model surpasses PiggyBack of more than 600 points, $\text{BA}^2$ of 300 and BAT simple of more than 200. It is worth highlighting that these results have been achieved \textit{without task-specific hyperparameter tuning}, differently from previous works e.g. \cite{rebuffi2017learning,rebuffi2018efficient,li2019efficient}.

Analyzing the $S_p$ score, BAT is the third best performing model, behind $\text{BA}^2$ and CovNorm. We highlight however that CovNorm requires a two-stage pipeline to reduce the amount of parameters needed, while $\text{BA}^2$ is explicitly designed with the purpose of limiting the budget (i.e. parameters, flops) required by the model.} 

\begin{table*}[t]
			\caption{Results in terms of $S$ and $S_p$ scores for the Visual Decathlon Challenge.} 
            
		\centering
		\setlength{\tabcolsep}{4pt}
        \scalebox{.82}{
		\begin{tabular}{ l | c || c  c  c  c  c  c  c  c  c  c | c c } 
			 Method&\#Par&ImN&Airc &C100&DP&DTD&GTS&Flw&Ogl&SVHN&UCF& Score & ${S}_p$\\\hline
            Feature \cite{rebuffi2017learning}&1&59.7	&23.3	&63.1	&80.3	&45.4	&68.2	&73.7	&58.8	&43.5	&26.8	&544&544\\
            Fine-tune \cite{rebuffi2017learning}&10&59.9	&60.3	&82.1	&92.8	&55.5	&97.5	&81.4	&87.7	&96.6	&51.2	&2500&250\\
      \hline     RA\cite{rebuffi2017learning}&2&59.7	&56.7	&81.2	&93.9	&50.9	&97.1	&66.2	&{89.6}	&96.1	&47.5	&2118&1059\\
           RA-decay\cite{rebuffi2017learning}&2&59.7	&61.9	&81.2	&93.9	&57.1	&97.6	&81.7	&{89.6}	&96.1	&{50.1}	&2621&1311\\
           RA-N\cite{rebuffi2018efficient}&2&{60.3}&61.9&81.2&93.9&57.1&{99.3}&81.7&{89.6}&96.6&{50.1}&3159&1580\\
            DAN \cite{rosenfeld2017incremental}&2.17&57.7	&64.1	&80.1	&91.3	&56.5	&98.5	&86.1	&{89.7}	&{96.8}	&49.4	&2852&1314\\
             PA \cite{rebuffi2018efficient}&2&{60.3}&{64.2}&81.9&94.7&58.8&{99.4}&84.7&89.2&96.5&{50.9}&{3412}&1706\\
            MTAN \cite{liu2019end}&1.74&{63.9}&{61.8}&81.6&91.6&56.4&{98.8}&81.0&89.8&96.9&{50.6}&{2941}&1690\\
            SpotTune \cite{guo2019spottune} & 11 & 60.3 &63.9 &80.5&96.5&57.1&99.5&85.2&88.8&96.7&52.3&{3612}&328\\
            CovNorm \cite{li2019efficient} &1.25&60.4&69.4&81.3&98.8&60.0&99.1&83.4&87.7&96.6&48.9&\textbf{3713}&{2970}\\
            PB \cite{mallya2018piggyback}&1.28&57.7	&{65.3}	&79.9	&{97.0}	&57.5	&97.3	&79.1	&87.6	&{97.2}	&47.5	&2838&2217\\
            PB ours &1.28&{60.8}&52.3&80.0&95.1&{59.6}&98.7&82.9&85.1&96.7&46.9&2805&2191\\
            $\text{BA}^2$ \cite{berriel2019budget}&1.03&56.9	&49.4&78.1&95.5&55.1&99.4&86.1&88.7&96.9&50.2&3199&\textbf{3106}\\
          	BAT (S) \cite{mancini2018adding}&1.29&{60.8}	&51.3	&{81.9}	&94.7	&{59.0}	&99.1	&{88.0}	&89.3	&96.5	&48.7	&3263&2529\\
            BAT (F)&1.29&{60.8}	&52.8	&{82.0}	&{96.2}	&58.7	&{99.2}	&{88.2}	&89.2	&{96.8}	&48.6	&{3497}&{2711}\\
            \hline
            PA-SVD\cite{rebuffi2018efficient}&1.5&60.3&66.0&81.9&94.2&57.8&99.2&85.7&89.3&96.6&52.5&3398&2265\\
            RA-joint\cite{rebuffi2017learning}&2&59.2	&63.7	&81.3	&93.3	&57.0	&97.5	&83.4	&89.8	&96.2	&50.3	&2643&1322\\
            \hline
             
		\end{tabular}}
		\label{MT-tab:vdc-accuracy}
\end{table*}

\myparagraph{Ablation Study}
\label{MT-sec:ablation}

In the following we will analyze the impact of the various components of our model. In particular we consider the impact of the parameters $k_0$, $k_1$, $k_2$, $k_3$ and the surrogate function $\tilde{h}$ on the final results of our model for the ResNet-50 and DenseNet-121 architectures in the ImageNet-to-Sketch scenario. Since the architectures contains batch-normalization layers, we set $k_0=1$ for our simple and full versions and $k_0=0$ when we analyze the special case \cite{mallya2018piggyback}. For the other parameters we adopt various choices: either we fix them to a constant in order not take into account their impact, or we train them, to assess their particular contribution to the model. The surrogate function we use is the identity function $\tilde{h}(x)=x$, unless otherwise stated (\ie \textit{with Sigmoid}). The results of our analysis are shown in Tables \ref{MT-tab:ablation-res} and \ref{MT-tab:ablation-dense}. 

As the Tables shows, while the BN parameters allow a boost in the performances of Piggyback, adding $k_1$ to the model does not provide further gain in performances. This does not happen for the simple version of our model: without $k_1$ our model is not able to fully exploit the presence of the binary masks, achieving comparable or even lower performances with respect to the Piggyback model. We also notice that a similar drop affecting our \emph{Simple} version when bias was omitted.  

Noticeable, the full versions with $k_2=0$ suffer a large decrease in performances in almost all settings (\eg ResNet-50 Flowers from 96.7\% to 91.0\%), showing that the component that brings the largest benefits to our algorithm is the addition of the binary mask itself scaled by $k_2$ (\ie $k_2 \cdot \mat M$). This explains also the reason why the simple version achieves a performance similar to the full version of our model. We finally note that there is a limited contribution brought by the standard Piggyback component (\ie $k_1\cdot \mat W\circ \mat M$), compared to the new components that we have introduced in the transformation:
in fact, there is a clear drop in performance in various scenarios (\eg CUBS, Cars) when we set either $k_1=0$ or $k_2=0$, thus highlighting the importance of those components. %
Consequently, as $k_1$ is introduced in our \emph{Simple} model, the boost of performances is significant such that neither the inclusion of $k_3$, nor considering channel-wise parameters $k_1$ provide further gains. Slightly better results are achieved in a larger datasets, such as WikiArt, with the additional parameters giving more capacity to the model, thus better handling the larger amount of information available in the dataset. 

As to what concerns the choice of the surrogate $\tilde{h}$, no particular advantage has been noted when $\tilde{h}(x)=\sigma(x)$ with respect to the standard straight-through estimator ($\tilde{h}(x)=x$). This may be caused by the noisy nature of the straight-through estimator, which has the positive effect of regularizing the parameters, as shown in previous works \cite{bengio2013estimating,neelakantan2015adding}.

We also note that for DenseNet-121, as opposed to ResNet-50, setting $k_1$ to zero degrades the performance only in 1 out of 5 datasets (\ie CUBS) while the other 4 are not affected, showing that the effectiveness of different components of the model is also dependent on the architecture used.

\begin{table*}[t]
			\caption{Impact of the parameters $k_0$, $k_1$, $k_2$ and $k_3$ of our model using the ResNet-50 architectures in the ImageNet-to-Sketch scenario. \ding{51} denotes a learned parameter, while~$^*$ denotes \cite{mallya2018piggyback} obtained as a special case of our model.} 

		\centering
		\scalebox{.85}{
		\begin{tabular}{ l | c | c | c | c | c | c | c | c | c |  } 
        Method &$k_0$ &$k_1$ & $k_2$&$k_3$& CUBS & CARS & Flowers & WikiArt & Sketch\\
         \hline
Piggyback \cite{mallya2018piggyback}&0&0&0&1&80.4&88.1&93.6&73.4&79.4\\
\hline
Piggyback$^*$ 
&0&0&0&1&80.4&87.8&93.1&72.5&78.6\\ 
Piggyback$^*$ with BN & 0&0&0&1 &82.1&90.6&95.2&74.1&79.4\\ 
Piggyback$^*$ with BN
&0&\ding{51}&0&1&81.9&89.9&94.8&73.7&79.9\\ 
        BAT (Simple, no bias)
&1&0&\ding{51}&0&80.8&90.3&96.1&73.5&80.0\\ 
        BAT (Simple) \cite{mancini2018adding}&1&\ding{51}&\ding{51}&0&82.6&91.5&96.5&74.8&80.2\\
        BAT (Simple with Sigmoid) &1&\ding{51}&\ding{51}&0&82.6&91.4&96.4&75.2&80.2\\
        BAT (Full, no bias)&1&0&\ding{51}&\ding{51}&80.7&90.2&96.0&72.0&78.8\\
        BAT (Full, no $k_2$)&1&\ding{51}&0&\ding{51}&80.6&87.5&91.0&73.0&78.4\\
        BAT (Full)&1&\ding{51}&\ding{51}&\ding{51}&82.4&91.4&96.7&75.3&80.2\\
         BAT (Full with Sigmoid)&1&\ding{51}&\ding{51}&\ding{51}&82.7&91.4&96.6&75.2& 80.2\\
        BAT (Full, channel-wise) &1&\ding{51}&\ding{51}&\ding{51}&82.0&91.0&96.3&74.8&80.0\\
        
			\hline
             
		\end{tabular}}
		\label{MT-tab:ablation-res}
\end{table*}

\begin{table*}[t]
			\caption{Impact of the parameters $k_0$, $k_1$, $k_2$ and $k_3$ of our model using the DenseNet-121 architectures in the ImageNet-to-Sketch scenario. \ding{51} denotes a learned parameter, while~$^*$ denotes \cite{mallya2018piggyback} obtained as a special case of our model.}

		\centering
		\scalebox{.85}{
		\begin{tabular}{ l | c | c | c | c | c | c | c | c | c |  } 
        Method &$k_0$ &$k_1$ & $k_2$&$k_3$& CUBS & CARS & Flowers & WikiArt & Sketch\\
         \hline
Piggyback \cite{mallya2018piggyback}&0&0&0&1&79.7&87.2&94.3&72.0&80.0\\
\hline
Piggyback$^*$ 
&0&0&0&1&80.0&86.6&94.4&71.9&78.7\\ 
Piggyback$^*$ with BN & 0&0&0&1 &81.4&90.1&95.5&73.9&79.1\\ 
Piggyback$^*$ with BN
&0&\ding{51}&0&1&81.9&90.1&95.4&72.6&79.9\\ 
        BAT (Simple, no bias)
&1&0&\ding{51}&0&80.4&91.4&96.7&75.0&79.7\\ 
        BAT (Simple) \cite{mancini2018adding}&1&\ding{51}&\ding{51}&0&81.5&91.7&96.7&75.5&79.9\\
        BAT (Simple with Sigmoid) &1&\ding{51}&\ding{51}&0&81.5&91.7&97.0&76.0&79.8\\
        BAT (Full, no bias)&1&0&\ding{51}&\ding{51}&80.2&91.1&96.5&75.1&79.2\\
        BAT (Full, no $k_2$)&1&\ding{51}&0&\ding{51}&79.8&87.2&91.8&73.2&78.1\\
        BAT (Full) \cite{mancini2020boostingmva}&1&\ding{51}&\ding{51}&\ding{51}&81.7&91.6&96.9&75.7&79.9\\
         BAT (Full with Sigmoid)&1&\ding{51}&\ding{51}&\ding{51}&82.0&91.7&97.0&76.0&79.9\\
        BAT (Full, channel-wise) &1&\ding{51}&\ding{51}&\ding{51}&81.4&91.6&96.5&75.5&79.9\\
        
			\hline
             
		\end{tabular}}
		\label{MT-tab:ablation-dense}
\end{table*}

\begin{figure*}[!b]
\centering
 \includegraphics[width=1.\textwidth,trim=2cm 0 1.5cm 0,clip]{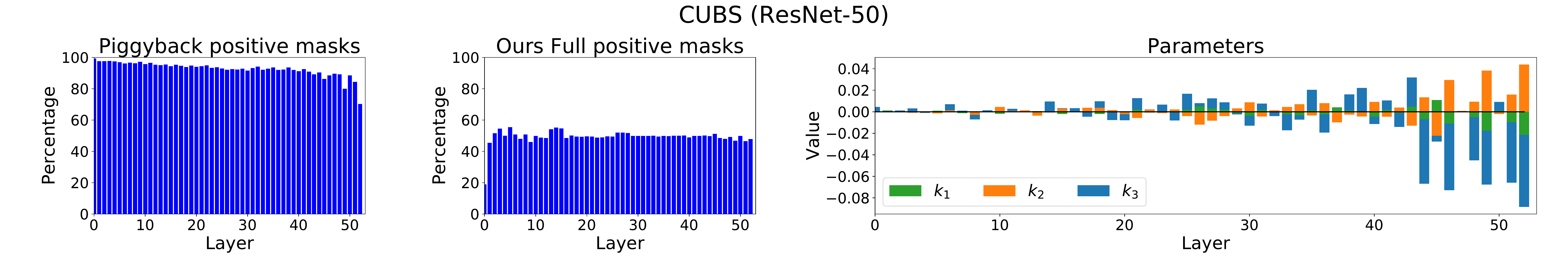}  \\
\includegraphics[width=1.\textwidth,trim=2cm 0 1.5cm 0,clip]{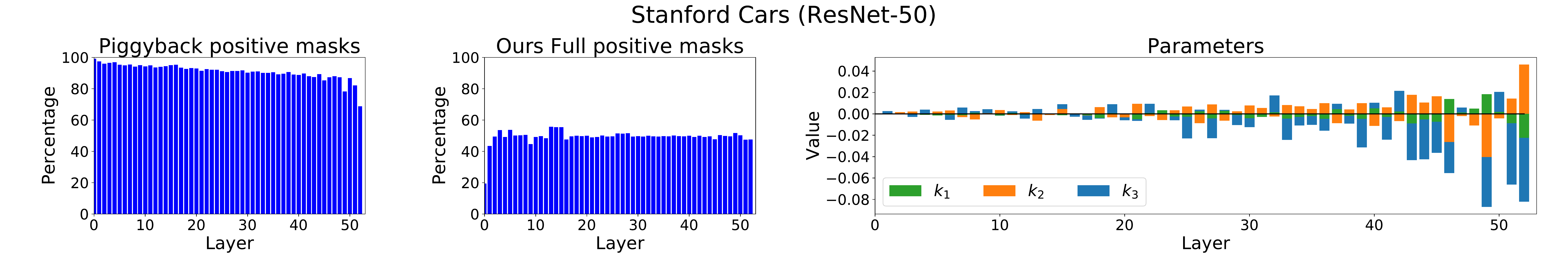} \\
\includegraphics[width=1.\textwidth,trim=2cm 0 1.5cm 0,clip]{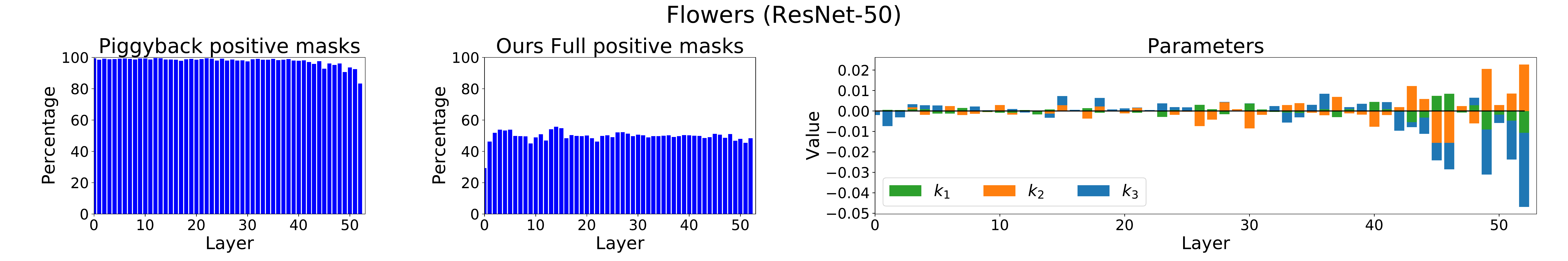} \\
\includegraphics[width=1.\textwidth,trim=2cm 0 1.5cm 0,clip]{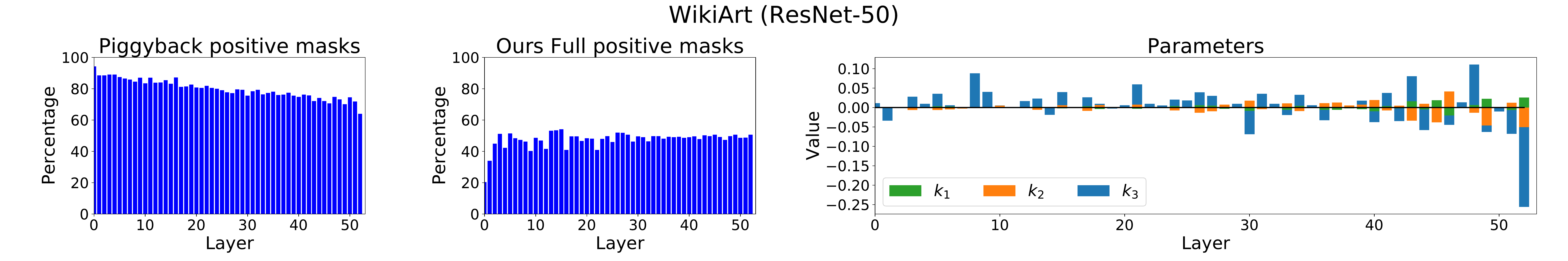} \\
\includegraphics[width=1.\textwidth,trim=2cm 0 1.5cm 0,clip]{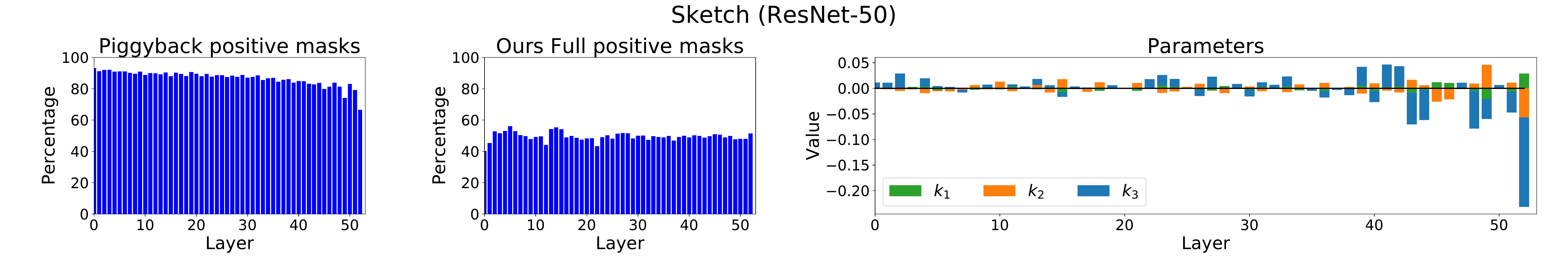} \\
  \caption{Percentage of 1s in the binary masks at different layers depth for Piggyback (left) and our full model (center) and values  of the parameters $k_1$, $k_2$, $k_3$ computed by our full model (right) for all datasets of the Imagenet-to-Sketch benchmark and the ResNet-50 architecture.
  }
  \label{MT-fig:res-params}
  \end{figure*}

\begin{figure*}[!b]
\centering
   \includegraphics[width=1.\textwidth,trim=2cm 0 1.5cm 0,clip]{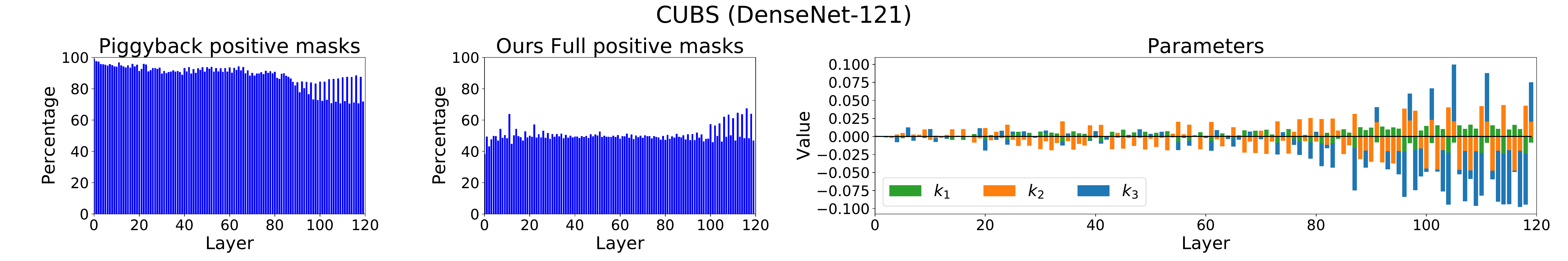}  \\
\includegraphics[width=1.\textwidth,trim=2cm 0 1.5cm 0,clip]{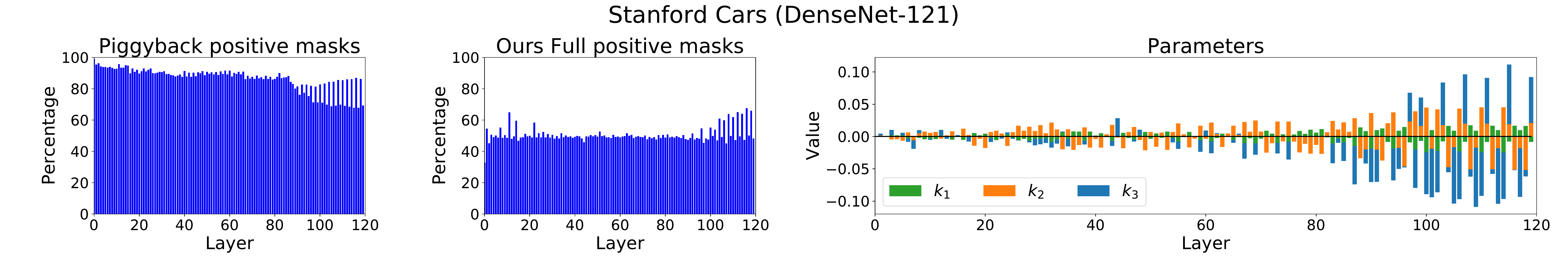} \\
\includegraphics[width=1.\textwidth,trim=2cm 0 1.5cm 0,clip]{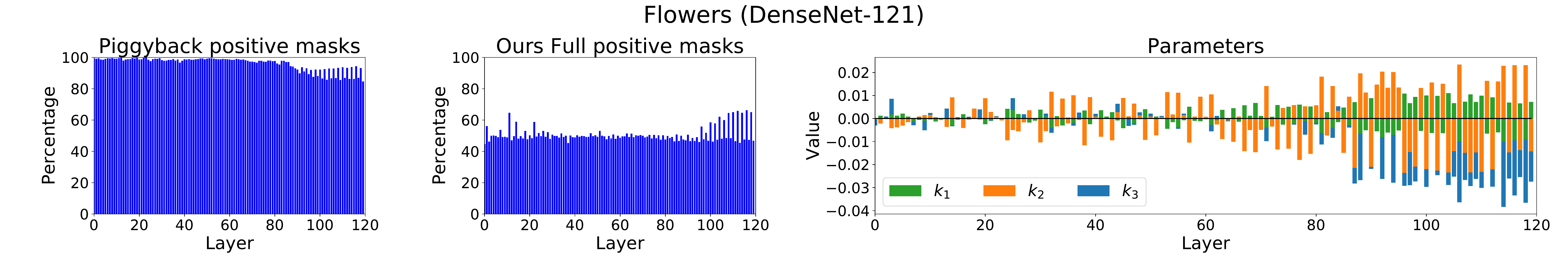} \\
\includegraphics[width=1.\textwidth,trim=2cm 0 1.5cm 0,clip]{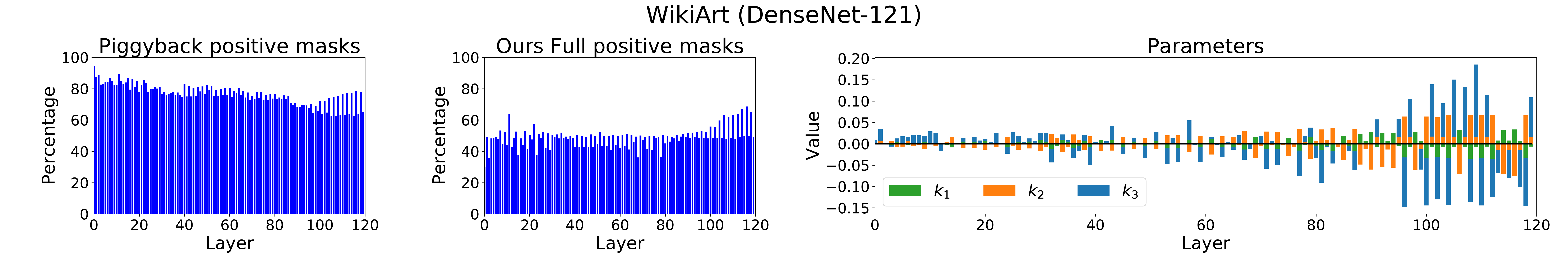} \\
\includegraphics[width=1.\textwidth,trim=2cm 0 1.5cm 0,clip]{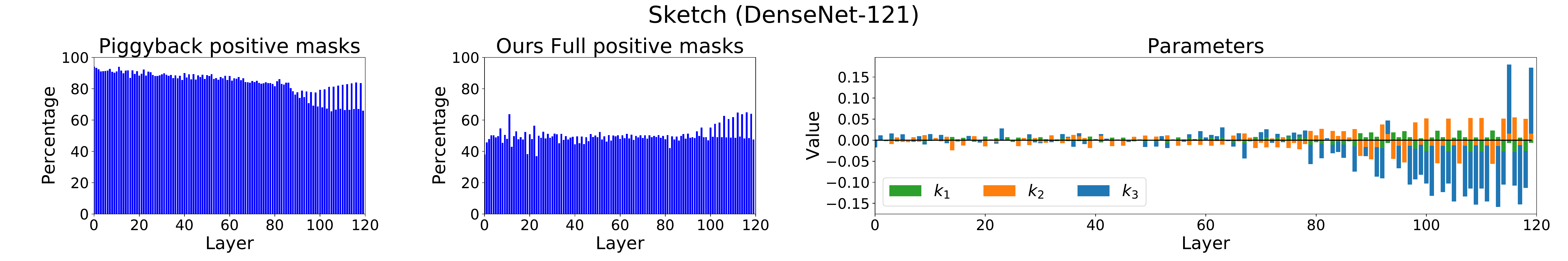} \\
  \caption{Percentage of 1s in the binary masks at different layers depth for Piggyback (left) and our full BAT model (center, \textit{ours}) and values  of the parameters $k_1$, $k_2$, $k_3$ computed by our full model (right) for all datasets of the Imagenet-to-Sketch benchmark with the DenseNet-121 architecture.
  }
    \label{MT-fig:dense-params}
  \end{figure*}

\myparagraph{Parameter Analysis}
We analyze the values of the parameters $k_1$, $k_2$ and $k_3$ of one instance of our full model in the ImageNet-to-Sketch benchmark. We use both the architectures employed in that scenario (\ie ResNet-50 and DenseNet-121) and we plot the values of  $k_1$, $k_2$ and $k_3$ as well as the percentage of 1s present inside the binary masks for different layers of the architectures. Together with those values we report the percentage of 1s for the masks obtained through our implementation of Piggyback. Both the models have been trained considering task-specific batch-normalization parameters. The results are shown in Figures \ref{MT-fig:res-params} and \ref{MT-fig:dense-params}. In all scenarios our model keeps almost half of the masks active across the whole architecture. Compared to the masks obtained by Piggyback, there are 2 differences: 1) Piggyback exhibits denser masks (\ie with a larger portion of 1s), 2) the density of the masks in Piggyback tends to decreases as the depth of the layer increases. Both these aspects may be linked to the nature of our model: by having more flexibility through the affine transformation adopted, there is less need to keep active large part of the network, since a loss of information can be recovered through the other components of the model, as well as constraining a particular part of the architecture. For what concerns the value of the parameters $k_1$, $k_2$ and $k_3$ for both architectures $k_2$ and $k_3$ tend to have larger magnitudes with respect to $k_1$. Also, the values of $k_2$ and $k_1$ tend to have a different sign, which allows the term $k_1\mat 1+k_2\mat M$ to span over positive and negative values. We also notice that the transformation of the weights are more prominent as the depth increases, which is intuitively explained by the fact that baseline network requires stronger adaptation to represent the higher-level concepts pertaining to different tasks. This is even more evident for WikiArt and Sketch due to the variability that these datasets contain with respect to standard natural images.

\subsection{Conclusions}
This section presented a simple yet powerful method for sequentially learning new tasks, given a fixed, pre-trained deep architecture. In particular, we generalize previous works on multi-domain learning applying binary masks to the original weights of the network \cite{mallya2018piggyback} by introducing an affine transformation that acts upon such weights and the masks themselves. Our generalization allows implementing a large variety of possible transformations, better adapting to the specific characteristics of each task. These advantages are shown experimentally on two public benchmarks fully confirm the power of our approach which fills the gap between the binary-mask based and state-of-the-art methods on the Visual Decathlon Challenge. 

Interesting future directions are extending this approach to several life-long learning scenarios (from incremental class learning to open-world recognition) and exploiting the relationship between different task through cross-task affine transformations, in order to reuse knowledge obtained from different tasks by the model. 

While in this section we considered multi-domain learning, an inherently multi-head problem, single-head incremental learning scenarios are considered more challenging in the community, due to the more severe presence of the catastrophic forgetting problem \cite{chaudhry2018riemannian}. In the next section, we will study the problem of incremental class learning in semantic segmentation, a single-head problem mostly unexplored in the community.

\section[Incremental Learning in Semantic Segmentation]{Incremental Learning in Semantic Segmentation \footnotemark\footnotetext{F. Cermelli, M. Mancini, E. Ricci, B. Caputo. {\sl Modeling the Background for Incremental Learning in Semantic Segmentation}. IEEE/CVF International Conference on Computer Vision and Pattern Recognition (CVPR) 2020.}}
\label{sec:IL-semantic-seg}




In Section \ref{sec:IL-multitask}, we focused on the problem of multi-domain learning, where the goal is to equip a model to tackle multiple tasks at the same time. Both in our BAT approach and previous works \cite{rebuffi2017learning,rebuffi2018efficient,rosenfeld2017incremental,li2019efficient} this is achieved by learning task-specific parameters which are included in the original pre-trained model. This kind of scenario falls into the multi-head incremental learning setting (i.e. one network per task/set of concepts) and it is considered to be an easier problem than the single-head counterpart \cite{chaudhry2018riemannian}. In the single-head scenario, we have a unique model classifying all semantic concepts together and, since all concepts share the same output space, this makes the catastrophic forgetting problem more severe. In this section, we will describe a solution to a classical single-head scenario, i.e. incremental class learning, for an unexplored task: semantic segmentation.

Semantic segmentation
is a fundamental problem in computer vision.  
In the last years, thanks to the emergence of deep neural networks and to the availability of large-scale human-annotated datasets \cite{pascal-voc-2012,zhou2017scene}, the state of the art 
has improved significantly \cite{long2015fully, chen2018encoder, zhao2017pyramid, lin2017refinenet, zhang2018exfuse}.
Current approaches 
are derived by extending deep architectures from image-level to pixel-level classification, taking advantage of Fully Convolutional Networks (FCNs) \cite{long2015fully}. Over the years, semantic segmentation models based on FCNs have been improved in several ways, \eg by exploiting multiscale representations \cite{lin2017refinenet,zhang2018exfuse}, modeling spatial dependencies and contextual cues \cite{chen2017rethinking, chen2017deeplab, chen2018encoder} or considering attention models \cite{chen2016attention}.

Still, existing semantic segmentation methods 
are not designed 
to incrementally update their inner classification model when new categories are discovered. 
While deep nets are undoubtedly powerful, it is well known that their capabilities in an incremental learning setting are limited \cite{kemker2018measuring}. In fact, deep architectures struggle in updating their parameters for learning new categories whilst preserving good performance on the old ones
(\textit{catastrophic forgetting} \cite{mccloskey1989catastrophic}).

\begin{figure}[t]
    \centering
    \includegraphics[width=\linewidth]{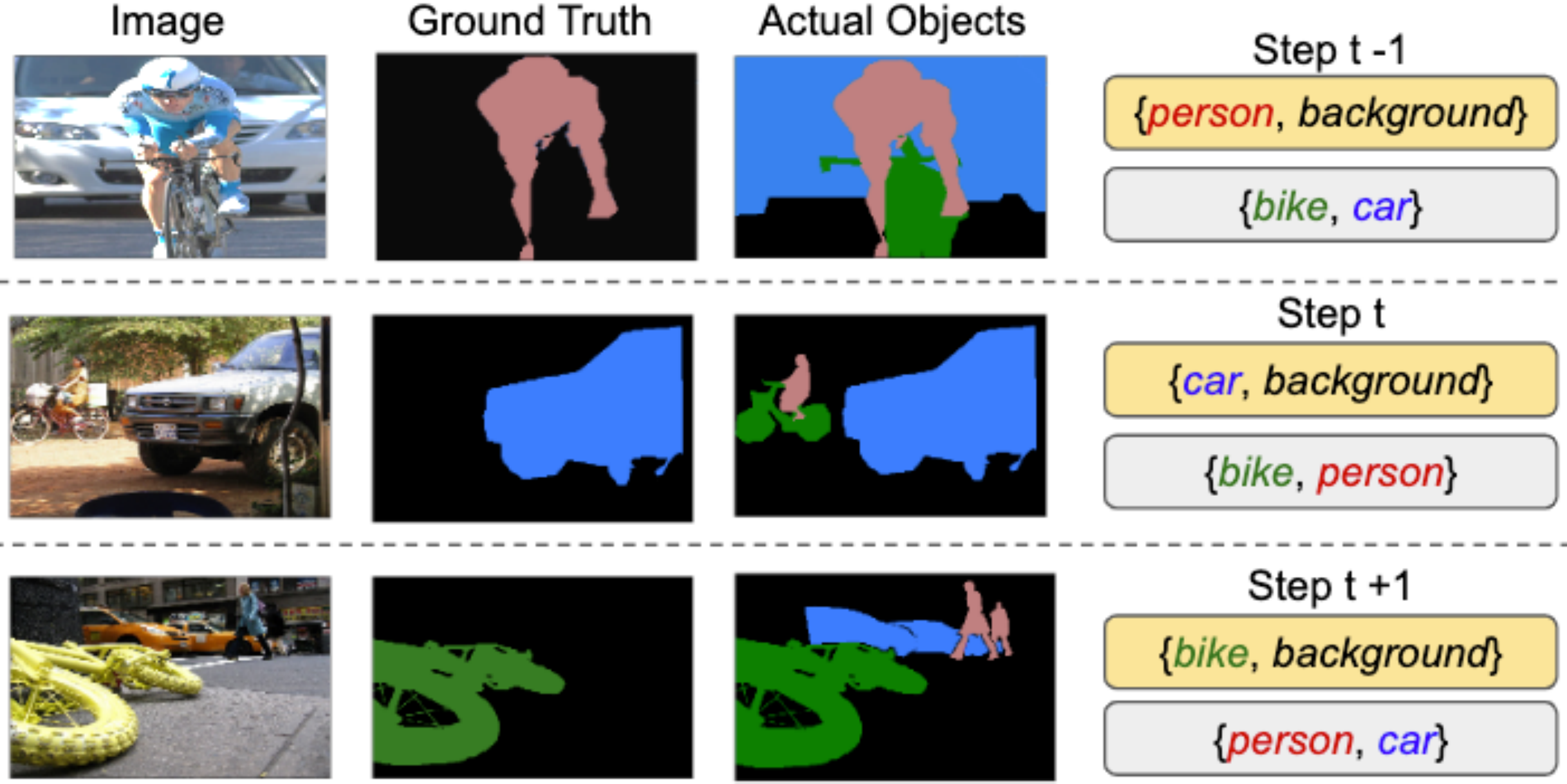}
    \caption{Illustration of the semantic shift of the background class in incremental learning for semantic segmentation. Yellow boxes denote the ground truth provided in the learning step, while grey boxes denote classes not labeled. As different learning steps have different label spaces, at step $t$ old classes (\eg \textit{person}) and unseen ones (\eg \textit{car}) might be labeled as background in the current ground truth. 
    Here we  show the specific case of single class learning steps, but we address the general case where an arbitrary number of classes is added.} 
    \label{fig:ILSS-teaser}
\end{figure}

As described in Section \ref{sec:IL-relateds}, the problem of incremental learning has been traditionally addressed in 
object recognition \cite{li2017learning,kirkpatrick2017overcoming,chaudhry2018riemannian,rebuffi2017icarl,hou2019learning} and detection \cite{shmelkov2017incremental}, but less attention has been devoted to semantic segmentation.
Here we fill this gap, 
proposing an incremental class learning (\icl) approach for semantic segmentation. Inspired by previous methods on image classification \cite{li2017learning,rebuffi2017icarl,castro2018end}, we 
cope with catastrophic forgetting by resorting to knowledge distillation \cite{hinton2015distilling}. However, we argue (and experimentally demonstrate) that a naive application of previous knowledge distillation strategies would not suffice in this setting. In fact, one peculiar aspect of semantic segmentation is the presence of a special class, the background class, indicating pixels not assigned to any of the given object categories. While the presence of this class marginally influences the design of traditional, offline semantic segmentation methods, this is not true in an incremental learning setting.
As illustrated in Fig. \ref{fig:ILSS-teaser}, it is reasonable to assume that the semantics associated to the background class changes over time. In other words, pixels associated to the background during a learning step may be assigned to a specific object class in subsequent steps or vice-versa, with the effect of exacerbating the catastrophic forgetting. 
To overcome this issue, we revisit the classical distillation-based framework for incremental learning \cite{li2017learning} by introducing two novel loss terms 
to properly account for the semantic distribution shift within the background class, thus introducing the first \icl\ approach tailored to semantic segmentation. We name this method as \ours (\expandednick). 
We extensively evaluate \ours on two datasets, Pascal-VOC \cite{pascal-voc-2012} and ADE20K \cite{zhou2017scene}, showing that our approach, coupled with a novel classifier initialization strategy, largely outperform traditional \icl\ methods. 


\noindent To summarize, the contributions described in this section are as follows:
\begin{itemize}
    \item We study the  task of incremental class learning for semantic segmentation, analyzing in particular the problem of distribution shift arising due to the presence of the background class. 
    \item We propose a new objective function 
    and introduce a specific classifier initialization strategy to explicitly cope with the evolving semantics of the background class. We show that our approach greatly alleviates the catastrophic forgetting, leading to the state of the art. 
    \item We benchmark \ours over several previous \icl\ methods on two popular semantic segmentation datasets, considering different experimental settings. We hope that our results will serve as a reference for future works.
\end{itemize}


\begin{figure*}
    \centering
    \includegraphics[width=\textwidth]{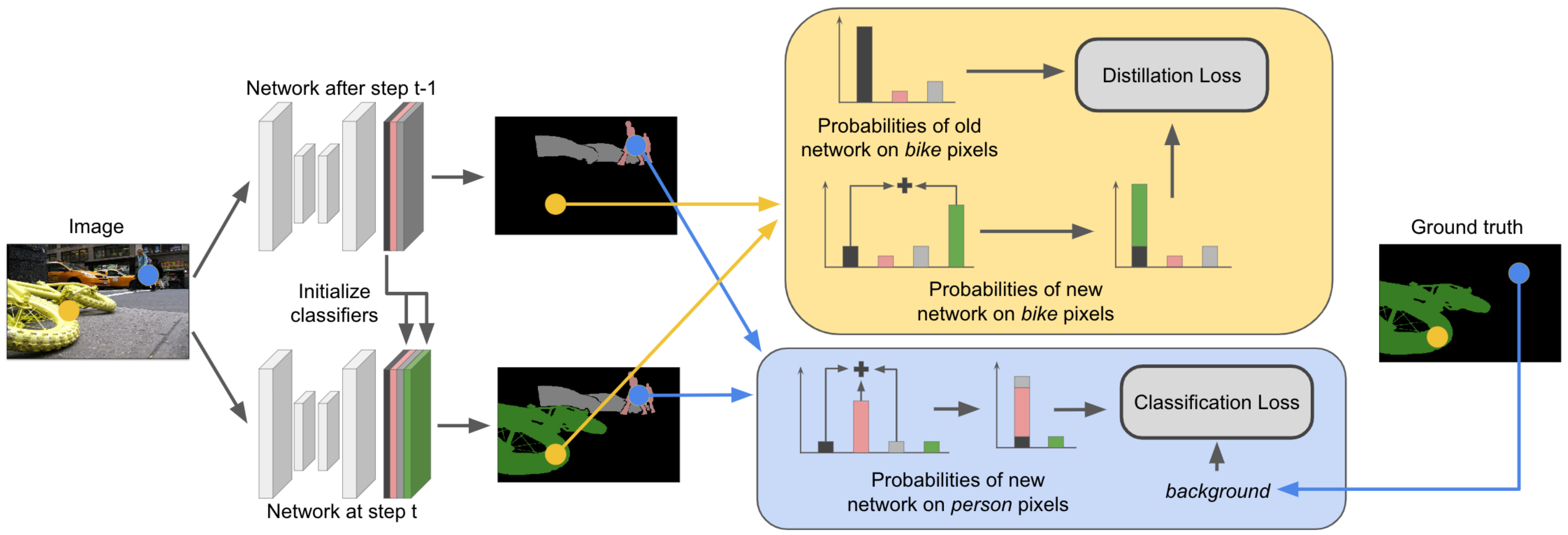}
    \caption{Overview of \ours. 
    At learning step $t$ an image is processed by the old (top) and current (bottom) models, mapping the image to their respective output spaces. As in standard \icl\ methods, we apply a cross-entropy loss to learn new classes (blue block) and a distillation loss to preserve old knowledge (yellow block). In this framework, we model the semantic changes of the background 
 across different learning steps by (i) initializing the new classifier using the weights of the old background one (left), (ii) comparing the pixel-level background ground truth in the cross-entropy with the probability of having either the background (black) or an old class (pink and grey bars) and (iii) relating the background probability given by the old model in the distillation loss with the probability of having either the background or a novel class (green bar).}
    \label{fig:ILSS-method}
\end{figure*}
 

\subsection{Problem Formulation}
\label{sec:ILSS-problem}

Before delving into the details of \icl\ for semantic segmentation, we first introduce the task of semantic segmentation.
Let us denote as $\mathcal{X}$ the input space (\ie the image space) and, without loss of generality, let us assume that each image $x\in\mathcal{X}$ is composed by a set of pixels $\set I$ with constant cardinality $|\set I|=N$. 
The output space is defined as $\mathcal{Y}^{N}$, with the latter denoting the product set of $N$-tuples with elements in a label space $\mathcal{Y}$. Given an image $x$ the goal of semantic segmentation is to assign each pixel $x_i$ of image $x$ a label $y_i \in \mathcal{Y}$, representing its semantic class. Out-of-class pixels can be assigned a special class, \ie the background class $\mathtt{b}\in\set Y$. 
{Given a training set $\mathcal{T} \subset \mathcal{X}\times \mathcal{Y}^N$, the mapping is realized by learning a model $f_{\theta}$ with parameters $\theta$ from the image space $\mathcal{X}$ to a pixel-wise class probability vector, \ie $f_{\theta} : \mathcal{X} \mapsto \mathcal{\real}^{N \times |\mathcal{Y}|}$.
The output segmentation mask is obtained as $y^* = \{ \argmax_{c\in\mathcal{Y}} f_{\theta}(x)[i,c]\}_{i=1}^{N}$, where $f_{\theta}(x)[i,c]$ is the probability for class $c$ in pixel $i$.}

In the \icl\ setting, training is realized over multiple phases, called \textit{learning steps}, and each step introduces novel categories to be learnt. In other terms, during the $t_{\text{th}}$ learning step, the previous label set $\set Y^{t-1}$ is expanded with a set of new {classes} $\set C^t$, yielding a new label set $\set Y^t=\set Y^{t-1}\cup\set C^{t}$.
Following the notation in Section \ref{sec:IL-ps}, at learning step $t$ we are also provided with a training set $\mathcal{T}^t \subset \mathcal{X}\times (\set {C}^{t})^N$ that is used in conjunction to the previous model $f_{\theta^{t-1}}:\set X\mapsto \real^{N\times |\set Y^{t-1}|}$ to train an updated model $f_{\theta^{t}}:\set X\mapsto \real^{N\times |\set Y^{t}|}$.
As in standard \icl, we assume the sets of labels $\mathcal{C}^t$ that we obtain at the different learning steps to be disjoint, except for the special void/background class $\con b$.

\subsection{Modeling the Background for Incremental Learning in Semantic Segmentation}
\label{sec:ILSS-our-method}
A naive approach to address the \icl\ problem consists in retraining the model $f_{\theta^t}$ on each set $\mathcal{T}^t$ sequentially. When the predictor $f_{\theta^t}$ is realized through a deep architecture, this corresponds to fine-tuning the network parameters on the training set $\mathcal{T}^t$ initialized with the parameters $\theta^{t-1}$ from the previous stage. This approach is simple, but it leads to catastrophic forgetting. Indeed, when training using $\mathcal{T}^t$ no samples from the previously seen object classes 
are provided. This biases the new predictor $f_{\theta^t}$ 
towards the novel set of categories in $\mathcal{C}^t$ to the detriment of the classes from the previous sets. In the context of \icl\ for image-level classification, a standard way to address this issue is coupling the supervised loss on $\mathcal{T}^t$ with a regularization term, either taking into account the importance of each parameter for previous tasks \cite{kirkpatrick2017overcoming,shin2017continual}, or by distilling the knowledge using the predictions of the old model $f_{\theta^{t-1}}$ \cite{li2017learning,rebuffi2017icarl,castro2018end}. We take inspiration from the latter solution to initialize the overall objective function of our problem. In particular, we  minimize a loss function of the form:
\begin{equation}
   \label{eq:ILSS-obj-general}
    \mathcal{L}(\theta^t)= \frac{1}{|\mathcal{T}^t| 
    }\sum_{(x,y)\in\mathcal{T}^t}
    \left(\ell^{\theta^t}_{ce}(x,y) + \lambda \ell^{\theta^t}_{kd}(x) \right)
\end{equation}
where $\ell_{ce}$ is a standard supervised loss (\eg cross-entropy loss), $\ell_{kd}$ is the distillation loss and $\lambda>0$ is a \hyper balancing the importance of the two terms. 

As stated in Sec. \ref{sec:ILSS-problem}, differently from standard \icl\ settings considered for image classification problems, in semantic segmentation we have that two different label sets $\mathcal{C}^s$ and $\mathcal{C}^u$  share the common void/background class $\mathtt{b}$. 
However, 
{the distribution of the background class} changes across different incremental steps. In fact, background annotations given in $\set T^t$ refer to classes not present in $\set C^t$, that might belong to the set of seen classes $\set Y^{t-1}$ and/or to still unseen classes \ie $\set C^{u}$ with $u>t$  (see Fig. \ref{fig:ILSS-teaser}). 
In the following, we show how we account for the semantic shift in the distribution of the background class by revisiting standard choices for the general objective defined in Eq. \eqref{eq:ILSS-obj-general}.

\myparagraph{Revisiting Cross-Entropy Loss. } In Eq.\eqref{eq:ILSS-obj-general}, a possible choice for $\ell_{ce}$ is the standard cross-entropy loss computed over all image pixels: 
\begin{equation}
   \label{eq:ILSS-CE}
 \ell^{\theta^t}_{ce}(x,y) = -\frac{1}{|\set I|}\sum_{i \in\set I}\log q_x^t(i,y_i)\,,
\end{equation} 
where $y_i \in \set Y^t$ is the ground truth label associated to pixel $i$ and $q_x^t(i,c)=f_{\theta^t}(x)[i,c]$. 

The problem with Eq.\eqref{eq:ILSS-CE} is that the training set $\mathcal{T}^t$ we use to update the model 
only contains information about novel classes in $\mathcal{C}^t$. However, the background class in $\mathcal{T}^t$ might include also pixels associated to the previously seen classes in $\mathcal{Y}^{t-1}$. Here we argue that, without explicitly taking into account this aspect, the catastrophic forgetting problem would be even more severe. In fact, we would drive our model to predict the background label $\mathtt{b}$ for pixels of old classes, further degrading the capability of the model to preserve semantic knowledge of past categories. To avoid this issue, we propose to modify the cross-entropy loss in Eq.\eqref{eq:ILSS-CE} as follows: 
\begin{equation}
   \label{eq:ILSS-our-CE}
    \ell^{\theta^t}_{ce}(x,y) = -\frac{1}{|\set I|}\sum_{i \in\set I}\log \tilde{q}_x^t(i,y_i)\,,
 \end{equation}
 where:
\begin{equation}
    \label{eq:ILSS-cases-ce}
    \tilde{q}_x^t(i,c) = \begin{cases}
      {q}_x^t(i,c)\;\;& \text{if}\ c\neq\mathtt{b}\\
      \sum_{k\in\mathcal{{Y}}^{t-1}}{q}_x^t(i,k)\;\;& \text{if}\ c=\mathtt{b}\,.
    \end{cases}
\end{equation}

Our intuition is that by using Eq.\eqref{eq:ILSS-our-CE} we can update the model to predict the new classes and, at the same time, account for the uncertainty over the actual content of the background class. In fact, in Eq.\eqref{eq:ILSS-our-CE} the background class ground truth is not directly compared with its probabilities ${q}_x^t(i,\con b) $ obtained from the current model $f_{\theta^t}$, but with the probability of having \textit{either an old class or the background}, as predicted by $f_{\theta^t}$ (Eq.\eqref{eq:ILSS-cases-ce}). {A schematic representation of this procedure is depicted in Fig.~\ref{fig:ILSS-method} (blue block).}
It is worth noting that the alternative of ignoring the background pixels within the cross-entropy loss is a sub-optimal solution. In fact, this would not allow to adapt the background classifier to its semantic shift and to exploit the information that new images might contain about old classes. 

\myparagraph{Revisiting Distillation Loss.} In the context of incremental learning, distillation loss \cite{hinton2015distilling} is a common strategy to transfer knowledge from the old model $f_{\theta^{t-1}}$ into the new one, preventing catastrophic forgetting. 
Formally, a standard choice for the distillation loss $\ell_{kd}$ is:
\begin{equation}
    \label{eq:ILSS-std-distill}
     \ell^{\theta^t}_{kd}(x,y) = -\frac{1}{|\set I|}\sum_{i \in\set I}\sum_{c \in\set Y^{t-1}} q_x^{t-1}(i,c)\log \hat{q}_x^{t}(i,c)\,,
\end{equation}
where $\hat{q}_x^{t}(i,c)$ is defined as the probability of class $c$ for pixel $i$ given by $f_{\theta^t}$ but re-normalized across all the classes in $\set Y^{t-1}$ \ie:
\begin{equation}
    \label{eq:ILSS-cases-kd}
    \hat{q}_x^{t}(i,c)= \begin{cases}
       0 \;\;& \text{if}\ c\in \set C^t\setminus\{\mathtt{b}\}\\
      {q}_x^{t}(i,c)/\sum_{k\in \set Y^{t-1}} q_x^{t}(i,k) \;\;& \text{if}\ c\in \set Y^{t-1}\,.
    \end{cases}
\end{equation}

The rationale behind $\ell_{kd}$ is that $f_{\theta^t}$ should produce activations close to the ones produced by $f_{\theta^{t-1}}$. This regularizes the training procedure in such a way that the parameters $\theta^t$ are still anchored to the solution found for recognizing pixels of the previous classes, \ie $\theta^{t-1}$. 

The loss defined in Eq.\eqref{eq:ILSS-std-distill} has been used either in its base form or variants in different contexts, from incremental task \cite{li2017learning} and class learning \cite{rebuffi2017icarl,castro2018end} in object classification to complex scenarios such as detection \cite{shmelkov2017incremental} and segmentation \cite{michieli2019incremental}. Despite its success, it has a fundamental drawback in semantic segmentation: it completely ignores the fact that the background class is shared among different learning steps. 
While with Eq.\eqref{eq:ILSS-our-CE} we tackled the first problem linked to the semantic shift of the background (\ie $\mathtt{b} \in \set T^{t}$ contains pixels of $\set Y^{t-1}$), we use the distillation loss to tackle the second: {annotations for background in $\set T^s$ with $s<t$ might include pixels of classes in $\set C^t$.}

{From the latter considerations, the background probabilities assigned to a pixel by the old predictor $f_{\theta^{t-1}}$ and by the current model $f_{\theta^{t}}$ do not share the same semantic content.} 
More importantly, $f_{\theta^{t-1}}$ might predict as background pixels of classes in $\set C^t$ that we are currently trying to learn. Notice that this aspect is peculiar to the segmentation task and it is not considered in previous incremental learning models. However, in our setting we must explicitly take it into account to perform a correct distillation of the old model into the new one. To this extent we define our novel distillation loss by rewriting $\hat{q}_x^t(i,c)$ in Eq.\eqref{eq:ILSS-cases-kd} as: 
\begin{equation}
    \label{eq:ILSS-cases-ukd}
    \hat{q}_x^{t}(i,c)= \begin{cases}
      {q}_x^{t}(i,c)\;\;& \text{if}\ c\neq\mathtt{b}\\
      \sum_{k\in \set C^t}q_x^{t}(i,k)\;\;& \text{if}\ c=\mathtt{b}\,.
    \end{cases}
\end{equation}
Similarly to Eq.\eqref{eq:ILSS-std-distill}, we still compare the probability of a pixel belonging to seen classes assigned by the old model, with its counterpart computed with the current parameters $\theta^t$. However, differently from classical distillation, in Eq.\eqref{eq:ILSS-cases-ukd} the probabilities obtained with the current model are kept unaltered, \ie normalized across the whole label space $\set {Y}^t$ and not with respect to the subset $\set Y^{t-1}$ (Eq.\eqref{eq:ILSS-cases-kd}). More importantly, the background class probability as given by $f_{\theta^{t-1}}$ is not directly compared with its counterpart in $f_{\theta^t}$, but with the probability of having \textit{either a new class or the background}, as predicted by $f_{\theta^t}$ ({see Fig. \ref{fig:ILSS-method}, yellow block}).  

We highlight that, with respect to Eq.\eqref{eq:ILSS-cases-kd} and other simple choices 
(\eg excluding $\con b$ from Eq.\eqref{eq:ILSS-cases-kd}) this solution has two advantages. First, we can still use the full output space of the old model to distill knowledge in the current one, without any constraint on pixels and classes. Second, we can propagate the uncertainty we have on the semantic content of the background in $f_{\theta^{t-1}}$ 
without penalizing the probabilities of new classes we are learning in the current step $t$. 


\myparagraph{Classifiers' Parameters Initialization.} As discussed above, the background class $\con b$ is a special class devoted to collect the probability that a pixel belongs to an unknown object class. 
In practice, at each learning step $t$, the novel categories in $\set C^t$ are unknowns for the old classifier $f_{\theta^{t-1}}$. As a consequence, unless the appearance of a class in $\set C^t$ is very similar to one in $\set Y^{t-1}$, it is reasonable to assume that $f_{\theta^{t-1}}$ will likely assign pixels of $\set C^t$ to $\con b$. 
Taking into account this initial bias on the predictions of $f_{\theta^{t}}$ on pixels of $\set C^t$, it is detrimental to randomly initialize the classifiers for the novel classes. In fact a random initialization would provoke a misalignment among the features extracted by the model (aligned with the background classifier) and the random parameters of the classifier itself. Notice that this could lead to possible training instabilities while learning novel classes since the network could initially assign high probabilities for pixels in $\set C^t$ to $\con b$. 


To address this issue, we propose to initialize the classifier's parameters for the novel classes in such a way that given an image $x$ and a pixel $i$, the probability of the background $q_x^{t-1}(i,\con b)$ is uniformly spread among the classes in $\set C^{t}$, \ie $q_x^{t}(i,c)=q_x^{t-1}(i,\con b)/|\set C^t|\; \forall c \in \set C^t$, where $|\mathcal{C}^t|$ is the number of new classes (notice that $\mathtt{b} \in \mathcal{C}^t$). 
To this extent, let us consider a standard fully connected classifier and let us denote as $\{\omega^{t}_c, \beta^{t}_c\}\in\theta^t$ the classifier parameters for a class $c$ at learning step $t$, with $\omega$ and $\beta$ denoting its weights and bias respectively. We can initialize $\{\omega^t_c, \beta^t_c\}$ as follows:
\begin{align}
    \label{eq:ILSS-init-cases}
    \omega_c^{t}&=\begin{cases}
     \omega_{\con b}^{t-1} \;\;& \text{if}\ c \in \mathcal{C}^t\\
      \omega^{t-1}_c \;\;& \text{otherwise}\\
    \end{cases}\\
        \label{eq:ILSS-init-cases2}
    \beta_c^{t}&=\begin{cases}
      \beta_{\con b}^{t-1} - \log(|\set C^t|)\;\;& \text{if}\ c \in \mathcal{C}^t\\
      \beta^{t-1}_c \;\;& \text{otherwise}\\
    \end{cases}
\end{align}
 where $\{\omega_\mathtt{b}^{t-1},\beta_\mathtt{b}^{t-1} \}$ are the weights and bias of the background classifier at the previous learning step. The fact that the initialization defined in Eq.\eqref{eq:ILSS-init-cases} and \eqref{eq:ILSS-init-cases2} leads to $q_x^{t}(i,c)=q_x^{t-1}(i,\con b)/|\set C^t|\; \forall c \in \set C^t$ is easy to obtain from $q_x^{t}(i,c)\propto \exp(\omega_\mathtt{b}^{t}\cdot x + \beta^{t}_\mathtt{b})$. 

As we will show in the experimental analysis, this simple initialization procedure brings benefits in terms of both improving the learning stability of the model and the final results, since it eases the role of the supervision imposed by Eq.\eqref{eq:ILSS-our-CE} while learning new classes and follows the same principles used to derive our distillation loss (Eq.\eqref{eq:ILSS-cases-ukd}). 



\subsection{Experimental results}
\label{sec:ILSS-exp}
\subsubsection{\icl\
 Baselines} \label{sec:ILSS-baselines}
We compare \ours against standard \icl\ baselines, 
 originally designed for classification tasks,
 on the considered segmentation task, 
 thus segmentation is treated as a pixel-level classification problem.
{Specifically, we report the results of six different regularization-based methods, three prior-focused and three data-focused.} 

In the first category, we chose Elastic Weight Consolidation (EWC) \cite{kirkpatrick2017overcoming}, Path Integral (PI) \cite{zenke2017continual}, and Riemannian Walks (RW) \cite{chaudhry2018riemannian}. They employ different strategies to compute the importance of each parameter for old classes: EWC uses the empirical Fisher matrix, PI uses the learning trajectory, while RW combines EWC and PI in a unique model. We choose EWC since it is a standard baseline employed also in \cite{shmelkov2017incremental} and PI and RW since they are two simple applications of the same principle. Since these methods act at the parameter level, to adapt them to the segmentation task we keep the loss in the output space unaltered (\ie standard cross-entropy across the whole segmentation mask), computing the parameters' importance by considering their effect on learning old classes.

For the data-focused methods, we chose Learning without forgetting (LwF) \cite{li2017learning}, LwF multi-class (LwF-MC) \cite{rebuffi2017icarl} and the segmentation method of \cite{michieli2019incremental} (ILT).
We denote as LwF the original distillation based objective as implemented in Eq.\eqref{eq:ILSS-obj-general} with basic cross-entropy and distillation losses, which is the same as \cite{li2017learning} except that distillation and cross-entropy share the same label space and classifier. LwF-MC is the single-head version of \cite{li2017learning} as adapted from \cite{rebuffi2017icarl}. It is based on multiple binary classifiers, with the target labels defined using the ground truth for novel classes (\ie $\set C^t$) and the probabilities given by the old model for the old ones (\ie $\set Y^{t-1}$). Since the background class is both in $\set C^t$ and $\set Y^{t-1}$ 
we implement LwF-MC by a weighted combination of two binary cross-entropy losses, on both the ground truth and the probabilities given by $f_{\theta^{t-1}}$. 
Finally, ILT \cite{michieli2019incremental} is the only method specifically proposed for \icl\ in semantic segmentation. It uses a distillation loss in the output space, as in our adapted version of LwF \cite{li2017learning} and/or another distillation loss in the features space, attached to the output of the network decoder. Here, we use the variant where both losses are employed.
As done by \cite{shmelkov2017incremental}, we do not compare with replay-based methods (\eg \cite{rebuffi2017icarl}) since they violate the standard \icl\ assumption regarding the unavailability of old data.

In all tables we report other two baselines: simple fine-tuning (FT) on each $\set T^t$ (\eg Eq.\eqref{eq:ILSS-CE}) and training on all classes offline (Joint). The latter can be regarded as an upper bound. All results are reported as mean Intersection-over-Union (mIoU) in percentage, averaged over all the classes of a learning step and all the steps.

\subsubsection{Implementation Details}\label{sec:ILSS-impdetails}
For all methods we use the Deeplab-v3 architecture \cite{chen2017rethinking} with a ResNet-101 \cite{he2016deep} backbone and output stride of 16. Since memory requirements are an important issue in semantic segmentation, we use in-place activated batch normalization, as proposed in \cite{rota2018place}. The backbone has been initialized using the ImageNet pre-trained model \cite{rota2018place}. 
We follow \cite{chen2017rethinking}, training the network with SGD and the same learning rate policy, momentum and weight decay. 
We use an initial learning rate of $10^{-2}$ for the first learning step and $10^{-3}$ for the followings, as in \cite{shmelkov2017incremental}.
We train the model with a batch size of 24 for 30 epochs for Pascal-VOC 2012 and 60 epochs for ADE20K in every learning step. 
We apply the same data augmentation of \cite{chen2017rethinking} 
and we crop the images to $512\times 512$ during both training and test.
For setting the \hypers of each method, we use the protocol of incremental learning defined in \cite{de2019continual}, using 20\% of the training set as validation.
The final results are reported on the standard validation set of the datasets.


\subsubsection{Pascal-VOC 2012} \label{sec:ILSS-pascal}
PASCAL-VOC 2012 \cite{pascal-voc-2012} is a widely used benchmark that includes 20 foreground object classes. 
Following \cite{michieli2019incremental,shmelkov2017incremental}, we define two experimental settings, depending on how we sample images to build the incremental datasets.
Following \cite{michieli2019incremental}, we define an experimental protocol called the \textit{disjoint} setup: each learning step contains a unique set of images, whose pixels belong to classes seen either in the current or in the previous learning steps. Differently from \cite{michieli2019incremental}, at each step we assume to have only labels for pixels of novel classes, while the old ones are labeled as background in the ground truth. 
The second setup, that we denote as \textit{overlapped}, follows what has been done in \cite{shmelkov2017incremental} for detection: each training step contains all the images that have at least one pixel of a novel class, with only the latter annotated. It is important to note a difference with respect to the previous setup: images may now contain pixels of classes that we will learn in the future, but labeled as background. This is a more realistic setup since it does not make any assumption on the objects present in the images.

Following previous works \cite{shmelkov2017incremental, michieli2019incremental}, we perform three different experiments concerning the addition of one class (\textit{19-1}), five classes all at once (\textit{15-5}), and five classes sequentially (\textit{15-1}), following the alphabetical order of the classes to split the content of each learning step. 

\myparagraph{Addition of one class \textit{(19-1)}.}
In this experiment, we perform two learning steps: the first in which we observe the first 19 classes, and the second where we learn the \textit{tv-monitor} class.
Results are reported in Table \ref{tab:ILSS-pascal-dis} for the \textit{disjoint} scenario and in Table \ref{tab:ILSS-pascal-ov} for the \textit{overlapped}. Without employing any regularization strategy, the performance on past classes drops significantly. FT, in fact, performs poorly, completely forgetting the first 19 classes. 
Unexpectedly, using PI as a regularization strategy does not provide benefits, while EWC and RW improve performance of nearly 15\%. However, prior-focused strategies are not competitive with data-focused ones. In fact, LwF, LwF-MC, and ILT, outperform them by a large margin, confirming the effectiveness of this approch on preventing catastrophic forgetting. While ILT surpasses standard \icl\ baselines, our model is able to obtain a further boost. This improvement is remarkable for new classes, where we gain $11\%$ in mIoU, while do not experience forgetting on old classes. It is especially interesting to compare \ours with the baseline LwF which uses the same principles of our method but without modeling the background. Compared to LwF we achieve an average improvement of about $15\%$, thus demonstrating the importance of modeling the background in \icl\ for semantic segmentation. These results are consistent in both the \textit{disjoint} and \textit{overlapped} scenarios.

\begin{table*}[t]
\centering
\setlength{\tabcolsep}{5pt} 
\caption{Mean IoU on the Pascal-VOC 2012 dataset for the \textit{disjoint} incremental class learning scenarios.}
\label{tab:ILSS-pascal-dis}
\begin{tabular}{l||cc|c||cc|c||cc|c}
\multicolumn{1}{c}{}    & \multicolumn{3}{c}{\textbf{{19-1}}}   & \multicolumn{3}{c}{{\textbf{15-5}}} & \multicolumn{3}{c}{{\textbf{15-1}}}    \\
\bf{Method} & \it{1-19}  & \it{20}   & \it{all}  & \it{1-15}  & \it{16-20}   & \it{all}  & \it{1-15}  & \it{16-20}   & \it{all}     \\ \hline
{FT }     & 5.8     & 12.3    & 6.2      & 1.1    & 33.6    & 9.2      & 0.2        & 1.8        & 0.6       \\
{PI \cite{zenke2017continual}}     & 5.4     & 14.1    & 5.9  & 1.3    & 34.1    & 9.5        & 0.0 &	1.8 &	0.4   \\
{EWC \cite{kirkpatrick2017overcoming}}    & 23.2    & 16.0    & 22.9     & 26.7   & 37.7    & 29.4      & 0.3        & 4.3        & 1.3        \\
{RW \cite{chaudhry2018riemannian}}     & 19.4    & 15.7    & 19.2    & 17.9   & 36.9    & 22.7       & 0.2        & 5.4  & 1.5\\
{LwF \cite{li2017learning} }    & 53.0    & 9.1     & 50.8    & 58.4   & 37.4    & 53.1    & 0.8        & 3.6        & 1.5            \\
{LwF-MC \cite{rebuffi2017icarl}} & 63.0    & 13.2    & 60.5   & 67.2   & 41.2    & 60.7      & 4.5        & 7.0        & 5.2   \\
{ILT \cite{michieli2019incremental}}    & 69.1    & 16.4    & 66.4  & 63.2   & 39.5    & 57.3     & 3.7        & 5.7        & 4.2         \\
\ours   & \bf{69.6} & \bf{25.6}   & \bf{67.4}   & \bf{71.8}     & \bf{43.3}    & \bf{64.7}  & \bf{46.2}       & \bf{12.9}       & \bf{37.9}      \\ \hline
{Joint} & 77.4&	78.0&	77.4&	79.1&	72.6&	77.4& 79.1       & 72.6       & 77.4
\end{tabular}
\end{table*}

\begin{table*}[t]
\centering
\setlength{\tabcolsep}{5pt} 
\caption{Mean IoU on the Pascal-VOC 2012 dataset for the \textit{overlapped} incremental class learning scenario.}
\label{tab:ILSS-pascal-ov}
\begin{tabular}{l||cc|c||cc|c||cc|c}
\multicolumn{1}{c}{}    & \multicolumn{3}{c}{\textbf{{19-1}}}   & \multicolumn{3}{c}{{\textbf{15-5}}} & \multicolumn{3}{c}{{\textbf{15-1}}}    \\
\bf{Method} & \it{1-19}  & \it{20}   & \it{all}  & \it{1-15}  & \it{16-20}   & \it{all}  & \it{1-15}  & \it{16-20}   & \it{all}     \\ \hline
{FT }    & 6.8       & 12.9      & 7.1& 2.1     & 33.1  & 9.8  & 0.2        & 1.8        & 0.6       \\
{PI \cite{zenke2017continual}}     & 7.5       & 14.0      & 7.8       & 1.6     & 33.3  & 9.5    & 0.0        & 1.8        & 0.5 \\
{EWC \cite{kirkpatrick2017overcoming}}   & 26.9      & 14.0      & 26.3     & 24.3    & 35.5  & 27.1   & 0.3        & 4.3        & 1.3       \\
{RW \cite{chaudhry2018riemannian}}    & 23.3      & 14.2      & 22.9     & 16.6    & 34.9  & 21.2      & 0.0        & 5.2        & 1.3 \\
{LwF \cite{li2017learning} }    & 51.2      & 8.5       & 49.1     & 58.9    & 36.6  & 53.3       & 1.0        & 3.9        & 1.8       \\
{LwF-MC \cite{rebuffi2017icarl}}  & 64.4      & 13.3      & 61.9       & 58.1    & 35.0  & 52.3  & 6.4        & 8.4        & 6.9 \\
{ILT \cite{michieli2019incremental}}   & 67.1      & 12.3      & 64.4     & 66.3    & 40.6  & 59.9   & 4.9        & 7.8        & 5.7       \\
\ours    & \bf{70.2}    & \bf{22.1}       & \bf{67.8}    &  \bf{75.5}     & \bf{49.4}  & \bf{69.0} & \bf{35.1}       & \textbf{13.5}       & \textbf{29.7}             \\ \hline
{Joint} &	77.4&	78.0&	77.4&	79.1&	72.6&	77.4 & 79.1       & 72.6       & 77.4
\end{tabular}
\end{table*}

\myparagraph{Single-step addition of five classes (\textit{15-5}).}
In this setting we add, after the first training set, the following classes: \textit{plant, sheep, sofa, train, tv-monitor}. As before, results are reported in Table \ref{tab:ILSS-pascal-dis} for the \textit{disjoint} scenario and in Table \ref{tab:ILSS-pascal-ov} for the \textit{overlapped}. 
Overall, the behavior on the first 15 classes is consistent with the 19-1 setting: FT and PI suffer a large performance drop, data-focused strategies (LwF, LwF-MC, ILT) outperform EWC and RW by far, while \ours gets the best results, obtaining performances closer to the joint training upper bound. 
For what concerns the \textit{disjoint} scenario, our method improves over the best baseline of $4.6\%$ on old classes, of $2\%$ on novel ones and of $4\%$ in all classes. These gaps increase in the \textit{overlapped} setting where \ours surpasses the baselines by nearly $10\%$ in all cases, clearly demonstrating its ability to take advantage of the information contained in the background class. 

\myparagraph{Multi-step addition of five classes (\textit{15-1}).}
This setting is similar to the previous one except that the last 5 classes are learned sequentially, one by one.
From Table \ref{tab:ILSS-pascal-dis} and Table \ref{tab:ILSS-pascal-ov}, we can observe that performing multiple steps is challenging and existing methods work poorly for this setting, reaching performance inferior to 7\% on both old and new classes.
In particular, FT and prior-focused methods are unable to prevent forgetting, biasing their prediction completely towards new classes and demonstrating performances close to 0\% on the first 15 classes.
Even data-focused methods suffer a dramatic loss in performances in this setting, decreasing their score from the single to the multi-step scenarios of more than 50\% on all classes. 
On the other side, \ours is still able to achieve good performances. 
Compared to the other approaches, \ours outperforms all baselines by a large margin in both old ($46.2\%$ on the \textit{disjoint} and $35.1\%$ on the \textit{overlapped}), and new (nearly $13\%$ on both setups) classes. As the overall performance drop ($11\%$ on all classes) shows, the \textit{overlapped} scenario is the most challenging one since it does not impose any constraint on which classes are present in the background. 
\begin{table}[t]
\centering
\setlength{\tabcolsep}{6pt} 
\caption{Ablation study of the proposed method on the Pascal-VOC 2012 \textit{overlapped} setup. \textit{CE} and \textit{KD} denote our cross-entropy and distillation losses, while \textit{init} our initialization strategy.}
\label{tab:ILSS-ablation}
\begin{tabular}{l||cc|c||cc|c||cc|c}
      \multicolumn{1}{c}{}& \multicolumn{3}{c}{\textbf{19-1}}          & \multicolumn{3}{c}{\textbf{15-5}}          & \multicolumn{3}{c}{\textbf{15-1}} \\
  &\it{1-19} & \it{20} & \it{all} & \it{1-15} & \it{16-20} & \it{all} & \it{1-15}  & \it{16-20}     & \it{all}     \\ \hline
LwF \cite{li2017learning}         & 51.2      & 8.5       & 49.1       & 58.9      & 36.6       & 53.3       & 1.0         & 3.9   & 1.8   \\
     + \textit{CE} & 57.6      & 9.9       & 55.2       & 63.2       & 38.1       & 57.0       & 12.0        & 3.7   & 9.9   \\
+ \textit{KD}&   66.0      & 11.9      & 63.3       & 72.9       & 46.3       & 66.3       & 34.8        & 4.5   & 27.2  \\
+ \textit{init} & \textbf{70.2}      & \textbf{22.1}      & \textbf{67.8}      & \textbf{75.5}      & \textbf{49.4}      & \textbf{69.0}      & \textbf{35.1}       & \textbf{13.5} & \textbf{29.7}
\end{tabular}
\end{table}

\myparagraph{Ablation Study.}
In Table \ref{tab:ILSS-ablation} we report a detailed analysis of our contributions, considering the \textit{overlapped} setup. 
We start from the baseline LwF \cite{li2017learning} which employs standard cross-entropy and distillation losses.
We first add to the baseline our modified cross-entropy (\textit{CE}): this increases the ability to preserve old knowledge in all settings without harming (\textit{15-1}) or even improving (\textit{19-1}, \textit{15-5}) performances on the new classes. 
Second, we add our distillation loss (\textit{KD}) to the model. 
Our \textit{KD} provides a boost on the performances for both old and new classes. The improvement on old classes is remarkable, especially in the 15-1 scenario (\ie 22.8\%). For the novel classes, the improvement is constant and is especially pronounced in the 15-5 scenario (7\%). Notice that this aspect is peculiar of our \textit{KD} since standard formulation work only on preserving old knowledge. 
This shows that the two losses provide mutual benefits. 
Finally, we add our classifiers' initialization strategy (\textit{init}). This component provides an improvement in every setting, especially on novel classes: it doubles the performance on the \textit{19-1} setting ($22.1\%$ vs $11.9\%$) and triplicates on the \textit{15-1} ($4.5\%$ vs $13.5\%$). This confirms the importance of accounting for the background shift at the initialization stage to facilitate the learning of new classes. 

\subsubsection{ADE20K}
\label{sec:ILSS-ade}
\begin{table*}[t]
\centering
\setlength{\tabcolsep}{6pt} 
\caption{Mean IoU on the ADE20K dataset for different incremental class learning scenarios, adding 50 classes at each step.}
\label{tab:ILSS-ade-50}
\begin{tabular}{l||cc|c||ccc|c}
\multicolumn{1}{c}{} & \multicolumn{3}{c}{{\textbf{100-50}}}  & \multicolumn{4}{c}{{\textbf{50-50}}} \\
\textbf{Method}       & \textit{1-100} & \textit{101-150} & \textit{all}  & \textit{1-50} & \textit{51-100} & \textit{101-150} & \textit{all}  \\ \hline
FT     & 0.0   & 24.9    & 8.3   & 0.0  & 0.0    & 22.0    & 7.3  \\
LwF \cite{li2017learning}   & 21.1  & 25.6    & 22.6 & 5.7  & 12.9   & 22.8    & 13.9 \\
LwF-MC \cite{rebuffi2017icarl} & 34.2  & 10.5    & 26.3 & 27.8 & 7.0    & 10.4    & 15.1 \\
ILT \cite{michieli2019incremental}  & 22.9  & 18.9    & 21.6 & 8.4  & 9.7    & 14.3    & 10.8 \\
MiB   & \textbf{37.9}  & \textbf{27.9}    & \textbf{34.6} &  \textbf{35.5} & \textbf{22.2}   & \textbf{23.6}    & \textbf{27.0} \\ \hline
Joint  & 44.3  & 28.2    & 38.9 &  51.1 & 38.3   & 28.2    & 38.9
\end{tabular}
\end{table*}

\begin{table*}[t]
\centering
\setlength{\tabcolsep}{5pt} 
\caption{Mean IoU on the ADE20K dataset for a multi-step incremental class learning scenario, adding 50 classes in 5 steps.}
\label{tab:ILSS-ade-10}
\begin{tabular}{l|cccccc|c}
\multicolumn{1}{c}{} &  \multicolumn{7}{c}{{\textbf{100-10}}} \\
\textbf{Method}       &  \textit{1-100} & \textit{100-110} & \textit{110-120} & \textit{120-130} & \textit{130-140} & \textit{140-150} & \textit{all}  \\ \hline
FT  & 0.0   & 0.0    & 0.0    & 0.0    & 0.0    & 16.6   & 1.1   \\
LwF \cite{li2017learning}  & 0.1   & 0.0    & 0.4    & 2.6    & 4.6    & 16.9   & 1.7  \\
LwF-MC \cite{rebuffi2017icarl} &  18.7  &	2.5 &	8.7 &	4.1 &	6.5 &	5.1 &	14.3 \\
ILT \cite{michieli2019incremental}   & 0.3   & 0.0    & 1.0    & 2.1    & 4.6    & 10.7   & 1.4 \\
MiB   & \textbf{31.8}  & \textbf{10.4}   & \textbf{14.8}   & \textbf{12.8}   & \textbf{13.6}   & \textbf{18.7}   & \textbf{25.9} \\ \hline
Joint  & 44.3  & 26.1   & 42.8   & 26.7   & 28.1   & 17.3   & 38.9 
\end{tabular}
\end{table*}
ADE20K \cite{zhou2017scene} is a large-scale dataset that contains 150 classes. Differently from Pascal-VOC 2012, this dataset contains both stuff (\eg sky, building, wall)
and object classes. 
We create the incremental datasets $\set T^t$ by splitting the whole dataset into disjoint image sets, without any constraint except ensuring a minimum number of images (\ie 50) where classes on $\set C^t$ have labeled pixels. 
Obviously, each $\set T^t$ provides annotations only for classes in $\set C^t$ while other classes (old or future) appear as background in the ground truth. 
In Table \ref{tab:ILSS-ade-50} and Table \ref{tab:ILSS-ade-10} we report the mean IoU obtained averaging the results on two different class orders: the order proposed by \cite{zhou2017scene} and a random one. In this experiments, we compare \ours with data-focused methods only (\ie LwF, LwF-MC, and ILT) due to their gap in performance with prior-focused ones.

\begin{figure*}[t]
     \centering
         \includegraphics[width=\textwidth]{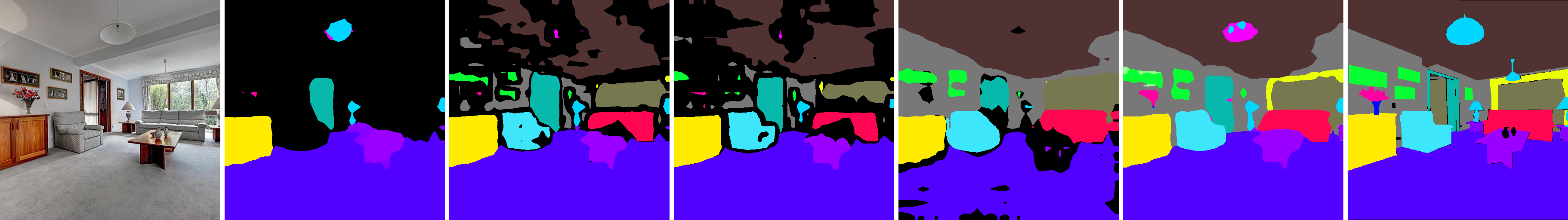}\\
         \includegraphics[width=\textwidth]{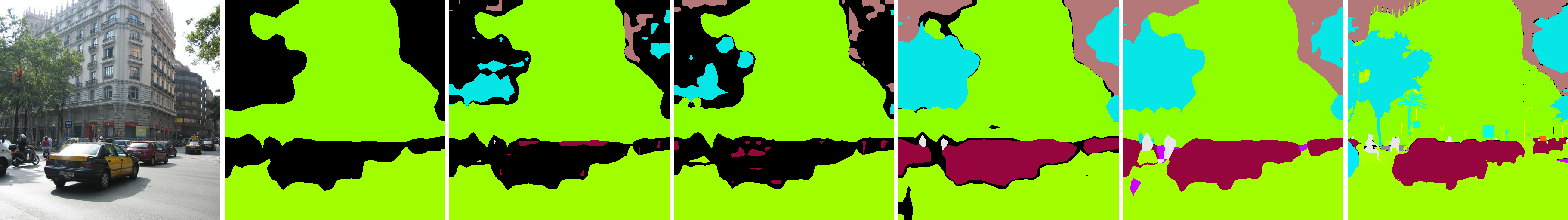}\\
         \includegraphics[width=\textwidth]{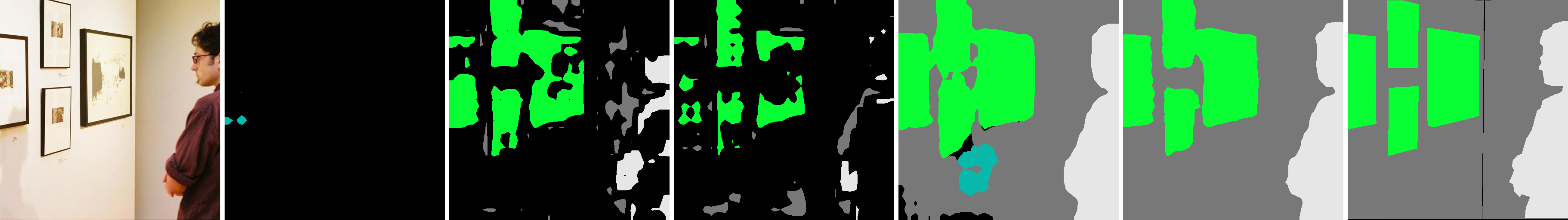}\\
         \includegraphics[width=\textwidth]{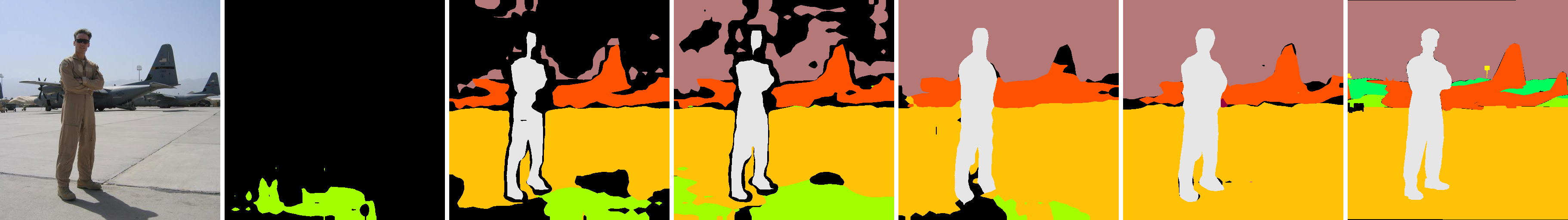}\\
     \caption{Qualitative results on the \textit{100-50} setting of the ADE20K dataset using different incremental methods. The image demonstrates the superiority of our approach on both new (\eg \textit{building}, \textit{floor}, \textit{table}) and old (\eg \textit{car}, \textit{wall}, \textit{person}) classes. From left to right: image, FT, LwF \cite{li2017learning}, ILT \cite{michieli2019incremental}, LwF-MC \cite{rebuffi2017icarl}, \ours, and the ground-truth. Best viewed in color.}
        \label{fig:ILSS-qualitative}
\end{figure*}

\myparagraph{Single-step addition of 50 classes (\textit{100-50}).} In the first experiment, we initially train the network on 100 classes and we add the remaining 50 all at once.
From Table \ref{tab:ILSS-ade-50} we can observe that FT is clearly a bad strategy on large scale settings since it completely forgets old knowledge. 
Using a distillation strategy enables the network to reduce the catastrophic forgetting: LwF obtains $21.1\%$ on past classes, ILT $22.9\%$, and LwF-MC $34.2\%$. Regarding new classes, LwF is the best strategy, exceeding LwF-MC by $18.9\%$ and ILT by $6.6\%$.
However, \ours is far superior to all others, improving on the first classes and on the new ones. Moreover, we can observe that we are close to the joint training upper bound, especially considering new classes, where the gap with respect to it is only $0.3\%$.
In Figure \ref{fig:ILSS-qualitative} we report some qualitative results which demonstrate the superiority of \ours compared to the baselines.

\myparagraph{Multi-step addition of 50 classes (\textit{100-10}).} We then evaluate the performance on multiple incremental steps: we start from 100 classes and we add the remaining classes 10 by 10, resulting in 5 incremental steps. In Table \ref{tab:ILSS-ade-10} we report the results on all sets of classes after the last learning step.
In this setting the performance of FT, LwF and ILT are very poor because they strongly suffers catastrophic forgetting. 
LwF-MC demonstrates a better ability to preserve knowledge on old classes, at the cost of a performance drop on new classes.
Again, \ours achieves the best trade-off between learning new classes and preserving past knowledge, outperforming LwF-MC by $11.6\%$ considering all classes. 

\myparagraph{Three steps of 50 classes (\textit{50-50}).} Finally, in Table \ref{tab:ILSS-ade-50} we analyze also the performance on three sequential steps of 50 classes.
Previous \icl\ methods achieve different trade-offs between learning new classes and not forgetting old ones. LwF and ILT obtain a good score on new classes, but they forget old knowledge. On the contrary, LwF-MC preserves knowledge on the first 50 classes without being able to learn new ones.
\ours outperforms all the baselines by a large margin with a gap of $11.9\%$ on the best performing baseline, achieving the highest mIoU on every step. Remarkably, the highest gap 
is on the intermediate step, where there are classes that we must 
 both learn incrementally and preserve from forgetting on the subsequent learning step. 
 
\subsection{Conclusions}
{
In this section, we studied the incremental class learning problem for semantic segmentation, analyzing the realistic scenario where the new training set does not provide annotations for old classes, leading to the semantic shift of the background class and exacerbating the catastrophic forgetting problem. We address this issue by proposing a novel objective function and a classifiers' initialization strategy which allows our network to explicitly model the semantic shift of the background, effectively learning new classes without deteriorating its ability to recognize old ones. Results show that \ours outperforms regularization-based \icl\ methods by a large margin, considering both small and large scale datasets. We believe that our problem formulation, our approach and our extensive comparison with previous methods will encourage future works on this novel research topic, especially in the direction of effectively including the semantic shift in the background class in \icl\  models in semantic segmentation.

In Sections \ref{sec:IL-multitask} and Sections \ref{sec:IL-semantic-seg}, we focused on the multi-domain and incremental learning problem respectively, incrementally adding new semantic task/concepts to a pre-trained model. However, in both these tasks, the underlying assumption is that the images will contain only objects we have seen during training or that we can safely consider as background. A more realistic problem is equipping models with the ability to not only recognizing semantic concepts and incrementally learn new ones, but also detecting if an image contains a previously unseen semantic category. In the next section, we will show how we can address this problem in the framework of open-world recognition.
}

\section[Open World Recognition]{Open World Recognition \footnotemark\footnotetext{M. Mancini, H. Karaoguz, E. Ricci, P. Jensfelt, B. Caputo. {\sl Knowledge is Never Enough: Towards Web Aided Deep Open World Recognition}. IEEE International Conference on Robotics and Automation (ICRA) 2019.}\footnotemark\footnotetext{D. Fontanel, F. Cermelli, M. Mancini, S. Rota Bul\'o, E. Ricci, B. Caputo. {\sl Boosting Deep Open World Recognition by Clustering}. IEEE Robotics and Automation Letters 2020. }}
\label{sec:IL-owr}

In the previous sections, we have discussed how new knowledge in terms of classification tasks (Section \ref{sec:IL-multitask}) and semantic concepts (Section \ref{sec:IL-semantic-seg}) can be added to a pre-trained model. In particular, in Section \ref{sec:IL-semantic-seg}, we showed how it is possible to have a model whose output space contains all the concepts incrementally learned by the model. However, all the models discussed so-far rely on a simple assumption: all the categories we are interested in recognize are contained in our output space. This closed-world assumption (CWA) is unrealistic for agents acting in the real-world. Indeed it is impossible to capture all existing semantic concepts in a single training set unless we are in a very constrained scenario. In this section, we take a step forward and we show how we can break the CWA developing two visual systems able to work in the \textit{open world}.

\begin{figure}[tb]
  \centering
  \includegraphics[width=\columnwidth]{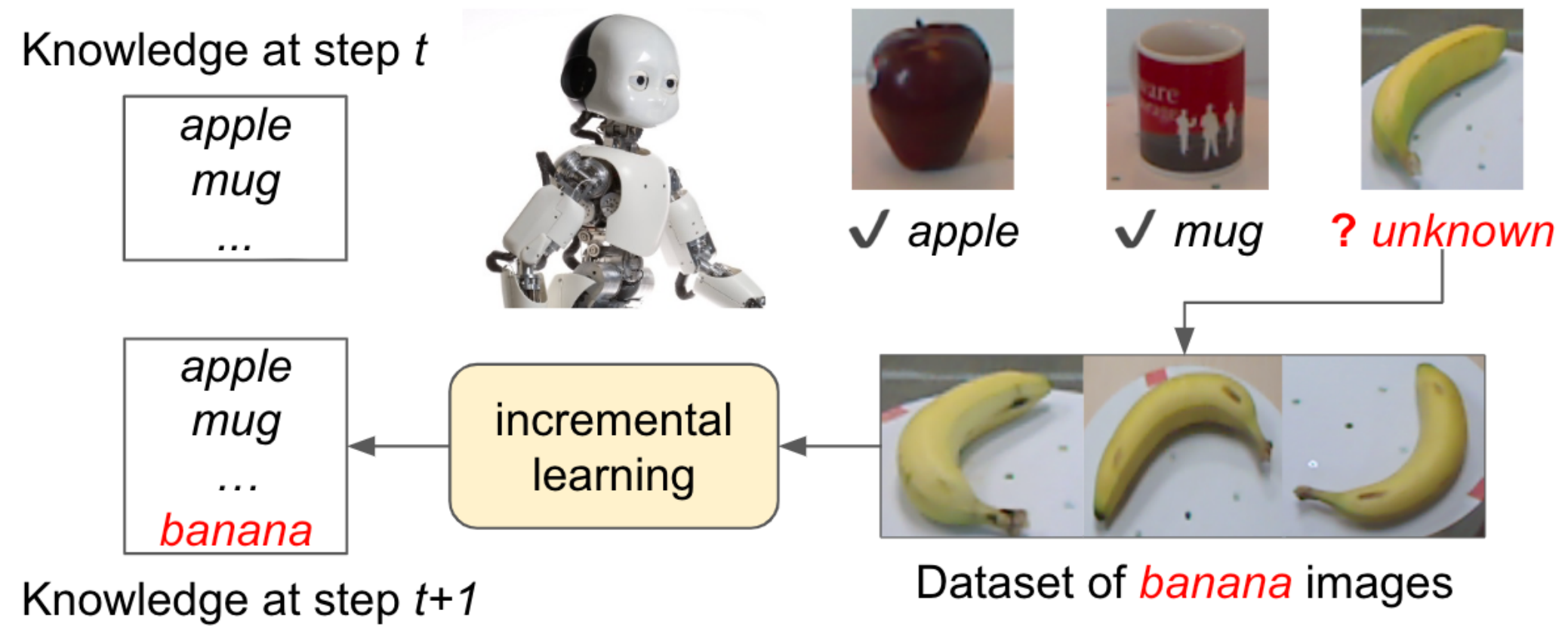} 
   \caption{In the open-world scenario a robot must be able to classify correctly known objects, (\textit{apple} and \textit{mug}), and detect novel semantic concepts (e.g. \textit{banana}). When a novel concept is detected, it should learn the new class from an auxiliary dataset, updating its internal knowledge.}
  \label{fig:teaser}
 \end{figure}

To clarify our goal, let us consider the example shown in Fig.~\ref{fig:teaser}. The robot has a knowledge base composed by a limited number of classes. Given an image containing an unknown concept (e.g. banana), we want the robot to detect it as unknown and being able to add it to its knowledge base in subsequent learning stages. 
To accomplish this goal, it is very important for a robot vision system to have two crucial abilities: (i) it must be able to recognize already seen concepts and detect unknown ones
(i.e. open set recognition), and (ii) it must be able to extend its knowledge base with new classes 
(i.e. incremental learning), without forgetting the already learned ones and without access to old training sets, avoiding \textit{catastrophic forgetting} \cite{mccloskey1989catastrophic}). 
While open set recognition \cite{scheirer2012toward, fragoso2013evsac, li2005open} and incremental learning \cite{rebuffi2017icarl,camoriano2016incremental,camoriano2017incremental,valipour2017incremental} are well-studied problems in the literature, few works proposed a solution to solve them together \cite{bendale2015towards,de2016online}.  Standard approaches for open world recognition (OWR) equip the nearest class mean (NCM) classification algorithm with a rejection option based on an estimated threshold. While standard approaches \cite{bendale2015towards,de2016online} use shallow features, in this section we take a step forward, proposing two deep models for open world recognition. 

The first model we will discuss builds on recent work by Guerriero et al. \cite{guerriero2018deep} and is (up to our knowledge) the first deep open world recognition architecture in the literature. This approach couples the flexibility of non-parametric classification methods, necessary to add incrementally new classes over time and able to estimate a probability score for each known class supporting the detection of new classes (Nearest Non Outlier, NNO \cite{bendale2015towards}), with the powerful intermediate representations learned by deep networks. We enable end-to-end training of the architecture through an online approximate estimate and update function for the mean prototype representing each known class and for the threshold allowing to detect novel classes in a life-long learning fashion. We name this approach \textbf{DeepNNO} (\textbf{Deep} \textbf{N}earest \textbf{N}on-\textbf{O}utlier)\cite{mancini2019knowledge}.

The second model improves DeepNNO by forcing the deep architecture used as feature extractor to cluster appropriately samples belonging to the same class, while pushing away samples of other classes. For this reason, it introduces a global clustering loss term that aims at keeping closer the features of samples belonging to the same class to their class centroid. Furthermore, we show how the soft nearest neighbor loss \cite{salakhutdinov2007learning, frosst2019analyzing} can be successfully employed as a local clustering loss term in order to force pair of samples of the same class to be closer in the learned metric space than relative sample points of other classes. 
Moreover, differently from DeepNNO and previous shallow works \cite{bendale2015towards} 
we avoid to estimate a global rejection threshold on the model predictions based on heuristic rules but we (i) define an independent threshold for each class and (ii) we explicitly learn the thresholds by using a margin-based loss function which balances rejection errors on samples of a reserved memory held-out from the training. We name this approach \textbf{\owr} (\owrFull) \cite{fontanel2020boosting}.

We evaluate DeepNNO and \owr on Core50 \cite{lomonaco2017core50}, RGB-D Object Dataset \cite{lai2011large} and CIFAR-100 \cite{krizhevsky2009learning} datasets, showing experimentally that DeepNNO outperforms previous OWR methods and \owr show increased effectiveness in both detecting novel classes and adding new classes to the set of known ones. 

The outline of this section is as follows. We start by giving a more formal definition of the OWR problem (Section \ref{sec:OWR-problem}) and some preliminaries on the NCM \cite{mensink2012metric,guerriero2018deep} and NNO \cite{bendale2015towards,de2016online} algorithms which serve as starting point for our approaches (Section \ref{sec:OWR-preliminaries}). We then describe DeepNNO (Section \ref{sec:OWR-preliminaries}) and \owr (Section \ref{sec:OWR-preliminaries}), showing their results on the aforementioned benchmarks (Section \ref{sec:OWR-preliminaries}). We conclude by providing a perspective toward autonomous visual systems with preliminary experiments on Web-aided OWR (Section \ref{sec:OWR-web}) and the conclusions (Section \ref{sec:OWR-concl}).

\subsection{Problem Formulation}
\label{sec:OWR-problem}
The goal of OWR is producing a model capable of (i) recognizing known concepts (i.e. classes seen during training), (ii) detecting unseen categories (i.e. classes not present in any training set used for training the model) and (iii) incrementally add new classes as new training data is available. 
Formally, let us denote as $\mathcal{X}$ and $\mathcal{Y}$ the input space (i.e. image space) and the closed world output space respectively (i.e. set of known classes).  Moreover, since our output space will change as we receive new data containing novel concepts, we will denote as $\mathcal{Y}_t$ the set of classes seen after the $t_{\text{th}}$ incremental step, with $\mathcal{Y}_0$ denoting the category present in the first training set. Additionally, since we aim to detect if an image contains an unknown concept, in the following we will denote as $unk$ the special unknown class, building the output space as $\mathcal{Y}_t \cup \{unk\}$. We assume that, at each incremental step, we have access to a training set $\mathcal{T}_t = \{(x^t_1 , c^t_1), \cdots, (x^t_{N_t}, c^t_{N_t}) \}$, with $N_t=|\mathcal{T}_t|$, $x^t \in \mathcal{X}$, and $c^t \in \mathcal{C}_t$, where $\mathcal{C}_t$ is the set of categories contained in the training set $\mathcal{T}_t$.
Note that, without loss of generality, in each incremental step, we assume to see a new set of classes $\mathcal{C}_i\cap\mathcal{C}_j = \emptyset$ if $i\neq j$. The set of known classes at step $t$ is computed as $\mathcal{Y}_t = \cup_{i=0}^t \mathcal{C}_i $ and given a sequence of $S$ incremental steps, 
our goal is to learn a model mapping input images to either their corresponding label in $\mathcal{Y}_S$ or to the special class $unk$. 
In the following we will split the classification model into two components: a feature extractor $f$ that maps the samples into a feature space and a classifier $g$ that maps the features into a class label, i.e. $g(f(x))=c$ with $c \in \{\mathcal{Y}_S,unk\}$.


\subsection{Preliminaries}
\label{sec:OWR-preliminaries}
Standard approaches to tackle the OWR problem apply non-parametric classification algorithms on top of learned metric spaces \cite{bendale2015towards, de2016online}. A common choice for the classifier $g$ is the Nearest Class Mean (NCM) \cite{mensink2012metric, guerriero2018deep}.
NCM works by computing a centroid for each class (i.e. the mean feature vector) and assigning a test sample to the closest centroid in the learned metric space. Formally, we have:
\begin{equation}
    g^{\text{NCM}}(x)=\argmin_{c \in \mathcal{C}_t} d(f(x), \mu_c)
\end{equation}
where $d(\cdot,\cdot)$ is a distance function (e.g. Euclidean) and $\mu_c$ is the mean feature vector for class $c$. The standard NCM formulation cannot be applied in the OWR setting since it lacks the inherent capability of detecting images belonging to unknown categories. To this extent, in \cite{bendale2015towards} the authors extend the NCM algorithm to the OWR setting by defining a rejection criterion for the unknowns. In this extension, called Nearest Non-Outlier (NNO), class scores are defined as: 
\begin{equation} 
\label{eq:nno-score} s_c^\mathtt{NNO}(x)=\mathcal{Z} (1-\frac{d(f(x), \mu_c)}{\tau}), 
\end{equation}
where $\tau$ is the rejection threshold and $\mathcal{Z}$ is a normalization factor. The final classification is held-out as:
\begin{equation}
\label{eq:rejection-nno}
    g(x) = 
  \begin{cases}
    unk & \text{if}\; s_c^\mathtt{NNO}(x) \leq 0\; \forall{c \in \mathcal{Y}_t},\\
    g^{\text{NCM}}(x) & \text{otherwise}.
  \end{cases}
\end{equation}
{Following \cite{mensink2012metric}, in \cite{bendale2015towards} the features are linearly projected into a metric space defined by a matrix $W$ (i.e. $f(x)=W\cdot x$)}, 
with $W$ learned on the first training set $\mathcal{T}_0$ and kept fixed during the successive learning steps.
The main limitation of this approach is that new knowledge will be incorporated in the classifier $g$ without updating the feature extractor $f$ accordingly. 
In the next section, we show how the performance of NNO can be significantly improved by using as $f$ a deep architecture trained end-to-end in each incremental step.

\subsection{Deep Nearest Non-Outlier}
\label{sec:OWR-deepnno}
The DeepNNO algorithm is obtained from NNO with the following modifications: (i)
the feature extractor function is replaced with deep representations derived from neural network layers; (ii)
an online update strategy is adopted for the mean vectors $\mu_c$; (iii) an appropriate loss is optimized 
using stochastic gradient descent (SGD) methods in order to 
compute the feature representations and the associated class specific means.

First, inspired by the recent work \cite{guerriero2018deep},
we 
replace the feature extractor function $f(\cdot)$ with deep representations derived from a neural network $f_\theta(\cdot)$
and define the class-specific probability scores as follows:
\begin{equation}
\label{eq:deep-nno-prob}
s_c^{\mathtt{DNNO}}(x)=\exp\left(-\frac{1}{2}||f_\theta(x) - \mu_c ||\right) .
\end{equation}
Note that, differently from \cite{bendale2015towards}, we do not consider explicitly the matrix $W$ since this is replaced by the network parameters $\theta$. 
Furthermore, we avoid to use a clamping function as this could hamper the gradient flow within the network. 
This formulation is similar to the NNO version proposed in \cite{de2016online} which have been showed to be more effective than that in \cite{bendale2015towards} for online scenarios. 

In OWR the classification model must be updated as new samples arrive. In DeepNNO this translates into incrementally updating the feature representations $f_\theta(x)$ and defining an appropriate strategy for updating the class mean vectors. 
Given a mini-batch of samples $\mathcal{B}=\{(x_1,c_1),\dots, (x_b,c_b)\}$, we compute the mean vectors through:
\begin{equation}
\label{eq:deep-nno-means}
\mu_c^{t+1}=\frac{n_c\cdot\mu_c^{t}+n_{c,\mathcal{B}}\cdot\mu_c^{\mathcal{B}}}{n_c+n_{c,\mathcal{B}}}
\end{equation}
where $n_{c}$ represents the number of samples belonging to class $c$ seen by the network until the current update step $t$, $n_{c,\mathcal{B}}$ represents the number of samples belonging to class $c$ in the current batch and $\mu_c^{\mathcal{B}}$ represents the current mini-batch mean vector
relative to the features of class $c$.

Given the class-probability scores in DeepNNO we define the following prediction function:
\begin{equation}
\label{eq:nno-predictiondeep}
c^*=\begin{cases}
      unk & \text{if}\ s_c^{\mathtt{DNNO}}(x) \leq \Delta \ \ \ \forall c \in \mathcal{Y}_t\\
      \argmax_{c\in\mathcal{Y}_t} s_c^{\mathtt{DNNO}}(x) & \text{otherwise}
    \end{cases}
\end{equation}
where $\Delta$ is a threshold which, similarly to the parameter $\tau$ in Eqn.(\ref{eq:nno-score}), regulates the number of samples that are assigned to a new class. While in \cite{bendale2015towards} $\tau$ is a user defined parameter which is kept fixed, in this subsection we argue that a better strategy is to dynamically update $\Delta$ since the feature extractor function and the mean vectors change during training. {Intuitively, while training the deep network, an estimate of $\Delta$ can be obtained by looking at the probability score given to the ground truth class. If the score is higher than the threshold, the value of $\Delta$ can be increased. Oppositely, the value of the threshold is decreased if the prediction is rejected. Specifically, given a mini-batch $\mathcal{B}$ we update $\Delta$ as follows:
\begin{equation}
\label{eq:deep-nno-tau-update}
\Delta^{t+1}=\frac{1}{t+1}\left(t\cdot\Delta^t+\frac{1}{C_\mathcal{B}}\sum_{c\in\mathcal{Y}_t}\bar{s}^{\mathtt{DNNO}}_{c,\mathcal{B}}\right)
\end{equation}
where 
$C_\mathcal{B}$ is the number of classes in $\mathcal{Y}_t$ represented by at least one sample in $\mathcal{B}$ and $\bar{s}^{\mathtt{DNNO}}_{c,\mathcal{B}}$ is the weighted average probability score of instances of class $c$ within the batch. Formally we consider:
\begin{equation}
\label{eq:deep-nno-tau-batch-avg}
\bar{s}^{\mathtt{DNNO}}_{c,\mathcal{B}}=\frac{1}{\eta_{\mathcal{B},k}}\sum_{i=1}^{\text{b}}w_{c,i}\cdot s^{\mathtt{DNNO}}_{c}(x_i)
\end{equation}
where $\eta_{\mathcal{B},k}=\sum_{i=1}^\text{b} w_{c,i}$ is a normalization factor and:
\begin{equation}
\label{eq:nno-ws}
w_{c,i}=\begin{cases}
     w^+ & \text{if}\ c_i=c \wedge s_c^{\mathtt{DNNO}}(x_i)>\Delta \\
     w^- & \text{if}\ c_i=c \wedge s_c^{\mathtt{DNNO}}(x_i)\leq \Delta \\
      0 & \text{otherwise}
    \end{cases}
\end{equation}
where $w^-$ and $w^+$ are scalar parameters which allow to assign different importance to samples for which the scores given to the ground truth class are respectively rejected or not by the current threshold $\Delta$. 

}

To train the network, we employ standard SGD optimization, minimizing the binary cross entropy loss over the training set:
\begin{equation}
 \mathcal{L}=\frac{1}{|\mathcal{T}_t|}\sum_i\ell_{CL}(x_i,c_i) 
\end{equation}
where:
\begin{equation}
 \ell_{{CL}}(x_i,c_i) =  -\log s^{\mathtt{DNNO}}_{c_i}(x_i) - \sum_{c \in \mathcal{Y}_t} \mathds{1}_{c\neq c_i} \log \left(1 - s_c^{\mathtt{DNNO}}(x_i)\right)
\end{equation}
After computing the loss, we use standard backpropagation to update the network parameters $\theta$. After updating $\theta$, we use the samples of the current batch to update both the class mean estimates $\mu_c$ and the threshold $\Delta$, 
using Eqn.\eqref{eq:deep-nno-means} and Eqn.\eqref{eq:deep-nno-tau-update} respectively.

To allow our model for incremental learning of our deep neural network, we exploit two additional components. Following standard rehearsal-based approaches for incremental learning \cite{rebuffi2017icarl, chaudhry2018riemannian, castro2018end}, the first is a memory which stores the most relevant samples of each class in $\mathcal{Y}_t$. The relevance of a sample $(x,k)$ is determined by its distance $d_c(x)$ to the class mean $\mu_c$ \ie the lower is the distance, the higher is the relevance of the sample. The memory is used to augment the training set $\mathcal{T}_{t+1}$, allowing to update the mean estimates of the classes in $\mathcal{Y}_t$ as the network is trained using samples of novel ones. In order to avoid an unbounded growth, the size of the memory is kept fixed and it is pruned after each incremental step to make room for instances of novel classes. The pruning is performed by removing, for each class in $\mathcal{Y}_t$, the instances with lowest relevance. 

The second component is a batch sampler 
which makes sure that a given ratio of the batch is composed by samples taken from the memory, independently from the memory size. This allows to avoid 
biasing the incremental learning procedure towards novel categories, in the case their number of samples is much larger than the memory size. {Additionally, we add a distillation loss \cite{hinton2015distilling} which act as regularizer and 
 avoids the forgetting of previously learned features.} Denoting as $f_\theta^{\mathcal{Y}_{t-1}}$ the network trained on the set of known classes, the distillation loss is defined as:
\begin{equation}
\label{eq:owr-distill}
\ell_{\text{DS}}(x_i) =||f_\theta(x)- f_\theta^{\mathcal{Y}_{t-1}}(x)||
\end{equation}
The overall loss is thus defined as:
\begin{equation}
\label{eq:owr-dnno-loss}
 \mathcal{L}_\mathtt{DNNO}=\frac{1}{|\mathcal{T}_t|}\sum_i \left( \ell_{\text{CL}}(x_i,c_i) + \lambda \ell_{{DS}}(x_i) \right)
\end{equation}
where $\lambda$ is an hyperparameter balancing the contribution of $\ell^{\text{distill}}$ within $\mathcal{L}$.

\subsection{Boosting Deep Open World Recognition}
\label{sec:OWR-dario}
Despite its experimental effectiveness (see Section \ref{sec:OWR-exps}), DeepNNO has two main drawbacks. First, the learned feature representation $f$ is not forced to produce predictions clearly \textit{localized} in a limited region of the metric space. 
{Indeed, constraining the feature representations of a given class to a limited region of the metric space allows to have both more confident predictions on seen classes and producing clearer rejections also for images of unseen concepts.} Second, having an heuristic strategy for setting the threshold is sub-optimal with no guarantees on the robustness of the choice. In the following, we will detail how we provide solutions to both problems in \owr.

\begin{figure}[t]
    \centering
    \centering
    \includegraphics[width=\linewidth]{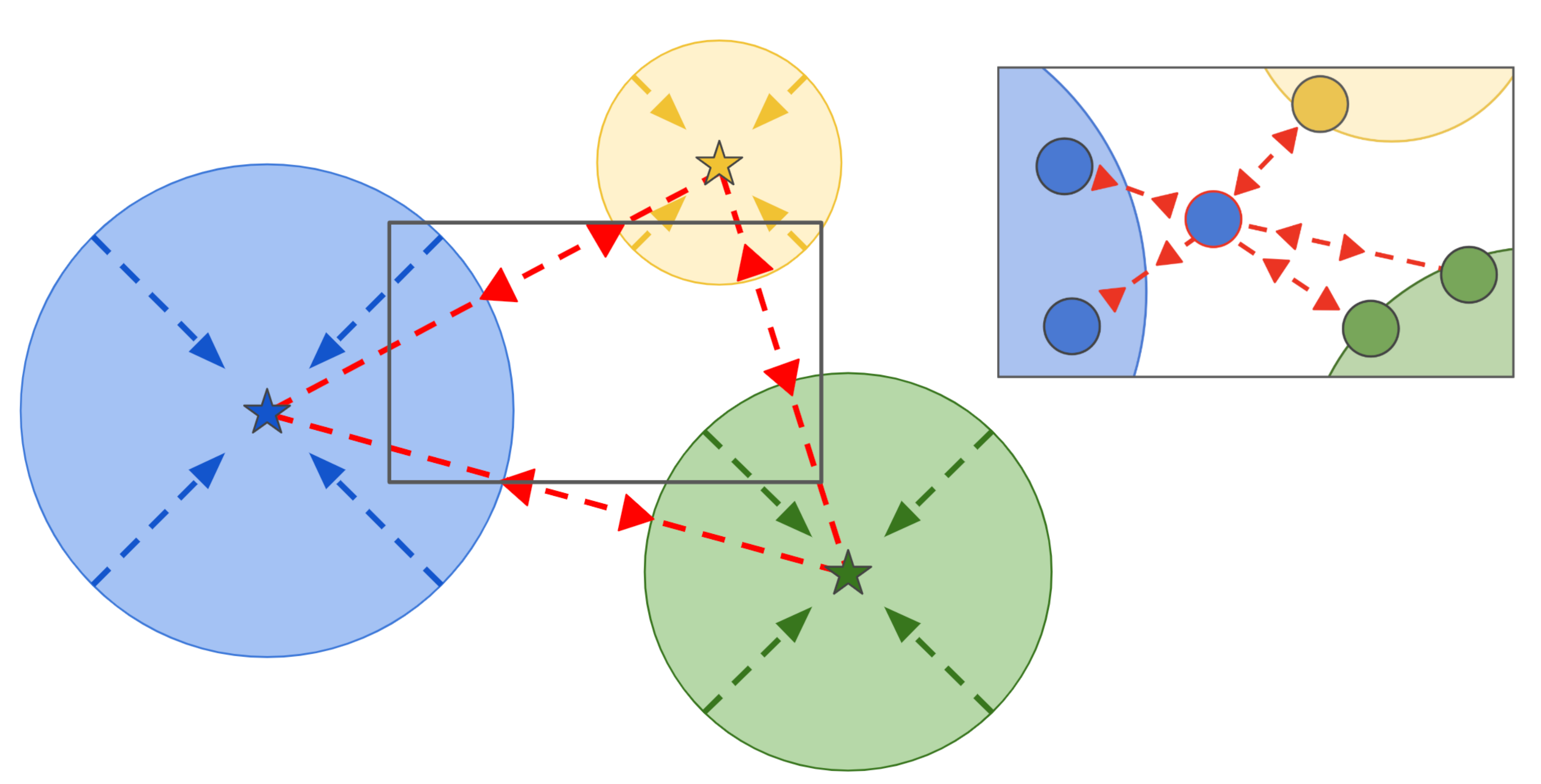} 
    \caption{Overview of the \owr global to local clustering. The global clustering (left) pushes sample representations closer to the centroid (star) of the class they belong to. 
    The local clustering (right), instead, forces the neighborhood of a sample in the representation space to be semantically consistent, pushing away samples of other classes.}
    \label{fig:owr-method1}
\end{figure}

{To obtain feature representations clearly localized in the metric space based on their semantic, 
we propose to use a pair of losses enforcing clustering.} 
In particular, we use a \textit{global} term which forces the network to map samples of the same class close to their centroid (Fig.\ref{fig:owr-method1}, left) and a \textit{local} clustering term which constrains the neighborhood of a sample to be semantically consistent, i.e. to contain samples of the same class (Fig.\ref{fig:owr-method1}, right). 
In the following we describe the two clustering terms. 

\myparagraph{Global Clustering}.
The global clustering term aims to reduce the distance between the features of a sample with the centroid of its class. To model this, we took inspiration from what has been proposed in \cite{mensink2012metric} and we employ a cross-entropy loss with the probabilities obtained through the distances among samples and class centroids.
Formally, given a sample $x$ and its class label $c$, we define the global clustering term as follows:
\begin{equation} 
    \ell_{GC}(x, c) = - \log \frac{s^{\mathtt{\owr}}_c(x)}{ \displaystyle\sum_{k \in \mathcal{Y}_t} s_c(x)}.
\end{equation}
The class-specific score $s_c(x)$ is defined as:
\begin{equation} \label{eq:prob-softmax} 
    s^{\mathtt{\owr}}_c(x) = \frac{e^{-\frac{1}{T}||f_\theta(x) - \mu_c||^2}}{ \displaystyle\sum_{k \in \mathcal{C}_t}  e^{- \frac{1}{T}||f_\theta(x) - \mu_k||^2}}
\end{equation}
where $T$ is a temperature value which allows us to control the behavior of the classifier. We set $T$ as the variance of the activations in the feature space, $\sigma^2$, in order to normalize the representation space and increase the stability of the system. 
During training, $\sigma^2$ is the variance of the features extracted from the current batch while, at the same time, we keep an online global estimate of $\sigma^2$ that we use at test time. 
The class mean vectors $\mu_i$ with $i \in \mathcal{Y}_t$ as well as $\sigma^2$ are computed in an online fashion, as in DeepNNO. 

\myparagraph{Local Clustering}. 
To enforce that the neighborhood of a sample in the feature space is semantically consistent (i.e. given a sample $x$ of a class $c$, the nearest neighbours of $f(x)$ belong to $c$), we employ the soft nearest neighbour loss \cite{salakhutdinov2007learning, frosst2019analyzing}. 
{This loss has been proposed to measure the class-conditional entanglement of features in the representation space}. 
In particular, it has been defined as: 
\begin{equation}
    \label{SNNL}
    \ell_{LC}(x, c,\mathcal{B})  = - \log\ \  
    \frac{ \displaystyle\sum_{\mathclap{x_j \in \mathcal{B}_c\setminus \{x\}}
                               }
                      e^{- \frac{1}{T}||f_\theta(x) - f_\theta(x_j)||^2}}{
          \displaystyle\sum_{\mathclap{x_k \in \mathcal{B}\setminus \{x\}}}
             e^{- \frac{1}{T}||f_\theta(x) - f_\theta(x_k)||^2}
          }
\end{equation}
where T refers to the temperature value, $\mathcal{B}$ is the current training batch, and $\mathcal{B}_c$ is the set of samples in the training batch belonging to class $c$. Instead of performing multiple learning steps to optimize the value of T as proposed in \cite{frosst2019analyzing}, we use as $T=\sigma^2$ as we do in Eq.~\ref{eq:prob-softmax}. 

Intuitively, given a sample $x$ of a class $c$, a low value of the loss indicates that the nearest neighbours of $f(x)$ belong to $c$, while high values indicates the opposite (i.e. nearest neighbours belong to classes $i\in\mathcal{Y}_t$ with $i\neq c$). 
Minimizing this objective allows to enforce the semantic consistency in the neighborhood of a sample in the feature space.

\myparagraph{Knowledge distillation and full objective.}
As highlighted in the Section \ref{sec:OWR-deepnno}, to avoid forgetting old knowledge, we want the feature extractor to preserve the behaviour learned in previous learning steps. 
To this extent, as in DeepNNO, we introduce (i) a memory which stores the most relevant samples for classes in $\mathcal{Y}_t$ and (ii) a distillation loss which enforces consistency among the features extracted by $f$ and ones obtained by the feature extractor of the previous learning step, $f_{t-1}$. 
The distillation loss is computed as in Eq.~\eqref{eq:owr-distill}. As before, this loss is minimized only for incremental training steps, hence, only when $t>1$. Additionally, we apply also the same balanced batch sampling scheme of DeepNNO.

Overall, given a batch of samples $\mathcal{B}={\{(x_1 , c_1), \cdots, (x_{b}, c_{b})}\}$, we train the network to minimize the following loss:
\begin{equation}
\label{eq:owr-bdoc-loss}
    \mathcal{L}_{\mathtt{\owr}}=\frac{1}{|\mathcal{B}|} \sum_{(x,c) \in \mathcal{B}} \ell_{GC}(x,c) + \lambda\ \ell_{LC}(x,c,\mathcal{B}) + \gamma\ \ell_{DS}(x)
\end{equation}
with $\lambda$ and $\gamma$ hyperparameters weighting the different components. 

\begin{figure}[t]
    \centering
    \includegraphics[width=0.7\linewidth]{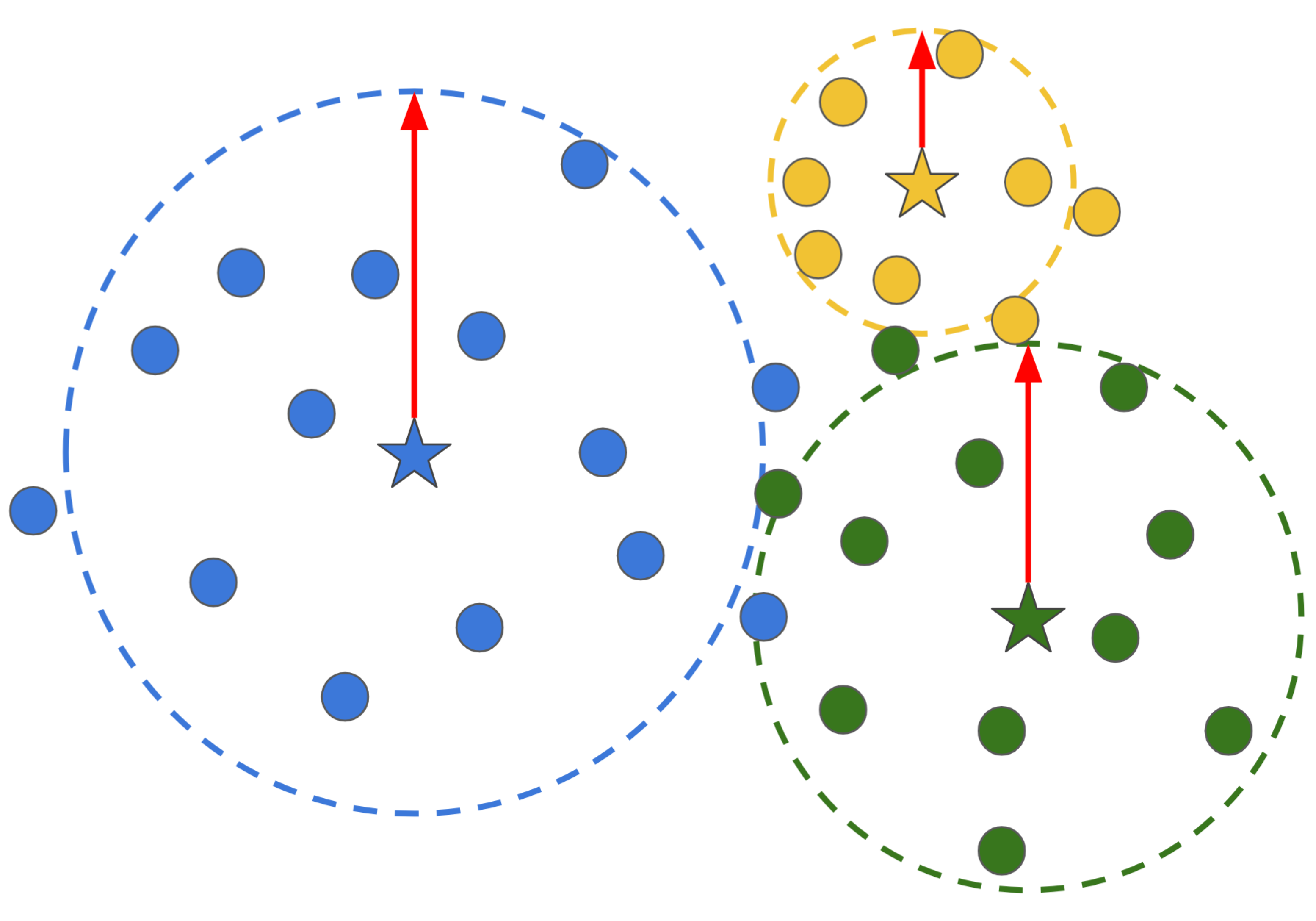} 
    \caption{
    Overview of how \owr learns the \textit{class-specific} rejection thresholds. The small circles represent the samples in the held out set. The dashed circles, having radius the maximal distance (red), represent the limits beyond which a sample is rejected as a member of that class. As it can be seen, the class-specific threshold is learned to reduce the rejection errors.} 
    \label{fig:method2}
\end{figure}

\myparagraph{Learning to detect the unknown}.
In order to extend the NCM-based classifier of \owr to work on the open set scenario, we explicitly learn class-specific rejection criterions. As illustrated in Fig.~\ref{fig:method2}, for each class $c$ we define the \textit{class-specific} threshold as the maximal distance $\Delta_c$ for which the sample belongs to $c$. Under this definition, the \owr classifier is:
\begin{equation}
\label{eq:rejection-our}
    g(x) = 
  \begin{cases}
    unk & \text{if}\, d(f_\theta(x),\mu_c)> \Delta_c,\; \forall{c \in \mathcal{Y}_t},\\
    \text{argmin}_c d(f_\theta(x),\mu_c)&\text{otherwise}
  \end{cases}
\end{equation}
with $d(x,y)=\frac{1}{\sigma^2}||x-y||^2$.
Instead of heuristically estimating or fixing a maximal distance, we explicitly learn it for each class be freezing the feature extractor $f_\theta$ and minimizing the following objective over the thresholds $\Delta_c$:
\begin{equation}
\label{eq:owr-bdoc-rejection-loss}
    \ell^{\Delta}_{MD}(x,c) = \sum_{k \in \mathcal{Y}_t} \max (0, m \cdot (\frac{1}{\sigma^2}||f_\theta(x) - \mu_k||^2 - \Delta_k))
\end{equation}
where $m=-1$ if $c=k$ and $m=1$ otherwise. 
The $\ell^\Delta_{MD}$ loss leads to an increase of $\Delta_c$ if the distance from a sample belonging to the class $c$ and the class centroid $\mu_c$ is greater than $\Delta_c$. Instead, if a sample not belonging to $c$ has a distance from $\mu_c$ less then $\Delta_c$, it increases the value of $\Delta_c$.

{Overall, the training procedure of \owr is made of two steps: in the first we train the feature extractor on the training set minimizing Eq.~\ref{eq:owr-bdoc-loss}, while in the second we learn the distances $\Delta_c$ on a set of samples which we held-out from training set.
To this extent, we split the samples of the memory in two parts, one used for updating the feature extractor $f$ and the centroids $\mu_c$ and the other part for learning the $\Delta_c$ values.}

\subsection{Experimental results}
\label{sec:OWR-exps}
In this subsection, we first introduce the experimental setting and the metrics used for the evaluation, then we report results of DeepNNO and \owr, showing ablation studies for each of their components.

 \begin{figure*}[t]
    \centering
    \subfloat[Closed World Without Rejection]{
        \includegraphics[width=0.48\textwidth]{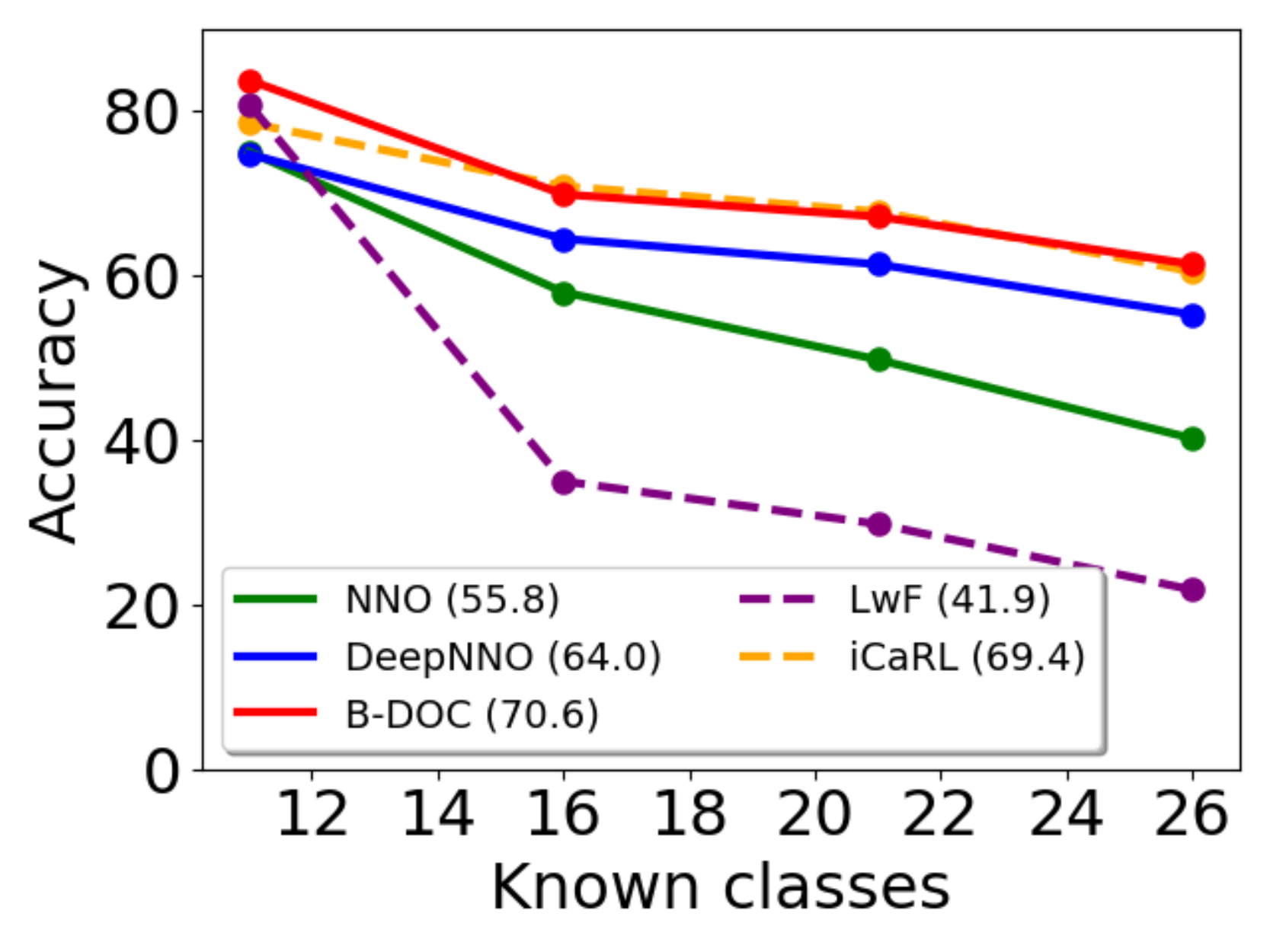}
        \label{fig:ROD-wor}
    }
    \hfill
    \subfloat[Closed World With Rejection]{
        \includegraphics[width=0.48\textwidth]{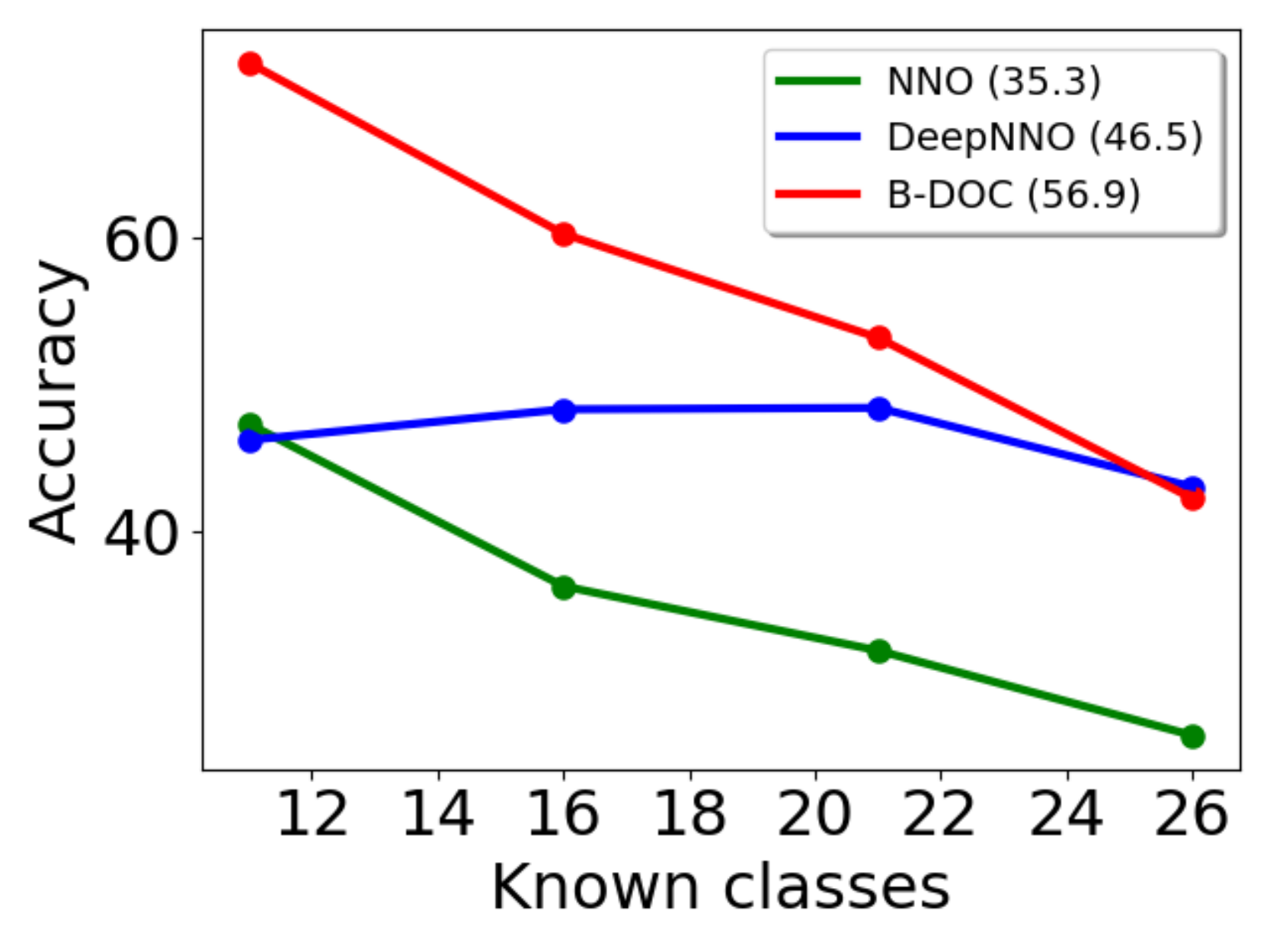}
        \label{fig:ROD-wir}
    }
    
    \subfloat[Open World Recognition Average]{
        \includegraphics[width=0.48\textwidth]{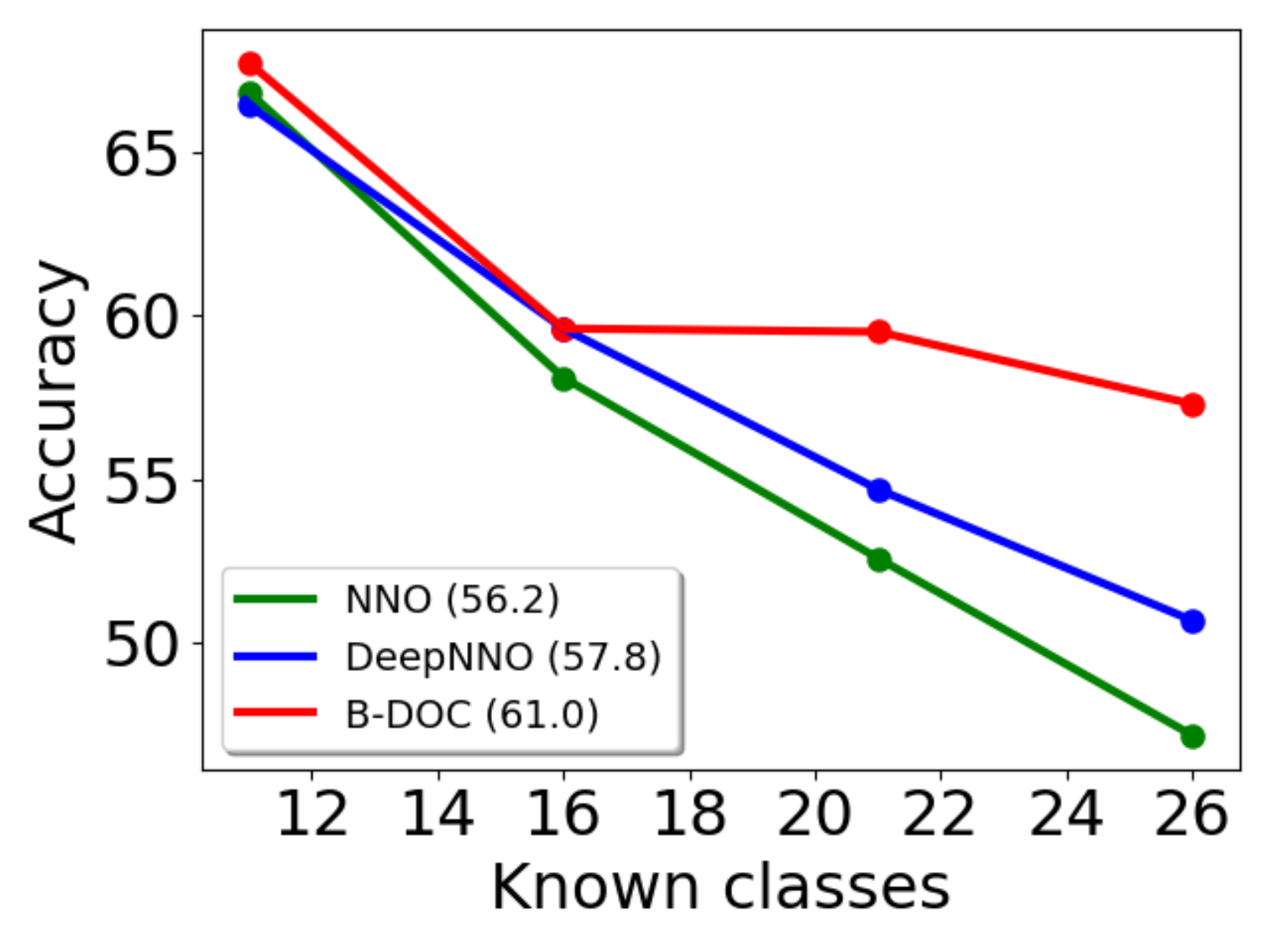}
        \label{fig:ROD-owra}
    }
   \hfill
    \subfloat[Open World Recognition Harmonic Mean]{
        \includegraphics[width=0.48\textwidth]{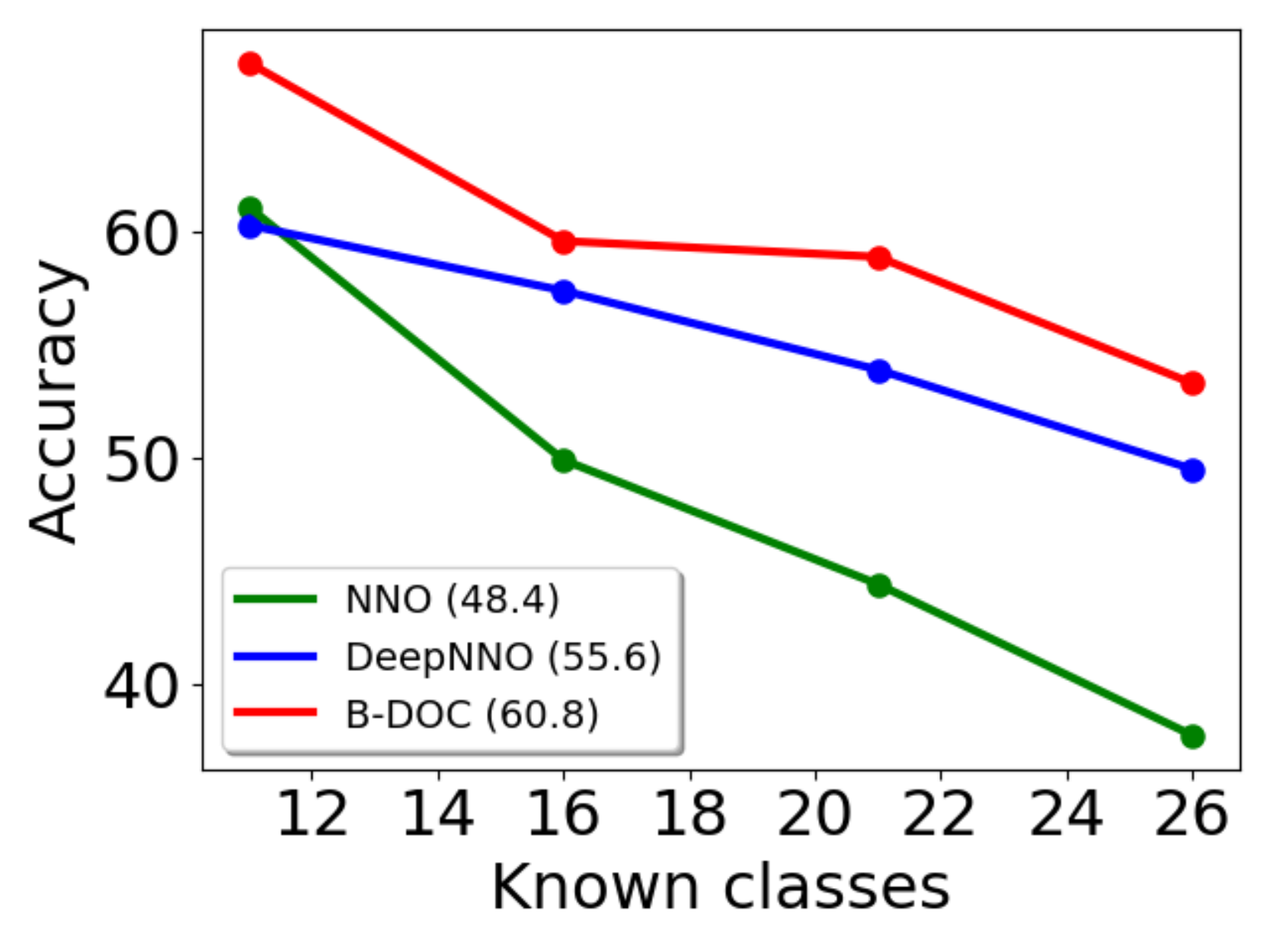} 
        \label{fig:ROD-owrh}
   }
     \caption{Comparison of NNO \cite{bendale2015towards}, DeepNNO and \owr on RGB-D Object dataset \cite{lai2011large}. The numbers in parenthesis denote the average accuracy among the different incremental steps.}
     \label{fig:ROD}
\end{figure*}

 \begin{figure*}[t]
    \centering
    \subfloat[Closed World Without Rejection]{
        \includegraphics[width=0.48\textwidth]{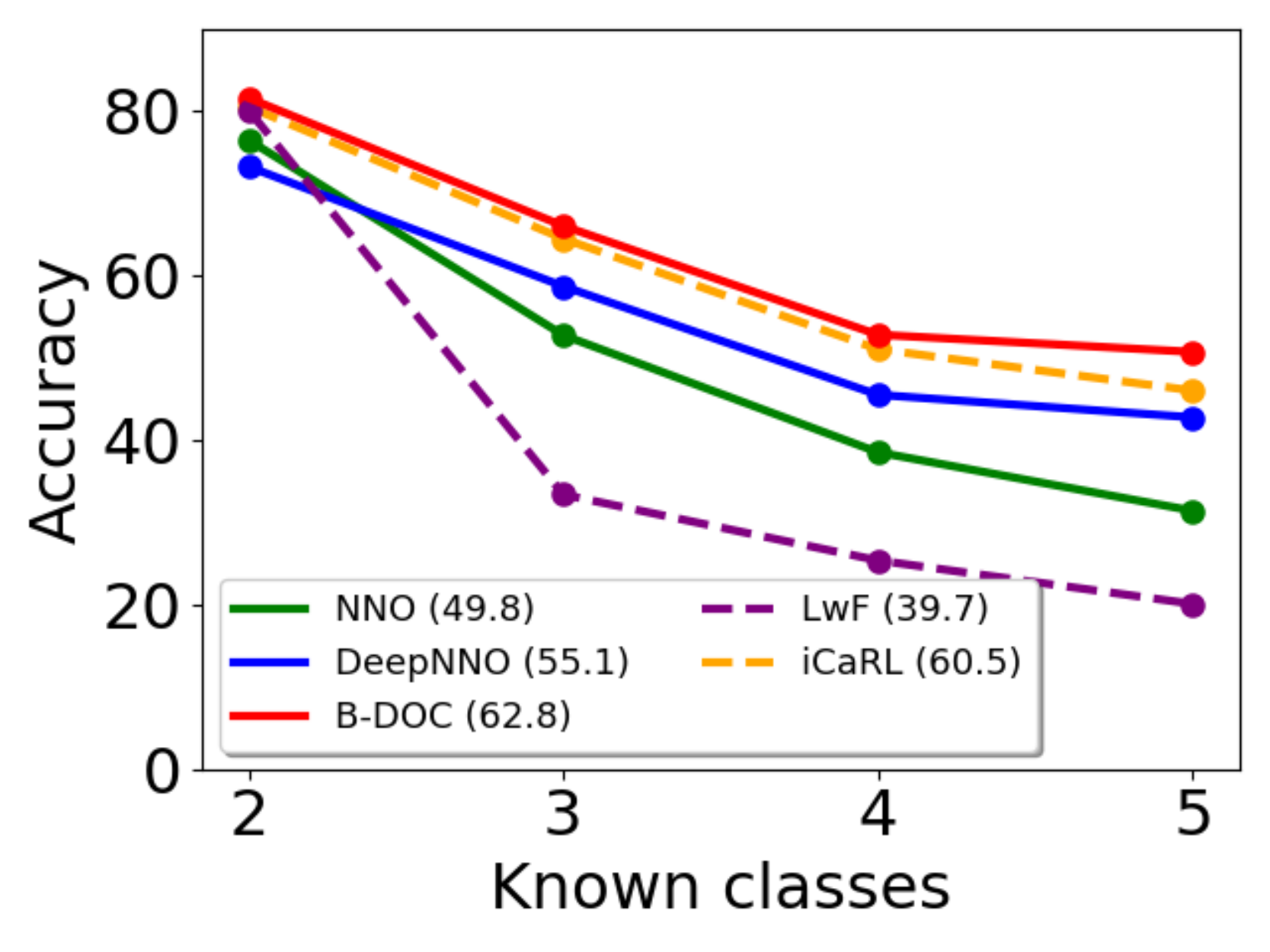}
        \label{fig:Core-wor}
    }
    \hfill
    \subfloat[Closed World With Rejection]{
        \includegraphics[width=0.48\textwidth]{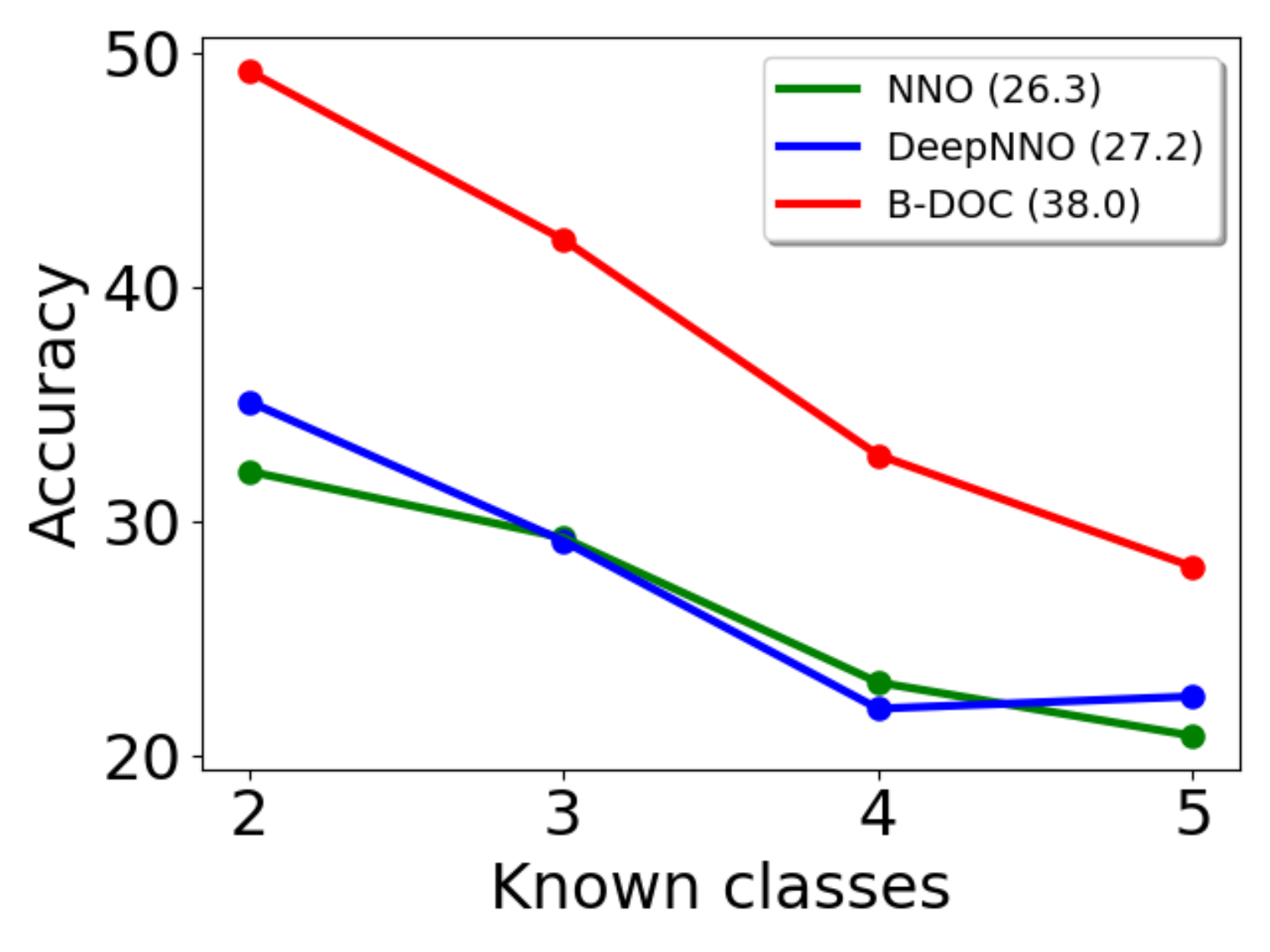}
        \label{fig:Core-wir}
    }
    
    \subfloat[Open World Recognition Average]{
        \includegraphics[width=0.48\textwidth]{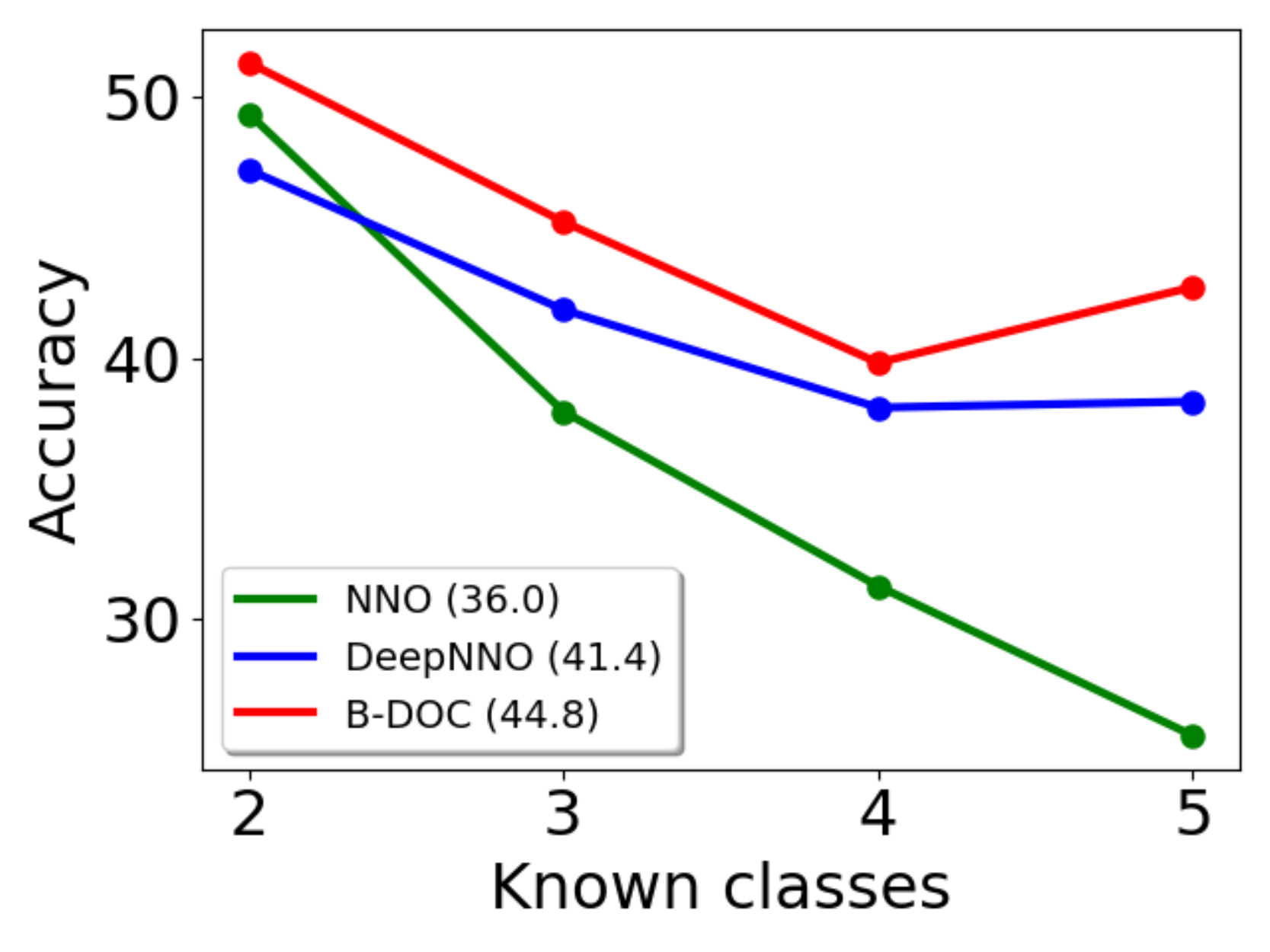}
        \label{fig:Core-owra}
    }
   \hfill
    \subfloat[Open World Recognition Harmonic Mean]{
        \includegraphics[width=0.48\textwidth]{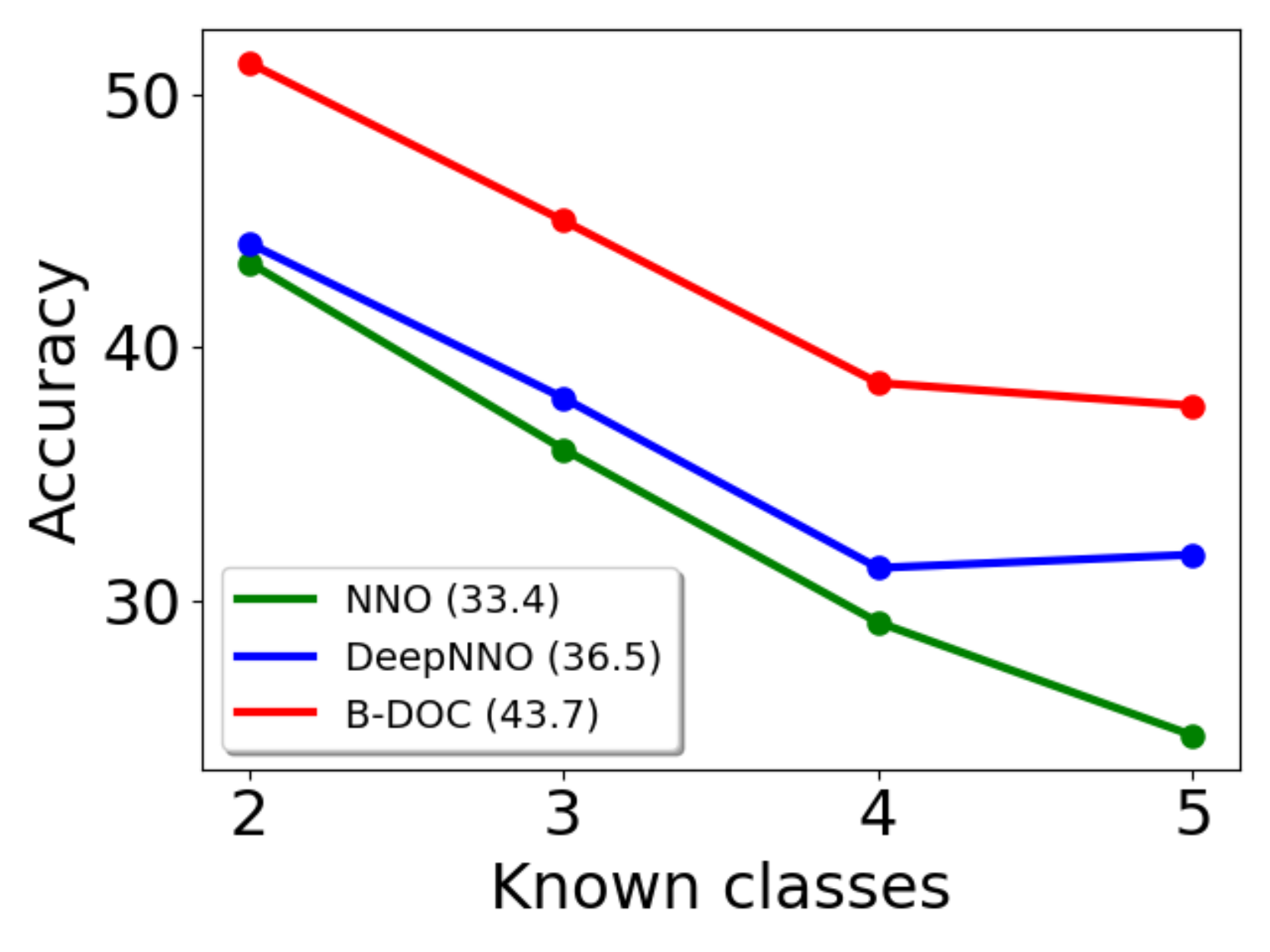} 
        \label{fig:Core-owrh}
   }
     \caption{Comparison of NNO \cite{bendale2015towards}, DeepNNO and \owr on Core50 \cite{lomonaco2017core50}. The numbers in parenthesis denote the average accuracy among the different incremental steps.}
     \label{fig:Core}
\end{figure*}

\begin{figure}[t]
    \centering
    \subfloat[Open World Recognition Average]{
        \centering
        \includegraphics[width=0.48\textwidth]{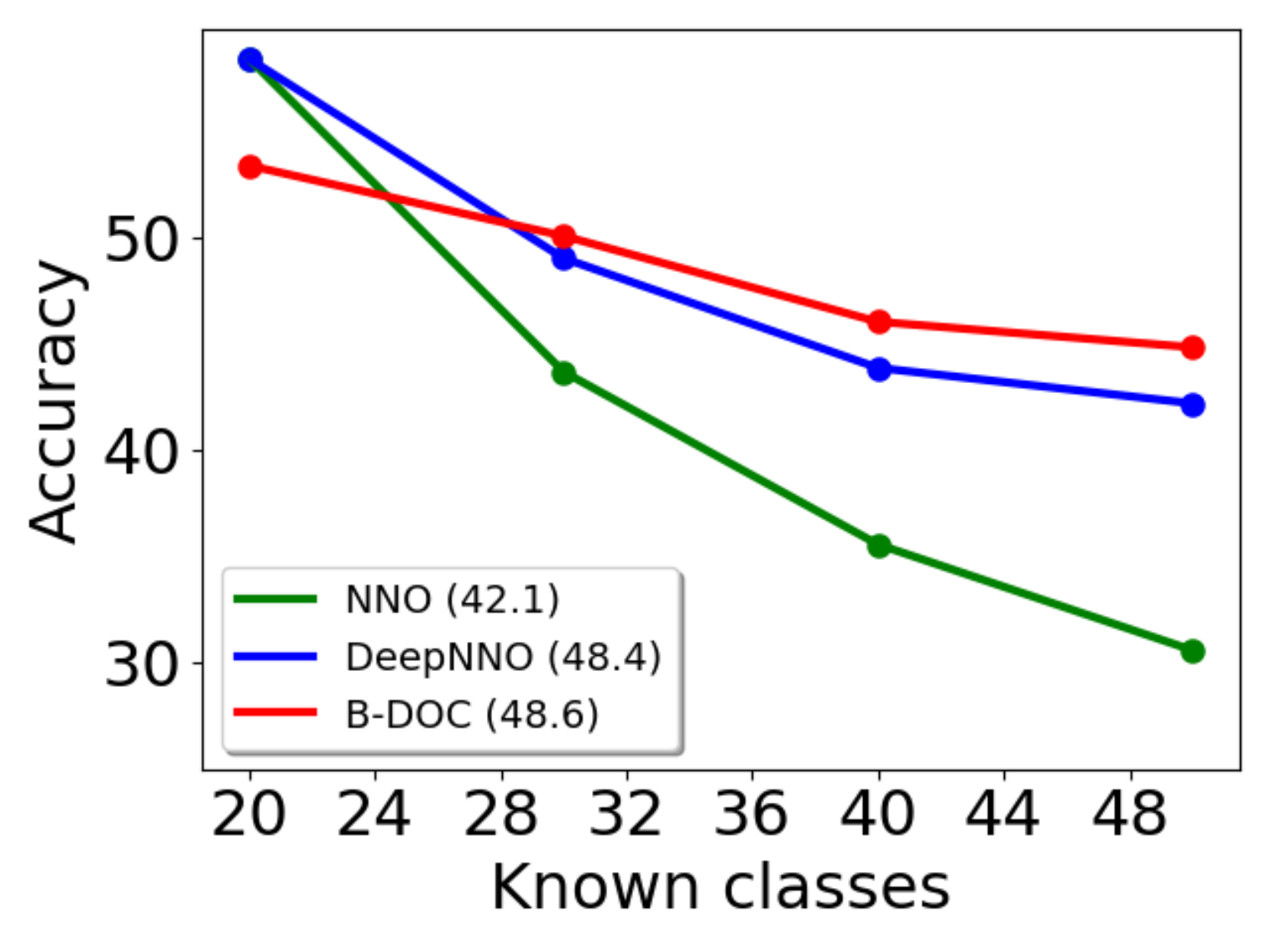}
        \label{fig:cifar-owra}    
        }
    \hfill
    \subfloat[Open World Harmonic Mean]{
        \centering
        \includegraphics[width=0.48\textwidth]{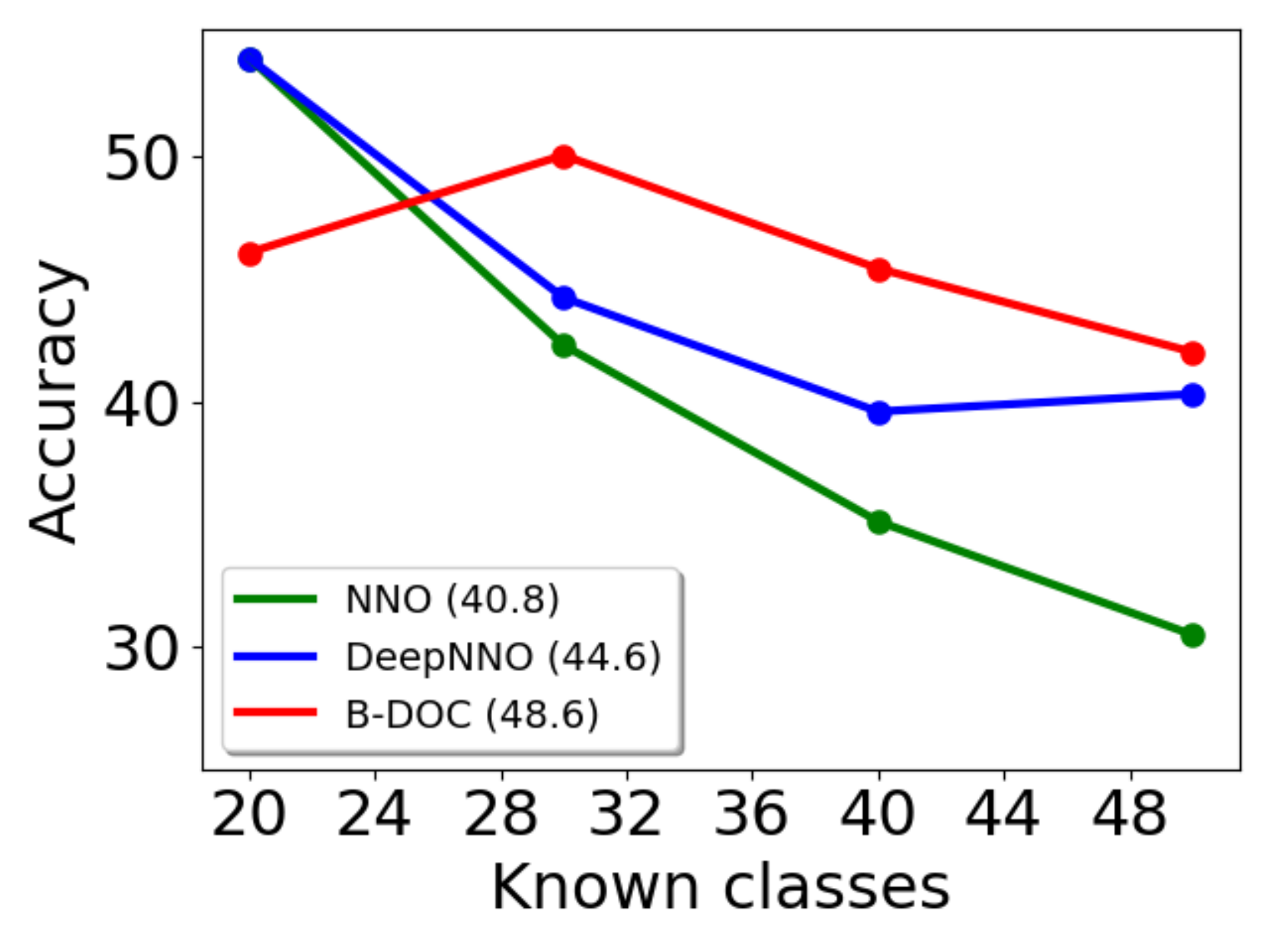} 
        \label{fig:cifar-owrh}
    }
    \caption{Comparison of NNO \cite{bendale2015towards}, DeepNNO and \owr on CIFAR-100 dataset \cite{krizhevsky2009learning}. The numbers in parenthesis denote the average accuracy among the different steps.}
    \label{fig:cifar}
\end{figure}

\myparagraph{Datasets and Baselines.} We assess the performance of our models on three datasets: RGB-D Object \cite{lai2011large} Core50 \cite{lomonaco2017core50} and CIFAR-100 \cite{krizhevsky2009learning}. The RGB-D Object dataset \cite{lai2011large} is one of the most used dataset to evaluate the ability of a model to recognize daily-life objects. It contains 51 different semantic categories that we split in two parts in our experiments: 26 classes are considered as known categories, while the other 25 are the set of unknown classes. Among the 26 classes, we consider the first 11 classes as the initial training set and we incrementally add the remaining classes in 4 steps of 5 class each. 
{As proposed in \cite{lai2011large}, we sub-sample the dataset taking one every fifth frame. For the experiments, we use the first train-test split among the original ones defined by the authors \cite{lai2011large}. {In each split one object instance from each class is chosen to be used in the test set and removed from the training set.} This split provides nearly 35,000 training images and 7,000 test images.}

Core50 \cite{lomonaco2017core50} is a recently introduced benchmark for testing continual learning methods in an egocentric setting. The dataset contains images of 50 objects grouped into 10 semantic categories. The images have been acquired on 11 different sequences with varying conditions. 
Following the standard protocol described in \cite{lomonaco2017core50}, we select the sequences 3, 7, 10 for the evaluation phase and use the remaining ones to train the model. {Due to these differences in conditions between the sequences, Core50 represents a very challenging benchmark for object recognition}. As in the RGB-D Object dataset, we split it into two parts: 5 classes are considered known and the other 5 as unknown. In the known set, the first 2 classes are considered as the initial training set. The others are incrementally added 1 class at a time. 

CIFAR-100 \cite{krizhevsky2009learning} is a standard benchmark for comparing
incremental class learning algorithms \cite{rebuffi2017icarl}. It contains 100 different semantic categories. We split the dataset into 50 known and 50 unknown classes and considering 20 classes as the initial training set. Then, we incrementally add the remaining ones in steps of 10 classes. We evaluate the performance of DeepNNO and \owr in the OWR scenario comparing it with NNO \cite{bendale2015towards}, using the simplified implementation in \cite{de2016online}. We further compare our methods with two standard incremental class learning algorithms, namely LwF \cite{li2017learning} (in the MC variant of \cite{rebuffi2017icarl}) and iCaRL \cite{rebuffi2017icarl}. Both LwF and iCaRL are designed for the closed world scenario, thus we use their performances as reference in that setting, without open-ended evaluation. 
For each dataset, we have randomly chosen five different sets of known and unknown classes. After fixing them, we run the experiments three times for each method. The results are obtained by averaging the results among each run and order.

\myparagraph{Networks architectures and training protocols.} We use a ResNet-18 architecture \cite{he2016deep} for all the experiments. For RGB-D Object dataset and Core50, we train the network from scratch on the initial classes for 12 epochs and for 4 epochs in the incremental steps. For CIFAR-100, instead, we set the epochs to 120 for the initial learning stage and to 40 for each incremental step. In the case of NNO we use the features extracted from the pre-trained network to compute the class-specific mean vectors of novel categories, but we do not update the weight matrix $W$ and the threshold parameter $\tau$, as in \cite{bendale2015towards}.
For DeepNNO we use an initial learning rate of $1.0$ in all settings, for \owr we use a learning rate of $0.1$ for the RGB-D Object dataset and CIFAR-100, and $0.01$ for Core50, with a batch size of 128 for RGB-D Object dataset and of 64 for CIFAR-100 and Core50. We train the networks using SGD with momentum $0.9$ and a weight decay of $10^{-3}$ on all datasets.  
{We resize the images of RGB-D Object dataset to $64 \times 64$ pixels, the ones of CIFAR-100 to 32x32 and the images of Core50 to $128 \times 128$ pixels.} We perform random cropping and mirroring for all the datasets. 
In all experiments, we set $\lambda=1$, $w^{+}=1$ and $w^{-}=3$ for DeepNNO, while $\lambda=\gamma=1$ for \owr. For both methods we consider a fixed size memory of 2000 samples, constructing each batch by drawing 40\% of the instances from the memory. Note that, in \owr 20\% of the samples present in the memory are never seen during training, but are used only to learn the class-specific threshold values $\Delta_c$. For this set of held-out samples, we also perform color jittering varying brightness, hue and saturation.


\myparagraph{Metrics}
We use 3 standard metrics for comparing the performances of OWR methods. For the closed world we show the global accuracy \textit{with} and \textit{without} rejection option. Specifically, in the closed world \textit{without rejection} setting, the model is tested only on the known set of classes, \textit{excluding} the possibility to classify a sample as \textit{unknown}. This scenario measures the ability of the model to correctly classify samples among the given set of classes. In the closed world \textit{with rejection} scenario, instead, the model can either classify a sample among the known set of classes or categorize it as \textit{unknown}. This scenario is more challenging than the previous one because samples belonging to the set of known classes might be misclassified as \textit{unknowns}. For the open world we use the standard OWR metric defined in \cite{bendale2015towards} as the average between the accuracy computed on the closed world \textit{with rejection} scenario and the accuracy computed on the open set scenario (i.e. the accuracy on rejecting samples of unknown classes). Since the latter metric creates biases on the final score (i.e. a method rejecting every sample will achieve a 50\% accuracy), we introduced the OWR-H as the harmonic mean between the accuracy on open set and the closed world with rejection scenarios to mitigate this bias.

\subsubsection{Quantitative results}
We report the results on the RGB-D Object dataset in Fig.~\ref{fig:ROD}. 
Considering 
the closed world without rejection, reported in Fig.~\ref{fig:ROD-wor}. This scenario is used to asses the ability of a method to learn novel classes while preserving old knowledge, without considering the open-set scenario. As a first observation, we note that both our deep methods outperform NNO by a large margin (i.e. 9.2\% DeepNNO and 14.8\% \owr in accuracy on average), showing the importance of end-to-end trained deep representations for OWR.  Remarkably, \owr outperforms DeepNNO by 5.6\% of accuracy on average. The reason for the improvement comes from the introduction of the global and local clustering loss terms, which allows the model to better aggregate samples of the same class and to better separate them from samples of other classes. Comparing our models with the incremental class learning approaches LwF and iCaRL, we can see that both of them are highly competitive, surpassing LwF with a large gap while being either comparable (\owr) or slightly inferior (DeepNNO) with the more effective iCaRL. We believe these are remarkable results given that the main goal of our models is not to purely extend their knowledge over time with new concepts. 

For what concerns the comparison on the closed world with rejection, shown in Fig.~\ref{fig:ROD-wir}, again DeepNNO and \owr surpass NNO in terms of performance. However, the results of \owr are remarkable, demonstrating how it is achieves higher confidence on the known classes, being able to reject a lower number of known samples. In particular, \owr is more confident on the first incremental steps, and obtains, on average, an accuracy of 10.3\% more than DeepNNO.

The findings are confirmed also on OWR metrics. Again, both DeepNNO and \owr surpass NNO, showing the importance of end-to-end trained representations and updated thresholds in achieving a higher performance, even in the presence of unknowns. Even on the OWR metrics, \owr surpasses DeepNNO. From the results of OWR, reported in Fig.~\ref{fig:ROD-owra}, we see that \owr reaches performance similar to DeepNNO in the first steps, while it outperforms it in the latest ones. However, considering the OWR-H (Fig.~\ref{fig:ROD-owrh}), \owr is better in all the incremental steps. This is because its learned rejection criterion, coupled with the clustering losses, allows \owr to achieve a better trade-off between the accuracy of open set and closed world with rejection. Overall, \owr improves on average by 4.8\% and 5.2\% with respect to DeepNNO in the OWR and OWR-H metrics respectively. We will provide a deeper analysis on the rejection criterion of DeepNN and \owr with ablation studies in the next subsections. 

In Fig.~\ref{fig:Core} we report the results on the Core50 \cite{lomonaco2017core50} dataset. Similarly to the RGB-D Object dataset, DeepNNO and \owr achieve very competitive results with respect to incremental class learning algorithms designed for the closed world scenario, with \owr remarkably outperforming iCaRL by 4.7\% of accuracy in the last incremental step. Similarly, \owr achieves a superior performance in both closed world, without and with rejection option with respect to the other OWR algorithms, outperforming NNO by 13.01\% and DeepNNO by 7.74\% on average in the first (Fig.~\ref{fig:Core-wor}) and by more than $10\%$ for both NNO and DeepNNO in the latter (Fig.~\ref{fig:Core-wir}).
In particular, it is worth noting how the challenges of Core50 (i.e. train and test acquisitions under different conditions) does not allow DeepNNO and NNO to properly model the confidence threshold, rejecting most of the sample of the known classes. Indeed, by including the rejection option the accuracy drops to 27.2\% and 26.3\% respectively for DeepNNO and NNO, while \owr reaches an average accuracy of 38.0\%. 

In Fig.~\ref{fig:Core-owra} and Fig.~\ref{fig:Core-owrh}, we report the OWR performances (standard and harmonic) on Core50. While DeepNNO surpasses the performance of NNO in both metrics (5.4\% in standard OWR and 3.1\% in OWR-H), \owr performs even better,  outperforming DeepNNO by 3.4\% and 7.2\% in average respectively in standard OWR and OWR-H metrics, confirming the effectiveness of the clustering losses and the learned class-specific maximal distances. 

Finally, in Fig. \ref{fig:cifar} we report the results on the CIFAR-100 dataset in terms of the OWR (Fig. \ref{fig:cifar-owra}) and OWR-H metrics (Fig. \ref{fig:cifar-owrh}). Even in this benchmark, confirms the finding of previous analysis: end-to-end trained methods with updated thresholds (DeepNNO and \owr) are more effective than shallow methods (NNO). Similarly to previous analyses, \owr outperforms, on average both DeepNNO and NNO, with lower performances only in the initial training stage. However, in the incremental learning steps \owr clearly outperforms both methods, demonstrating its ability to learning and recognizing in an open-world without forgetting old classes. The relative gaps are still remarkable. DeepNNO improves over NNO by 6.3\% and ~4\% in OWR and OWR-H metrics respectively. However, in the incremental steps, the average improvement of \owr over NNO are of ~10\% in both OWR and OWR-H metrics, while over DeepNNO are of 2\% for the OWR and 4.5\% for the OWR-H metric.

\begin{figure*}[t]
\minipage{0.48\textwidth}
   \includegraphics[width=1.0\textwidth]{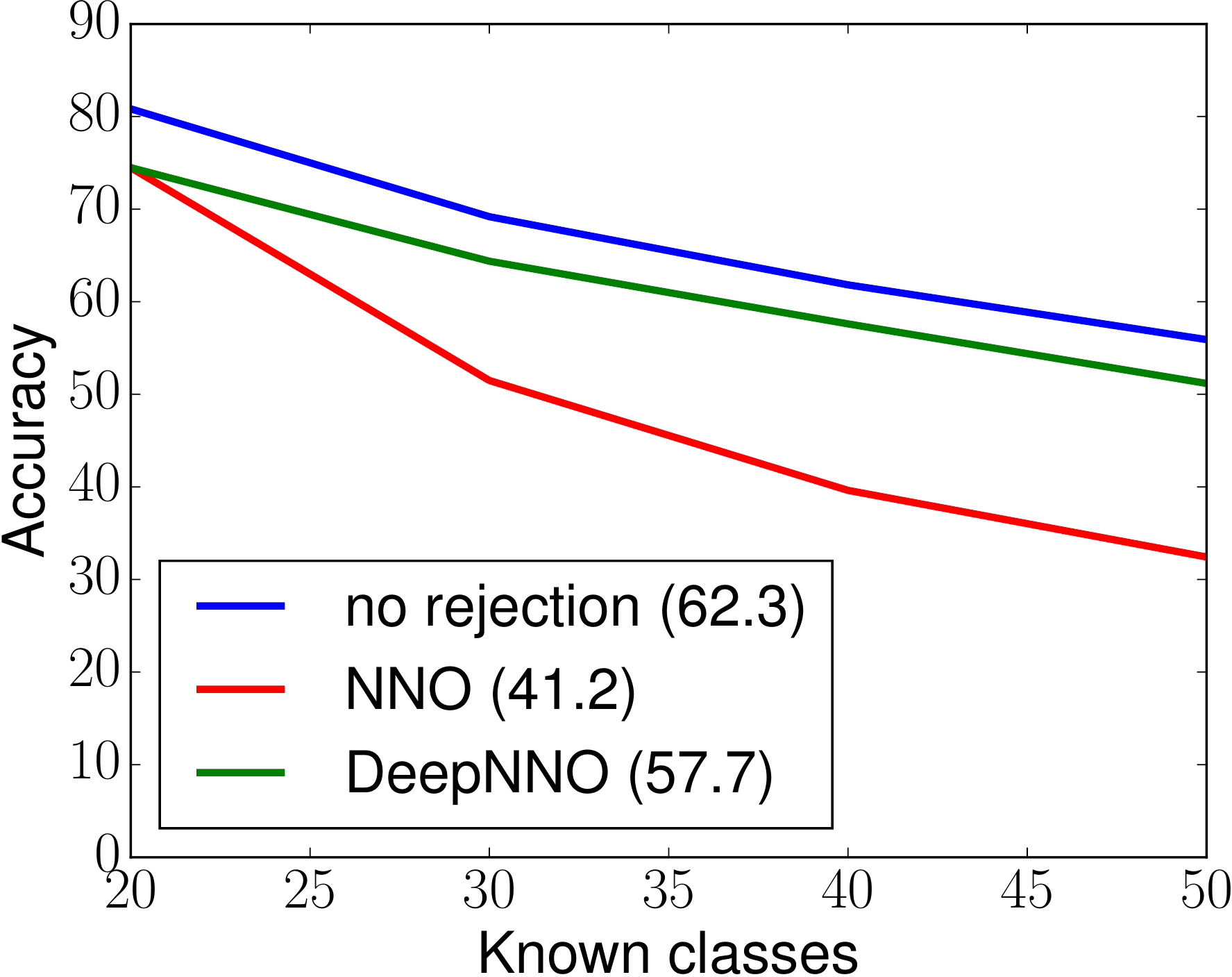}
   \caption{CIFAR-100 results in the closed world scenario.
 }
  \label{fig:res-cifar100-cw}
\endminipage\hfill
\minipage{0.48\textwidth}%
    \includegraphics[width=\textwidth]{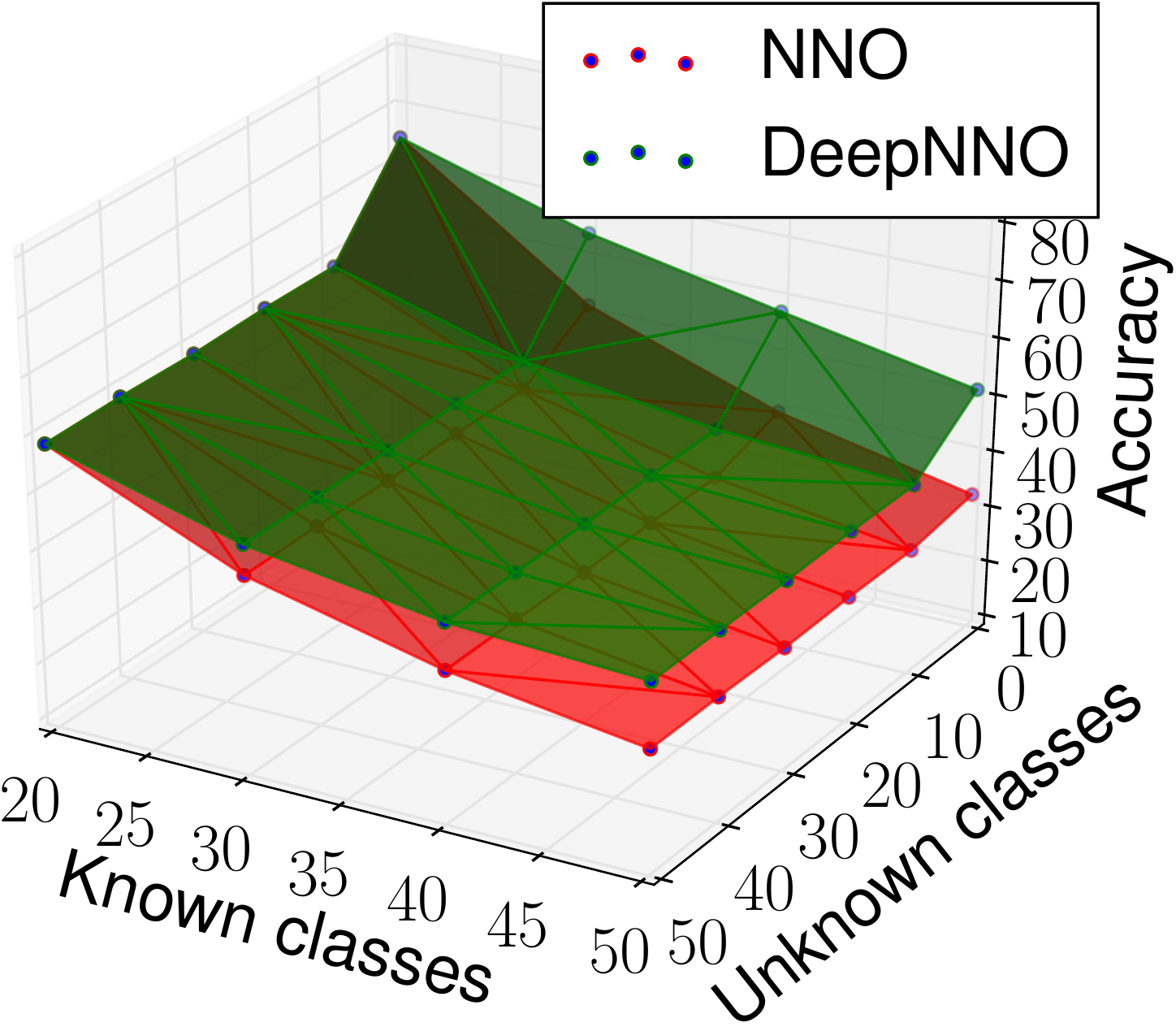}
   \caption{CIFAR-100: open world performances varying the number of known and unknown classes.}
  \label{fig:res-cifar100-ow3d}
  \endminipage
\end{figure*}

\begin{figure*}[t]
\minipage{0.48\textwidth}
   \includegraphics[width=\textwidth]{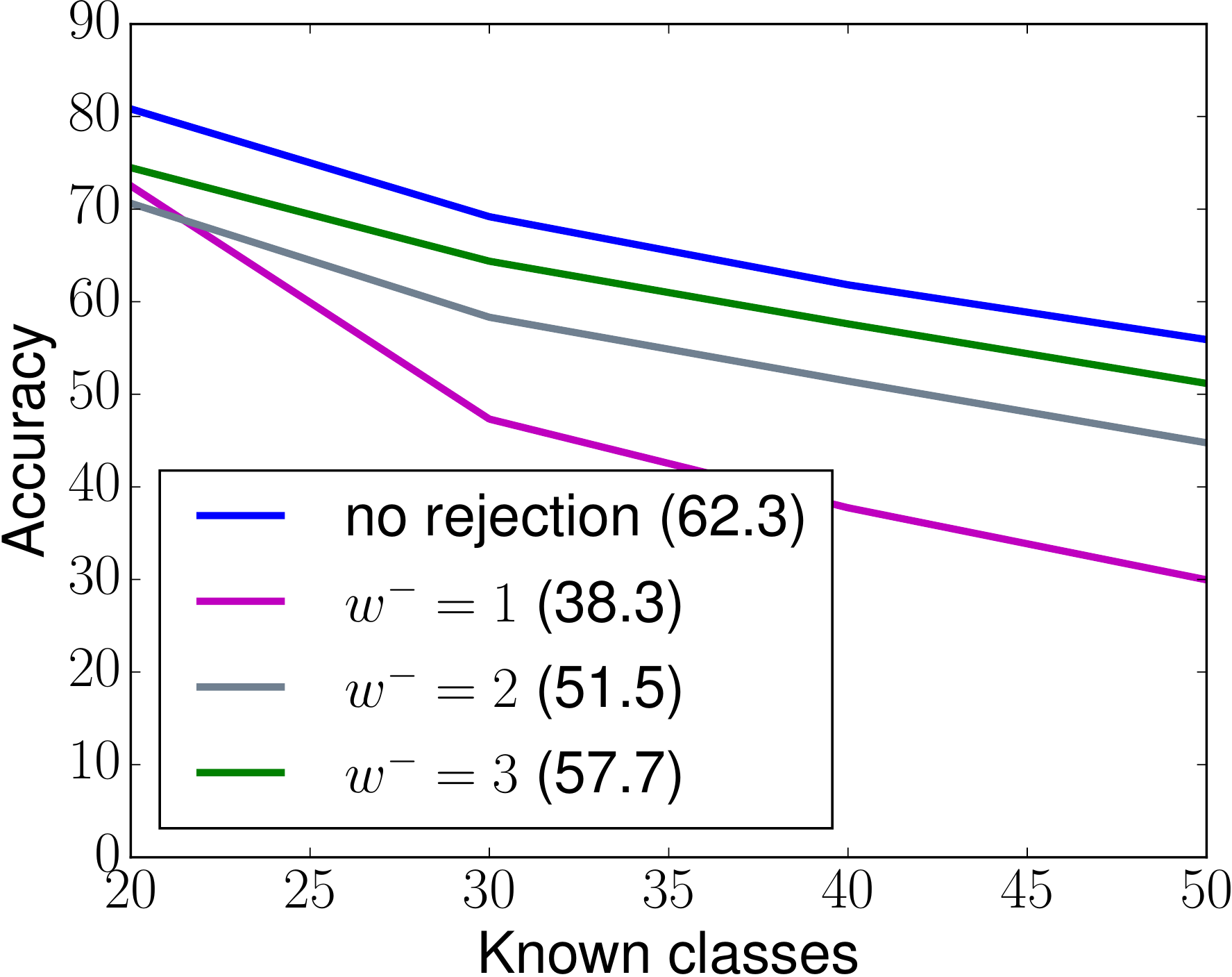}
   \caption{CIFAR-100 results of DeepNNO in the closed world scenario for different values of $w^-$.
  }
  \label{fig:ablation-margin}
\endminipage\hfill
\centering
\minipage{0.48\textwidth}
\centering
\includegraphics[width=\textwidth]{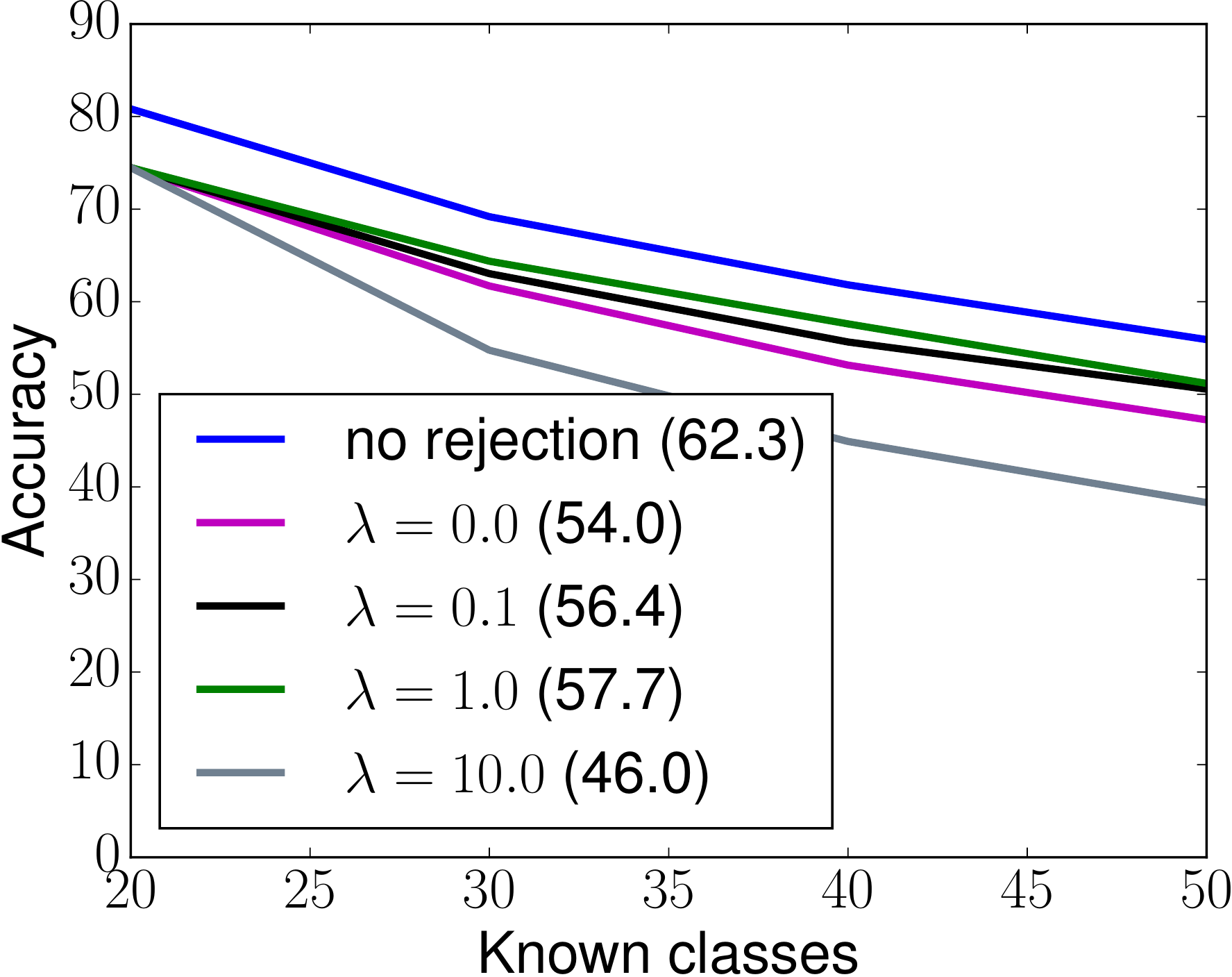}
   \caption{CIFAR-100 results of DeepNNO in the closed world scenario for different values of $\lambda$.
   }
   \label{fig:ablation-distill}
\endminipage
\end{figure*}

\subsubsection{Ablation study of DeepNNO}
DeepNNO improves over NNO by introducing two main aspects: end-to-end trained deep representations with an updated rejection threshold, and a distillation loss to preserve old knowledge. In the following we analyze in detail the reasons behind the improvement of DeepNNO with respect to NNO on the CIFAR-100 dataset, focusing first on the importance of learning deep representations with updated thresholds and then on the impact of the distillation loss.

\myparagraph{Deep representation and updated threshold.} We start by performing experiments in the closed world scenario, \ie measuring the performances considering only the set of known classes.  
In particualr, we compare the performance of DeepNNO with NNO and DeepNNO without rejection option (\ie \textit{DeepNNO-no rejection}). The latter baseline method is the upper bound of DeepNNO in terms of performances in the closed world, since it does not reject any instance of known classes (\ie it does not identify samples of known classes as unknowns). This baseline is used to demonstrate the validity of the method in Eq.~\eqref{eq:deep-nno-tau-update} for setting the threshold $\Delta$. The results are shown in Fig. \ref{fig:res-cifar100-cw} where the numbers between parenthesis denote the average accuracy among the different incremental steps. From Fig. \ref{fig:res-cifar100-cw} it is possible to draw two observations. First, there is a large gap between the performances of DeepNNO and NNO, with our model outperforming its non-deep counterpart by more than $16\%$ on average and by more than $20\%$ after all the incremental steps. 

The improved performance of DeepNNO can be ascribed to the fact that, by dynamically updating the learned feature representations, DeepNNO is able to better adapt the learned classifier to novel semantic concepts. Second, DeepNNO achieves results close to DeepNNO without rejection. This indicates that, thanks to the proposed approach for setting the threshold $\Delta$, DeepNNO only rarely identifies samples of known classes as belonging to an unknown category. We believe this is mainly due to the introduction of the different weighting factors $w^-$ and $w^+$ while updating $\Delta$. This observation is confirmed by results shown in Fig.~\ref{fig:ablation-margin} 
 which analyzes the effect of varying $w^-$ with $w^+$ fixed to 1. As $w^-$ decreases, the accuracy decreases as well, due to the higher value reached by $\Delta$ which leads to wrongly reject many samples, classified as instances of unknown classes. We want to highlight however that for more complex and realistic scenario, the threshold obtained by DeepNNO does not generalize as well and more the more principled strategy of \owr results more effective, as we will show in the next subsection. 
 
 As a second experiment, we compare the performances of DeepNNO and NNO in the open world recognition scenario varying the number of known and unknown classes. The results are shown in Fig. \ref{fig:res-cifar100-ow3d}, from which it is easy to see that DeepNNO outperforms its non-deep counterpart by a large margin. 
In fact, in this scenario, our model achieves a standard OWR accuracy 9\% higher than standard NNO on average, considering 50 unknown classes. Moreover, this margin increases during the training: after all the incremental steps our model outperforms NNO by a margin close to 15\%.
It is worth noting that the advantages of our model are independent on the number of unknown classes, since DeepNNO constantly outperforms NNO in all settings.
 
\myparagraph{Distillation loss.} Another important component of DeepNNO is the distillation loss. This loss guarantees the right balance between learning novel concepts and preserving old features.
To analyze its impact, in Fig.~\ref{fig:ablation-distill} we report the performances of DeepNNO in the closed world scenario for different values of $\lambda$. From the figure it is clear that, without the regularization effect of the distillation loss, the accuracy significantly drops.
On the other hand, a high value of $\lambda$ leads to poor performances and low confidence on the novel categories. Properly balancing the contribution of classification and distillation loss the best performance can be achieved. The use of the distillation loss is thus crucial for limiting the catastrophic forgetting, as previously verified in \cite{li2017learning,rebuffi2017icarl}.

\subsubsection{Ablation study of \owr} \owr is mainly built on three components, i.e. global clustering loss (GC), local clustering loss (LC) and the learned class-specific {rejection thresholds}.
In the following we analyze the contribution of each of them. We start from the two clustering losses and then we compare the choice we made for the rejection with other common choices.

\begin{table}[t]
\centering
\begin{tabular}{l||llll||l|l}
\multicolumn{1}{c||}{\textbf{Method}} & \multicolumn{4}{c||}{\textbf{Known Classes}} & \multicolumn{2}{c}{\textbf{OWR}} \\
 & \multicolumn{1}{c}{11} & \multicolumn{1}{c}{16} & \multicolumn{1}{c}{21} & \multicolumn{1}{c||}{26} & \multicolumn{1}{c}{{[}20{]}} & \multicolumn{1}{c}{H} \\\hline 
\multicolumn{1}{l||}{GC} & \multicolumn{1}{l|}{66.0} & \multicolumn{1}{l}{57.3} & \multicolumn{1}{l|}{58.6} & 53.3 & 58.8 & \multicolumn{1}{l}{58.7} \\ 
\multicolumn{1}{l||}{LC} & \multicolumn{1}{l|}{64.1} & \multicolumn{1}{l}{56.0} & \multicolumn{1}{l|}{57.9} & 56.4 & 58.6 & \multicolumn{1}{l}{58.4} \\\multicolumn{1}{l||}{Triplet} & \multicolumn{1}{l|}{62.1} & \multicolumn{1}{l}{54.9} & \multicolumn{1}{l|}{54.8} & 49.5 & 55.4 & \multicolumn{1}{l}{55.4} \\  
\multicolumn{1}{l||}{GC + LC} & \multicolumn{1}{l|}{\textbf{67.7}} & \multicolumn{1}{l}{\textbf{59.6}} & \multicolumn{1}{l|}{\textbf{59.5}} & \textbf{57.3} & \textbf{61.0} & \multicolumn{1}{l}{\textbf{60.8}} \\ 
\end{tabular}
\caption{Ablation study of \owr on the global (GC), local clustering (LC) and {Triplet} loss on the OWR metric. The right column shows the average OWR-H over all steps. 
}
\label{tab:ablation_clustering}
\end{table}

\begin{table}[t]
\centering
\begin{tabular}{l|cc|ll|l}
\multirow{2}{*}{\textbf{Method}} &\textbf{Class} &\textbf{Multi} & \multirow{2}{*}{\textbf{Known}} & \multirow{2}{*}{\textbf{Unknown}} & \multirow{2}{*}{\textbf{Diff.}} \\ 
&\textbf{specific} & \textbf{stage}& & & \\\hline
 DeepNNO& & & 84.4 & 98.8 & 14.4 \\\hline
\multirow{3}{*}{B-DOC} & \cmark & & 83.0 & 98.6 & 15.6 \\
& &\cmark & 4.4 & 26.9 & 22.6 \\
 & \cmark& \cmark& 27.4 & 65.2 & \textbf{37.8} \\  \hline
\end{tabular}
\caption{Rejection rates of different techniques for detecting the unknowns. The results are computed using the same feature extractor on the RGB-D Object dataset.
}
\label{tab:ablation_unknown}
\end{table}

\myparagraph{Global and local clustering.} In Table \ref{tab:ablation_clustering} we compare the two clustering terms considering the open world recognition metrics in the RGB-D Object dataset. By analyzing the two loss terms separately we see that, on average, they show similar performance. In particular, using only the global clustering (GC) term we achieve slightly better performance on the first three incremental steps, while on the fourth the local clustering (LC) term is better.
However, the best performance on every step is achieved by combining the global and local clustering terms (GC + LC). This demonstrates that the two losses provide different contributions, being complementary to learn a representation space which properly clusters samples of the same classes while better detecting unknowns.

Lastly, since the \owr loss functions and triplet loss \cite{balntas2016learning} share the same objective, i.e. building a metric space where samples sharing the same semantic are closer then ones with different semantics, we report in Table \ref{tab:ablation_clustering} also the results achieved by replacing our loss terms with a triplet loss \cite{balntas2016learning}. 
As the Table shows, the triplet loss formulation (Triplet) fails in reaching competitive results with respect to our full objective function in Eq.~\eqref{eq:owr-bdoc-loss}, with a gap of more than 5\% in both standard OWR metric and OWR harmonic mean. Notably, it achieves lower results also with respect to all of the loss terms in isolation and the superior performances of LC confirm the advantages of SNNL-based loss functions with respect to triplets, as shown in \cite{frosst2019analyzing}.

\myparagraph{Detecting the Unknowns.} In Table \ref{tab:ablation_unknown} we report a comparison of different strategies to reject samples on the RGB-D Object dataset \cite{lai2011large}. In particular, using the same feature extractor, we compare the proposed method to learn the class-specific maximal distances (i.e. Eq.~\eqref{eq:owr-bdoc-rejection-loss}) with three baselines: (i) the online update strategy of DeepNNO (Eq.~\eqref{eq:deep-nno-tau-update}), (ii) we learn class-specific maximal distances but during training (i.e. without our two-stage pipeline) and (iii) we learn a single maximal distance which applies to all classes using our two-stage training strategy. 

The comparison is performed considering the difference of the rejection rates on the known and unknown samples. For the known class samples, we report the percentage of correctly classified samples in the closed-world that are rejected when the rejection option is included.
We intentionally remove the wrongly classified samples since we want to isolate rejection mistakes from classification ones. On the unknown samples, we report the open-set accuracy, i.e. the percentage of rejected samples among all the unknown ones. In the third column, we report the difference among the open-set accuracy and the rejection rate on known samples. Ideally, the difference should be as close as possible to 100\%, since we want a 100\% rejection rate on unknown class samples and 0\% on the known class ones.

From the table, we can see that the highest gap is achieved by the class-specific maximal distance with the two-stage pipeline we proposed, which rejects 27.4\% of known class samples and 65.2\% on the unknown ones. The gap with the other strategies is remarkable. Using the {two stage-pipeline but a class-generic maximal distance} leads to a low rejection rate, both on known and unknown samples, achieving a difference of 22.6\%, which is 15.2\% less than using a class-specific distance. On the other hand, estimating the confidence threshold as proposed in DeepNNO or without our two-stage pipeline provides a very high rejection rate, both on known and unknown classes, which lead to a difference of 14.4\% and 15.6\% for DeepNNO and the single-stage strategy respectively,  the lowest two among the four strategies.  
In fact, computing the thresholds using only the training set biases the rejection criterion on the overconfidence that the method has acquired on this set. Consequently, this makes the model considering the different test data distribution (caused by e.g. different object instances) as a source for rejection even if the actual concept present in the input is known. Using the two-stage process we can overcome this bias, tuning the rejection criterion on unseen data on which the model cannot be overconfident.

\subsection{Towards Autonomous Visual Systems: Web-aided OWR}
\label{sec:OWR-web}

\begin{figure}[t]
  \centering
  \includegraphics[width=\textwidth]{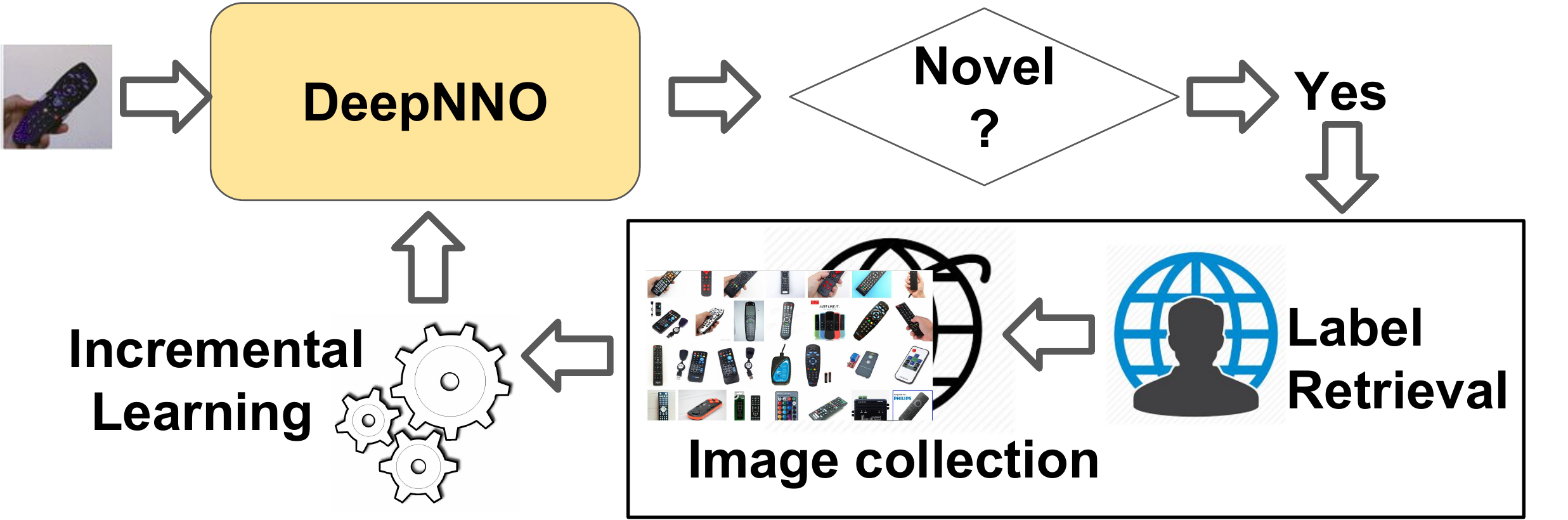}
 \caption{Overview of the open world recognition task within a robotic platform. Given an image of an object, a classification algorithm assigns to it a class label. If the object is recognized as novel, the object label and relative are obtained through external resource (\eg a human and/or the Web). Finally, the images are used to incrementally updated the knowledge base of the robot.
 }
   \label{fig:owr-framework}
\end{figure}

 The OWR frameworks considered so far assumes the existence of an 'oracle', providing annotated images for each new class. In a robotic scenario, this has been often translated into having a human in the loop, with the robot asking for images and labels. This scenario somehow limits the autonomy of a robot system, that without the presence of a teacher, would find itself stuck when detecting a new object. 
 Moreover, especially in robotics applications, this assumption is highly unrealistic since: i) the labels of samples of unknown categories are, by definition, unknown; 
ii) images of the unknown classes for incrementally updating the model are usually unavailable, since it is impossible to have a pre-loaded database containing all possible classes existing in the real world. In the last part of this section we want to describe a simple general pipeline to address the aforementioned issues with first pilot experiments showing its possible application.

We start by considering the problem of retrieving the correct label of an unknown object. To this extent, we exploit standard search tools used by humans. First, once an object is recognized as unknown, we query the Google Image Search engine\footnote{https://images.google.com/} to retrieve the closest keyword to the current image. Obviously the retrieved label might not be correct \eg due to low resolution of the image or a non canonical pose of the object. We tackle this issue through an additional human verification step, 
leaving the investigation of this problem to future works. As subsequent step, we use the retrieved keyword to automatically download images from the web. These weakly-annotated and noisy images represent new training data for the novel category which can be used to incrementally train the deep network. Fig.~\ref{fig:owr-framework} shows an overview of our pipeline. Interestingly, this simple framework mimics the human ability to learn not only from situated experiences, but also from visual knowledge externalized on artifacts (e.g. like drawings), or indeed Web resources.

\begin{figure*}[t]
\centering
\minipage{0.48\textwidth}
\centering
    \includegraphics[width=0.98\textwidth]{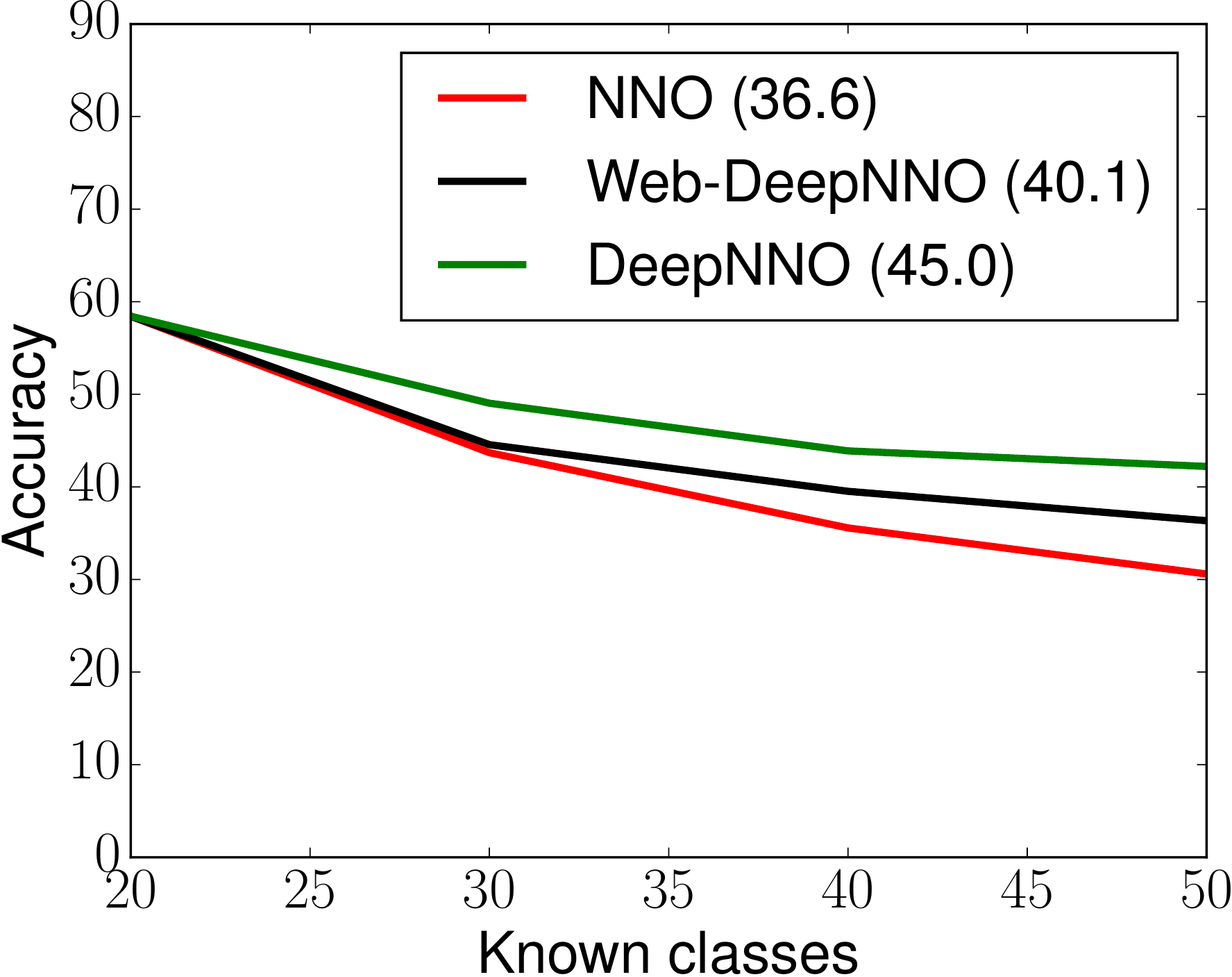}
      \caption{CIFAR-100: performances of Web-aided OWR in the open world scenario, with 50 unknown classes.}
   \label{fig:web-cifar100}
\endminipage\hfill
\minipage{0.48\textwidth}%
    \includegraphics[width=\textwidth]{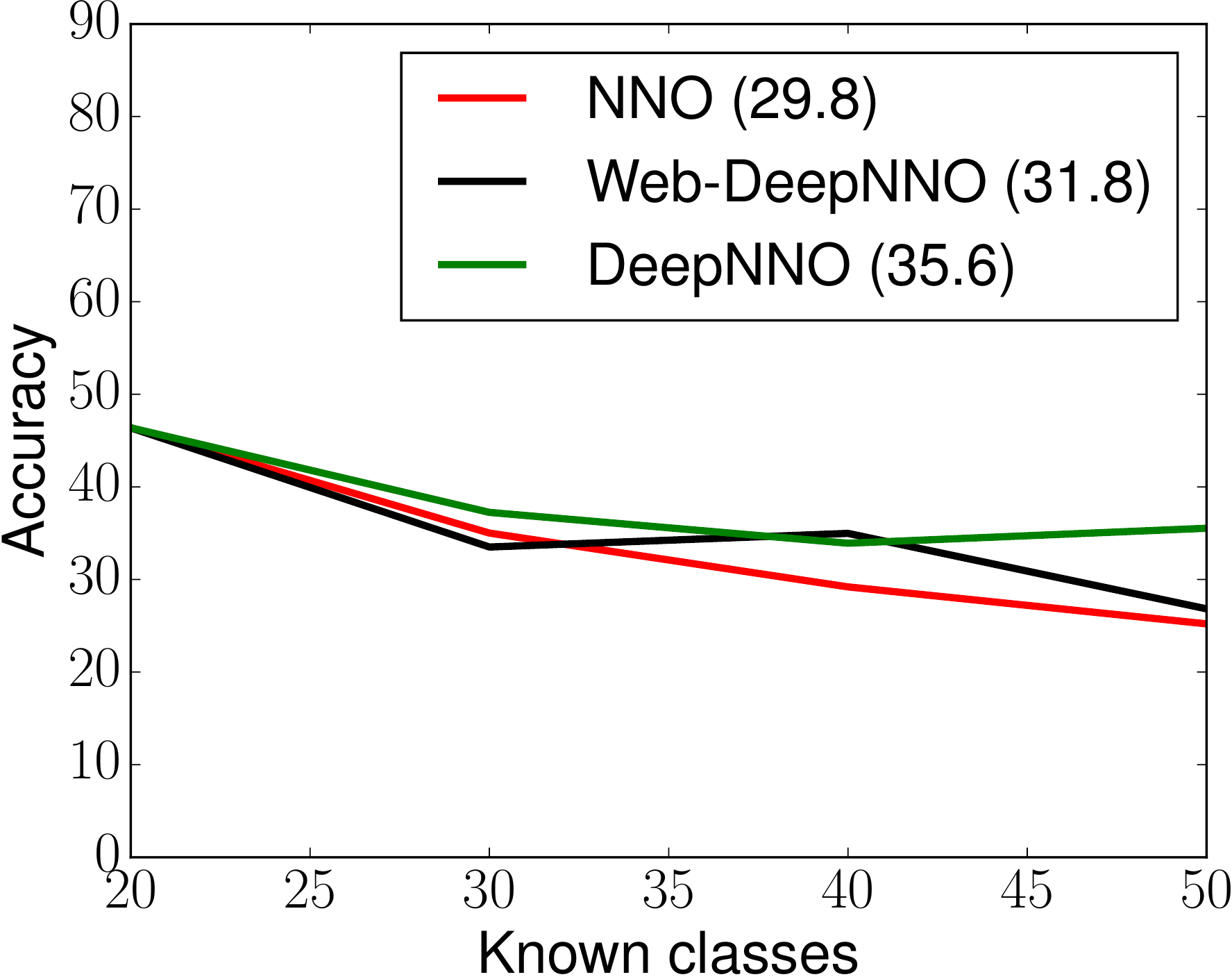}
      \caption{Core50 dataset: performances of Web-aided OWR in the open world scenario, with 5 unknown classes.} 
   \label{fig:web-core50}
\endminipage
\end{figure*}

We conduct a first series of preliminary experiments, using web-images in the incremental learning steps of DeepNNO, to validate the feasibility of this pipeline. The results of our experiments are shown in Fig.~\ref{fig:web-cifar100} for CIFAR-100 and in Fig.~\ref{fig:web-core50} for Core50. As expected, considering images from the Web instead of images from the datasets lead to a decrease in terms of performance. However, the accuracy of the Web-based DeepNNO is still good, especially when compared with its non-deep counterpart. 

On the CIFAR-100 experiments we achieve a remarkable performance, with Web DeepNNO outperforming NNO by 3.5\% on average and by more than 5\% after all the incremental steps, with respect to the standard OWR metric. We highlight that these results have been achieved exploiting only noisy and weakly labeled Web images, without any filtering procedure or additional optimization constraints. 
On the Core50 experiments, the gap Between DeepNNO and NNO is lower, as shown in Fig, \ref{fig:Core-owra} and \ref{fig:Core-owrh} and this impacts also the results of the  Web-based version of DeepNNO, achieving a modest improvement with respect to NNO. We ascribe this behavior to the fact that there is a large appearance gap between Core50 images gathered in an egocentric setting and Web images, thus both the rejection threshold and the semantic centroids of new classes are not able to well model the underline data distribution, with deteriorated final results. We believe that this issue can be addressed in future works by e.g. imposing some constraints on the quality of downloaded images and by coupling DeepNNO with domain adaptation techniques~\cite{patel2015visual,carlucci2017autodial,mancini2018kitting,mancini2018boosting} in order to reduce the domain shift between downloaded images and training data.

\begin{figure}[t]
    \centering
    \includegraphics[width=1.\columnwidth]{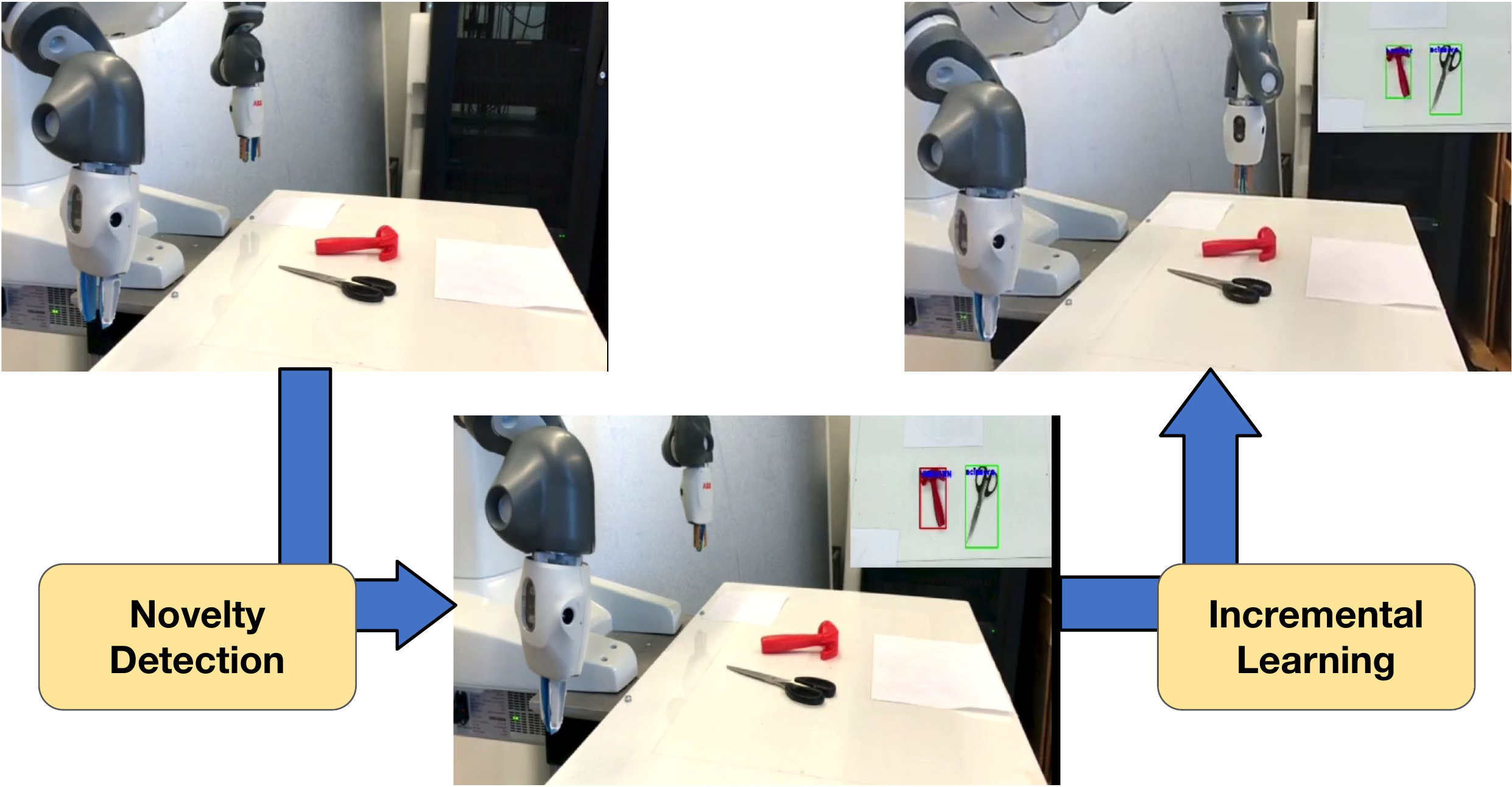}
    \caption{{Qualitative results of deployment of DeepNNO on a robotic platform. The robot recognizes an object as unknown (\ie the red hammer, bottom) and adds it to the knowledge base through the incremental learning procedure (top right).}}
    \label{fig:owr-robot-task}
  \end{figure}

To validate the applicability of the pipeline in a real scenario, we tested the Web-aided version of DeepNNO by integrating it into a visual object detection framework and running it on a Yumi 2-arm manipulator equipped with a Kinect. We have used the Faster-RCNN framework in~\cite{ren2015faster} with the ResNet-101 architecture~\cite{he2016identity} as backbone. We pre-trained the network on the COCO dataset \cite{lin2014microsoft}, 
after replacing the standard fully-connected classifier with the proposed DeepNNO. We performed an open world detection experiment by placing multiple objects (known and unknowns) in the workspace of the robot. Whenever a novel object is detected, the robot tries to get the corresponding label from Google Image Search, using the cropped image of the unknown object. In case the label is not correct, a human operator cooperates with the robot and provides the right label. The provided label is used by the robot to automatically download the images associated to the novel class from the Web sources. These images and the original one where the object has been detected in the workspace, are then used to update the classification model.

{Figure~\ref{fig:owr-robot-task} shows a qualitative result associated to our experiment. The robot was able to correctly detect the red hammer as unknown, add it in its knowledge base and recognize it in subsequent learning steps. \footnote{A full example is available in the supplementary material of \cite{mancini2019knowledge}.}} Despite the simplicity of the workspace, we want to highlight that the robot was able to recognize the hammer without any explicitly labeled training data for the class of interest.

We want to point out that here we are not claiming that our framework is incorporating new knowledge into a visual robotic system in a completely autonomous and fully effective way. Indeed (i) the human verification step on the retrieved keyword is necessary and (ii) web supervision \cite{divvala2014webly,chen2015webly} requires to address challenges such as noisy labels \cite{niu2018noisy} and domain shift \cite{xu2016weblyda}, which we did not take into account. Nevertheless, we still believe our experiments show how our pipeline is a feasible starting point which is worth exploring in future research directions toward autonomous learners in the real world. 

\subsection{Conclusions}
\label{sec:OWR-concl}
In this section, we presented two approaches to tackle the open world recognition problem in robot vision. We base our approaches on an NCM classifier built on top of end-to-end trainable deep features (DeepNNO), and we further boost the OWR performances of this framework by training the deep architecture to minimize a global to local semantic clustering loss (\owr) which allows reducing distances of samples of the same class in the feature space while separating them from points belonging to other classes, better detecting unknown concepts. In \owr we also avoid heuristic estimates of a rejection criterion for detecting unknowns by explicitly learning class-specific distances beyond which a sample is rejected. Quantitative and qualitative analysis on standard recognition benchmarks shows the efficacy of the proposed approaches and choices, outperforming the previous state-of-the-art OWR algorithm. Finally, we also showed preliminary experiments with a simple pipeline for allowing the robot to autonomously learn new semantic concepts, without the aid of an oracle providing it with a training set containing the desired target classes. 
 
  Future works will further investigate webly supervised approaches with the goal of pushing the envelope in life-long learning of autonomous systems. In particular, when training images are autonomously retrieved from the Web, they come with inherent noisy labeling (e.g. wrong semantic) and domain shift (e.g. white backgrounds). Attacking all these problems would allow active visual systems to get closer to full autonomy. In an intermediate direction, it would be interesting to analyze the OWR problem in an active learning context \cite{parisi2019rethinking}, letting the robot decide when to ask for human help for either collecting data or label new concepts.
  
  This section concludes our line of works on incrementally injecting new knowledge in a pre-trained deep model under various scenarios, with (ICL) or without (multi-domain learning) shared output spaces with old knowledge, and with (ICL, multi-domain learning) or without (OWR) closed-world assumption. Additionally, we identified problems (e.g. semantic shift of the background class) and posed challenges (web-aided OWR) not tackled in the community. Nevertheless, differently from Chapter \ref{chap:da}, here we consider the training and test distributions to belong to be equal, without any domain shift problem. On the other hand, differently from the techniques presented in Chapter \ref{chap:da}, this chapter described techniques that allow modifying the output space of a pre-trained architecture. In the next chapter, we will merge these two worlds together, describing the first method capable of recognizing unseen semantic concepts in unseen visual domains.

\chapter{Towards Recognizing Unseen Categories in Unseen Domains}
\label{chap:both}
\textit{While in the previous chapters we considered methods extending a pretrained model either to new input distributions or to new semantic concepts, an open research question is whether we can address the two problems together, producing a deep model able to recognize new semantic concepts (i.e. addressing the semantic shift) in possibly unseen domains (i.e. addressing the domain shift). In this chapter, we start analyzing how we can merge these two worlds, providing a first attempt in this direction in an offline but quite extreme setting. In particular, we considered a scenario where, during training, we are given a set of images of multiple domains
and semantic categories and our goal is to build a model able to recognize images of unseen concepts, as in zero-shot learning (ZSL), in unseen domains, as in domain-generalization (DG). This novel problem, which we called ZSL under DG (ZSL+DG), poses novel research questions going beyond the ones posed by the DG and ZSL problems if taken in isolation. For instance, similarly to DG, we can rely on the fact that the multiple source domains permit to disentangle semantic and domain-specific information. However, differently from DG, we have no guarantee that the disentanglement will hold for the unseen semantic
categories at test time. Moreover, while in ZSL it is reasonable to assume that the learned mapping between images and semantic attributes will generalize also to test images of the unseen concepts, in ZSL+DG we have no guarantee that this will happen for images of unseen domains. In Section \ref{both:problem} we provide a formal definition of the problem, while in Sec. \ref{both:related} {we review the related works in the zero-shot learning literature} and domain generalization. In Section \ref{both:dgzsl} we provide a first solution to this problem by designing a curriculum strategy based on the \textit{mixup} \cite{zhang2017mixup} algorithm. In particular, we use \textit{mixup} both at the input and feature level to simulate the domain shift and semantic shift the network will encounter at test time. Experiments show how this approach is effective in both ZSL, DG, and the two tasks together, producing one of the first attempts for recognizing unseen categories in unseen domains.  }

\section{Problem statement}
\label{both:problem}
\myparagraph{Overview.}  
As highlighted in Chapter \ref{chap:intro}, most existing deep visual models are based on the assumptions that (a) training and test data come from the same underlying distribution, i.e. domain shift, and (b) the set of classes seen during training constitute the only classes that will be seen at test time, i.e. semantic shift.  
These assumptions rarely hold in practice and, in addition to depicting different semantic categories, training and test images may differ significantly in terms of visual appearance in the real world. 

Up to now, we have presented approaches that tackle these problems in isolation. In particular, in Chapter \ref{chap:da}, we have considered the case where training and test distribution changes, addressing the domain shift problem, starting from the assumption of having target data available (Section \ref{sec:da-latent}), and removing it in the more complex domain generalization (Section \ref{sec:da-dg}), continuous (Section \ref{sec:da-continuous}) and predictive domain adaptation (Section \ref{sec:da-predictive}). However, in all the works we assumed the output space to be constant after the initial training stage and shared between training and test times.


On the other hand, in Chapter \ref{chap:icl}, we considered the case where the semantic space of a model is extended over time, as new training data arrives, but without the presence of the domain shift problem. In fact, while in Multi-Domain Learning (Section \ref{sec:IL-multitask}), a single model is asked to tackle different classification tasks in different visual domains, we have full supervision in each of the domains, and no unseen domain is received at test time. Similarly, in Incremental Learning (Section \ref{sec:IL-semantic-seg}) and Open World Recognition (Section \ref{sec:IL-owr}), we consider a single data distribution during all training steps and little to no shift at test time.  

In this chapter we focus on a different problem, considering the two shifts occurring jointly at test time. In particular, our goal is recognizing new semantic categories in new domains, without any of the categories and domains being present in our initial training set. 
In terms of the domain shift, we will consider the problem from a DG perspective (i.e. data of the target domain are not present during training while multiple sources are available). For the semantic shift,  we will consider the problem as Zero-Shot Learning (ZSL) \cite{xian2018zeroshotgood}. In ZSL, the goal is to recognize objects unseen during training given \textit{no data} but external information about the novel classes provided in forms of semantic attributes \cite{lampert2013awa}, visual descriptions \cite{akata2015evaluation} or word embeddings \cite{mikolov2013efficient}. We consider this problem because allows us to decouple semantic and domain shift, without considering other problems (e.g. catastrophic forgetting, see Section \ref{sec:IL-relateds}). Moreover, we will start by considering a ZSL scenario (i.e. at test time we want to recognize only unseen classes) and not the generalized ZSL one \cite{xian2018zeroshotgood} (where both seen and unseen categories must be recognized) because this allows us to sidestep the inherent bias our model would have on seen classes, focusing solely on the domain and semantic shifts.

\begin{figure}[tb]
  \centering
  \includegraphics[width=0.98\columnwidth,trim=0 5pt 0 0,clip]{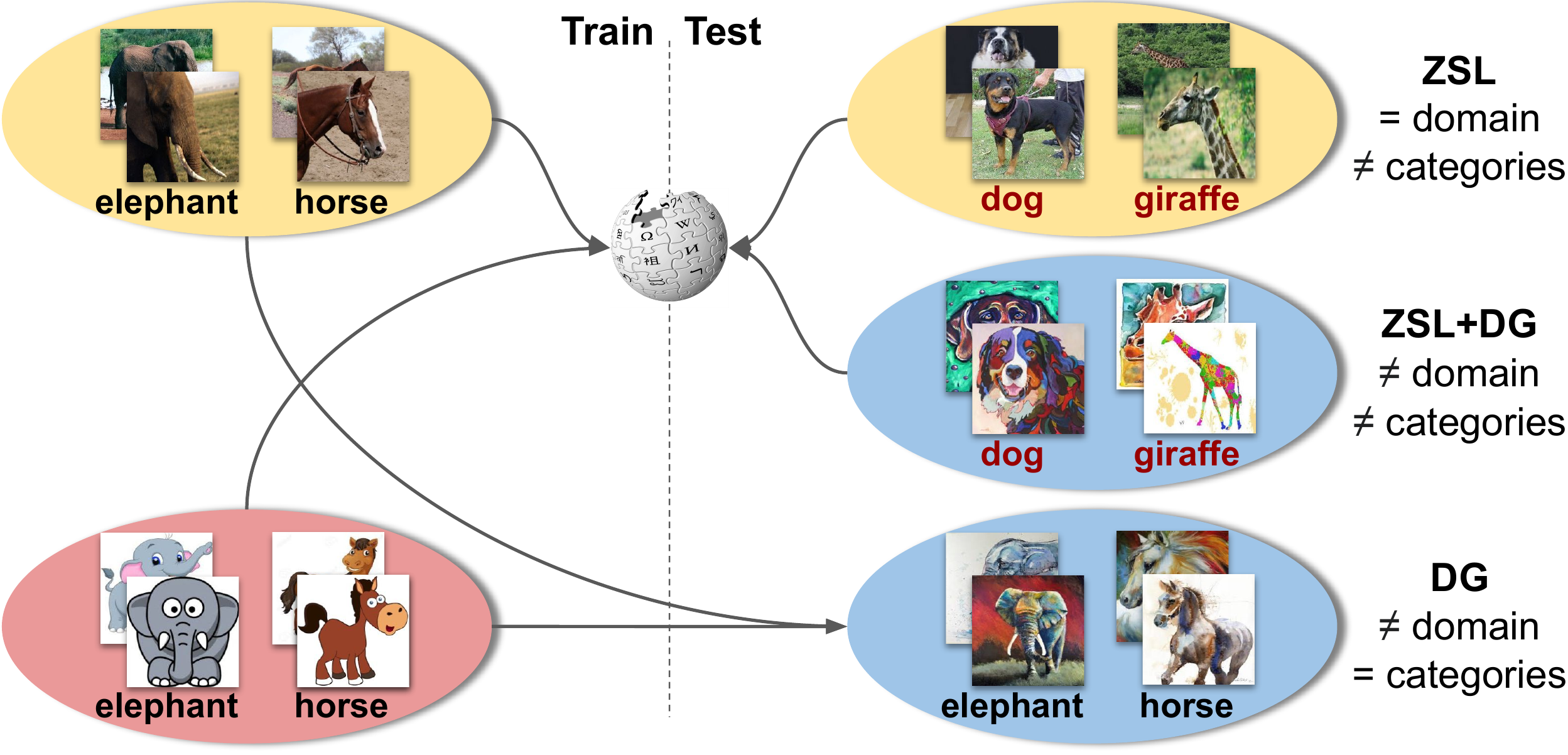}
   \caption{Our ZSL+DG problem. During training we have images of multiple categories (\eg \textit{elephant},\textit{horse}) and domains (\eg \textit{photo}, \textit{cartoon}). At test time, we want to recognize unseen categories (\eg \textit{dog}, \textit{giraffe}), as in ZSL, in unseen domains (\eg \textit{paintings}), as in DG, exploiting side information describing seen and unseen categories. }
  \label{fig:DGZSL-teaser}
 \end{figure}

To clarify the setting, let us consider the case depicted in Fig.~\ref{fig:DGZSL-teaser}. A system trained to recognize elephants and horses from realistic images and cartoons might be able to recognize the same categories in another visual domain, like art paintings (Fig.~\ref{fig:DGZSL-teaser}, bottom) or it might be able to describe other quadrupeds in the same training visual domains (Fig.~\ref{fig:DGZSL-teaser}, top). On the other hand, how to deal with the case where new animals are shown in a new visual domain is not clear. 
We want to remark that, while the one of Fig. \ref{fig:DGZSL-teaser} is a toy example, the need for a holistic approach jointly recognizing unseen categories in unseen domains comes from the large variability of the real world itself. Since it is impossible to construct a training set containing such variability, we cannot train a model to be robust to all the possible environments and semantic inputs it might encounter. Addressing these two problems together, allows our models to be more robust to these variabilities. Applications, where we need such robustness, are countless. For example, given a robot manipulation task we cannot forecast a priori all the possible conditions (e.g. environments, lighting) it will be employed in. Moreover, we might have data only for a subset of objects we want to recognize while only descriptions for the others.

To our knowledge, our work \cite{mancini2020dgzsl} is the first attempt to answer this question, proposing a method that is able to \textit{recognize unseen semantic categories in unseen domains}. In particular, our goal is to jointly tackle ZSL and DG (see Fig.\ref{fig:DGZSL-teaser}). ZSL algorithms usually receive as input a set of images with their associated semantic descriptions, and learn the relationship between an image and its semantic attributes. Likewise, DG approaches are trained on multiple source domains and at test time are asked to classify images, assigning labels within the same set of source categories but in an unseen target domain. 
Here we want to address the scenario where, during training, 
\emph{we are given a set of images of multiple domains and semantic categories and our goal is to build a model able to recognize images of unseen concepts, as in ZSL, in unseen domains, as in DG.} 

To achieve this, we need to address challenges usually not present  
when these two classical tasks, i.e. ZSL and DG, are considered in isolation. For instance, while in DG we can rely on the fact that the multiple source domains permit to disentangle semantic and domain-specific information, in ZSL+DG we have no guarantee that the disentanglement will hold for the unseen semantic categories at test time. Moreover, while in ZSL it is reasonable to assume that the learned mapping between images and semantic attributes will generalize also to test images of the unseen concepts, 
in ZSL+DG we have no guarantee that this will happen for images of unseen domains. 

{To overcome these issues, during training we 
simulate
both the semantic and the domain shift we will encounter at test time. Since explicitly generating images of unseen domains and concepts is an ill-posed problem, we sidestep this issue and we synthesize unseen domains and concepts by interpolating existing ones. To do so, we revisit the \textit{mixup} \cite{zhang2017mixup} algorithm as a tool to obtain partially unseen categories and domains. Indeed, by randomly mixing samples of different categories we obtain new samples which do not belong to a single one of the available categories during training. Similarly, by mixing samples of different domains, we obtain new samples which do not belong to a single source domain available during training.

Under this perspective, mixing samples of both different domains and classes allows to obtain samples that cannot be categorized in a single class and domain of the one available during training, thus they are \textit{novel} both for the semantic and their visual representation. Since higher levels of abstraction contain more task-related information, we perform \textit{mixup} at both image and feature level, showing experimentally the need for this choice. {Moreover, we introduce  a  curriculum-based  mixing  strategy to generate increasingly complex training samples. } 
We show that our \methodName (\methodNameFull) model obtains state-of-the-art performances in both ZSL and DG in standard benchmarks and it can be effectively applied to the combination of the two tasks, recognizing unseen categories in unseen domains.\footnote{The code is available at \textit{\url{https://github.com/mancinimassimiliano/CuMix}}}} 

To summarize, the contributions of this chapter are:
(i) We introduce the ZSL+DG scenario, a first step towards recognizing unseen categories in unseen domains.
(ii) We describe \methodName, the first holistic method able to address ZSL, DG, and the two tasks together. Our method is based on simulating new domains and categories during training by mixing the available training domains and classes both at image and feature level. The mixing strategy becomes increasingly more challenging during training, in a curriculum fashion.
(iii) Through our extensive evaluations and analysis, we show the effectiveness of \methodName in all three settings: namely ZSL, DG and ZSL+DG.

\myparagraph{Problem statement.}
In this chapter, we will considered the ZSL+DG problem. Differently from the incremental learning methods presented in \ref{chap:icl}, here we assume that our new semantic concepts is not available in the form of a training set, but is contained in a semantic descriptor which we receive at test time. Using the semantic descriptors for training classes, we can learn how to match visual features, generalizing their available during training, we can match In the ZSL+DG problem, the goal is to recognize unseen categories (as in ZSL) in unseen domains (as in DG). Formally, let $\mathcal{X}$ denote the input space (e.g. the image space), $\mathcal{Y}$ the set of possible classes and $\mathcal{D}$ the set of possible domains. During training, we are given a set  $\mathcal{S}=\{(x_i,y_i,d_i)\}_{i=1}^n$ where $x_i\in\mathcal{X}$, $y_i \in \mathcal{Y}^s$ and $d_i \in \mathcal{D}^s$. 
 Note that $\mathcal{Y}^s\subset \mathcal{Y}$ and $\mathcal{D}^s\subset \mathcal{D}$ and, as in standard DG, {we have multiple source domains (\ie $\mathcal{D}^s=\cup_{j=1}^m d_j$, with $m>1$) with 
 } different distributions \ie $p_\mathcal{X}(x|d_i)\neq p_\mathcal{X}(x|d_j)$, $\forall i\neq j$. For simplicity, in this section we assume to have exact knowledge about the domain label of each sample. 
 
 In the ZSL+DG problem, the goal is to recognize unseen categories (as in ZSL) in unseen domains (as in DG). Formally, let $\mathcal{X}$ denote the input space (e.g. the image space), $\mathcal{Y}$ the set of possible classes and $\mathcal{D}$ the set of possible domains. During training, we are given a set  $\mathcal{S}=\{(x_i,y_i,d_i)\}_{i=1}^n$ where $x_i\in\mathcal{X}$, $y_i \in \mathcal{Y}^s$ and $d_i \in \mathcal{D}^s$. Note that $\mathcal{Y}^s\subset \mathcal{Y}$ and $\mathcal{D}^s\subset \mathcal{D}$ and, as in standard DG, {we have multiple source domains (\ie $\mathcal{D}^s=\cup_{j=1}^m d_j$, with $m>1$) with} different distributions \ie $p_\mathcal{X}(x|d_i)\neq p_\mathcal{X}(x|d_j)$, $\forall i\neq j$.
Given $\mathcal{S}$ our goal is to learn a function $h$ mapping an image $x$ of 
domains $\mathcal{D}^u\subset \mathcal{D}$ to its corresponding label in a set of 
classes $\mathcal{Y}^u\subset \mathcal{Y}$. Note that in standard ZSL, {while the set of train and test domains are shared, \ie $\mathcal{D}^s\equiv\mathcal{D}^u$}, the label sets are disjoint \ie $\mathcal{Y}^s\cap\mathcal{Y}^u \equiv \emptyset$, {thus $\mathcal{Y}^u$ is a set of \textit{unseen} classes. } On the other hand, in DG we have {a shared output space}, \ie $\mathcal{Y}^s\equiv\mathcal{Y}^u$, but a disjoint set of domains between training and test \ie $\mathcal{D}^s\cap\mathcal{D}^u \equiv \emptyset$, 
{thus $\mathcal{D}^u$ is a set of \textit{unseen} domains}. Since the goal of our work is to recognize unseen classes in unseen domains, we unify the settings of DG and ZSL, considering both semantic- and domain shift at test time \ie $\mathcal{Y}^s\cap\mathcal{Y}^u \equiv \emptyset$ and $\mathcal{D}^s\cap\mathcal{D}^u \equiv \emptyset$.

\section{Related Works}
\label{both:related}
In this section, we review related works in ZSL, and works trying to perform DA and/or DG with techniques linked to the \textit{mixup} algorithm which serves as the base for our method. We will also describe works addressing ZSL under domain shift and/or DG with different semantic spaces, highlighting the differences with our setting.  

\myparagraph{Zero-Shot Learning (ZSL).} 
Traditional ZSL approaches attempt to learn a projection function mapping images/visual features to a semantic embedding space where  classification is performed. This idea is achieved by directly predicting image attributes e.g. \cite{lampert2013awa} or by learning a linear mapping through margin-based objective functions \cite{akata2013label,akata2015evaluation}. Other approaches explored the use of non-linear multi-modal embeddings \cite{xian2016latent}, intermediate projection spaces \cite{zhang2015zero,zhang2016zero} or similarity-based interpolation of base classifiers \cite{changpinyo2016synthesized}. Recently, various methods tackled ZSL from a generative point of view considering Generative Adversarial Networks  \cite{xian2018feature}, Variational Autoencoders (VAE) \cite{schonfeld2019generalized} or both of them \cite{xian2019fvaegan}. While none of these approaches explicitly tackled the domain shift, i.e. visual appearance changes among different domains/datasets, various methods proposed to use domain adaptation technique, e.g. to refine the semantic embedding space, aligning semantic and projected visual features \cite{schonfeld2019generalized} or, in transductive scenarios, to cope with the inherent domain shift existing among the appearance of attributes in different categories \cite{Kodirov_2015_ICCV,fu2015transductive,gan2016learning}. For instance, in \cite{schonfeld2019generalized} a distance among visual and semantic embedding projected in the VAE latent space is minimized. In \cite{Kodirov_2015_ICCV} the problem is addressed through a regularised sparse coding framework, while in \cite{fu2015transductive} a multi-view hypergraph label propagation framework is introduced.

Recently, works have considered also coupling ZSL and DA in a transductive setting. For instance, in \cite{zhuo2019unsupervised} a semantic guided discrepancy measure is employed to cope with the asymmetric label space among source and target domains. In the context of image retrieval, multiple works addressed the sketch-based image retrieval problem \cite{yelamarthi2018sketch,Dutta_2019_CVPR}, even across multiple domains. In \cite{thong2019open} the authors proposed a method to perform cross-domain image retrieval by training domain-specific experts. While these approaches integrated DA and ZSL, none of them considered the more complex scenario of DG, where \textit{no target data} are available.

\myparagraph{Simulating the Domain Shift for Domain Generalization.} As highlighted in Section \ref{sec:da-related}, multiple research efforts have been recently devoted into addressing the domain generalization problem. Here we will recall some of them that are linked to the idea behind of the approach we will present in the next section. For a more detailed overview of DG works, we ask the reader to refer to Section \ref{sec:da-related}. 

In particular, since we mix samples to simulate new domains, our approach is linked with data and feature augmentation strategies for DG \cite{shankar2018generalizing,volpi2018generalizing,volpi2019addressing}. Among them, we can distinguish two main categories: adversarial-based \cite{shankar2018generalizing,volpi2018generalizing,zhou2020deep,zhou2020learning}, trying to simulate novel domains through adversarial perturbations of the original input, and data augmentation-based \cite{volpi2019addressing}, which determines which augmentations to perform in order to improve the generalization capabilities of the model. Differently from these methods, we will specifically employ mixup to perturb input and feature representations.

Similarly, the fact that mixed samples are made increasingly more difficult during training, has a link with episodic strategies for domain generalization, such as \cite{li2019episodic}. In \cite{li2019episodic}, the authors describe a DG procedure which is based on multiple domain-specific and one domain-agnostic networks. During training, a domain-specific feature extractor receives as input images of different domains (i.e. with a different distributions) that the domain agnostic predictor is asked to correctly classify. Vice-versa, the domain-agnostic feature extractor must learn to extract features which even a domain-specific classifier of a different domain (with respect to the one of the input image) should correctly classify. In this way, the domain-agnostic components learn to cope with domain shift in their inputs, similarly to what they will experience at test time. In our method, we will not require domain-specific components, but we will simulate the domain shift by gradually increasing the challenge posed by the mixed samples.

Recently, works have considered mixup in the context of domain adaptation {\cite{xu2019adversarial}} to \eg reinforce the judgments of a domain discrimination. However, we employ mixup from a different perspective \ie simulating semantic and domain shift we will encounter at test time. To this extent, we are not aware of previous methods using mixup for DG and ZSL.

Finally, works have recently considered the heterogeneous domain generalization (HDG) problem \cite{li2019episodic,li2019feature}. The goal of HDG is to train a feature extractor able to produce useful representations for novel
domains and novel categories \cite{li2019feature}. The novel domains have their specific output space (as in MDL, see Section \ref{sec:IL-multitask}). Despite data of novel domains and classes are not present during the feature extractor training phase, data of the novel domains are required to train a classifier for the new domains/categories on top of the agnostic feature extractor. Our ZSL+DG is different since we assume that a model is trained once and uses side information (e.g. word embeddings) to classify unseen categories in unseen domains at the test time, \textit{without} any training samples for new domains and categories.

\section[Recognizing Unseen Categories in Unseen Domains]{Recognizing Unseen Categories in Unseen Domains\footnotemark\footnotetext{M. Mancini, Z. Akata, E. Ricci, B. Caputo. {\sl Towards Recognizing Unseen Categories in Unseen Domains}. European Computer Vision Conference (ECCV) 2020.}}
\label{both:dgzsl}
\subsection{Preliminaries}
From the definitions of Section \ref{both:problem}, we recall that our goal is to learn a function $h$ mapping an image $x$ of unseen domains $\mathcal{D}^u\subset \mathcal{D}$ to its corresponding label in a set of 
unseen classes $\mathcal{Y}^u\subset \mathcal{Y}$.

In the following we divide the function $h$ into three parts: $f$, mapping images into a feature space $\mathcal{Z}$, \ie $f:\mathcal{X}\rightarrow \mathcal{Z}$, $g$ going from $\mathcal{Z}$ to a semantic embedding space $\mathcal{E}$, \ie $g:\mathcal{Z}\rightarrow \mathcal{E}$, and an embedding function {$\omega:\mathcal{Y}^t\rightarrow\mathcal{E}$} 
where $\mathcal{Y}^t\equiv\mathcal{Y}^s$ during training and $\mathcal{Y}^t\equiv\mathcal{Y}^u$ at test time. Note that $\omega$ is a learned classifier for DG while it is a fixed semantic embedding function in ZSL, mapping classes into their vectorized representation extracted from external sources.
Given an image $x$, the final class prediction is obtained as follows:
\begin{equation}
    y^{*} = \text{argmax}_{y} {\omega(y)}^{\intercal}g(f(x)).
\end{equation}
In this formulation, $f$ can be any learnable feature extractor (\eg a deep neural network), while $g$ any ZSL predictor {(\eg a semantic projection layer, as in \cite{xian2019semantic} or a compatibility function among visual features and labels, as in \cite{akata2013label,akata2015evaluation})}. 
The first solution {to address the ZSL+DG problem} could be training a classifier using the aggregation of data from all source domains. In particular, for each sample we could minimize a loss function of the form:
\begin{equation}
    \label{eq:DGZSL-:aggregation}
    \mathcal{L}_{\text{AGG}}(x_i,y_i) = \sum_{y\in\mathcal{Y}^s} \ell(\omega(y)^\intercal g(f(x_i)),y_i)
\end{equation}
with $\ell$ an arbitrary loss function, \eg the cross-entropy loss. In the following, we show how we can use the input to Eq.~\eqref{eq:DGZSL-:aggregation} to effectively recognize unseen categories in unseen domains.

\subsection{Simulating Unseen Domains and Concepts through \textit{Mixup}}

\begin{figure}[tb]
  \centering
  \includegraphics[width=1.\columnwidth]{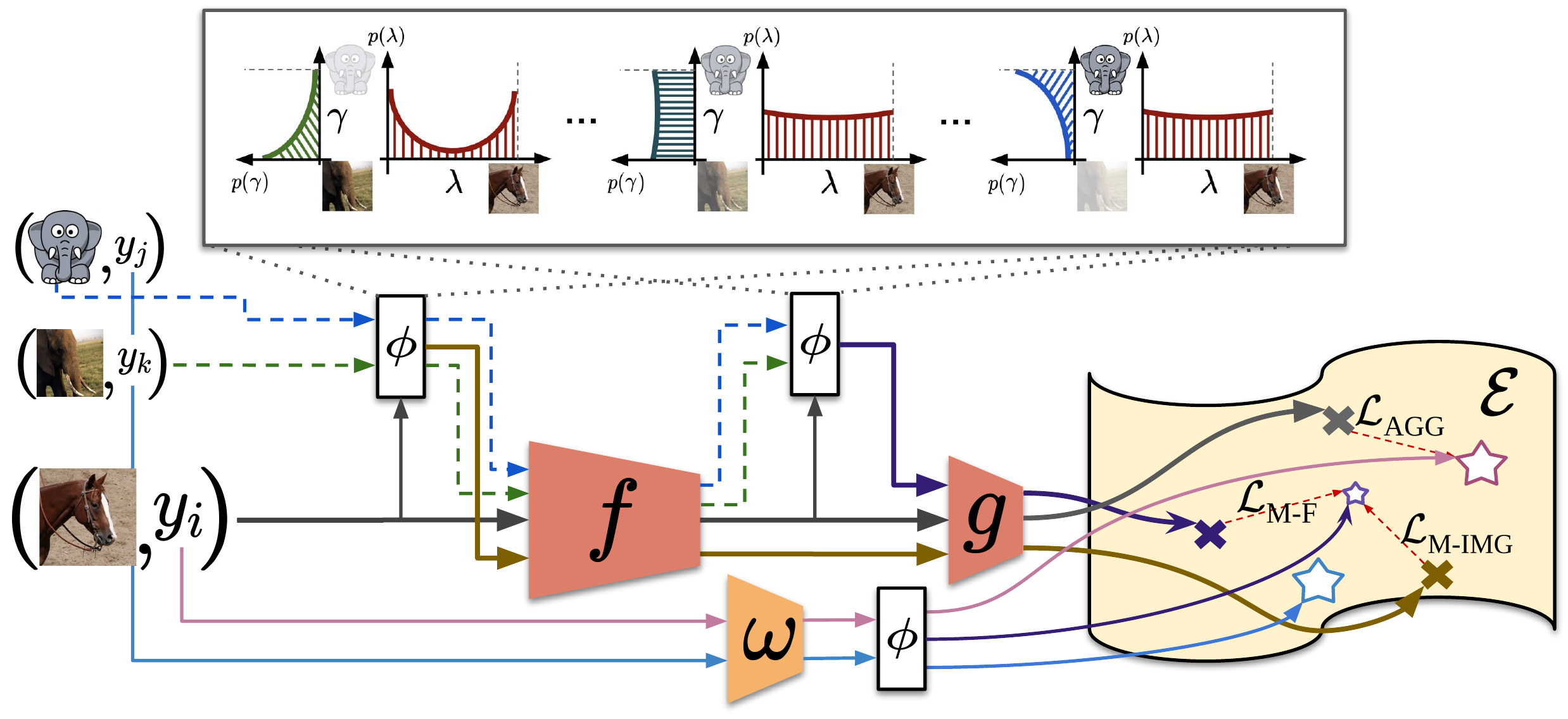} 
  \caption{Our \methodName Framework. Given an image (bottom, \textit{horse}, \textit{photo}), we randomly sample one image from the same (middle, \textit{photo}) and one from another (top, \textit{cartoon}) domain. The samples are mixed through $\phi$ (white blocks) both at image and feature level, with their features and labels projected into the embedding space $\mathcal{E}$ (by $g$ and $\omega$ respectively) and there compared to compute the final objective. Note that $\phi$ varies during training (top part), changing the mixing ratios in and across domains. }
  \label{fig:DGZSL-:method}
 \end{figure}
 
The fundamental problem of ZSL+DG is that, during training, we have neither access to visual data associated to categories in $\mathcal{Y}^u$ nor to data of the unseen domains $\mathcal{D}^u$. One way to overcome this issue in ZSL is to generate samples of unseen classes by learning a generative function conditioned on the semantic embeddings in {$\mathcal{W}=\{\omega(y) | y\in\mathcal{Y}^{s}\}$} \cite{xian2018feature,xian2019fvaegan}. However, since no {description} is available for the unseen target domain(s) in $\mathcal{D}^u$, this strategy is not feasible in ZSL+DG. On the other hand, previous works on DG proposed to synthesize images of unseen domains through adversarial strategies of data augmentation  {\cite{volpi2018generalizing,shankar2018generalizing}}. However, these strategies are not applied to ZSL since they cannot easily be extended to generate data for unseen semantic categories 
$\mathcal{Y}^u$. 

To circumvent this issue, we introduce a strategy to simulate, during training, novel domains and semantic concepts by interpolating from the ones available in $\mathcal{D}^s$ and $\mathcal{Y}^s$. Simulating novel domains and classes allows to train the network to cope with both semantic- and domain shift, the same situation our model will face at test time. Since explicitly generating inputs of novel domains and categories is a complex task, in this section we propose to achieve this goal, by \textit{mixing} images and features of different classes and domains, revisiting the popular \textit{mixup} \cite{zhang2017mixup} strategy. 

In practice, given two elements $a_i$ and $a_j$ of the same space (\eg $a_i,a_j\in\mathcal{X}$), \textit{mixup} \cite{zhang2017mixup} defines a mixing function $\varphi$ as follows: 
\begin{equation}
    \label{eq:DGZSL-:phi-mixup}
    \varphi(a_i,a_j) =\lambda\cdot a_i + (1-\lambda) \cdot a_j
\end{equation}
with $\lambda$ sampled from a beta distribution, \ie $\lambda\sim \text{Beta}(\beta,\beta)$, with $\beta$ an hyperparameter. Given two samples $(x_i,y_i)$ and $(x_j,y_j)$ randomly drawn from a training set $\mathcal{T}$, a new loss term is defined as:
\begin{equation}
    \label{eq:DGZSL-:mixuploss}
    \mathcal{L}_{\text{MIXUP}}((x_i,y_i),(x_j,y_j)) = \mathcal{L}_{\text{AGG}}(\varphi(x_i,x_j),\varphi(\bar{y}_i,\bar{y}_j))
\end{equation}
where $\bar{y}_i\in \Re^{|\mathcal{Y}^s|}$ is the one-hot vectorized representation of label $y_i$. Note that, when mixing two samples and label vectors with $\varphi$, a single $\lambda$ is drawn and applied within $\varphi$ in both image and label spaces. The loss defined in Eq.\eqref{eq:DGZSL-:mixuploss} forces the network to disentangle the various semantic components (\ie $y_i$ and $y_j$) contained in the mixed inputs (\ie $x_i$ and $x_j$) plus the ratio $\lambda$ used to mix them.
This auxiliar task acts as a strong regularizer that helps the network to \eg being more robust against adversarial examples \cite{zhang2017mixup}. Note however that the function $\varphi$ creates input and targets which {do not represent} a single semantic concept in $\mathcal{T}$ but contains characteristics taken from multiple samples and categories, synthesising a \textit{new} semantic concept from the interpolation of existing ones. 

For recognizing unseen concepts in unseen domains at test time, we revisit $\varphi$ to obtain both cross-domain and cross-semantic mixes during training, simulating both semantic- and domain shift. While simulating the semantic shift is a by-product of the original \textit{mixup} formulation, here we explicitly revisit $\varphi$ in order to perform cross-domain mixups. In particular, instead of considering a pair of samples from our training set, we consider a triplet $(x_i,y_i,d_i)$, $(x_j,y_j,d_j)$ and $(x_k,y_k,d_k)$. Given $(x_i,y_i,d_i)$, the other two elements of the triplet are randomly sampled from $\mathcal{S}$, with the only constraint that $d_i=d_k,i\neq k$ and $d_j\neq d_i$. In this way, the triplet contains two samples of the same domain (\ie $d_i$) and a third of a different one (\ie $d_j$). Then, our mixing function $\phi$ is defined as follows: 
\begin{equation}
    \label{eq:DGZSL-:our-phi}
    \phi(a_i,a_j,a_k) = \lambda a_i + (1-\lambda)( \gamma a_j + (1-\gamma) a_k)
\end{equation}
with $\gamma$ sampled from a Bernoulli distribution $\gamma \sim \mathcal{B}(\alpha)$ and $a$ representing either the input $x$ or the vectorized version of the label $y$, \ie $\bar{y}$. Note that we introduced a term $\gamma$ which allows to perform either intra-domain (with $\gamma=0$) or cross-domain (with $\gamma=1$) mixes.  

To learn a feature extractor $f$ and a semantic projection layer $g$ robust to domain- and semantic shift, we propose to use  $\phi$ to simulate both samples and features of novel domains and classes during training. Namely, we simulate the semantic- and domain shift at two levels, i.e. image and class levels. 
 Given a sample $(x_i,y_i,d_i)\in\mathcal{S}$ we define the following loss:
\begin{equation}
    \label{eq:DGZSL-:mixuploss-img}
    \mathcal{L}_{\text{M-IMG}}(x_i,y_i,d_i) = \mathcal{L}_{\text{AGG}}(\phi(x_i,x_j,x_k),\phi(\bar{y}_i,\bar{y}_j,\bar{y}_k)).
\end{equation}
where $(x_i,y_i,d_i)$,$(x_j,y_j,d_j)$,$(x_k,y_k,d_k)$ are randomly sampled from $\mathcal{S}$, with $d_i=d_k$ and $d_j\neq d_k$. 
The loss term in Eq. \eqref{eq:DGZSL-:mixuploss-img} enforces the feature extractor to effectively process inputs of mixed domains/semantics obtained through $\phi$. Inspired by \cite{verma2019manifold}, we design an additional loss acting at the classification level, by enforcing the semantic consistency of mixed features in $\mathcal{E}$. This loss term is defined as: 
\begin{equation}
    \label{eq:DGZSL-:mixuploss-feats}
    \mathcal{L}_{\text{M-F}}(x_i,y_i,d_i) =  \sum_{y\in\mathcal{Y}^s} \ell\biggl(\omega(y)^\intercal g\bigl(\phi(f(x_i),f(x_j),f(x_k))\bigr),\phi(\bar{y}_i,\bar{y}_j,\bar{y}_k)\biggr)
\end{equation} 
where, as before, $(x_j,y_j,d_j),(x_k,y_k,d_k)\sim \mathcal{S}$, with $d_i=d_k,i\neq k$ and $d_j\neq d_k$ and $\ell$ is a generic loss function \eg the cross-entropy loss. 
This second loss term forces the classifier $\omega$ and the semantic projection layer $g$ to be robust to features with mixed domains and semantics. 

While we can simply use a fixed mixing function $\phi$, as defined in Eq.~\eqref{eq:DGZSL-:our-phi}, for Eq.~\eqref{eq:DGZSL-:mixuploss-img} and Eq.~\eqref{eq:DGZSL-:mixuploss-feats}, we found that it is more beneficial to devise a dynamic $\phi$ which changes its behaviour during training, in a curriculum fashion. Intuitively, minimizing the two objectives defined in Eq.\eqref{eq:DGZSL-:mixuploss-img} and Eq.\eqref{eq:DGZSL-:mixuploss-feats} {requires our model to correctly disentangle the various semantic components used to form the mixed samples.} While this is a complex task even for intra-domain mixes (\ie when only the semantic is mixed), mixing samples across domains makes the task even harder, requiring to isolate also domain-specific factors. 
To effectively tackle this task, we choose to act on the mixing function $\phi$. {In particular, we want our $\phi$ to create mixed samples with progressively increased degree of mixing both with respect to content and domain, in a curriculum-based fashion}. 

During training we regulate both $\alpha$ (weighting the probability of cross-domain mixes) and $\beta$ (modifying the probability distribution of the mix ratio $\lambda$), changing the probability distribution of the mixing ratio $\lambda$ and of the cross-domain mix $\gamma$. In particular, given a warm-up step of $N$ epochs and being $s$ the current epoch we set $\beta=\text{min}(\frac{s}{N}\beta_{max},\beta_{max}))$, with $\beta_{max}$ as hyperparameter, while $\alpha=\text{max}(0,\text{min}(\frac{s-N}{N},1)$. As a consequence, the learning process is made of three phases, with a smooth transition among them. We start by solving the plain classification task on a single domain (\ie $s<N$,$\alpha=0$,$\beta=\frac{s}{N}\beta_{max},$). In the subsequent step ($N\leq s<2N$) samples of the same domains are mixed randomly, with possibly different semantics (\ie $\alpha=\frac{s-N}{N}$, $\beta=\beta_{max}$). In the third phase ($s\geq 2N$), we mix up samples of different domains (\ie $\alpha=1$), simulating the domain shift the predictor will face at test time. Figure \ref{fig:DGZSL-:method}, shows a representation of how $\phi$ varies during training (top, white block).   

\myparagraph{Final objective.} {The full training procedure, is represented in Figure \ref{fig:DGZSL-:method}. Given a training sample $(x_i,y_i,d_i)$, we randomly draw other two samples, $(x_j,y_j,d_j)$ and $(x_k,y_k,d_k)$, with $d_i=d_k,i\neq k$ and $d_j\neq d_i$, feed them to $\phi$ and obtain the first mixed input. We then feed $x_i$, $x_j$, $x_k$ and the mixed sample through $f$, to extract their respective features. At this point we use features extracted from other two randomly drawn samples (in the figure, and just for simplicity, $x_j$ and $x_k$ with same mixing ratios $\lambda$ and $\gamma$), to obtain the feature level mixed features needed to build the objective in Eq.\eqref{eq:DGZSL-:mixuploss-feats}. Finally, the features of $x_i$ and the two mixed variants at image and feature level, are fed to the semantic projection layer $g$, which maps them to the embedding space $\mathcal{E}$. At the same time, the labels in $\mathcal{Y}^s$ are projected in $\mathcal{E}$ through $\omega$. Finally, the objectives defined in Eq.\eqref{eq:DGZSL-:aggregation},Eq.\eqref{eq:DGZSL-:mixuploss-img} and Eq.\eqref{eq:DGZSL-:mixuploss-feats} functions are then computed in the semantic embedding space. } 
Our final objective is:
  \begin{equation}
 \label{eq:DGZSL-:final-objective}
 \mathcal{L}_{\text{CuMIX}}(\mathcal{S}) = {|\mathcal{S}|^{-1}}\sum_{\mathclap{(x_i,y_i,d_i)\in\mathcal{S}}}\mathcal{L}_{\text{AGG}}(x_i,y_i)+\\\eta_{\text{I}}\mathcal{L}_{\text{M-IMG}}(x_i,y_i,d_i)+\eta_{\text{F}}\mathcal{L}_{\text{M-F}}(x_i,y_i,d_i) 
 \end{equation}
with $\eta_{\text{I}}$ and $\eta_{\text{F}}$ hyperparameters weighting the importance of the two terms. As $\ell(x,y)$ in both $\mathcal{L}_{\text{AGG}}$, $\mathcal{L}_{\text{M-IMG}}$ and $\mathcal{L}_{\text{M-F}}$, we use the standard cross-entropy loss, even if any ZSL objective can be applied. Finally, we highlight that the optimization is performed batch-wise, thus also the sampling of the triplet considers the current batch and not the full training set $\mathcal{S}$. Moreover, while in Figure \ref{fig:DGZSL-:method} we show for simplicity that the same samples are drawn for $\mathcal{L}_{\text{M-IMG}}$ and $\mathcal{L}_{\text{M-F}}$, in practice, given a sample, the random sampling procedure of the other two members of the triplet is held-out twice, one at the image level and one at the feature level. Similarly, the sampling of the mixing ratios $\lambda$ and cross domain factor $\gamma$ of $\phi$ is held-out sample-wise and twice, one at image level and one at feature level. As in Eq.~\eqref{eq:DGZSL-:phi-mixup}, $\lambda$ and $\gamma$ are kept fixed across mixed inputs/features and their respective targets in the label space. 

\myparagraph{Discussion.} We present similarities between our \methodName framework with DG and ZSL methods. In particular, presenting the classifier with noisy features extracted by a non-domain specialist network, has a similar goal as the episodic strategy for DG described in \cite{li2019episodic}. On the other hand, here we sidestep the need to train domain experts by directly presenting as input to our classifier features of novel domains that we obtain by interpolating the available sources samples.  
Our method is also linked to \textit{mixup} approaches developed in DA {\cite{xu2019adversarial}}. Differently from them, we use \textit{mixup} to simulate unseen domains rather then to progressively align the source to the given target data.

Our method is also related to ZSL frameworks based on feature generation \cite{xian2018feature,xian2019fvaegan}. While the quality of our synthesized samples is lower since we do not exploit attributes for conditional generation, we have a lower computational cost. {In fact, during training we simulate the test-time semantic shift without generating samples of unseen classes. Moreover, we do not require additional training phases on the generated samples or the availability of unseen class attributes to be available beforehand.}

\subsection{Experimental results}
\begin{figure}[tb]
  \centering
  \includegraphics[width=1.\columnwidth]{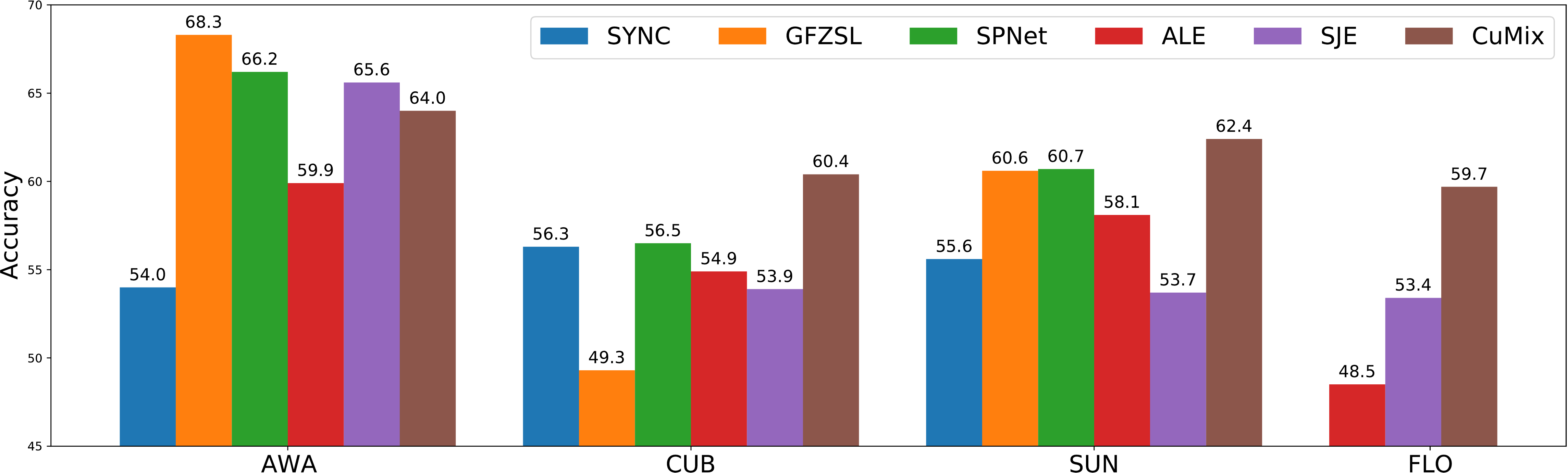} 
   \caption{ZSL results on CUB, SUN, AWA and FLO datasets with ResNet-101 features.}
  \label{fig:dgzls-zsl-results}
 \end{figure}

\subsubsection{Datasets and implementation details}
We assess \methodName in three scenarios: ZSL, DG and the proposed ZSL+DG setting. 

\myparagraph{ZSL}. We conduct experiments on {four} standard benchmarks: Caltech-UCSD-Birds 200-2011 (CUB)~\cite{welinder2010cub}, SUN attribute (SUN)~\cite{patterson2012sun}, Animals with Attributes (AWA)~\cite{lampert2013awa} and Oxford Flowers (FLO)~\cite{nilsback2008flo}. CUB contains 11,788 images of 200 bird species, with 312 attributes, SUN 14,430 images of 717 scenes annotated with 102 attributes, and AWA 30,475 images of 50 animal categories with 85 attributes. Finally, FLO is a fine-grained dataset of flowers, containing 8,189 images of 102 categories. As semantic representation, we use the visual descriptions of \cite{reed2016learning}, following 
\cite{xian2018feature,xian2019semantic}. For each dataset, we use the train, validation and test split provided by \cite{xian2018zeroshotgood}. In all the settings we employ features extracted from the second-last layer of a ResNet-101 \cite{he2016deep} pre-trained on ImageNet as image representation. For \methodName, we consider $f$ as the identity function and as $g$ a simple fully connected layer, performing our version of mixup directly at the feature level while applying our alignment loss in the embedding space. All hyperparameters have been set following \cite{xian2018zeroshotgood}.

\myparagraph{DG}. We perform experiments on the PACS dataset \cite{li2017deeper}with 9,991 images of 7 semantic classes in 4 different visual domains, \textit{art paintings}, \textit{cartoons}, \textit{photos} and \textit{sketches}. For this experiment we use the standard train and test split defined in \cite{li2017deeper}, with the same validation protocol. We use as base architecture a ResNet-18 \cite{he2016deep} pre-trained on ImageNet. For our model, we consider $f$ to be the ResNet-18 while $g$ to be the identity function. We use the same training hyperparameters and protocol of \cite{li2019episodic}. 

\myparagraph{ZSL+DG}. Since no previous work addressed the problem of ZSL+DG, there is no benchmark on this task. As a valuable benchmark, we choose DomainNet~\cite{peng2019moment}, a recently introduced dataset for multi-source domain adaptation \cite{peng2019moment} with a large variety of domains, visual concepts and possible descriptions. It contains approximately 600'000 images from 345 categories and 6 domains,  \textit{clipart}, \textit{infograph}, \textit{painting}, \textit{quickdraw}, \textit{real} and \textit{sketch}. 

To convert this dataset from a DA to a ZSL scenario, we need to define an unseen set of classes. Since \methodName uses a network pre-trained on ImageNet \cite{russakovsky2015imagenet}, the set of unseen classes can not contain any of the classes present in ImageNet following the good practices in \cite{xian2017zero}. We build our validation + test set with 100 classes that contain at least 40 images per domain and that has no overlap with ImageNet. We reserve 45 of these classes for the unseen test set, matching the number used in \cite{thong2019open}, and the remaining 55 classes for the unseen validation set. The remaining 245 classes are used as seen classes during training. 

We set the hyperparameters of each method by training on all the images of the seen classes on a \textit{subset} of the source domains and validating on all the images of the validation set from the held-out source domain. After the hyperparameters are set, we retrain the model on the training set, i.e. 245 classes, and validation set, i.e. 55 classes, of a total number of 300 classes. Finally, we report the final results on the 45 unseen classes. As semantic representation we use word2vec embeddings \cite{mikolov2013efficient} extracted from the Google News corpus and \textit{L2}-normalized, following \cite{thong2019open}.  For all the baselines and our method, we employ as base architecture a ResNet-50 \cite{he2016deep} pre-trained on ImageNet, using the same number of epochs and SGD with momentum as optimizer, with the same hyperparameters of \cite{thong2019open}.

\subsubsection{Results}

\myparagraph{ZSL.} 
In the ZSL scenario, we choose as baselines standard inductive methods plus more recent approaches. In particular we report the results of ALE~\cite{akata2013label}, SJE~\cite{akata2015evaluation}, SYNC~\cite{changpinyo2016synthesized}, GFZSL~\cite{verma2017simple} and SPNet~\cite{xian2019semantic}. ALE~\cite{akata2013label} and SJE~\cite{akata2015evaluation} are linear compatibility methods using a ranking loss and the structural SVM loss respectively. SYNC~\cite{changpinyo2016synthesized} learns a mapping from the feature space and the semantic embedding space by means of phantom classes and a weighted graph. GFZSL~\cite{verma2017simple} employs a generative framework where each class-conditional distribution is modeled as a multivariate Gaussian. Finally, SPNet~\cite{xian2019semantic} learns a semantic projection function from the feature space through the image embedding space by minimizing the standard cross-entropy loss. 

Our results grouped by datasets are reported in Figure \ref{fig:dgzls-zsl-results}. 
Our model achieves performance either superior or comparable to the state of the art in all benchmarks but AWA. We believe that in AWA learning a better alignment between visual features and attributes may not be as effective as improving the quality of the visual features. Especially, although the names of the test classes do not appear in the training set of ImageNet, for AWA being a non-fine-grained dataset, the information content of the test classes is likely represented by the ImageNet training classes. Moreover, for non-fine-grained datasets, finding labeled training data may not be as challenging as it is in fine-grained datasets. Hence, we argue that zero-shot learning is of higher practical interest in fine-grained settings. Indeed \methodName is effective in fine-grained scenarios (\ie CUB, SUN, FLO) where it consistently outperforms the state-of-the-art approaches.

\begin{table*}[t]
		\caption{Domain Generalization accuracies on PACS with ResNet-18.} 
		\centering
		\scalebox{0.8}{
		{
		\begin{tabular}{ p{4em}| >{\centering}p{3.7em} >{\centering}p{3.7em} >{\centering}p{3.7em} >{\centering}p{4.6em} >{\centering}p{4.3em} >{\centering}p{3.7em} >{\centering}p{4em} | >{\centering\arraybackslash}p{3.7em}}
		&AGG&DANN&MLDG&CrossGrad& MetaReg&JiGen&Epi-FCR&\methodName\\
		Target&&\cite{ganin2016domain}&\cite{li2018learning}&\cite{shankar2018generalizing}&\cite{balaji2018metareg}&\cite{carlucci2019domain}&\cite{li2019episodic}&\\
		\hline
		Photo   &  94.9 & 94.0  & 94.3  &94.0   &94.3   & \textbf{96.0} &93.9 & 95.1\\   
		Art & 76.1  &81.3   &79.5   &78.7   &79.5   &79.4   &82.1   &\textbf{82.3}\\
		Cartoon & 73.8&73.8& \textbf{77.3}&73.3&75.4&75.3&77.0&76.5\\
		Sketch &69.4&\textbf{74.3}&71.5&65.1&72.2&71.4&73.0&72.6\\
		\hline
		Average&78.5&80.8&80.7&80.7&77.8&80.4&81.5&\textbf{81.6}
		\end{tabular}}
		\label{tab:DGZSL-dg-results}}
\end{table*}

These results show that our model based on \textit{mixup} achieves competitive performances on ZSL by simulating the semantic shift the classifier will experience at test time. To this extent, our approach is the first to show that mixup can be a powerful regularization strategy for the challenging ZSL setting.

\myparagraph{DG.} The second series of experiments consider the standard DG scenario. Here we test our model on the PACS dataset using a ResNet-18 architecture. As baselines for DG we consider the standard model trained on all source domains together (AGG), the adversarial strategies in \cite{ganin2016domain} (DANN) and \cite{shankar2018generalizing}, the meta learning-based strategy MLDG \cite{li2018learning} and MetaReg \cite{balaji2018metareg}. Moreover we consider the episodic strategy presented in \cite{li2019episodic} (Epi-FCR). 

As shown in Table \ref{tab:DGZSL-dg-results}, our model achieves competitive results comparable to the state-of-the-art episodic strategy Epi-FCR \cite{li2019episodic}. Remarkable is the gain obtained with respect to the adversarial augmentation strategy CrossGrad \cite{shankar2018generalizing}. Indeed, synthesizing novel domains for domain generalization is an ill-posed problem, since the concept of unseen domain is hard to capture. However, with \methodName we are able to simulate inputs/features of novel domains by simply interpolating the information available in the samples of our sources. Despite containing information available in the original sources, our approach produces a model more robust to domain shift. 

Another interesting comparison is against the self-supervised approach JiGen \cite{carlucci2019domain}. Similarly to \cite{carlucci2019domain} we employ an additional task to achieve higher generalization abilities to unseen domains. While in \cite{carlucci2019domain} the JigSaw puzzles \cite{noroozi2016unsupervised} are used as a secondary self-supervised task, here we employ the mixed samples/features in the same manner. 
The improvement in the performances of \methodName highlights that recognizing the semantic of mixed samples acts as a more powerful secondary task to improve robustness to unseen domains.

Finally, it is worth noting that \methodName performs a form of episodic training, similar to Epi-FCR \cite{li2019episodic}. However, while Epi-FCR considers multiple domain-specific architectures (to simulate the domain experts needed to build the episodes), we require a single domain agnostic architecture. We build  
our episodes by making the \textit{mixup} among images/features of different domains increasingly more drastic. Despite not requiring any domain experts, \methodName achieves comparable performances to Epi-FCR, showing the efficacy of our strategy to simulate unseen domain shifts.

\myparagraph{Ablation study.} In this section, we ablate the various components of CuMix. We performed the ablation on the PACS benchmark for DG, since this allows us to show how different choices act on the generalization to unseen domains. In particular, we ablate the following implementation choices: 1) mixing samples at the image level, feature level or both 2) impact of our curriculum-based strategy for mixing features and samples. 

As shown in Table \ref{tab:DGZSL-ablation-study-pacs}, mixing samples at feature level produces a clear gain on the results with respect to the baseline, while mixing samples only at image level can even harm the performance. This happens particularly in the \textit{sketch} domain, where mixing samples at feature level produces a gain of ~2\% while at image level we observe a drop of ~10\% with respect to the baseline. This could be explained by mixing samples at image level producing inputs that are too noisy for the network and not representative of the actual shift experienced at test time. Mixing samples at feature level instead, after multiple layers of abstractions, allows to better synthesize the information contained in the different samples, leading to more reliable features for the classifier. Using both of them we obtain higher results in almost all domains. 

Finally, we analyze the impact of the curriculum-based strategy for mixing samples and features.  
As the table shows, adding the curriculum strategy allows to boost the performances for the most difficult cases (i.e. sketches) producing a further accuracy boost. Moreover, applying this strategy allows to stabilize the training procedure, as demonstrated experimentally. 

\begin{table*}[t]
			\caption{Ablation on PACS dataset with ResNet-18 as backbone.}
		\centering
		{
		\begin{tabular}{ c c c c | c  c  c  c | c}
		$\mathcal{L}_{\text{AGG}}$ &$\mathcal{L}_{\text{M-IMG}}$&$\mathcal{L}_{\text{M-F}}$&Curriculum&Art&Cartoon&Photo&Sketch&Avg.\\
		\hline
		\ding{51} & & & & 76.1  & 73.8  & 94.9 & 69.4 & 78.5  \\\hline
		 \ding{51}& \ding{51} & &   &78.4 &72.7   &94.7  &59.5  & 76.3  \\
	\ding{51}  &  & \ding{51} & &81.8&\textbf{76.5} &{94.9}  &71.2  &81.1  \\ 
	 \ding{51}& \ding{51} & \ding{51}   & & \textbf{82.7}& 75.4& \textbf{95.4} & 71.5 & 81.2\\ \hline
    	\ding{51}	 &\ding{51} &\ding{51}  & \ding{51}&82.3&\textbf{76.5}& 95.1 & \textbf{72.6} &\textbf{81.6} \\ 
		\end{tabular}
		}
		\label{tab:DGZSL-ablation-study-pacs}
\end{table*}

\myparagraph{ZSL+DG.} On the proposed ZSL+DG setting we use the DomainNet dataset, training on five out of six domains and reporting the average per-class accuracy on the held-out one. We report the results for all possible target domains but one, \ie real photos, since our backbone has been pre-trained on ImageNet, thus the photo domain is not an unseen one. Since no previous methods addressed the ZSL+DG problem, in this section we consider simple baselines derived from the literature of both ZSL and DG. The first baseline is a standard ZSL model without any DG algorithm (i.e. the  standard AGG): as ZSL method we consider SPNet \cite{xian2019semantic}. The second baseline is a DG approach coupled with a ZSL algorithm. To this extent we select the state-of-the-art Epi-FCR as the DG approach, coupling it with SPNet. As reference, we also evaluate the performance of standard \textit{mixup} coupled with SPNet.

As shown in Table \ref{tab:DGZSL-domainnet}, CuMix achieves competitive performances in ZSL+DG setting when compared to a state-of-the-art approach for DG (Epi-FCR) coupled with a state-of-the-art one for ZSL (SPNet), outperforming this baseline in almost all settings but \textit{sketch} and, in average by almost $1\%$. Particularly interesting are the results on the \textit{infograph} and \textit{quickdraw} domains. These two domains are the ones where the shift is more evident as highlighted by the lower results of the baseline. In these scenarios, our model consistently outperforms the competitors, with a remarkable gain of more than 1.5\% in average accuracy per class with respect to the ZSL only baseline. We want to highlight also that DomainNet is a challenging dataset, where almost all standard DA approaches are ineffective or can even lead to negative transfer \cite{peng2019moment}. CuMix however is able to overcome the unseen domain shift at test time, improving the performance of the baselines in all scenarios. Our model consistently outperforms SPNet coupled with the standard \textit{mixup} strategy in every scenario. This demonstrates the efficacy of the choices in \methodName for revisiting \textit{mixup} in order to recognize unseen categories in unseen domains.

\begin{table*}[t]
			\caption{ZSL+DG scenario on the DomainNet dataset with ResNet-50 as backbone.} 
		\centering
		
		{
		\begin{tabular}{ l |  c  c  c  c  c | c   }
		Method&Clipart&Infograph&Painting&Quickdraw&Sketch&Avg.\\
		\hline
                             SPNet    &{26.0}  &16.9  & 23.8 & 8.2  & 21.8  &19.4 \\
		                           \textit{mixup}+SPNet    &27.2  &16.9  & 24.7 & 8.5  & 21.3  & 19.7\\
		
        \multirow{1}{*}{Epi-FCR+SPNet}
                               &{26.4} & 16.7  & 24.6  & 9.2 & \textbf{23.2}    &20.0\\
        \hline
        \methodName    
                                     &\textbf{27.6} &\textbf{17.8}  & \textbf{25.5}  & \textbf{9.9}  &{22.6}  & \textbf{20.7}  \\
		\end{tabular}}
		\label{tab:DGZSL-domainnet}
\end{table*}

\subsection{Conclusions}
In this section, we proposed the novel ZSL+DG scenario. In this setting, during training, we are given a set of images of multiple domains and semantic categories and our goal is to build a model able to recognize unseen concepts, as in ZSL, in unseen domains, as in DG. To solve this problem we design CuMix, the first algorithm which can be holistically and effectively applied to DG, ZSL, and ZSL+DG. \methodName is based on simulating inputs and features of new domains and categories during training by mixing the available source domains and classes, both at image and feature level. 
Experiments on public benchmarks show the effectiveness of CuMix, achieving state-of-the-art performances in almost all settings in all tasks. Future works might investigate alternative data-augmentation schemes in the ZSL+DG setting as well as the use of novel formulations of the mixing functions. Moreover, it would be interesting to extend CuMix to the more realistic Generalized-ZSL scenario, where the model must recognize both seen and unseen categories.

\chapter{Conclusions and Future Works}
\label{chap:conclusions}
\section{Summary of contributions}
In this thesis, we analyzed the capability of deep neural networks to generalize to unseen input distributions and to include knowledge not present in their initial training set, with the final goal of building deep models able to recognize new/unseen categories in unseen visual domains. 

In Chapter \ref{chap:da}, we started by analyzing the problem from a perspective of the input the network receives, considering scenarios where training (source) and test (target) output spaces do not change but their input distribution does. In particular, in Section \ref{sec:da-latent} we considered the problem of latent domain discovery in domain adaptation. In this setting, we assume the availability of unlabeled target data during training and that either source or target domains (or both) are a mixture of multiple latent domains. 
In this context, we proposed the first deep neural network able to work in this scenario. Our architecture is made of two main components, namely novel multi-domain alignment layers (mDA) and a domain prediction branch. The mDA layers perform batch-normalization (BN) \cite{ioffe2015batch}, extending previous works on domain adaptation \cite{carlucci2017autodial,carlucci2017just,li2016revisiting}, through weighted statistics, computed using the domain probabilities extracted by the domain prediction branch. The domain prediction branch relies on the assumption that similar inputs should produce similar activations, and it is trained through a simple entropy-loss, without requiring any domain label. Our results show that our framework can successfully enable the deep model to discover latent domains and outperform standard single-source methods.

As a second step, we removed the assumption of having target data available during training by considering the domain generalization (DG) scenario (Section \ref{sec:da-dg}). Here we build on the idea that improvements on the performances of a DG model can be achieved by modeling the similarity of a target sample to the available source domains. We thus develop a simple extension to the latent domain discovery framework which makes use of domain labels (if available) and of the domain prediction branch at test time, to decide which of the source domains should contribute more in the final decision. In particular, we use domain-specific BN layers, weighting their activations using the similarity that the target sample has with the source domains. Experiments on robotics scenarios show the effectiveness of the approach in multiple place categorization benchmarks under various domain shifts (e.g. light conditions, seasons, environments) with and without the presence of domain labels (Section \ref{sec:wbn}). Subsequently, we extend this solution to levels of the network different from BN layers. In particular, we consider merging the activation of domain-specific classifiers at test time, with their importance weighted again by the similarity of a target sample to the source domains. We also explore the use of different kinds of merging strategies, balancing domain-specific features with domain-agnostic ones. Results show the effectiveness of the approach against domain generalization baselines in standard benchmarks (Section \ref{sec:bsf}).

While DG is one of the domain adaptation scenario where target data are not present, other settings can be considered, depending on the amount of information we have about our target domain. Studying other aspects of the problem, in Section \ref{sec:da-continuous} we focused on the continuous domain adaptation scenario, where a single source domain is available during training (without any target data) and adaptation must be performed exploiting the incoming stream of target samples at test time, without access to the original training set. In this context, we develop an extension to the domain alignment layers \cite{carlucci2017autodial,carlucci2017just,li2016revisiting} to tackle this problem. In particular, we show how updating the statistics of BN layers using the incoming stream of target data can be a simple yet effective strategy for tackling this problem. We assess the performance of our model, ONDA, on a robotic object classification task, collecting and releasing a dataset for studying this unexplored problem, containing multiple objects in various acquisition conditions.

Finally, we considered the predictive domain adaptation (PDA) scenario where, during training, we have a single labeled source domain and multiple unlabeled auxiliary domains, each of them with a description (i.e. metadata) attached. The goal of this problem is to build a model able to address the classification task in the target domain by using just the target-specific metadata. We develop the first deep learning model for this problem, AdaGraph (Section \ref{sec:da-predictive}), which builds on a graph, where each node is a domain with attached its domain-specific parameters, and its edge is the distance among the metadata of the two connected domains. At test time, given the target domain metadata, we obtain the target-specific parameters through a weighted combination of its closest nodes in the graph. Due to their simplicity and the easiness of linearly combining them, we use domain-specific BN layers as domain-specific parameters. To improve the estimated target statistics, we also incorporate a continuous domain adaptation strategy into the framework, extending the previously described ONDA algorithm. Experiments show how our model outperforms standard PDA approaches, with the continuous update strategy being surpassing state-of-the-art approaches in continuous domain adaptation.

In Chapter \ref{chap:icl}, we moved to the problem of extending the output space of a pre-trained architecture to new semantic categories. We started by analyzing the problem of multi-domain learning, where the goal is to add new (classification) tasks to a pre-trained model without harming the performance on old tasks and with as few task-specific parameters as possible. Our contribution (Section \ref{sec:IL-multitask}) has been showing how affinely transformed task-specific binary masks applied to the original network weights can allow a network to learn multiple models with (i) performances close to networks fine-tuned for the specific tasks and (ii) with very little overhead in terms of the number of parameters required for each task. We assess the performance of our model on the challenging Visual Domain Decathlon, showing performance comparable with more complicated/multi-stages state-of-the-art approaches.

In Section \ref{sec:IL-semantic-seg}, we focused on a different problem, incremental class learning. In this task, we want to add new knowledge to a pre-trained model without having access to the original training set, thus addressing the catastrophic forgetting problem. We analyzed this task in semantic segmentation discovering how the performance of standard incremental learning algorithms is hampered by the change of the semantic of the background class in different learning steps, a problem which we named the background shift. Indeed, in a given learning step, the background might contain pixels of classes learned in previous steps as well as pixels of classes we will learn in future ones. We showed how a simple modification of standard cross-entropy and distillation losses, taking explicitly into account the different meaning of the background across different learning steps and coupled with an ad-hoc initialization procedure, can effectively address both catastrophic forgetting and background shift, even in large-scale scenarios (e.g. ADE-20k).

In the final section of Chapter \ref{chap:icl}, we studied a more challenging problem, open-world recognition (OWR). In this task, we must not only be able to add new concepts to a pre-trained model, but also to detect unknown concepts if received as inputs. In this scenario, we developed DeepNNO (Section \ref{sec:OWR-deepnno}), the first end-to-end trainable model in OWR, extending standard non-parametric algorithms \cite{bendale2015towards} with losses and training schemes preventing catastrophic forgetting. We then showed how clustering-based objectives and trainable class-specific rejection thresholds can further boost the performances of deep OWR model (Section \ref{sec:OWR-dario}). The experiments on standard datasets and robotics scenarios showed the efficacy of the two approaches and the importance of each design choice. Moreover, we described and tested a simple pipeline for Web-aided OWR, where knowledge about new classes is not given by an external 'oracle' but automatically retrieved from web queries (Section \ref{sec:OWR-web}). We believe our algorithms and our web-based pipeline constitute a first meaningful step towards autonomously learning real-world agents.

Finally, in Chapter \ref{chap:both}, we merged the two worlds, analyzing whether it is possible to build a model recognizing unseen classes in unseen domains. In particular, we described the problem of Zero-Shot Learning (ZSL) under Domain Generalization (DG), where, during training, we are given images of a set of classes in multiple source domains and, at test time, we are asked to recognize different, unseen categories depicted in unseen visual domains. In this scenario, we must learn how to map images into a semantic embedding space of class descriptions (e.g. word embeddings) making sure that the mapping generalizes both to unseen semantic classes (addressing the semantic shift problem) and to unseen domains (addressing the domain shift problem). We developed the first simple solution to this problem based on \textit{mixup} \cite{zhang2017mixup}. In particular, our idea was to simulate the shifts we will encounter at test time by simulating samples (and features) of new domains and/or categories by mixing the domains and classes available at training time. Moreover, we made the mixes increasingly more challenging during training by increasing both the probability of having a high mixing ratio and of cross-domain mixing. Our approach, named CuMix, showed remarkable results on ZSL, DG, and the proposed ZSL+DG, being not only the first holistic approach for ZSL and DG but also the first model that effectively recognizes unseen categories in unseen domains.

\section{Open problems and future directions}
While in this thesis we studied how to build deep learning models generalizing to either new visual domains (Chapter \ref{chap:da}) or new semantic concepts (Chapter \ref{chap:icl}) or both (Chapter \ref{chap:both}), multiple problems remain to be addressed and multiple directions to be explored towards having visual systems recognizing new semantic concepts in arbitrary visual domains.

Starting with the proposed solutions, for each of the proposed algorithms, we briefly discussed possible immediate extensions as well as interesting research directions worth to be explored. For instance, in Chapter \ref{chap:da} we discussed multiple solutions involving domain-specific parameters whose activations are merged by using the weights obtained a domain prediction branch. For all these solutions, it might be interesting to investigate how to strengthen the domain classifier. For instance, one can avoid the use of a domain-specific classifier but rely on the distances of the activations from the various domain-specific distributions (e.g. statistics of BN layers) as a measure of domain similarity. On the other hand, stronger clustering objectives could be applied as objectives for the domain prediction branch to strengthen the discovering of the latent domains. Moreover, it would be interesting to investigate if the algorithms of Chapter \ref{chap:da} can be extended to use parameters beyond standard BN layers, as preliminary experiments with classifiers in Section \ref{sec:bsf}. To this extent, also the affinely transformed binary masks of Section \ref{sec:IL-multitask} can be a good starting point for having simple and easy to combine domain-specific parameters.

When tackling Predictive DA with AdaGraph (Section \ref{sec:da-predictive}), we assign to each domain a node in our graph. However, the current formulation has two main drawbacks. First, it is not scalable if the set of possible metadata descriptors increases; second it considers all the metadata equally important for addressing the domain shift problem. Future works might explore different strategies for including metadata-specific information as well as modeling the importance of the different metadata components. For instance, a possible solution could be to employ domain-specific alignment layers per each metadata group (e.g. viewpoint, year of production) and learning how to optimally recombine their activations for the final prediction. 

Other interesting research directions can be drawn from the works in Chapter \ref{chap:icl}. For instance, the first question is whether the multi-domain algorithm presented in Section \ref{sec:IL-multitask} can be applied to other scenarios (e.g. incremental class learning) and even to tackle the domain shift problem (e.g. PDA, DG). In the latter case, one can use one binary-mask per network parameter per domain, combining them at test time based on the target domain metadata (PDA) or similarity to the source domains (DG). Another interesting question is whether the model can exploit the relationships among the single task/domains through side connections, in such a way that each task/domain benefits from the others.

For what concerns incremental learning in semantic segmentation,  it would be interesting to quantify the background shift and understanding whether the different kinds of shift (e.g. background containing old classes vs background containing classes we will learn in the future) require more specific solutions than the general one we designed in Section \ref{sec:IL-semantic-seg}. On the other hand, it would be interesting to verify if the effectiveness of our MiB algorithm (or simple extensions) generalizes to other problems where the semantic of the background is uncertain, such as incremental learning in object detection \cite{shmelkov2017incremental} and instance segmentation \cite{porzi2019seamless} as well as non-incremental tasks such as weakly-supervised semantic segmentation \cite{bearman2016whats}, generalized zero-shot learning \cite{xian2019semantic}, and dataset merging \cite{dmitriev2019learning}.

In OWR, an interesting future work would be to quantify the robustness of OWR algorithms to the domain shift. In this context, we could verify how much their capabilities of detecting unknowns and recognizing known concepts are affected by changes in the input distributions. Moreover, a very important research direction would be improving the various components of the web-based pipeline sketched in Section \ref{sec:OWR-web}. For instance, we could develop a tool for automatic and robust labeling of the detected unknowns. Moreover, we could design algorithms for filtering the noisy web images retrieved and/or dealing with both noisy labels and domain shift while learning the new categories. We believe the latter being a promising direction towards having robotics visual systems learning fully autonomously from the environment they interact with.

Finally, in Chapter \ref{both:dgzsl}, we introduce a new research problem (ZSL+DG) and algorithm (CuMix) with the aim of encouraging the community towards developing models tackling both domain and semantic shift together. However, we believe that our ZSL+DG problem is just the beginning of this journey. Indeed, in principle, we would like the semantic space of our models to consider both seen and unseen categories (as in Generalized ZSL \cite{xian2018zeroshotgood}) in arbitrary domains, while requiring the minimum number of source domains possible (even just one, as recent works in single-source domain generalization \cite{volpi2018generalizing,volpi2019addressing,qiao2020learning}). Moreover, we might receive data for new classes over time, as in incremental learning. In this case, we would like our model to recognize old seen, new seen, and still unseen categories at test time. This would require our model to address both semantic shift, with the relative bias among the set of classes (as in GZSL), the catastrophic forgetting problem \cite{french1999catastrophic} and related ones, such as our identified background shift. Moreover, if the data for the new classes come from new domains, we want our model to also address the domain shift problem. Despite that with CuMix and ZSL+DG we focused on a subset of these problems, We believe that the contributions of Chapter \ref{chap:both} and the findings of this thesis, will push researchers into exploring ways to overcome both domain and semantic shift together, towards building visual algorithms able to cope with the large and unpredictable variability of the real world.

\appendix
\chapter{Recognition across New Visual Domains}
\section{Latent Domain Discovery}
\subsection{mDA layers formulas}
\label{sec:formulas}
From Section \ref{sec:da-latent}, we have the output of our mDA layer denoted by
\begin{equation}
y_{i}=\mDA(x_i, \vct{w}_i; \vct{\hat{\mu}}, \vct{\hat{\sigma}})=\sum_{d\in \set{D}} w_{i,d} \hat{x}_{i,d},
\end{equation}
where, for simplicity:
\begin{equation}
\hat{x}_{i,d}=\frac{x_i - \hat{\mu}_d}{\sqrt{\hat{\sigma}_d^2 + \epsilon}},
\end{equation}
and the statistics are given by
\begin{equation}
\begin{aligned}
  \hat{\mu}_d &= \sum_{i=1}^{\con{b}} \hat{w}_{i,d} x_i, \\
  \hat{\sigma}_d^2&=\sum_{i=1}^{\con{b}} \hat{w}_{i,d} (x_i-\hat{\mu}_d)^2,
\end{aligned}
\end{equation}
where
$\hat{w}_{i,d}=w_{i,d}/\sum_{j=1}^{\con{b}}w_{j,d}$.

From the previous equations we can derive the partial derivative of the loss function with respect to both the input $x_i$ and the domain assignment probabilities $w_{i,d}$.
Let us denote $\frac{\partial L}{\partial y_i}$ the partial derivative of the loss function $L$ with respect to the output $y_i$ of the mDA layer.
We have:
\begin{equation}
\begin{aligned}
  \frac{\partial \hat{x}_{i,d}}{\partial \hat{\sigma}_{d^*}^2} &=
    -\ind{d=d^*} \frac{1}{2} (x_i - \hat{\mu}_{d^*}) \cdot (\hat{\sigma}_{d^*}^2 + \epsilon)^{-\frac{3}{2}}, \\
  \frac{\partial \hat{x}_{i,d}}{\partial \hat{\mu}_{d^*}} &=
    -\ind{d=d^*} (\hat{\sigma}_{d^*}^2 + \epsilon)^{-\frac{1}{2}},
\end{aligned}
\end{equation}
and
\begin{equation}
\begin{aligned}
  \frac{\partial \hat{\sigma}_d^2}{\partial x_{i}} &=
    2\,\hat{w}_{i,d}\cdot(x_i-\hat{\mu}_d), &
  \frac{\partial \hat{\mu}_d}{\partial x_{i}} &=
    \hat{w}_{i,d}.
\end{aligned}
\end{equation}
Thus, the partial derivative of $L$ w.r.t. the input $x_i$ is:
\begin{equation}
  \frac{\partial L}{\partial x_{i^*}} =
    \sum_{d\in \set{D}} \frac{w_{i^*,d}}{\sqrt{\hat{\sigma}_d^2 + \epsilon}} \left[
      \frac{\partial L}{\partial y_{i^*}} - A_d - \hat{x}_{i^*,d} B_d
    \right],
\end{equation}
where:
\begin{equation}
\label{eqn:a-b}
\begin{aligned}
  A_d &= \sum_{i=1}^\con{b} \hat{w}_{i,d} \frac{\partial L}{\partial y_{i}}, \\
  B_d &= \sum_{i=1}^\con{b} \hat{w}_{i,d} \hat{x}_{i,d} \frac{\partial L}{\partial y_{i}}.
\end{aligned}
\end{equation}
For the domain assignment probabilities $w_{i,d}$ we have:
\begin{align}
  \frac{\partial \hat{\mu}_d}{\partial \hat{w}_{i,d^*}} &= \ind{d=d^*} x_{i}, \\
  \frac{\partial \hat{\sigma}_d^2}{\partial \hat{w}_{i,d^*}} &= \ind{d=d^*} (x_i-\hat{\mu}_d)^2, \\
  \frac{\partial \hat{w}_{i,d}}{\partial w_{i^*,d^*}} &=
    \ind{d=d^*} \frac{\ind{i=i^*} \sum_{j=1}^{\con{b}}w_{j,d} - w_{i,d}}{(\sum_{j=1}^{\con{b}}w_{j,d})^2}.
\end{align}
Thus, the partial derivative of $L$ w.r.t. $w_{i,d}$ is:
\begin{equation}
  \frac{\partial L}{\partial w_{i^*,d}} = \hat{x}_{i^*,d} \left(
    \frac{\partial L}{\partial y_{i^*}} - A_d
  \right) - \frac{1}{2} \left(
    \hat{x}_{i^*,d}^2 - \frac{\sigma_d^2}{\sigma_d^2 + \epsilon}
  \right) B_d,
\end{equation}
where $A_d$ and $B_d$ are defined as in \eqref{eqn:a-b}.

\subsection{Training loss progress}
\begin{figure}[t]
 \centering
 \subfloat[MNIST-m as target]
  {\includegraphics[width=0.48\textwidth]{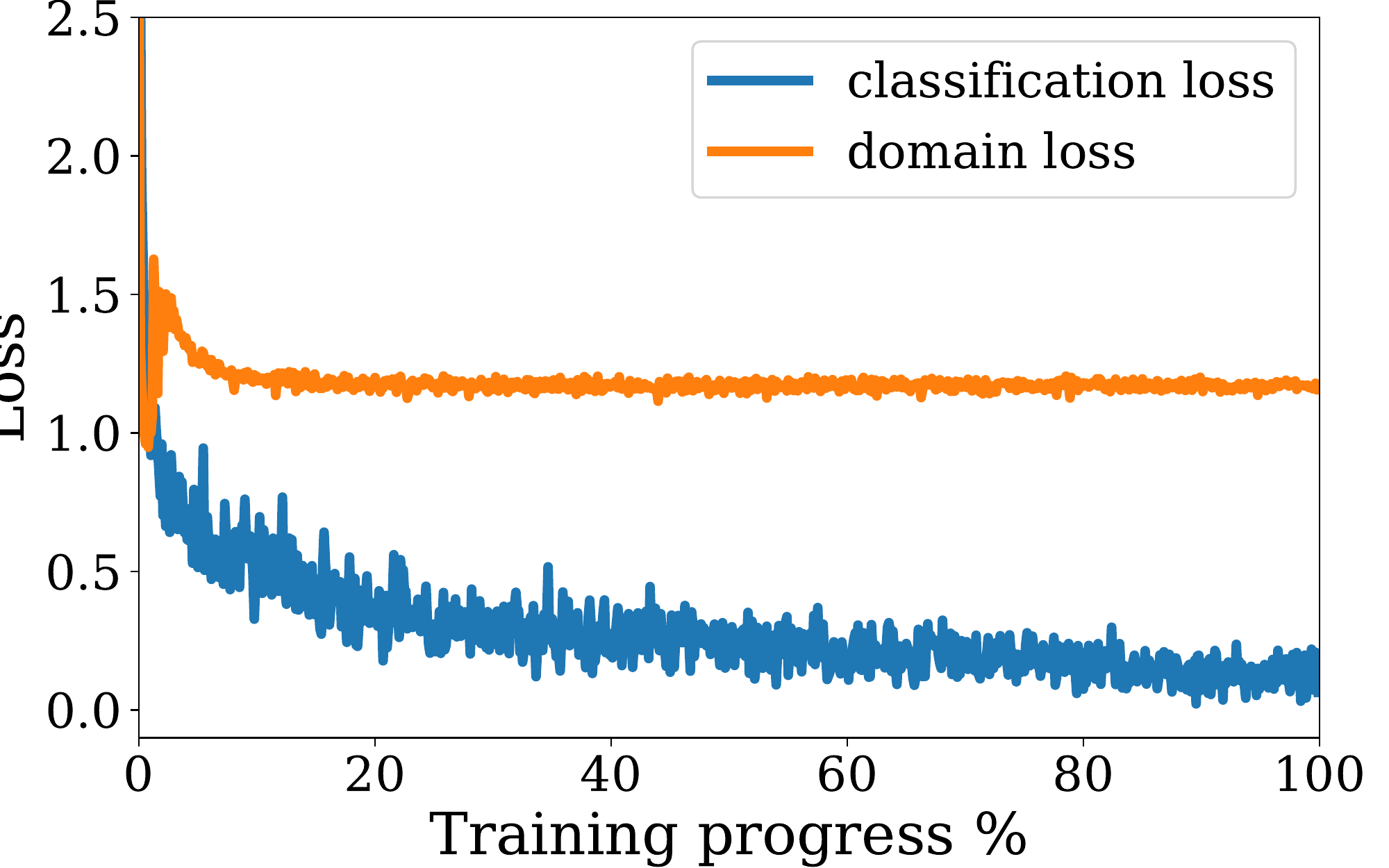}}
 \subfloat[SVHN as target]
  {\includegraphics[width=0.48\textwidth]{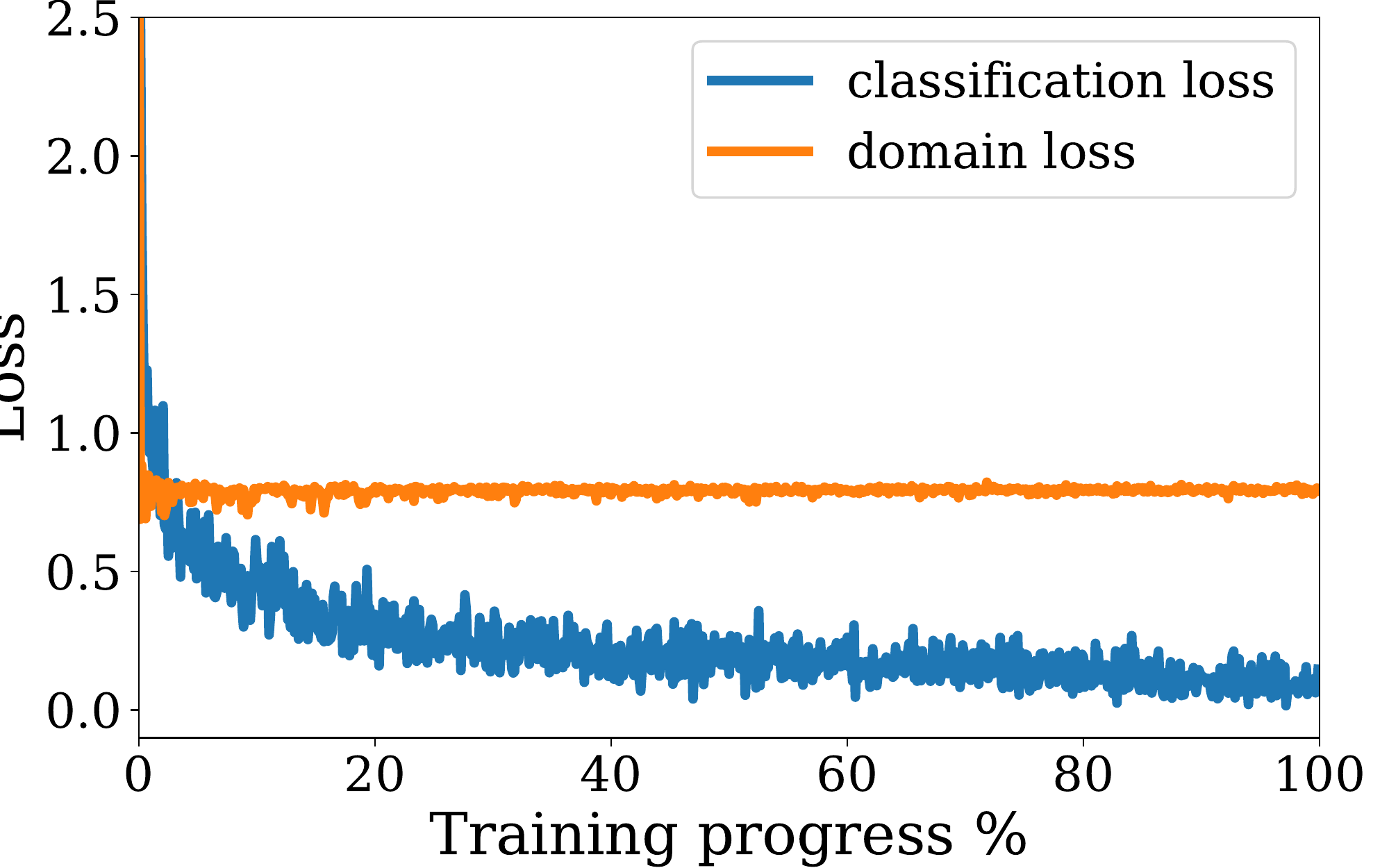}}
  \caption{{Digits-five: plots of the domain (orange) and classification (blue) losses during the training phase.}
  }
  \label{fig:losses-unified}
\end{figure}

\begin{figure}[t]
 \centering
 \subfloat[MNIST-m as target]
  {\includegraphics[width=0.48\textwidth]{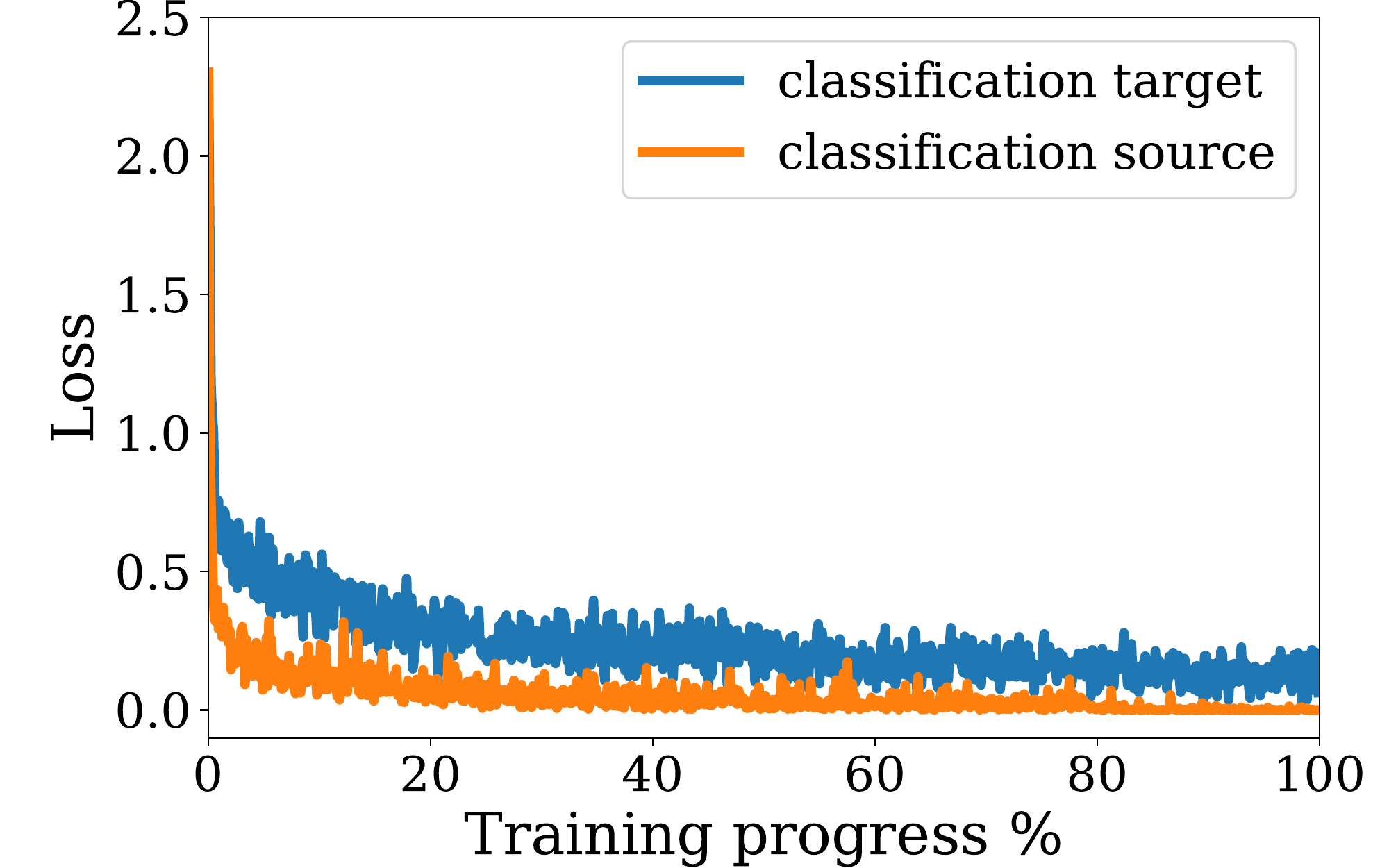}}
 \subfloat[SVHN as target]
  {\includegraphics[width=0.48\textwidth]{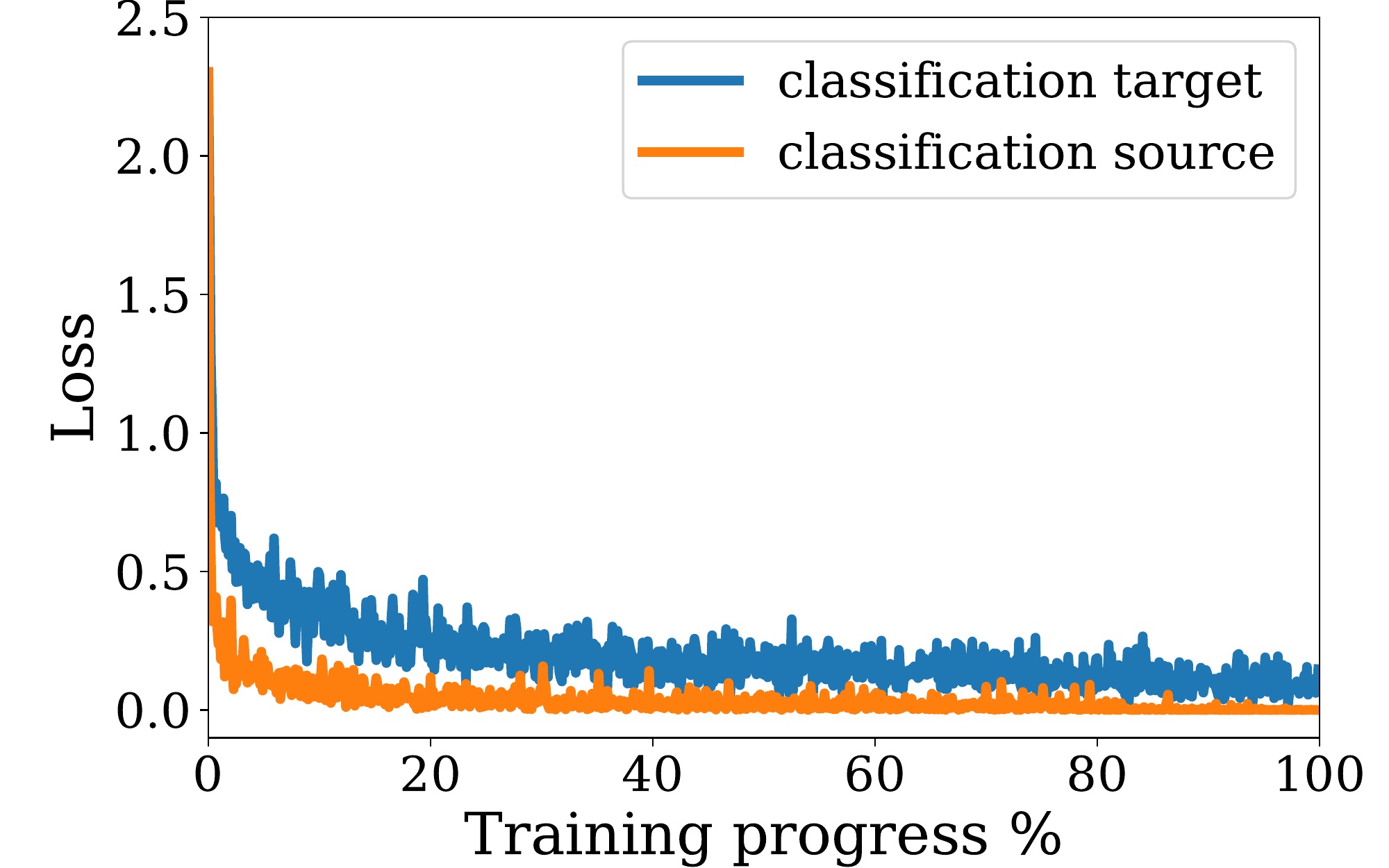}}
  \caption{{Digits-five: plots of the cross-entropy loss on source samples (orange) and entropy loss on target sample (blue) for the semantic classifier during the training phase.}
  }
  \label{fig:losses-semantic}
\end{figure}

\begin{figure}[t]
 \centering
 \subfloat[MNIST-m as target]
  {\includegraphics[width=0.48\textwidth]{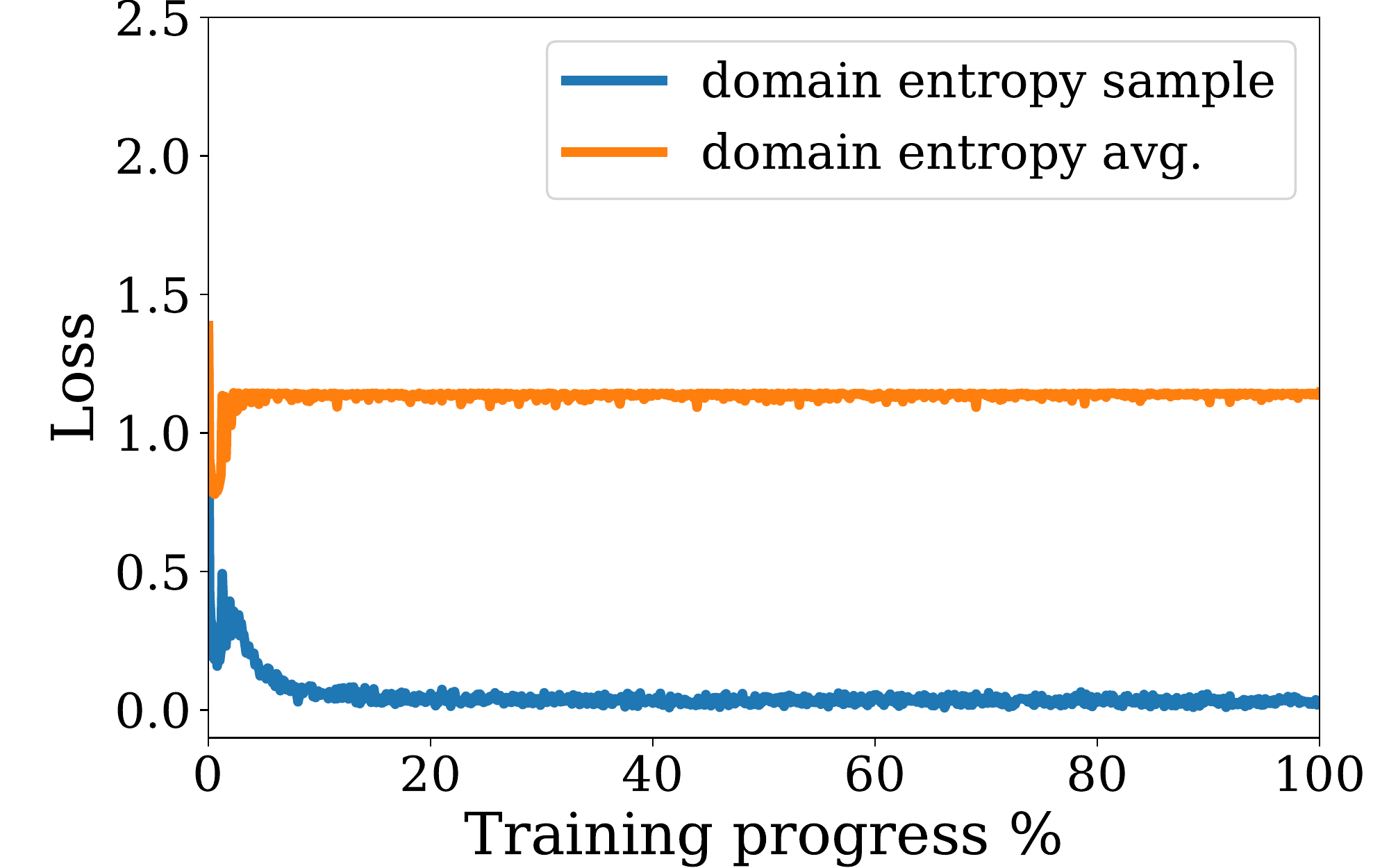}}
 \subfloat[SVHN as target]
  {\includegraphics[width=0.48\textwidth]{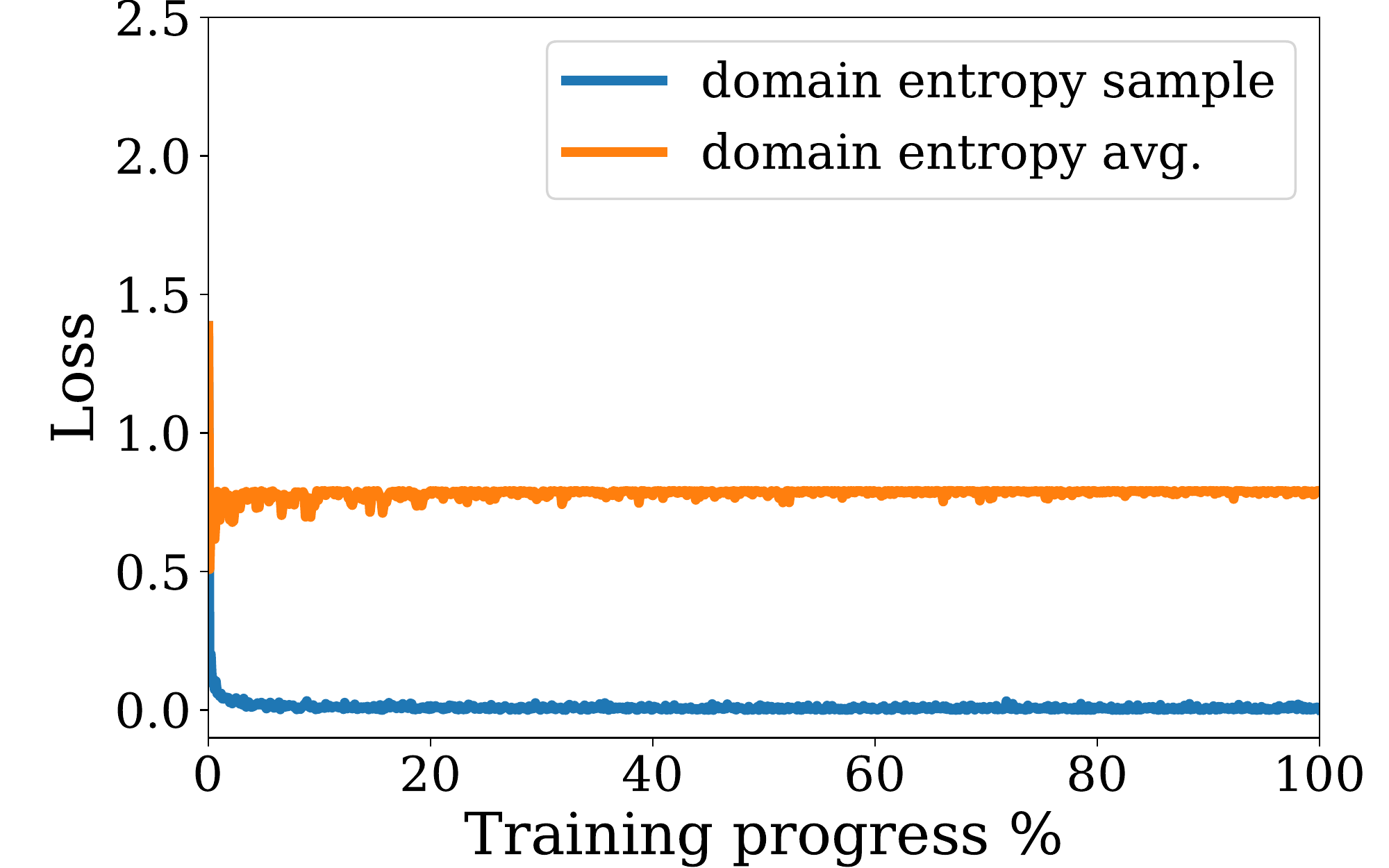}}
  \caption{{Digits-five: plots of the entropy loss on single sample (blue) and on the average batch assignments (orange) for the domain classifier during the training phase.}
  }
  \label{fig:losses-domain}
\end{figure}

{In this section, we plot the losses as the training progresses for the Digits-five experiments. The plots are shown in Figure \ref{fig:losses-unified}. For both MNIST-m and SVHN, the classification loss smoothly decreases, while the domain loss first decreases and then stabilizes around a fixed value. This is a consequence of the introduced balancing term on the domain assignments, which enforces the entropy to be low for the assignment of a single sample, but high for the assignments averaged across the entire batch. In Figures \ref{fig:losses-semantic} and \ref{fig:losses-domain} we plot the single components of the classification and domain loss respectively. For the semantic part (Figure \ref{fig:losses-semantic}), both the entropy loss on target sample and the cross-entropy loss on source samples decrease smoothly. For the domain assignment part (Figure \ref{fig:losses-domain}), we can see how the entropy loss on single samples rapidly decreases, while the average batch assignment keeps an high entropy, as expected. We highlight that when SVHN is used as target, the source domains are a bit closer to each other in appearance, thus the average batch entropy has a slightly lower value (\ie the assignments are less balanced) with respect to the MNIST-m as target case.

Finally, it is worth noticing that the domain loss reaches a stable value earlier than the classification components. This is a design choice, since we want to learn a semantic predictor on stable and confident domain assignments.}

\subsection{Additional Results on PACS}
{
A crucial problem in domain adaptation rarely addressed in the literature is how to tune model hyper-parameters. In fact, setting the hyper-parameters values based on the performance on the source domain is sub-optimal, due to the domain shift. Furthermore, assuming the presence of a validation set for the target domain is not realistic in practice \cite{morerio2017minimal}: in unsupervised domain adaptation we only assume the presence of a set of unlabelled target data. Despite recent research in this direction \cite{morerio2017minimal}, there is no clear solution to this problem in the literature. This problem is more severe in our case, since it is not trivial to define a validation set for the latent domain discovery problem, due to the assumption that multiple source and target domains are mixed.

\begin{table}[t]
			\caption{PACS dataset: comparison of different methods using the ResNet architecture. The first row indicates the target domain, while all the others are considered as sources. {The numbers in parenthesis indicate the results using a target validation set for model selection.}} 
		\centering
		\scalebox{0.8}{
		\begin{tabular}{ l | c  c  c  c | c  } 
			\hline
			Method & Sketch & Photo & Art & Cartoon & Mean \\
            	\hline
                ResNet \cite{he2016deep} &60.1&92.9&74.7&72.4&75.0\\
DIAL \cite{carlucci2017just} &66.8 (71.3)&\textbf{97.0} (\textbf{97.4})&87.3 (87.5)&85.5 (87.0)&84.2 (85.8)\\
mDA   &\textbf{70.7} (\textbf{75.2})&\textbf{97.0} (97.3)&87.4 (\textbf{87.7})&86.3 (\textbf{87.2})&\textbf{85.4} (\textbf{86.9})\\\hline\hline
Multi-source DA & 71.6 (78.1)& 96.6 (97.2) & 87.5 (88.7) & 87.0 (87.4) & 85.7 (87.9) \\ \hline 
\hline
		\end{tabular}
        }
		\label{tab:pacs-val}
\end{table}

\begin{table*}[ht]
			\caption{PACS dataset: comparison of different methods using the ResNet architecture on the multi-source multi-target setting. The first row indicates the two target domains. {The numbers in parenthesis indicate the results using a target validation set for model selection.} }
		\centering
		\scalebox{.65}{
		\begin{tabular}{ l | c  c c  c c c | c } 
			\hline
			Method  & Photo-Art & Photo-Cartoon & Photo-Sketch&Art-Cartoon&Art-Sketch&Cartoon-Sketch & Mean\\
            	\hline
                ResNet \cite{he2016deep} &71.4&84.2&81.4&62.2&70.3&54.2&70.6\\
DIAL \cite{carlucci2017just} &{86.7} (87.5)&86.5 (87.1)&86.8 (88.2)&77.1 (78.7)&72.1 (74.2)&67.7 (70.4)&79.5 (81.0)\\


mDA &\textbf{87.2} (\textbf{87.7}) &\textbf{88.1} (\textbf{88.5})&\textbf{88.7} (\textbf{89.7})&77.7 (\textbf{79.6})&\textbf{81.3} (\textbf{82.2})&\textbf{77.0} (\textbf{79.3})&\textbf{83.3} (\textbf{84.5})\\\hline\hline
Multi-source/& \multirow{2}{*}{87.7 (88.8)}&\multirow{2}{*}{88.9 (89.8)}&\multirow{2}{*}{86.8 (88.3)}&\multirow{2}{*}{79.0 (79.5)}&\multirow{2}{*}{79.8 (82.2)}&\multirow{2}{*}{75.6 (79.1)}&\multirow{2}{*}{83.0 (84.6)}\\
target DA &&&&&&\\ \hline 
		\end{tabular}
        }
		\label{tab:pacs-multitarget-val}
\end{table*}

Nonetheless, for the sake of completeness, we analyze the performances of our model and the baselines if we assume the presence of a target validation set  to perform model selection. We consider the PACS dataset, in both the single and multi-target scenarios. The results are reported in parenthesis in Table \ref{tab:pacs-val} and in Table \ref{tab:pacs-multitarget-val}. While both our model and the baselines obviously benefit from the validation set, the overall trends remain the same, with our model achieving higher performances with respect to the baseline and close to the multi-source upper bound. Notice that a validation set is especially beneficial in the case of consistent domain shift: for instance, all the methods increase their results by almost 5\% in Table \ref{tab:pacs-val} when Sketch is the target domain.

As a final note, we underline that the use of a validation set on the target domain for unsupervised domain adaptation is not a common practice in the community, thus these results can be regarded as an upper bound with respect to our model. 
}

\section{Predictive Domain Adaptation}

\subsection{Metadata Details}
\myparagraph{CompCars.} For the experiments with the CompCars dataset \cite{yang2015large}, we have two domain information: the car production year and the viewpoint. We encode the metadata through a 2-dimensional integer vector where the first integer encodes the year of production (between 2009 and 2014) and the second the viewpoint. While encoding the production year is straightforward, for the viewpoint we use the same criterion adopted in \cite{yang2016multivariate}, \ie we encode the viewpoint through integers between 1-5 in the order: \textit{Front}, \textit{Front-Side}, \textit{Side}, \textit{Rear-Side}, \textit{Rear}.

\myparagraph{Portraits.} For the experiments with the Portraits dataset \cite{ginosar2015century}, we have again two domain information: the year and the region where the picture has been taken. To allow for a bit more precise geographical information we encode the metadata through a 3-dimensional integer vector.

As for the CompCars dataset, the first integer encodes the decade of the image (8 decades between 1934 and 2014), while the second and third the geographical position. For the geographical position we simplify the representation through a coarse encoding involving 2 directions: est-west (from 0 to 1) and north-south (from 0 to 3). In particular we assign the following value pairs ([north-south, east-west]): \textit{Mid-Atlantic} $\rightarrow[0,1]$, \textit{Midwestern} $\rightarrow[0,2]$, \textit{New England} $\rightarrow[0,0]$, \textit{Pacific} $\rightarrow[0,3]$ and \textit{Southern} $\rightarrow[1,1]$. Each component of the vector has been normalized in the range 0-1.

\subsection{Additional Analysis}
\subsubsection{ResNet-18 on CompCars}
Here we apply \textit{AdaGraph} to the ResNet-18 architecture in the CompCars dataset \cite{yang2015large}. As for the other experiments, we apply \textit{AdaGraph} by replacing each BN layer of the network with its GBN counterpart. 

The network is initialized with the weights of the model pretrained on ImageNet. We train the network for 6 epochs on the source dataset, employing Adam as optimizer with a weight decay of $10^{-6}$ and a batch-size of 16. The learning rate is set to $10^{-3}$ for the classifier and $10^{-4}$ for the rest of the network and it is decayed by a factor of 10 after 4 epochs. We extract domain-specific parameters by training the network for 1 epoch on the union of source and auxiliary domains, keeping the same optimizer and hyperparameters. The batch size is kept to 16, building each batch with elements of a single pair
production year-viewpoint belonging to one of the domains available during training (either auxiliary or source). 

{The results are shown in Table \ref{tab:compcars-resnet}. As the table shows, \textit{AdaGraph} largely increases the performance of the \textit{Baseline} model}. Coherently with previous experiments, our refinement strategy is able to further increase the performances of \textit{AdaGraph}, filling almost entirely the gap with the DA upper bound.

\begin{table}[t]
			\caption{CompCars dataset \cite{yang2015large}. Results with ResNet-18 architecture.} 
		\centering
		\scalebox{1.}{
		\begin{tabular}{ l  | c } 
			\hline
			Method & Avg. Accuracy\\\hline
           Baseline& 56.8\\
           AdaGraph & \textbf{65.1} \\
           \hline
           Baseline + Refinement & 65.3 \\
           AdaGraph + Refinement & \textbf{66.7} \\
        \hline\hline
        DA upper bound & 66.9\\	\hline
        \end{tabular}
        }
        \label{tab:compcars-resnet}
\end{table}

\subsubsection{Performances vs Number of Auxiliary Domains}



In this section, we analyze the impact of varying the number of available auxiliary domains on the performances of our model. We employ the ResNet-18 architecture on the Portraits dataset, with the same setting and set of hyperparameters described in the experimental section. However, differently from the previous experiments, we vary the number of available auxiliary domains, from 1 to 38. 
We repeat the experiments 20 times, randomly sampling the available auxiliary domains each time.

\begin{figure*}[t!]
  \centering
    \subfloat[From 1954 Mid-Atlantic to 1984 Pacific.]{
    \includegraphics[width=0.7\textwidth]{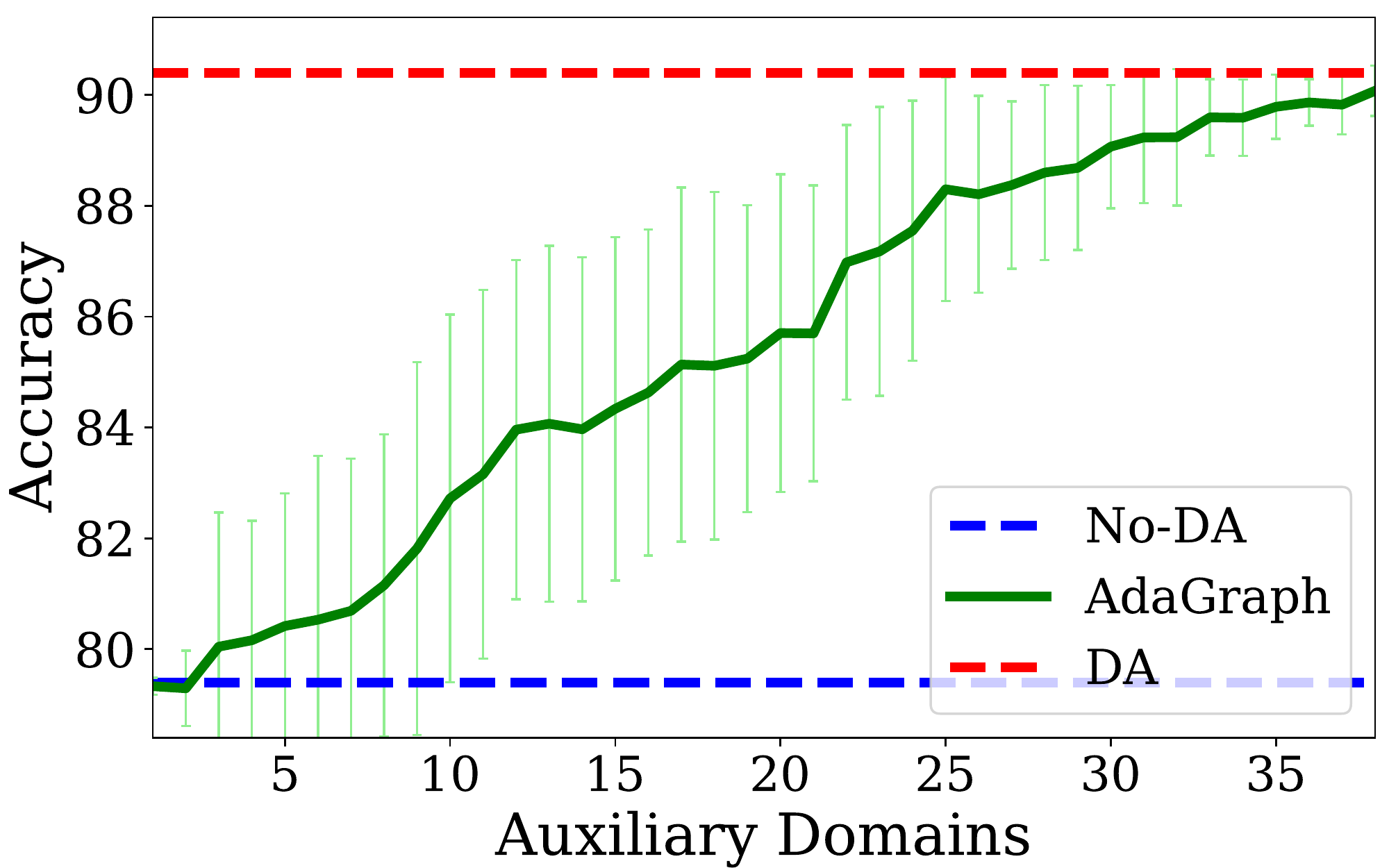}
    \label{fig-a-faces}}\\
  \subfloat[From 2004 Midwestern to 1944 Southern.]{
    \includegraphics[width=0.7\textwidth]{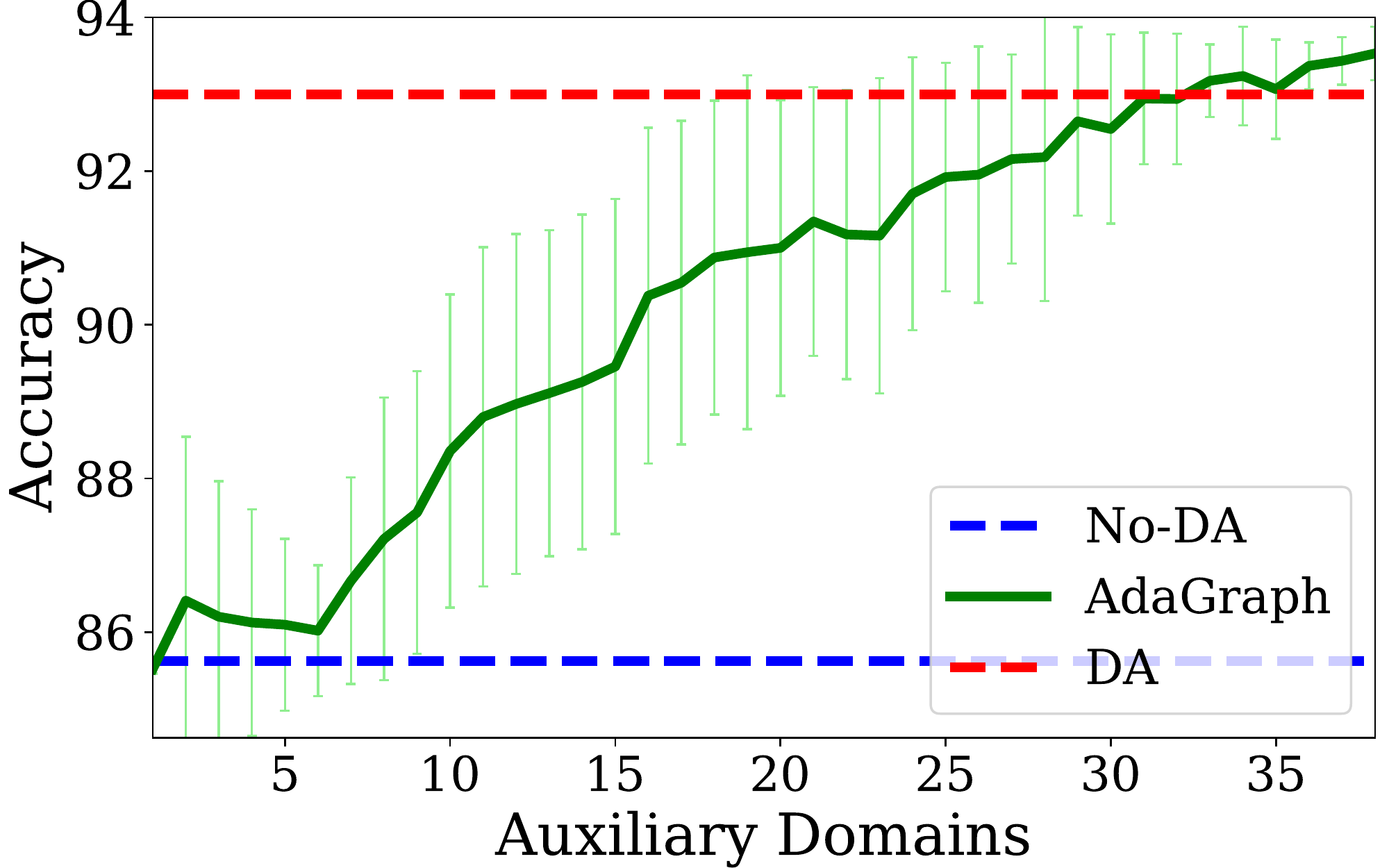}
    \label{fig-b-faces}}
  \\
    \subfloat[From 1974 Mid-Atlantic to 1994 New England.]{
    \includegraphics[width=0.7\textwidth]{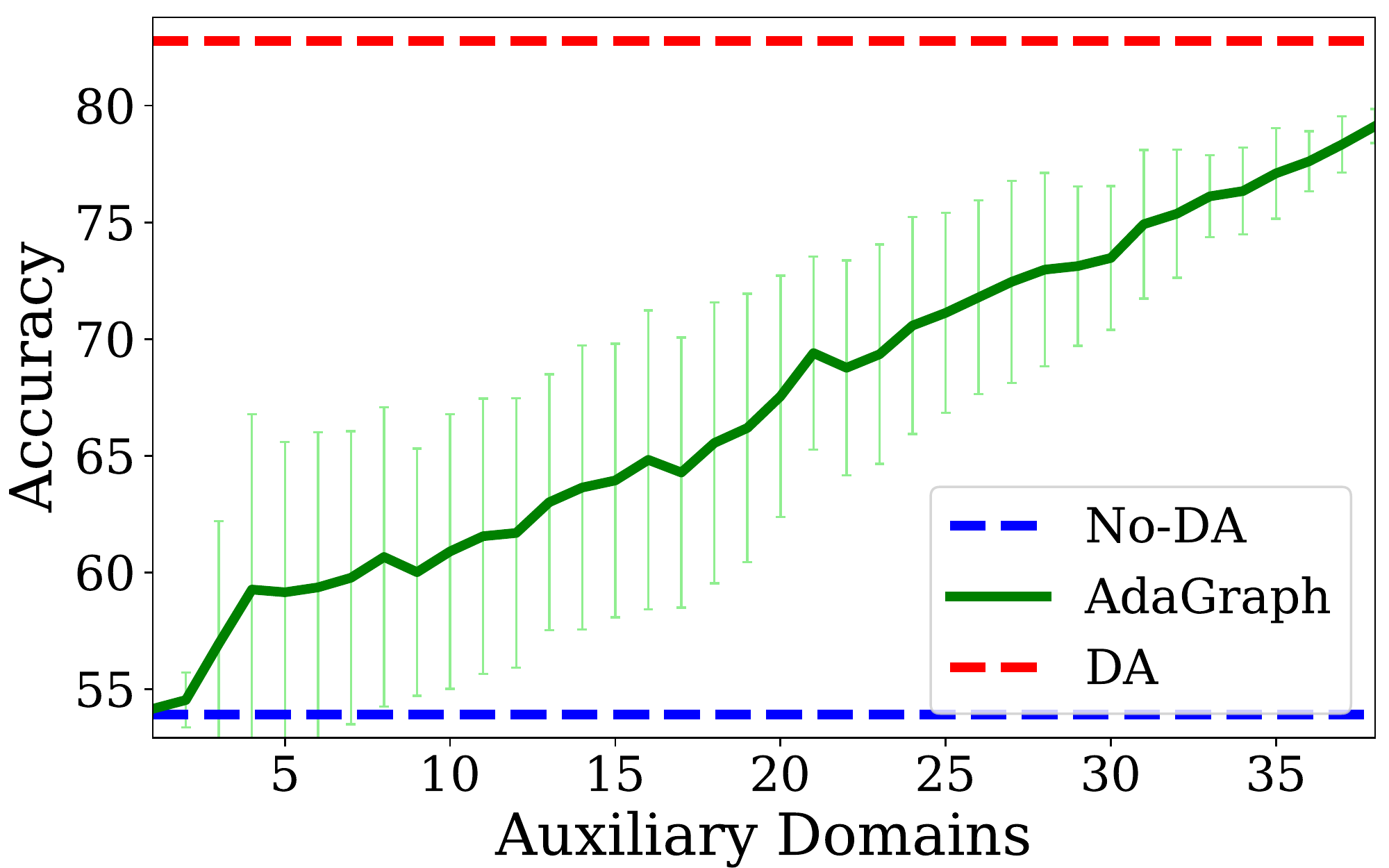}
    \label{fig-c-faces}}
  \caption{Portraits dataset: performances of \textit{AdaGraph} with respect to the number of auxiliary domains available for different source-target pairs. The years reported in the captions indicate the starting year of source and target decades.} 
  \label{fig:accuracy-vs-domains-portraits}
\end{figure*}

The results are shown in Figure \ref{fig:accuracy-vs-domains-portraits}. As expected, increasing the number of auxiliary domains leads to an increase in the performance of the model. In general, as we have more than 20 domains available, the performance of our model are close to the DA upper bound. While these results obviously depend on the relatedness between the auxiliary domains and the target, the plots show that having a large set of auxiliary domains may not be strictly necessary for achieving good performances.

\chapter{Recognizing New Semantic Concepts}
\section{Incremental Learning in Semantic Segmentation}

\subsection{How should we use the background?}
As highlighted in Section \ref{sec:IL-semantic-seg}, an important design choice for incremental learning in semantic segmentation is how to use the background. In particular, since the background class is present both in old and new classes, it can be considered either in the supervised cross-entropy loss, in the distillation component or in both. For our MiB method and all the baselines (LwF \cite{li2017learning}, Lwf-MC \cite{rebuffi2017icarl}, ILT \cite{michieli2019incremental}), we considered the latter case (\ie background in both). However, a natural question arises on how different choices for the background would impact the final results. In this section we investigate this point. 

We start from the LwF-MC \cite{rebuffi2017icarl} baseline, since it is composed of multiple binary classifiers and allows to easy decouple modifications on the background from the other classes. We then test two variants: 
 \begin{itemize}
    \item \textbf{LwF-MC-D} ignores the background in the classification loss, using as 
    target for the background the probability given by $f_{\theta^{t-1}}$.
    \item \textbf{LwF-MC-C} ignores the background in the distillation loss, using only the supervised signal from the ground-truth.
\end{itemize}
In Table \ref{tab:lwf-voc} and \ref{tab:lwf-ade} we report the results of the two variants for the overlapped scenarios of the Pascal VOC dataset and the \textit{50-50} scenario of ADE20K respectively. Together with the two variants, we report the results of our method (MiB), the offline training upper-bound (Joint) and the LwF-MC version employed in Section \ref{sec:ILSS-baselines} which uses the background in both binary cross-entropy and distillation, blending the two components with a hyper-parameter. 

As the tables show, the three variants of Lwf-MC exhibit different trade-offs among learning new knowledge and remembering the past one.
In particular, LwF-MC-C learns very well new classes, being always the most performing variant on the last incremental step. However, it suffers a significant drop in the old knowledge, showing its inability to tackle the catastrophic forgetting problem.

LwF-MC-D shows the opposite trend. It maintains very well the old knowledge, being the best variant in old classes for every setting. However, it is very intransigent \cite{chaudhry2018riemannian} \ie it is not able to correctly learn new classes, thus obtaining the worst performances on them. 

As expected, LwF-MC which considers the background in both cross-entropy and distillation achieves a trade-off among learning new knowledge, as in LwF-MC-C, while preserving the old one, as in LwF-MC-D. 

As the tables show, our MiB approach models the background more effectively, achieving the best trade-off among learning new knowledge and preserving old concepts. In particular, our method is the best by a margin in all scenarios for the new classes, while for old ones it is either better or comparable to the performance of the intransigent LwF-MC-D method. The only scenarios where it shows lower performances are the multi-step ones. Indeed in these scenarios, the multiple learning episodes make preserving old knowledge harder, and an intransigent method is less prone to forgetting since it is biased to old classes. However, the intransigence is not the right solution if the number of old and new classes are balanced, as in the \textit{50-50} scenario of ADE20k, since the overall performances will be damaged.




\begin{table}[t]
\small
\centering
\setlength{\tabcolsep}{2pt} 
\caption{Comparison of different implementations of LwF-MC on the Pascal-VOC 2012 \textit{overlapped} setup.}
\label{tab:lwf-voc}
\begin{tabular}{l||rr|r||rr|r||rr|r}
\multicolumn{1}{c}{}& \multicolumn{3}{c}{\textbf{19-1}}          & \multicolumn{3}{c}{\textbf{15-5}}          & \multicolumn{3}{c}{\textbf{15-1}} \\
\textbf{Method} & \it{1-19} & \it{20} & \it{all} & \it{1-15} & \it{16-20} & \it{all} & \it{1-15}  & \it{16-20}     & \it{all}     \\ \hline
LwF-MC-C & 44.6 & \underline{17.6} & 43.2 & 41.6 & \underline{42.2} & 41.8 & 4.4  & \underline{8.6}  & 5.4  \\
LwF-MC   & 64.4 & 13.3 & 61.9 & 58.1 & 35.0 & 52.3 & 6.4  & 8.4  & 6.9  \\
LwF-MC-D & \textbf{71.3} & 3.6  & \textbf{68.0} & \underline{73.7} & 21.0 & \underline{60.5} & \textbf{41.1} & 3.1  & \textbf{31.6} \\ 
MiB   & \underline{70.2}      & \textbf{22.1}      & \underline{67.8}      & \textbf{75.5}      & \textbf{49.4}      & \textbf{69.0}      & \underline{35.1}       & \textbf{13.5} & \underline{29.7}\\
\hline
Joint      & 77.4 & 78.0 & 77.4 & 79.1 & 72.6 & 77.4 & 79.1 & 72.6 & 77.4 

\end{tabular}
\end{table}


\begin{table}[h]
\small
\centering
\caption{Comparison of different implementations of LwF-MC on the \textit{50-50} setting of the ADE20K dataset.}
\label{tab:lwf-ade}
\begin{tabular}{l||rrr|r}
\textbf{Method}& \textit{1-50} & \textit{51-100} & \textit{101-150} & \textit{all}  \\ \hline
LwF-MC-C  & 8.0  & 7.2    & \underline{19.3}    & 11.5 \\
LwF-MC    & 27.8 & 7.0    & 10.4    & 15.1 \\
LwF-MC-D  & \textbf{39.1} & \underline{10.9}   & 6.7     & \underline{18.7} \\ 
MiB & \underline{35.5} & \textbf{22.2}   & \textbf{23.6}    & \textbf{27.0} \\\hline
Joint     & 51.1 & 38.3   & 28.2    & 38.9
\end{tabular}
\end{table}


\begin{table*}[t]
\centering
\caption{Per Class Mean IoU on 19-1 setting of Pascal-VOC 2012. disjoint setup}
\resizebox{\textwidth}{!}{
\label{tab:voc-19-1-d}
\begin{tabular}{l|rrrrrrrrrrrrrrrrrrr|r||r|r}
\textbf{Method} & aero & bike & bird & boat & bottle & bus  & car  & cat  & chair & cow  & table & dog  & horse & mbike & persn & plant & sheep & sofa & train & tv   & \textbf{1-19} & \textbf{all} \\ \hline
FT              & 11.9 & 2.1  & 1.1  & 11.6 & 4.8    & 6.9  & 13.5 & 0.2  & 0.0   & 3.8  & 14.4  & 0.5  & 1.5   & 4.7   & 0.0   & 15.8  & 2.8   & 1.8  & 13.5  & 12.3 & 5.8  & 6.2          \\
PI~\cite{zenke2017continual}              & 22.3 & 1.9  & 3.4  & 4.9  & 2.1    & 10.6 & 8.5  & 0.1  & 0.1   & 3.1  & 12.8  & 0.2  & 3.8   & 4.6   & 0.0   & 10.0  & 5.0   & 1.1  & 8.5   & 14.1 & 5.4  & 5.9          \\
EWC~\cite{kirkpatrick2017overcoming}             & 50.7 & 7.7  & 21.0 & 24.1 & 21.8   & 35.8 & 43.9 & 11.6 & 2.0   & 27.0 & 21.1  & 23.0 & 18.7  & 19.4  & 1.5   & 27.8  & 41.5  & 5.6  & 37.4  & 16.0 & 23.2 & 22.9         \\
RW~\cite{chaudhry2018riemannian}             & 45.8 & 5.3  & 15.1 & 22.8 & 17.8   & 28.9 & 40.9 & 7.5  & 1.3   & 22.4 & 20.3  & 14.5 & 13.7  & 16.3  & 0.8   & 25.3  & 31.8  & 4.8  & 33.3  & 15.7 & 19.4 & 19.2         \\
LwF~\cite{li2017learning}            & 28.1 & 40.5 & 53.1 & 38.8 & 47.4   & 46.4 & 63.6 & 83.5 & 35.8  & 60.1 & 48.8  & 76.5 & 65.3  & 67.1  & 83.2  & 50.2  & 61.2  & 42.5 & 14.2  & 9.1  & 53.0 & 50.8         \\
LwF-MC~\cite{rebuffi2017icarl}         & {79.4} & \textbf{41.3} & 75.6 & 47.9 & 51.0   & 69.6 & 75.4 & 78.5 & 35.1  & 66.6 & 49.0  & 72.7 & 73.8  & 71.6  & \textbf{84.9}  & \textbf{57.5}  & 67.7  & 42.7 & 56.8  & 13.2 & 63.0 & 60.5         \\
ILT~\cite{michieli2019incremental}            & \textbf{83.7} & 40.8 & 80.8 & \textbf{59.1} & 58.4   & 77.6 & \textbf{82.4} & 82.3 & 38.9  & 81.7 & 50.8  & 84.8 & \textbf{86.6}  & \textbf{81.0}  & 83.3  & 56.4  & 82.2  & 43.8 & 57.5  & 16.4 & 69.1 & 66.4         \\
MiB             & 78.0 & 40.5 & \textbf{85.7} & 51.6 & \textbf{64.4}   & \textbf{79.1} & {77.8} & \textbf{89.9} & \textbf{39.2} & \textbf{82.3} & \textbf{55.4}  & \textbf{86.2} & 82.7  & 72.2  & 83.6  & 56.6  & \textbf{86.2} & \textbf{45.1} & \textbf{65.0}  & \textbf{25.6} & \textbf{69.6} & \textbf{67.4}         \\ \hline
Joint           & 90.2 & 42.2 & 89.5 & 69.1 & 82.3   & 92.5 & 90.0 & 94.2 & 39.2  & 87.6 & 56.4  & 91.2 & 86.8  & 88.0  & 86.8  & 62.3  & 88.4  & 49.5 & 85.0  & 78.0 & 77.4 & 77.4        
\end{tabular}}
\end{table*}

\begin{table*}[t]
\centering
\caption{Per Class Mean IoU on 19-1 setting of Pascal-VOC 2012. overlapped setup}
\resizebox{\textwidth}{!}{
\label{tab:voc-19-1-o}
\begin{tabular}{l|rrrrrrrrrrrrrrrrrrr|r||r|r}
\textbf{Method} & aero & bike & bird & boat & bottle & bus  & car  & cat  & chair & cow  & table & dog  & horse & mbike & persn & plant & sheep & sofa & train & tv   & \textbf{1-19} & \textbf{all} \\ \hline
FT     & 23.7 & 1.9  & 1.5  & 9.3  & 6.9    & 16.9 & 8.5  & 0.0  & 0.0   & 9.5  & 5.3      & 0.1  & 2.9   & 8.8   & 0.0    & 15.1  & 1.0   & 0.7  & 16.0  & 12.9 & 6.8  & 7.1  \\
PI~\cite{zenke2017continual}     & 33.1 & 4.1  & 3.6  & 10.5 & 8.4    & 14.7 & 13.3 & 0.0  & 0.1   & 2.4  & 4.7      & 0.1  & 3.3   & 7.9   & 0.0    & 14.7  & 0.8   & 2.7  & 17.8  & 14.0 & 7.5  & 7.8  \\
EWC~\cite{kirkpatrick2017overcoming}   & 60.7 & 14.8 & 21.2 & 33.8 & 36.9   & 54.4 & 45.6 & 2.6  & 1.4   & 33.0 & 13.3     & 19.1 & 23.8  & 39.2  & 2.2    & 34.6  & 21.8  & 6.4  & 47.1  & 14.0 & 26.9 & 26.3 \\
RW~\cite{chaudhry2018riemannian}     & 57.5 & 12.1 & 15.4 & 29.6 & 32.9   & 50.7 & 40.0 & 1.3  & 0.8   & 30.7 & 10.7     & 12.6 & 18.6  & 32.9  & 0.8    & 30.7  & 17.5  & 5.5  & 42.7  & 14.2 & 23.3 & 22.9 \\
LwF~\cite{li2017learning}    & 36.6 & 35.1 & 62.0 & 32.9 & 47.5   & 31.6 & 51.5 & 77.9 & 36.5  & 67.7 & 44.3     & 71.4 & 68.6  & 66.2  & 82.2   & 49.6  & 58.7  & 41.1 & 11.9  & 8.5  & 51.2 & 49.1 \\
LwF-MC~\cite{rebuffi2017icarl} & 67.2 & 37.9 & 77.8 & 40.6 & 57.0   & 54.5 & 77.4 & 88.4 & 37.2  & 76.8 & 49.1     & 83.4 & 82.3  & 71.0  & \textbf{85.2}   & 55.6  & 81.9  & \textbf{46.0} & 54.9  & 13.3 & 64.4 & 61.9 \\
ILT~\cite{michieli2019incremental}    & \textbf{87.2} & \textbf{39.0} & 80.6 & \textbf{53.5} & 57.0   & 80.3 & 76.0 & 74.3 & 37.6  & 81.1 & 44.6     & 83.1 & \textbf{84.4}  & 81.6  & 82.4   & 54.5  & 82.7  & 38.9 & 56.1  & 12.3 & 67.1 & 64.4 \\
MiB    & 78.1 & 36.2 & \textbf{86.8} & 49.4 & \textbf{72.7}   & \textbf{80.8} & \textbf{78.2} & \textbf{90.8} & \textbf{38.3}  & \textbf{82.0} & \textbf{51.9}     & \textbf{86.7} & 82.8  & \textbf{76.9}  & 83.8   & \textbf{58.8}  & \textbf{84.4}  & 45.7 & \textbf{68.5}  & \textbf{22.1} & \textbf{70.2} & \textbf{67.8} \\ \hline
Joint  & 90.2 & 42.2 & 89.5 & 69.1 & 82.3   & 92.5 & 90.0 & 94.2 & 39.2  & 87.6 & 56.4     & 91.2 & 86.8  & 88.0  & 86.8   & 62.3  & 88.4  & 49.5 & 85.0  & 78.0 & 77.4 & 77.4
\end{tabular}}
\end{table*}

\begin{table*}[t]
\centering
\caption{Per Class Mean IoU on 15-5 setting of Pascal-VOC 2012. disjoint setup}
\resizebox{\textwidth}{!}{
\label{tab:voc-15-5-d}
\begin{tabular}{l|rrrrrrrrrrrrrrr|rrrrr||r|r|r}
\textbf{Method} & aero & bike & bird & boat & bottle & bus  & car  & cat  & chair & cow  & table & dog  & horse & mbike & persn & plant & sheep & sofa & train & tv   & \textbf{1-15} & \textbf{16-20} & \textbf{all} \\ \hline
FT     & 6.1  & 0.0  & 0.2  & 8.3  & 0.1    & 0.0  & 0.1  & 0.0  & 0.0   & 0.0  & 0.0   & 0.0  & 1.8   & 0.0   & 0.0   & 24.6  & 24.3  & 36.2 & 32.5  & 50.2 & 1.1  & 33.6  & 9.2  \\
PI~\cite{zenke2017continual}     & 8.8  & 0.0  & 0.2  & 10.5 & 0.0    & 0.0  & 0.1  & 0.0  & 0.0   & 0.0  & 0.0   & 0.0  & 0.4   & 0.0   & 0.0   & 25.6  & 24.7  & 34.3 & 34.1  & 52.0 & 1.3  & 34.1  & 9.5  \\
EWC~\cite{kirkpatrick2017overcoming}   & 58.8 & 4.1  & 56.4 & 46.2 & 44.4   & 4.3  & 67.4 & 3.6  & 2.3   & 14.8 & 10.3  & 12.4 & 51.6  & 20.4  & 2.9   & 28.8  & 32.2  & 35.6 & 35.5  & 56.3 & 26.7 & 37.7  & 29.4 \\
RW~\cite{chaudhry2018riemannian}    & 51.1 & 1.5  & 36.9 & 42.9 & 27.5   & 2.1  & 47.4 & 1.1  & 1.2   & 6.1  & 5.3   & 3.1  & 31.2  & 10.5  & 1.0   & 27.7  & 29.8  & 35.7 & 34.7  & \textbf{56.6} & 17.9 & 36.9  & 22.7 \\
LwF~\cite{li2017learning}   & 63.1 & 40.1 & 72.4 & 52.1 & 67.0   & 6.7  & 80.3 & 84.2 & 31.1  & 5.7  & 51.3  & 82.0 & 75.0  & 79.4  & 85.6  & 35.3  & 27.1  & 37.0 & 37.0  & 50.5 & 58.4 & 37.4  & 53.1 \\
LwF-MC~\cite{rebuffi2017icarl}& 78.1 & \textbf{42.3} & 78.9 & 62.1 & 78.6   & 47.3 & 84.6 & 89.1 & 35.0  & 26.2 & 50.5  & 86.6 & 77.6  & \textbf{84.9}  & \textbf{86.0}  & 35.0  & 35.2  & \textbf{40.8} & 49.2  & 45.9 & 67.2 & 41.2  & 60.7 \\
ILT~\cite{michieli2019incremental}   & 79.4 & 42.0 & 80.5 & 63.9 & \textbf{80.4}   & 12.8 & \textbf{86.0} & 90.2 & 30.7  & 6.7  & \textbf{53.3}  & 83.2 & 73.0  & 80.7  & 85.0  & \textbf{36.9}  & 29.9  & 36.8 & 38.3  & 55.7 & 63.2 & 39.5  & 57.3 \\
MiB    & \textbf{84.4} & 39.4 & \textbf{87.5} & \textbf{65.2} & 77.8   & \textbf{61.0} & \textbf{86.0} & \textbf{90.9 }& \textbf{35.3}  & \textbf{60.3} & 53.0  & \textbf{88.2} & \textbf{80.4}  & 82.4  & 85.3  & 28.7  & \textbf{46.0}  & 34.7 & \textbf{54.4}  & 52.7 & \textbf{71.8} & \textbf{43.3}  & \textbf{64.7} \\ \hline
Joint  & 90.2 & 42.2 & 89.5 & 69.1 & 82.3   & 92.5 & 90.0 & 94.2 & 39.2  & 87.6 & 56.4  & 91.2 & 86.8  & 88.0  & 86.8  & 62.3  & 88.4  & 49.5 & 85.0  & 78.0 & 79.1 & 72.6  & 77.4
\end{tabular}}
\end{table*}

\begin{table*}[t]
\centering
\caption{Per Class Mean IoU on 15-5 setting of Pascal-VOC 2012. overlapped setup}
\resizebox{\textwidth}{!}{
\label{tabvoc-:15-5-o}
\begin{tabular}{l|rrrrrrrrrrrrrrr|rrrrr||r|r|r}
\textbf{Method} & aero & bike & bird & boat & bottle & bus  & car  & cat  & chair & cow  & table & dog  & horse & mbike & persn & plant & sheep & sofa & train & tv   & \textbf{1-15} & \textbf{16-20} & \textbf{all} \\ \hline
FT     & 13.4 & 0.1  & 0.0  & 15.6 & 0.8    & 0.0  & 0.3  & 0.0  & 0.0   & 0.0  & 0.0   & 0.0  & 0.9   & 0.0   & 0.0   & 30.9  & 21.6  & 32.8 & 34.9  & 45.1 & 2.1  & 33.1  & 9.8  \\
PI~\cite{zenke2017continual}     & 7.8  & 0.0  & 0.0  & 12.9 & 0.3    & 0.0  & 0.3  & 0.0  & 0.0   & 0.0  & 0.0   & 0.0  & 2.7   & 0.0   & 0.0   & 33.2  & 22.2  & 33.2 & 36.1  & 42.0 & 1.6  & 33.3  & 9.5  \\
EWC~\cite{kirkpatrick2017overcoming}   & 67.3 & 12.8 & 50.5 & 52.9 & 35.0   & 24.7 & 41.7 & 1.2  & 1.0   & 9.8  & 5.7   & 3.7  & 42.9  & 15.4  & 0.6   & 31.8  & 26.3  & 32.1 & 42.0  & 45.0 & 24.3 & 35.5  & 27.1 \\
RW~\cite{chaudhry2018riemannian}    & 61.2 & 6.7  & 33.8 & 48.1 & 24.4   & 9.3  & 22.3 & 0.3  & 0.5   & 3.5  & 0.2   & 1.1  & 31.8  & 6.4   & 0.1   & 32.1  & 25.8  & 31.9 & 38.7  & 45.9 & 16.6 & 34.9  & 21.2 \\
LwF~\cite{li2017learning}   & 64.5 & 40.2 & 72.8 & 56.9 & 57.3   & 9.5  & 82.6 & 88.6 & 33.2  & 8.9  & 48.4  & 81.9 & 75.0  & 78.2  & 84.9  & 34.7  & 27.8  & 33.1 & 39.6  & 48.0 & 58.9 & 36.6  & 53.3 \\
LwF-MC~\cite{rebuffi2017icarl}& 60.6 & 38.9 & 74.7 & 41.6 & 67.2   & 10.8 & 81.4 & 88.8 & \textbf{38.7}  & 4.3  & 47.4  & 82.2 & 69.9  & 78.9  & \textbf{85.8}  & 28.4  & 28.5  & \textbf{34.1} & 36.4  & 47.8 & 58.1 & 35.0  & 52.3 \\
ILT~\cite{michieli2019incremental}   & 77.4 & \textbf{40.3} & 78.9 & 61.9 & 78.7   & 53.5 & 86.1 & 88.7 & 33.8  & 15.9 & 51.1  & 83.2 & 80.2  & 79.8  & 85.0  & \textbf{39.5}  & 30.9  & 31.0 & 49.3  & 52.6 & 66.3 & 40.6  & 59.9 \\
MiB    & \textbf{86.6} & 39.3 & \textbf{88.9} & \textbf{66.1} & \textbf{80.8}   & \textbf{86.6 }& \textbf{90.1} & \textbf{92.5} & 38.0  & \textbf{64.6} & \textbf{56.4}  & \textbf{89.6} & \textbf{80.5}  & \textbf{86.5}  & 85.7  & 30.2  & \textbf{52.9}  & 31.3 & \textbf{73.2}  & \textbf{59.5} & \textbf{75.5} & \textbf{49.4}  & \textbf{69.0} \\ \hline
Joint  & 90.2 & 42.2 & 89.5 & 69.1 & 82.3   & 92.5 & 90.0 & 94.2 & 39.2  & 87.6 & 56.4  & 91.2 & 86.8  & 88.0  & 86.8  & 62.3  & 88.4  & 49.5 & 85.0  & 78.0 & 79.1 & 72.6  & 77.4
\end{tabular}}
\end{table*}

\begin{table*}[t]
\centering
\caption{Per Class Mean IoU on 15-1 setting of Pascal-VOC 2012. disjoint setup}
\resizebox{\textwidth}{!}{
\label{tab:voc-15-1-d}
\begin{tabular}{l|rrrrrrrrrrrrrrr|r|r|r|r|r||r|r|r}
\textbf{Method} & aero & bike & bird & boat & bottle & bus  & car  & cat  & chair & cow  & table & dog  & horse & mbike & persn & plant & sheep & sofa & train & tv   & \textbf{1-15} & \textbf{16-20} & \textbf{all} \\ \hline
FT     & 0.3  & 0.0  & 0.0  & 2.5  & 0.0    & 0.0  & 0.0  & 0.0  & 0.0   & 0.0  & 0.0   & 0.0  & 0.0   & 0.0   & 0.0   & 0.0   & 0.0   & 0.0  & 0.0   & 8.8  & 0.2  & 1.8   & 0.6  \\
PI~\cite{zenke2017continual}     & 0.0  & 0.0  & 0.0  & 0.0  & 0.0    & 0.0  & 0.0  & 0.0  & 0.0   & 0.0  & 0.0   & 0.0  & 0.0   & 0.0   & 0.0   & 0.0   & 0.0   & 0.0  & 0.3   & 8.6  & 0.0  & 1.8   & 0.4  \\
EWC~\cite{kirkpatrick2017overcoming}   & 0.0  & 0.0  & 0.0  & 1.0  & 0.0    & 0.0  & 0.0  & 0.0  & 0.0   & 0.0  & 3.6   & 0.0  & 0.0   & 0.0   & 0.0   & 0.0   & 0.0   & 7.3  & 7.0   & 7.4  & 0.3  & 4.3   & 1.3  \\
RW~\cite{chaudhry2018riemannian}    & 0.0  & 0.0  & 0.0  & 0.2  & 0.0    & 0.0  & 0.0  & 0.0  & 0.0   & 0.0  & 2.2   & 0.0  & 0.0   & 0.0   & 0.0   & 0.0   & 0.0   & 8.1  & 10.5  & 8.2  & 0.2  & 5.4   & 1.5  \\
LwF~\cite{li2017learning}   & 0.0  & 0.0  & 0.0  & 0.0  & 0.6    & 0.0  & 0.0  & 0.0  & 0.0   & 0.0  & 0.0   & 0.0  & 0.0   & 0.0   & 10.7  & 0.0   & 0.0   & 1.9  & 8.2   & 7.9  & 0.8  & 3.6   & 1.5  \\
LwF-MC~\cite{rebuffi2017icarl}& 0.0  & 6.3  & 0.8  & 0.0  & 1.1    & 0.0  & 0.1  & 0.3  & 0.0   & 0.0  & 0.0   & 0.0  & 0.2   & 0.0   & 59.0  & 0.0   & 9.5   & 2.9  & 11.9  & \textbf{11.0} & 4.5  & 7.0   & 5.2  \\
ILT~\cite{michieli2019incremental}   & 3.7  & 0.0  & 2.9  & 0.0  & 12.8   & 0.0  & 0.0  & 0.1  & 0.0   & 0.0  & \textbf{21.2}  & 0.1  & 0.4   & 0.6   & 13.6  & 0.0   & 0.0   & 11.6 & 8.3   & 8.5  & 3.7  & 5.7   & 4.2  \\
MiB    & \textbf{53.6} & \textbf{38.9} & \textbf{53.6} & \textbf{17.7} & \textbf{62.7}   & \textbf{36.5} & \textbf{71.2} & \textbf{60.1} & \textbf{1.1}   & \textbf{35.2} & 8.1   & \textbf{57.6} & \textbf{55.0}  & \textbf{62.1}  & \textbf{79.4}  & \textbf{10.2}  & \textbf{14.2}  & \textbf{11.9} & \textbf{18.2} & 10.1 & \textbf{46.2} & \textbf{12.9}  & \textbf{37.9} \\ \hline
Joint  & 90.2 & 42.2 & 89.5 & 69.1 & 82.3   & 92.5 & 90.0 & 94.2 & 39.2  & 87.6 & 56.4  & 91.2 & 86.8  & 88.0  & 86.8  & 62.3  & 88.4  & 49.5 & 85.0  & 78.0 & 79.1 & 72.6  & 77.4
\end{tabular}}
\end{table*}

\begin{table*}[t]
\centering
\caption{Per Class Mean IoU on 15-1 setting of Pascal-VOC 2012. overlapped setup}
\resizebox{\textwidth}{!}{
\label{tab:voc-15-1-o}
\begin{tabular}{l|rrrrrrrrrrrrrrr|r|r|r|r|r||r|r|r}
\textbf{Method} & aero & bike & bird & boat & bottle & bus  & car  & cat  & chair & cow  & table & dog  & horse & mbike & persn & plant & sheep & sofa & train & tv   & \textbf{1-15} & \textbf{16-20} & \textbf{all} \\ \hline
FT     & 2.6  & 0.0  & 0.0  & 0.7  & 0.0    & 0.1  & 0.0  & 0.0  & 0.0   & 0.0  & 0.0   & 0.0  & 0.0   & 0.0   & 0.0   & 0.0   & 0.0   & 0.0  & 0.0   & 9.2  & 0.2  & 1.8   & 0.6  \\
PI~\cite{zenke2017continual}     & 0.0  & 0.0  & 0.0  & 0.0  & 0.0    & 0.0  & 0.0  & 0.0  & 0.0   & 0.0  & 0.0   & 0.0  & 0.0   & 0.0   & 0.0   & 0.0   & 0.0   & 0.0  & 0.2   & 9.1  & 0.0  & 1.8   & 0.5  \\
EWC~\cite{kirkpatrick2017overcoming}   & 0.0  & 0.0  & 0.0  & 1.0  & 0.0    & 0.0  & 0.0  & 0.0  & 0.0   & 0.0  & 3.6   & 0.0  & 0.0   & 0.0   & 0.0   & 0.0   & 0.0   & 7.3  & 7.0   & 7.4  & 0.3  & 4.3   & 1.3  \\
RW~\cite{chaudhry2018riemannian}    & 0.1  & 0.0  & 0.0  & 0.0  & 0.0    & 0.0  & 0.0  & 0.0  & 0.0   & 0.0  & 0.0   & 0.0  & 0.0   & 0.0   & 0.0   & 0.0   & 0.0   & 8.7  & {11.2}  & 6.3  & 0.0  & 5.2   & 1.3  \\
LwF~\cite{li2017learning}   & 3.7  & 0.1  & 0.0  & 2.5  & 0.2    & 0.0  & 0.0  & 0.0  & 0.0   & 0.0  & 0.0   & 0.0  & 0.1   & 0.0   & 9.0   & 0.0   & 0.0   & 1.6  & 8.9   & 8.8  & 1.0  & 3.9   & 1.8  \\
LwF-MC~\cite{rebuffi2017icarl}& 0.0  & 7.2  & 5.2  & 0.0  & 25.5   & 0.0  & 0.0  & 0.0  & 0.0   & 0.0  & 0.0   & 0.0  & 1.2   & 1.3   & 56.2  & 0.0   & 4.9   & 0.2  & 8.6   & \textbf{28.2} & 6.4  & 8.4   & 6.9  \\
ILT~\cite{michieli2019incremental}   & 20.0 & 0.0  & 3.2  & 6.3  & 2.3    & 0.0  & 0.0  & 0.0  & \textbf{0.3}   & 5.1  & \textbf{19.0}  & 0.0  & 9.1   & 0.0   & 8.7   & 0.0   & 0.0   & \textbf{21.0} & 9.9   & 8.1  & 4.9  & 7.8   & 5.7  \\
MiB    & \textbf{31.3} &\textbf{ 25.4} & \textbf{26.7} & \textbf{26.9} & \textbf{46.1}   & \textbf{31.0} & \textbf{63.6} & \textbf{52.8} & 0.1   & \textbf{11.0} & 9.4   & \textbf{52.4} & \textbf{41.2}  & \textbf{28.1}  & \textbf{80.7}  & \textbf{17.6}  & \textbf{13.1}  & 15.3 & \textbf{15.3}  & 6.2  & \textbf{35.1} & \textbf{13.5}  & \textbf{29.7} \\ \hline
Joint  & 90.2 & 42.2 & 89.5 & 69.1 & 82.3   & 92.5 & 90.0 & 94.2 & 39.2  & 87.6 & 56.4  & 91.2 & 86.8  & 88.0  & 86.8  & 62.3  & 88.4  & 49.5 & 85.0  & 78.0 & 79.1 & 72.6  & 77.4
\end{tabular}}
\end{table*}

\subsection{Per class results on Pascal-VOC 2012}

From Table \ref{tab:voc-19-1-d} to \ref{tab:voc-15-1-o}, we report the results for all classes of the Pascal-VOC 2012 dataset. As the tables show, MiB achieves the best results in the majority of classes (i.e. at least 14/20 in the 19-1 scenarios, 13/20 in the 15-5 and 16/20 in the 15-1 ones) being either the second best or comparable to the top two in all the others. Remarkable cases are the ones where we learn classes that are either similar in appearance (\eg bus and train) or appear in similar contexts (\eg sheep and cow): for those pairs, our model outperforms the competitors by a margin in both old classes (\ie bus and cow in the 15-5 and 15-1 scenarios) and new ones (\ie sheep and train). These results show the capability of MiB to not only learn new knowledge while preserving the old one, but also to learn discriminative features for difficult cases during different learning steps.

\FloatBarrier

\subsection{Validation protocol and \hypers}
In this work, we follow the protocol of \cite{de2019continual} for setting the \hypers in continual learning. The protocol works in three steps and does not require \textit{any} data of old tasks. First, we split the training set of the current learning step into train and validation sets. We use $80\%$ of the data for training and $20\%$ for validation. Note that the validation set contains only labels for the current learning step. 

Second, we set general \hypers values (\eg learning rate) as the ones achieving the highest accuracy in the new set of classes with the fine-tuned model. Since we tested multiple methods, we wanted to ensure fairness in terms of \hypers used, without producing biased results. To this extent, this step is held out only once starting from the fine-tuned model and fixing the \hypers for all the methods. In particular, we set the learning rate as $10^{-3}$ for the incremental steps in all datasets and settings.

As a final step, we set the \hypers specific of the continual learning method as the highest values (to ensure minimum forgetting) with a tolerated decay on the performance on the new classes with respect to the ones achieved by the fine-tuned model (to ensure maximum learning). We set the tolerated decay as $20\%$ of the original performances, exploring \hypers values of the form $A\cdot 10^{B}$, with $A\in\{1,5\}$ and $B\in\{-3,\dots,3\}$. We perform this validation procedure in the first learning step of each scenario, keeping the \hypers fixed for the subsequent ones. Since this procedure is costly, we perform it only for the Pascal-VOC dataset, keeping the \hypers for the large-scale ADE20k. As a result, for the prior focused methods, we obtain a weight of $500$ for EWC \cite{kirkpatrick2017overcoming} and PI \cite{zenke2017continual} and $100$ for RW \cite{chaudhry2018riemannian} in all scenarios. For the data-focused methods we obtain a weight of $100$ for the distillation loss of LwF \cite{li2017learning}, $10$ for the one in LwF-MC \cite{rebuffi2017icarl} and $100$ for both distillation losses in ILT \cite{michieli2019incremental}, in all settings. For our MiB method, we obtain a distillation loss weight of $10$ for all scenarios except for the \textit{15-1} in Pascal VOC, where the weight is set to $100$. 



\chapter{Towards Recognizing Unseen Categories in Unseen Domains}
\section{Recognizing Unseen Categories in Unseen Domains}
\subsection{Hyperparameter choices}

In this section, we will provide additional details on the hyperparameter choices and validation protocols, not included in Section \ref{both:dgzsl}.

\myparagraph{ZSL}. 
For each dataset, we use the train, validation and test split provided by \cite{xian2018zeroshotgood}. In all the settings we employ features extracted from the second-last layer of a ResNet-101 \cite{he2016deep} pretrained on ImageNet as image representation, \textit{without} end-to-end training. For \methodName, we consider $f$ as the identity function and as $g$ a simple fully connected layer, performing the mixing directly at the feature-level while applying our alignment loss in the embedding space (\ie $\mathcal{L}_{\text{M-IMG}}$ and $\mathcal{L}_{\text{M-F}}$ coincide in this case and are applied only once.) All hyperparameters have been set dataset-wise following \cite{xian2018zeroshotgood}, using the available validation sets. For all the experiments, we use SGD as optimizer with an initial learning rate equal to 0.1, momentum equal to 0.9, a weight-decay set to 0.001 for all settings but AWA, where is set 0. The learning-rate is downscaled by a factor of ten after 2/3 of the total number of epochs and $N=30$. In particular, for CUB and FLO we train our model for $90$ epochs, setting $\beta_{\text{max}}=0.8$ and $\eta_{\text{I}}=\eta_{\text{F}}=10.0$ for CUB, and $\beta_{\text{max}}=0.4$ and $\eta_{\text{I}}=\eta_{\text{F}}=4.0$ for FLO. For AWA, we train our network for $30$ epochs, with $\beta_{\text{max}}=0.2$ and $\eta_{\text{I}}=\eta_{\text{F}}=1.0$. For SUN, we train our network for $60$ epochs, with $\beta_{\text{max}}=0.8$ and $\eta_{\text{I}}=\eta_{\text{F}}=10$. In all settings, the batch-size is set to 128.

\myparagraph{DG.}  We use as base architecture a ResNet-18 \cite{he2016deep} pretrained on ImageNet. For our model, we consider $f$ to be the ResNet-18, $g$ to be the identity function and $\omega$ will be a learned, fully-connected classifier. We use the same training hyperparameters and protocol of \cite{li2019episodic}, setting $\beta_{\text{max}}=0.6$, $\eta_{\text{I}}=0.1$, $\eta_{\text{F}}=3$ and $N=10$.

\myparagraph{ZSL+DG.} For all the baselines and our method we employ as base architecture a ResNet-50 \cite{he2016deep} pretrained on ImageNet, using SGD with momentum as optimizer, with a learning rate of $0.001$ for the ZSL classifier and $0.0001$ for the ResNet-50 backbone, a weight decay of $5\cdot10^{-5}$ and momentum $0.9$. We train the models for 8 epochs (each epoch counted on the smallest source dataset), with a batch-size containing 24 sample per domain. We decrease the learning rates by a factor of $10$ after 6 epochs. For our model, we consider the backbone as $f$ and a simple fully-connected layer as $g$. We set $N=2$, $\eta_{\text{I}}=10^{-3}$ for all the experiments, while $\beta_{\text{max}}$ in $\{1,2\}$ and $\eta_{\text{F}}$ in $\{0.5,1,2\}$ depending on the scenario.

\subsection{ZSL+DG: analysis of additional baselines}
 In Table \ref{tab:DGZSL-domainnet}, we showed the performance of our method in the new ZSL+DG scenario on the DomainNet dataset \cite{peng2019moment}, comparing it with three baselines: SPNet \cite{xian2019semantic}, simple \textit{mixup} \cite{zhang2017mixup} coupled with SPNet and SPNet coupled with EpiFCR \cite{li2019episodic}, an episodic-based method for DG. We reported the results of these baselines to show 1) the performance of a state-of-the-art ZSL method (SPNet), 2) the impact of \textit{mixup} alone (\textit{mixup}+SPNet) and 3) the results obtained by coupling state-of-the-art models for DG and for ZSL together (EpiFCR+SPNet). We chose SPNet and EpiFCR as state-of-the-art references for ZSL and DG respectively because they are very recent approaches achieving high performances on their respective scenarios. 

In this section, we motivate our choices by showing that other baselines of ZSL and DG achieve lower performances in this new scenario. In particular we show the performances of two standard ZSL methods, ALE \cite{akata2013label} and DEVISE \cite{frome2013devise} and a standard DG/DA method, DANN \cite{ganin2016domain}. We choose DANN since it is a strong baseline for DG on residual architectures, as shown in \cite{li2019episodic}. As in Section \ref{both:dgzsl}, we show the performances of the ZSL methods alone, ZSL methods coupled with DANN, and with EpiFCR. For all methods, we keep the same training hyperparameters, tuning only the method-specific ones. The results are reported in Table \ref{tab:domainnet-additional}. As the table shows, \methodName achieves superior performances even compared to these additional baselines. Moreover, these baselines achieve lower results than the EpiFCR method coupled with SPNet, as expected. It is also worth highlighting how coupling ZSL methods with DANN for DG achieves lower performances than the ZSL methods alone in this scenario. This is in line with the results reported in \cite{peng2019moment}, where standard domain alignment-based methods are shown to be not effective in the DomainNet dataset, leading also to negative transfer in some cases \cite{peng2019moment}.

\begin{table*}[t]
			\caption{ZSL+DG scenario on the DomainNet dataset with ResNet-50 as backbone.} 
		\centering\scalebox{0.8}{
		\begin{tabular}{ >{\centering}p{6em} | >{\centering}p{6em} | >{\centering}p{4em} >{\centering}p{4em} >{\centering}p{4em} >{\centering}p{4.em} >{\centering}p{4.em}| >{\centering\arraybackslash}p{4em} }
		\multicolumn{2}{c |}{Method} & \multicolumn{5}{c |}{Target Domain} &\\
		DG & ZSL &clipart&infograph&painting&quickdraw&sketch&avg.\\
		\hline
                      \multirow{3}{*}{-}&       DEVISE~\cite{frome2013devise}    & 20.1 &11.7  &17.6  &6.1   &16.7  &14.4 \\
                     &        ALE~\cite{akata2013label}    & 22.7&12.7  &20.2  &6.8  &18.5   &16.2 \\
                      &       SPNet~\cite{xian2019semantic}     &{26.0}  &16.9  & 23.8 & 8.2  & 21.8  &19.4 \\
                             
                             		\hline
		
                             \multirow{3}{*}{DANN~\cite{ganin2016domain}}&DEVISE~\cite{frome2013devise}    & 20.5&10.4&16.4&7.1&15.1&13.9 \\
                             
                             &ALE~\cite{akata2013label}     & 21.2&12.5&19.7&7.4&17.9&15.7 \\
         &SPNet~\cite{xian2019semantic}&25.9&15.8&24.1&8.4&21.3&19.1\\
		\hline
		
                             \multirow{3}{*}{EpiFCR~\cite{li2019episodic}}&DEVISE~\cite{frome2013devise}    & 21.6& 13.9 &19.3  &7.3  &17.2   &15.9 \\
                             
                             &ALE~\cite{akata2013label}     & 23.2&  14.1&21.4  &7.8  &20.9   &17.5 \\
       &SPNet~\cite{xian2019semantic} 
                               &{26.4} & 16.7  & 24.6  & 9.2 & \textbf{23.2}    &20.0\\
        \hline
        \multicolumn{2}{c |}{\methodName}    
                                     &\textbf{27.6} &\textbf{17.8}  & \textbf{25.5}  & \textbf{9.9}  &{22.6}  & \textbf{20.7}  \\
		\end{tabular}
		}
		\label{tab:domainnet-additional}
\end{table*}

Finally, we want to highlight that coupling EpiFCR with any of the ZSL baselines, is not a straightforward approach, but requires to actually adapt this method, re-structuring the losses. In particular, we substitute the classifier originally designed for EpiFCR with the classifier specific of the ZSL method we apply on top of the backbone. Moreover, we additionally replace the classification loss with the loss devised for the particular ZSL method. For instance, for EpiFCR+SPNet, we use as classifier the semantic projection network, using the cross-entropy loss in \cite{xian2019semantic} as classification loss. Similarly, for EpiFCR+DEVISE and EpiFCR+ALE, we use as classifier a bi-linear compatibility function \cite{xian2018zeroshotgood} coupled with a pairwise ranking objective \cite{frome2013devise} and with a weighted pairwise ranking objective \cite{akata2013label} respectively.

\subsection{ZSL+DG: ablation study}
 In order to further investigate our design choices on the ZSL+DG setting, we conducted experiments on a challenging scenario where we consider just two domains as sources, i.e. Real and Painting. The results are shown in Table \ref{tab:ablation-zsldg}. On average our model improves SPNet by 2\% and SPNet + Epi-FCR by 1.1\%. Our approach without curriculum largely outperforms standard image-level mixup \cite{zhang2017mixup} (more than 2\%). Applying mixup at both feature and image level but without curriculum is effective but achieves still lower results with respect to our CuMix strategy (as in Tab. 2).  Interestingly, if we apply the curriculum strategy but switching the order of semantic and domain mixing (CuMix reverse), this achieves lower performances with respect to CuMix, which considers domain mixing harder than semantic ones. This shows that, in this setting, it is important to correctly tackle intra-domain semantic mixing before including inter-domain ones.
 
 \begin{table*}[t]
 			\caption{Results on DomainNet dataset with \textit{Real-Painting} as sources and ResNet-50 as backbone.} 
 			 		\centering
 		\begin{tabular}{ l | c  c  c  c | c}
 		Method/Target & Clipart& Infograph& Sketch& Quickdraw& Avg.\\\hline 
 SPNet& 21.5$\pm{0.6}$& 14.1$\pm{0.2}$& 17.3$\pm{0.3}$& 4.8$\pm{0.4}$& 14.4\\
 Epi-FCR+SPNet& 22.5$\pm{0.5}$& 14.9$\pm{0.7}$& 18.7$\pm{0.6}$& \textbf{5.6}$\pm{0.4}$& 15.4\\
 \hline
 MixUp img only & 21.2$\pm{0.4}$& 14.0$\pm{0.7}$& 17.3$\pm{0.3}$& 4.8$\pm{0.1}$& 14.3\\
 MixUp two-level& 22.7$\pm{0.3}$& 16.5$\pm{0.4}$& 19.1$\pm{0.4}$& 4.9$\pm{0.3}$& 15.8\\
 CuMix reverse& 22.9$\pm{0.3}$& 15.8$\pm{0.2}$& 18.2$\pm{0.3}$& 4.8$\pm{0.5}$& 15.4\\
 \hline
 CuMix & \textbf{23.7$\pm{0.3}$}& \textbf{17.1$\pm{0.2}$}& \textbf{19.7$\pm 0.3$}& {5.5$\pm{0.3}$}& \textbf{16.5} \\
 		\end{tabular}
           \label{tab:ablation-zsldg}
 \end{table*}

 \subsection{ZSL results}
 In this section, we report the ZSL results of Figure \ref{fig:dgzls-zsl-results} in tabular form. The results are shown in Table \ref{tab:zsl-results-additional}. Here, we also report the results of a baseline which uses just the cross-entropy loss term (similarly to \cite{xian2019semantic}), without the mixing term employed in our \methodName method. As the table shows, our baseline is weak, performing below most of the ZSL methods in all scenarios but FLO. However, adding our mixing strategy allows to boost the performances in all scenarios, achieving state-of-the-art performances in most of them. We also want to highlight that in Table \ref{tab:zsl-results-additional}, as in Figure \ref{fig:dgzls-zsl-results}, we do not report the results of methods based on generating features of unseen classes for ZSL \cite{xian2018feature,xian2019fvaegan}. This choice is linked to the fact that these methods can be used as data augmentation strategies to improve the performances of any ZSL method, as shown in \cite{xian2018feature}. While using them can improve the results of all the baselines as well as \methodName, this falls out of the scope of our work.

\begin{table*}[t]
			\caption{ZSL results.}
		\centering
		\begin{tabular}{ l |  c  c  c  c}
		Method&CUB&SUN&AWA1&FLO\\
		\hline
		ALE  \cite{akata2013label}   & 54.9  & 58.1  & 59.9& 48.5   \\
		SJE   \cite{akata2015evaluation}  &  53.9 & 53.7  &   65.6& 53.4 \\
		SYNC   \cite{changpinyo2016synthesized}  &  56.3 & 55.6   & 54.0& -  \\
		GFZSL \cite{verma2017simple}    &  49.3 & 60.6  & \textbf{68.3}&  -  \\
		SPNet \cite{xian2019semantic}    & 56.5  & 60.7  & 66.2& -  \\
		\hline
		Baseline  & 52.4  & 58.2 & 62.5 & 58.4 \\
		\methodName      & \textbf{{60.4}} & \textbf{62.4} & 64.0  &\textbf{ 59.7 }\\
		\end{tabular}
		\label{tab:zsl-results-additional}
\end{table*}

\backmatter
\phantomsection

\bibliographystyle{abbrv}
\bibliography{egbib}

\end{document}